%% file: main.tex
\documentclass{bmvc2k}

\newcommand{\OURS}{FlowObject\xspace} 

\title{\OURS: Flow Steering for Bridging Generative Priors and Reconstruction Fidelity}

\addauthor{Yuchen Rao}{yuchen.rao@tugraz.at}{1}
\addauthor{Xuqian Ren}{xuqian.ren@tuni.fi}{2}
\addauthor{Yinyu Nie}{yinyu.nie@tum.de}{3}
\addauthor{Sayan Deb Sarkar}{sdsarkar@stanford.edu}{4}
\addauthor{Biao Zhang}{biao.z@outlook.com}{5}
\addauthor{Vincent Lepetit}{vincent.lepetit@enpc.fr}{6}
\addauthor{Friedrich Fraundorfer}{friedrich.fraundorfer@tugraz.at}{1}

\addinstitution{
 Graz University of Technology \\ 
 Austria
}
\addinstitution{
Tampere University\\ Finland
}
\addinstitution{
Technical University of Munich\\
Germany}
\addinstitution{Stanford University\\ The United States of America}
\addinstitution{Xi'an Jiaotong University\\ China}
\addinstitution{
École des Ponts ParisTech \\
France}
\runninghead{Rao ET AL.}{FlowObject}

\usepackage{graphicx}
\usepackage{booktabs}
\usepackage{wrapfig}
\usepackage{booktabs}
\usepackage[table]{xcolor}
\usepackage{amssymb}
\usepackage{makecell}

\input{vincents_commands}
\begin{document}
\maketitle
\input{sections/0b_teaser}

\begin{abstract}
\input{sections/0_abstract}  
\end{abstract}

\input{sections/1_intro}
\input{sections/2_related_work}
\input{sections/4_method}
\input{sections/5_exp}

\input{sections/6_conclusion}
{
    \small
    \bibliography{main}
}

\input{X_suppl}

\end{document}

%% file: vincents_commands.tex
\usepackage{dsfont}
\usepackage{etoolbox}
\usepackage{color}

\newif\ifshowedits

\newcommand{\addeditor}[3]{%
  \definecolor{#1color}{rgb}{#3}
  \expandafter\newcommand\csname #1\endcsname[1]{%
  \ifshowedits
    {\color{#1color} ##1}%
  \else
    {##1}%
  \fi
  }%
  \expandafter\newcommand\csname #1rmk\endcsname[1]{%
  \ifshowedits
    {\color{#1color} {\bf [#2: ##1]}}
  \fi
  }%
  \expandafter\newcommand\csname #1rpl\endcsname[2]{%
  \ifshowedits
    {\color{#1color} ##1 \sout{##2}}
  \else
    {##1}
  \fi
  }%
}

\newcommand{\createtextvar}[1]{
  \expandafter\newcommand\csname #1\endcsname{%
  {\text{#1}}
}%
}
\newcommand{\textvars}[1]{\forcsvlist{\createtextvar}{#1}}

\newcommand{\mycomment}[1]{}

\newcommand{\calL}{{\cal L}}
\newcommand{\calM}{{\cal M}}

\newcommand{\calO}{{\cal O}}

\newcommand{\bv}{{\bf v}}

\newcommand{\bz}{{\bf z}}

\newcommand{\vcomment}[1]{}

%% file: sections/0b_teaser.tex
\begin{figure}[t]
\vspace{-1mm}
\centering
\includegraphics[width=0.9\linewidth]{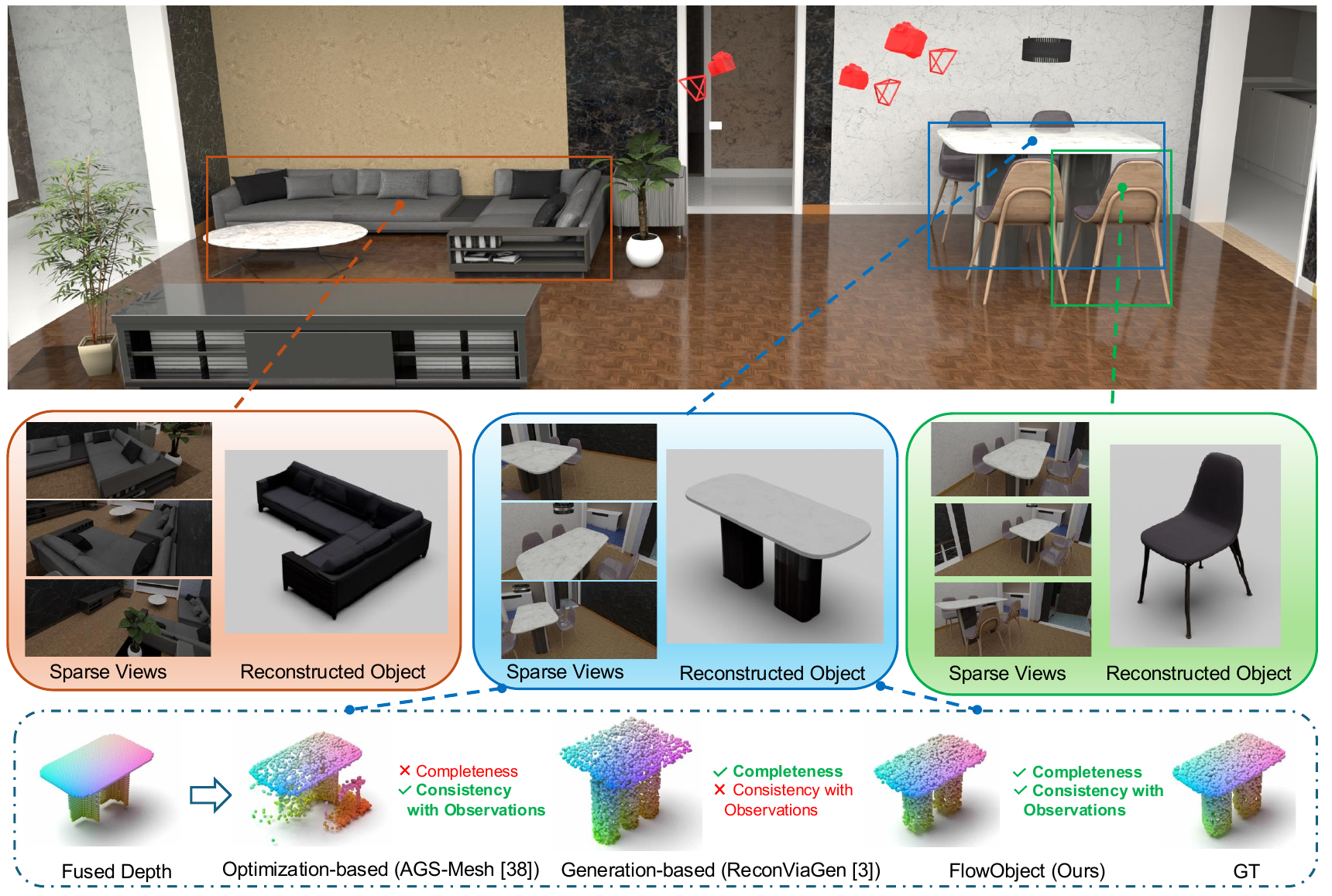}
\vspace{-2mm}
\caption{\textbf{Overview of \OURS.} 
\OURS provides a complete and high-fidelity 3D reconstruction of objects from few casual RGB-D images. Existing optimization-based 3D reconstruction methods fail to recover the unobserved geometry; generation-based methods produce complete object representations but in general a geometry that differs from the real object. \OURS brings the best of the two worlds by combining 3D reconstruction from the visible parts and prior knowledge from generative methods to obtain the complete geometry with high-fidelity.
}

\label{fig:teaser}
\vspace{-2mm}
\end{figure}

%% file: sections/0_abstract.tex
Recovering complete 3D representations of objects from few casual image captures remains a significant challenge.
Recent 3D generative models, particularly those based on Flow-Matching~(FM), can synthesize high-quality textured assets; however, they often suffer from ``synthetic bias'' where learned priors override observational evidence, alongside a lack of alignment with the observed instance. Conversely, optimization-based methods like 3D Gaussian Splatting~(3DGS) provide high fidelity on visible surfaces but fail to reason about unobserved geometry. In this paper, we present \OURS, a framework that reformulates sparse-view 3D reconstruction as a training-free, guided inverse problem. 
Our approach applies a dual-space guidance strategy to steer the Ordinary Differential Equation~(ODE) trajectory of a flow-matching model, enabling the completion of unseen regions through learned generative priors while enforcing strict consistency with real-world observations.
By integrating a 3DGS refinement stage, \OURS further bridges the gap between ``synthetic-looking'' generative outputs and photorealistic reconstructions. Comprehensive benchmarks on synthetic and real-world datasets demonstrate that current state-of-the-art methods often struggle to achieve geometric completeness and observational consistency simultaneously, especially under severe occlusions. In contrast, our method significantly outperforms state-of-the-art generative models and optimization-based frameworks in both geometric completeness and view-dependent appearance fidelity.

%% file: sections/1_intro.tex
\section{Introduction}
\label{sect:intro}
Recovering complete, reusable, and realistic 3D representations from real-world captures remains a fundamental challenge in computer vision. Casual real-world acquisition often produces fragmented, sparse, and noisy measurements due to sensor limitations, occlusions, and incomplete scene coverage. Reconstructing a coherent 3D scene thus requires a principled synthesis of the unobserved geometric and appearance data that is consistent with the captured evidence.

Optimization-based methods, such as standard 3D Gaussian Splatting (3DGS)~\cite{huang20242dgs,ren2025ags,xu2024grm,kerbl20233d,guedon2023sugar,yu2024gof,xu2024grm,yang2025instascene} and neural radiance fields (NeRF)~\cite{hong2023lrm,liu2023zero,chen2024mesh2nerf,zhang2024clay,huang2025cupid,xia2026holoscene,ni2025decompositional}, rely strictly on view-space photometric consistency constraints, rendering them inherently incapable of hallucinating or completing geometry and appearance in unobserved or heavily occluded regions.
To resolve these structural and appearance gaps, recent approaches have turned to 3D generative models, particularly Flow-Matching~(FM) frameworks, which exhibit a powerful capacity to synthesize complete, textured 3D assets~\cite{xiang2025structured,wu2025unilat3d,lu2025orientation,xiang2025native,chang2025reconviagen,siddiqui2026shaper} conditioned on text or images. 
Unlike previous generative methods that require extensive training or struggle with scale and translation ambiguities inherent in unconstrained latent sampling, we present \OURS, a framework that bypasses the need for per-instance retraining and ensures that generative completion remains strictly anchored to physical observations, enabling a robust, zero-shot paradigm for high-fidelity 3D reconstruction. \OURS reformulates sparse-view 3D modeling as an inference-time inverse problem, which can recover high-fidelity, complete 3D models of objects, including occluded regions, from casual real-world acquisitions, as illustrated in Fig.~\ref{fig:teaser}. Specifically, it steers the Ordinary Differential Equation (ODE) trajectory of the flow-matching process toward the intersection of two critical manifolds: the data manifold~($\mathcal{M}_{\text{data}}$), which provides a pretrained generative prior to resolve geometric and textural incompleteness, and the instance manifold~($\mathcal{M}_{\text{real}}$), which enforces strict adherence to real-world observations. 

Our core technical contribution is a dual-space guidance strategy that resolves the tension between generative priors and real-world observational evidence. By simultaneously backpropagating an implicit \textbf{latent-space} alignment loss and an explicit \textbf{observation-space} measurement-consistency loss through the optimized object features, we steer the velocity field of the generative model to populate the occluded regions while strictly anchoring visible structures to real-world captures. This ensures that the generative prior complements rather than overrides the input evidence, yielding 3D representations that are both geometrically complete and instance-faithful. By leveraging these reconstructions for 3DGS-based refinement, \OURS handles complex view-dependent lighting, effectively bridging the gap between ``synthetic-looking'' generative outputs and photorealistic reconstructions.

Experiments show that \OURS significantly outperforms state-of-the-art methods. While existing 3D generative models~\cite{xiang2025structured,lu2025orientation,chang2025reconviagen,chen2025sam} excel at synthesizing complete assets, they struggle with real-world reconstruction due to ``synthetic bias'' and a lack of consistency with sparse observations. Conversely, optimization-based approaches~\cite{kerbl20233d,ren2025ags} provide high fidelity on visible surfaces but lack the generative capacity to reason about occluded geometry, leading to incomplete reconstructions. Existing methods fail to achieve sufficient geometric completeness and observational consistency simultaneously, particularly when reconstructing objects under severe occlusions, which is highlighted in the last row of Fig.~\ref{fig:teaser}. By leveraging priors to reconstruct unobserved regions while enforcing observational constraints, \OURS achieves geometric completeness and appearance fidelity that transcend the ``synthetic-looking'' outputs of generative models.

To summarize, our main contributions are as follows:
\begin{itemize}
\item We reformulate sparse-view 3D reconstruction as a guided inverse problem, steering the flow-matching trajectory towards the intersection of the data manifold of a generative model and the instance manifold defined by real-world observations.
\item We introduce dual-space guidance to jointly optimize latent-space alignment and measurement consistency in the observation space, reconciling generative priors with observations to enable high-fidelity completion of occluded regions.
\item We integrate a 3DGS refinement stage to bridge the gap between generative synthesis and photorealism, rectifying ``synthetic-looking'' appearance and capturing complex view-dependent appearance.
\end{itemize}

%% file: sections/2_related_work.tex
\section{Related Work}
\subsection{Prior-based Object Reconstruction and Novel View Synthesis}
Traditional object reconstruction often relies on predefined CAD databases~\cite{gumeli2022roca,avetisyan2019scan2cad} or learned geometric priors~\cite{rao2022patchcomplete,dai2017shape} to achieve shape recovery. Although these methods provide strong structural constraints, they are fundamentally limited by the diversity of available libraries and struggle to generalize to unique or out-of-distribution real-world objects. While some approaches attempt to map textures onto water-tight meshes to incorporate appearance~\cite{michel2022text2mesh,chen2023text2tex,lin2023magic3d}, they frequently fail to simultaneously achieve refined geometry and high-fidelity rendering, often constrained by fixed topology or simplified surface assumptions. 

To bridge this gap between structural integrity and visual realism, 3D Gaussian Splatting (3DGS)~\cite{kerbl20233d} has emerged as a powerful representation. 
To enhance its robustness in sparse-view scenarios, several extensions have been proposed: some incorporate sensor depth as geometric supervision~\cite{turkulainen2025dn,ren2025ags}, others represent surfaces using 2D surfels for better geometric alignment~\cite{guedon2023sugar,Dai2024GaussianSurfels,huang20242dgs}, and others establish an opacity field from these Gaussians to facilitate mesh extraction via Marching Tetrahedra~\cite{Radl2025SOF,yu2024gof}.
Many works have also adopted multi-view geometric constraints~\cite{chen2024pgsr,hosseinzadeh2025g3splat} or monocular priors~\cite{chen2024vcr,chung2024depth} to enforce consistency between rendered meshes and Gaussians~\cite{guedon2025milo}.
However, such GS-based methods remain strictly observation-dependent and cannot recover structural information in occluded areas. This motivates the need for a more holistic, generative approach that can synthesize both complete geometry and consistent appearance by leveraging deep learned priors.

\subsection{Generative Priors as Guided Inverse Problems}
To recover complete geometry and appearance under extreme occlusions, 3D reconstruction is increasingly reformulated as a conditioned generative problem, using Diffusion~\cite{ho2020denoising} and Rectified Flow~\cite{lipman2022flow} models. Some methods train feed-forward large reconstruction models~\cite{hong2023lrm,wu2025unilat3d,zhang2024clay,xia2026holoscene,huang2025cupid} and multi-view diffusion models~\cite{liu2023zero,lai2025hunyuan3d} to enable instantaneous synthesis. Others use 3D-native priors for occlusion-aware geometry recovery~\cite{wu2025amodal3r,chou2023diffusion,yang2025instascene,ni2025decompositional}, leverage 3D priors from synthetic mesh data~\cite{chen2024mesh2nerf}, or predict holistic shapes from cluttered inputs~\cite{chen2025sam,siddiqui2026shaper}. Despite their completion capabilities, these models often exhibit a ``synthetic bias'', where learned priors override specific real-world observations. Unlike these approaches, \OURS{} treats the prior integration as a steerable process, ensuring that generative completion remains strictly anchored to the observational evidence.

From a theoretical perspective, we cast 3D reconstruction as a guided inverse problem solved through pretrained generative priors. Technically, ICTM~\cite{zhang2024flow} approximates the MAP estimator to solve linear inverse problems; DPS~\cite{chung2022diffusion} and Universal Guidance~\cite{bansal2023universal,sarkar2025} extend these solvers to non-linear problems via inference-time trajectory steering. 
However, lifting these strategies to structured 3D latent spaces~\cite{xiang2025structured,chang2025reconviagen,wu2025unilat3d,lu2025orientation} remains challenging for real-world scenarios due to high dimensionality and manifold complexity. 
To address this, \OURS leverages pretrained generative priors together with implicit latent-space alignment and explicit observation constraints. It ensures that unobserved regions are completed according to the learned prior, while visible surfaces remain strictly anchored to real-world observations, achieving a convergence between global plausibility and local fidelity. 

%% file: sections/4_method.tex
\textvars{data,guide,impl,expl}

\section{Method}
\label{sect:Method}

\subsection{Overview}

Given registered sparse-view RGB-D observations, \OURS reconstructs complete objects by steering a generative prior with real-world observations.
We perform Manifold-Guided Latent Optimization at each individual sampling step of the flow process using a dual-space objective. Explicit guidance enforces consistency with the observationsal measurements, while implicit guidance aligns the latent trajectory with observation-derived features extracted by a pretrained encoder.

This unified strategy, applied iteratively throughout the entire generation across both geometry~(\ref{sect:vox_generation}) and appearance~(\ref{sect:gs_generation}), leverages the prior to rationalize unobserved regions while maintaining strict adherence to real-world observation constraints eliminating the need for retraining. 
To overcome the ``synthetic bias'' of generative outputs, we further refine the 3D representation by optimizing 3DGS parameters~(\ref{sect:gs_opt}).
This step recovers high-frequency and view-dependent radiance that the prior cannot model directly.

\subsection{Problem Formulation}
\label{problem_formulation} 

We formulate sparse RGB-D object reconstruction under occlusions as a \textbf{dual-space guided inverse problem}. 
Given a sparse set of observed RGB-D images, we seek to recover a 3D object that minimizes the discrepancy between a prior provided by the generative model and evidence from the observations $\calO$.

\paragraph{Guided Latent Optimization.} 
The generative process~\cite{lipman2022flow}, as used in TRELLIS~\cite{xiang2025structured}, for example, is defined by a velocity field $\bv_t$ that transports noise at $t=1$ to the data manifold $\calM_\data$ at $t=0$, following the ODE: $\mathrm{d}\bz_t/\mathrm{d}t = \bv_t$.

We perform inference-time steering by treating the initial state noise $\mathbf{z}_{t=1}$, which structures the object's latent features, as the sole optimizable parameter. To ensure continuous alignment throughout the sampling process, we apply our guidance at every sampling step. Specifically, at each timestep $t$, we first ``look ahead'' by predicting state $\bz_{t-\Delta{t}}$ using the velocity field $\bv_{\theta}$ from the pretrained generative model~\cite{xiang2025structured}, conditioned on inputs $\mathbf{y}$ and directed toward the data manifold $\mathcal{M}_{\text{data}}$:
\begin{equation}
\bv_{t} = \bv_{\theta}(\bz_t, t, \mathbf{y}) \text{, with }
\bz_{t-\Delta{t}} = \bz_t + \bv_{t} \cdot (-\Delta t) \> .
\end{equation}

We then follow~\cite{bansal2023universal,sarkar2025} to minimize a guidance loss $\calL_\guide$ with respect to state $\bz_{t-\Delta{t}}$, which evaluates the consistency 
of the predicted state $\bz_{t-\Delta{t}}$ with the observations $\calO$ to satisfy the constraints of the instance manifold $\mathcal{M}_{\text{real}}$:
\begin{equation}
\calL_\guide(\bz_{t-\Delta t}) = \calL_\expl(\mathrm{Dec}(\bz_{t-\Delta t}), \calO) + \lambda_i\, \calL_\impl(\bz_{t-\Delta t}, \mathrm{Enc}(\calO)) \> ,
\label{eq:general_inverse}
\end{equation}
where $\calL_\expl$ and $\calL_\impl$ enforce explicit observation-space measurement-consistency and implicit latent-space alignment, respectively. In this formulation, $\mathrm{Dec}(\cdot)$ denotes the differentiable decoder that maps the latent representation $\bz$ to the real-world observations $\calO^\prime$ (i.e., voxels or 3D Gaussians). Conversely, $\mathrm{Enc}(\cdot)$ acts as the encoder, projecting real-world observations $\calO$ into the latent space to obtain the reference latent representation.

By backpropagating the gradients of $\mathcal{L}_{\mathrm{guide}}(\bz_{t-\Delta{t}})$, we update the state: 
\begin{equation}
\bz'_{t-\Delta{t}} = \bz_{t-\Delta{t}} - \eta \nabla_{\bz_{t-\Delta{t}}} \mathcal{L}_{\mathrm{guide}}(\bz_{t-\Delta{t}}) \> ,
\label{eq:backpropagate}
\end{equation}
where $\bz'_{t-\Delta{t}}$ is used as $\bz_{t}$ in the next iteration.

This steering mechanism enables the pretrained prior to complete occluded geometry and appearance at each sampling step; sequentially, the dual-space guidance anchors visible surfaces to real-world observations. Unlike direct backpropagation from sparse views, which often produces localized, unstable gradients that corrupt latent semantics, \OURS{} balances the generative prior with observational evidence. By leveraging $M_{data}$ to infer unobserved geometry while imposing $M_{real}$ constraints on visible surfaces, our approach achieves a reconstruction that is globally complete yet locally faithful."

\paragraph{Gaussian Parameter Refinement.}
Following the object-level completion, we move from the latent state $\mathbf{z}$ to the Gaussian parameter space 
$\Theta = \{ \mathbf{x}, \mathbf{s}, \linebreak \mathbf{q}, \mathbf{\alpha}, \mathbf{c} \}$. We then refine the reconstruction by jointly optimizing photometric and geometric consistency for the composed scene:
\begin{equation}
\min_{\Theta} 
\mathcal{L}_{\mathrm{refine}} 
\text{ with }
\mathcal{L}_{\mathrm{refine}} 
= \mathcal{L}_{\mathrm{rgb}}(G(\Theta), \mathcal{I}) + \lambda_d\, \mathcal{L}_{\mathrm{depth}}(D(\Theta), \mathcal{D}) + \lambda_r\, \mathcal{L}_{\mathrm{reg}}(\Theta),
\label{eq:gs_refine_equ}
\end{equation}
where $G(\cdot)$ and $D(\cdot)$ denote the differentiable Gaussian rendering for RGB and depth maps, respectively. Both $\mathcal{L}_{\mathrm{rgb}}$ and $\mathcal{L}_{\mathrm{depth}}$ are implemented as Mean Squared Error~(MSE) losses relative to the reference images $\mathcal{I}$ and depth maps $\mathcal{D}$. The regularization term $\mathcal{L}_{\mathrm{reg}}$ enforces geometric consistency and suppress artifacts during optimization.

By optimizing $\mathcal{L}_\mathrm{refine}$ with respect to the Gaussian parameters, we recalibrate the generative output to the scene's intrinsic lighting, effectively resolving the ``synthetic bias'' inherent in the generative prior, while ensuring that the globally complete geometry and appearance from the previous stage remain locally anchored to high-fidelity real-world observations.

\begin{figure}[!t]
\vspace{-1mm}
\centering
\includegraphics[width=0.85\linewidth]{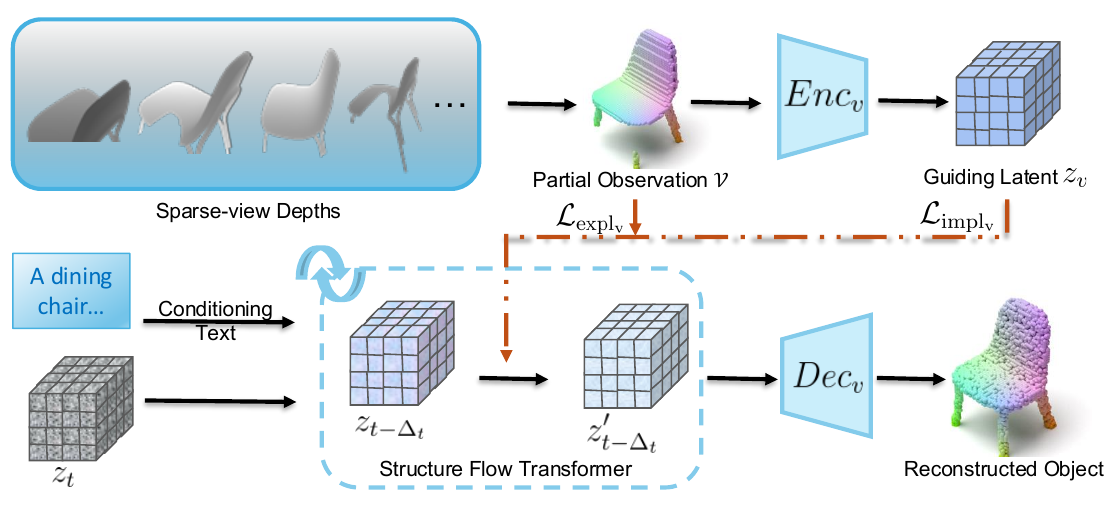}
\vspace{-4mm}
\caption{\textbf{Overview of the Guided Voxel Generation pipeline.} Partial depth inputs are fused into a voxel occupancy grid $\mathcal{V}$, denoted as the partial observation, from which the guiding latent $\mathbf{z}_v$ is extracted. During the structure flow process, our framework employs a combination of latent ($\mathcal{L}_{\mathrm{impl_v}}$) and observation ($\mathcal{L}_{\mathrm{expl_v}}$) constraints to iteratively refine the latent state $\mathbf{z}$, ultimately reconstructing a complete and high-fidelity 3D structure. }
\label{fig:vox_generation}
\vspace{-2mm}
\end{figure}
\subsection{Guided Object Reconstruction}
\label{sect:obj_rec}
We achieve object-level reconstruction from sparse-view RGB-D observations by combining a fixed generative prior with real-world observation constraints. Our approach leverages the pretrained data manifold~\cite{xiang2025structured} as the foundational prior, while utilizing real-world observations as steering gradients during the sampling process. 
By enforcing observation constraints from voxel-scale geometry to high-fidelity Gaussian appearance, the resulting reconstructions are both globally complete and locally faithful to the input observations.

\subsubsection{Guided Voxel Generation}
\label{sect:vox_generation}

\paragraph{Guidance Preparation.}
For each object $\mathcal{Q}$, we are given $N$ depth observations $\{\mathcal{D}_i\}_{i=1}^N$ along with their camera poses. We back-project these depth maps into a canonical object coordinate system and aggregate them into a voxel occupancy grid $\mathcal{V}$ as the partial observation, which serves as the explicit guidance. To provide implicit latent-space guidance, we encode $\mathcal{V}$ using a pretrained voxel encoder $\mathrm{Enc}_v(\cdot)$ to extract a guiding latent $\mathbf{z}_v = \mathrm{Enc}_v(\mathcal{V})$.

\paragraph{Voxel Reconstruction with Guidance.}
We instantiate the object's geometric structure using a pretrained rectified flow model, conditioned on a semantic text prompt. While the text prior provides a high-level bias to stabilize geometric completion in unobserved regions (maintaining alignment with $\mathcal{M}_{\text{data}}$), we update the trainable latent variables $\mathbf{z}$ by minimizing the voxel-scale formulation of the dual-space objective (Eq.~\eqref{eq:general_inverse}) at each integration step during the Structure Flow Transformer sampling process toward $\mathcal{M}_{\text{real}}$. Fig.~\ref{fig:vox_generation} illustrates this guided generation process. 

The objective $\mathcal{L}_{\mathrm{v}}$ combines explicit and implicit constraints. Since the partial voxel $\mathcal{V}$ only provides evidence for \emph{observed occupied} surfaces, we apply a masked binary cross-entropy (BCE) loss for explicit geometric supervision:
\begin{equation}
\mathcal{L}_{\mathrm{expl_v}}(\mathbf{z}; \mathcal{V}) = \mathrm{BCE}\!\left(\mathrm{Dec}_v(\mathbf{z}) \odot \mathcal{V},\, \mathcal{V}\right) \> ,
\end{equation}
where $\odot$ denotes element-wise multiplication, and $\mathrm{Dec}_v(\cdot)$ denotes the voxel-decoding branch of the pretrained generative model. To ensure $\mathbf{z}$ remains in high-probability regions of the pretrained latent space, we impose a latent alignment term:
\begin{equation}
\mathcal{L}_{\mathrm{impl_v}}(\mathbf{z}; \mathcal{V}) = \left\lVert \mathbf{z} - \mathbf{z}_v \right\rVert_1 \> .
\end{equation}

The total objective for geometric reconstruction is:
\begin{equation}
\mathcal{L}_{\mathrm{v}}(\mathbf{z}; \mathcal{V}) = \mathcal{L}_{\mathrm{expl_v}}(\mathbf{z}; \mathcal{V}) + \lambda_v\,\mathcal{L}_{\mathrm{impl_v}}(\mathbf{z}; \mathcal{V}) \>.
\label{eq:voxel_loss}
\end{equation}

In practice, we employ an annealed scheduling of guidance to ensure a stable optimization process. $\mathcal{L}_{\mathrm{impl_v}}$ is prioritized during early iterations to regularize the latent states within a physically plausible distribution, providing a stable global initialization. Subsequently, $\mathcal{L}_{\mathrm{expl_v}}$ is progressively introduced to enforce strict voxel-level measurement fidelity to the real-world instance. This staged transition allows the model to first ``rationalize'' the coarse object configuration, then precisely anchor the visible surfaces to the target measurements.

\subsubsection{Guided 3D Gaussian Generation}
\label{sect:gs_generation}
\begin{figure}[!t]
\vspace{-1mm}
\centering
\includegraphics[width=0.85\linewidth]{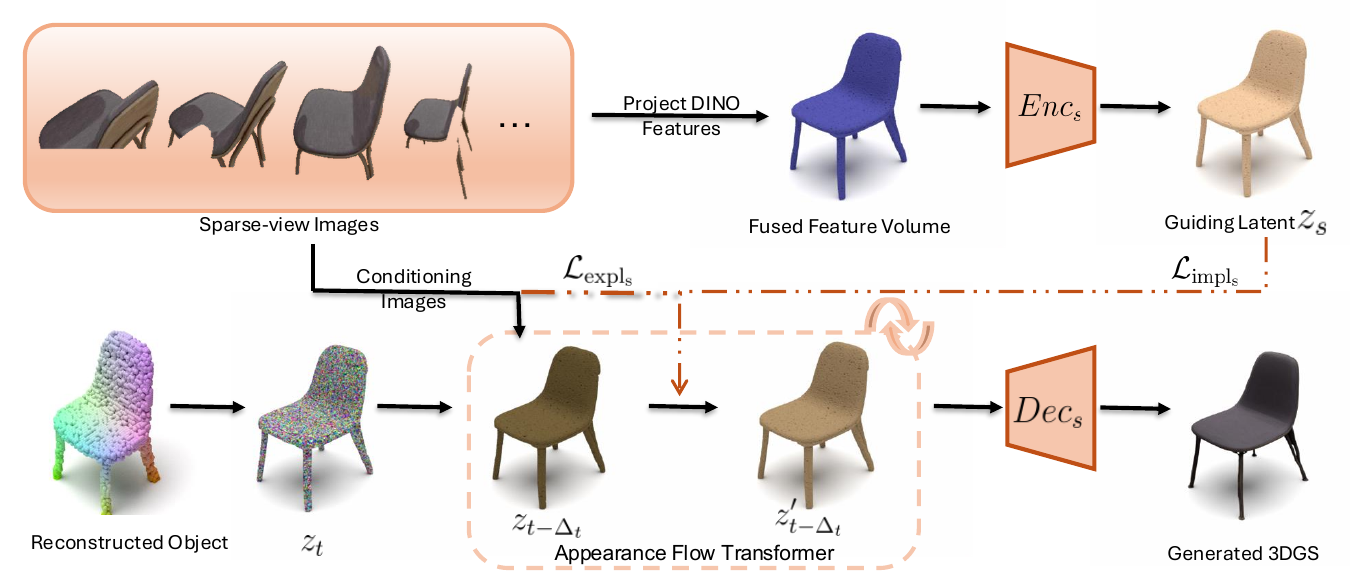}
\vspace{-4mm}
\caption{\textbf{Overview of the Guided 3DGS Generation pipeline.}  Our Gaussian synthesis pipeline consolidates sparse-view RGB images $\{\mathcal{I}_i\}$ into a photometric-derived guiding latent $\mathbf{z}_s$, built on the pre-generated voxel grid. By integrating latent ($\mathcal{L}_{\mathrm{impl_s}}$) and photometric ($\mathcal{L}_{\mathrm{expl_s}}$) constraints during the appearance flow process, the framework iteratively optimizes the latent state $\mathbf{z}$ to reconstruct a complete, view-consistent 3D Gaussian representation. }
\label{fig:gs_generation}
\vspace{-2mm}
\end{figure}
\paragraph{Guidance Preparation.}
For each object $\mathcal{Q}$, we are given $N$ RGB observations $\{\mathcal{I}_i\}_{i=1}^N$ with corresponding camera poses. We extract DINO features from each image and project them into the voxel grid generated in the previous stage. These multi-view features are statistically aggregated to form a fused feature volume $\mathbf{z}_f$, which is subsequently mapped to an appearance latent representation $\mathbf{z}_s$ using a pretrained encoder $\mathbf{z}_s = \mathrm{Enc}_s(\mathbf{z}_f)$. This latent $\mathbf{z}_s$ is spatially aligned with the fixed voxel grid and serves as the guiding latent for the appearance flow process.
\paragraph{3D Gaussian Reconstruction with Guidance.}
The final 3D Gaussian representation is synthesized using a pretrained rectified flow model. Guided by the appearance guiding latent $\mathbf{z}_s$ and differentiable rendering, this inference process~(Fig.~\ref{fig:gs_generation}) augments each sampling step with the iterative optimization of trainable latent $\mathbf{z}$ to minimize the Gaussian-scale objective~(Eq.~\eqref{eq:general_inverse}). This progressively aligns the 3DGS appearance with the sparse-view RGB observations, ensuring high-fidelity and view-consistent results.

The objective $\mathcal{L}_{\mathrm{s}}$ combines explicit and implicit constraints to drive the latent state toward $\mathcal{M}_{\text{real}}$. Specifically, $\mathbf{z}$ is mapped by the fixed appearance decoder $\mathrm{Dec}_s(\cdot)$ to a 3D Gaussian representation. We enforce an explicit masked photometric loss between the rendered views and input RGB observations:
\begin{equation}
\mathcal{L}_{\mathrm{expl_s}}(\mathbf{z}; \mathcal{I}) = \left\lVert \Pi\!\left(\mathrm{Dec}_s(\mathbf{z})\right) \odot \mathcal{M} - \mathcal{I} \odot \mathcal{M} \right\rVert_1,
\end{equation}
where $\Pi(\cdot)$ denotes differentiable rendering and $\mathcal{M}$ represents the visibility mask. 
Simultaneously, we impose an implicit alignment term:
\begin{equation}
\mathcal{L}_{\mathrm{impl_s}}(\mathbf{z}; \mathcal{I}) = \left\lVert \mathbf{z} - \mathbf{z}_s \right\rVert_1 \> .
\end{equation}
The total objective for Gaussian reconstruction is:
\begin{equation}
\mathcal{L}_{\mathrm{s}}(\mathbf{z}; \mathcal{I}) = \mathcal{L}_{\mathrm{expl_s}}(\mathbf{z}; \mathcal{I}) + \lambda_s\,\mathcal{L}_{\mathrm{impl_s}}(\mathbf{z}; \mathcal{I }) \> .
\label{eq:gs_loss}
\end{equation}

In practice, we adopt a staged conditioning and guidance strategy to manage the high degrees of freedom in appearance synthesis. The rectified flow model's conditioning is first set to a semantic text prompt and subsequently switched to the reference images $\{\mathcal{I}_i\}$ at a predefined iteration. Independent of this conditioning shift, the explicit photometric loss $\mathcal{L}_{\mathrm{expl_s}}$ is progressively introduced according to an annealed optimization schedule.
This decoupling of input conditioning from optimization supervision allows the prior to rationalize global distributions before strictly anchoring appearance to local RGB measurements.

\subsection{Photorealistic Radiance Refinement}
\label{sect:gs_opt}
We transfer the reconstructed objects into the scene coordinate system, integrating them with a background Gaussian representation initialized from complementary masks.
While generative stages ensure structural completeness and coarse appearance fidelity, the synthesized representation may still exhibit artifacts or misalignments in complex details.
The resulting joint radiance refinement enforces photorealistic NVS across the entire environment while preserving instance-level independence for downstream editing.

To resolve this and achieve high-fidelity photorealism, we unfreeze the full set of parameters for all 3D Gaussians, and optimize them jointly across the entire scene. We minimize a global photometric loss $\mathcal{L}_{\mathrm{scene}}$, as defined in Eq.~\eqref{eq:gs_refine_equ}, by rendering the scene from input viewpoints. In order to enhance the geometric quality and surface sparsity, $\mathcal{L}_{\mathrm{opac}} = \bar{\alpha}(1-\bar{\alpha})$ is employed as the regularizer $\mathcal{L}_{\mathrm{reg}}$ to encourage the opacity $\alpha$ of Gaussian primitives to converge toward binary values (0 or 1), thereby reducing semi-transparent artifacts. 
We also apply a recent densification method~\cite{zhou2025gradient} to enable gradient-direction-aware optimization. This final refinement stage serves as a fidelity anchoring phase, polishing surface details and recovering intricate radiance to resolve the ``synthetic bias'' of the generative prior. 

%% file: sections/5_exp.tex
\section{Experiments and Analysis}
\label{sect:experiments}

\begin{table}[t]
\centering
\scalebox{0.9}{
\begin{tabular}{l ccc ccc ccc}
\toprule
& \multicolumn{3}{c}{\makebox[0.2\textwidth][c]{\textbf{3D-FRONT}~\cite{fu20213d}}} & \multicolumn{3}{c}{\makebox[0.2\textwidth][c]{\textbf{ScanNet++}~\cite{yeshwanth2023scannet++}}} & \multicolumn{3}{c}{\makebox[0.2\textwidth][c]{\textbf{ShapeR}~\cite{siddiqui2026shaper}}}\\
\cmidrule(lr){2-4} \cmidrule(lr){5-7} \cmidrule(lr){8-10}

Method & 
\makebox[0.065\textwidth][c]{CD\textdownarrow} & 
\makebox[0.065\textwidth][c]{Comp.\textuparrow} & 
\makebox[0.065\textwidth][c]{F-Score\textuparrow} & 
\makebox[0.065\textwidth][c]{CD \textdownarrow} & 
\makebox[0.065\textwidth][c]{Comp.\textuparrow} & 
\makebox[0.065\textwidth][c]{F-Score\textuparrow} &
\makebox[0.065\textwidth][c]{CD\textdownarrow} & 
\makebox[0.065\textwidth][c]{Comp.\textuparrow} & 
\makebox[0.065\textwidth][c]{F-Score\textuparrow}\\

\midrule
3DGS-D~\cite{kerbl20233d}       & 5.49  & 40.31 & 40.16 & 5.77  & 28.39 & 28.79 & 7.31 & 36.99 & 43.78 \\
AGS-Mesh~\cite{ren2025ags}        & 5.87  & 38.51 & 41.10 & 5.80  & 21.93 & 23.62 & 7.24 & 39.08 & 46.12 \\
OM-GSD$^{\ast}$~\cite{lu2025orientation} & 12.57 & 23.47 & 23.45 & 11.91 & 16.92 & 19.37 & 12.67 & 18.53 & 19.36 \\
RVG-GSD$^{\ast}$~\cite{chang2025reconviagen} & 10.31 & 19.73 & 19.89 & 12.60 & 15.89 & 17.01 & 14.58 & 20.99 & 21.31 \\
SAM3D-GSD$^{\ast}$~\cite{chen2025sam} & 4.92 & 42.76 & 35.80 & 3.81 & 47.13 & 35.09 & 4.00 & 49.57 & 38.69 \\
\textbf{Ours} & \textbf{2.68} & \textbf{65.57} & \textbf{62.87} & \textbf{3.29} & \textbf{52.38} & \textbf{46.38} & \textbf{3.84} & \textbf{66.83} & \textbf{64.43} \\
\bottomrule
\end{tabular}
}
\caption{\textbf{Quantitative Evaluation of Geometric Fidelity.} We report Unidirectional Chamfer Distance (CD $\times 10^2$), Completion Score (Comp. \%), and F-Score (\%) on 3D-FRONT (synthetic data), ScanNet++ and ShapeR Evaluation Dataset~(real data). Threshold $\tau$ is set to $1\%$ of $d_{diag}$. Our method demonstrates superior
robustness in reconstructing complete geometry from sparse observations across both synthetic and real-world scenes. ($^{\ast}$: Results by OM-GSD and RVG-GSD are aligned via uniform scaling and rigid transformation and rotation. Results by SAM3D-GSD are obtained using its multi-view implementation.)}
\label{tab:combined_geometry_results}
\vspace{-5mm}
\end{table}

\subsection{Experimental Setup} 

\paragraph{\textbf{Datasets.}} 
To demonstrate the robustness of our framework, we evaluate our method on both complex synthetic environments and real-world captures. Our evaluation primarily focuses on representative indoor categories, ranging from structured furniture to geometrically complex electronic devices, as well as miscellaneous household objects, thereby validating our method’s capacity to handle varying scales and diverse geometric complexities.

\textbf{3D-FRONT}~\cite{fu20213d}. We utilize 3D-FRONT as our synthetic benchmark, which provides a vast diversity of 3D furniture layouts with ground-truth (GT) meshes. To simulate realistic partial observations, we adopt a human-centric viewpoint sampling strategy, described in~\cite{nie2023learning}, to generate camera trajectories.  Specifically, we constrain the training camera poses within a localized spatial region relative to the object center, ensuring that the initial input captures only a limited perspective of the target object.

\textbf{ScanNet++}~\cite{yeshwanth2023scannet++}. ScanNet++ provides a challenging real-world benchmark to evaluate the generalization of our method. By testing on these casual handheld RGB-D sequences involving motion blur and irregular sampling, we demonstrate our method's robust reconstruction from sparse, consumer-grade inputs. For quantitative assessment and spatial alignment, we utilize the object-level ground-truth meshes, oriented bounding boxes, and object poses provided by SCANnotate++~\cite{rao2025leveraging}.

\textbf{ShapeR Evaluation Dataset}~\cite{siddiqui2026shaper}.
This benchmark presents a challenging, in-the-wild collection of multi-view real-world imagery, calibrated camera parameters, and ground-truth (GT) 3D meshes in cluttered scenes. To adapt this data to our pipeline, foreground silhouette masks are obtained by projecting the GT 3D meshes onto the calibrated camera frames. Furthermore, we generate hybrid depth maps by rendering the GT object depth for the foreground and leveraging Depth Anything 3 (DA3)~\cite{lin2025depth} for the background.

\paragraph{\textbf{Evaluation Metrics.}}
We systematically assess our framework across two distinct dimensions: geometric fidelity and photorealistic synthesis. This dual-perspective evaluation is necessary to verify that our method not only recovers the physical structure of the object but also "rationalizes" its appearance in unobserved regions where observation data is absent.

\textbf{Geometry Evaluation.} The goal of geometry evaluation is to quantify structural completeness and surface accuracy. In sparse-view scenarios, capturing the object's true topology, especially in occluded areas, is the primary challenge.

We consider three key metrics: \textbf{Unidirectional Chamfer Distance ($L_1$)} from the ground truth (GT) to the predicted point cloud, \textbf{Completion Score}, and \textbf{F-Score}. Notably, our method disentangles the target object from the entire scene, whereas GS-based baselines rely on bounding box cropping, which often leaves residual background points. To ensure a fair comparison, we adopt a unidirectional evaluation protocol (GT $\rightarrow$ Prediction). Both Completion Score and F-Score are evaluated using a scale-adaptive threshold $\tau = 0.01 \cdot d_\text{diag}$, where $d_\text{diag}$ denotes the object's bounding box diagonal. This threshold identifies valid correspondences based on the object's scale, thereby ensuring a fair comparison across scenes of different sizes. We provide additional evaluation metrics in the supplementary material.

\textbf{Rendering Evaluation.} 
To assess the effectiveness of our manifold-guided synthesis, we evaluate rendering quality under challenging conditions characterized by high occlusion and extreme sparsity. Specifically, we select only a small subset of partial observations as training frames, intentionally leaving large angular gaps in the viewpoint coverage.
\begin{figure*}[t] 
    \centering
    \setlength{\tabcolsep}{2pt}
    \resizebox{\textwidth}{!}{
    \begin{tabular}{cccccccc}
        \includegraphics[trim={15cm 0.0cm 15cm 0.0cm},clip,width=0.12\textwidth]{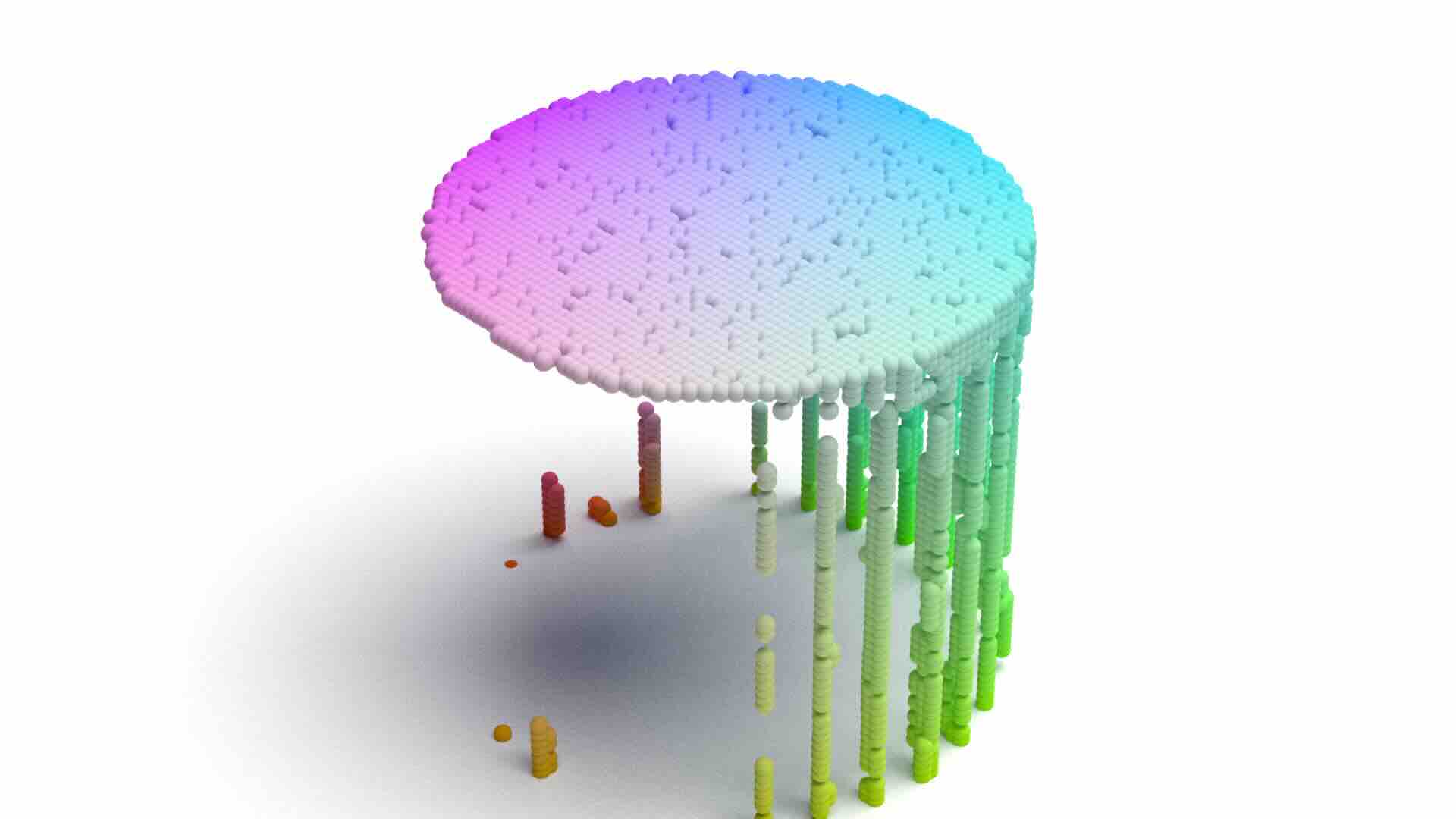} 
        &
        \includegraphics[trim={15cm 0.0cm 15cm 0.0cm},clip, width=0.12\textwidth]{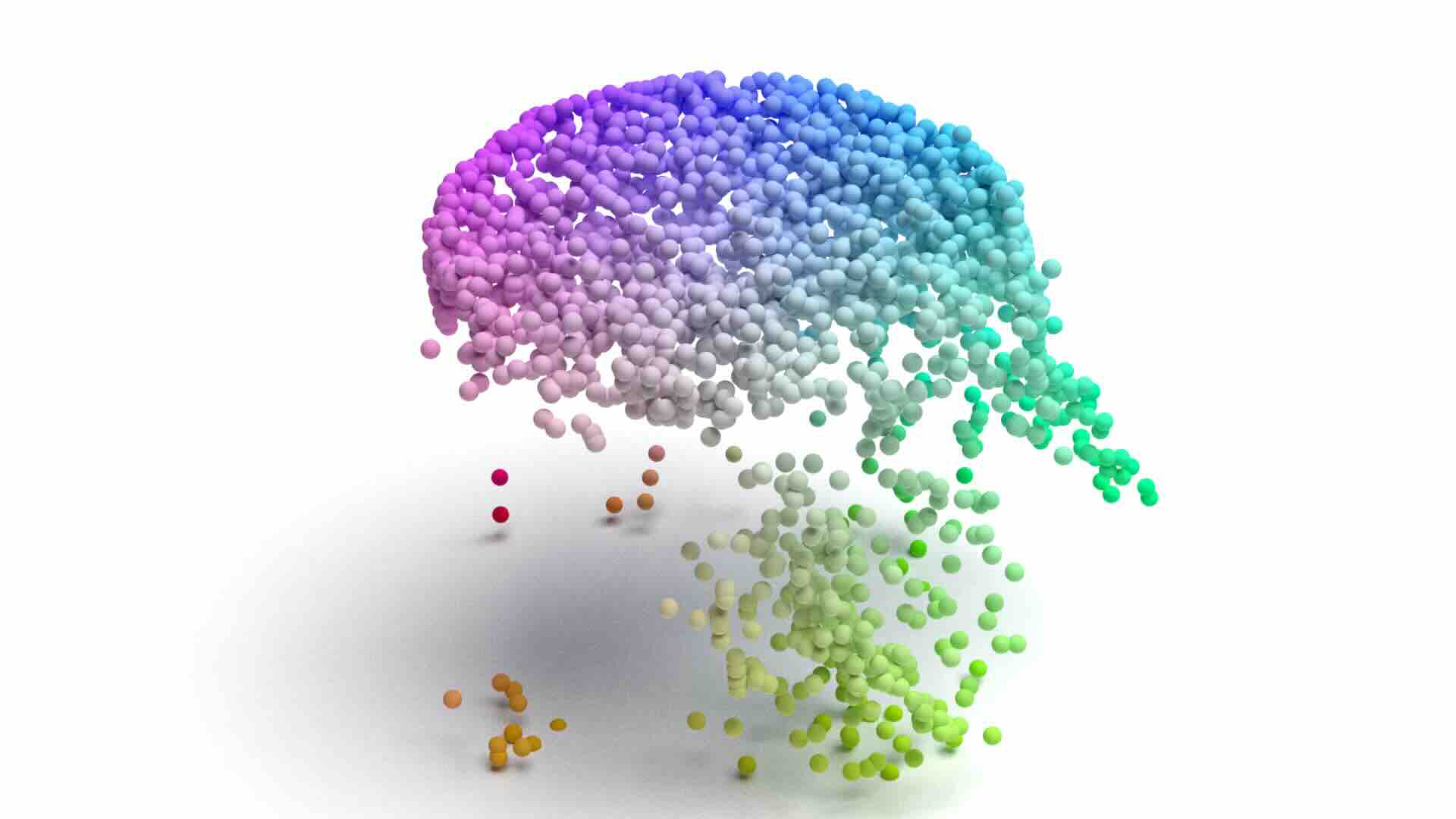} &
        \includegraphics[trim={15cm 0.0cm 15cm 0.0cm},clip, width=0.12\textwidth]{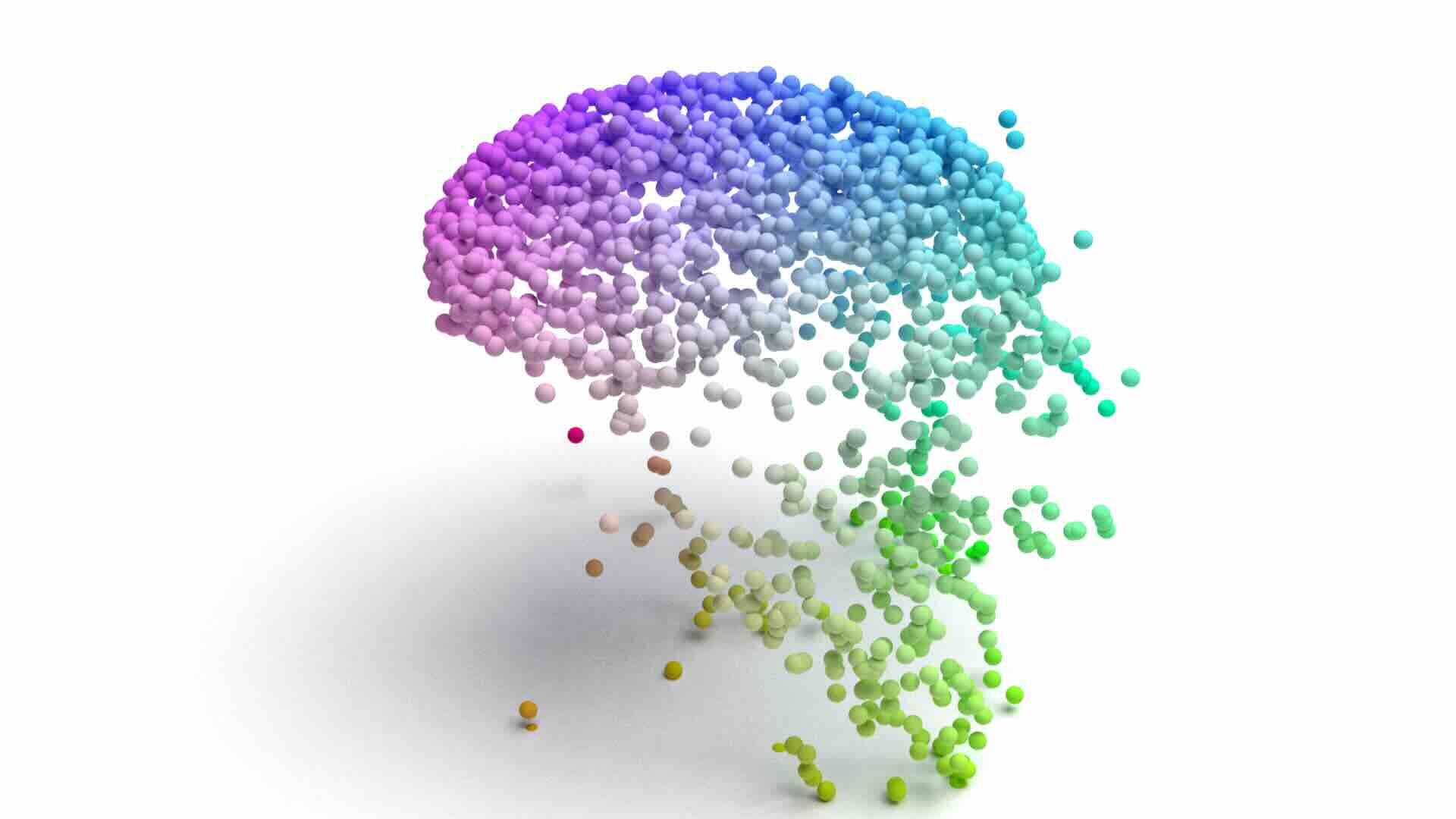} &
        \includegraphics[trim={15cm 0.0cm 15cm 0.0cm},clip, width=0.12\textwidth]{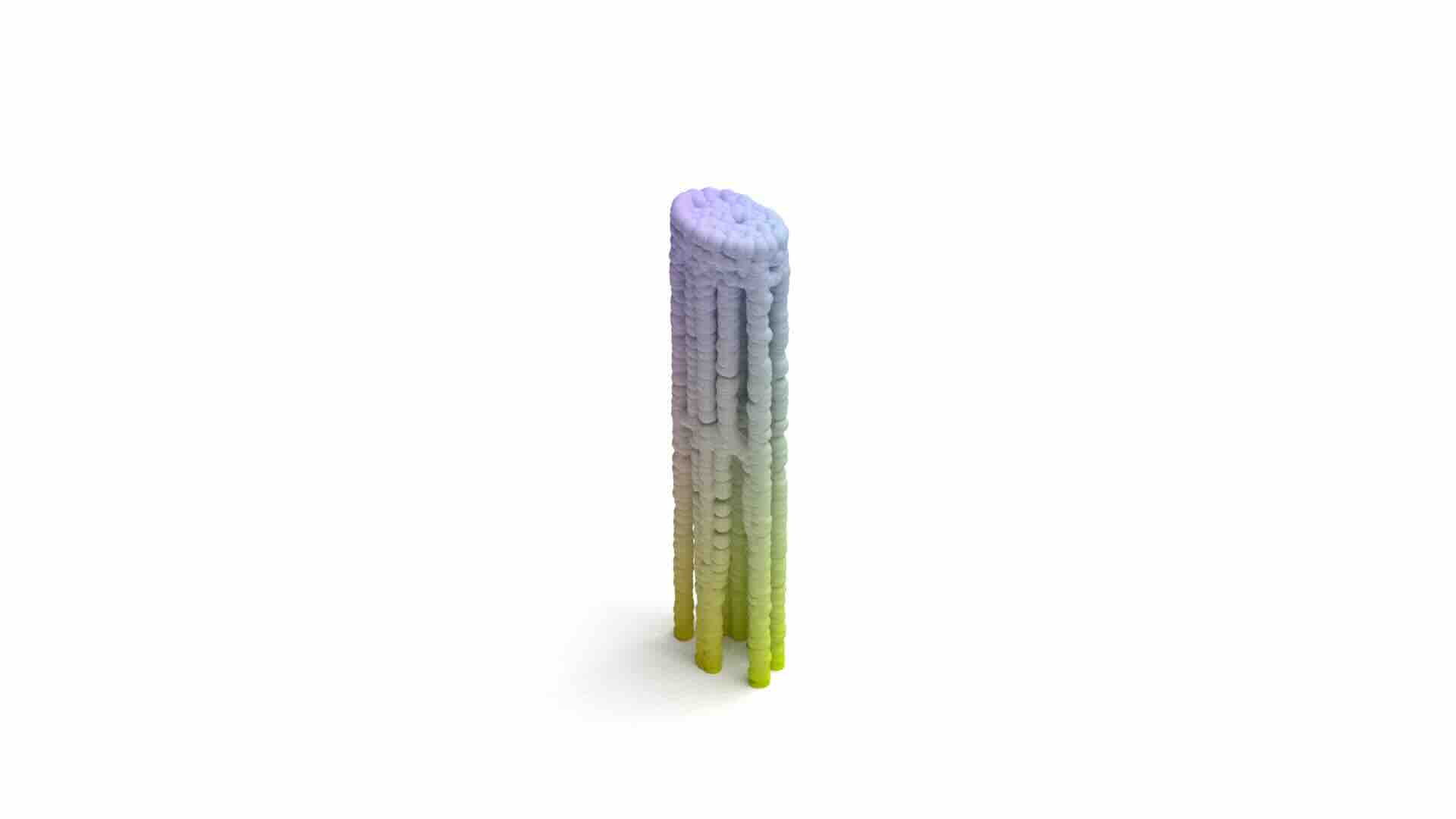} &
         \includegraphics[trim={15cm 0.0cm 15cm 0.0cm},clip,width=0.12\textwidth]{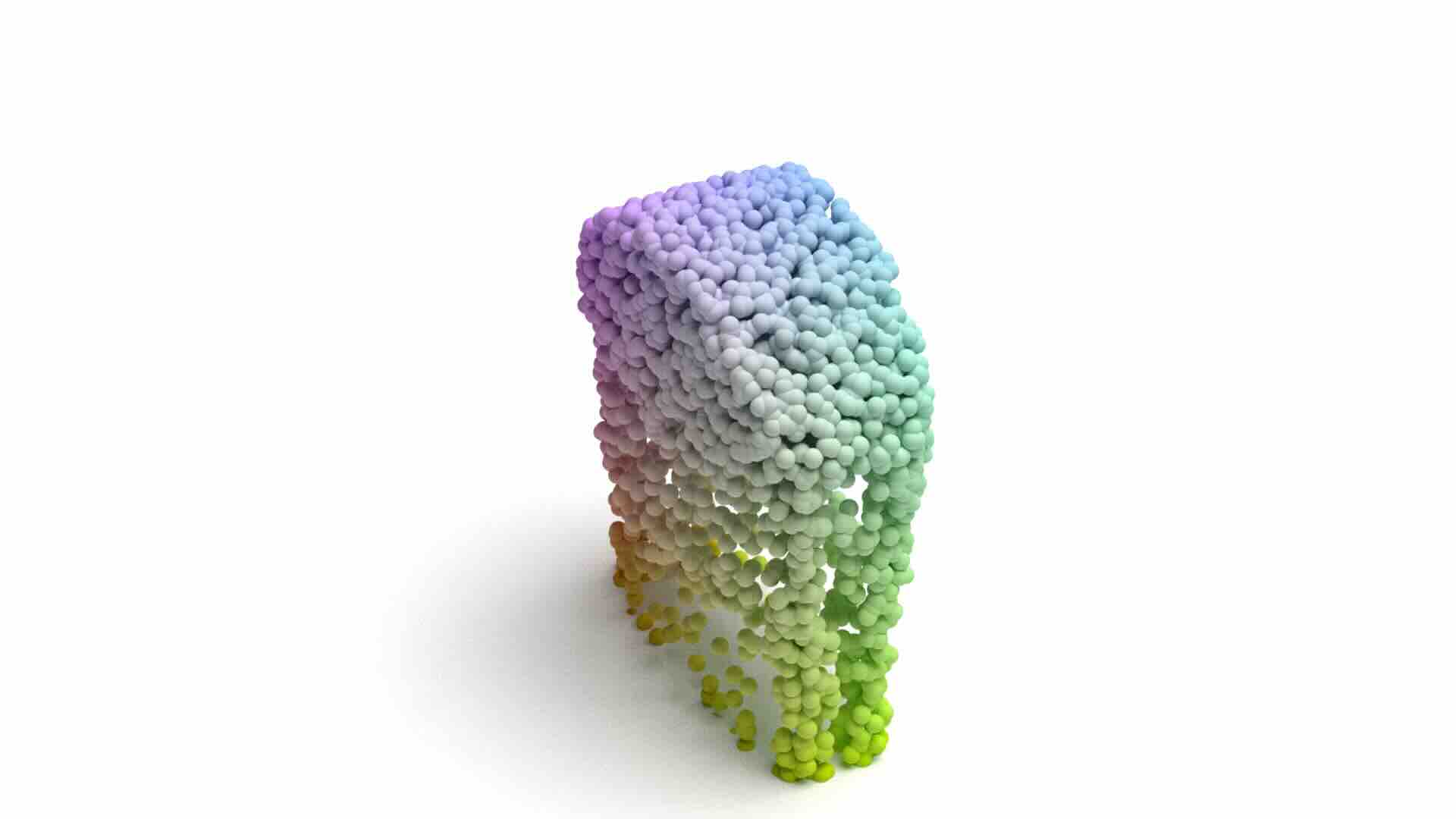} &
          \includegraphics[trim={15cm 0.0cm 15cm 0.0cm},clip,width=0.12\textwidth]{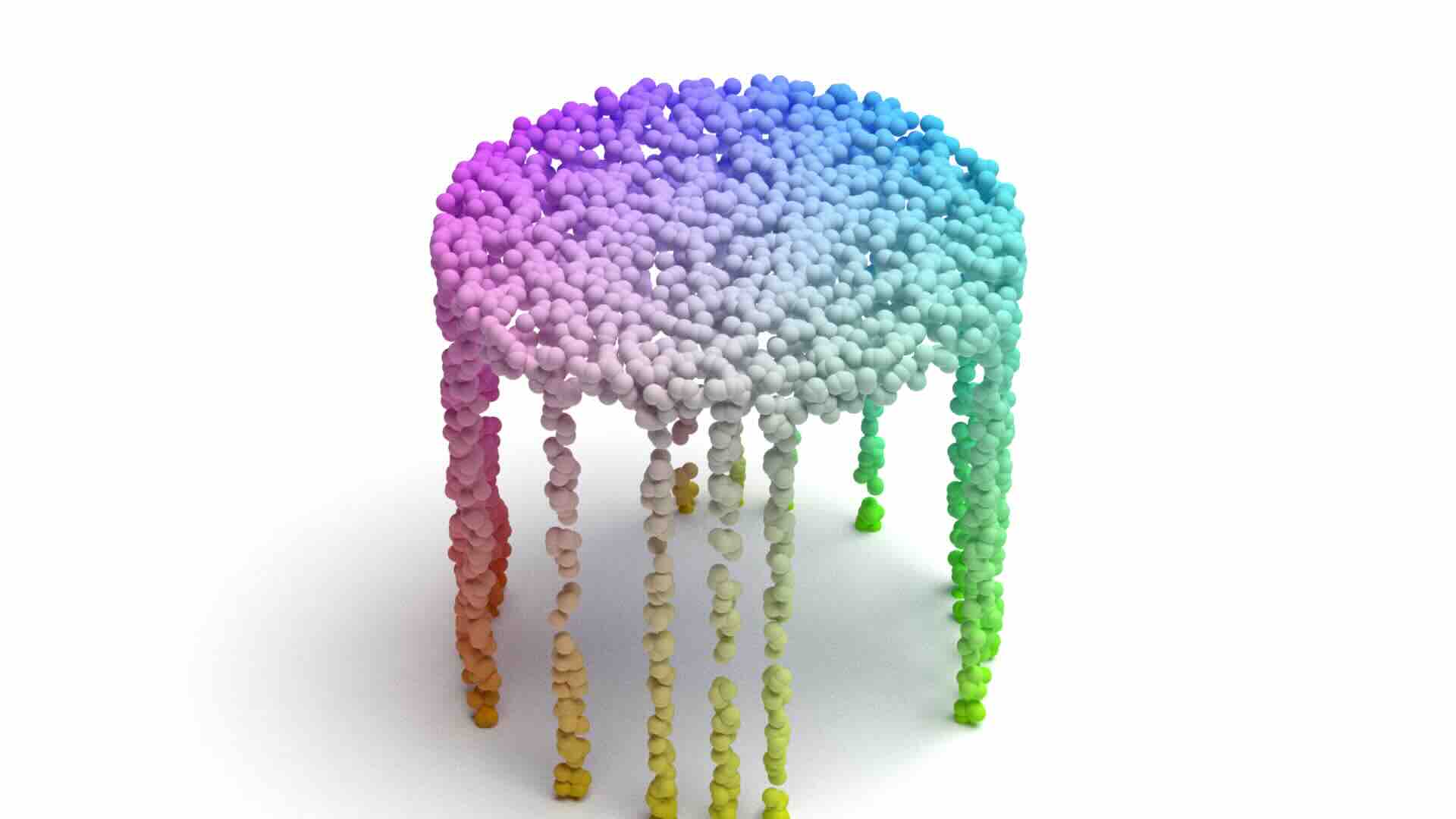} &
        \includegraphics[trim={15cm 0.0cm 15cm 0.0cm},clip,width=0.12\textwidth]{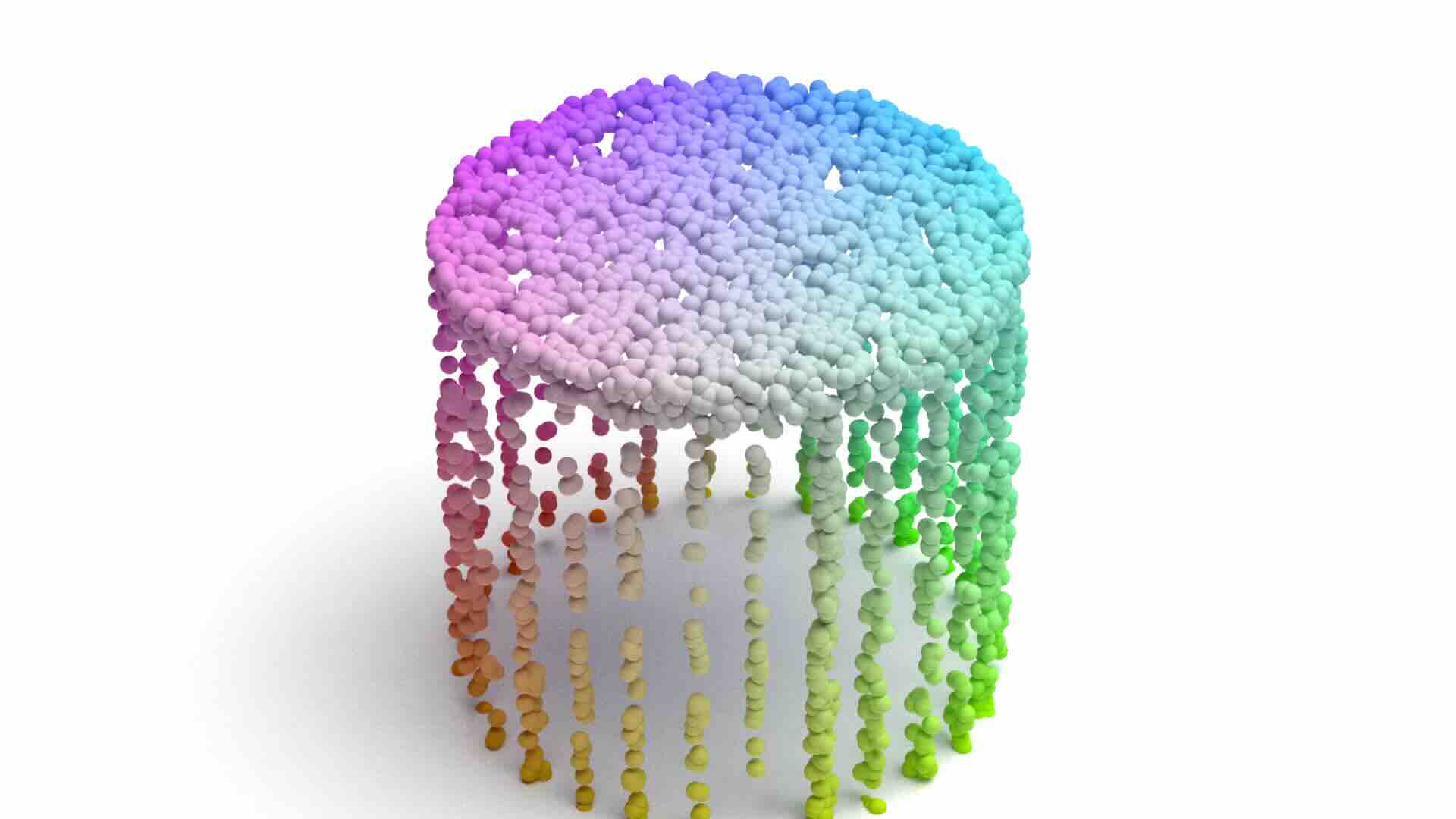} &
        \includegraphics[trim={15cm 0.0cm 15cm 0.0cm},clip,width=0.12\textwidth]{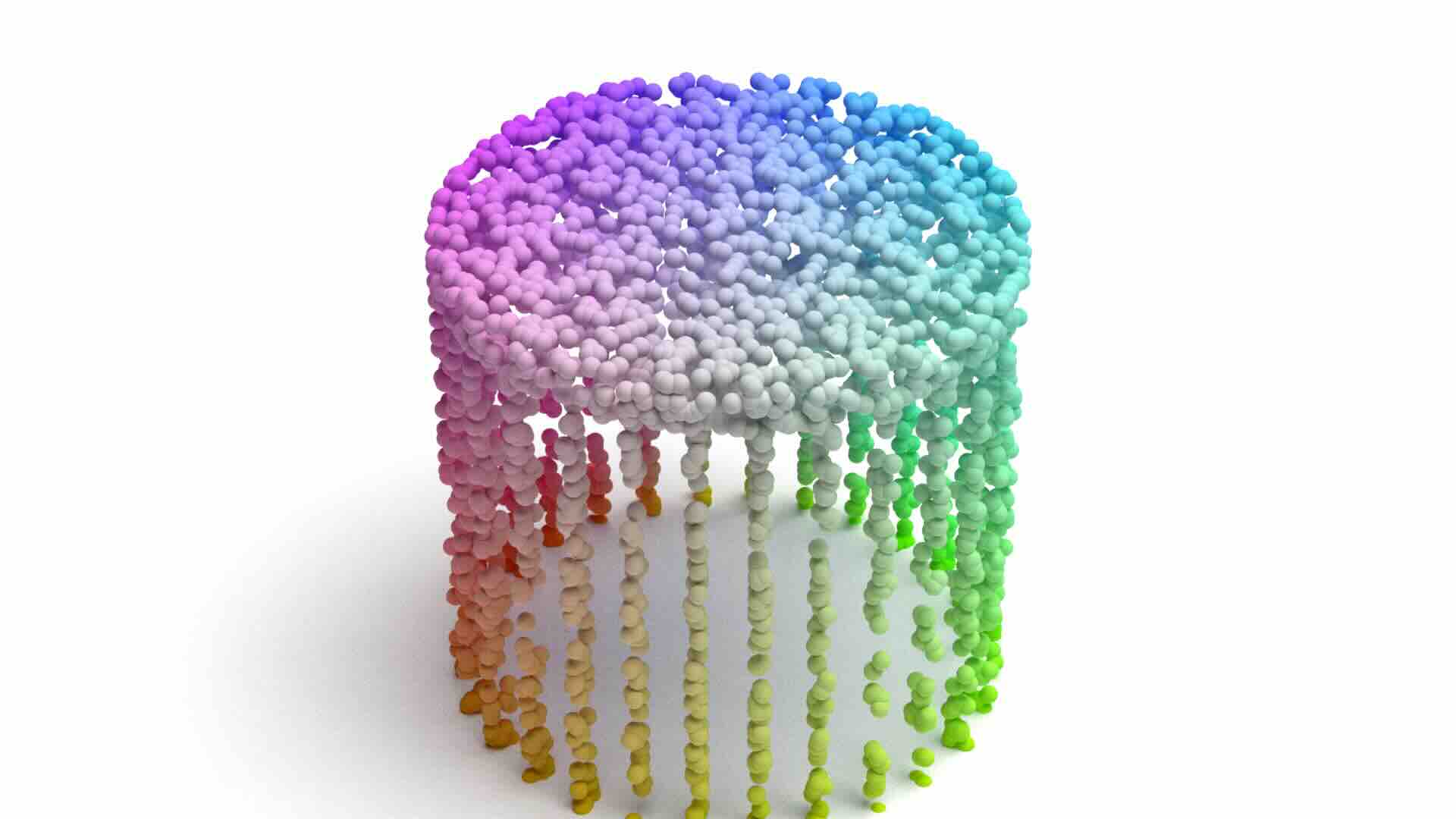}
        \\
        \includegraphics[trim={15cm 0.0cm 15cm 0.0cm},clip,width=0.12\textwidth]{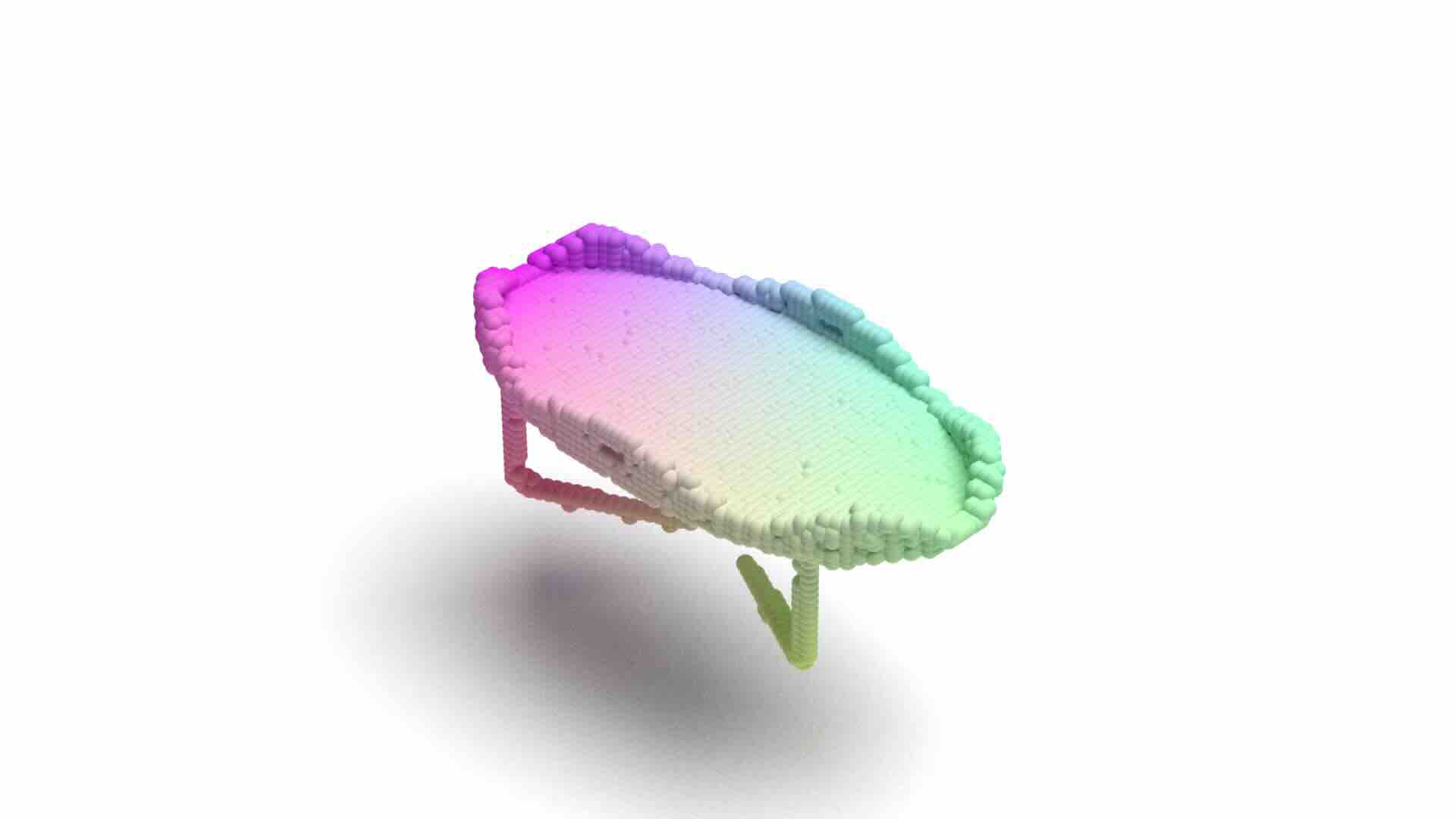} 
        &
        \includegraphics[trim={15cm 0.0cm 15cm 0.0cm},clip, width=0.12\textwidth]{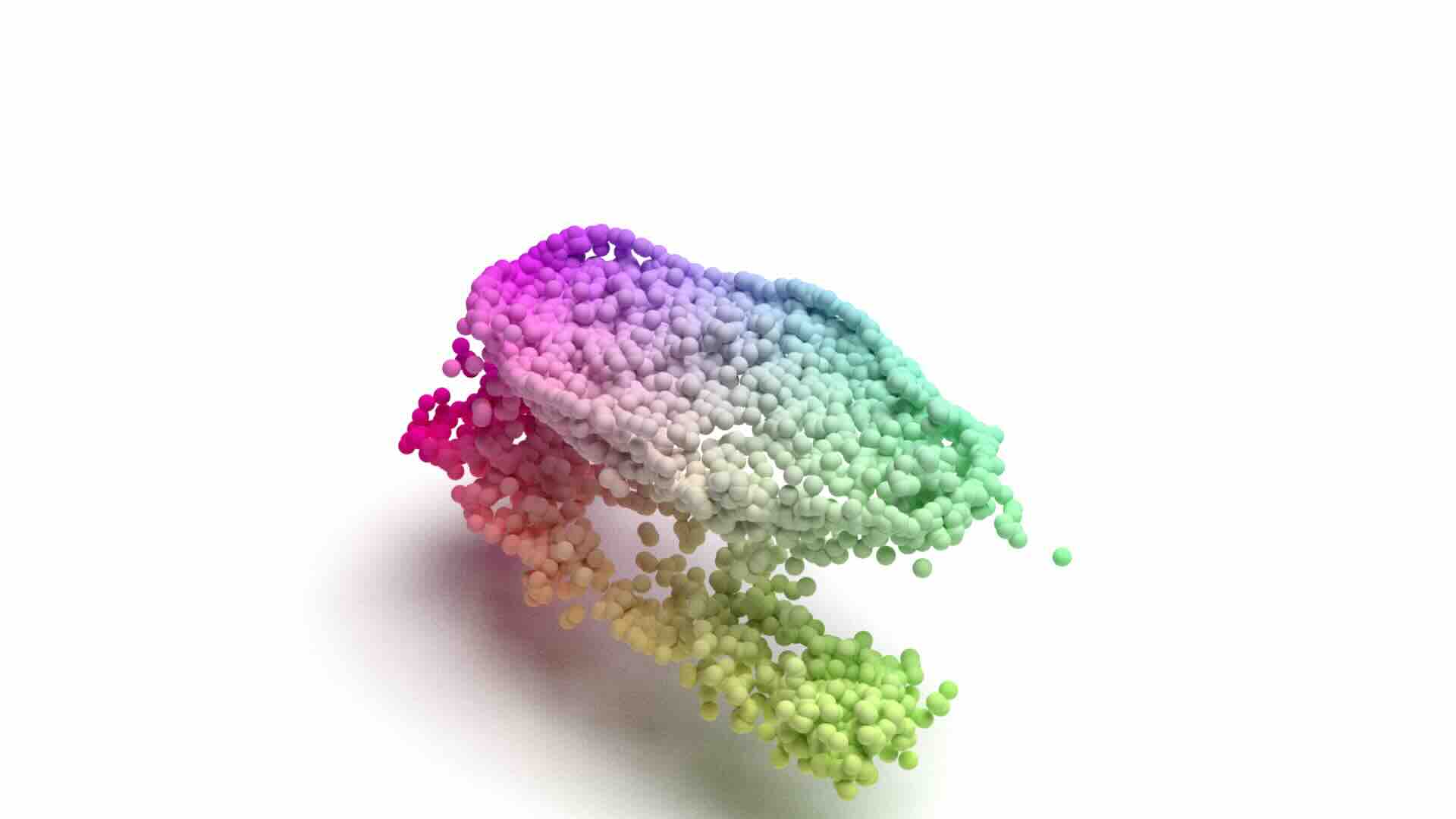} &
        \includegraphics[trim={15cm 0.0cm 15cm 0.0cm},clip, width=0.12\textwidth]{figures/3dfront_pts/54_ags.jpg} &
        \includegraphics[trim={15cm 0.0cm 15cm 0.0cm},clip, width=0.12\textwidth]{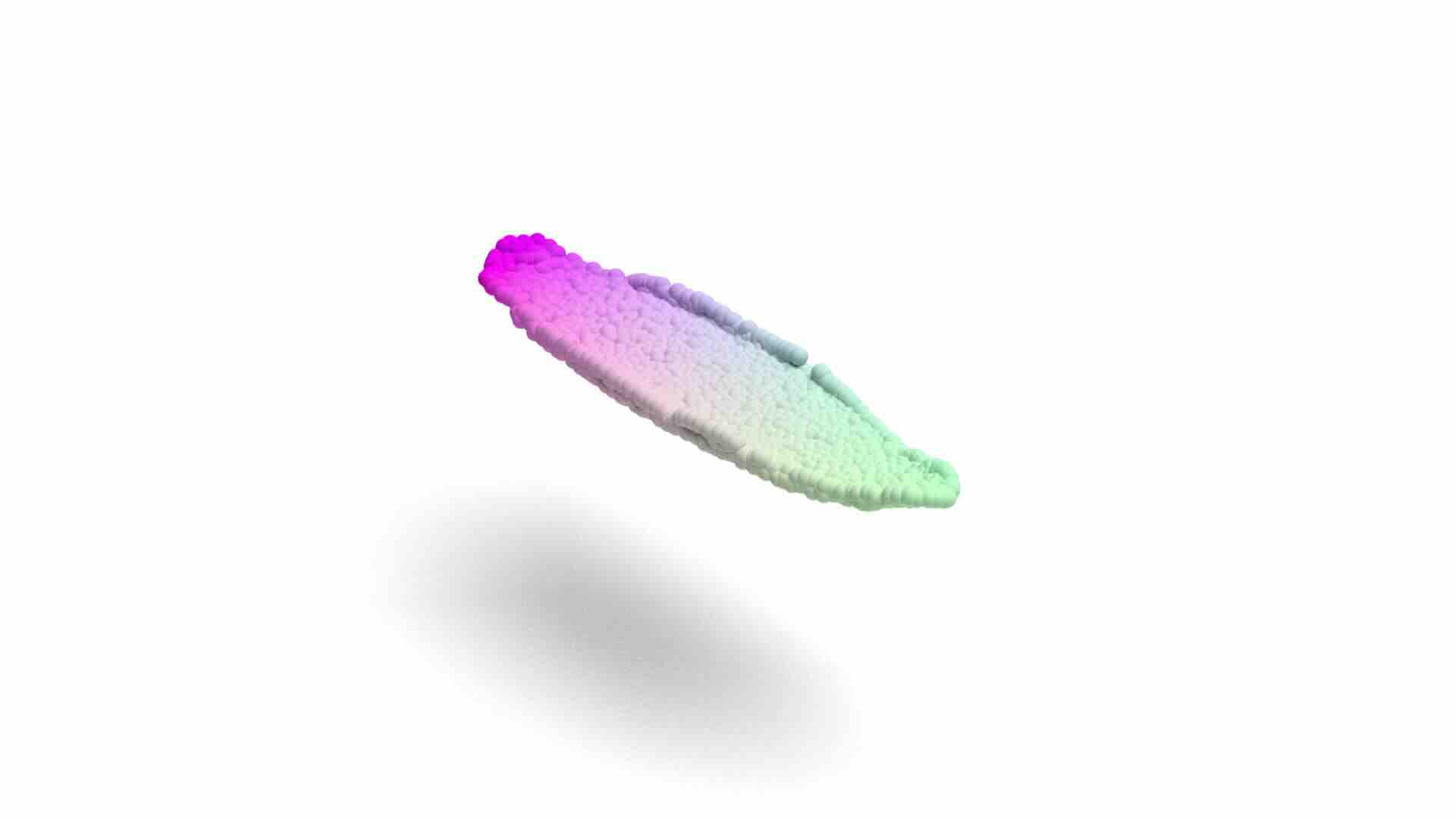} &
        \includegraphics[trim={15cm 0.0cm 15cm 0.0cm},clip, width=0.12\textwidth]{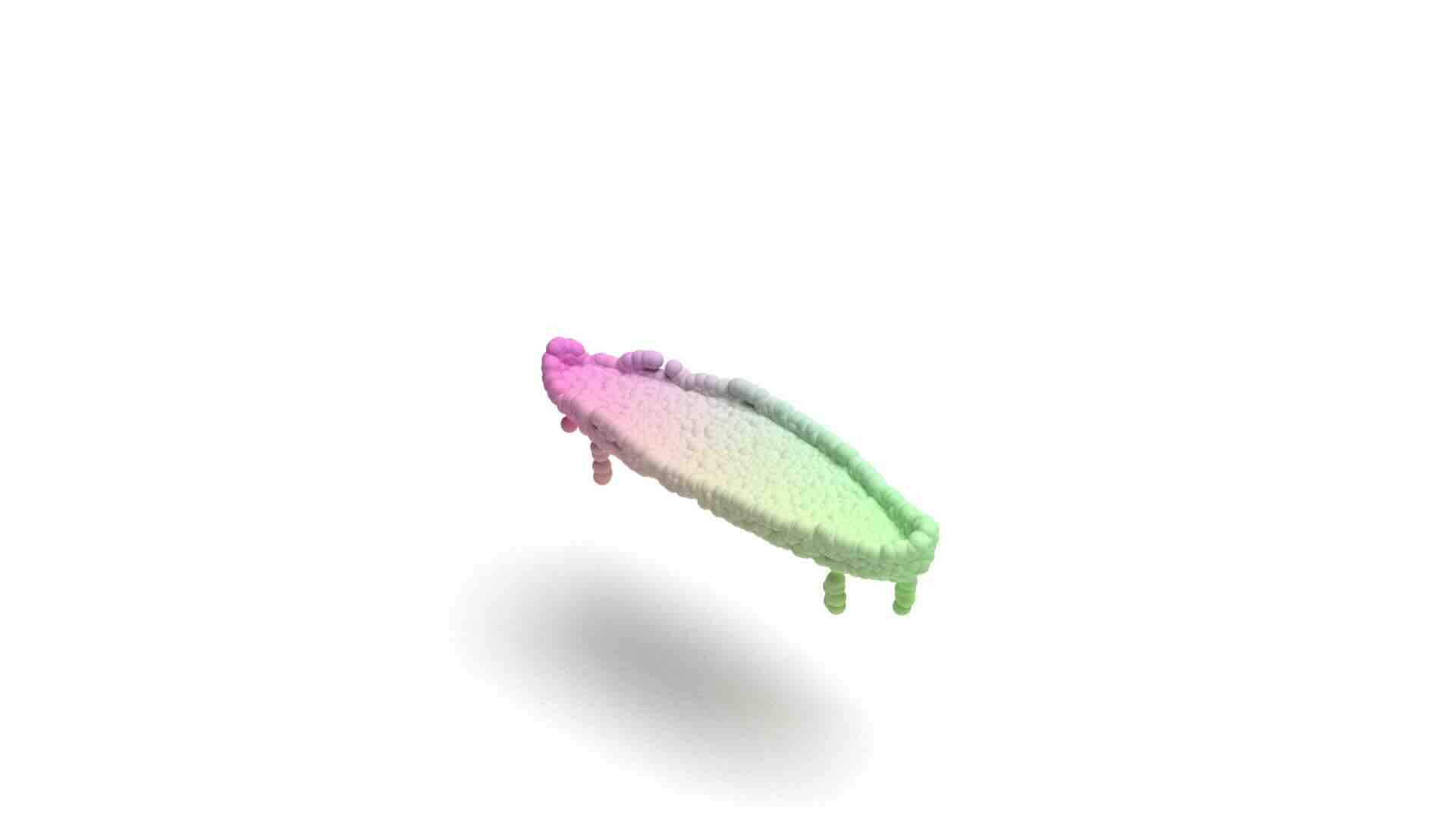} &
        \includegraphics[trim={15cm 0.0cm 15cm 0.0cm},clip, width=0.12\textwidth]{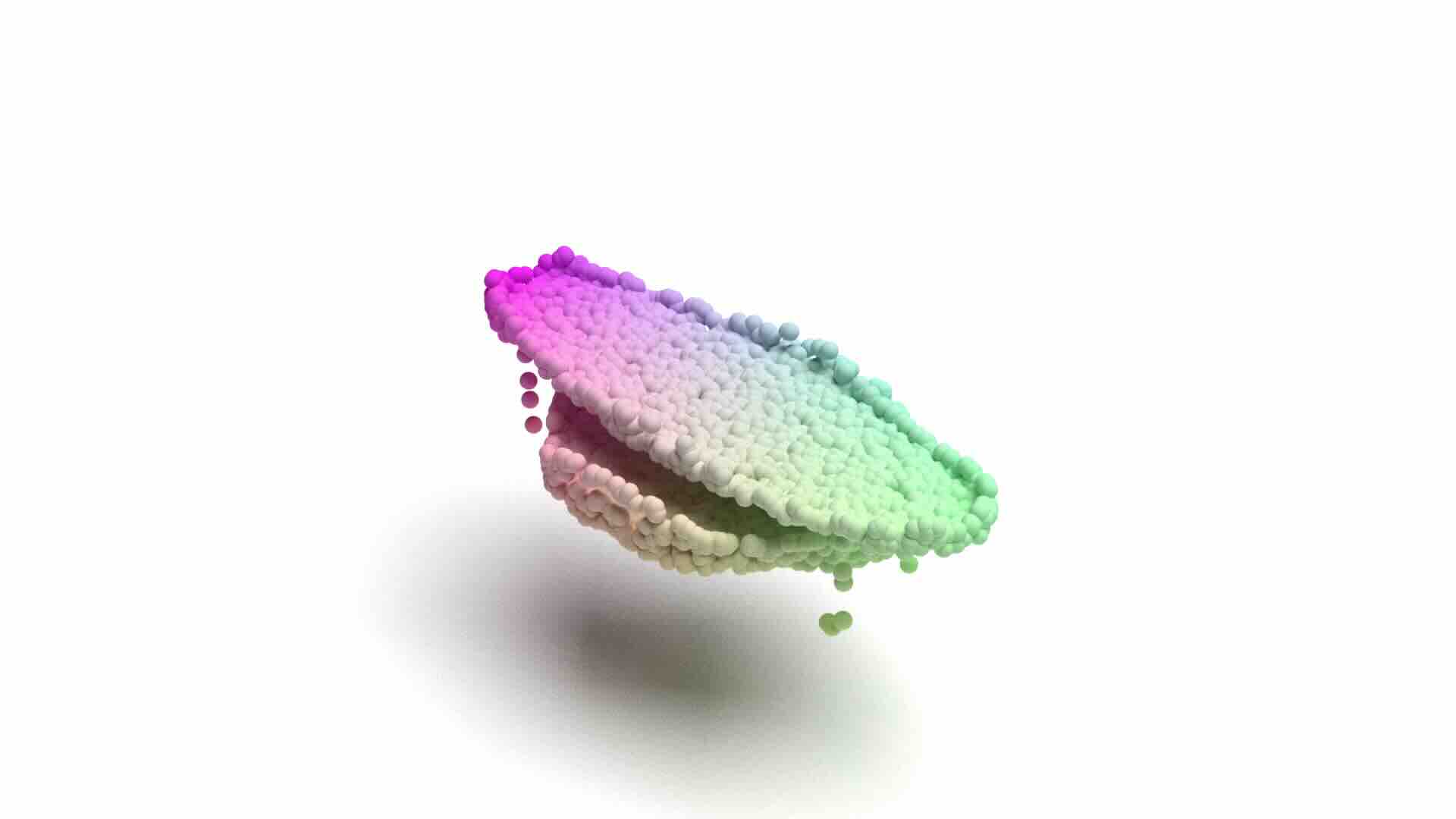} &
        \includegraphics[trim={15cm 0.0cm 15cm 0.0cm},clip, width=0.12\textwidth]{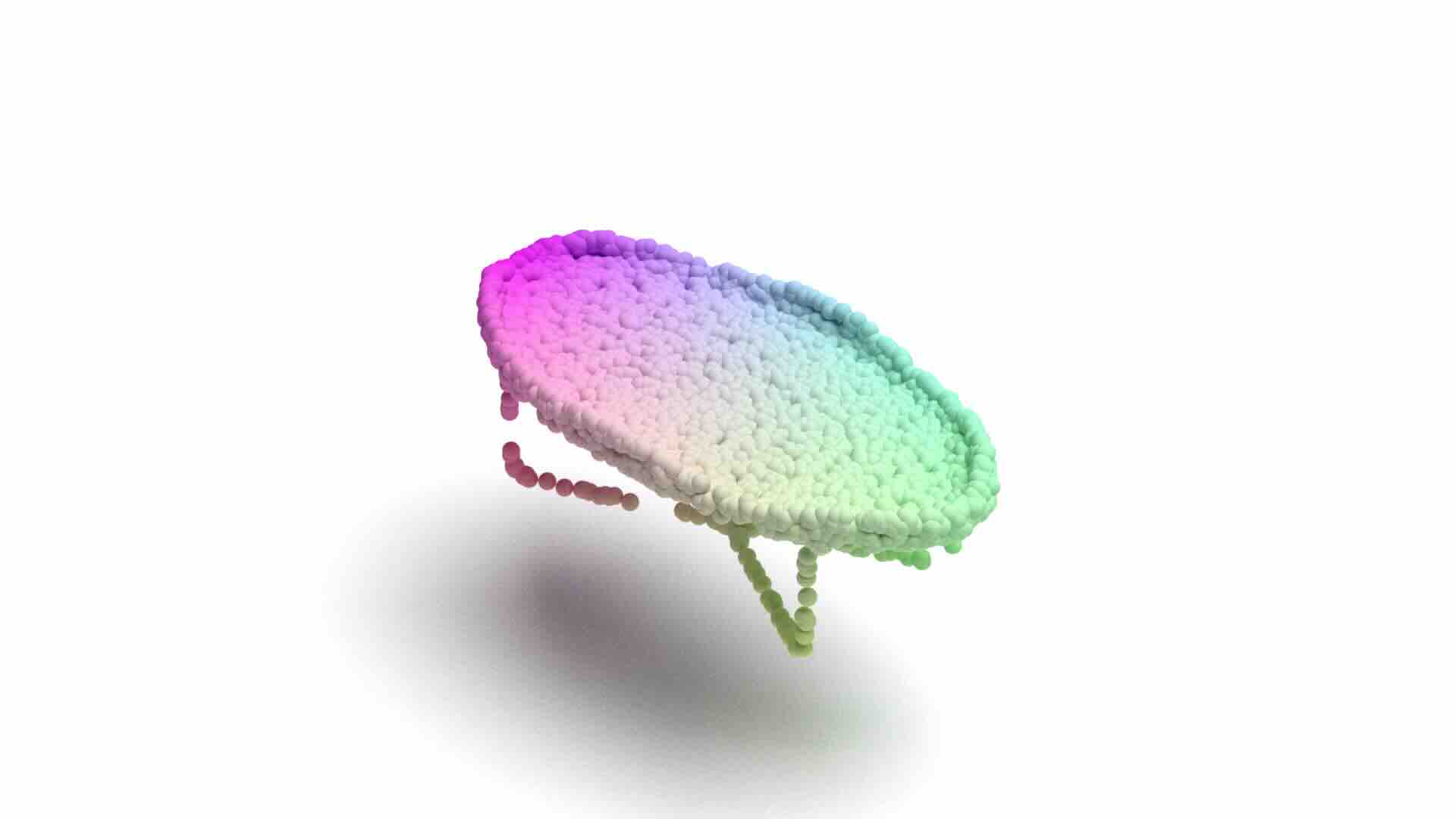} &
        \includegraphics[trim={15cm 0.0cm 15cm 0.0cm},clip,width=0.12\textwidth]{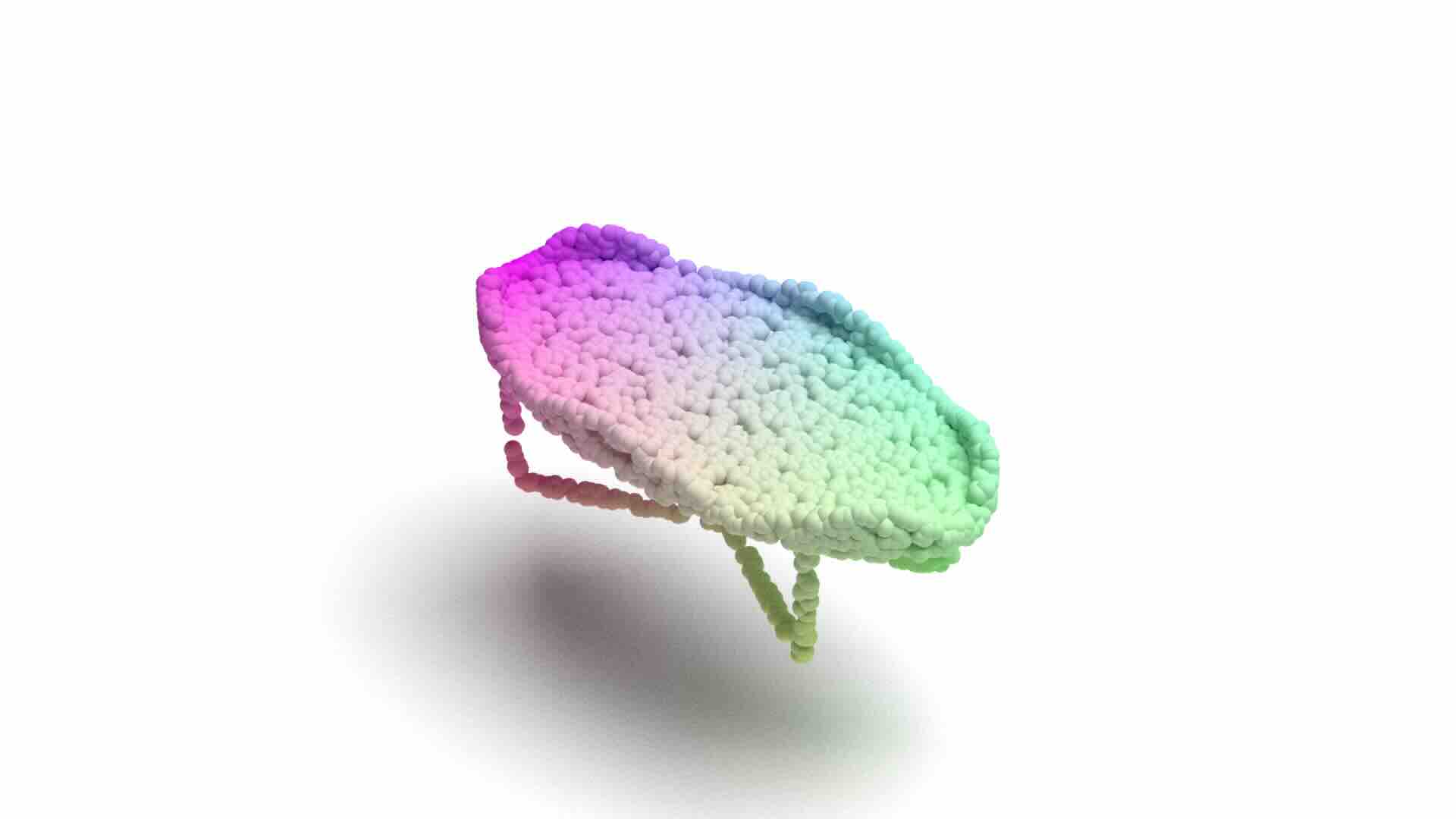} \\

        \includegraphics[trim={15cm 0.0cm 15cm 0.0cm},clip,width=0.12\textwidth]{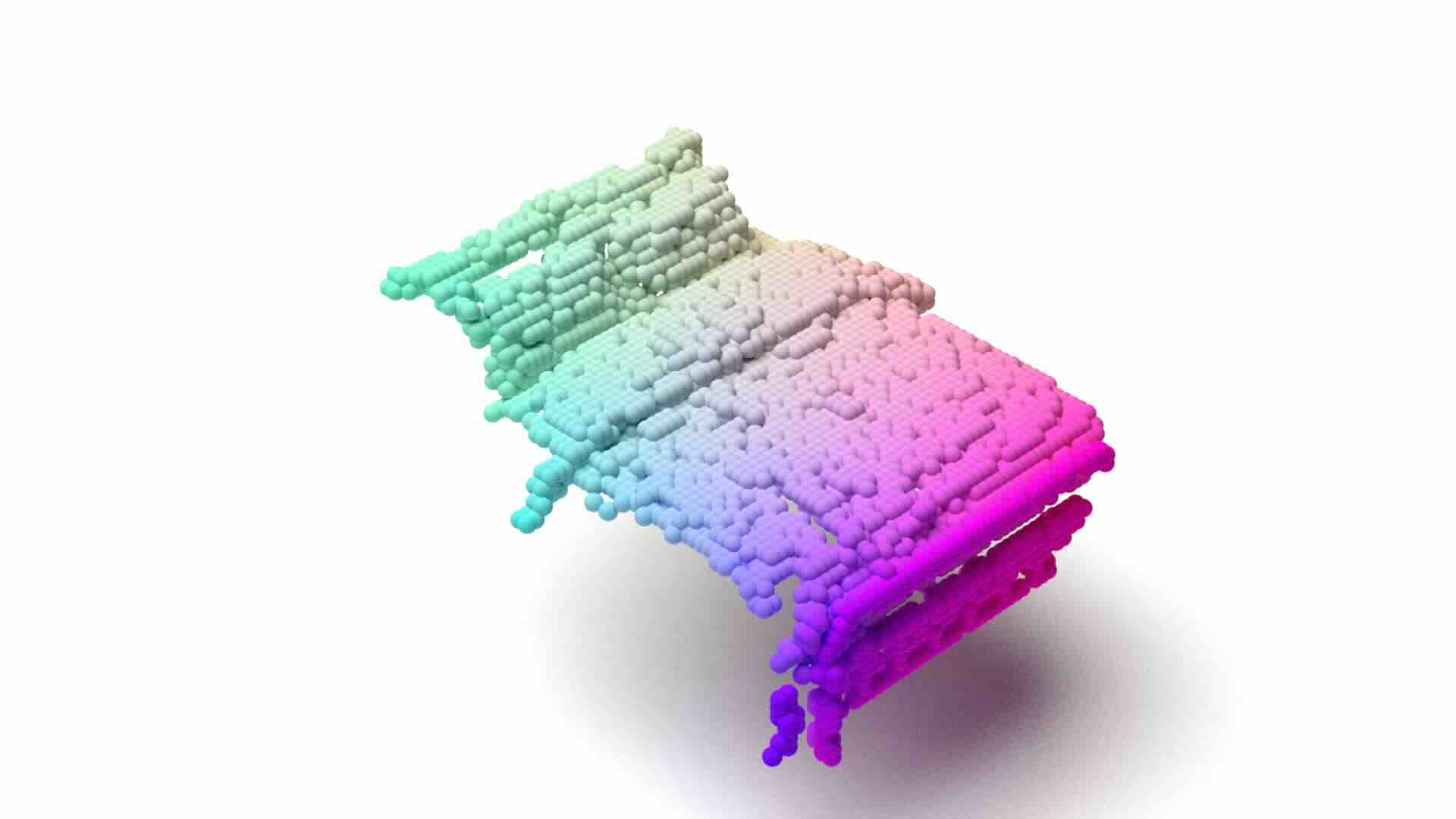} &
        
        \includegraphics[trim={15cm 0.0cm 15cm 0.0cm},clip,width=0.12\textwidth]{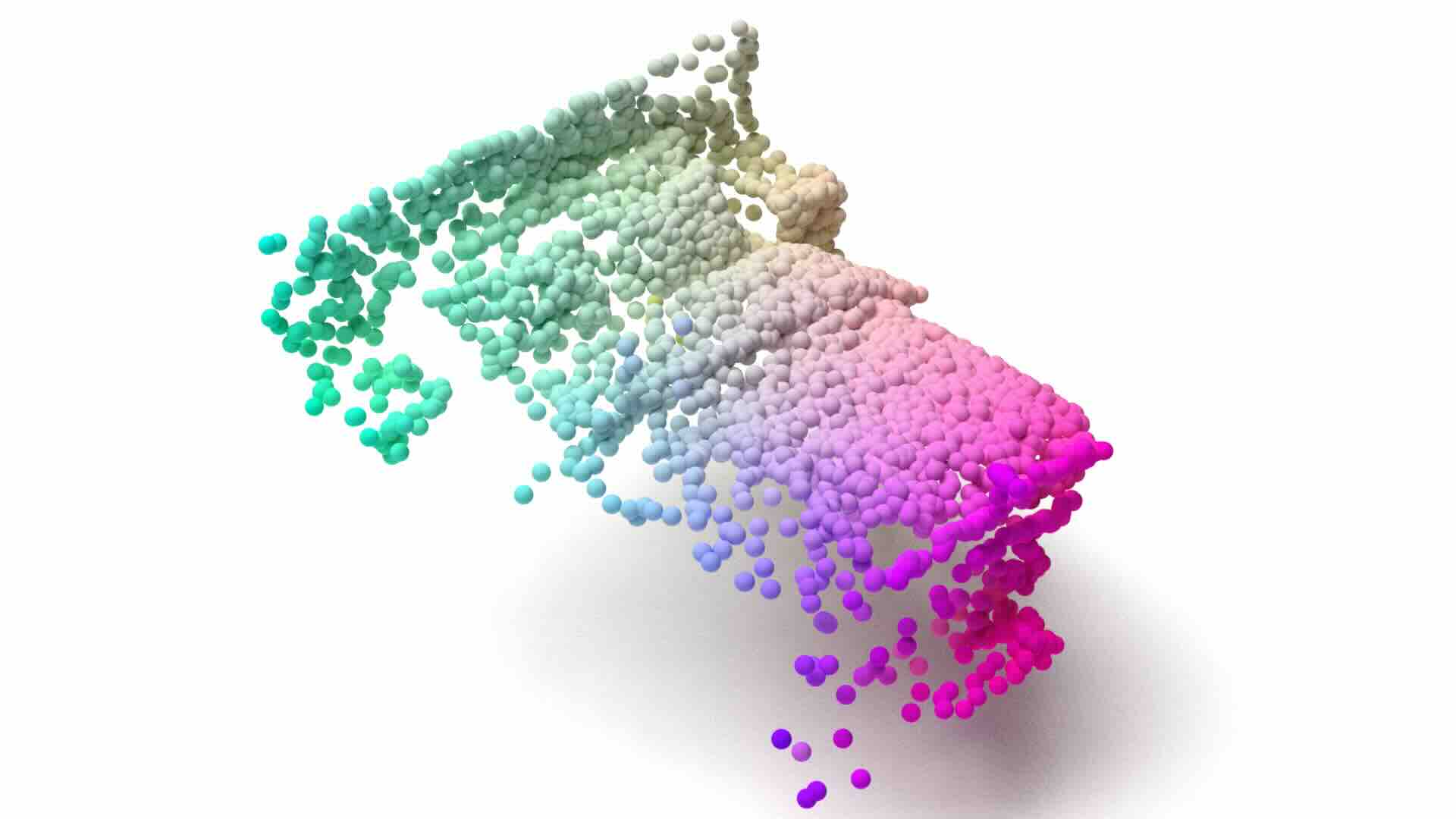} &
        \includegraphics[trim={15cm 0.0cm 15cm 0.0cm},clip, width=0.12\textwidth]{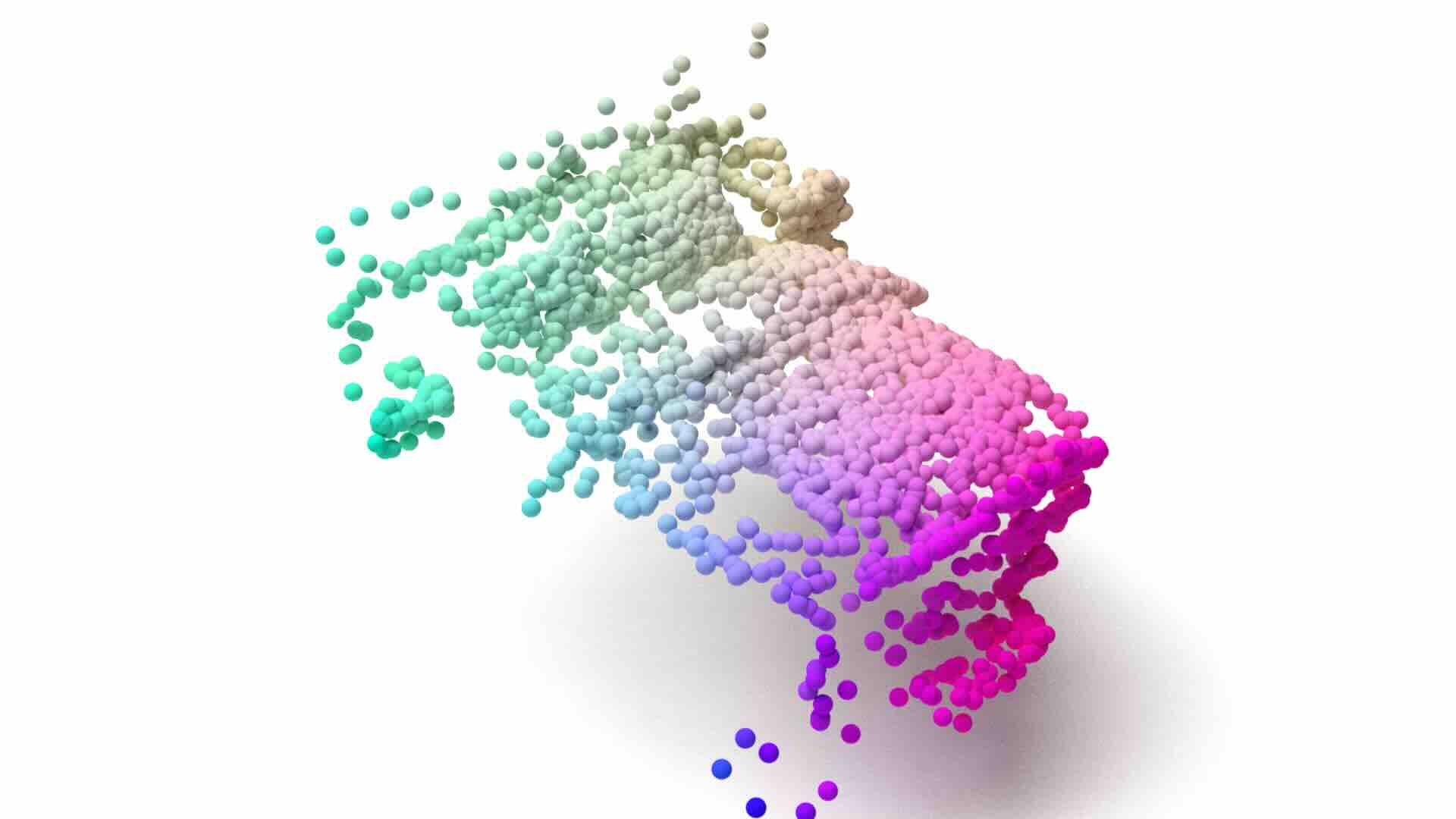} &
        \includegraphics[trim={15cm 0.0cm 15cm 0.0cm},clip, width=0.12\textwidth]{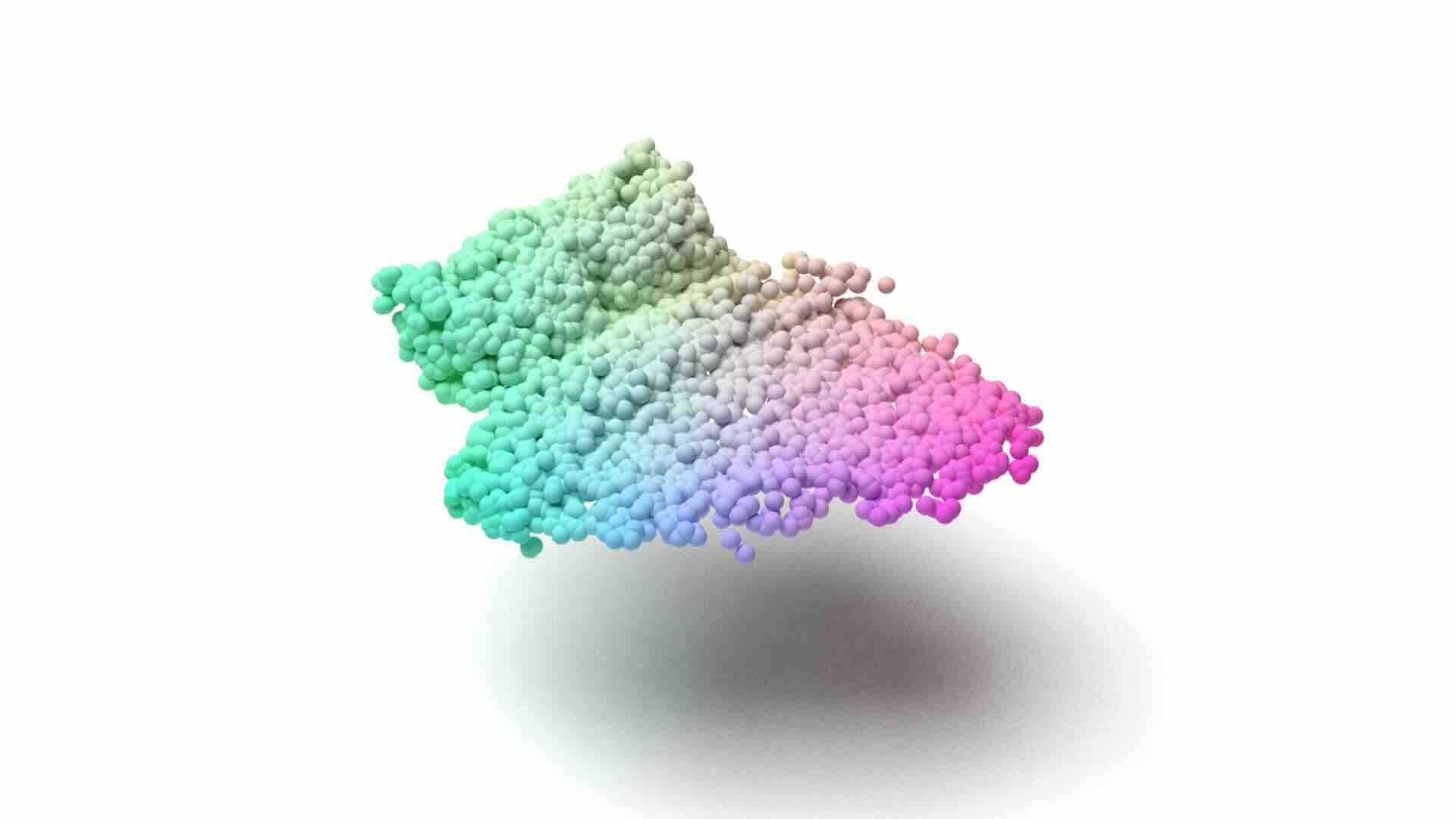} &
        \includegraphics[trim={15cm 0.0cm 15cm 0.0cm},clip, width=0.12\textwidth]{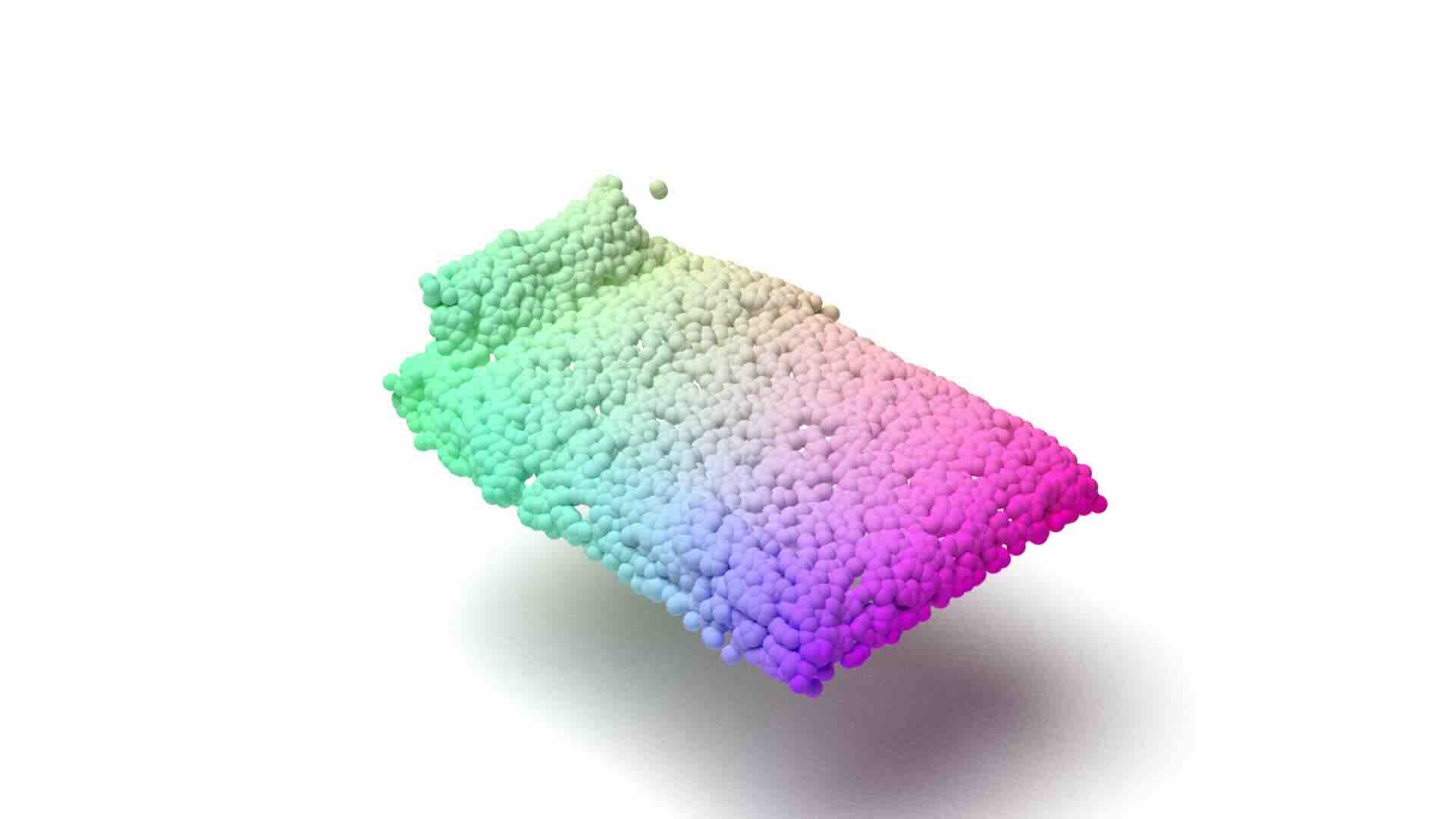} &
        \includegraphics[trim={15cm 0.0cm 15cm 0.0cm},clip, width=0.12\textwidth]{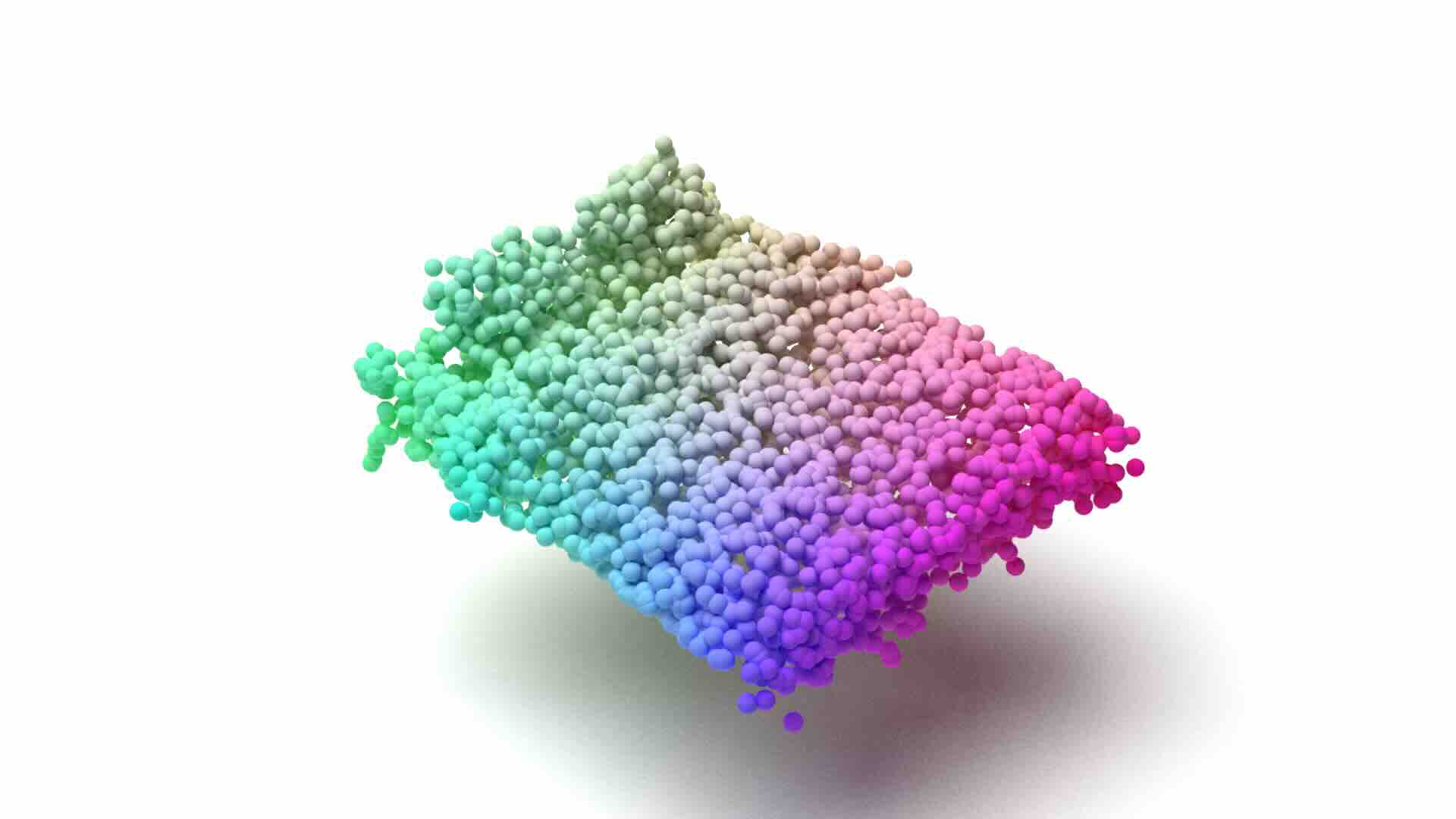} &
        \includegraphics[trim={15cm 0.0cm 15cm 0.0cm},clip, width=0.12\textwidth]{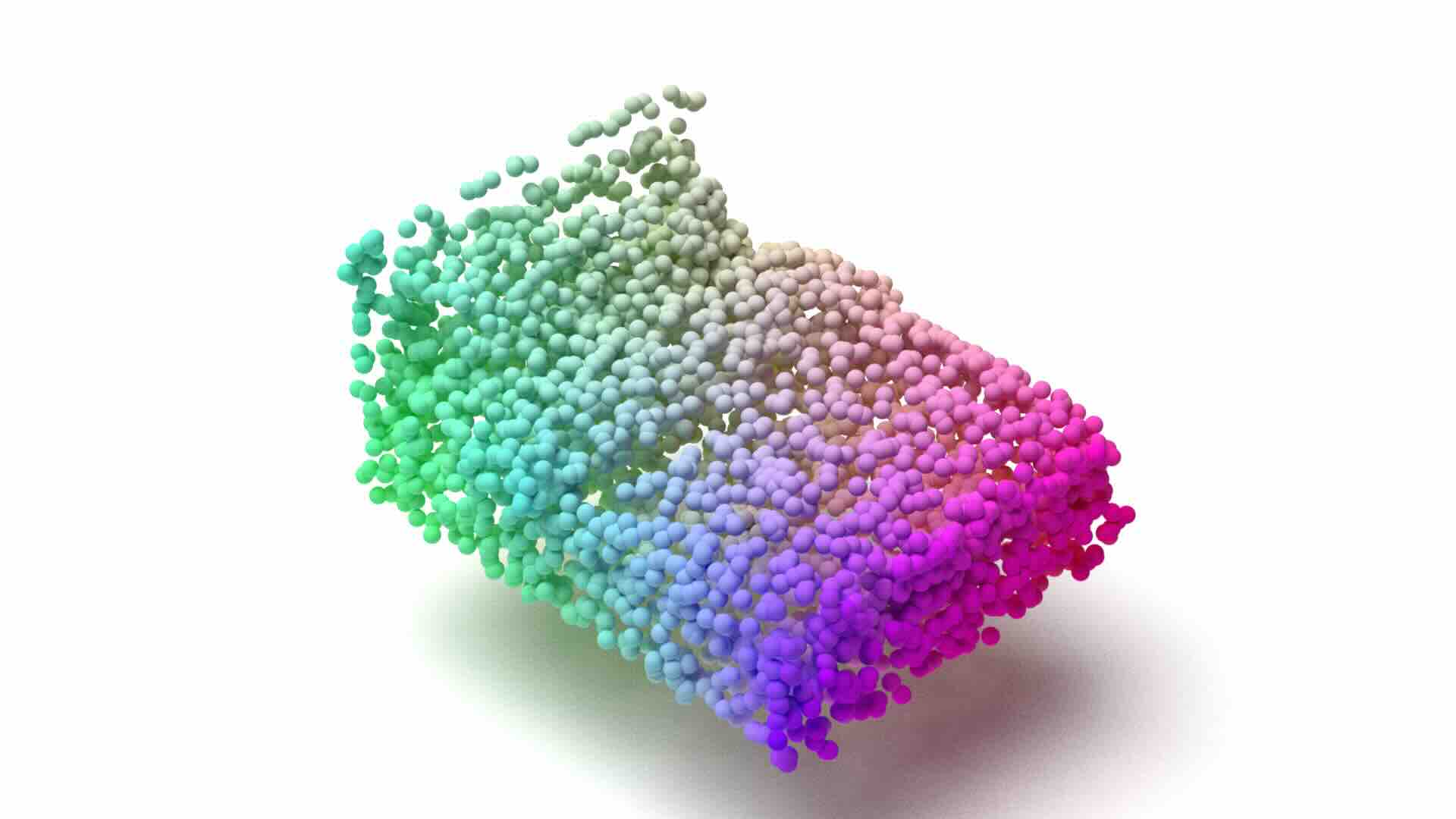} &
        \includegraphics[trim={15cm 0.0cm 15cm 0.0cm},clip,width=0.12\textwidth]{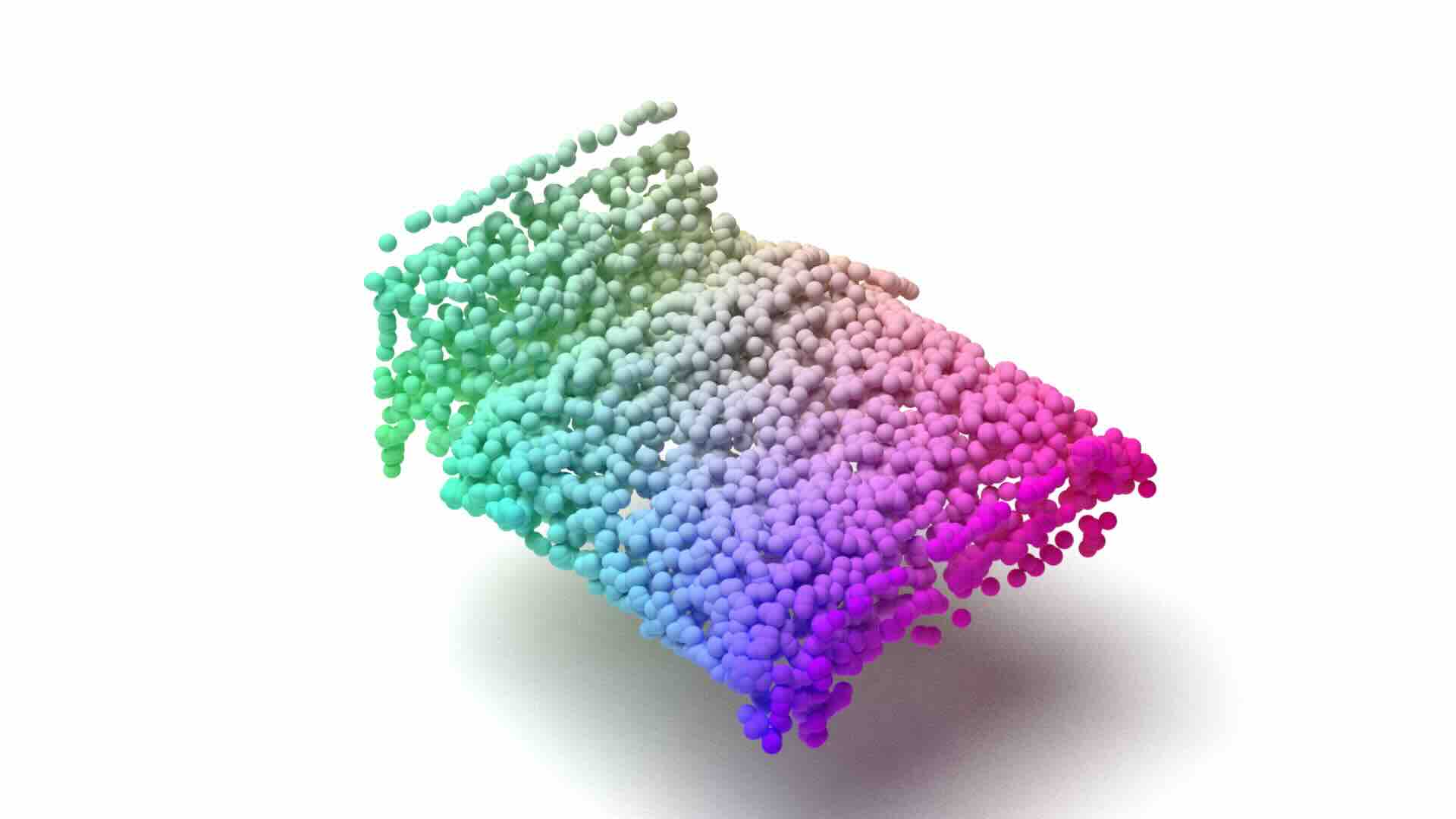} \\
        \scriptsize Fused Depths & 
\shortstack[c]{\scriptsize 3DGS-D \\ \scriptsize~\cite{kerbl20233d}} & 
\shortstack[c]{\scriptsize AGS-Mesh \\ \scriptsize~\cite{ren2025ags}} & 
\shortstack[c]{\scriptsize OM-GSD$^{\ast}$ \\ \scriptsize~\cite{lu2025orientation}} & 
\shortstack[c]{\scriptsize RVG-GSD$^{\ast}$ \\ \scriptsize~\cite{chang2025reconviagen}} & 
\shortstack[c]{\scriptsize SAM3D-GSD$^{\ast}$ \\ \scriptsize~\cite{chen2025sam}} & 
\scriptsize Ours & 
\scriptsize GT \\

        \end{tabular}
        }
    \caption{\textbf{Qualitative Comparison of Object Reconstruction on the 3D-FRONT.} Our method leverages generative priors to recover complete structures under heavy occlusion. In contrast, 3DGS optimization fails to reconstruct severely occluded furniture parts, while diffusion-based methods struggle to preserve structural integrity and topological consistency. } 
    \label{fig:main_results_3dfront}
    \vspace{-2mm}
\end{figure*}

We report \textbf{PSNR}, \textbf{SSIM}, and \textbf{LPIPS} on the selected test views for real-world datasets (ScanNet++~\cite{yeshwanth2023scannet++} and ShapeR Evaluation Dataset~\cite{siddiqui2026shaper}). The highly occluded training views highlight our method's ability to reconstruct occluded parts, while 3DGS-based baselines suffer from severe artifacts in unobserved regions and the generative method tends to produce results that diverge from the observations, particularly in maintaining texture consistency across the object surface. Our generative prior encourages plausible, high-fidelity synthesis for unseen views, resulting in a consistent representation.

\begin{figure*}[t] 
    \centering
    \setlength{\tabcolsep}{2pt}
    \vspace{-1mm}
    \resizebox{\textwidth}{!}{
    \begin{tabular}{cccccccc}
        \includegraphics[trim={15cm 0.0cm 15cm 0.0cm},clip,width=0.12\textwidth]{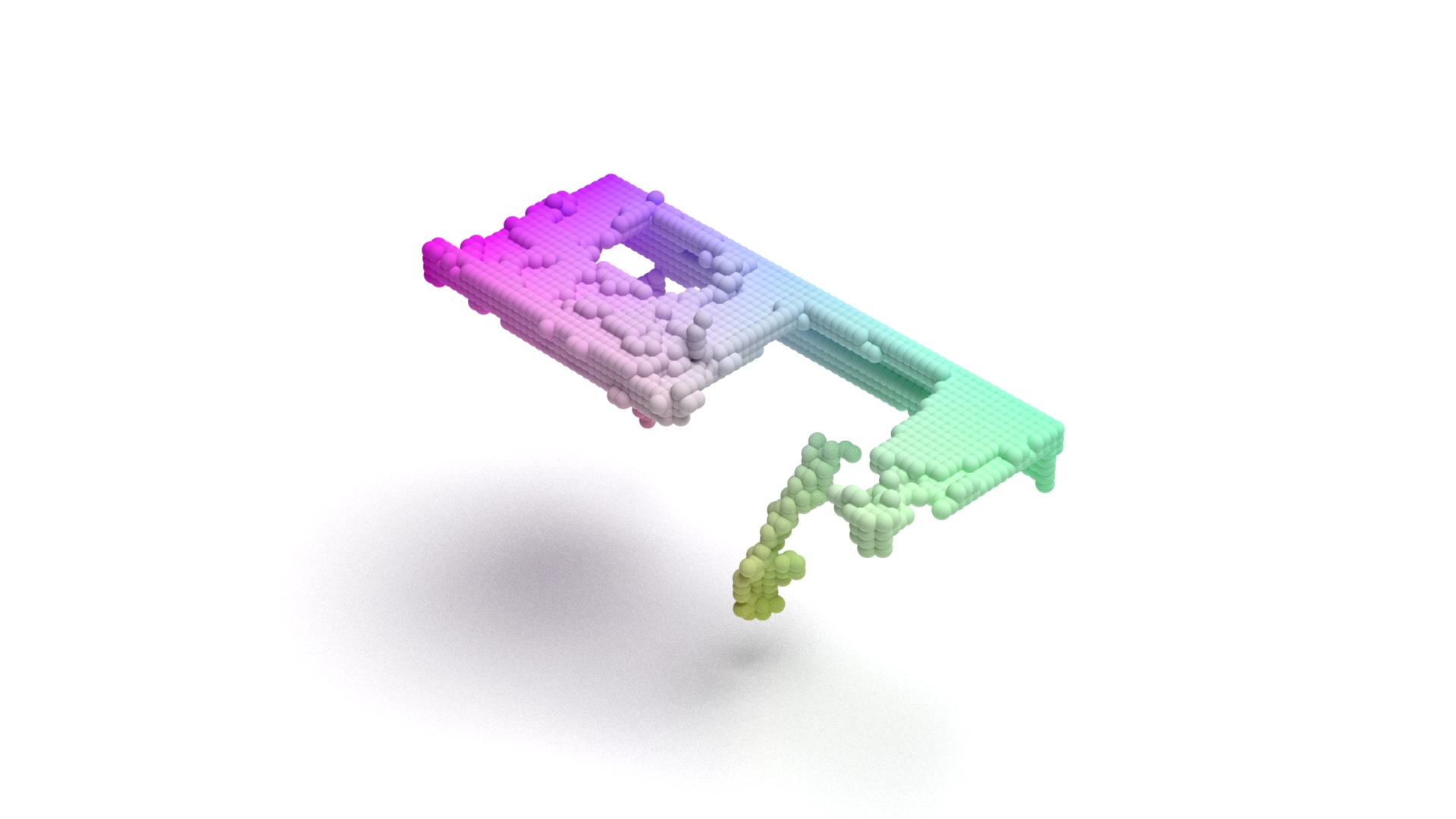} 
        &
        \includegraphics[trim={15cm 0.0cm 15cm 0.0cm},clip, width=0.12\textwidth]{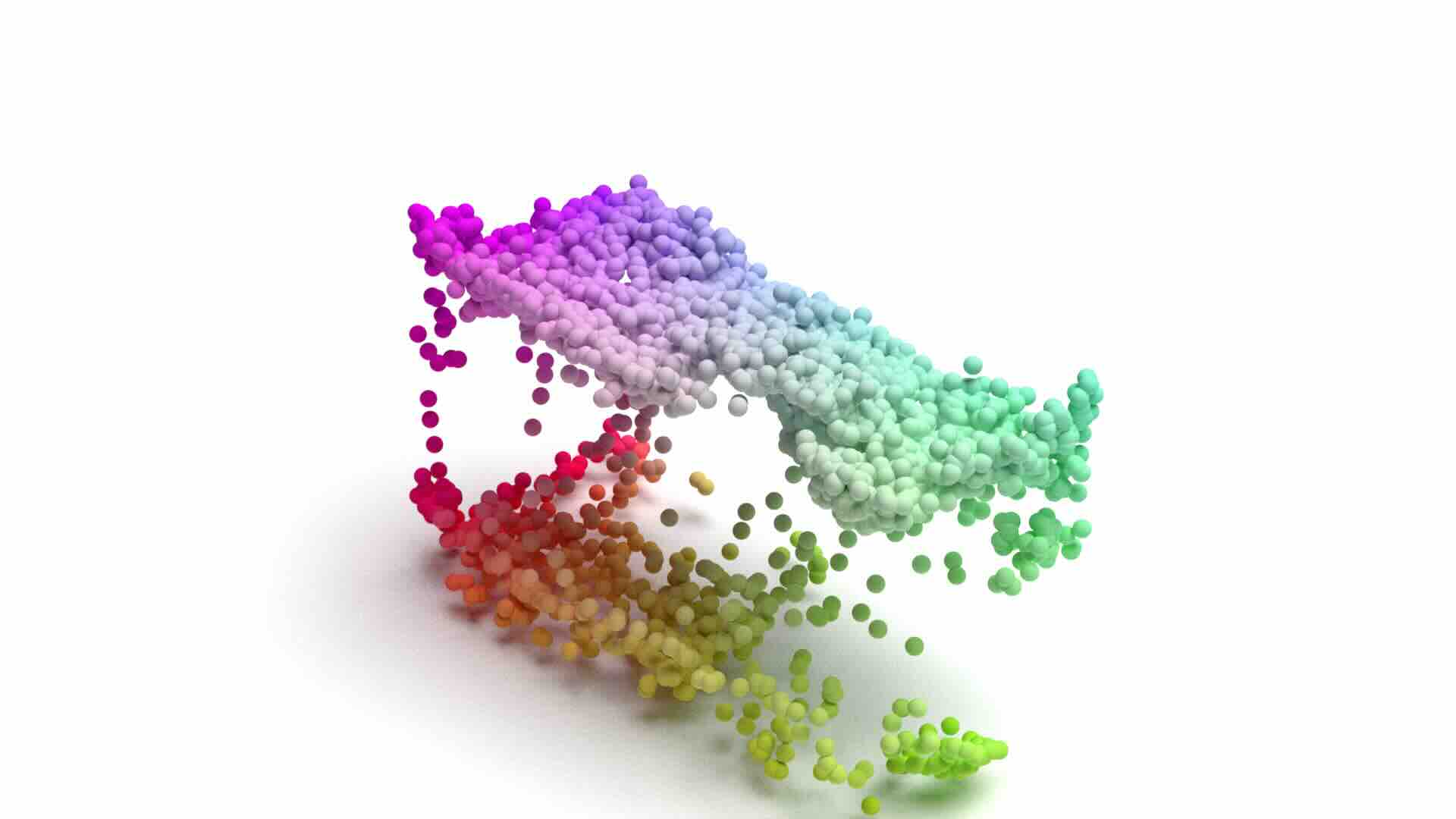} &
        \includegraphics[trim={15cm 0.0cm 15cm 0.0cm},clip, width=0.12\textwidth]{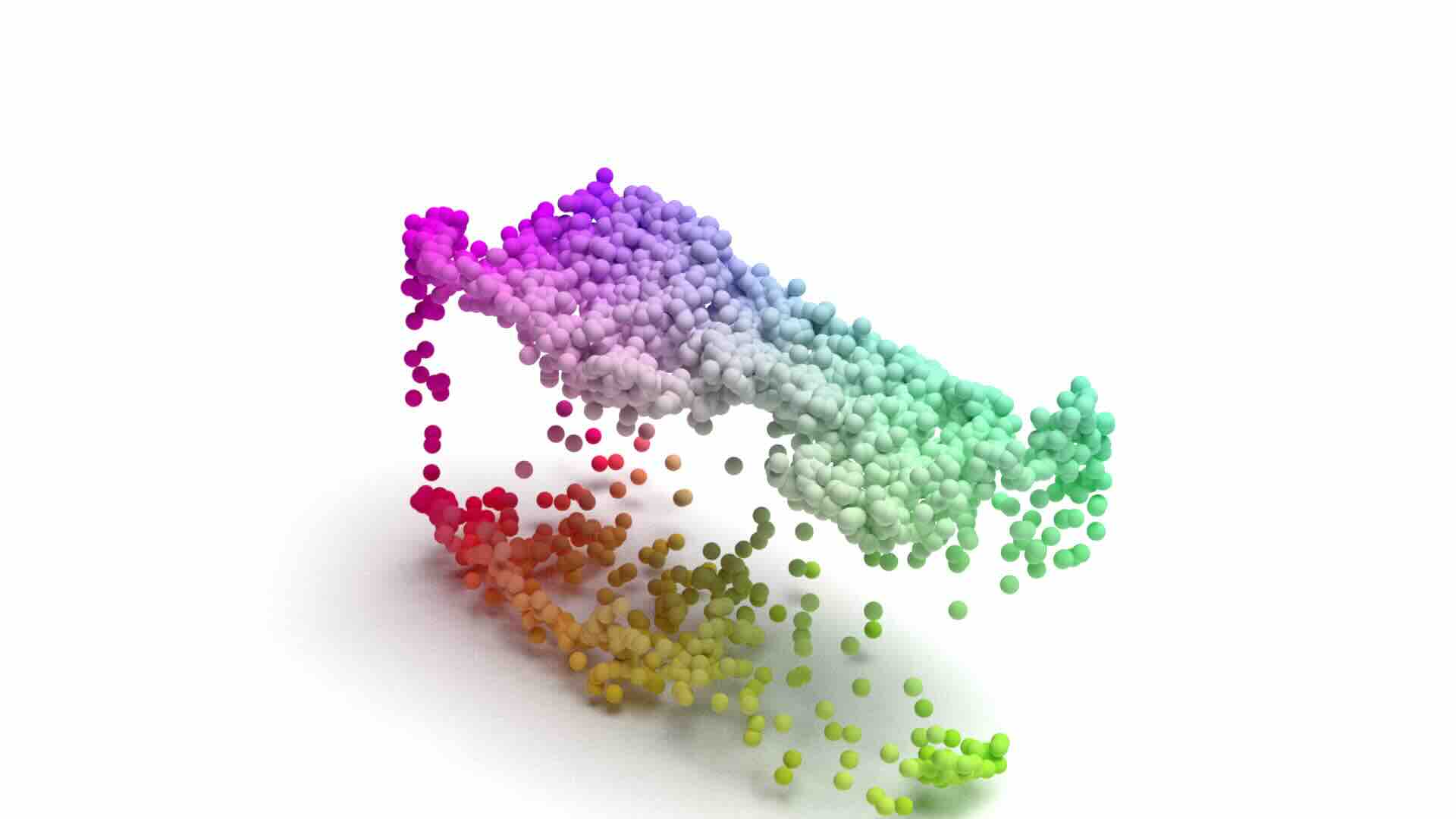} 
        &
        \includegraphics[trim={15cm 0.0cm 15cm 0.0cm},clip, width=0.12\textwidth]{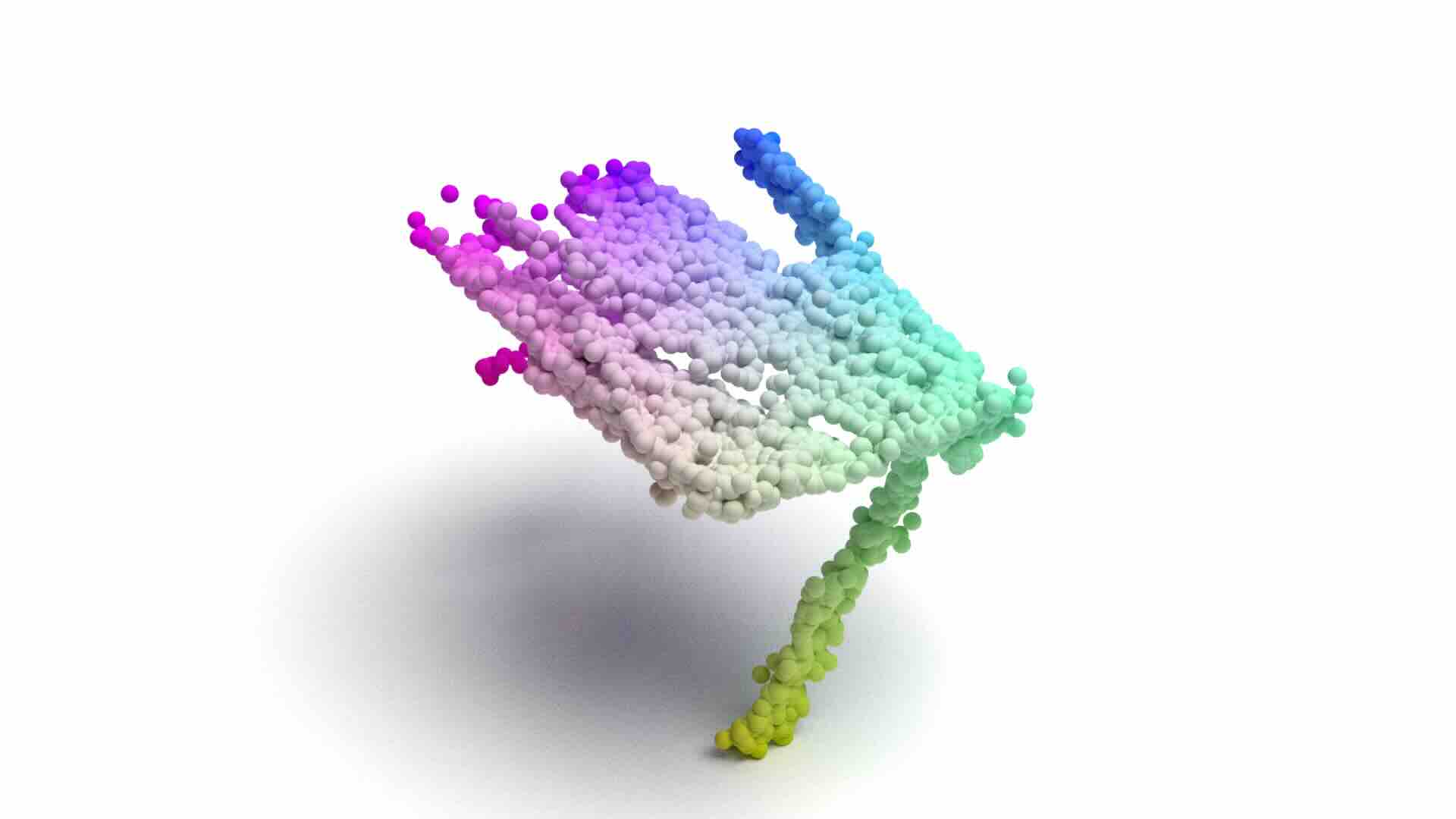} &
        \includegraphics[trim={15cm 0.0cm 15cm 0.0cm},clip,width=0.12\textwidth]{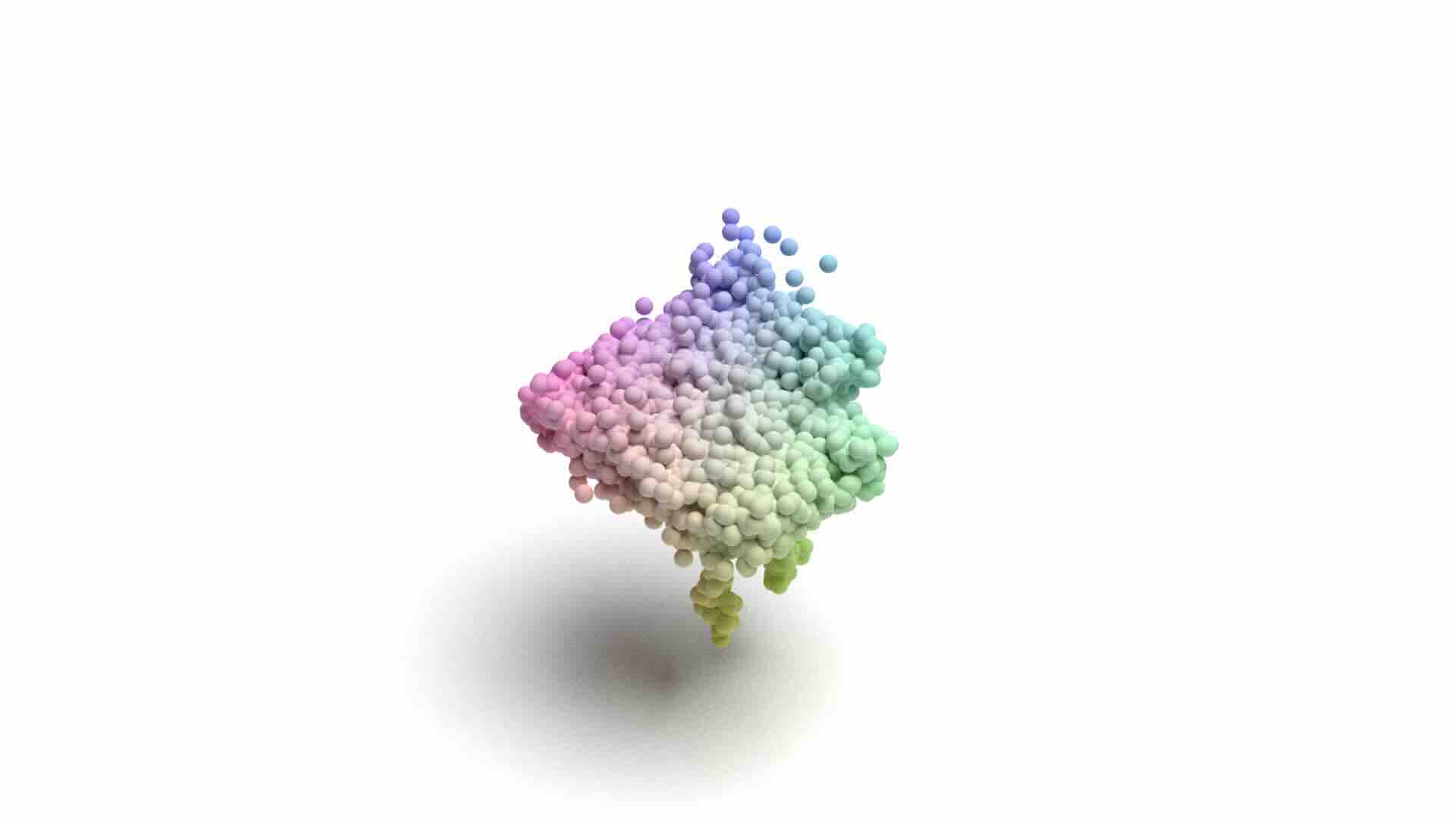} &
        \includegraphics[trim={15cm 0.0cm 15cm 0.0cm},clip,width=0.12\textwidth]{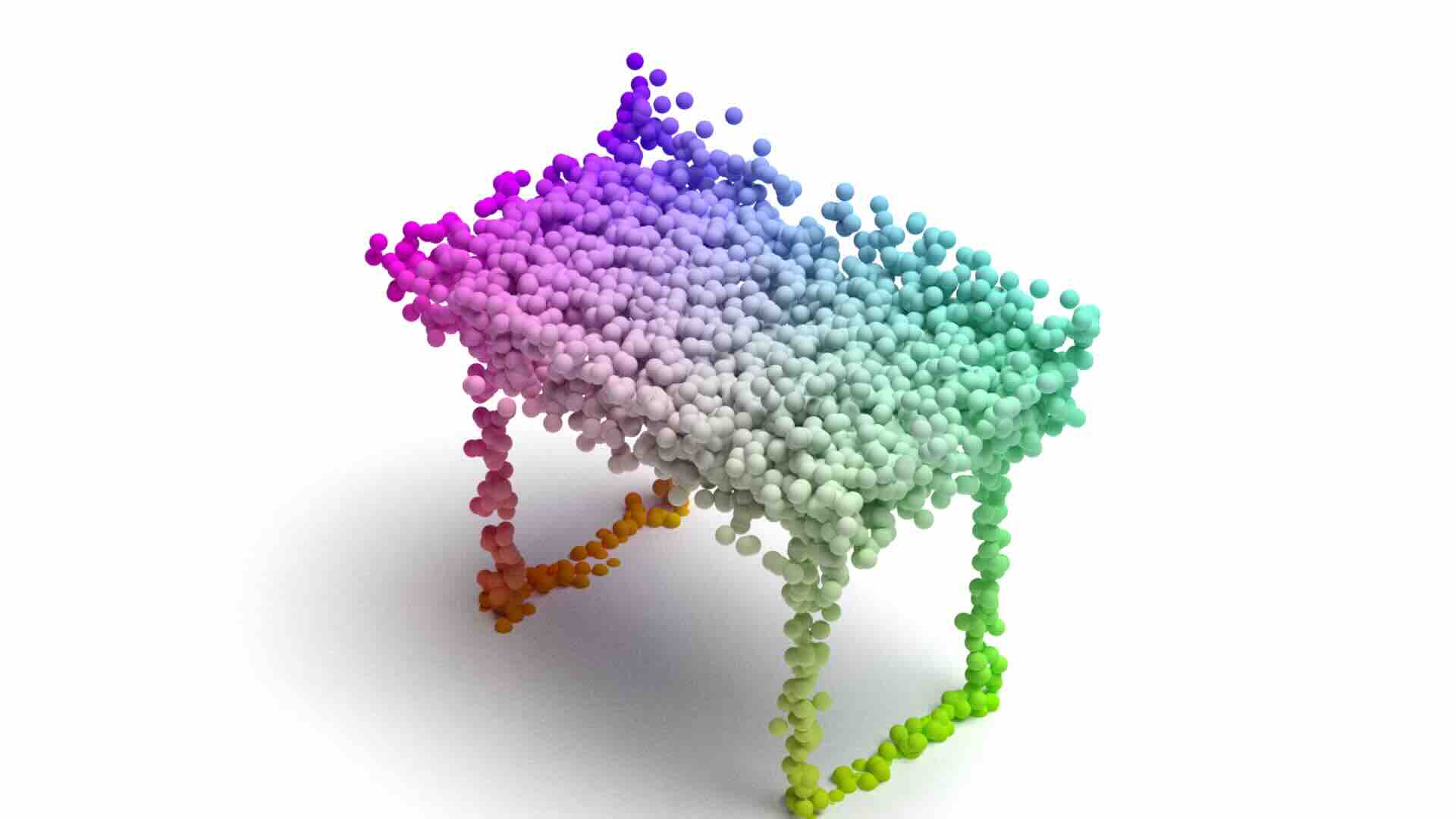} &
        \includegraphics[trim={15cm 0.0cm 15cm 0.0cm},clip,width=0.12\textwidth]{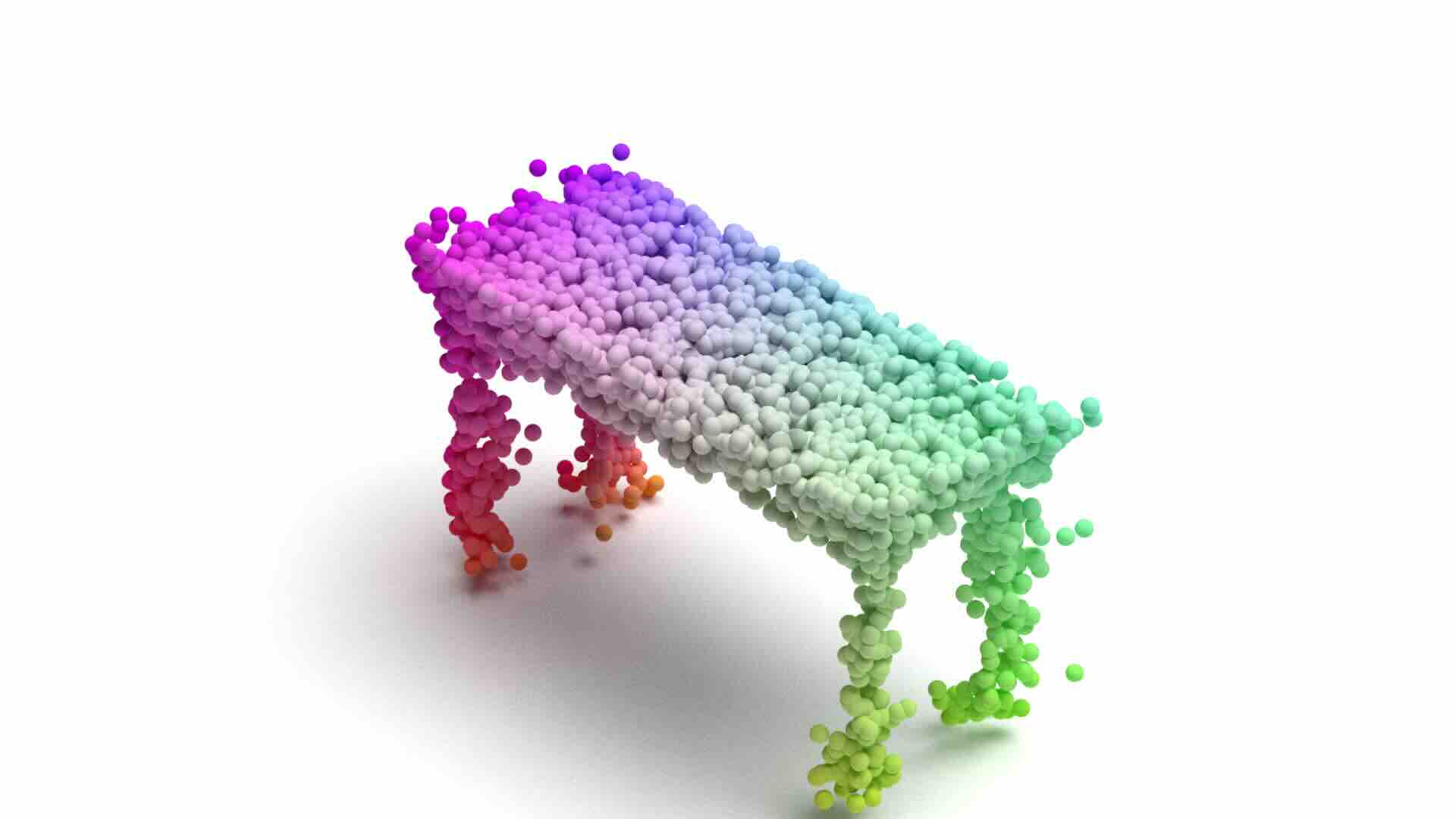} &
        \includegraphics[trim={15cm 0.0cm 15cm 0.0cm},clip,width=0.12\textwidth]{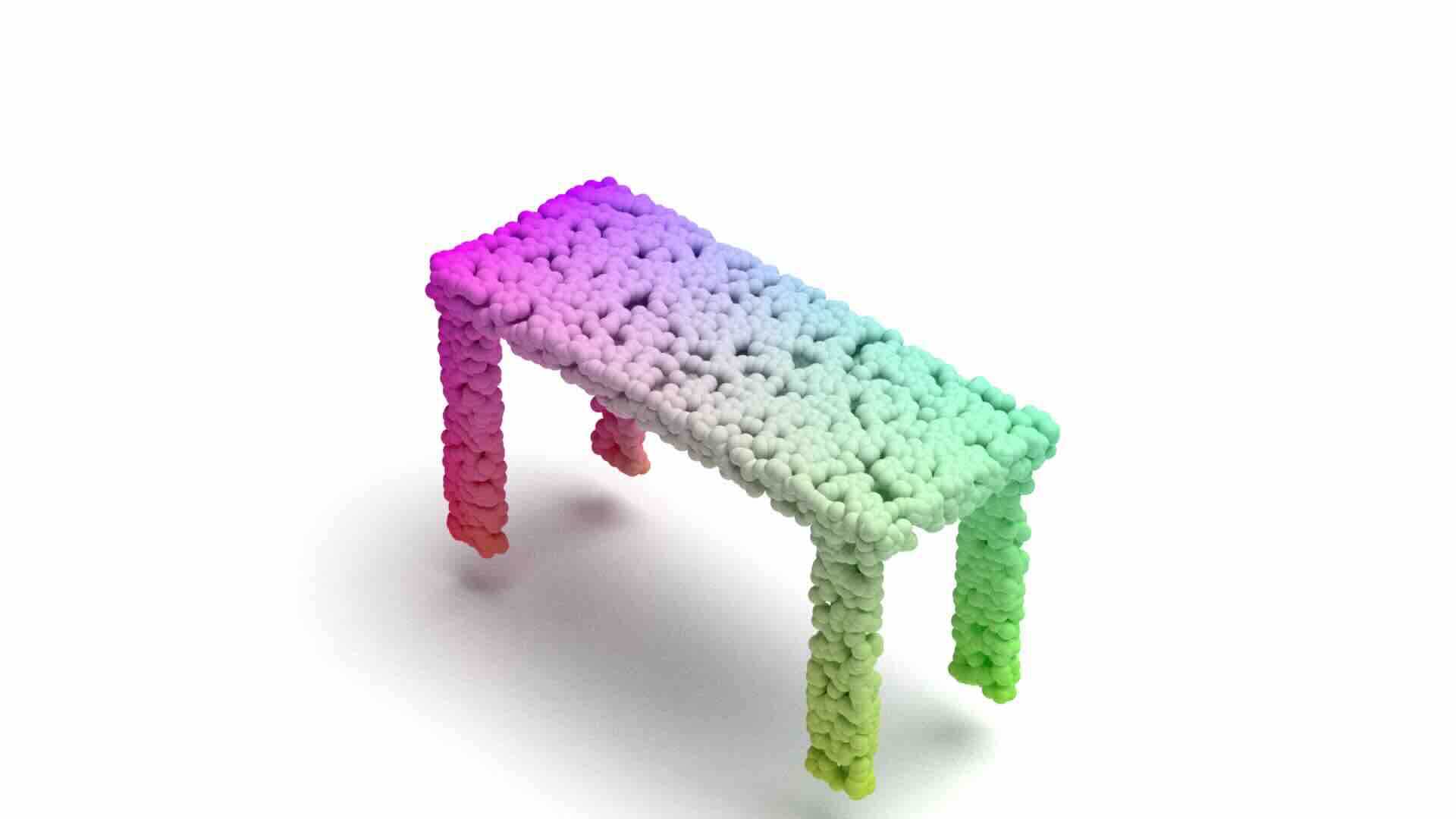}
        \\
        \includegraphics[trim={15cm 0.0cm 15cm 0.0cm},clip,width=0.12\textwidth]{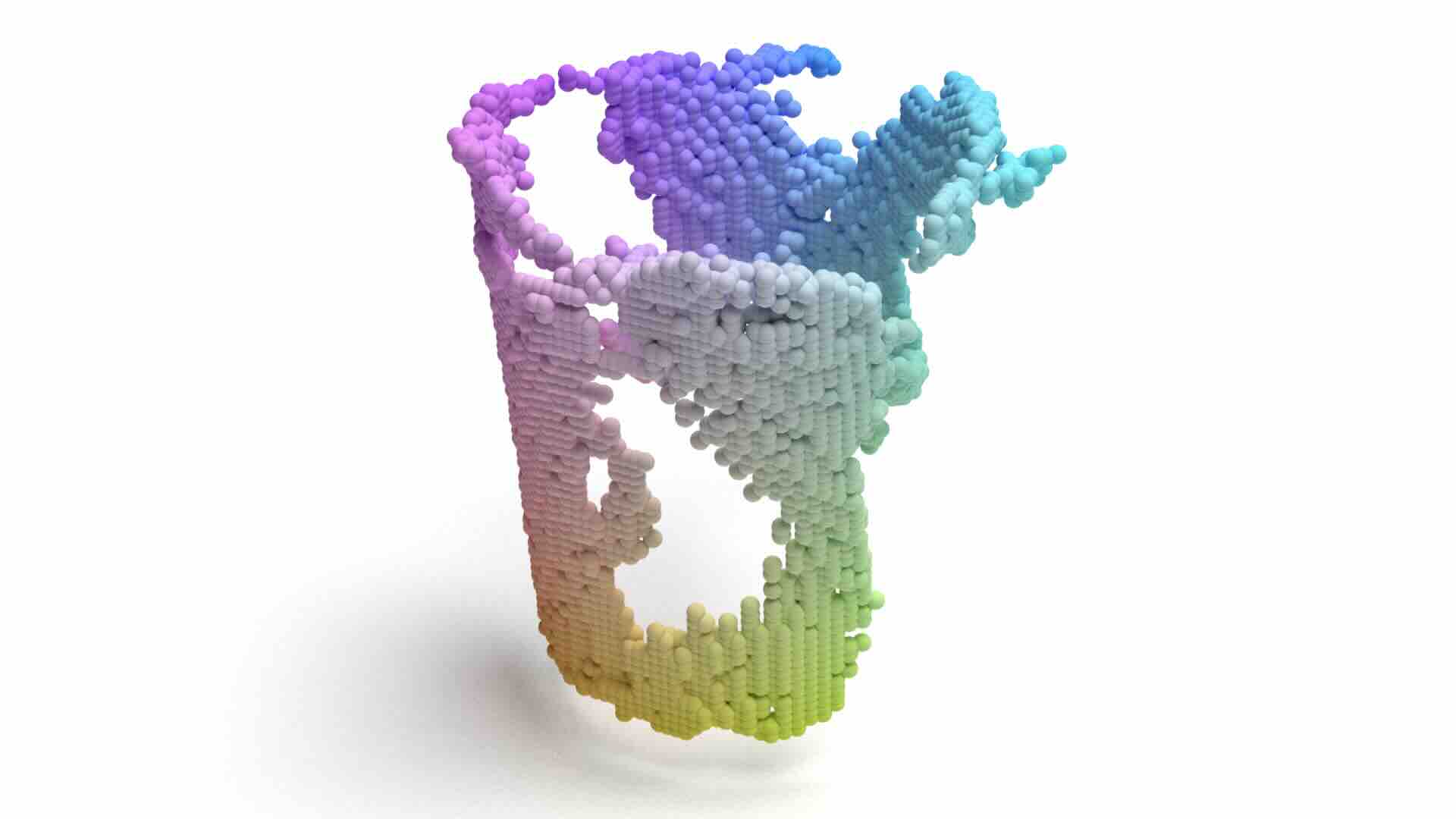} &
        \includegraphics[trim={15cm 0.0cm 15cm 0.0cm},clip, width=0.12\textwidth]{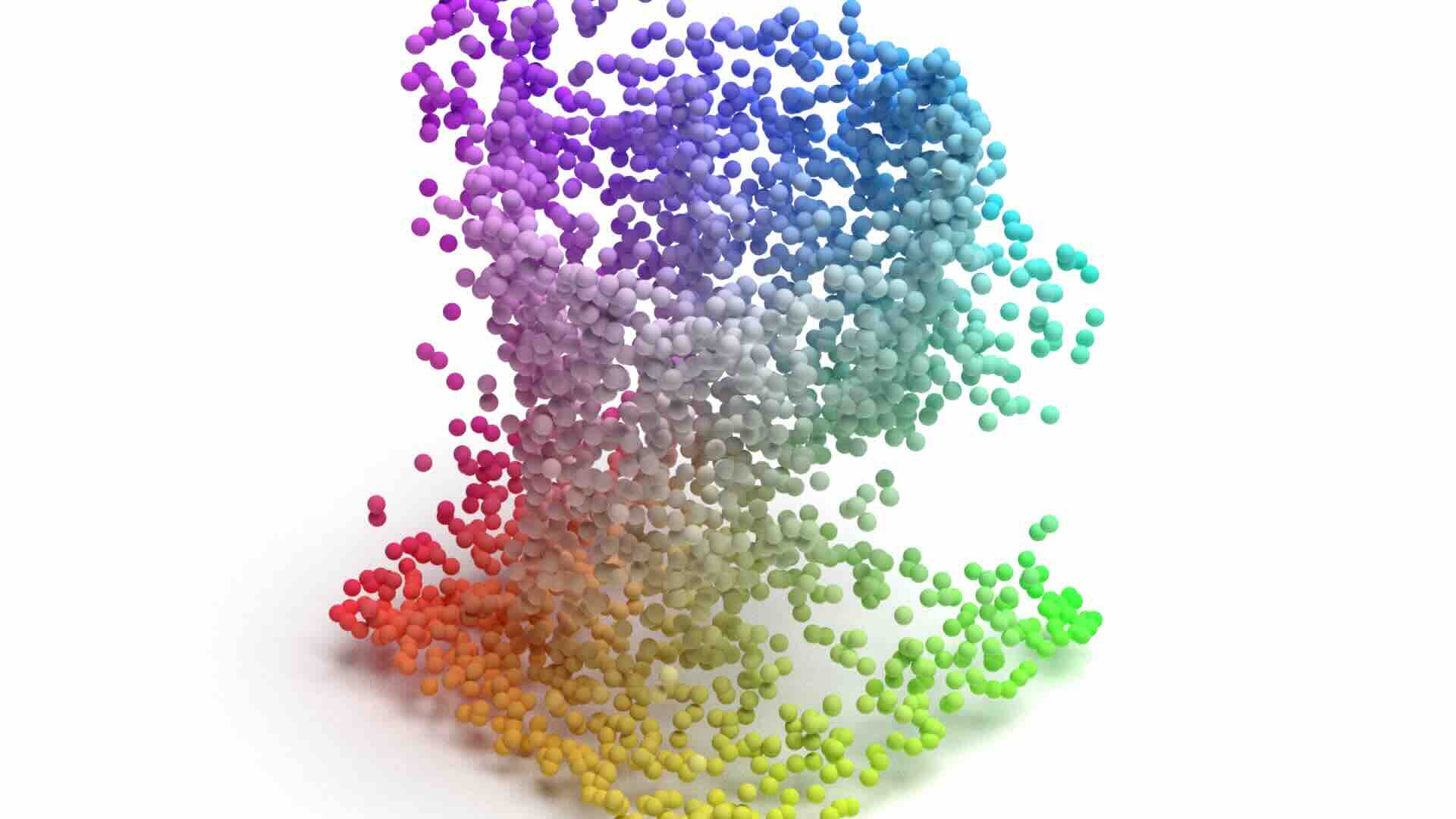} &
        \includegraphics[trim={15cm 0.0cm 15cm 0.0cm},clip, width=0.12\textwidth]{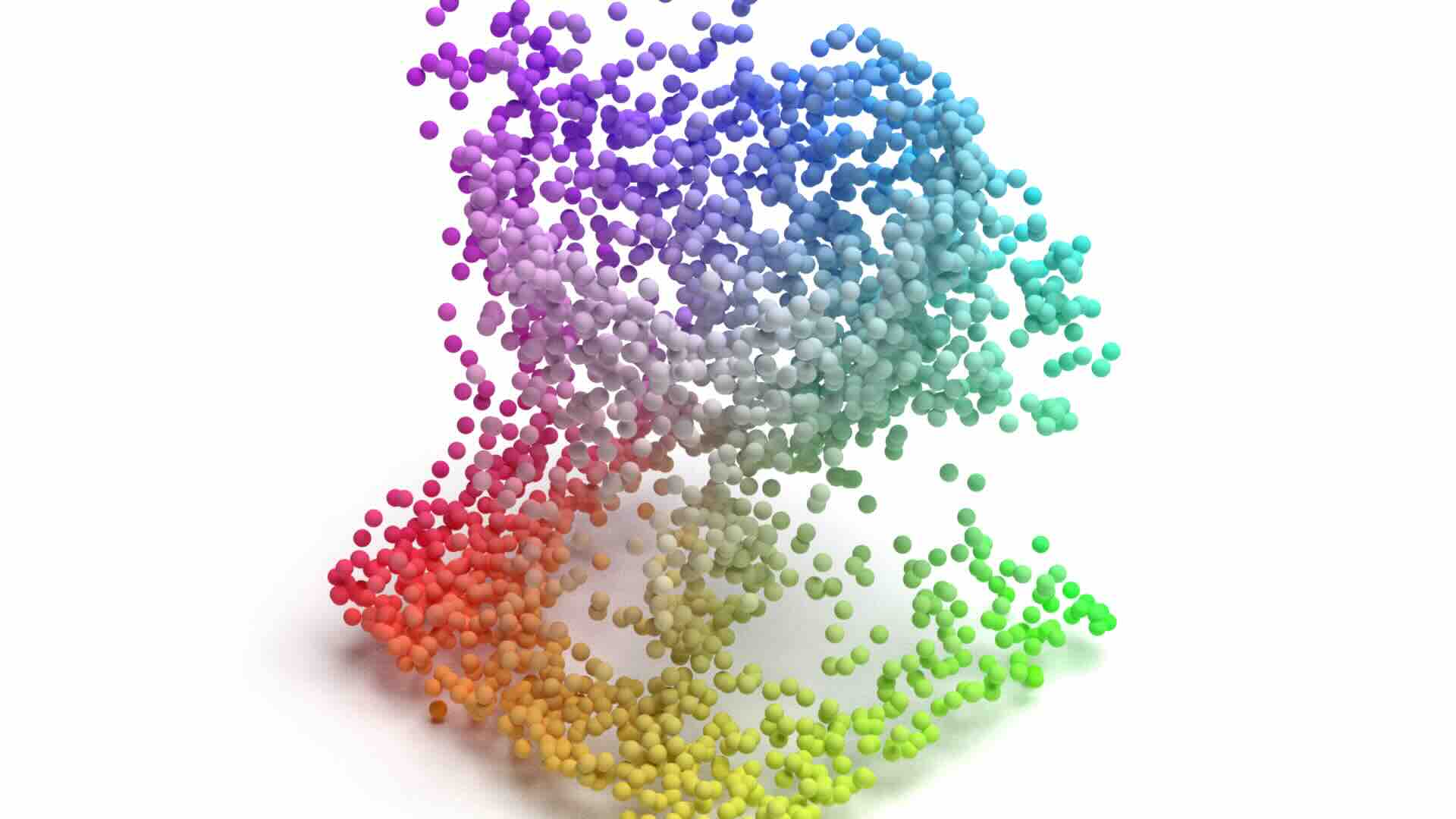} &
        \includegraphics[trim={15cm 0.0cm 15cm 0.0cm},clip, width=0.12\textwidth]{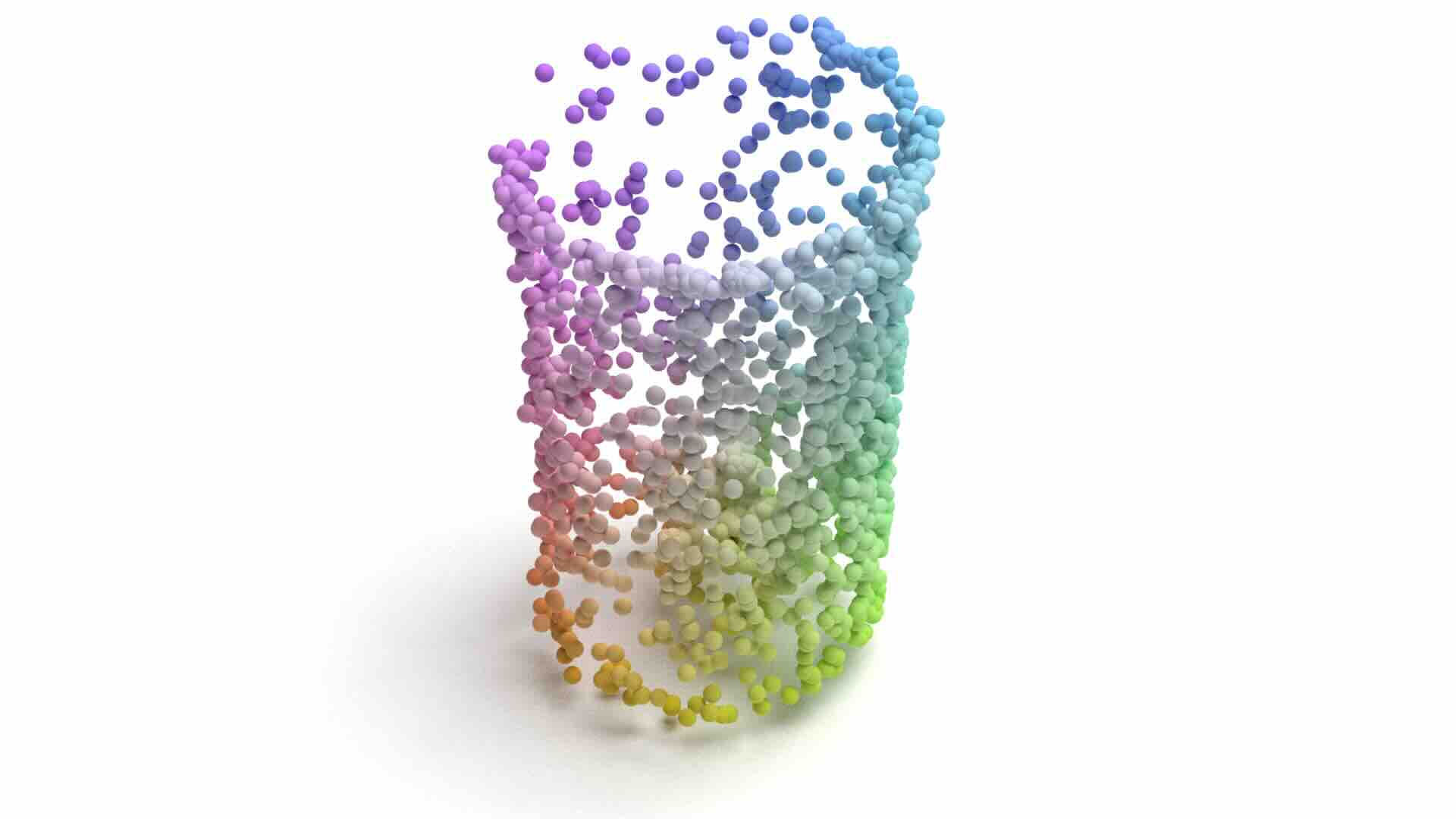} &
        \includegraphics[trim={15cm 0.0cm 15cm 0.0cm},clip, width=0.12\textwidth]{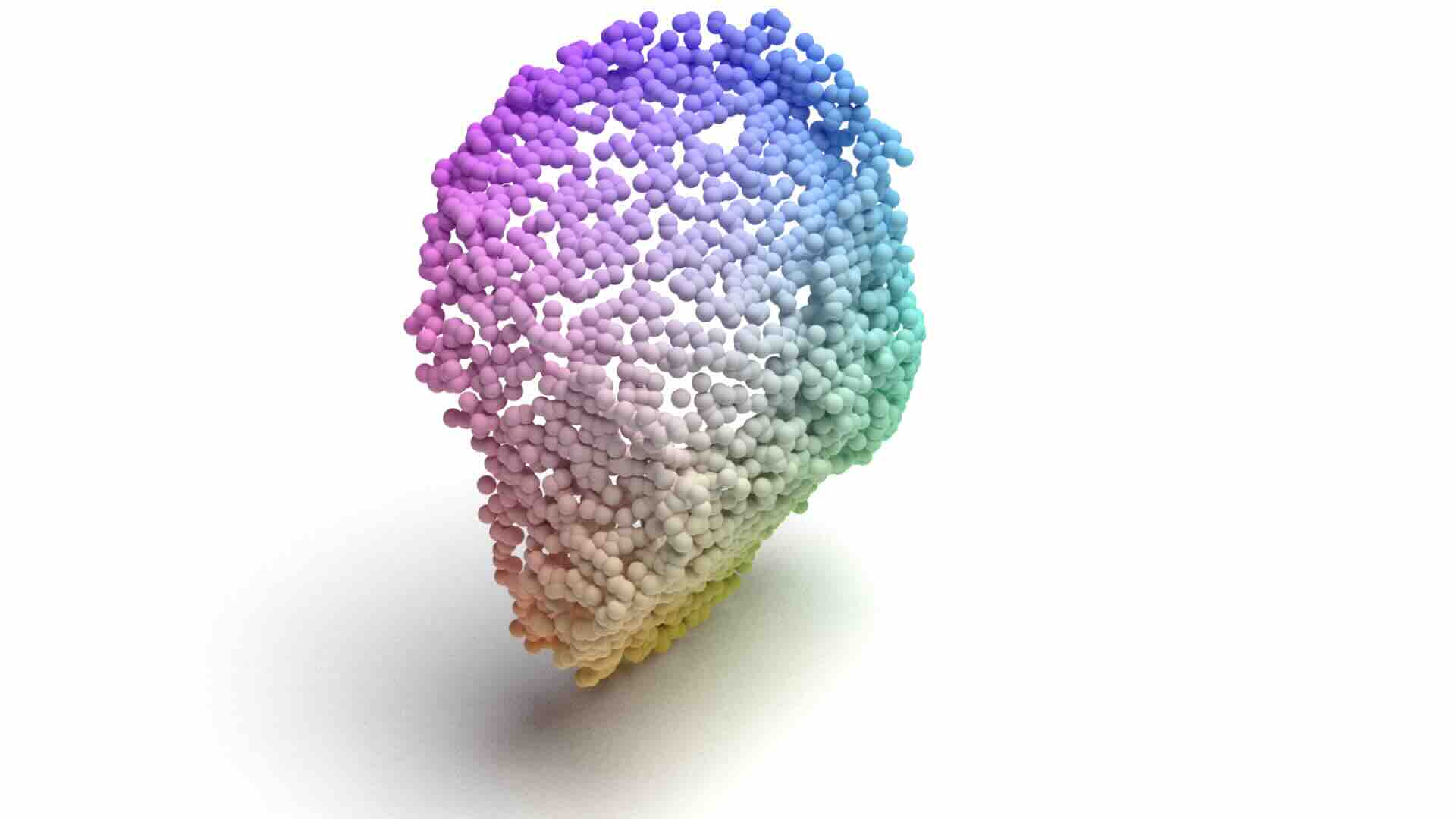} &
        \includegraphics[trim={15cm 0.0cm 15cm 0.0cm},clip, width=0.12\textwidth]{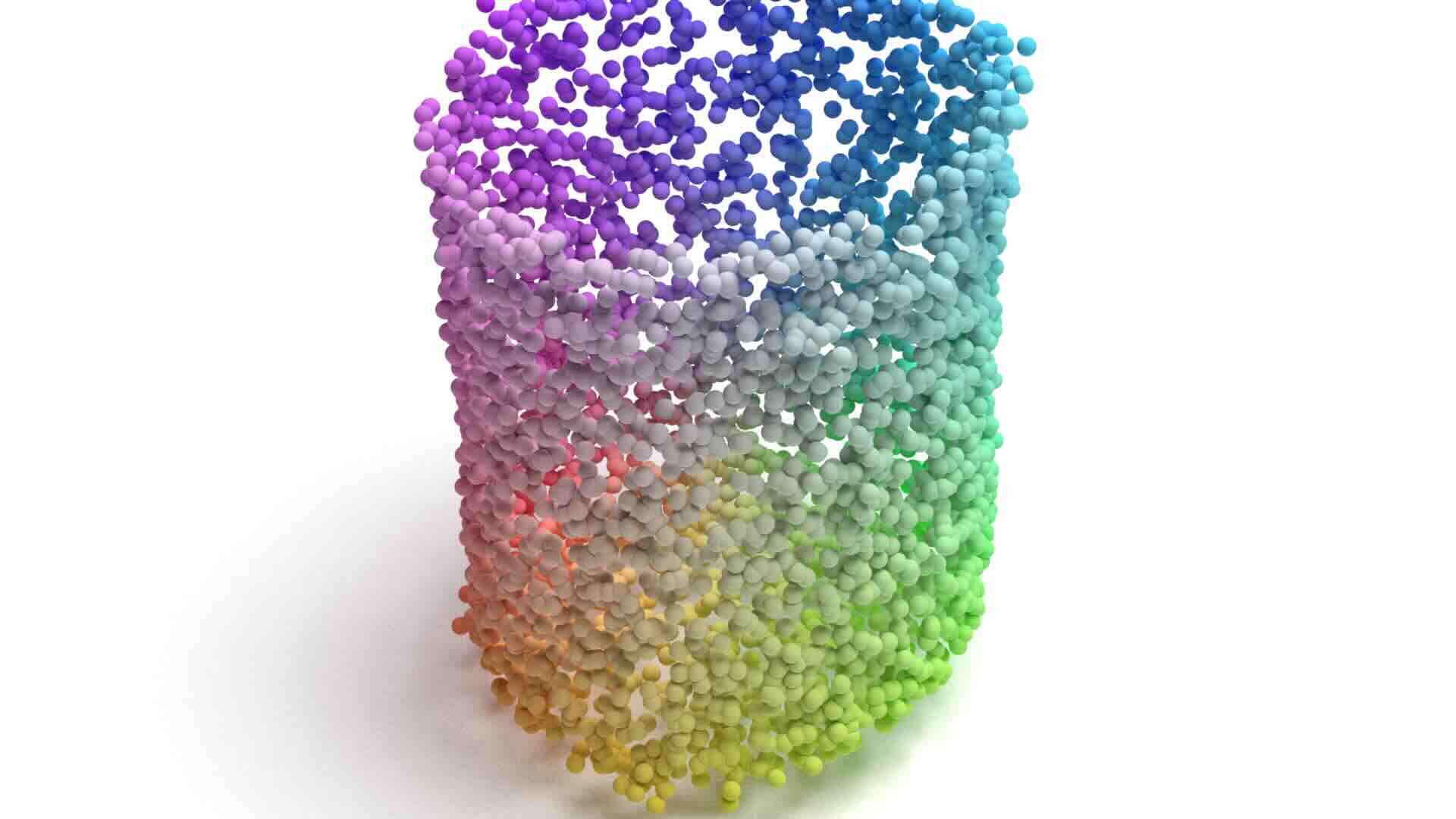} &
        \includegraphics[trim={15cm 0.0cm 15cm 0.0cm},clip, width=0.12\textwidth]{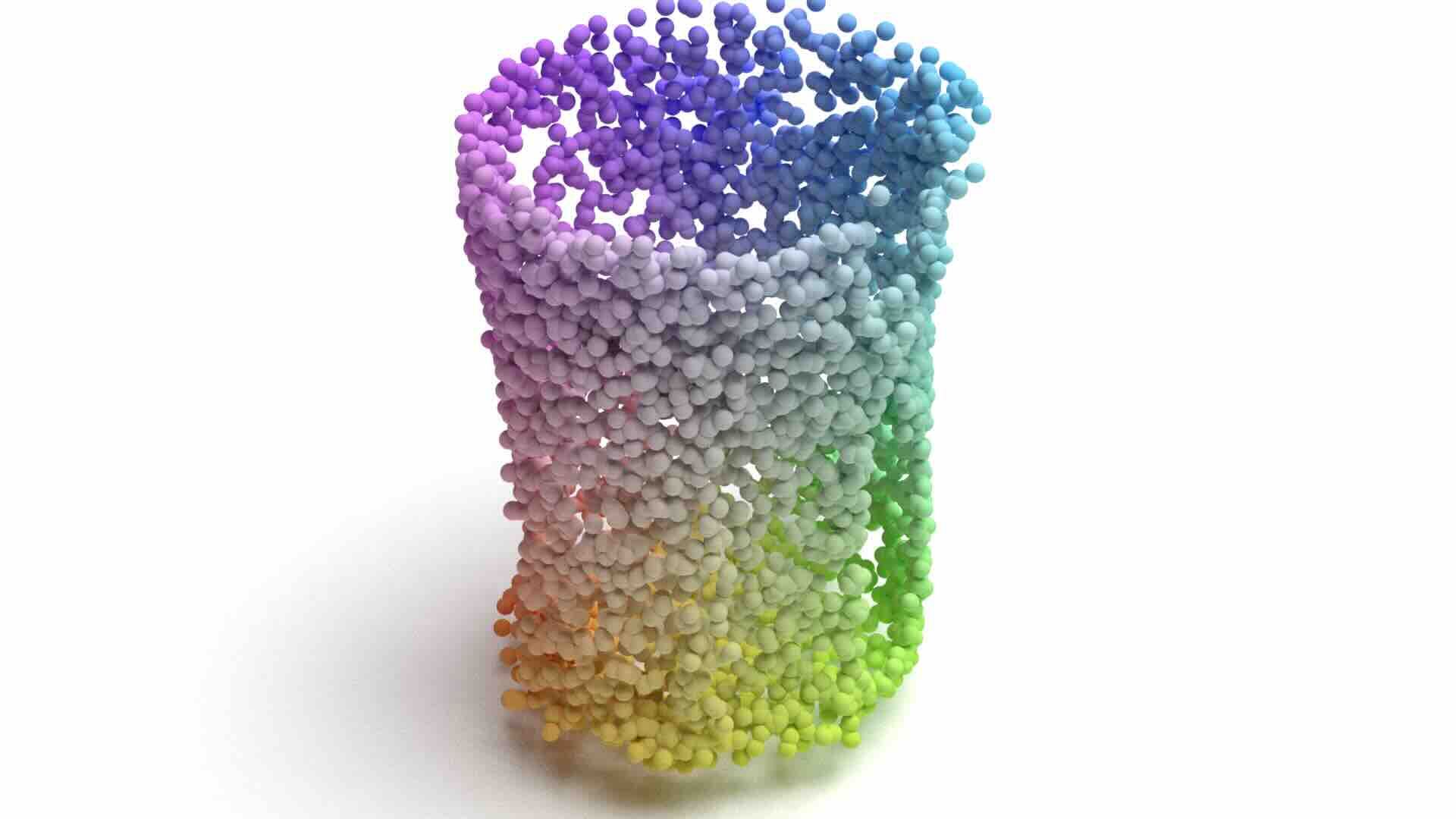} &
        \includegraphics[trim={15cm 0.0cm 15cm 0.0cm},clip,width=0.12\textwidth]{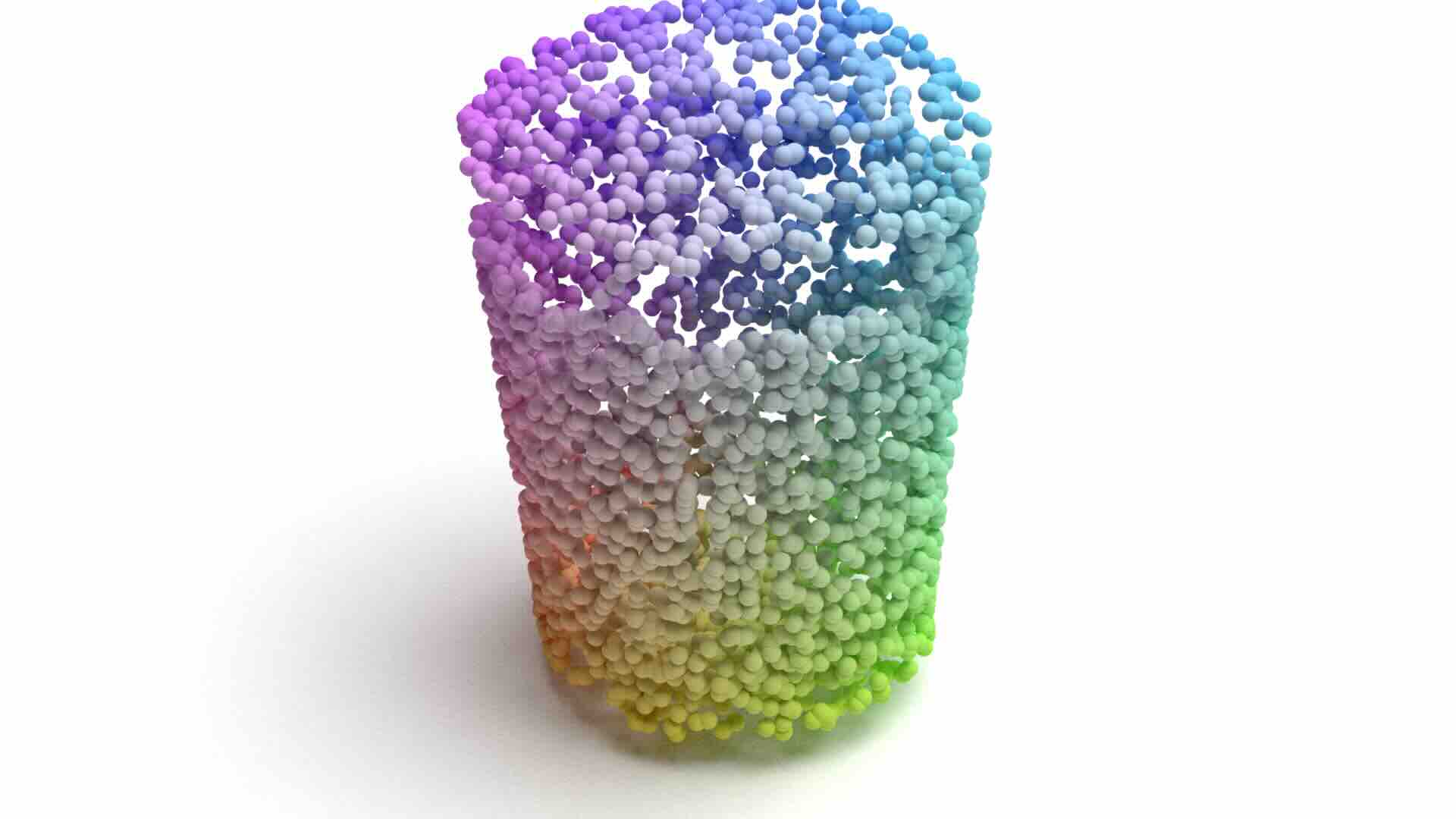} \\

        \includegraphics[trim={15cm 0.0cm 15cm 0.0cm},clip,width=0.12\textwidth]{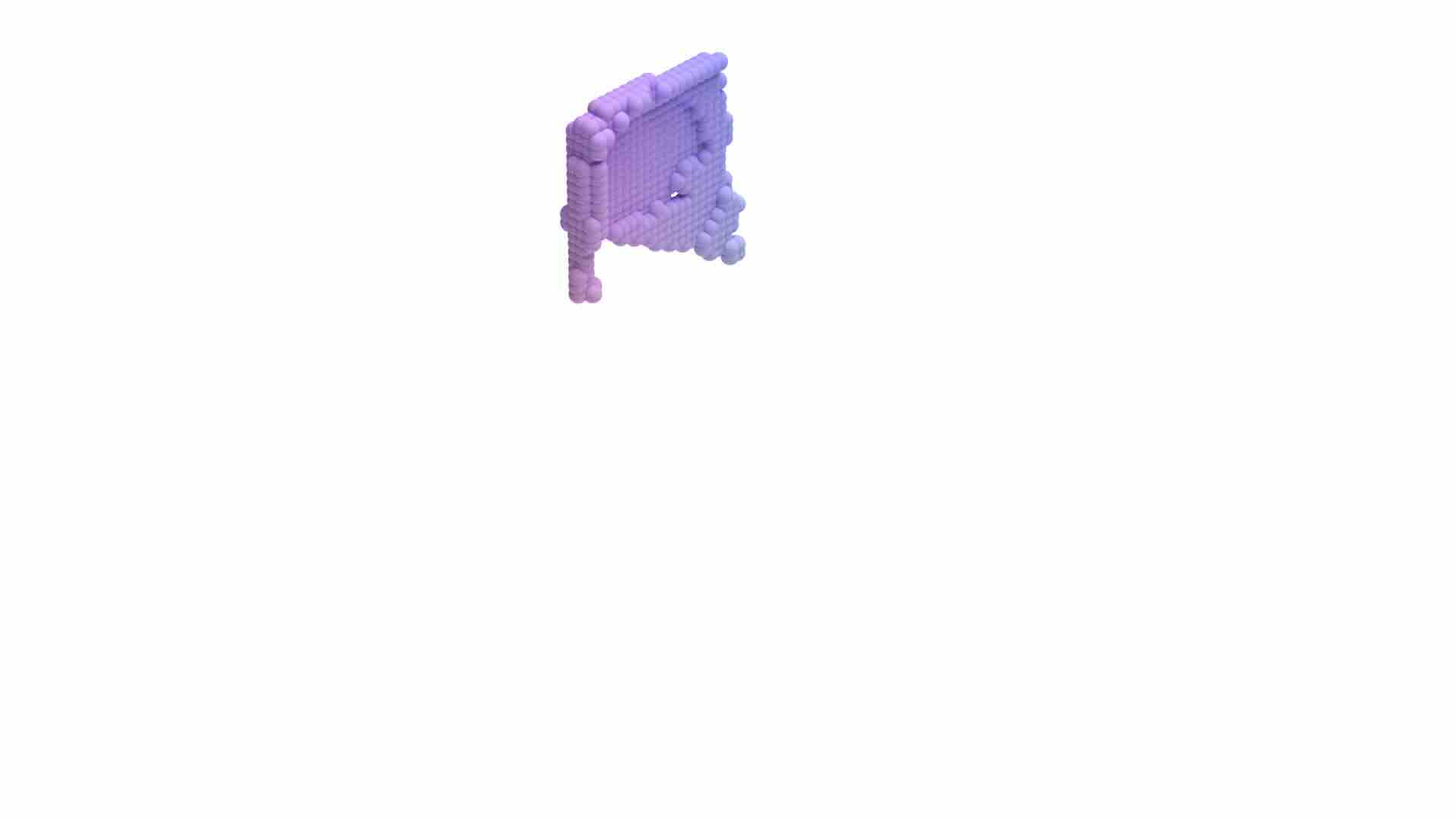} 
        &
        
        \includegraphics[trim={15cm 0.0cm 15cm 0.0cm},clip,width=0.12\textwidth]{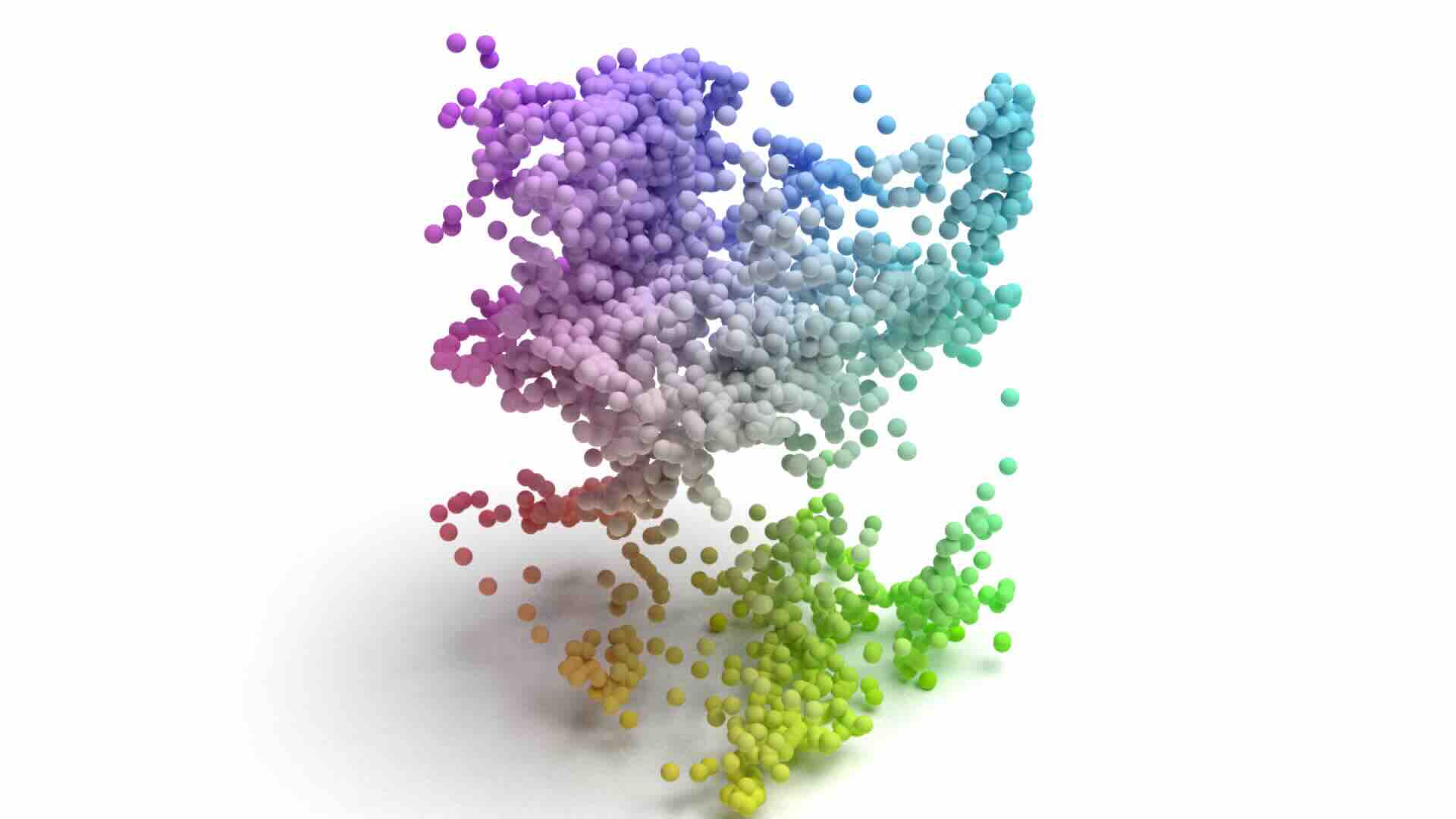} &
        \includegraphics[trim={15cm 0.0cm 15cm 0.0cm},clip, width=0.12\textwidth]{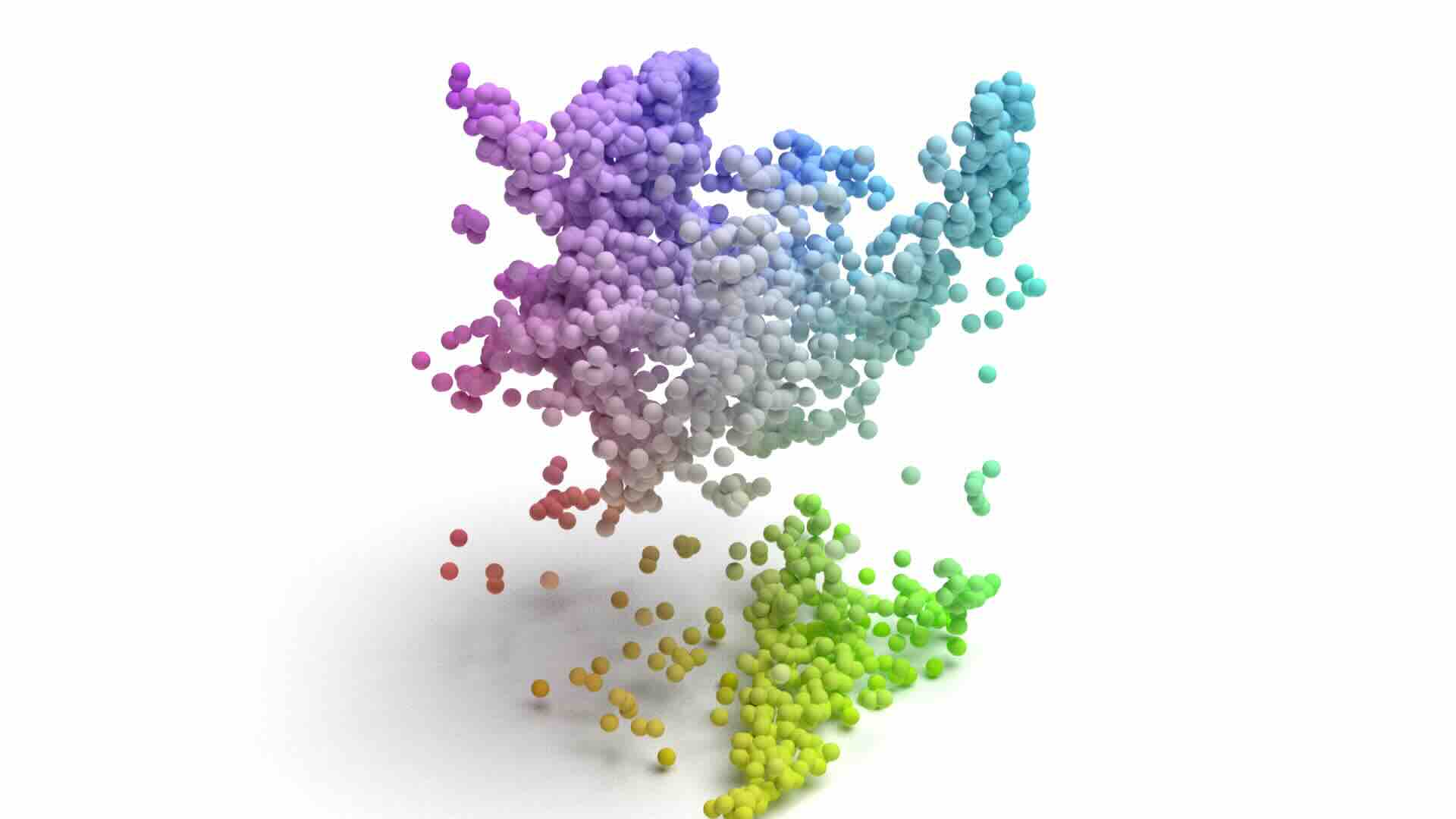} 
        &
        \includegraphics[trim={15cm 0.0cm 15cm 0.0cm},clip, width=0.12\textwidth]{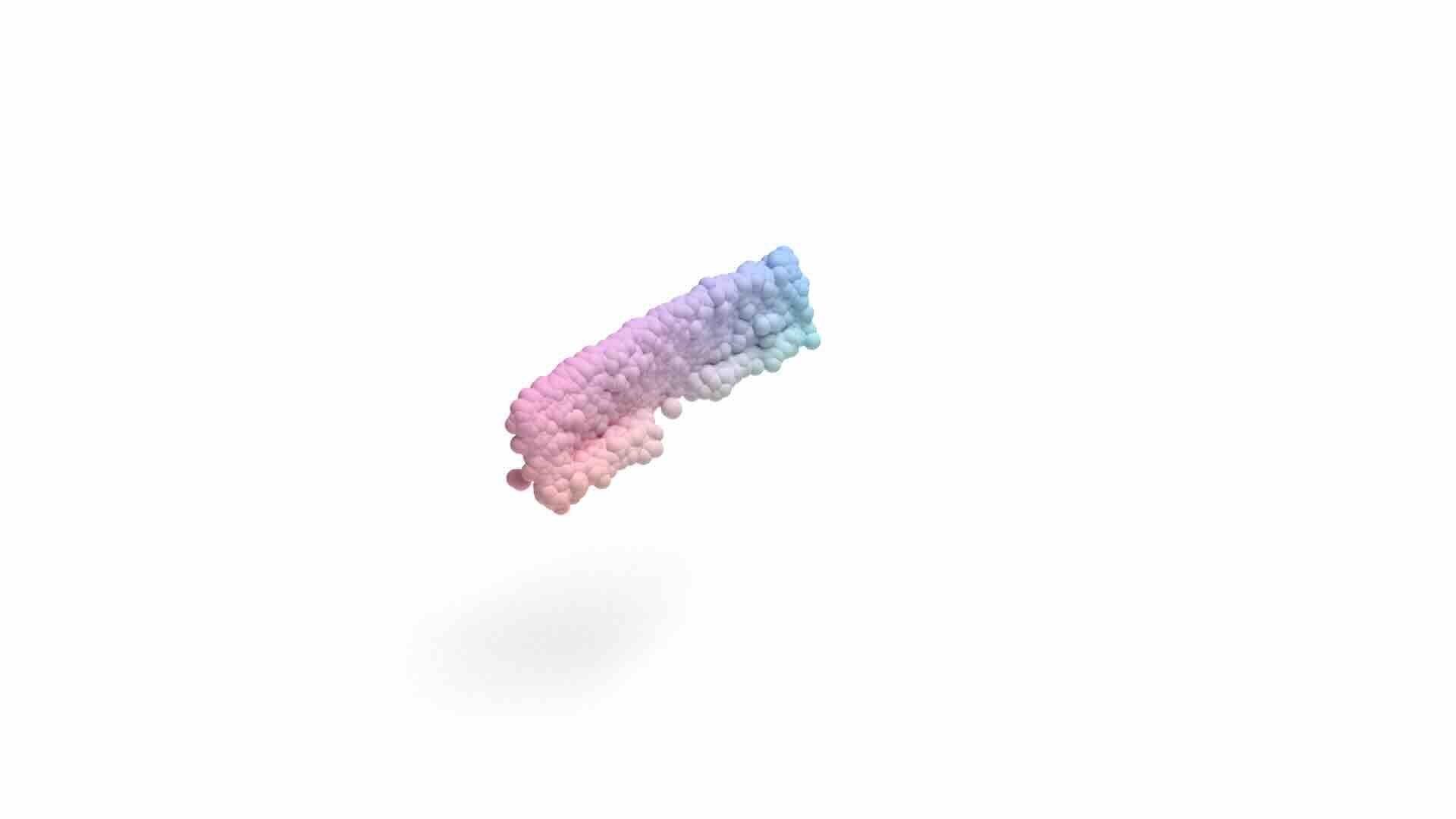} &
        \includegraphics[trim={15cm 0.0cm 15cm 0.0cm},clip, width=0.12\textwidth]{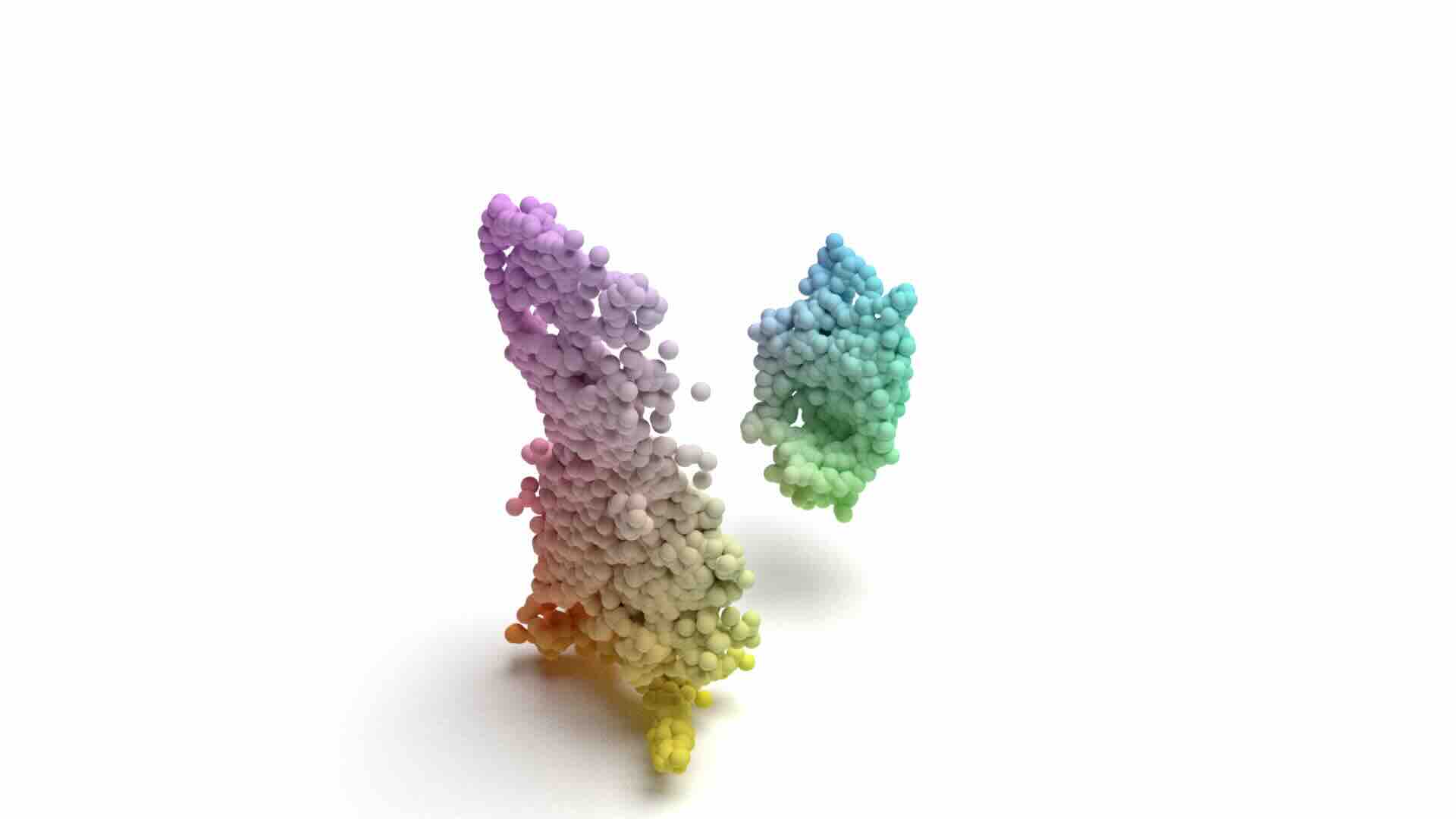} &
        \includegraphics[trim={15cm 0.0cm 15cm 0.0cm},clip, width=0.12\textwidth]{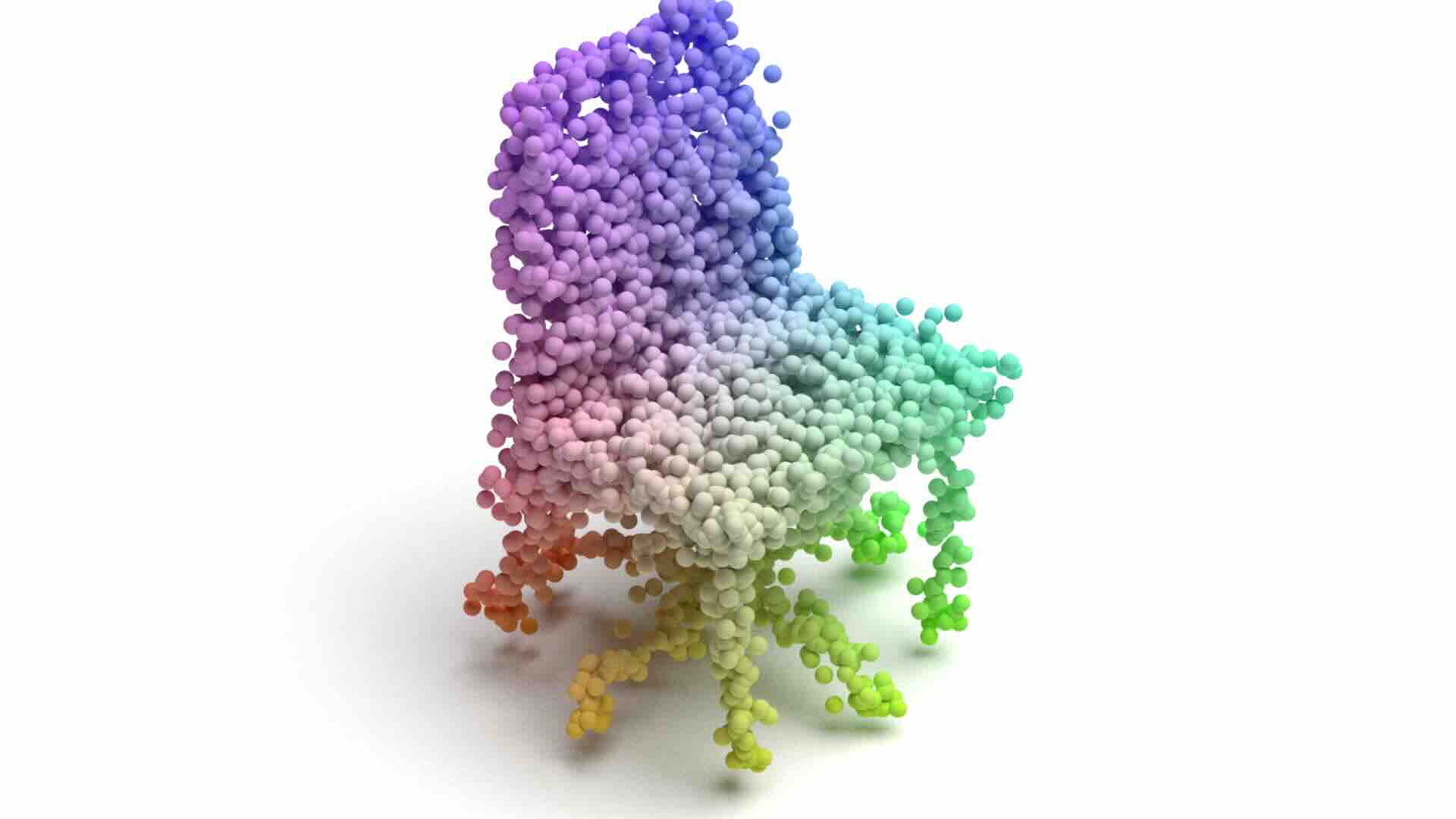} &
        \includegraphics[trim={15cm 0.0cm 15cm 0.0cm},clip, width=0.12\textwidth]{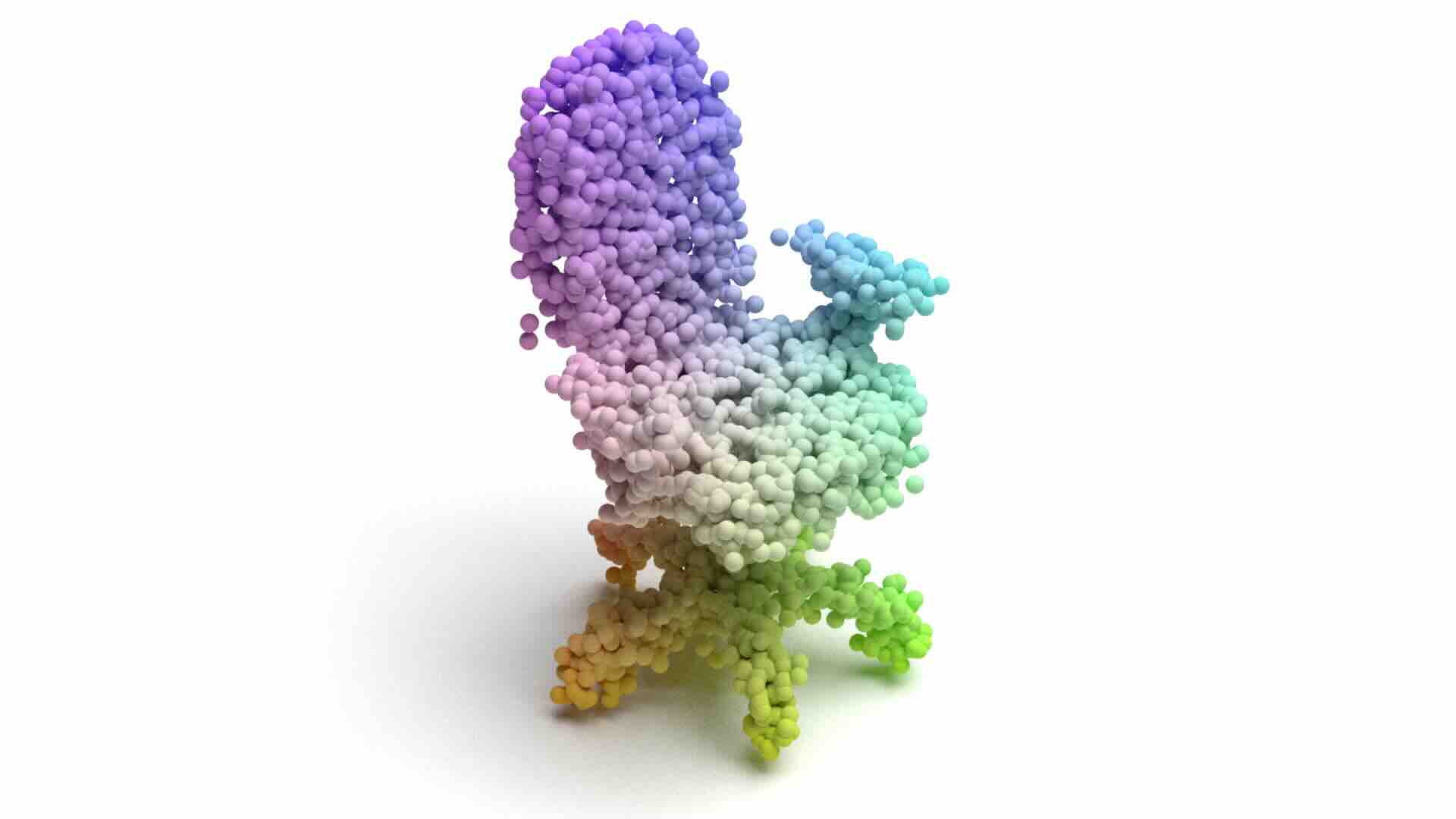} &
        \includegraphics[trim={15cm 0.0cm 15cm 0.0cm},clip,width=0.12\textwidth]{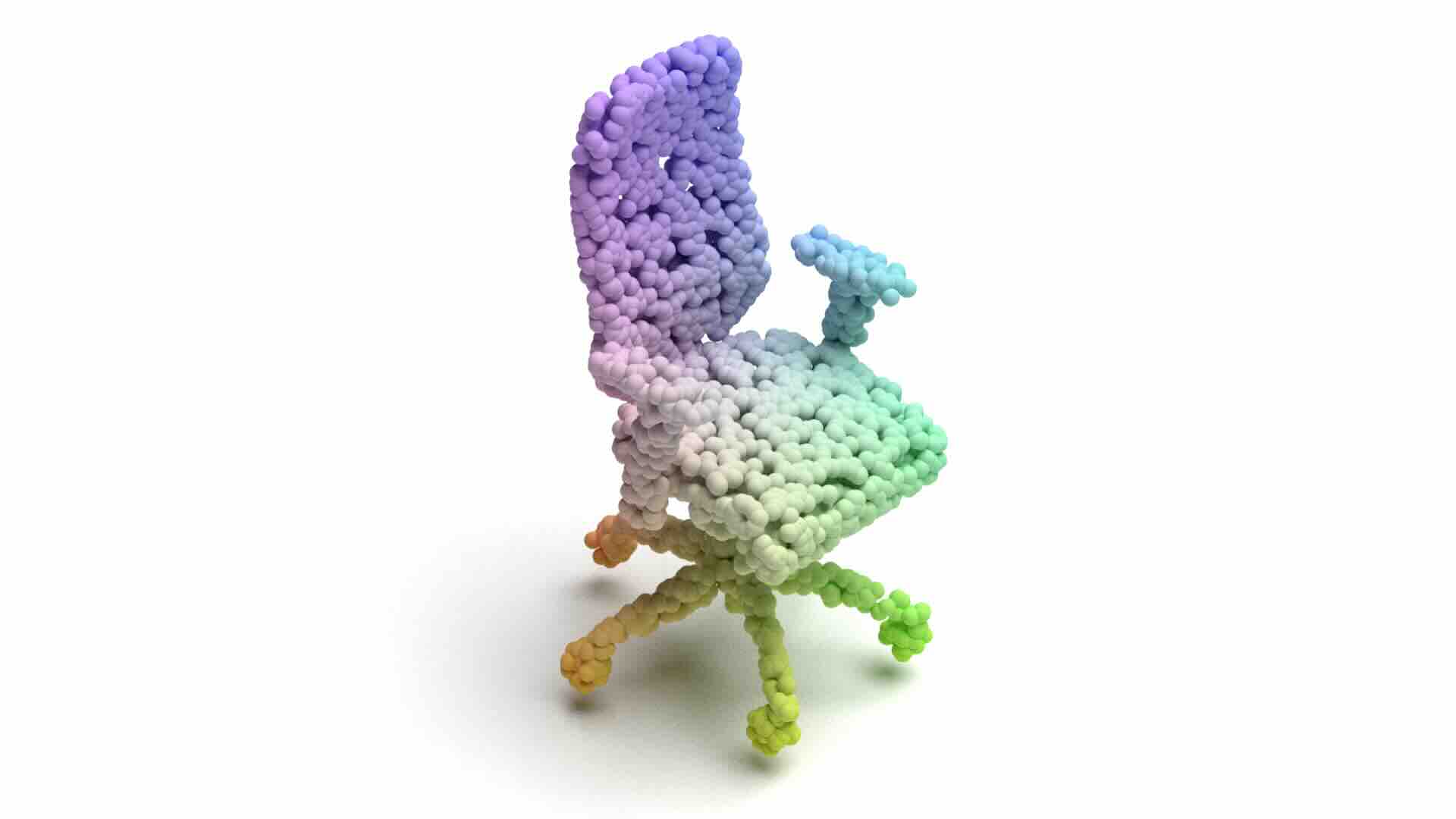} \\
        \includegraphics[trim={15cm 0.0cm 15cm 0.0cm},clip,width=0.12\textwidth]{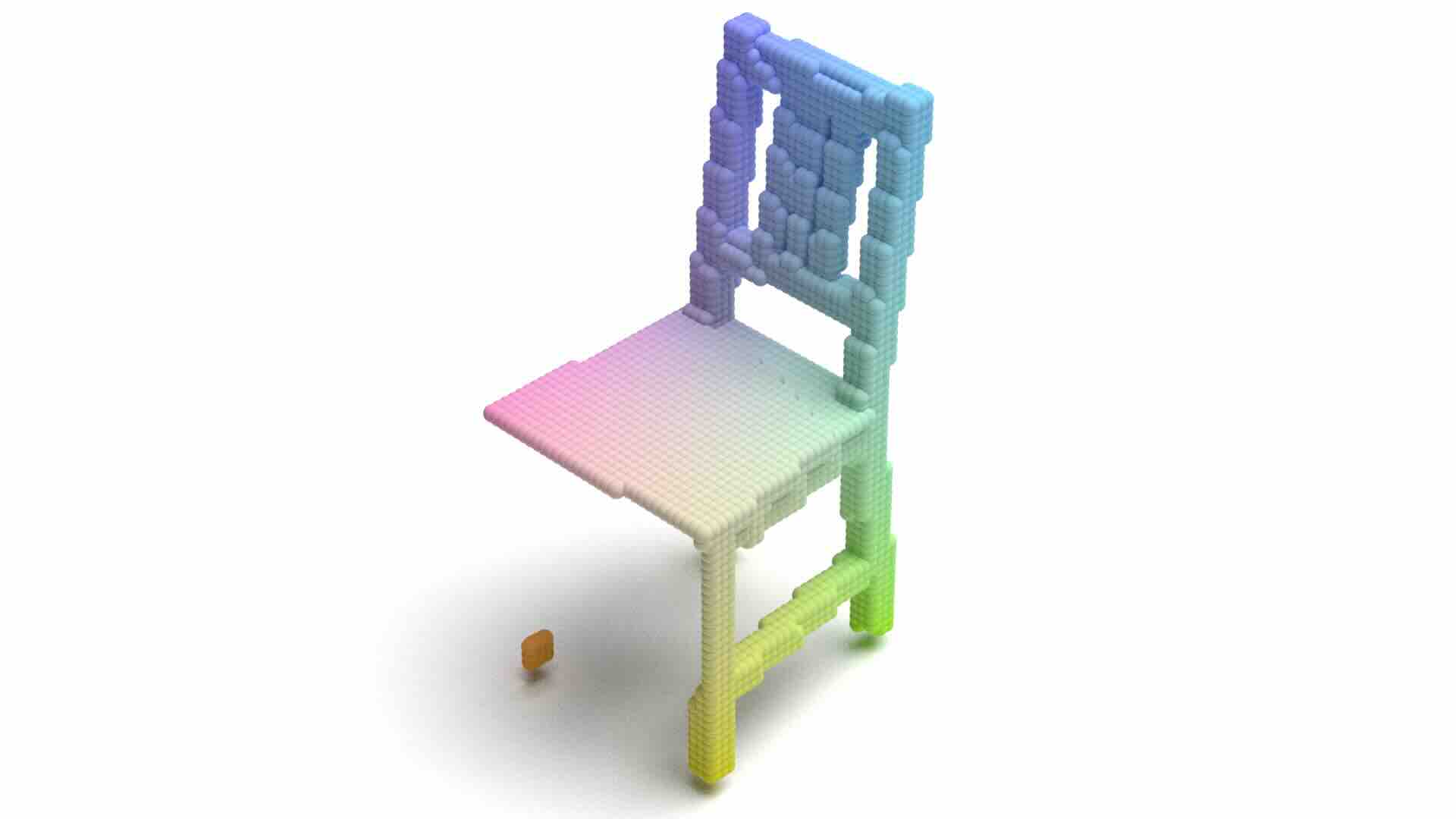} 
        &
        \includegraphics[trim={15cm 0.0cm 15cm 0.0cm},clip, width=0.12\textwidth]{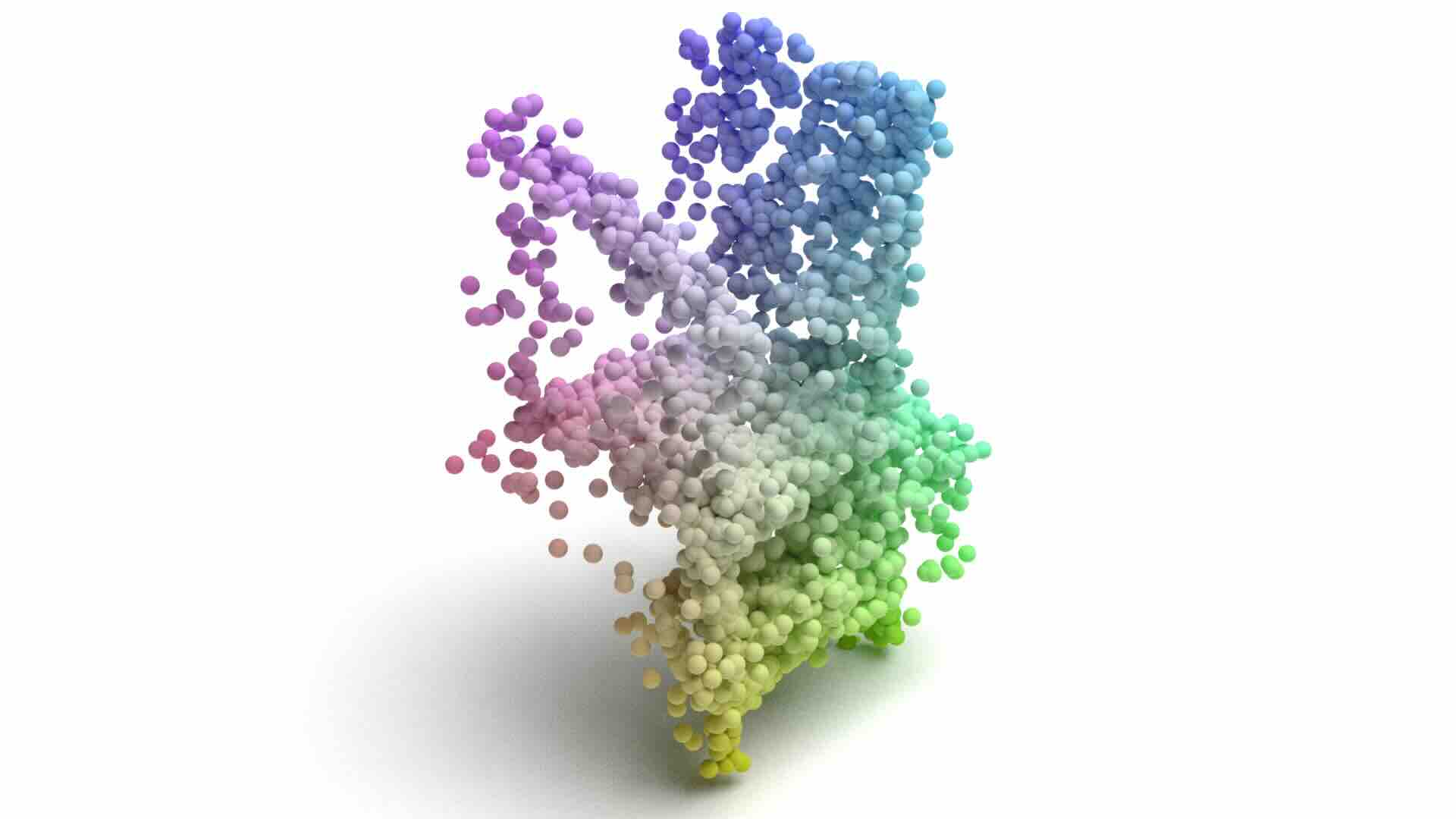} &
        \includegraphics[trim={15cm 0.0cm 15cm 0.0cm},clip, width=0.12\textwidth]{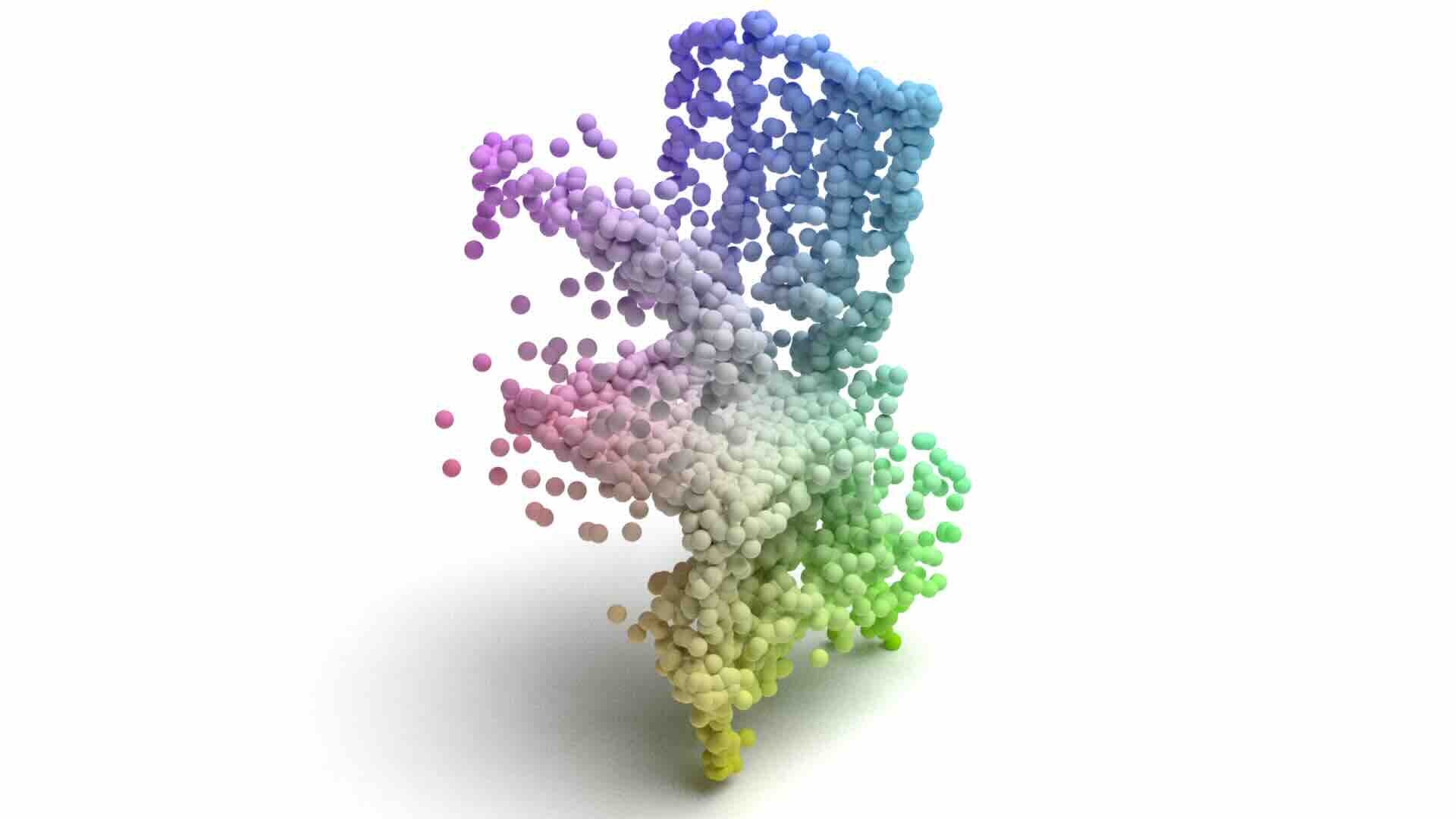} &
        \includegraphics[trim={15cm 0.0cm 15cm 0.0cm},clip, width=0.12\textwidth]{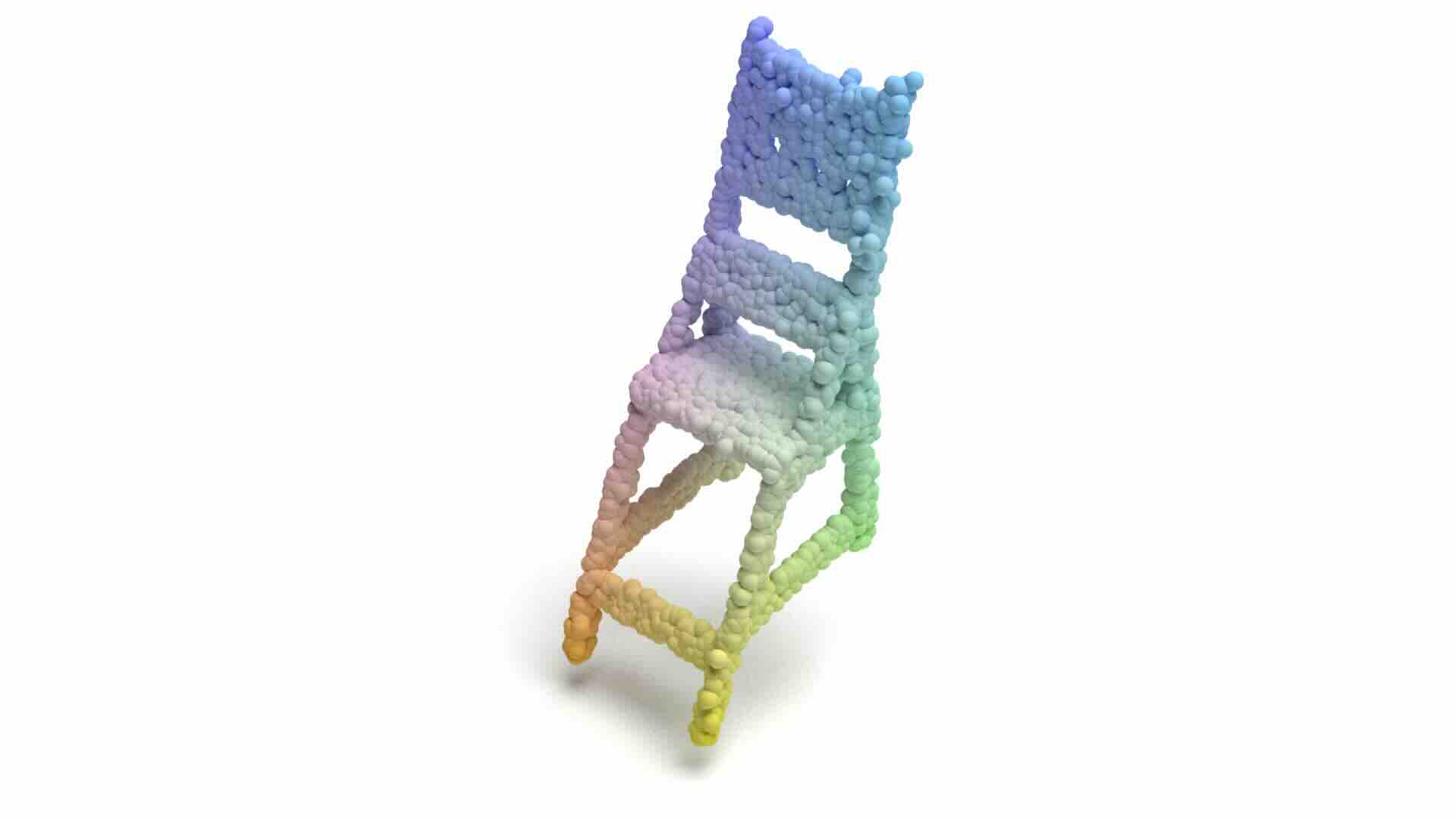} &
        \includegraphics[trim={15cm 0.0cm 15cm 0.0cm},clip,width=0.12\textwidth]{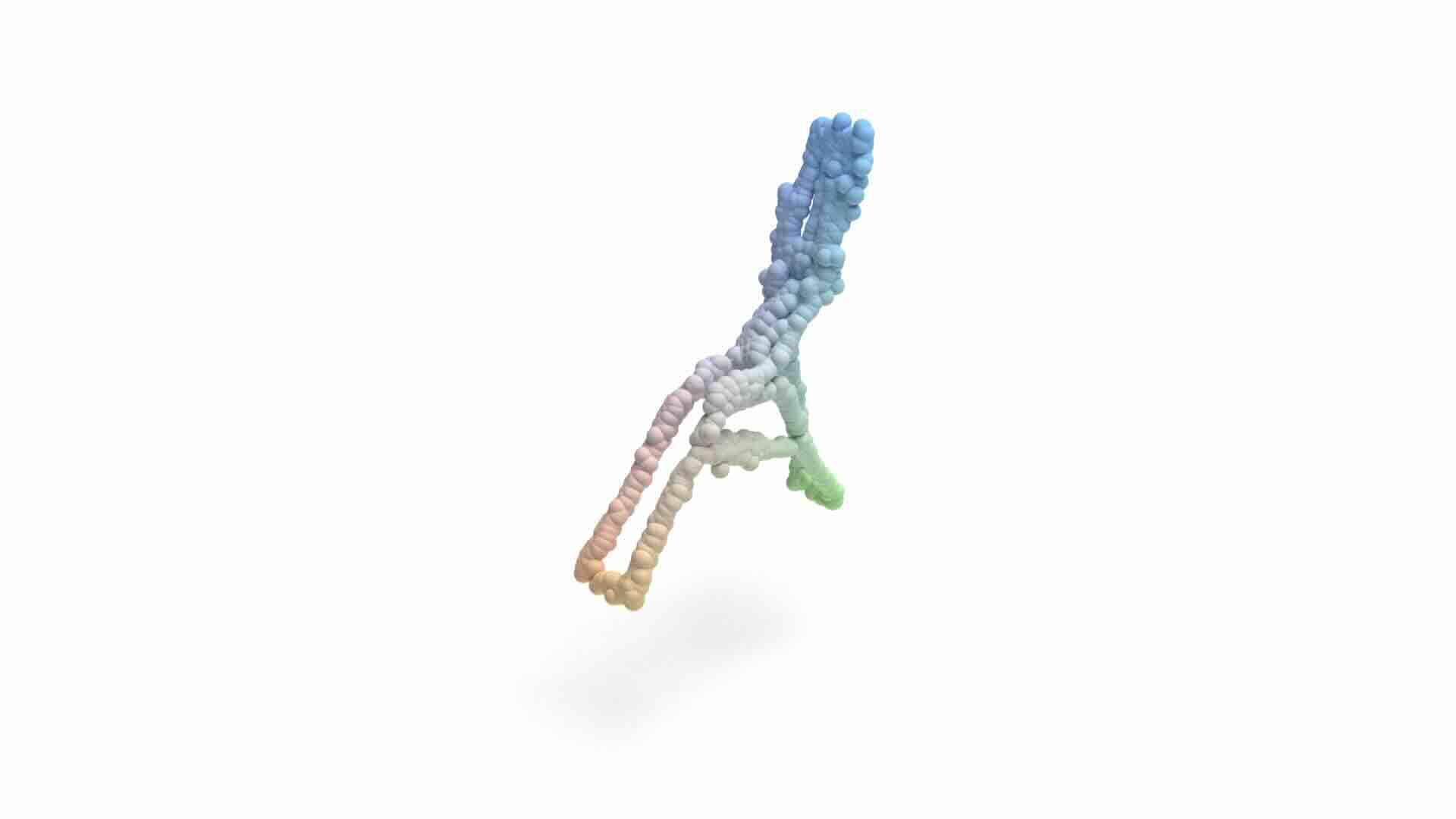} &
        \includegraphics[trim={15cm 0.0cm 15cm 0.0cm},clip,width=0.12\textwidth]{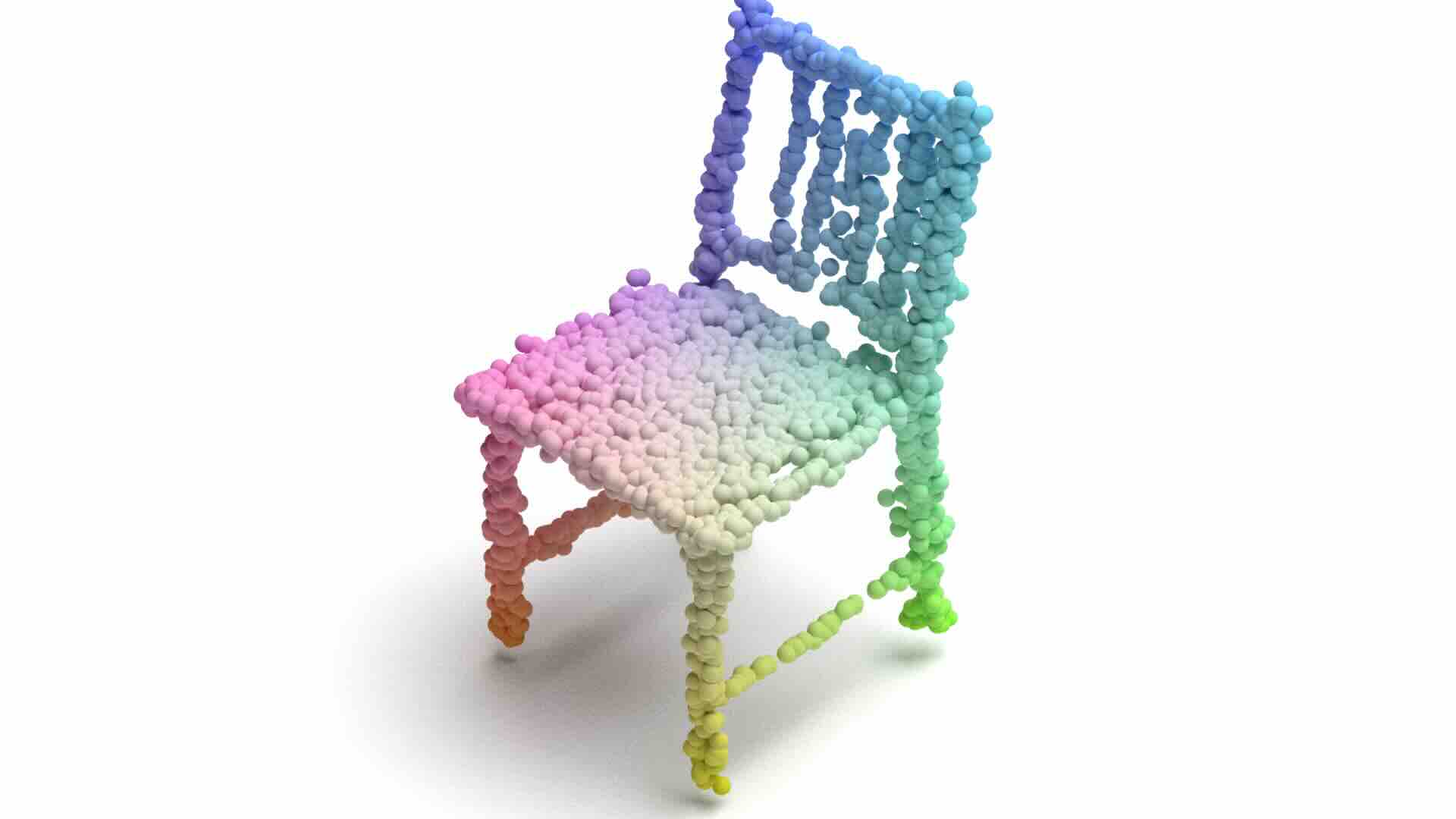} &
        \includegraphics[trim={15cm 0.0cm 15cm 0.0cm},clip,width=0.12\textwidth]{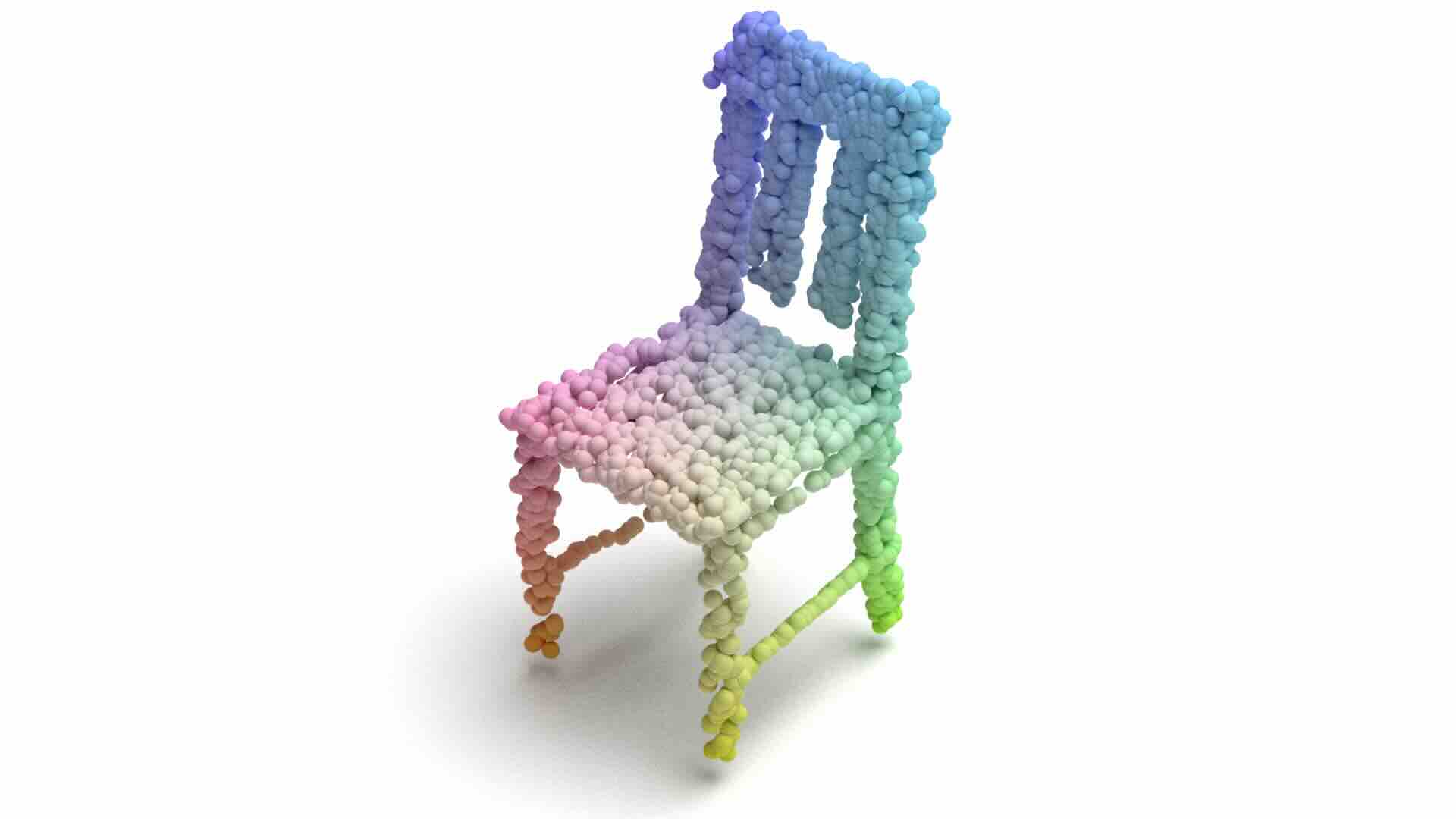} &
        \includegraphics[trim={15cm 0.0cm 15cm 0.0cm},clip,width=0.12\textwidth]{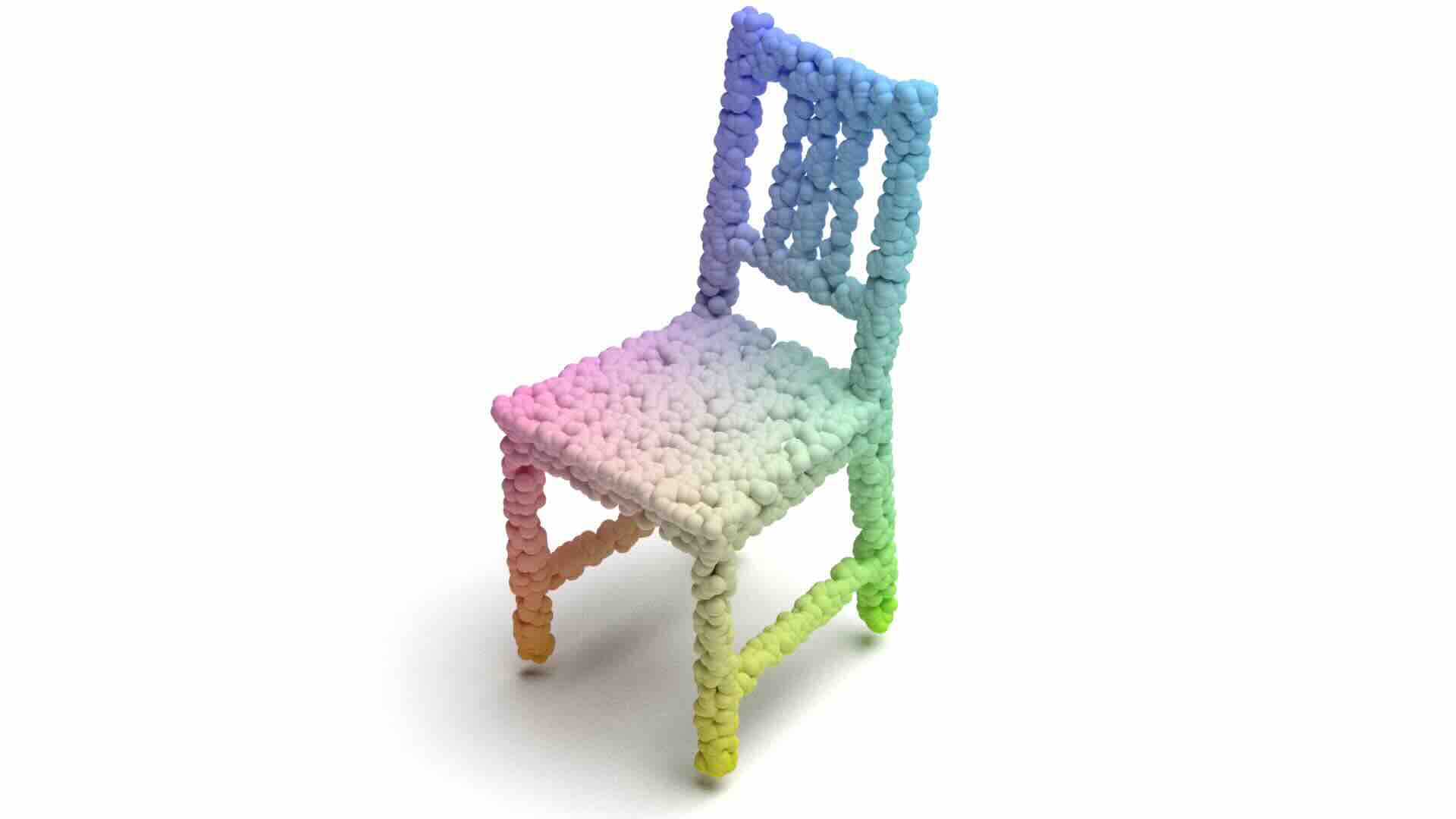}
        \\
        \includegraphics[trim={15cm 0.0cm 15cm 10.0cm},clip,width=0.12\textwidth]{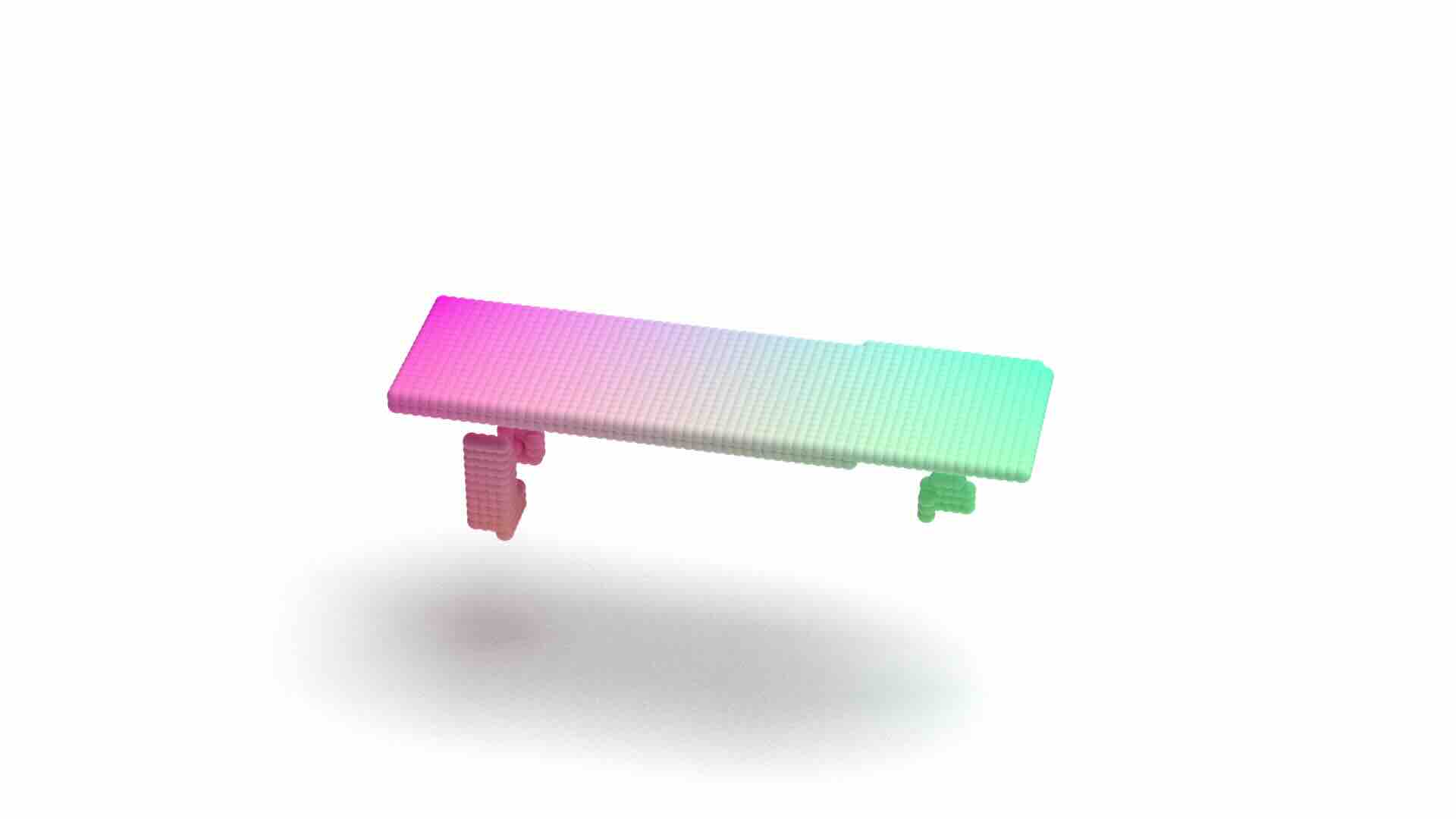} 
        &
        \includegraphics[trim={15cm 0.0cm 15cm 10.0cm},clip, width=0.12\textwidth]{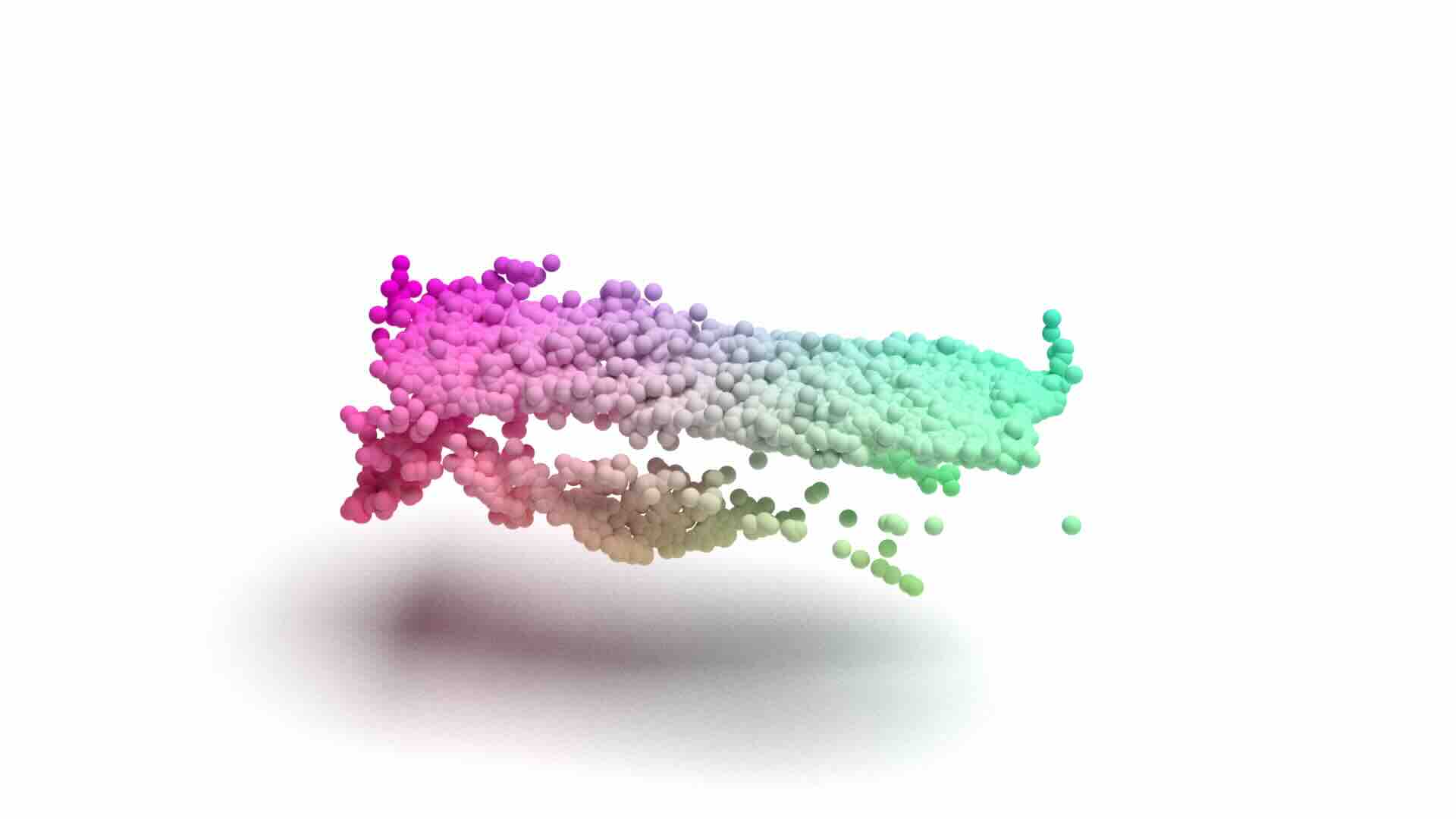} &
        \includegraphics[trim={15cm 0.0cm 15cm 5.0cm},clip, width=0.12\textwidth]{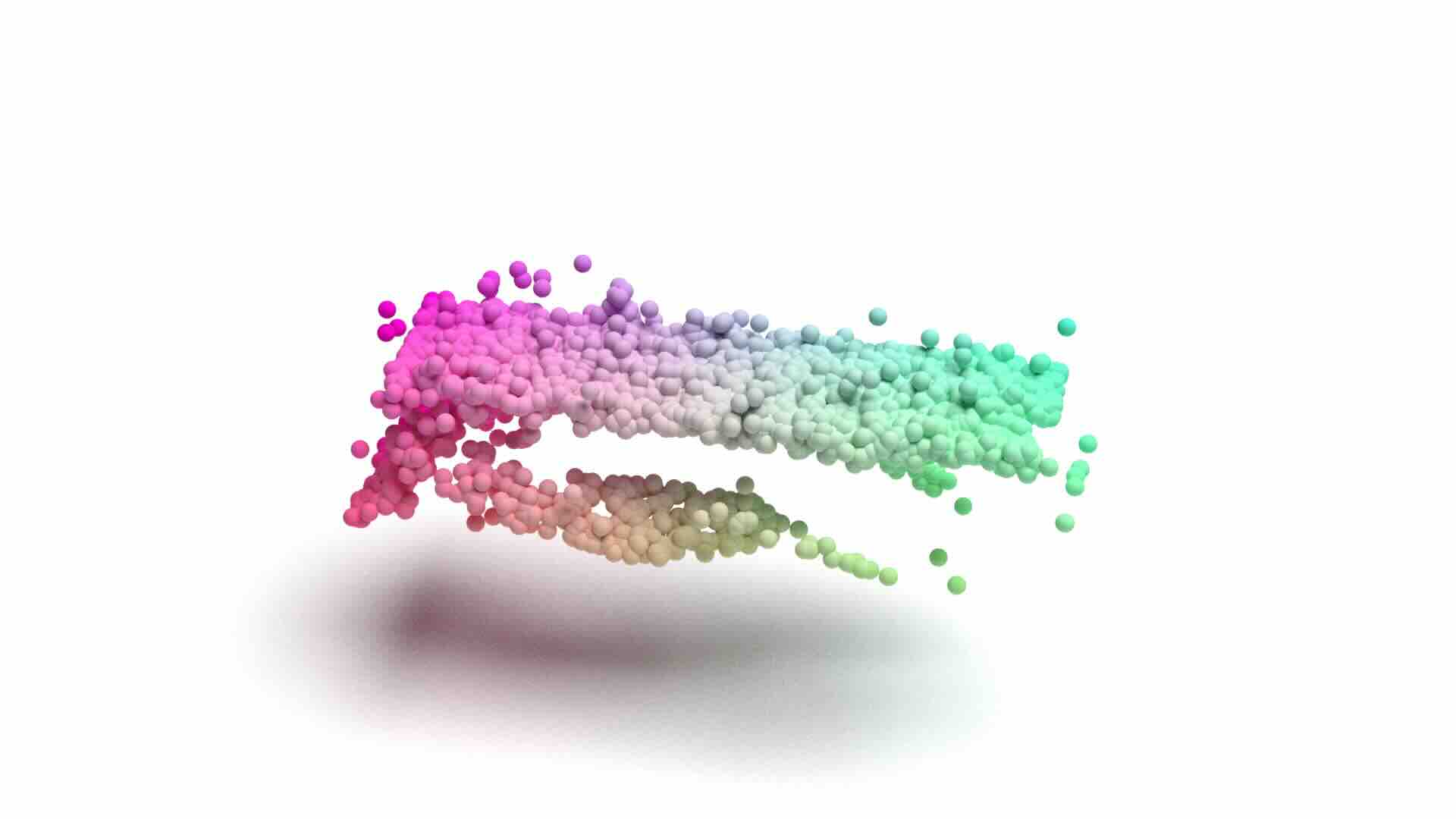} &
        \includegraphics[trim={15cm 0.0cm 15cm 10.0cm},clip, width=0.12\textwidth]{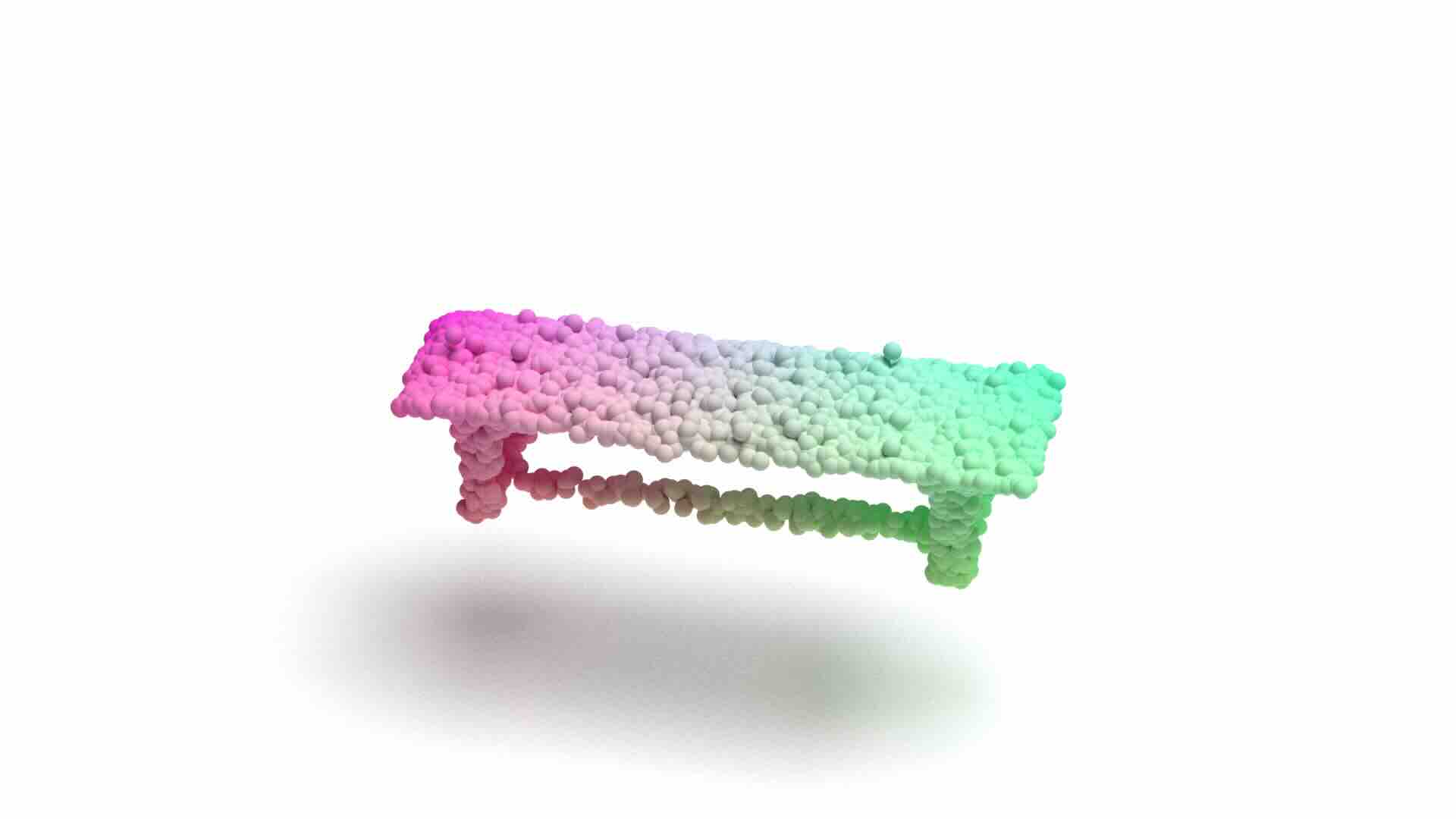} &
        \includegraphics[trim={15cm 0.0cm 15cm 10.0cm},clip,width=0.12\textwidth]{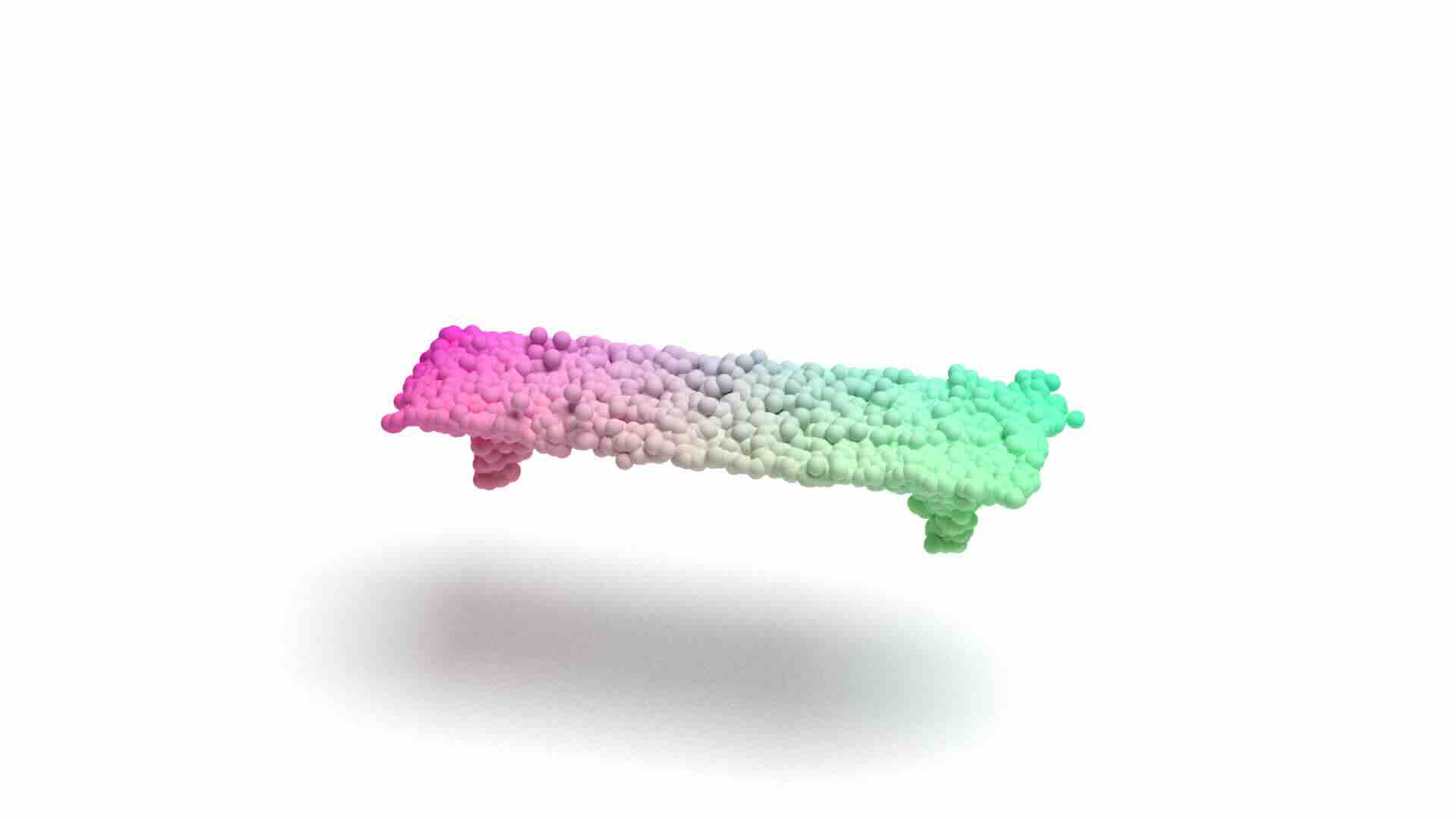} &
        \includegraphics[trim={15cm 0.0cm 15cm 10.0cm},clip,width=0.12\textwidth]{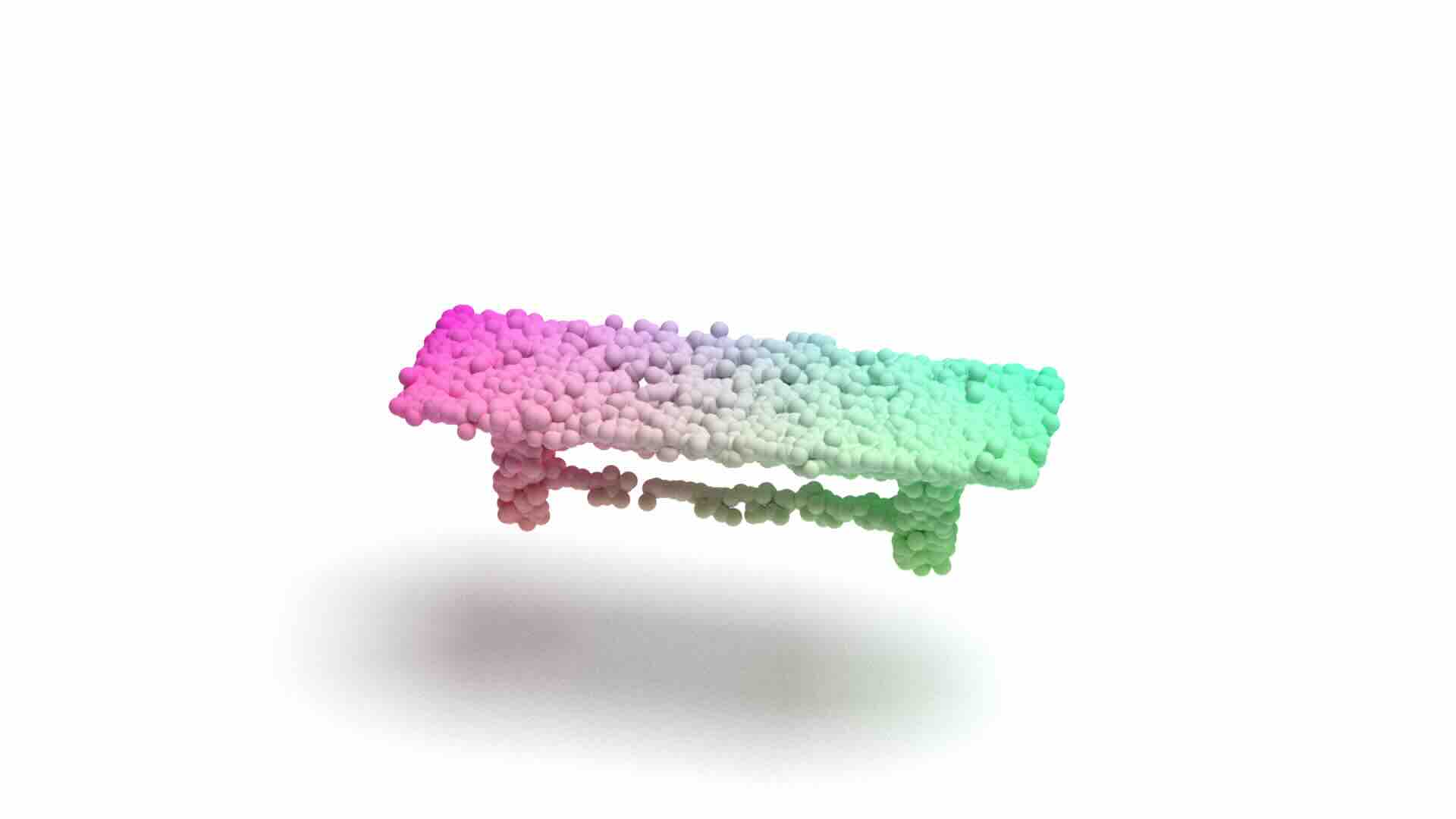} &
        \includegraphics[trim={15cm 0.0cm 15cm 10.0cm},clip,width=0.12\textwidth]{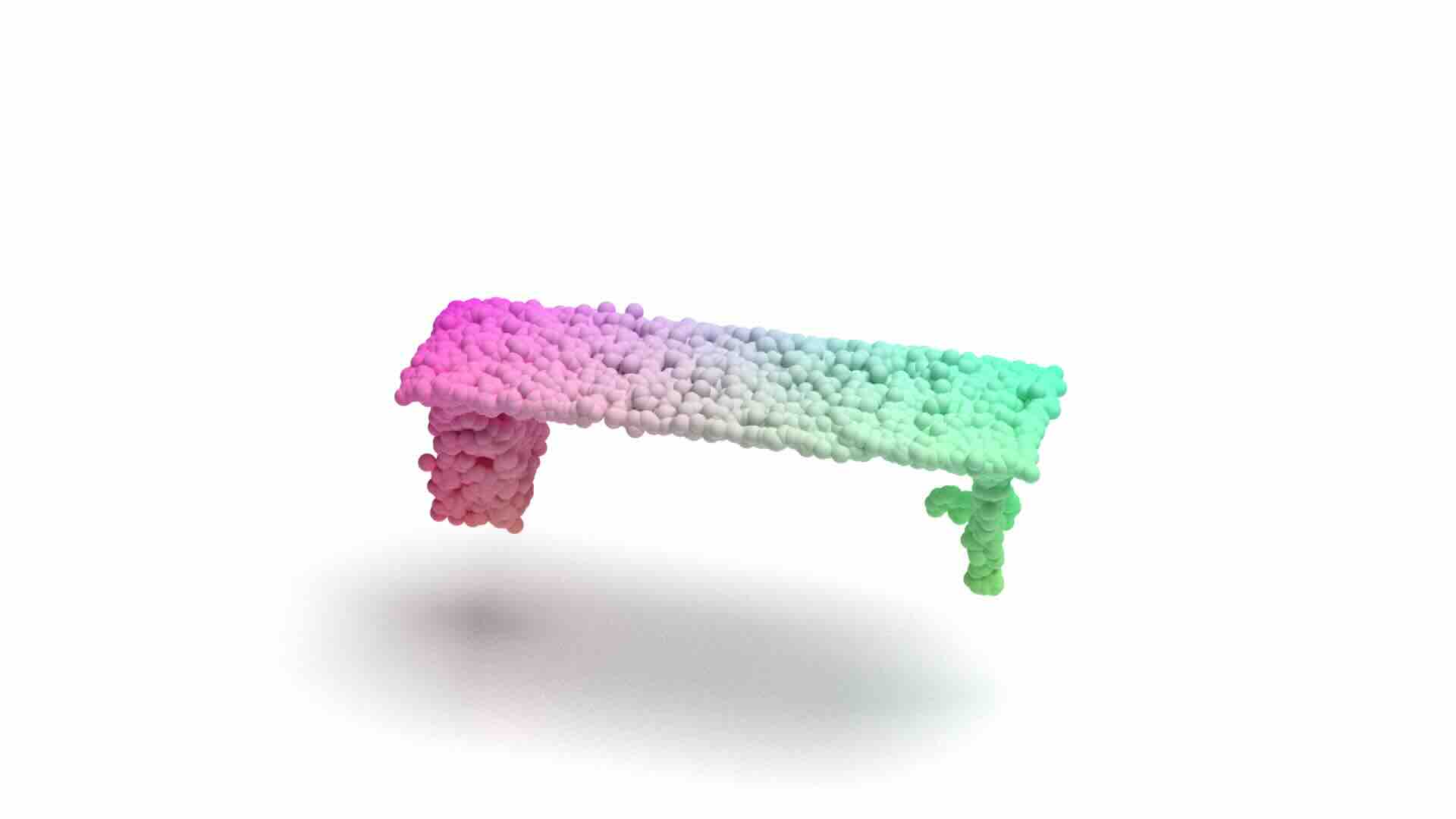} &
        \includegraphics[trim={15cm 0.0cm 15cm 10.0cm},clip,width=0.12\textwidth]{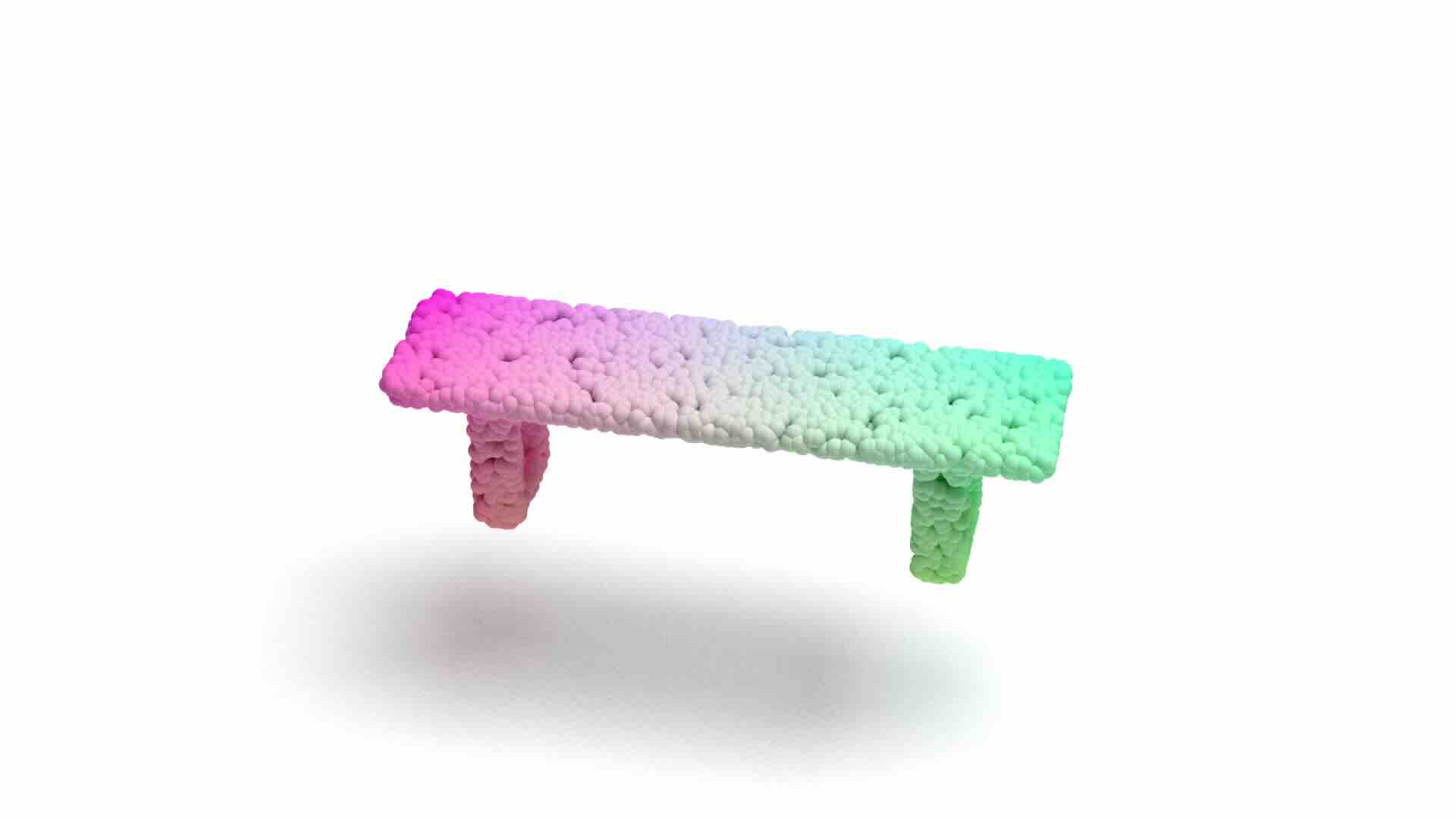}
        \\
        \includegraphics[trim={15cm 0.0cm 15cm 0.0cm},clip,width=0.12\textwidth]{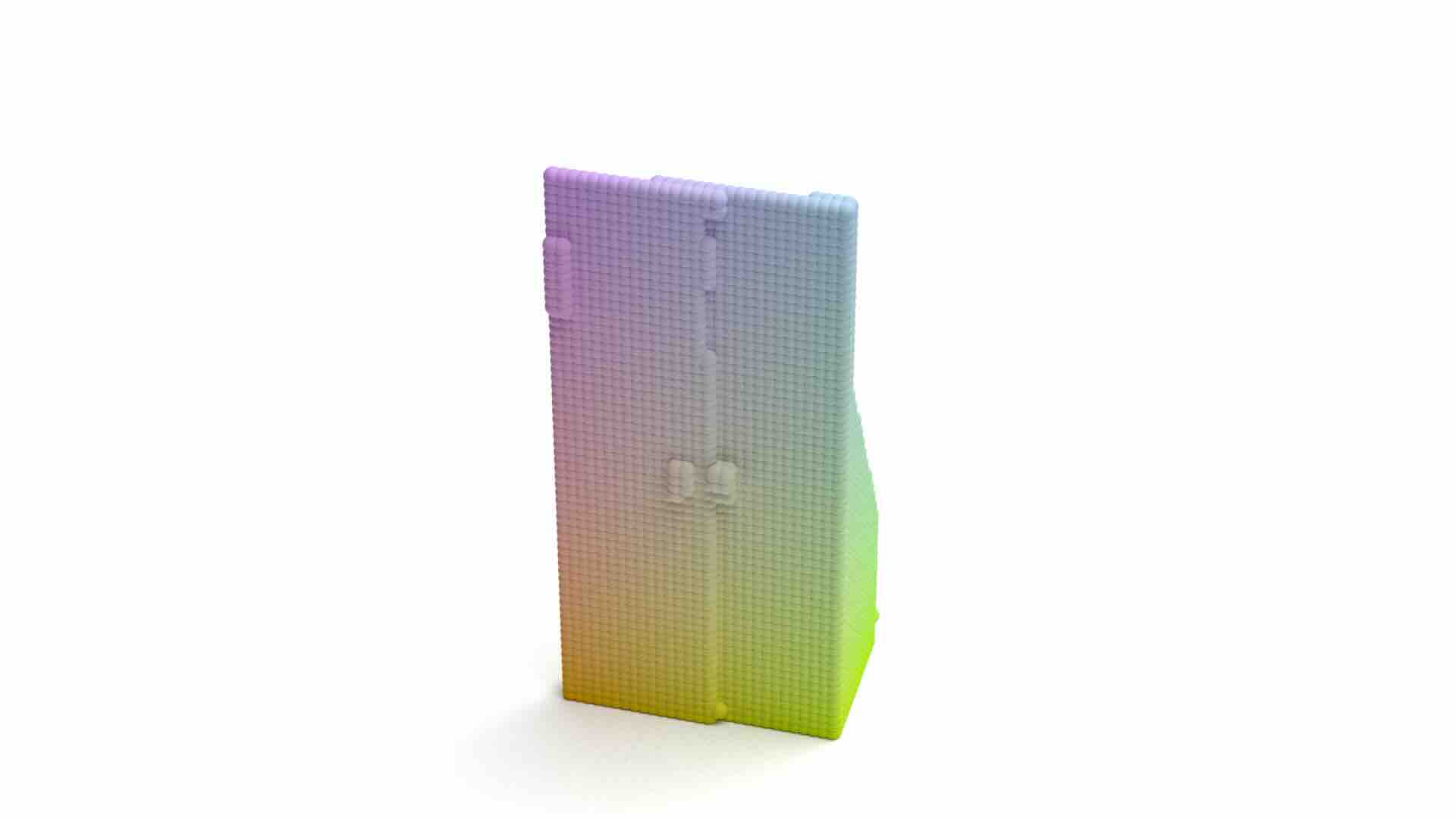} 
        &
        \includegraphics[trim={15cm 0.0cm 15cm 0.0cm},clip, width=0.12\textwidth]{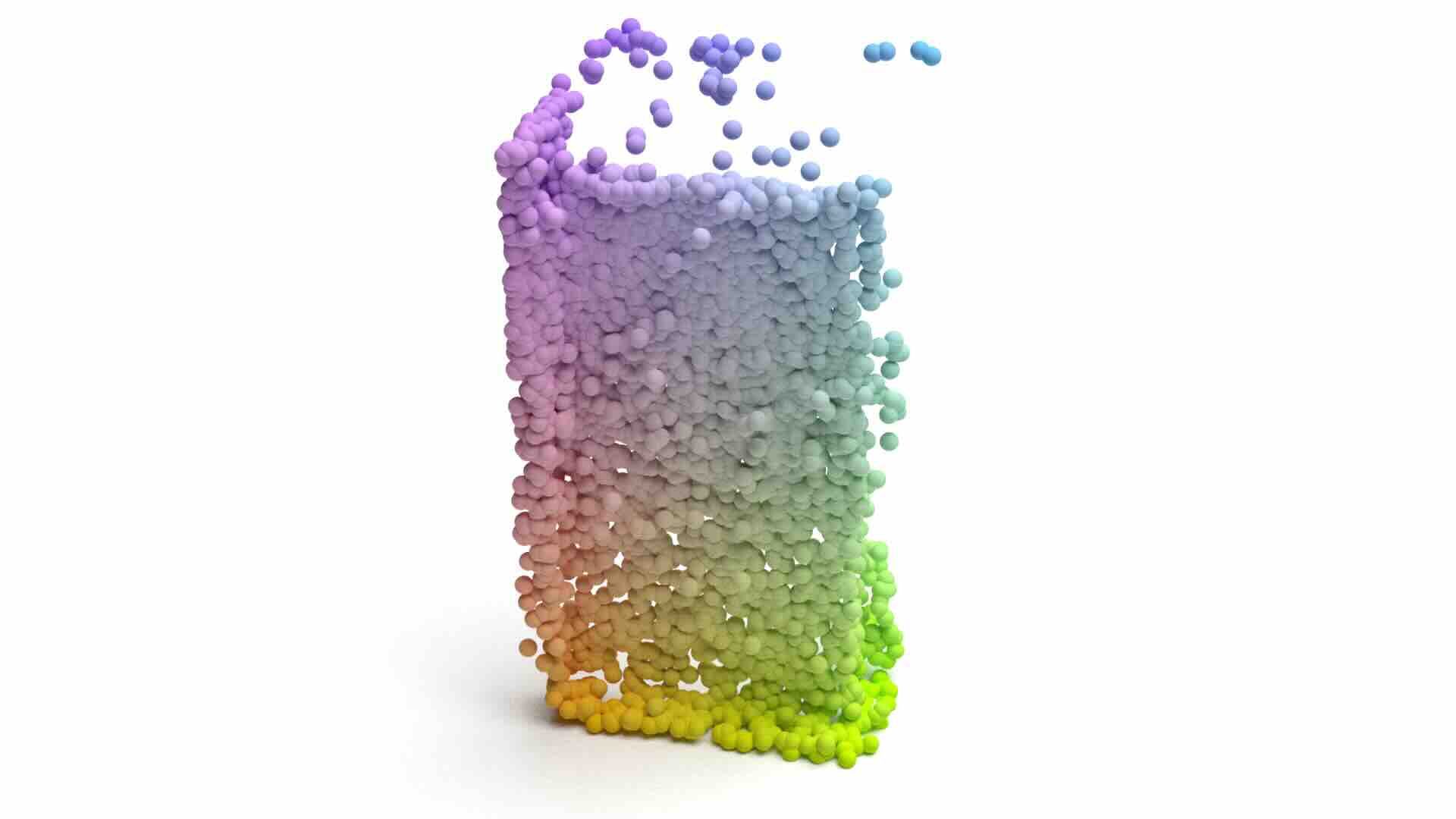} &
        \includegraphics[trim={15cm 0.0cm 15cm 0.0cm},clip, width=0.12\textwidth]{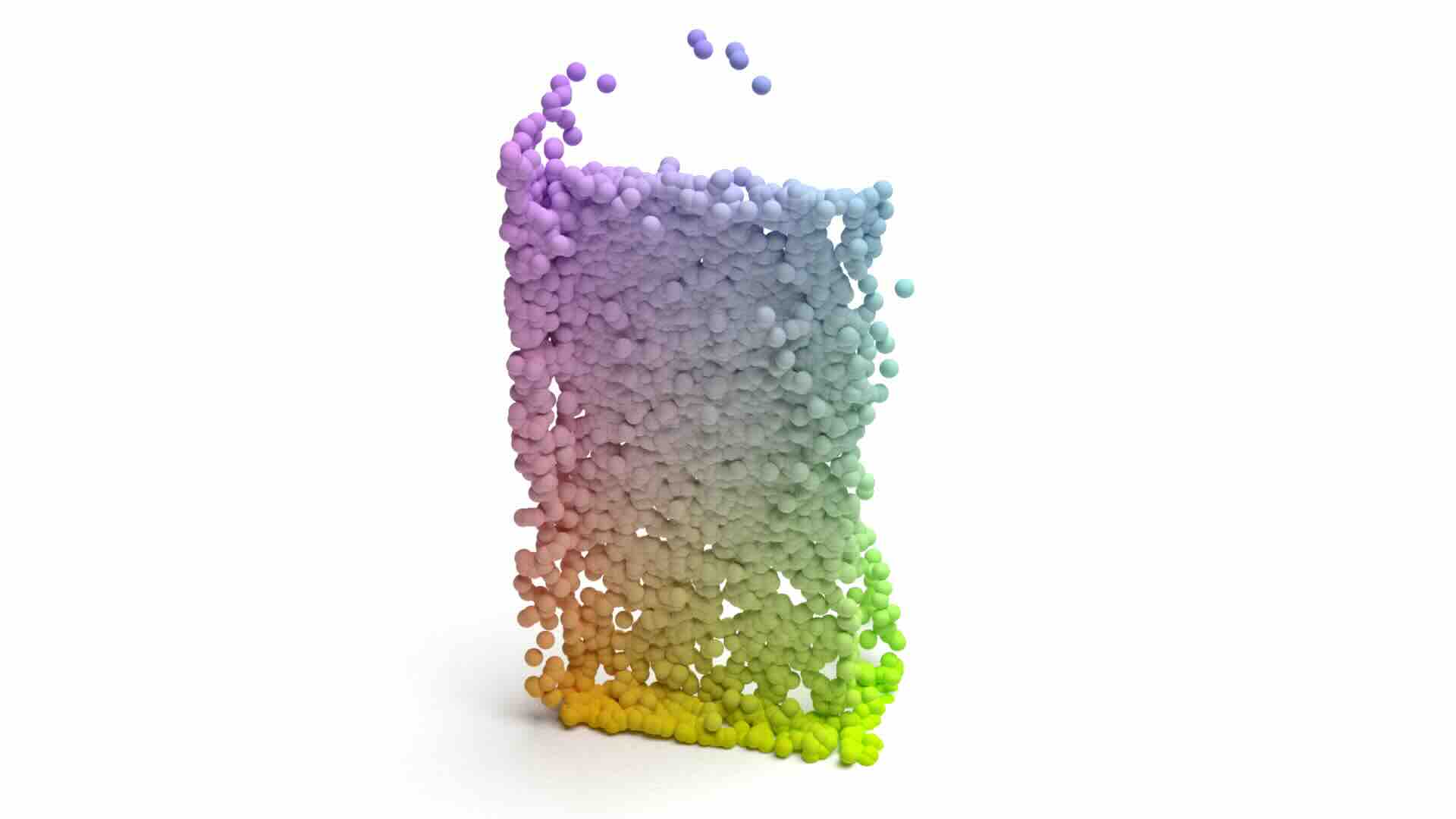} &
        \includegraphics[trim={15cm 0.0cm 15cm 0.0cm},clip, width=0.12\textwidth]{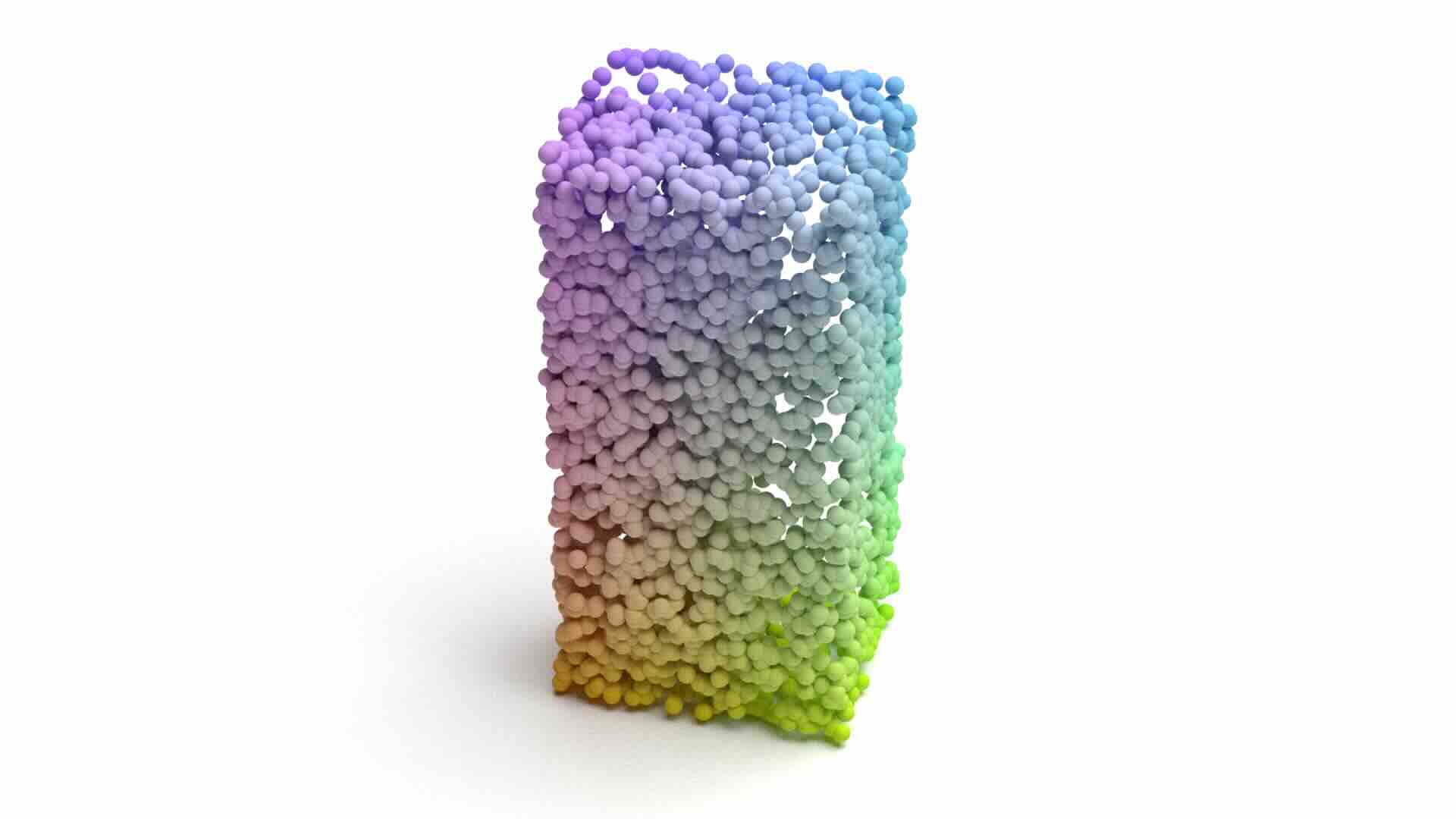} &
        \includegraphics[trim={15cm 0.0cm 15cm 0.0cm},clip,width=0.12\textwidth]{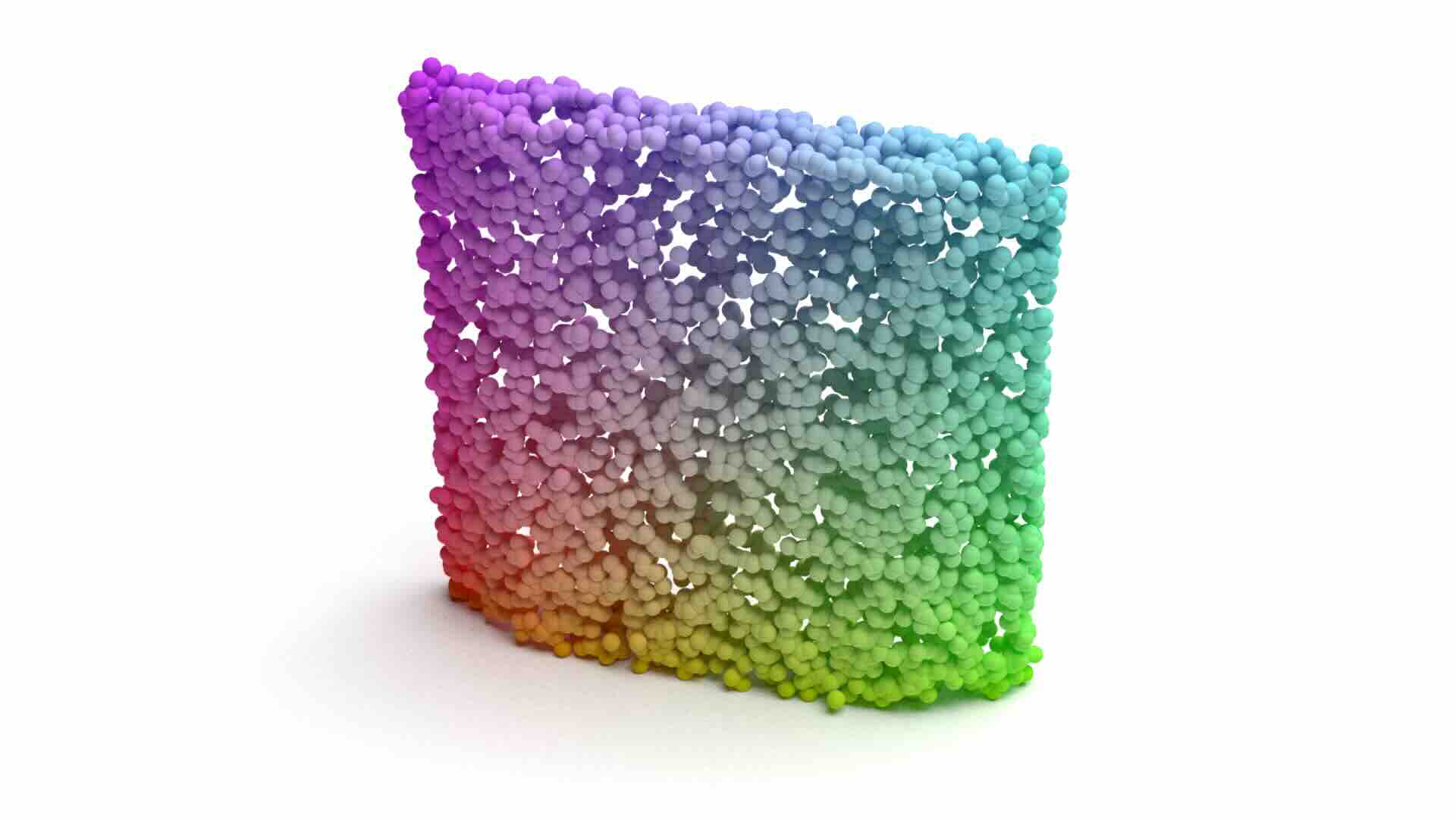} &
        \includegraphics[trim={15cm 0.0cm 15cm 0.0cm},clip,width=0.12\textwidth]{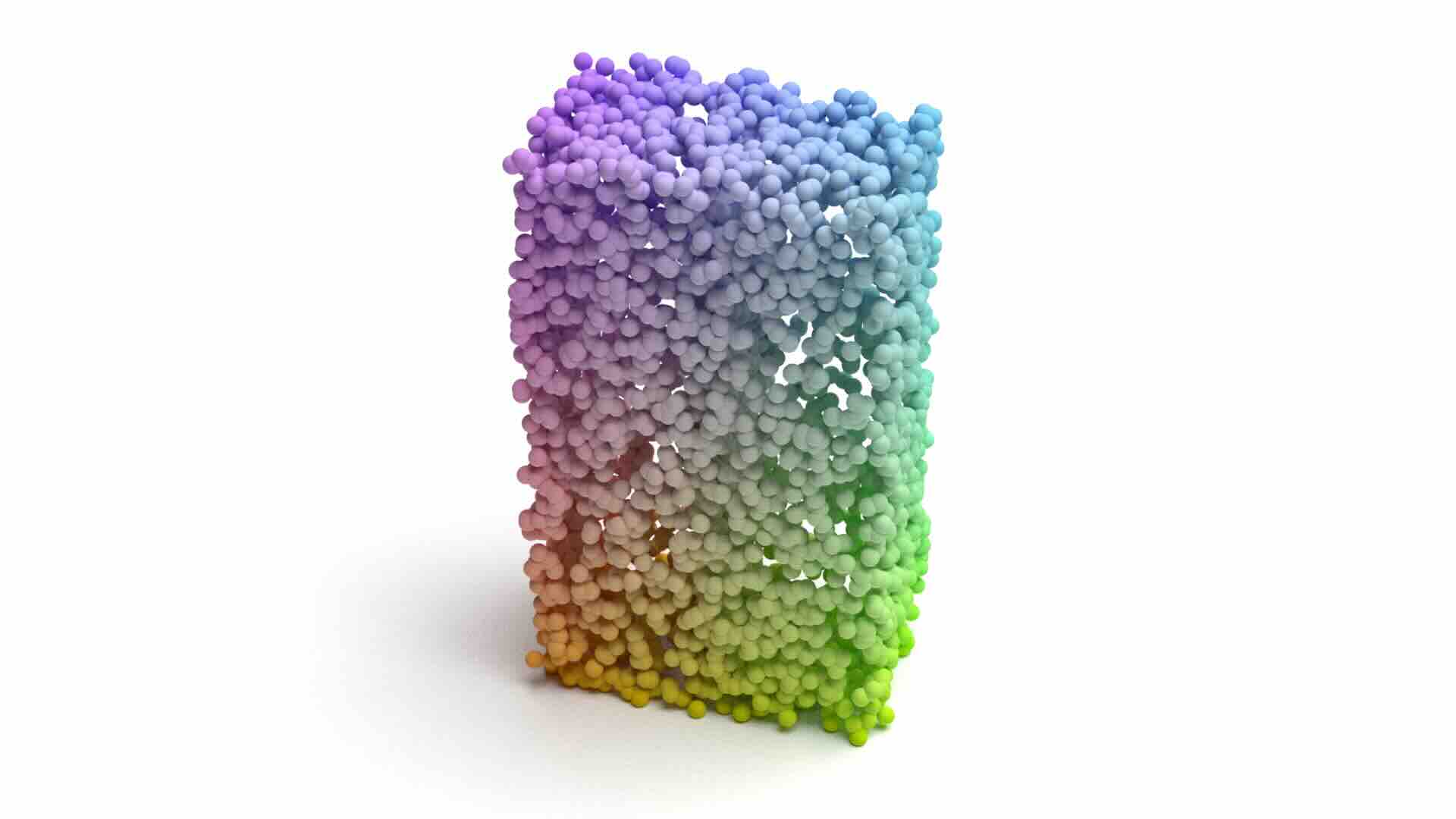} &
        \includegraphics[trim={15cm 0.0cm 15cm 0.0cm},clip,width=0.12\textwidth]{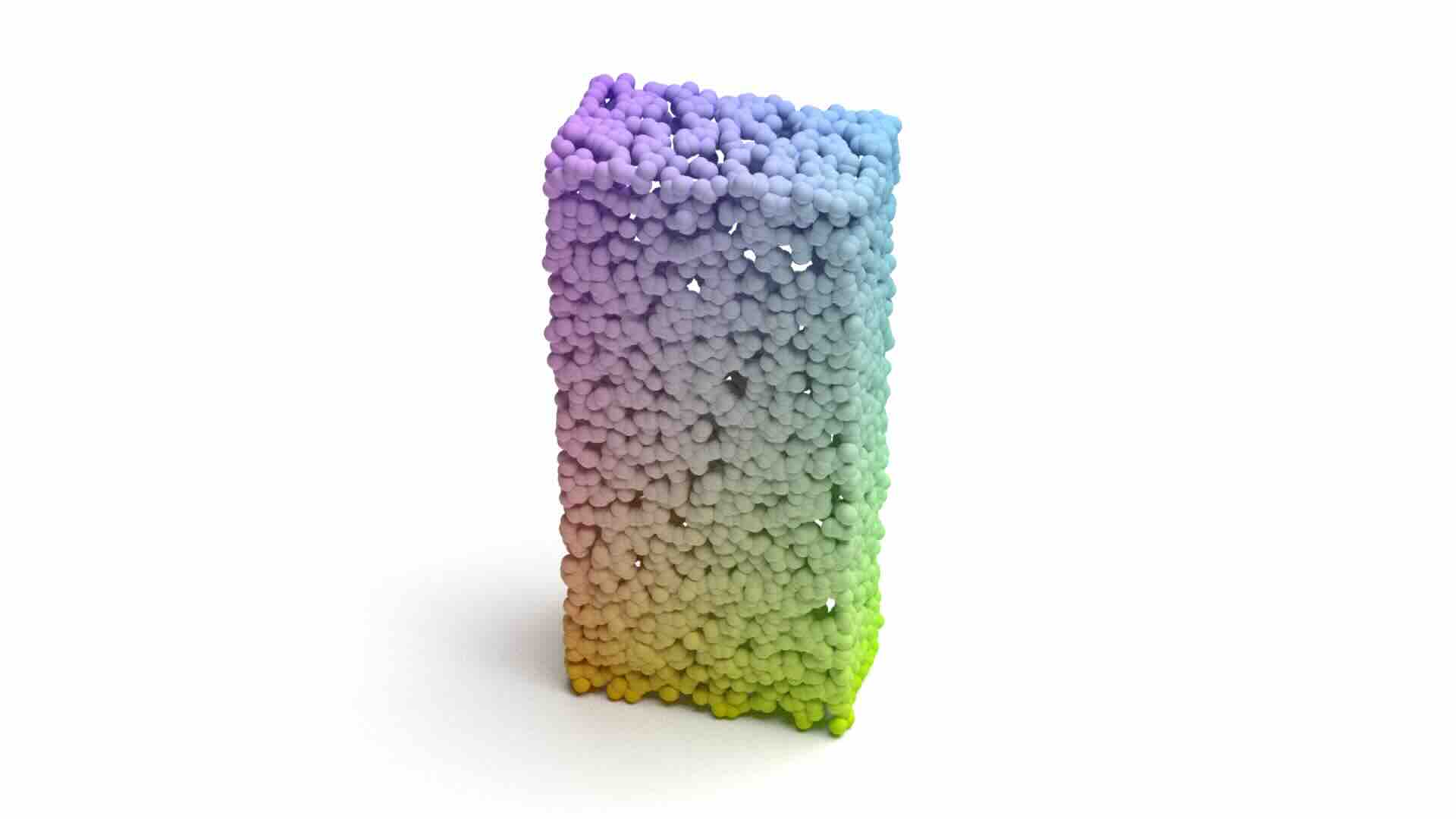} &
        \includegraphics[trim={15cm 0.0cm 15cm 0.0cm},clip,width=0.12\textwidth]{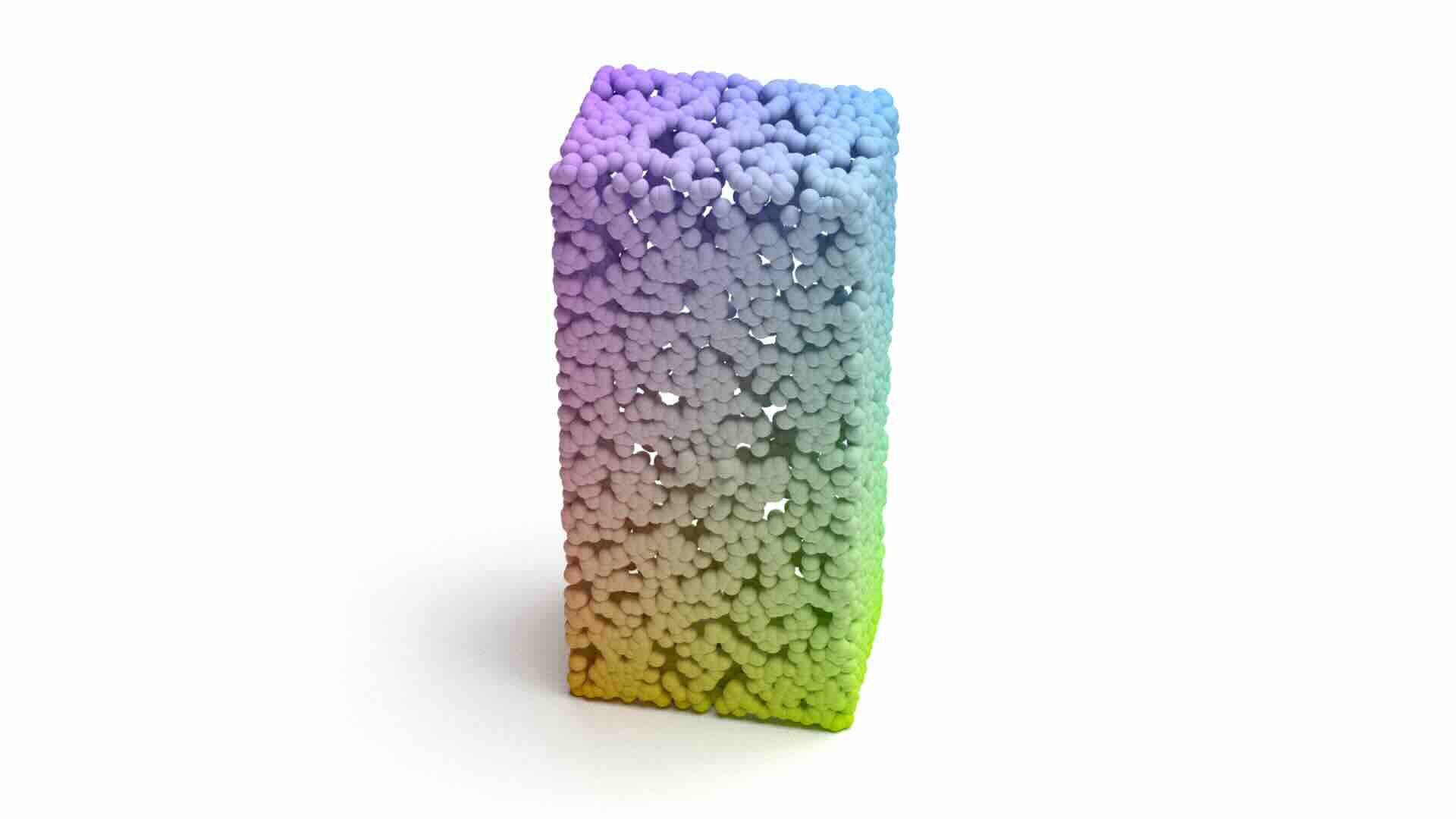}
        \\
               \scriptsize Fused Depths & 
\shortstack[c]{\scriptsize 3DGS-D \\ \scriptsize~\cite{kerbl20233d}} & 
\shortstack[c]{\scriptsize AGS-Mesh \\ \scriptsize~\cite{ren2025ags}} & 
\shortstack[c]{\scriptsize OM-GSD$^{\ast}$ \\ \scriptsize~\cite{lu2025orientation}} & 
\shortstack[c]{\scriptsize RVG-GSD$^{\ast}$ \\ \scriptsize~\cite{chang2025reconviagen}} & 
\shortstack[c]{\scriptsize SAM3D-GSD$^{\ast}$ \\ \scriptsize~\cite{chen2025sam}} & 
\scriptsize Ours & 
\scriptsize GT \\
        \end{tabular}
        }
    \caption{\textbf{Qualitative Comparison of Object Reconstruction on Real-World Datasets.} The first three samples (top row) originate from the ScanNet++ dataset~\cite{yeshwanth2023scannet++}, while the remaining three (bottom row) are drawn from ShapeR Evaluation Dataset~\cite{siddiqui2026shaper}. Our method maintains global structural and topological consistency, whereas other baselines exhibit significant geometric incompleteness or perceptual artifacts on real-world dataset.} \label{fig:main_results_realworlddata}
    \vspace{-2mm}
\end{figure*}

\paragraph{\textbf{Baselines.}} 
We benchmark \OURS{} against state-of-the-art frameworks. (1) \textbf{3DGS-D}: A standard 3D Gaussian Splatting~\cite{kerbl20233d} baseline augmented with depth-guided optimization for indoor scenes. (2) \textbf{AGS-Mesh}~\cite{ren2025ags}: A contemporary surface-based reconstruction method utilizing 2D Gaussian primitives for indoor scenes. (3) \textbf{OM-GSD}: A pipeline that initializes 3DGS from a diffusion-based method \textit{OrientationMatters}~\cite{lu2025orientation}. It inherits from TRELLIS~\cite{xiang2025structured} while specifically providing the consistent, object-centric orientation from sparse multi-view images that the original TRELLIS architecture lacks. (4) \textbf{RVG-GSD}: ReconViaGen~\cite{chang2025reconviagen} integrates VGGT geometric reconstruction priors into a 3D diffusion backbone~\cite{xiang2025structured} via coarse-to-fine multi-view conditioning. To ensure a fair comparison, we initialize the 3D Gaussians by feeding our camera poses and depth maps directly into ReconViaGen's pipeline.  
(5) \textbf{SAM3D-GSD}: SAM 3D~\cite{chen2025sam} is a visually grounded 3D reconstruction foundation model trained via a human-in-the-loop pipeline on unprecedented real-world annotations to jointly predict geometry, texture, and layout under severe occlusion and scene clutter. For a fair comparison, we extract the 3D Gaussian generated by SAM 3D with its multi-view implementation~\footnote{https://github.com/devinli123/multi-view-sam-3d-objects}, supported by our camera poses and depth maps. The generated 3D Gaussians from (3)–(5) are then refined under the standard GSD protocol, providing a comparative analysis between our dual-space gradient steering and feed-forward shape priors.

\subsection{Evaluation on Synthetic Data}
As summarized in Tab.~\ref{tab:combined_geometry_results}, we evaluate our framework on 3D-FRONT~\cite{fu20213d} synthetic dataset. Our approach significantly outperforms existing 3DGS-based baselines across all geometric metrics. Traditional optimization-based GS methods, such as 3DGS-D~\cite{kerbl20233d} and AGS-Mesh~\cite{ren2025ags}, exhibit a tendency to overfit observed views, accurately reconstructing visible regions but failing to generalize to unseen areas. In contrast, our method leverages a generative prior to achieve holistic surface completion, maintaining a superior balance between the pretrained data manifold ($\mathcal{M}_{\text{data}}$) and the real-world observation instance manifold ($\mathcal{M}_{\text{real}}$).

The advantage of our approach is further evidenced when compared to the generative baselines (OM-GSD~\cite{lu2025orientation}, RVG-GSD~\cite{chang2025reconviagen} and SAM3D-GSD~\cite{chen2025sam}). While diffusion-based priors often suffer from scale and translation ambiguities inherent in unconstrained latent sampling, our dual-space guided method enforces stringent 3D constraints by anchoring the flow trajectory to physical measurements. This ensures precise scale alignment between the generative prior and the real-world scene, directly contributing to the superior geometric fidelity illustrated in Fig.~\ref{fig:main_results_3dfront}. Consequently, \OURS{} reconstructs more geometrically consistent and topologically sound structures, particularly for occluded object components where standard optimization and unconstrained generation typically fail.

\subsection{Evaluation on Real-World Datasets} 

\begin{table}[t]
\centering
\scalebox{0.9}{
\begin{tabular}{l ccc ccc}
\toprule 
& \multicolumn{3}{c}{\makebox[0.33\textwidth][c]{\textbf{ScanNet++}~\cite{yeshwanth2023scannet++}}} 
& \multicolumn{3}{c}{\makebox[0.33\textwidth][c]{\textbf{ShapeR}~\cite{siddiqui2026shaper}}}\\
\cmidrule(lr){2-4} \cmidrule(lr){5-7}

Method & 
\makebox[0.1\textwidth][c]{PSNR \textuparrow} & 
\makebox[0.1\textwidth][c]{SSIM \textuparrow} & 
\makebox[0.1\textwidth][c]{LPIPS \textdownarrow} & 
\makebox[0.1\textwidth][c]{PSNR \textuparrow} & 
\makebox[0.1\textwidth][c]{SSIM \textuparrow} & 
\makebox[0.1\textwidth][c]{LPIPS \textdownarrow}  \\

\midrule
3DGS-D~\cite{kerbl20233d} & 25.35 & 0.967 & 3.69 & 26.21 & 0.971 & 3.89 \\
        AGS-Mesh~\cite{ren2025ags} & 26.24 & 0.969 & 3.52 & 28.09 & 0.974 & 3.75 \\
        OM-GSD$^{\ast}$~\cite{lu2025orientation} & 26.89 & 0.965 & 3.73 & 27.40 & 0.971 & 3.74 \\
        RVG-GSD$^{\ast}$~\cite{chang2025reconviagen} & 27.17 & 0.966 & 3.64 & 28.05 & 0.972 & 4.03 \\
        SAM3D-GSD$^{\ast}$~\cite{chen2025sam} & 27.88 & \textbf{0.972} & 3.51 & 27.40 & 0.971 & 3.74 \\
        \textbf{Ours} & \textbf{28.73} & \textbf{0.972} & \textbf{3.49} & \textbf{28.54}& \textbf{0.975} & \textbf{3.73}\\
\bottomrule
\end{tabular}
}
\caption{\textbf{Quantitative Rendering Comparison on Challenging ScanNet++~\cite{yeshwanth2023scannet++} and ShapeR Evaluation Dataset~\cite{siddiqui2026shaper} Samples.} We evaluate novel-view synthesis specifically on challenging, partially unobserved regions under extremely sparse viewing conditions. Our method demonstrates significant robustness, providing high-quality reconstructions in challenging regions where existing baselines exhibit limited performance. (LPIPS $\times 10^2$)}
\label{tab:main_rendering_scannetpp}
\vspace{-2mm}
\end{table}

To demonstrate the effectiveness of our framework in practical applications, we conduct extensive evaluations on real-world data using the ScanNet++ dataset~\cite{yeshwanth2023scannet++} and ShapeR Evaluation Dataset~\cite{siddiqui2026shaper}. These datasets cover a range of real-world acquisition scenarios and challenge our model with varying degrees of observation uncertainty and geometric complexity. As demonstrated in Tab.~\ref{tab:combined_geometry_results} and Fig.~\ref{fig:main_results_realworlddata}, our approach consistently produces more coherent and structurally sound shape completions compared to baselines, particularly when operating under sparse and partial observations.

\begin{figure*}[t] 
    \centering
    \vspace{-1mm}
    \setlength{\tabcolsep}{2pt}
    \resizebox{\textwidth}{!}{
    \begin{tabular}{cccccccc}
        \includegraphics[width=0.12\textwidth]{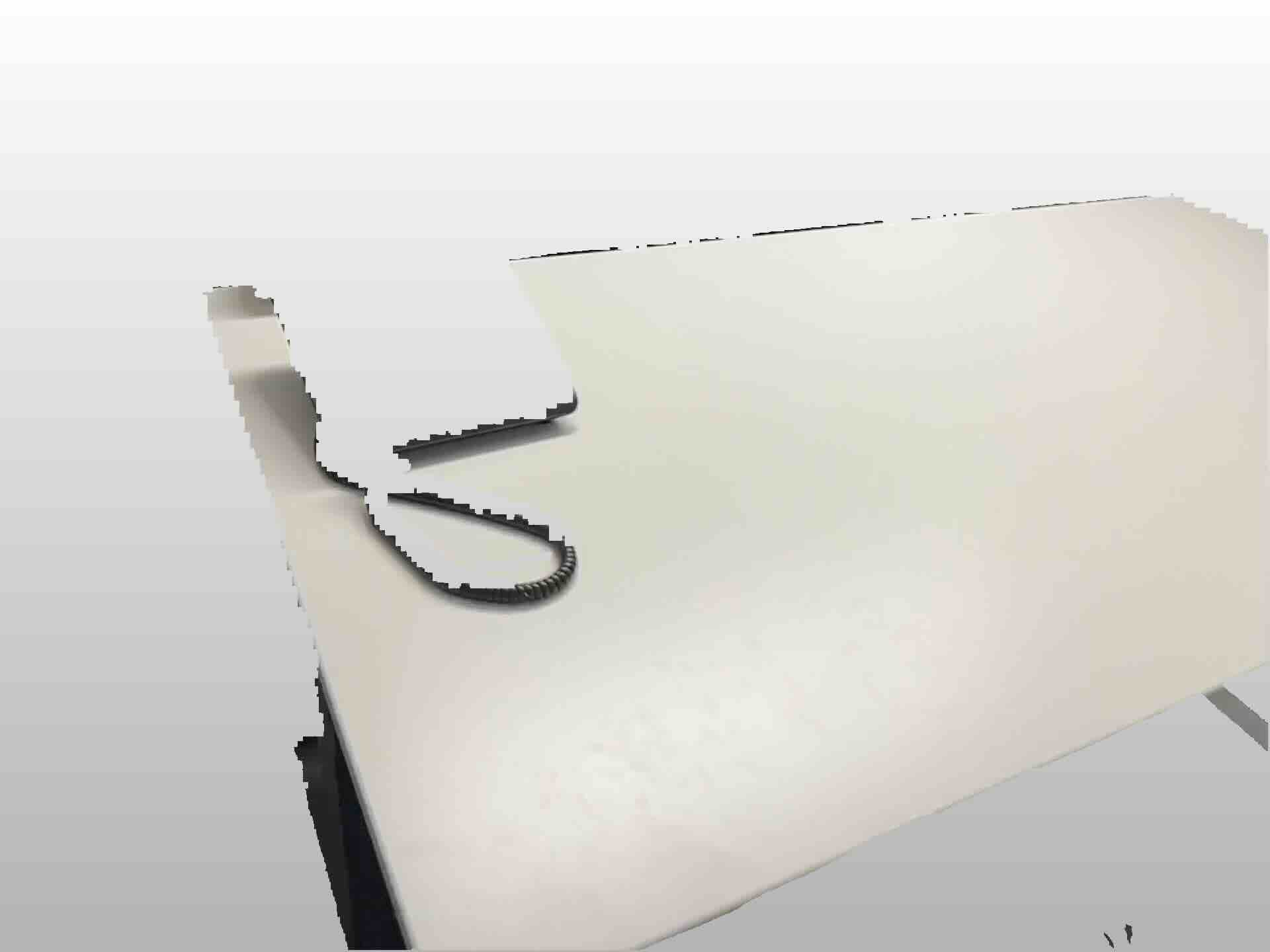} &
        \includegraphics[width=0.12\textwidth]{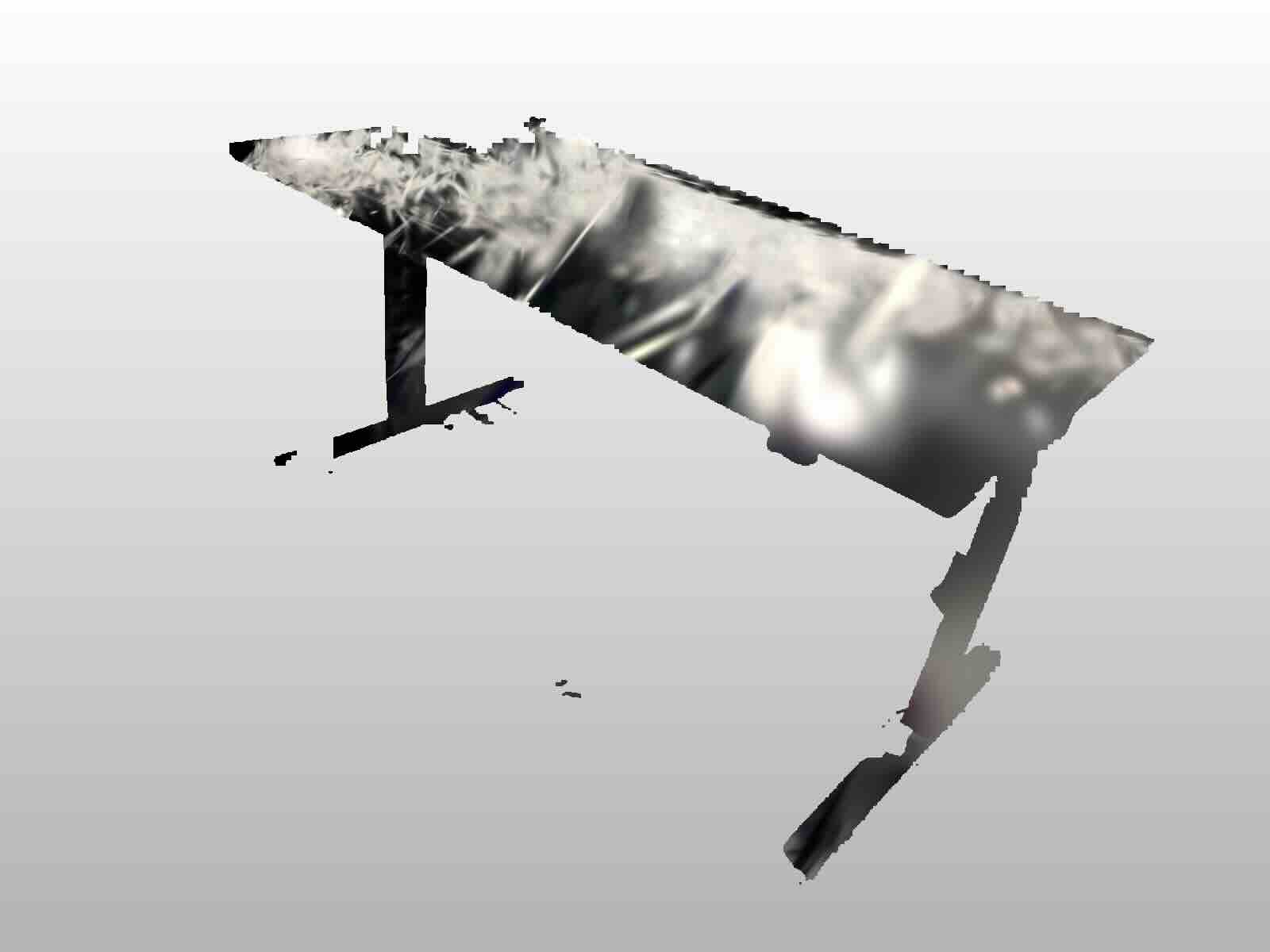} &
        \includegraphics[ width=0.12\textwidth]{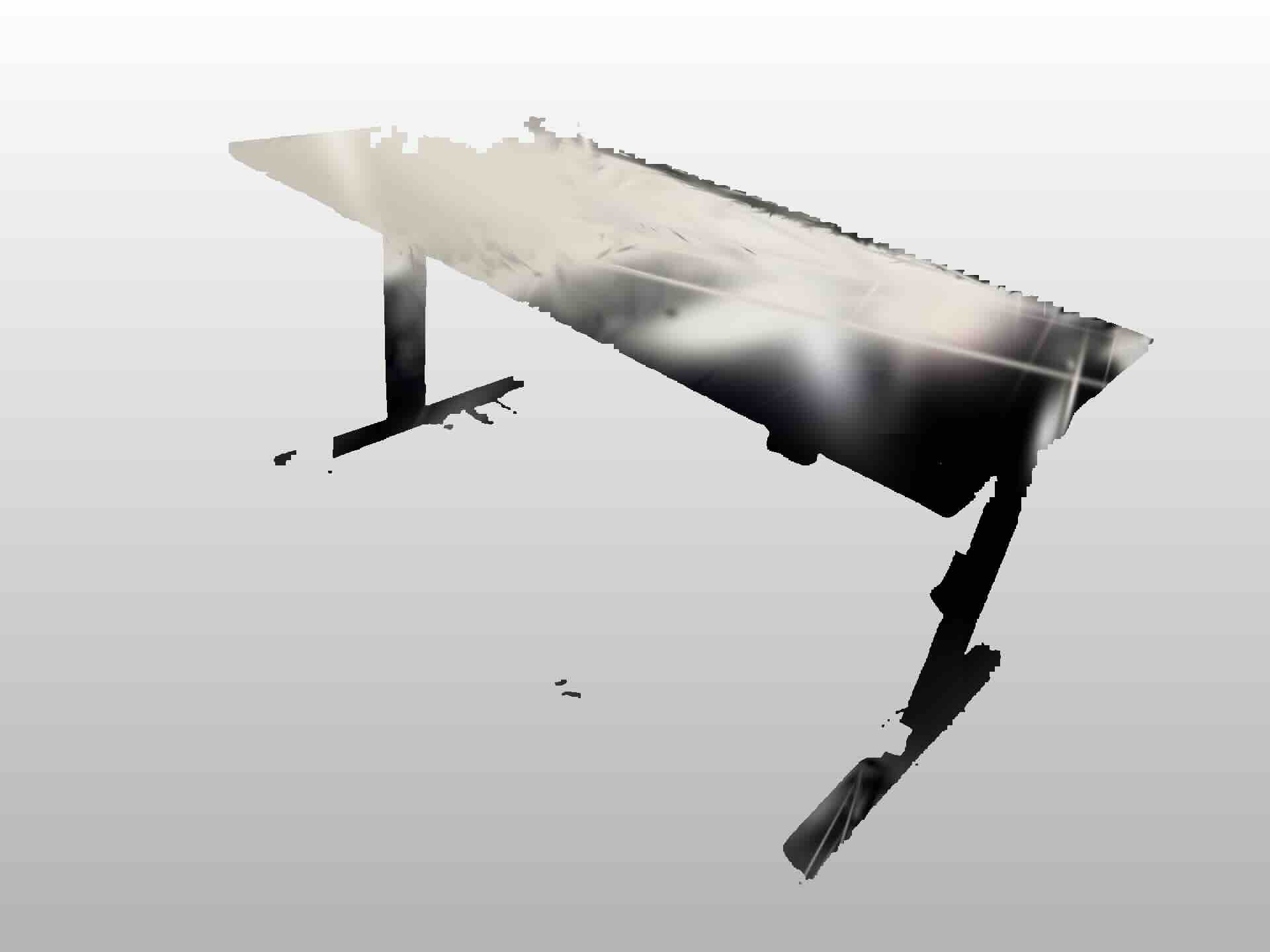} &
        \includegraphics[ width=0.12\textwidth]{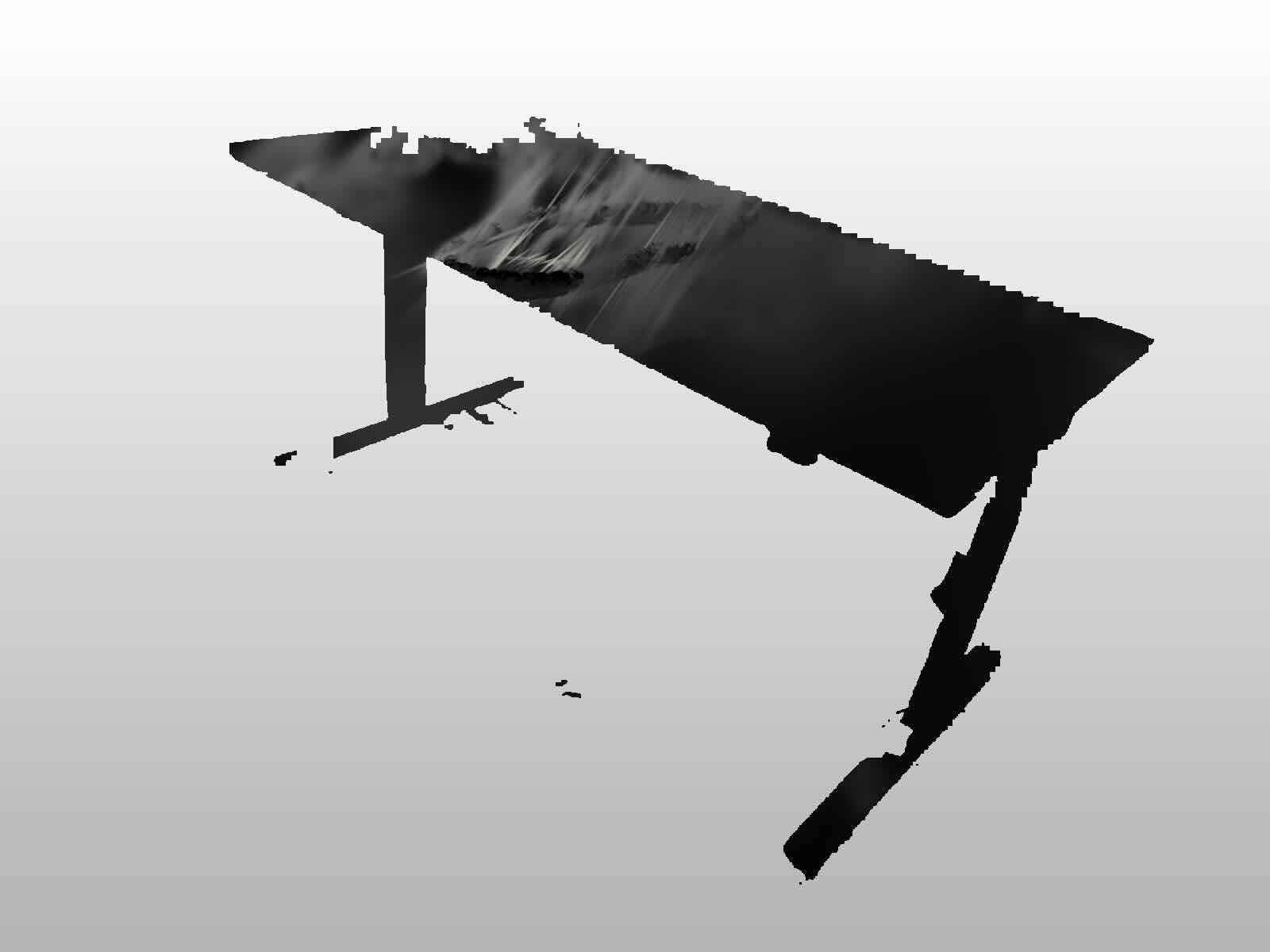} &
        \includegraphics[width=0.12\textwidth]{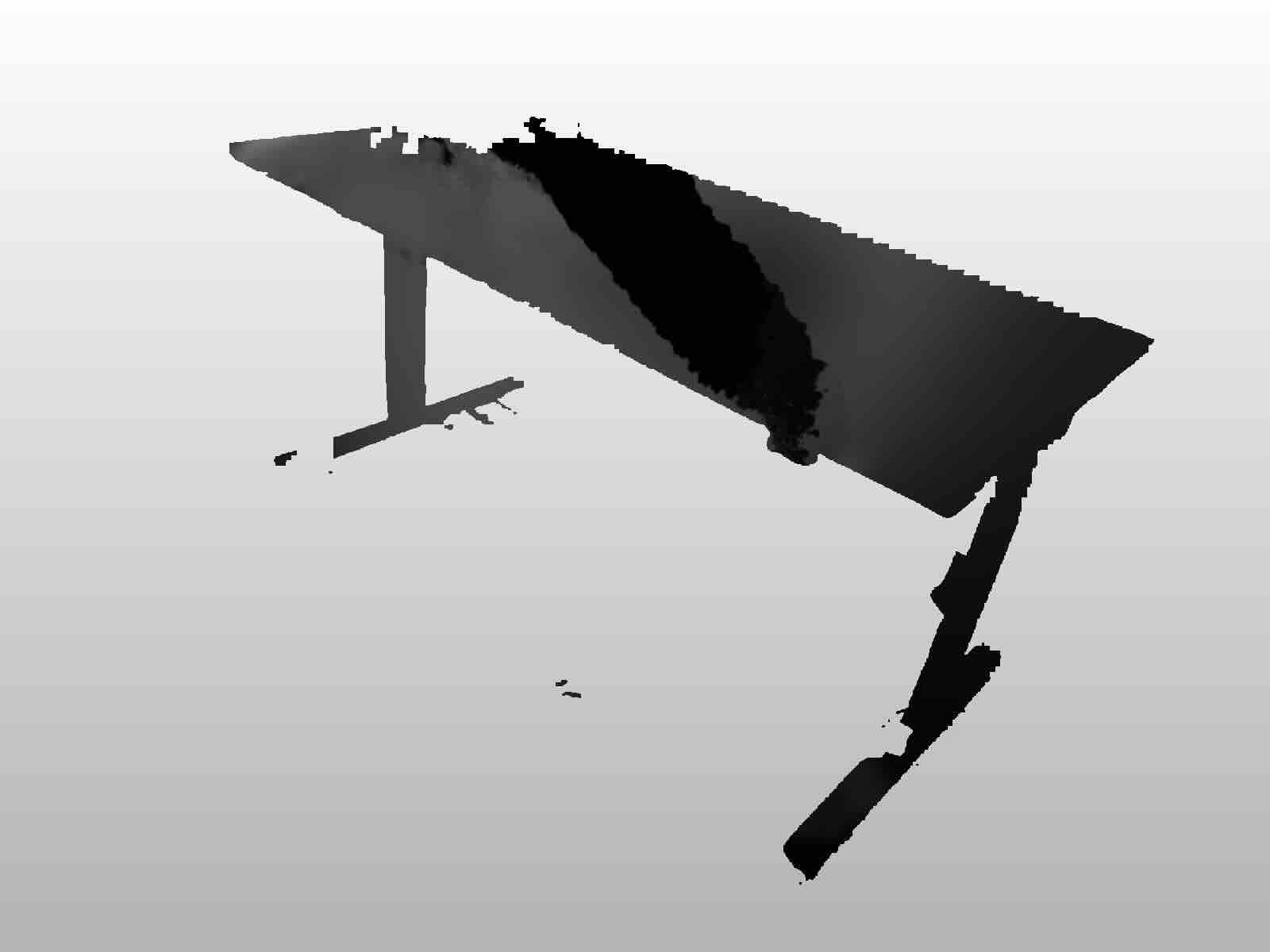} &
        \includegraphics[width=0.12\textwidth]{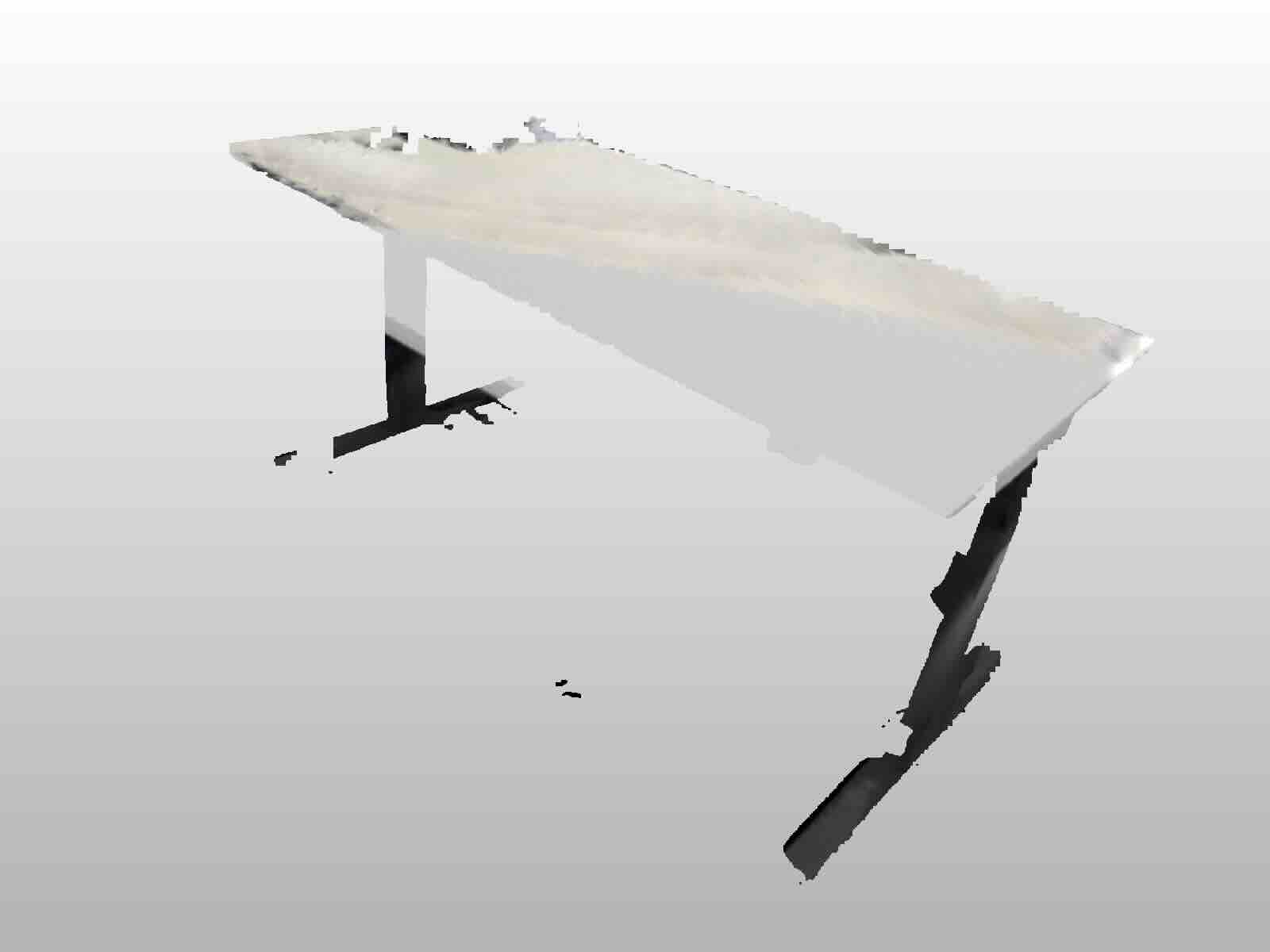} &
        \includegraphics[width=0.12\textwidth]{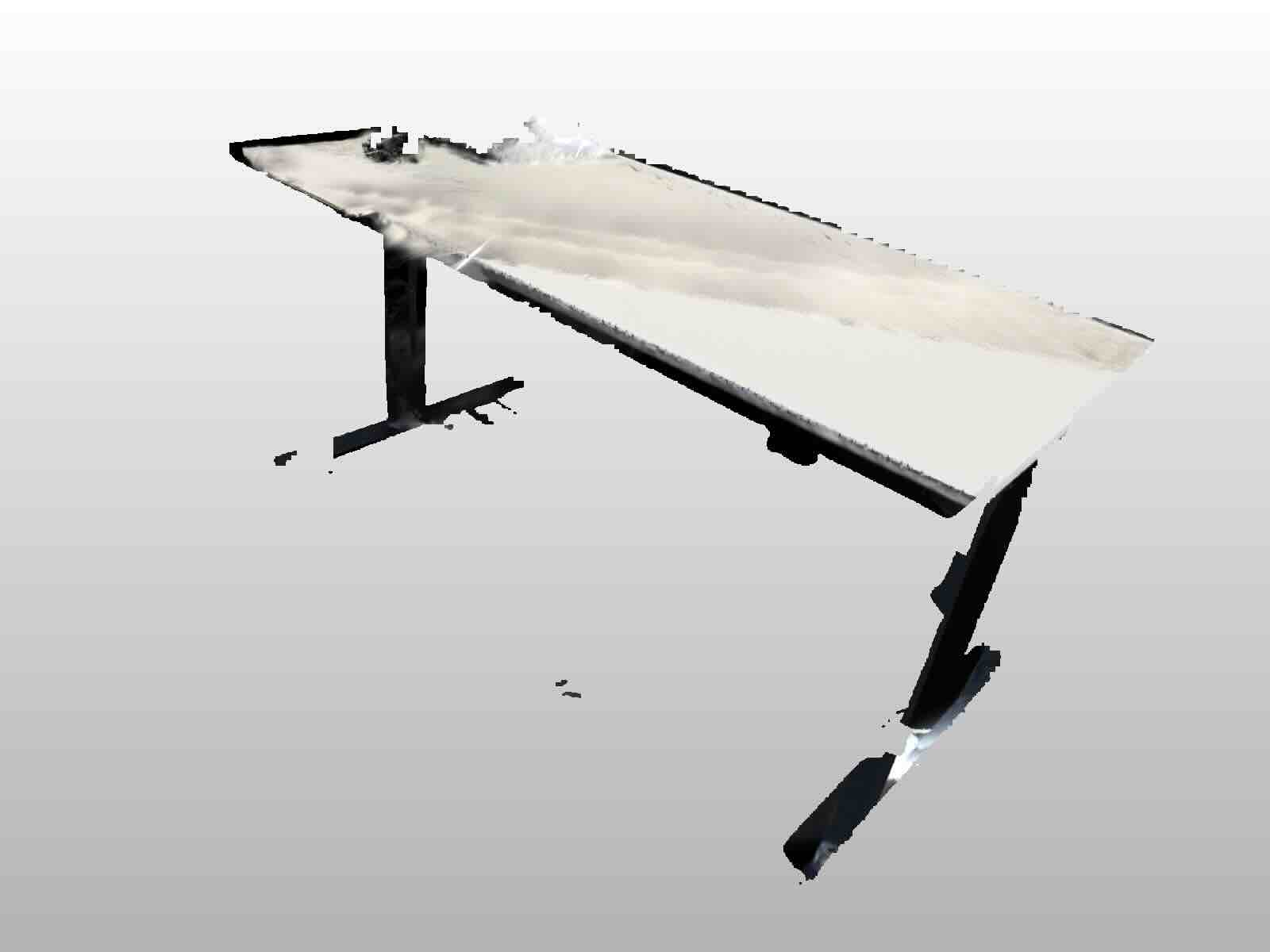} &
        \includegraphics[width=0.12\textwidth]{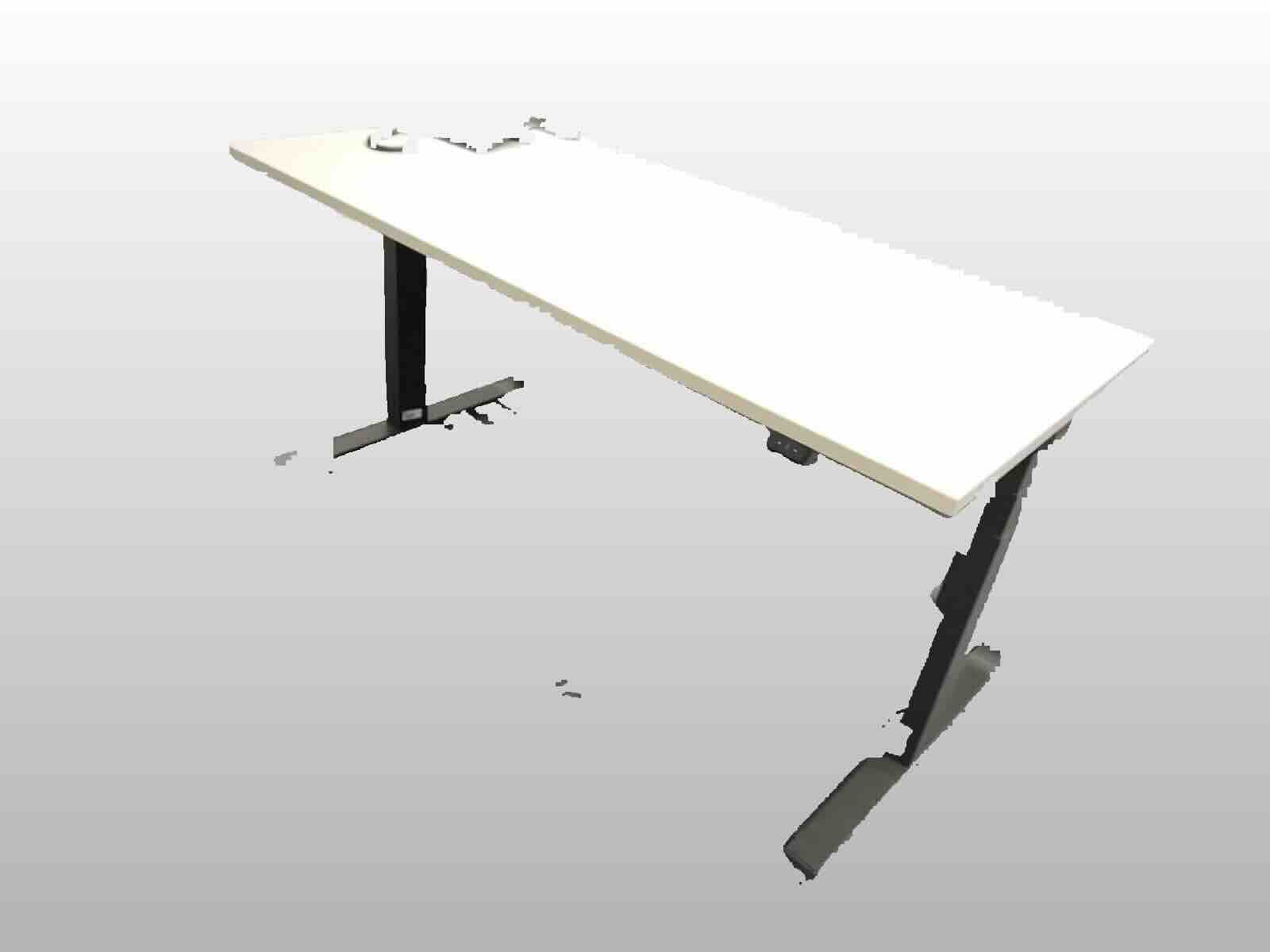}
        \\
        \includegraphics[trim={10.0cm 0.0cm 0.0cm 2.5cm},clip,width=0.12\textwidth]{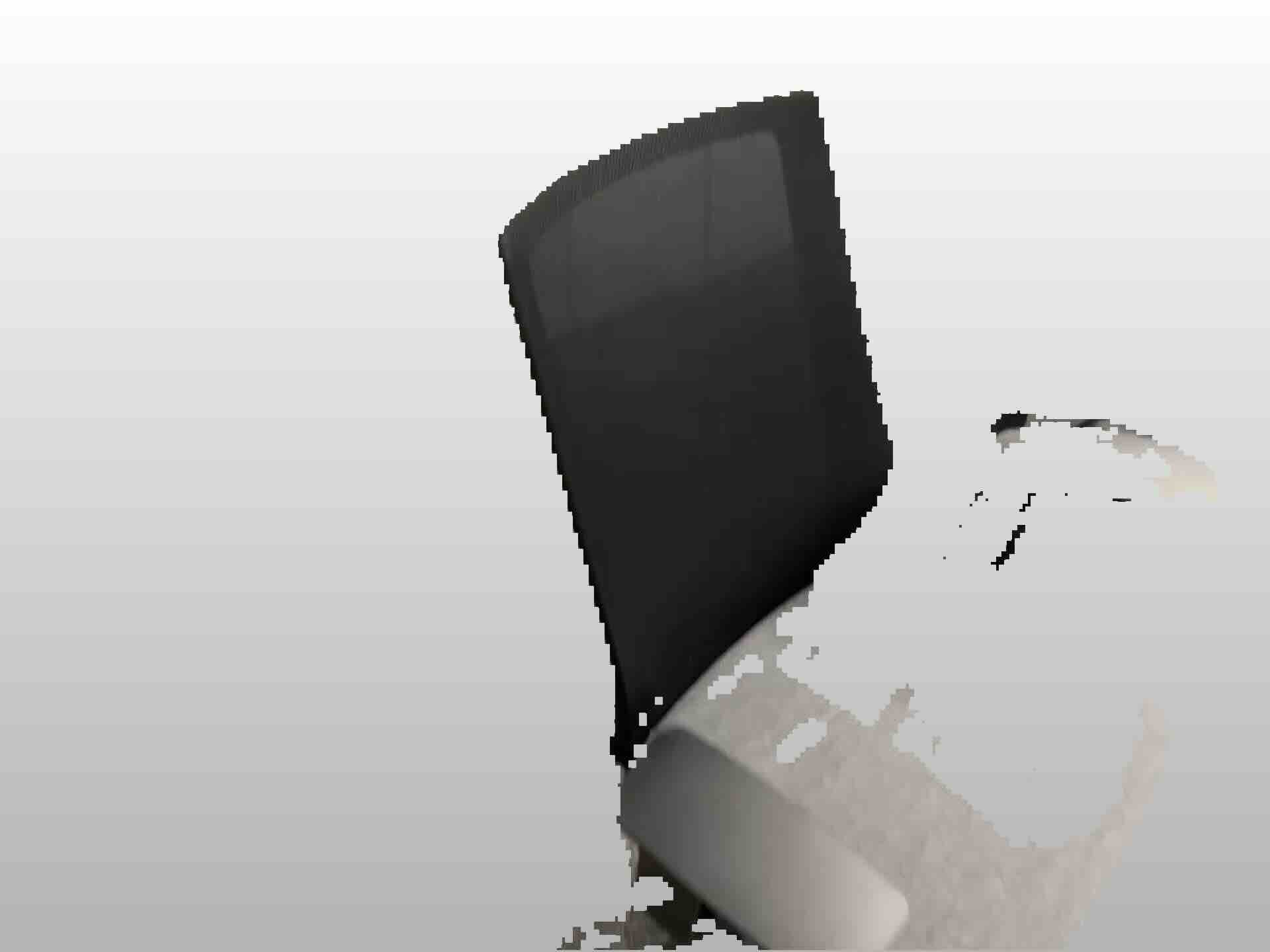} &
        \includegraphics[trim={0.0cm 11.5cm 20cm 0.0cm},clip,width=0.12\textwidth]{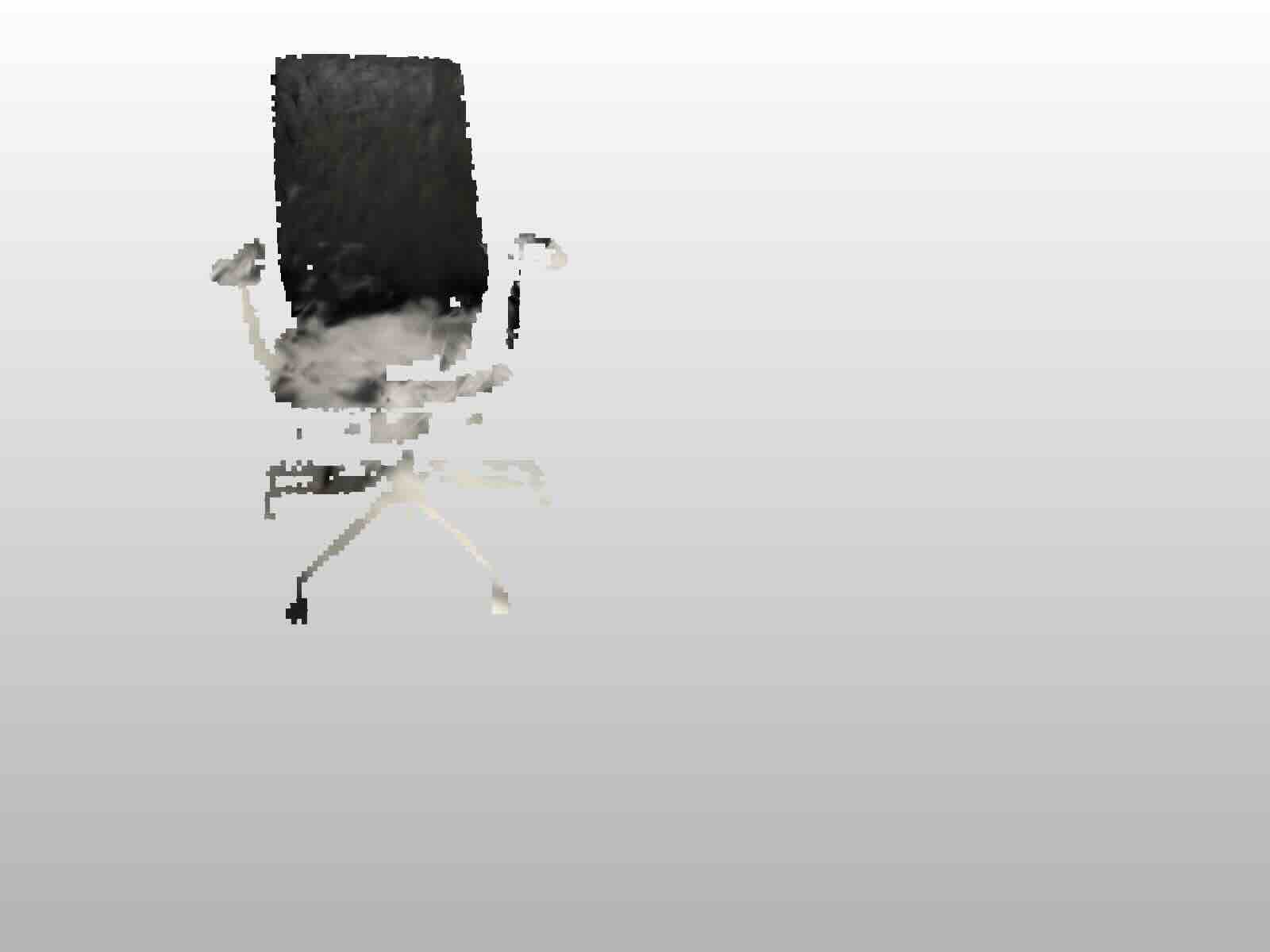} &
        \includegraphics[trim={0.0cm 11.5cm 20cm 0.0cm},clip,width=0.12\textwidth]{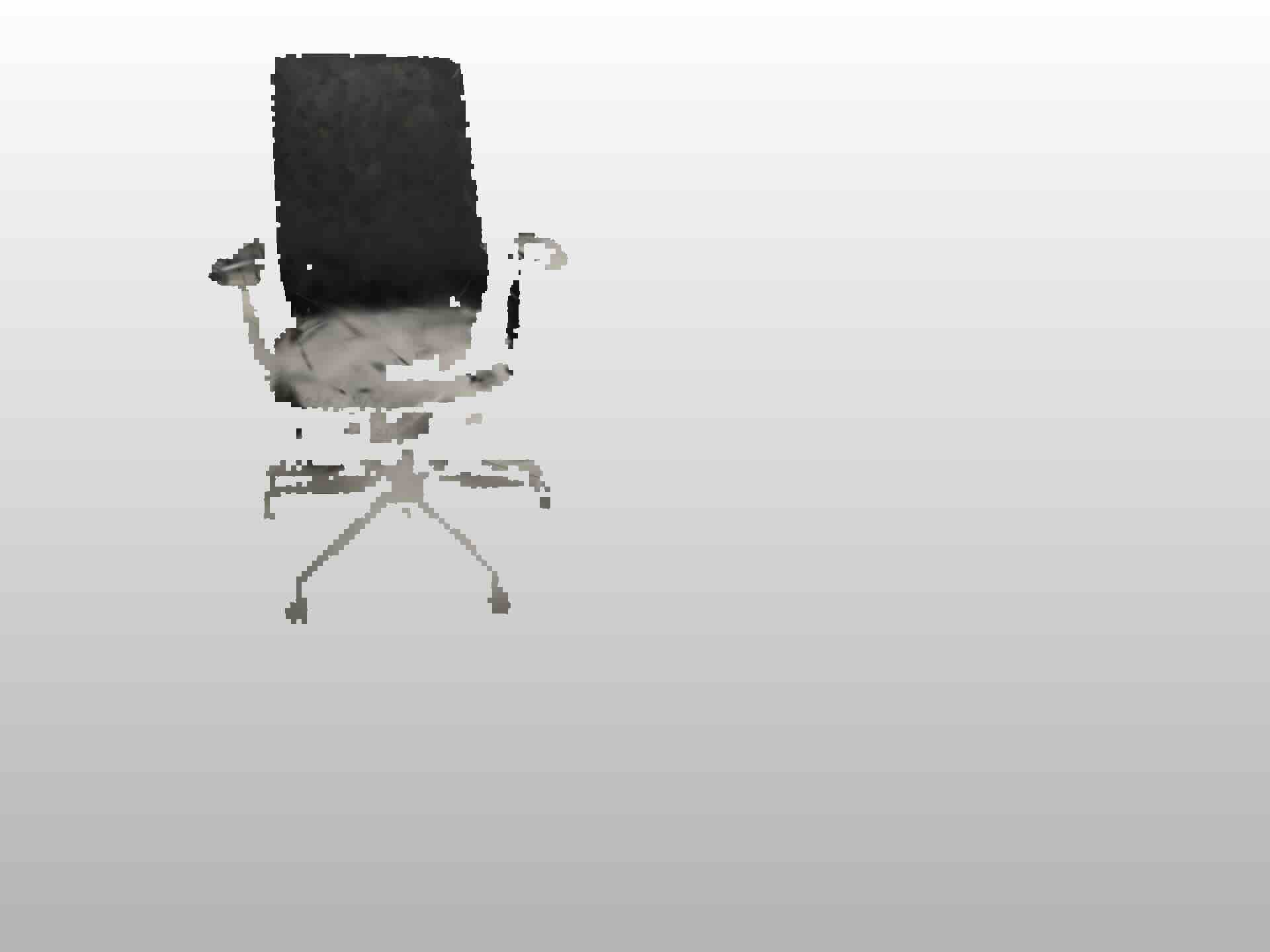} &
        \includegraphics[trim={0.0cm 11.5cm 20cm 0.0cm},clip, width=0.12\textwidth]{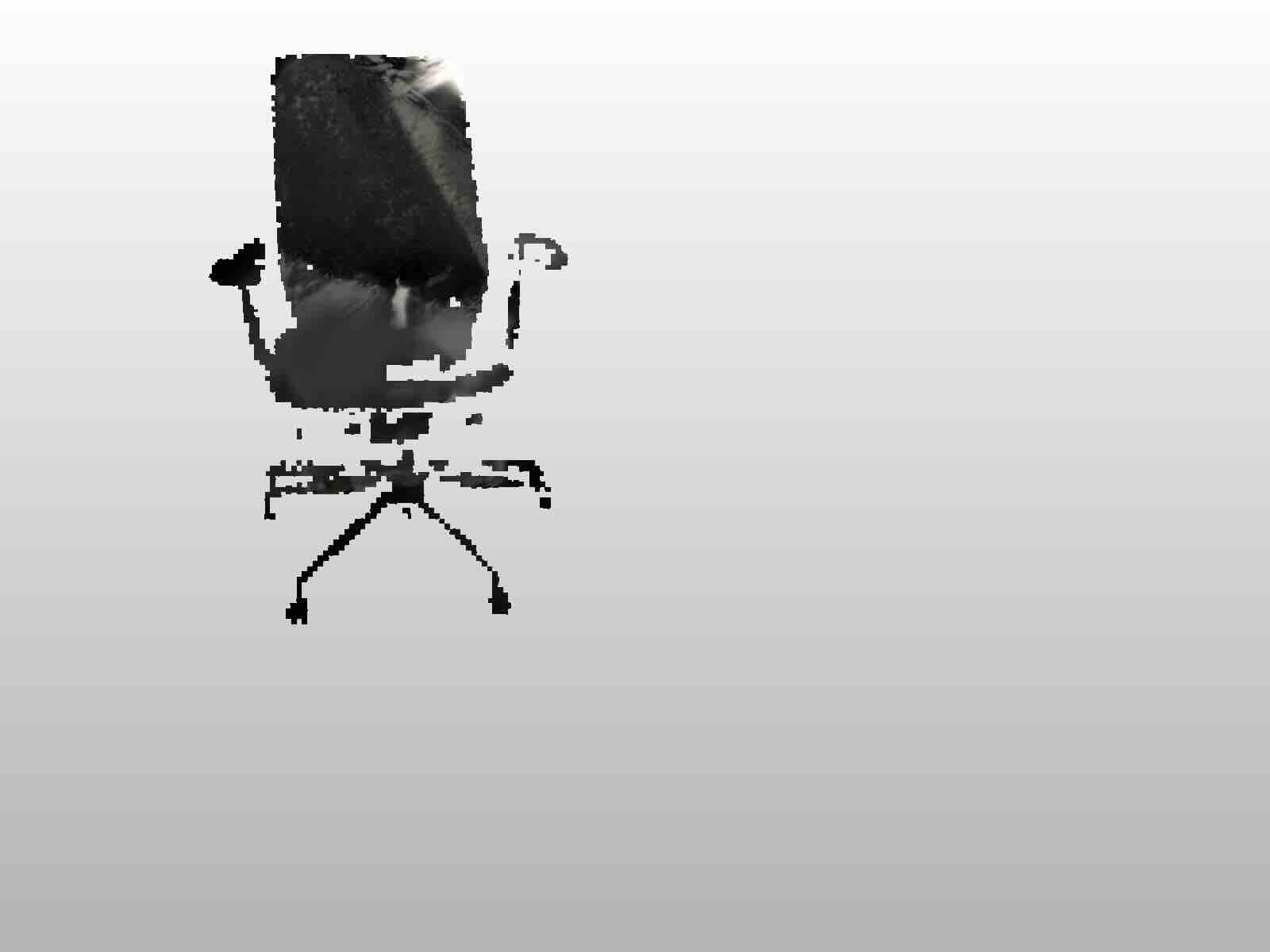} &
        \includegraphics[trim={0.0cm 11.5cm 20cm 0.0cm},clip,width=0.12\textwidth]{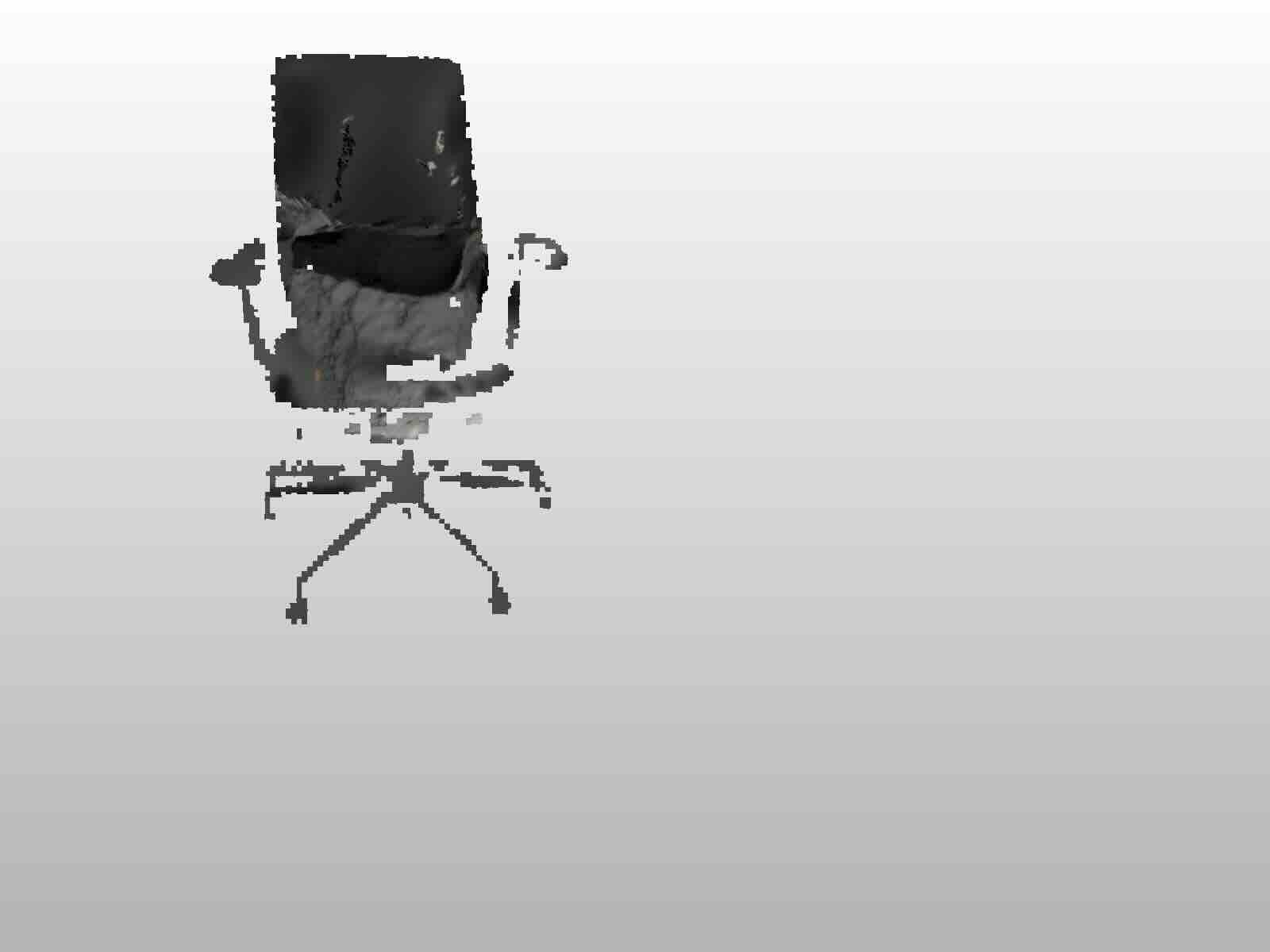} &
        \includegraphics[trim={0.0cm 11.5cm 20cm 0.0cm},clip,width=0.12\textwidth]{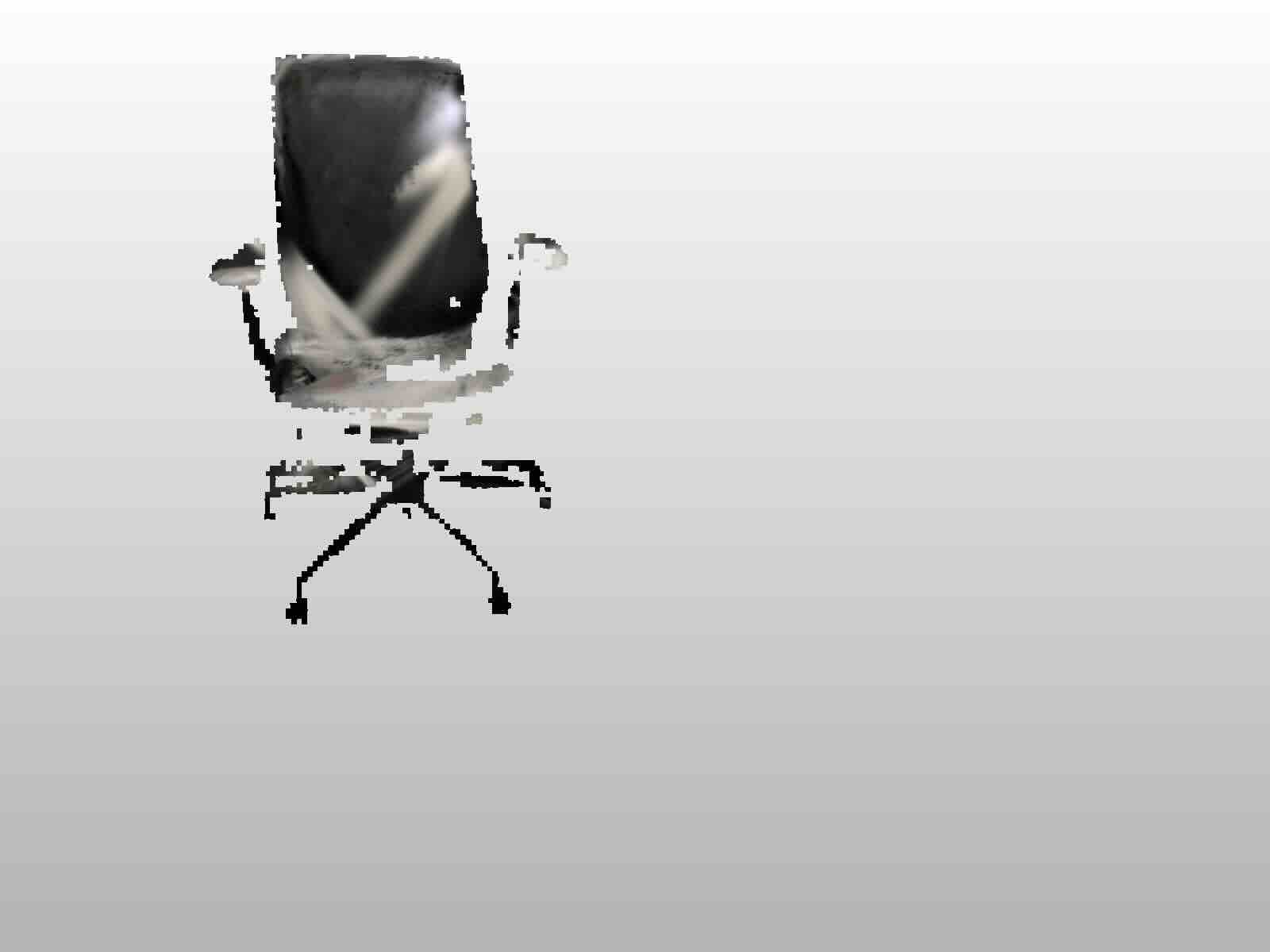} &
        \includegraphics[trim={0.0cm 11.5cm 20cm 0.0cm},clip,width=0.12\textwidth]{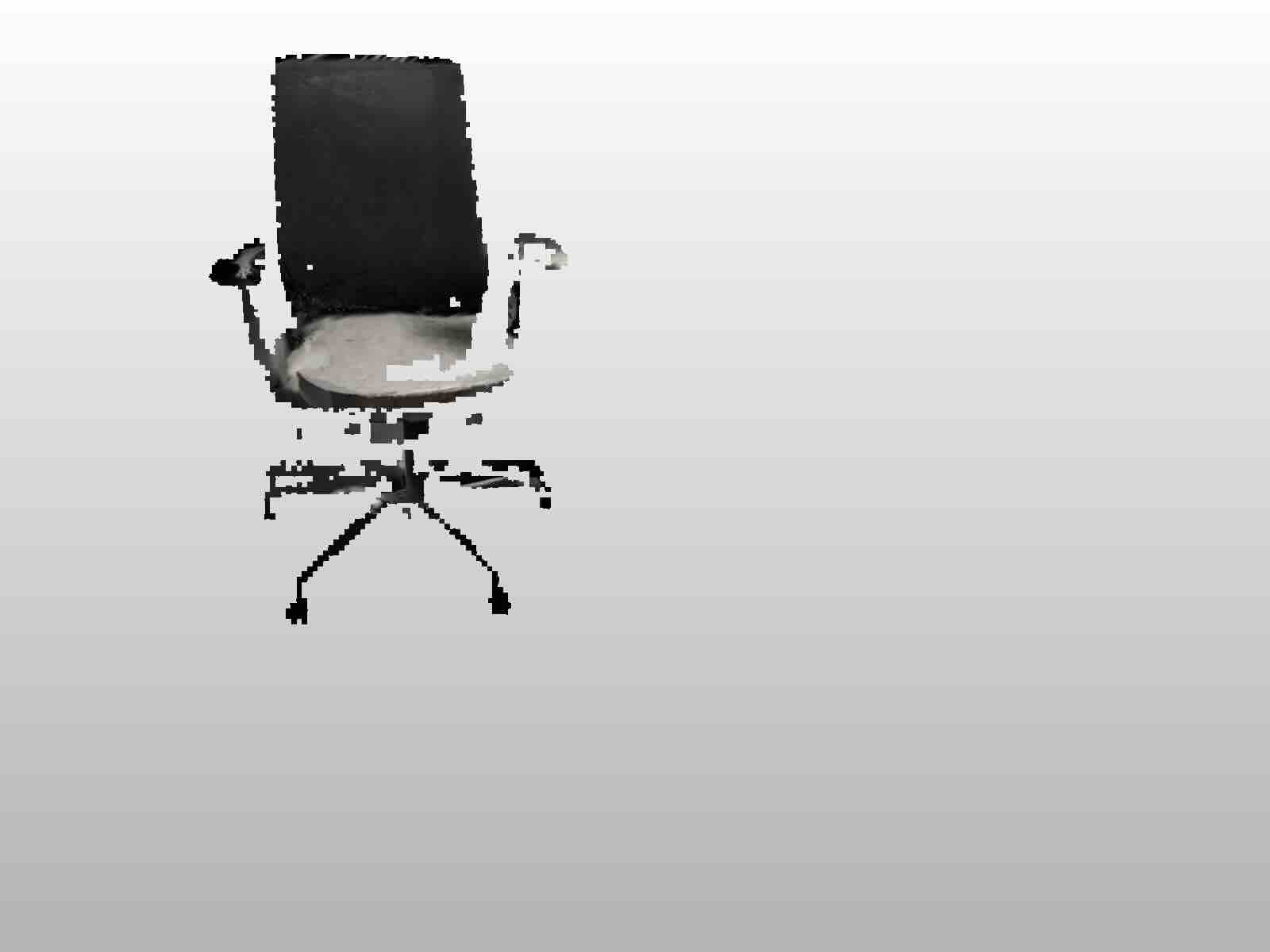} &
        \includegraphics[trim={0.0cm 11.5cm 20cm 0.0cm},clip,width=0.12\textwidth]{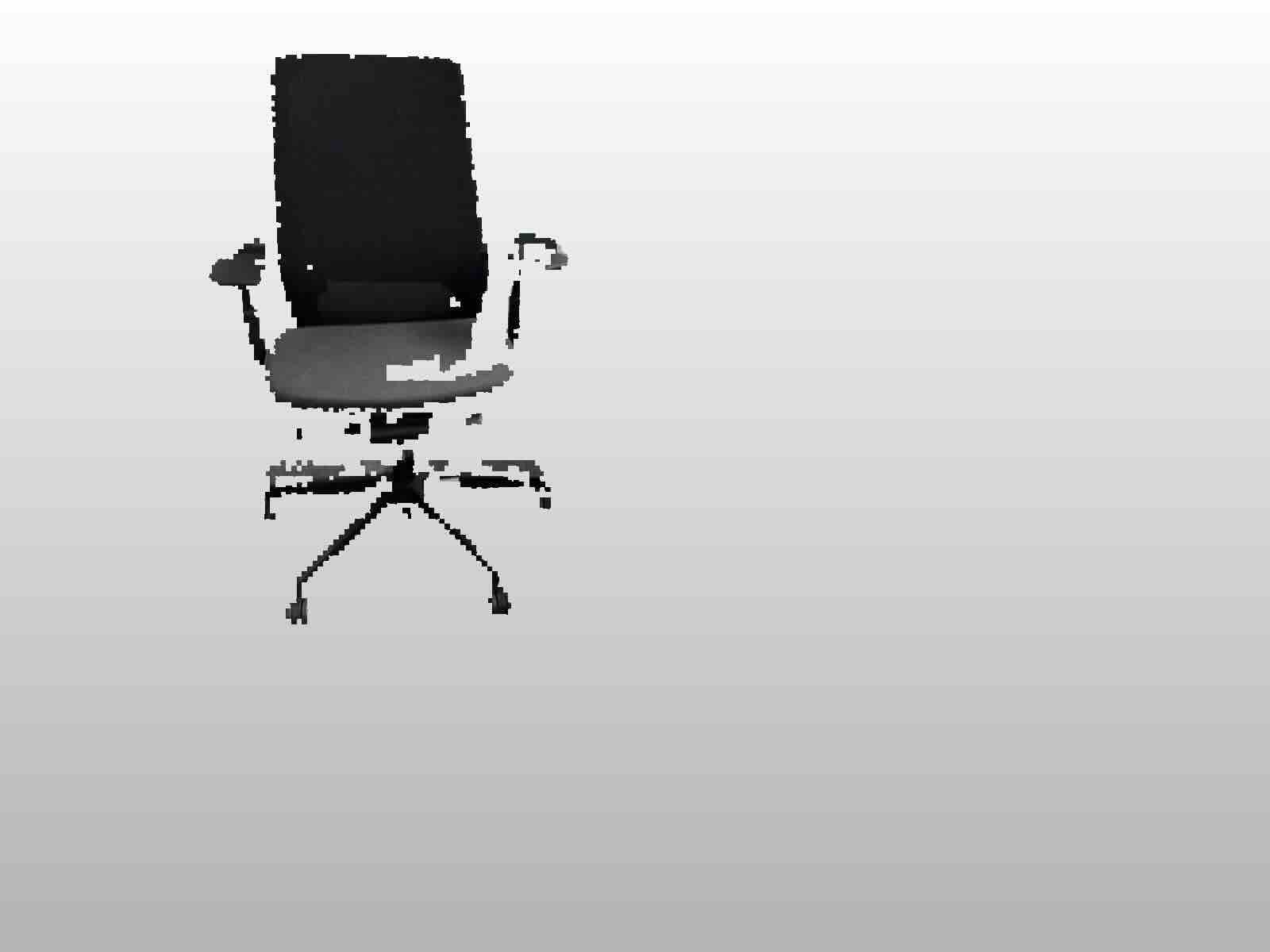}
        \\
        \includegraphics[trim={0.0cm 0.0cm 15cm 20.0cm},clip,width=0.12\textwidth]{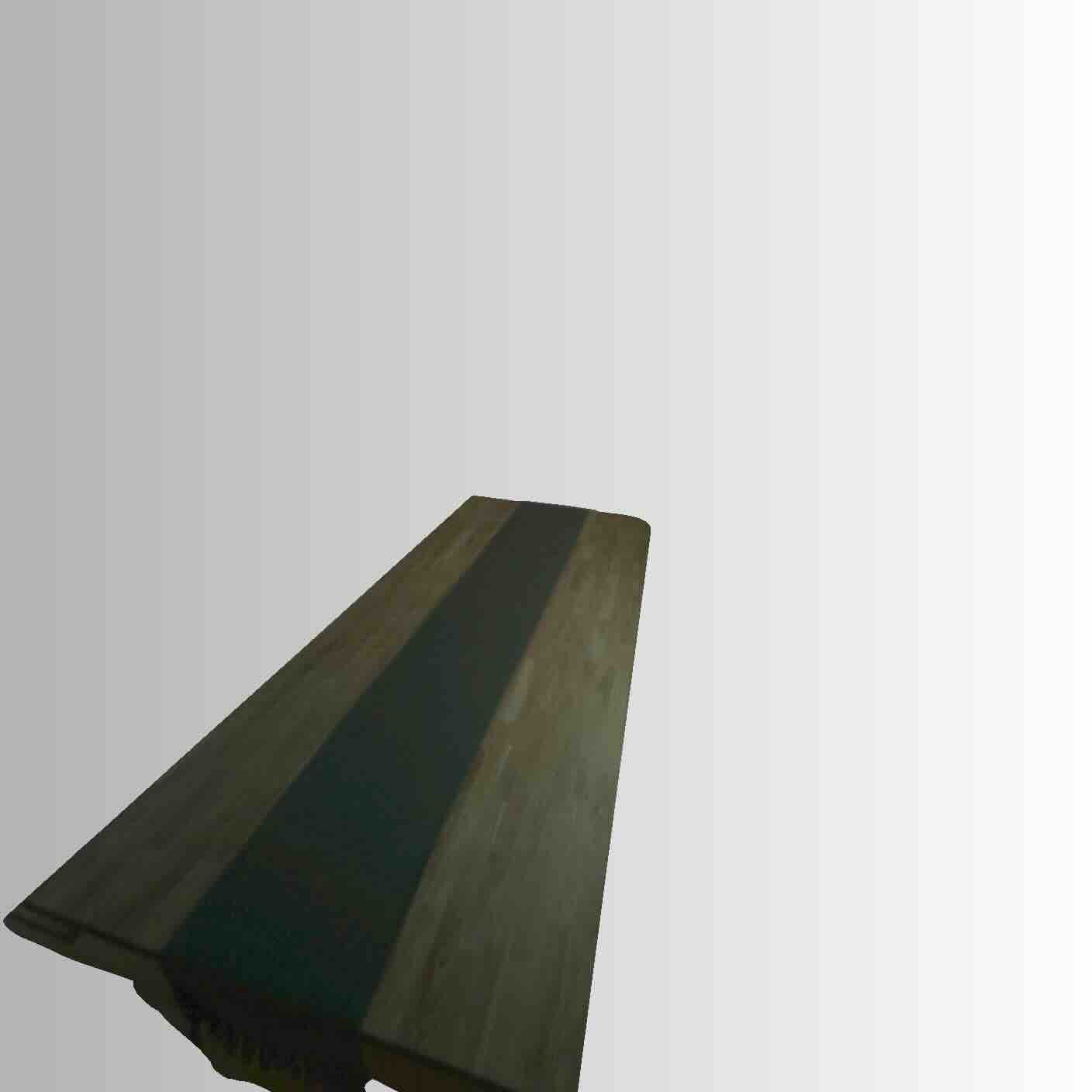} 
        &
        \includegraphics[trim={0.0cm 0.0cm 15cm 20.0cm},clip,width=0.12\textwidth]{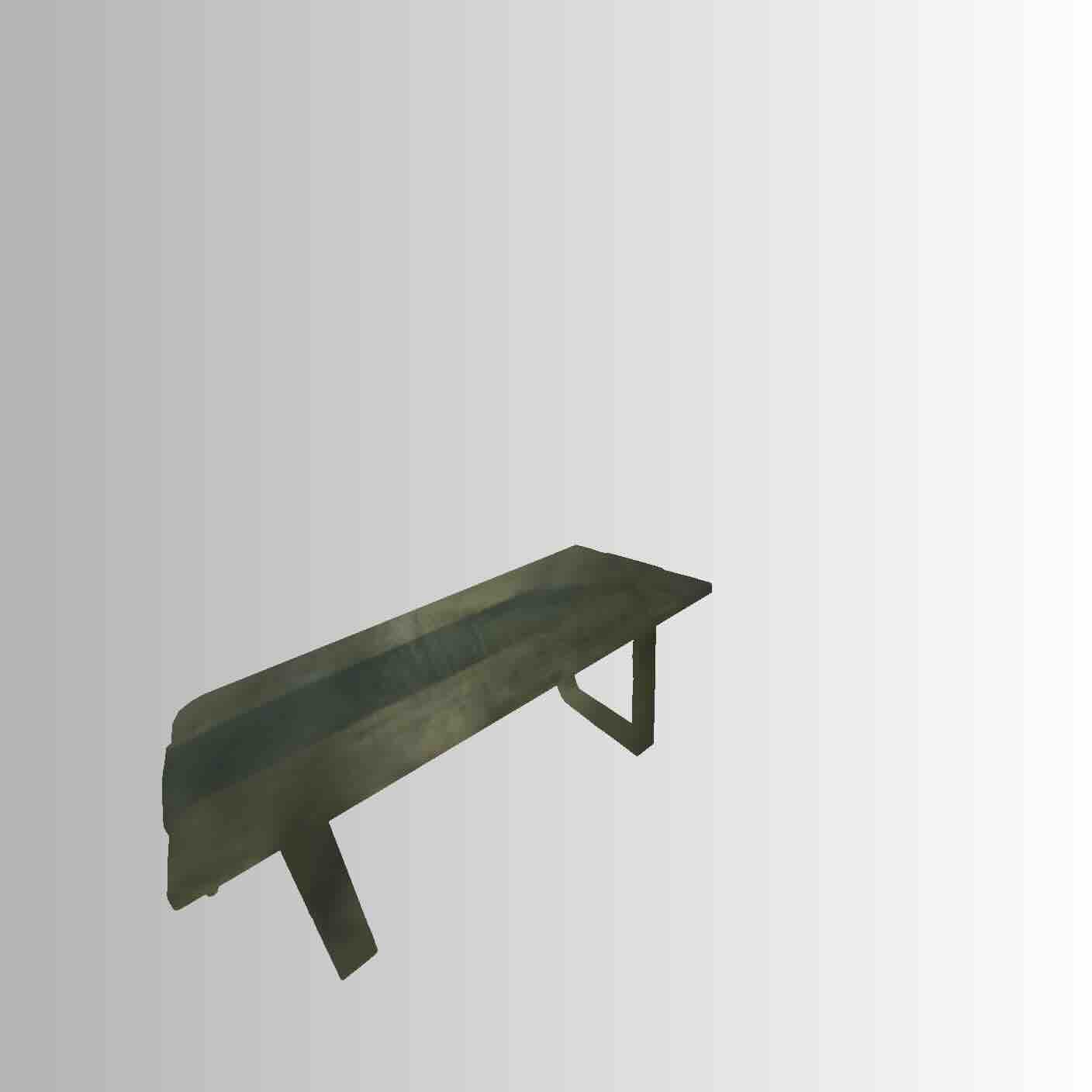} &
        \includegraphics[trim={0.0cm 0.0cm 15cm 20.0cm},clip,width=0.12\textwidth]{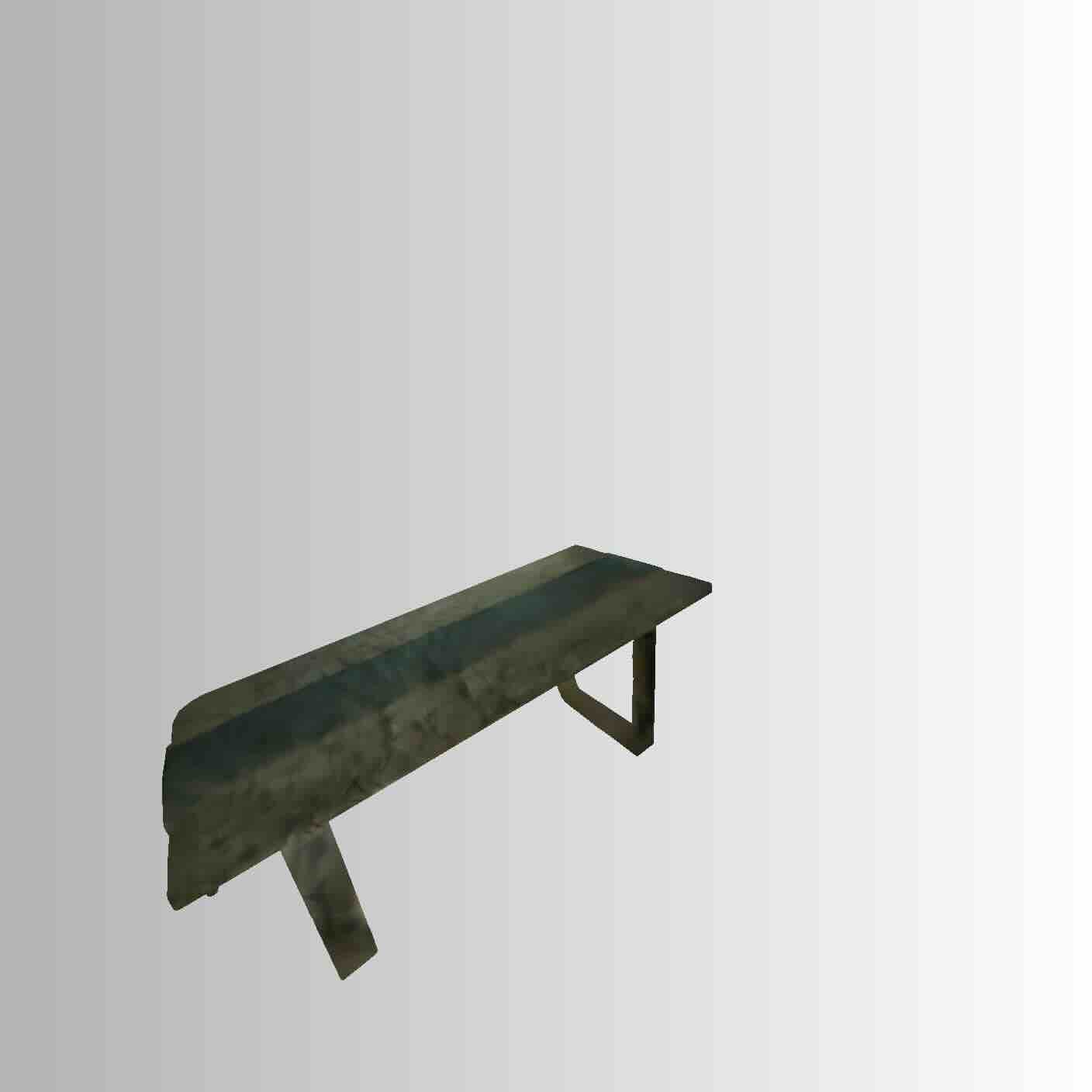} &
        \includegraphics[trim={0.0cm 0.0cm 15cm 20.0cm},clip,width=0.12\textwidth]{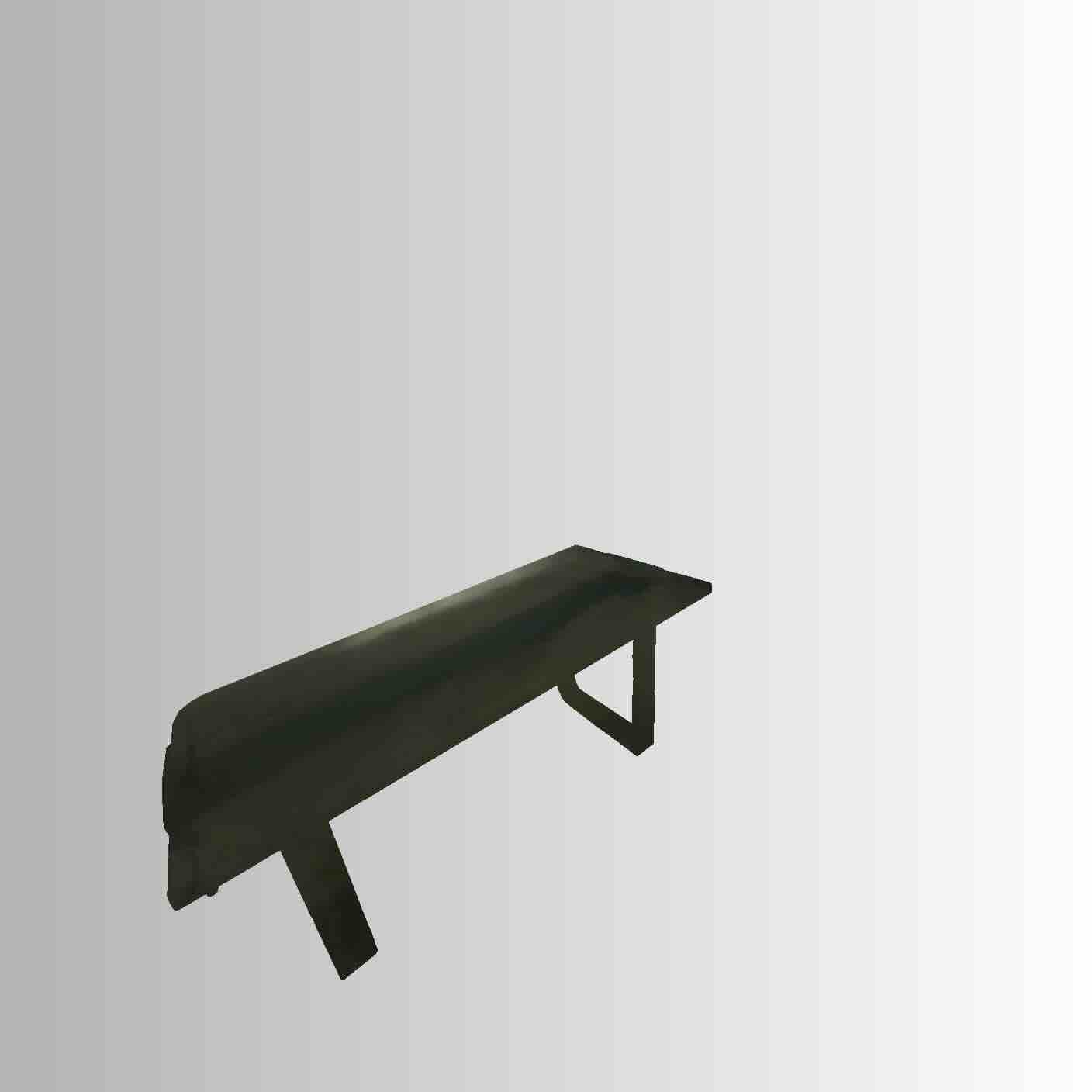} &
        \includegraphics[trim={0.0cm 0.0cm 15cm 20.0cm},clip,width=0.12\textwidth]{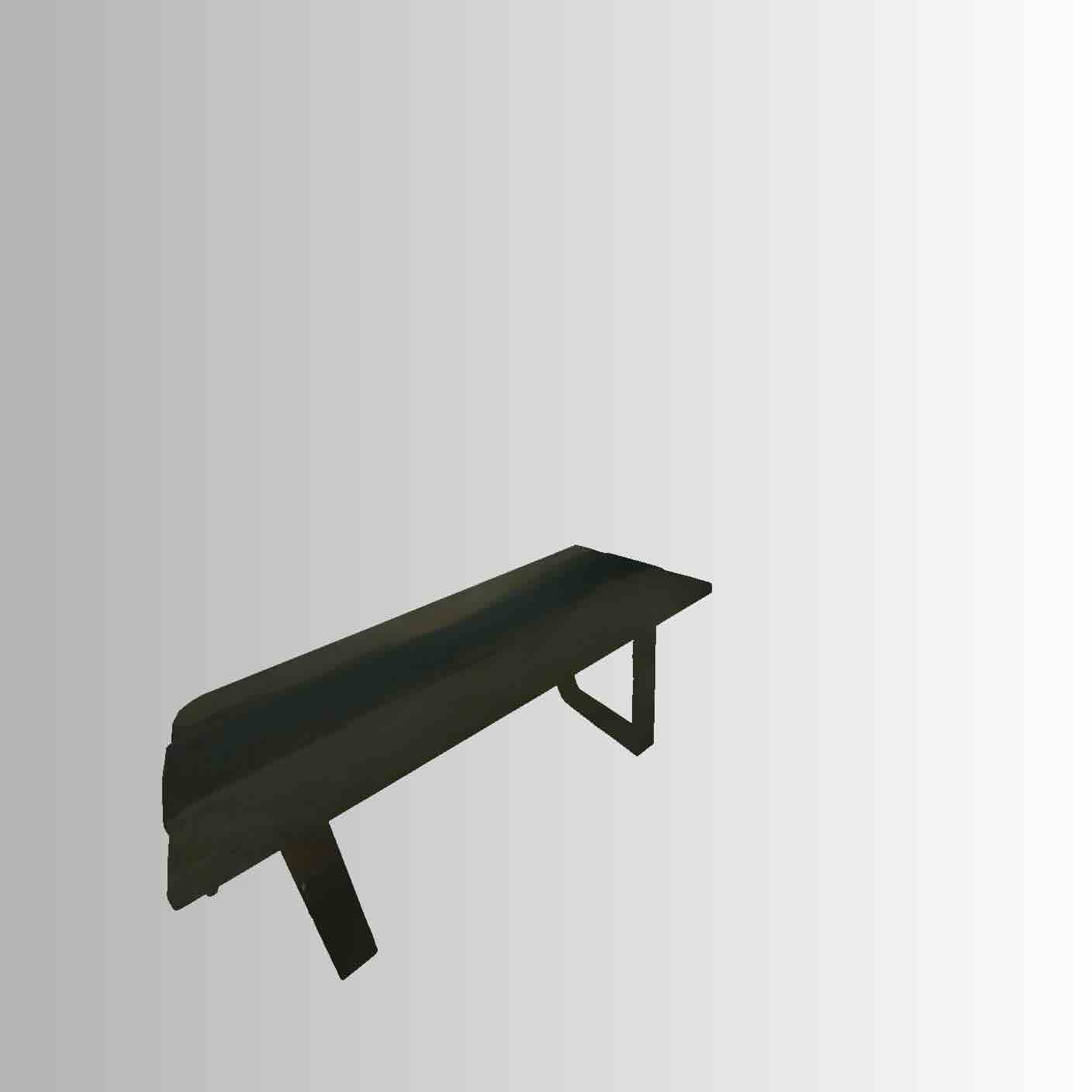} &
        \includegraphics[trim={0.0cm 0.0cm 15cm 20.0cm},clip,width=0.12\textwidth]{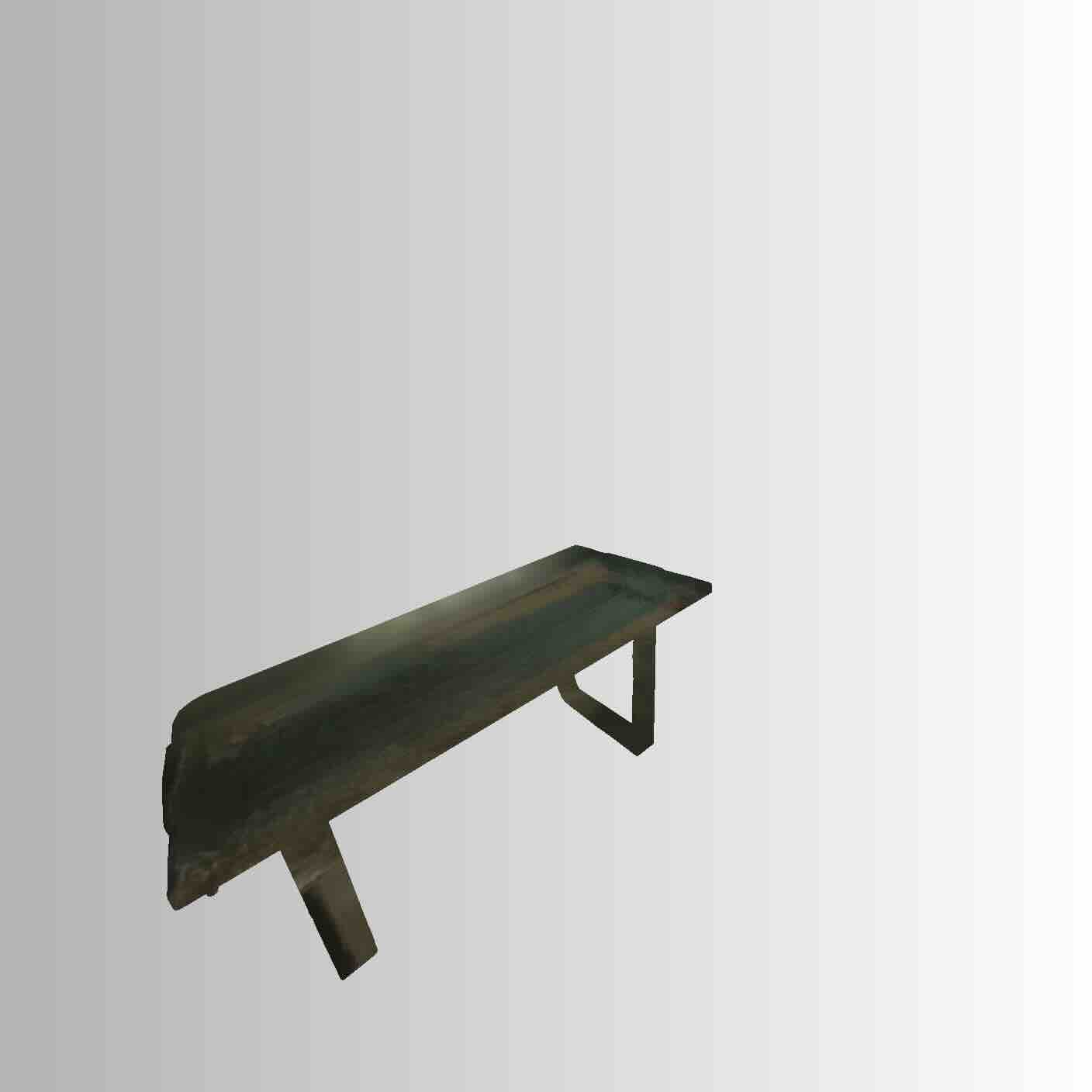} &
        \includegraphics[trim={0.0cm 0.0cm 15cm 20.0cm},clip,width=0.12\textwidth]{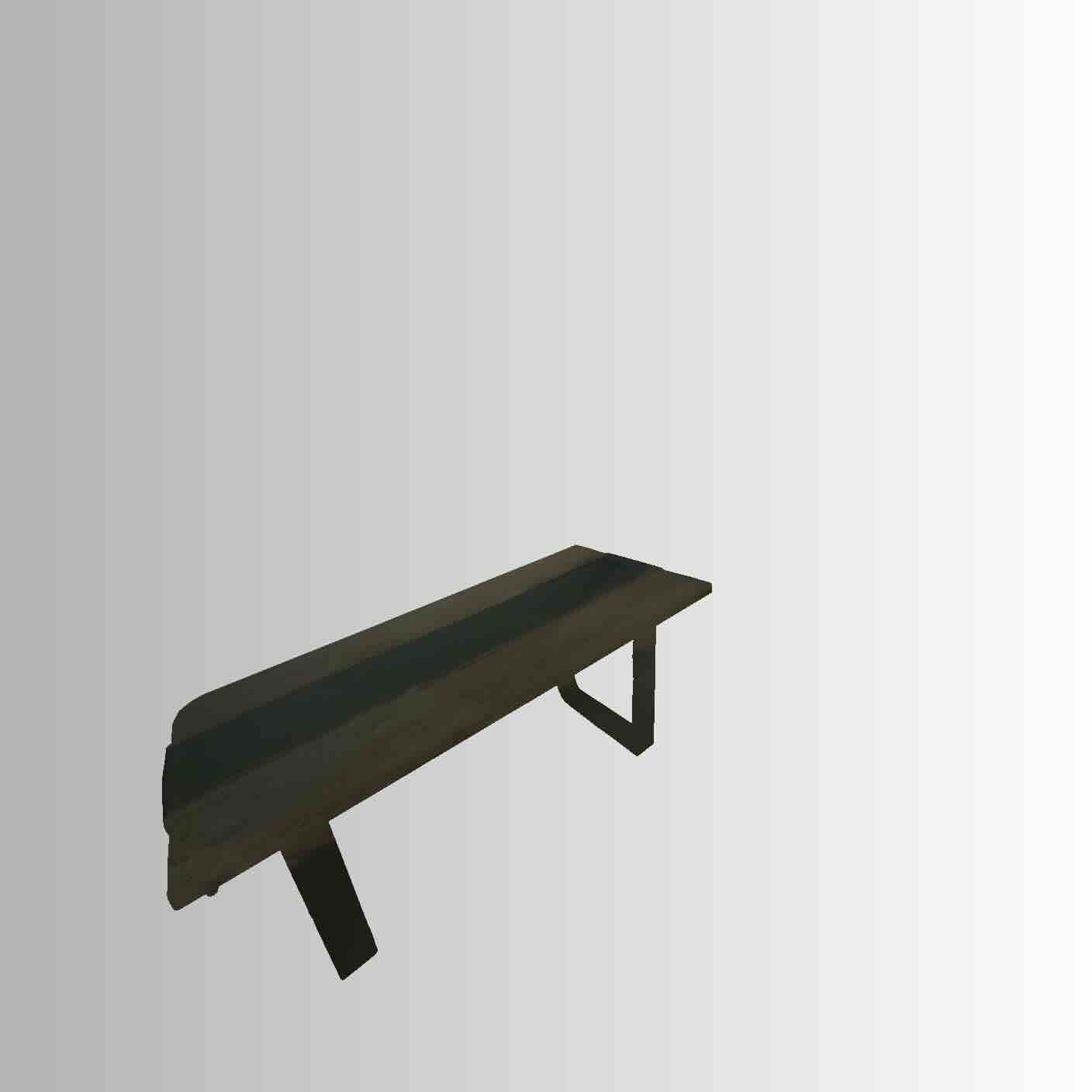} &
        \includegraphics[trim={0.0cm 0.0cm 15cm 20.0cm},clip,width=0.12\textwidth]{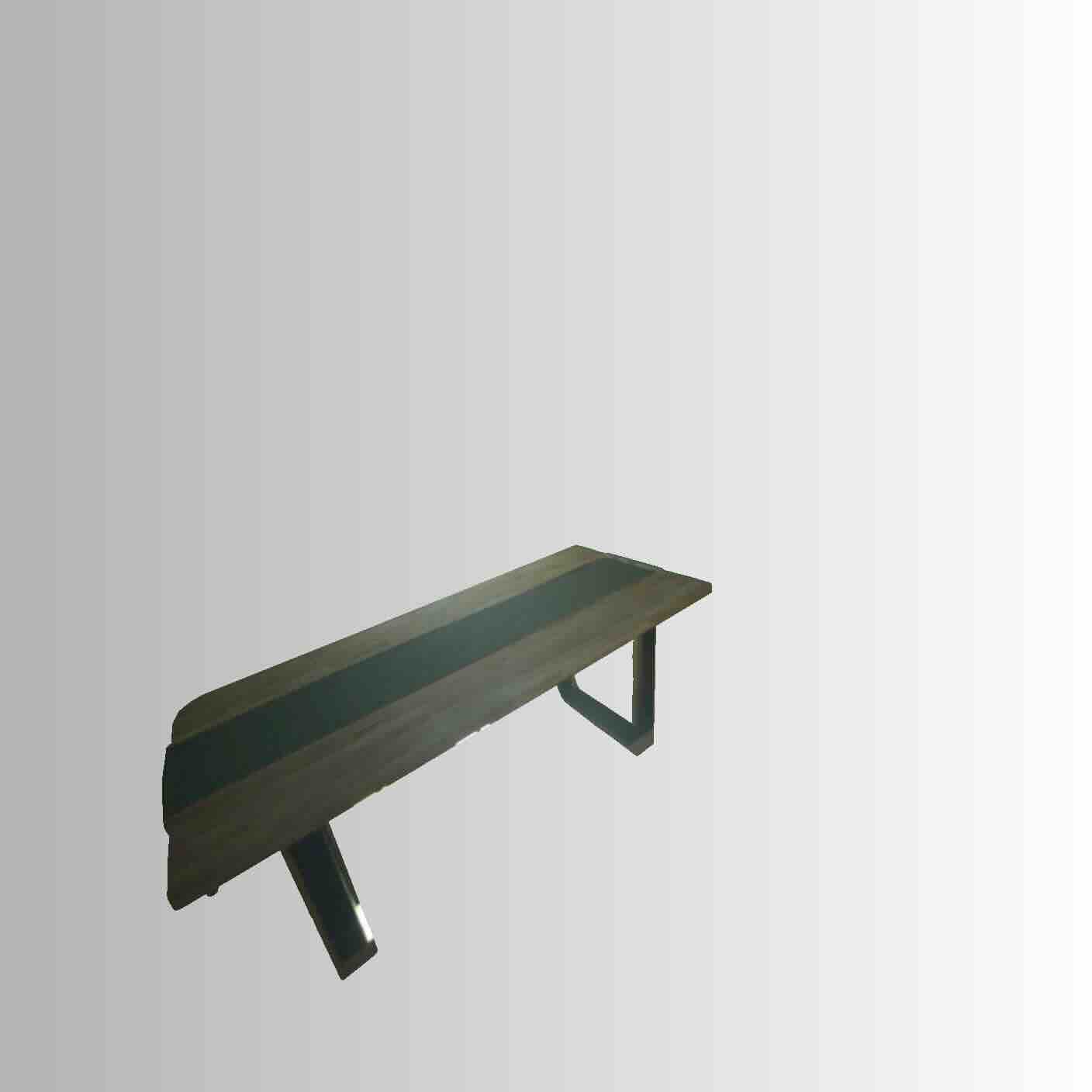}
        \\
        \includegraphics[trim={10.0cm 0.0cm 5.0cm 18cm},clip,width=0.12\textwidth]{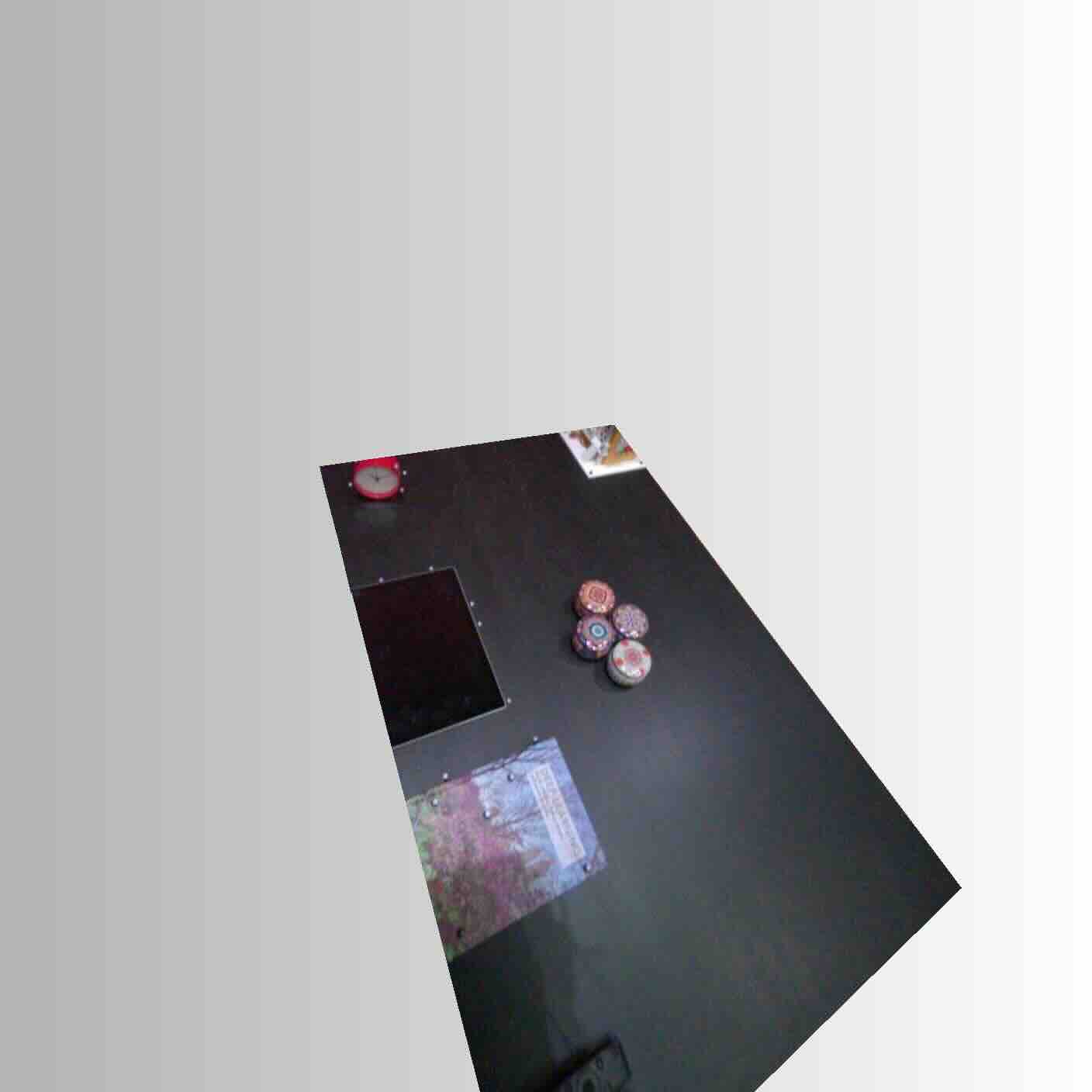} 
        &
        \includegraphics[trim={2.5cm 2.5cm 17.5cm 20cm},clip,width=0.12\textwidth]{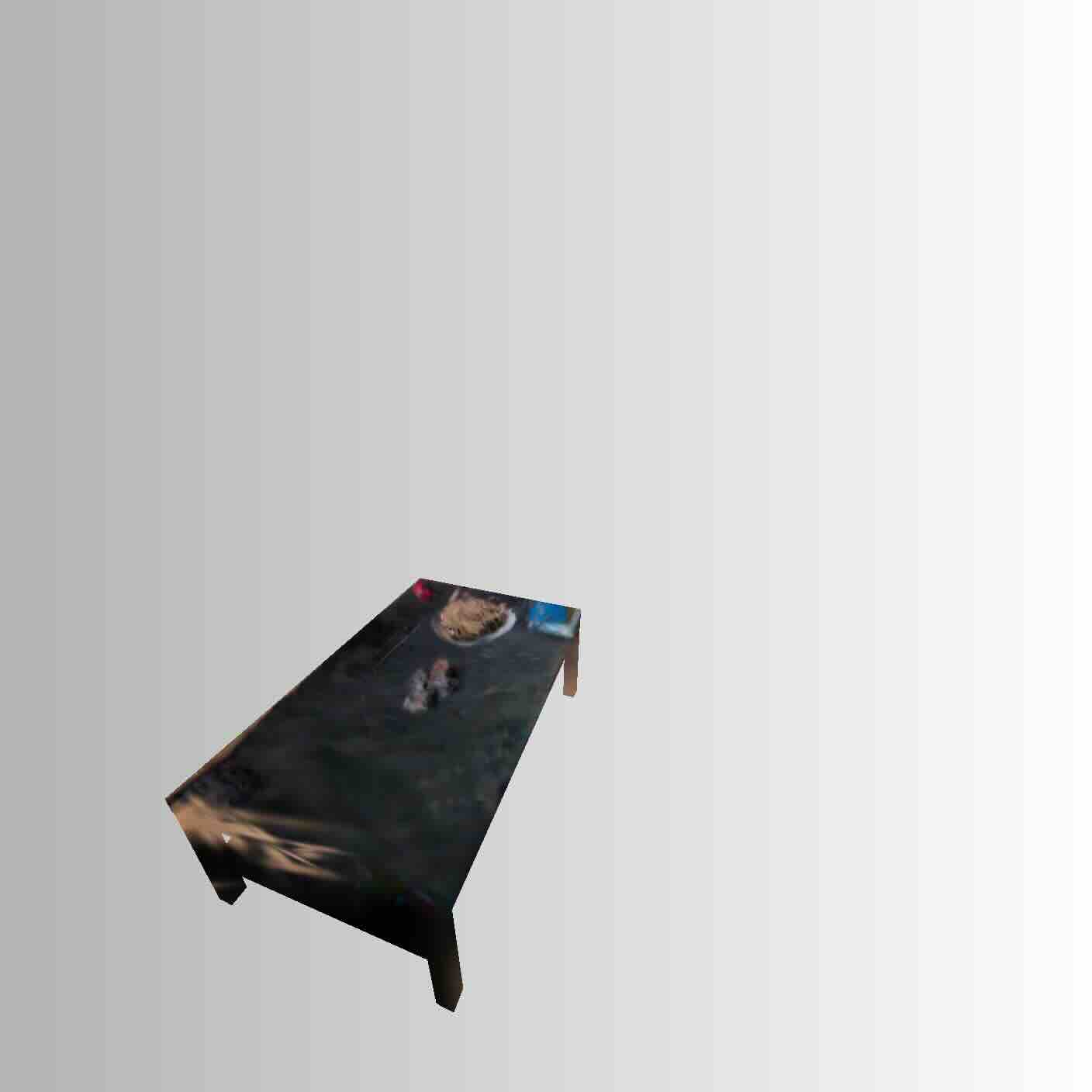} &
        \includegraphics[trim={2.5cm 2.5cm 17.5cm 20cm},clip,width=0.12\textwidth]{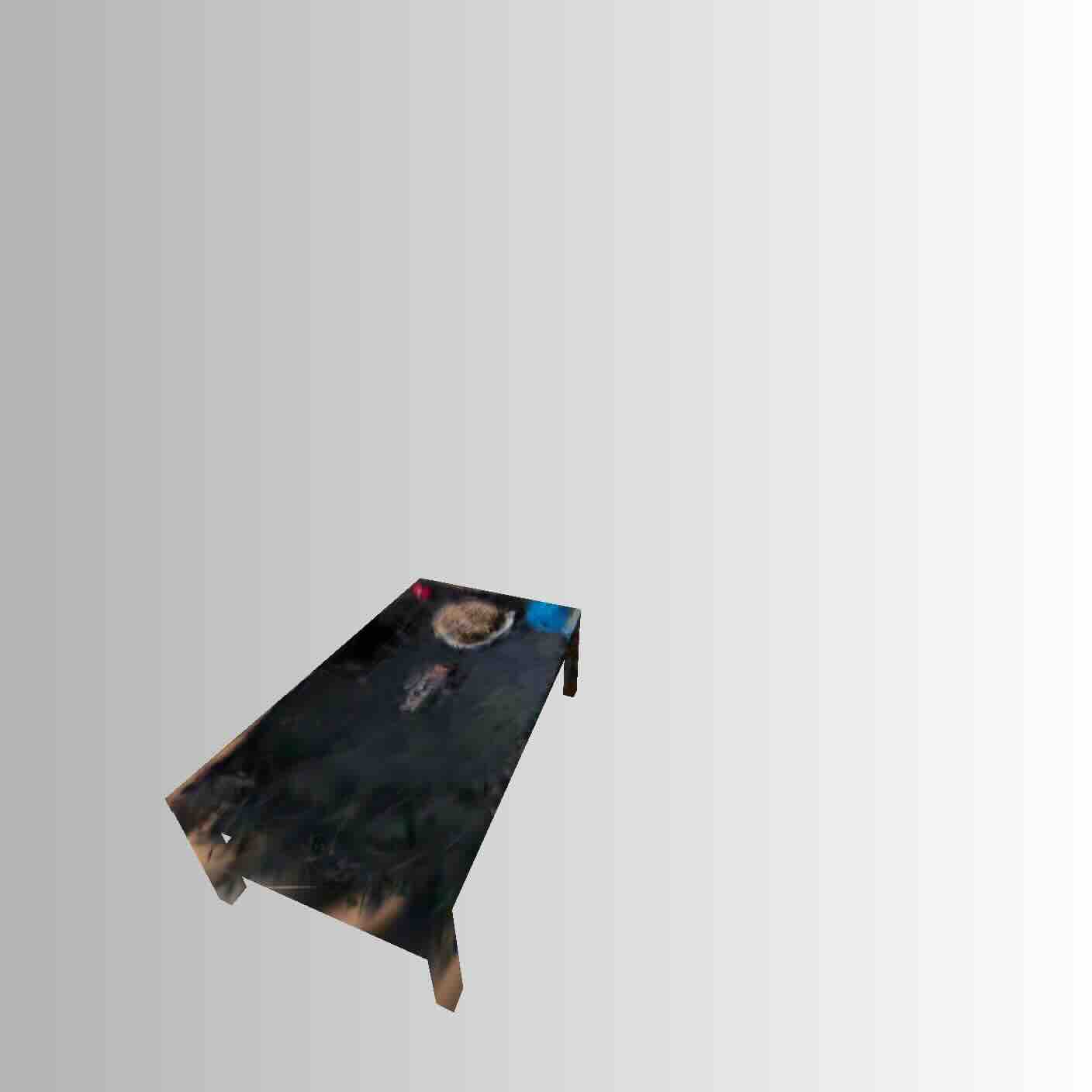} &
        \includegraphics[trim={2.5cm 2.5cm 17.5cm 20cm},clip,width=0.12\textwidth]{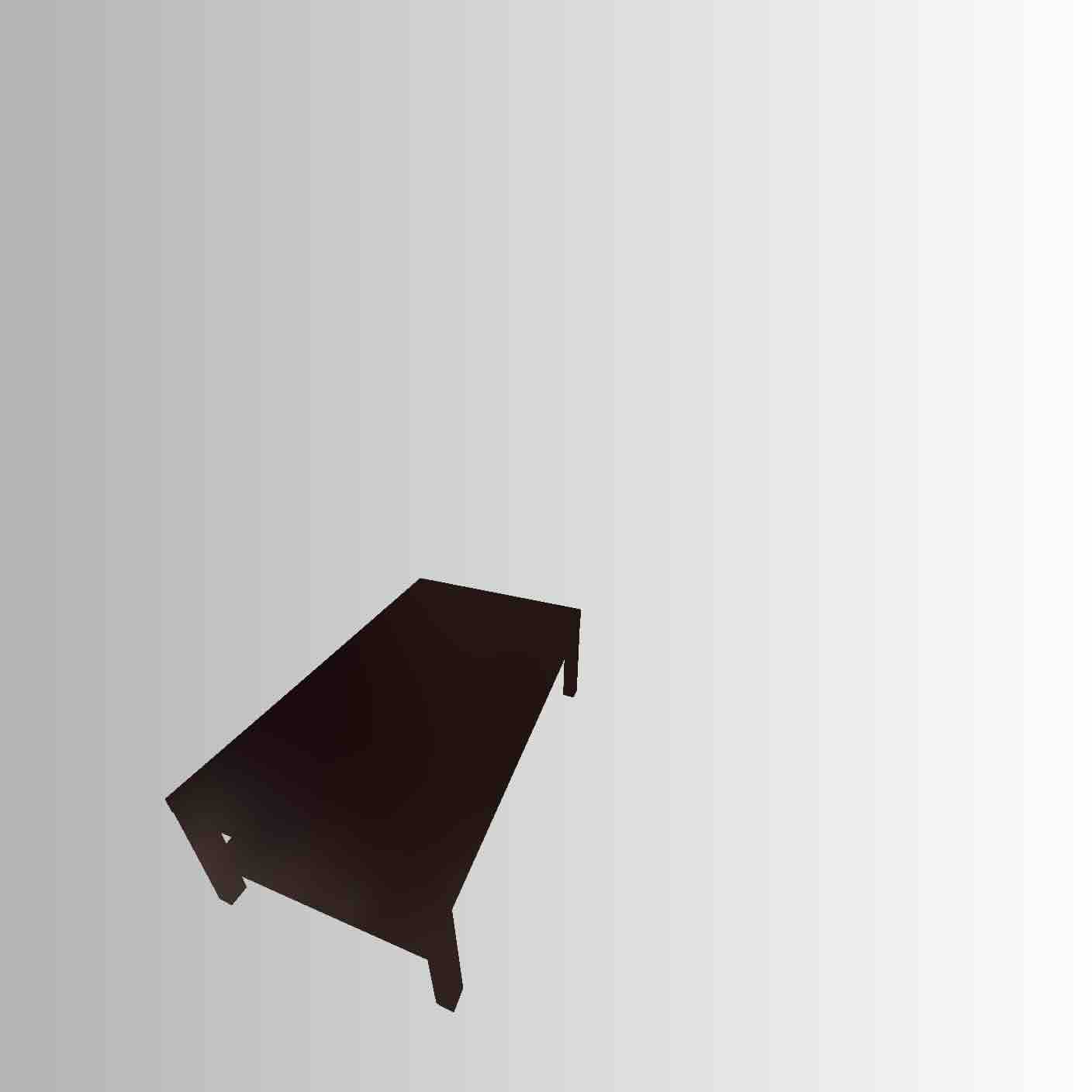} &
        \includegraphics[trim={2.5cm 2.5cm 17.5cm 20cm},clip,width=0.12\textwidth]{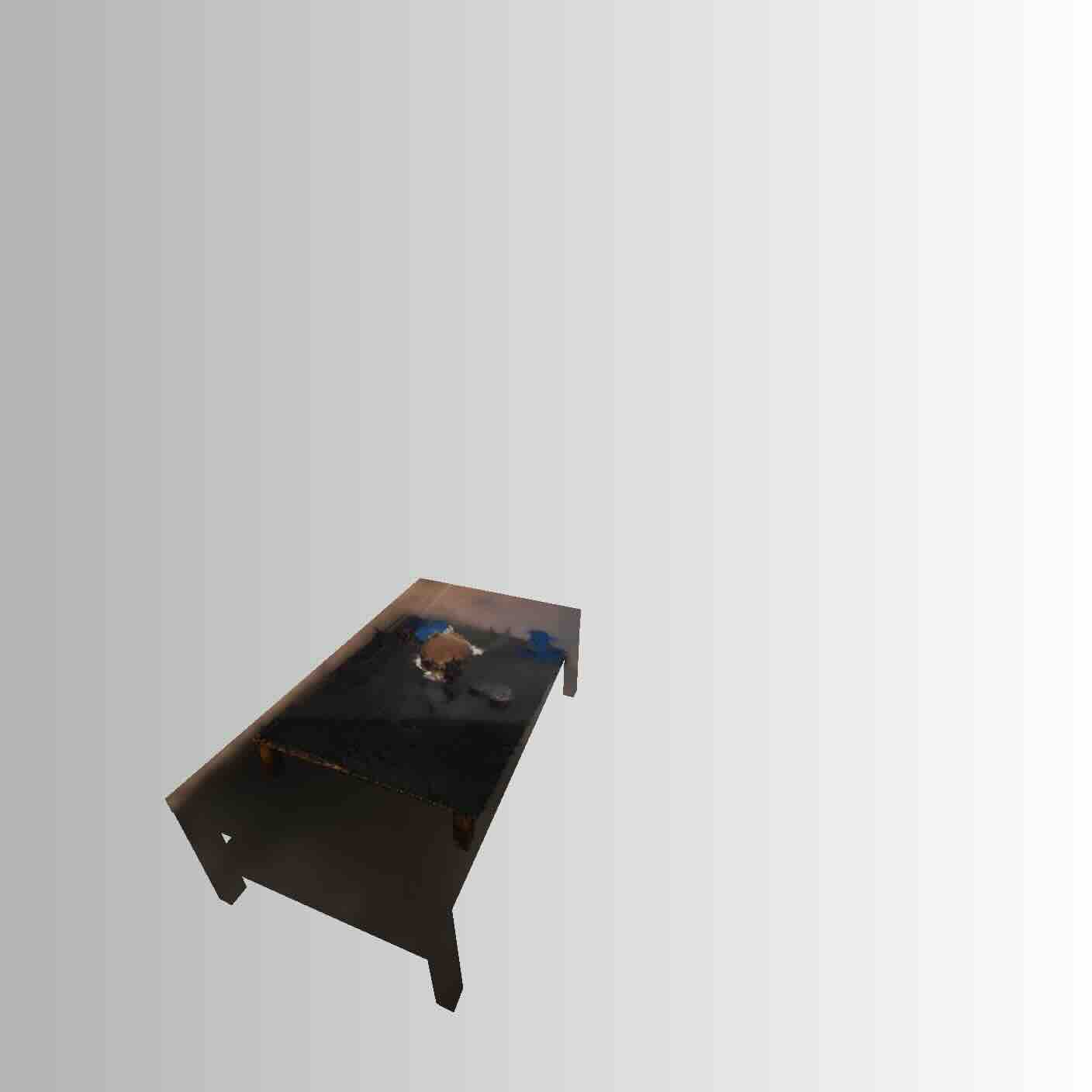} &
        \includegraphics[trim={2.5cm 2.5cm 17.5cm 20cm},clip,width=0.12\textwidth]{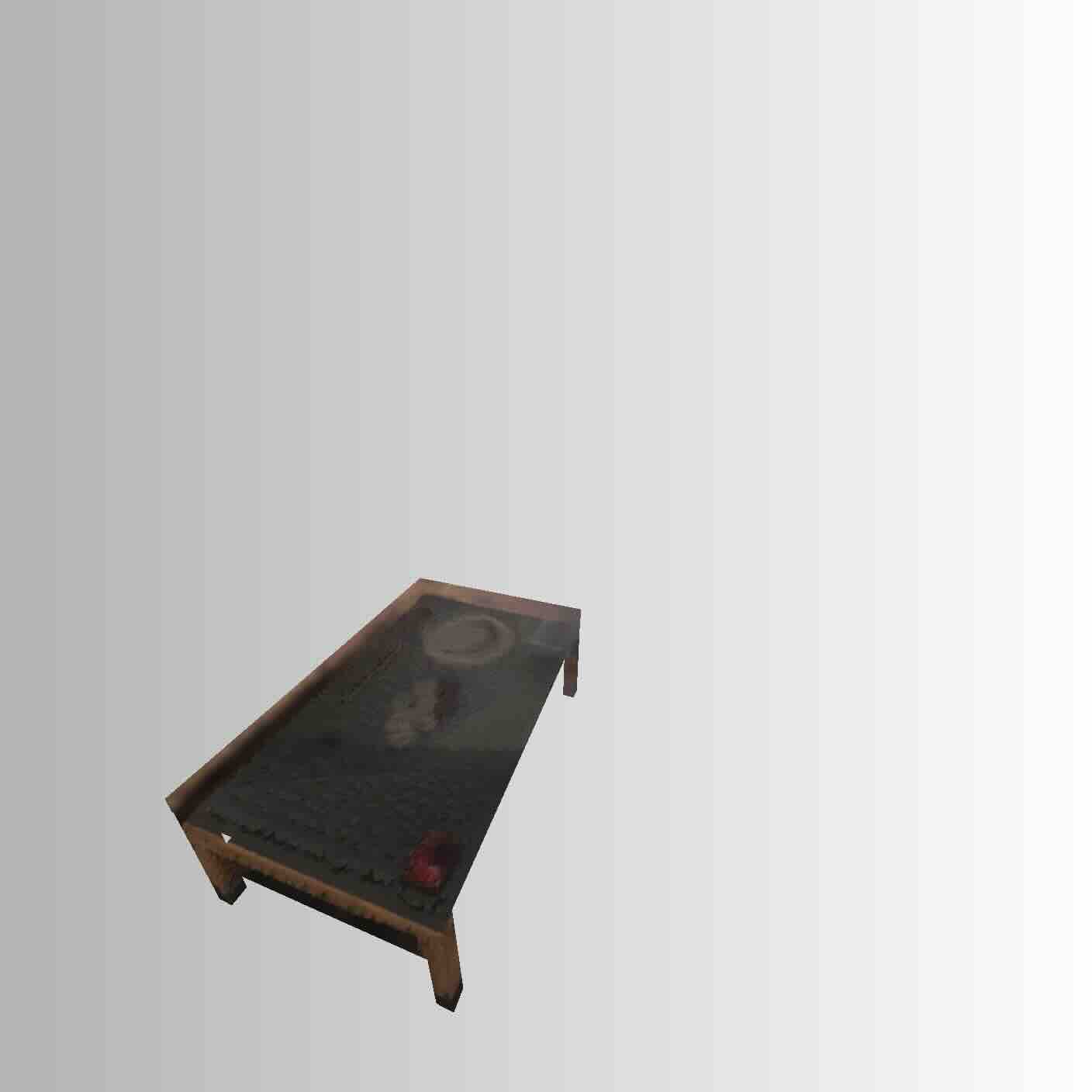} &
        \includegraphics[trim={2.5cm 2.5cm 17.5cm 20cm},clip,width=0.12\textwidth]{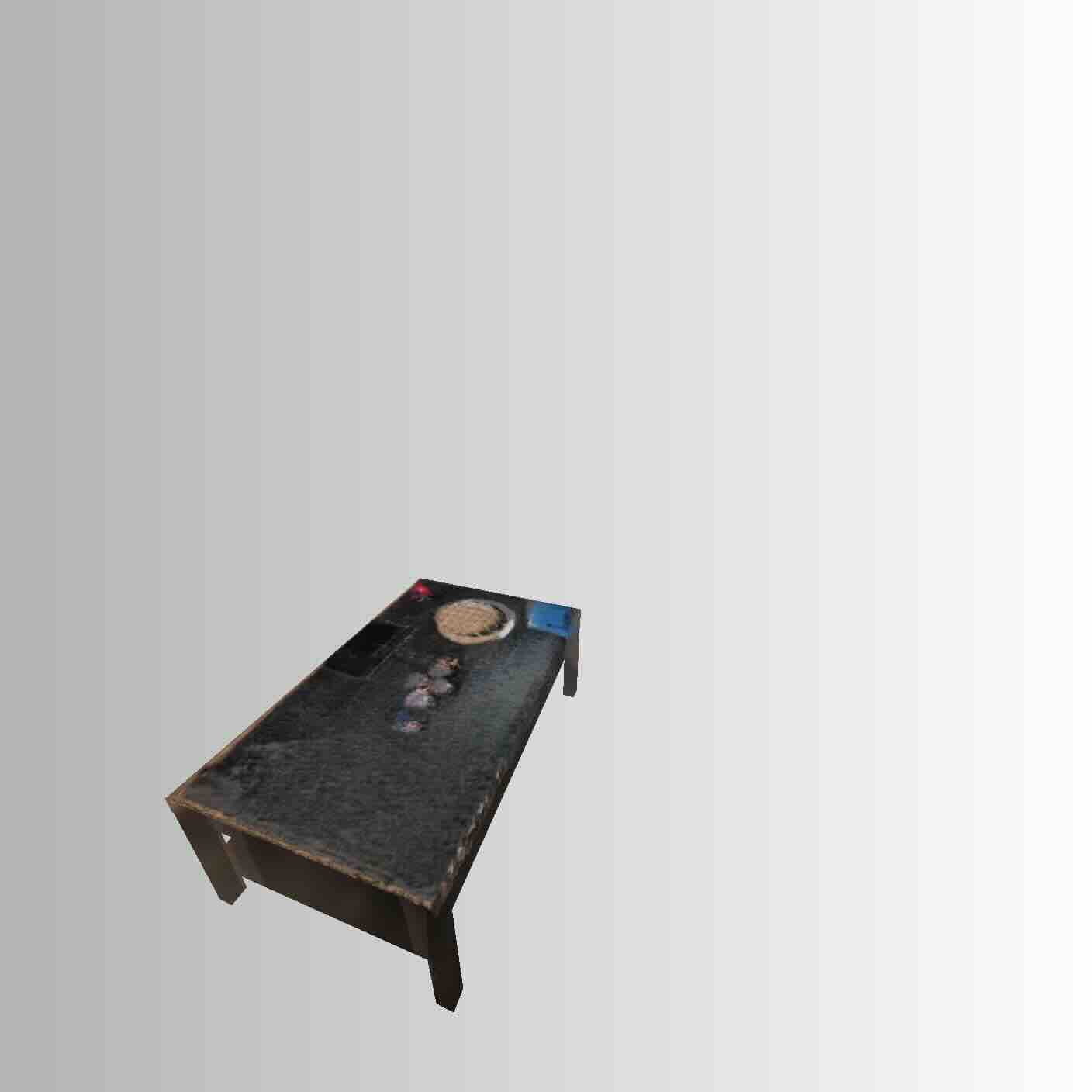} &
        \includegraphics[trim={2.5cm 2.5cm 17.5cm 20cm},clip,width=0.12\textwidth]{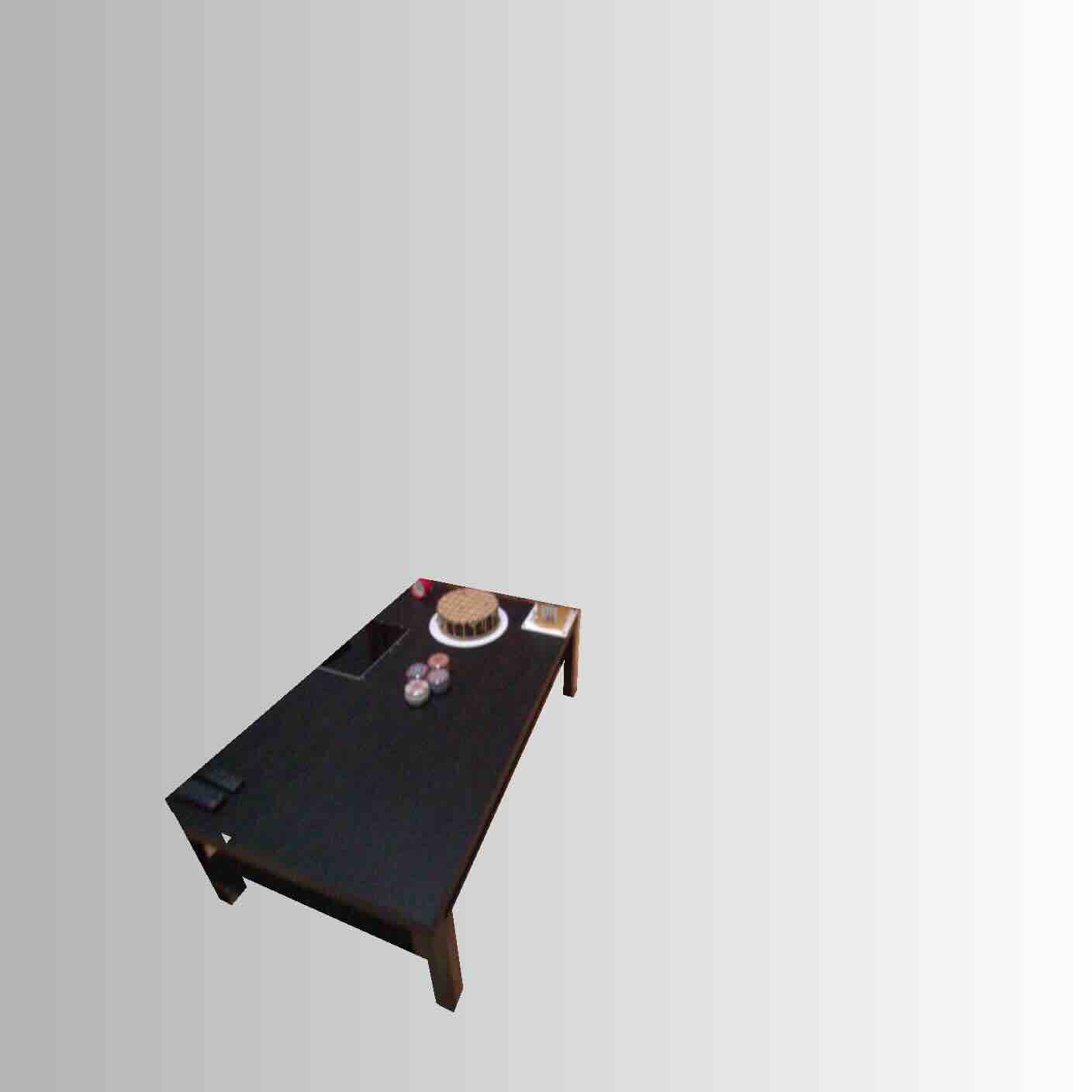}

        \\
                \scriptsize Input Images & 
\shortstack[c]{\scriptsize 3DGS-D \\ \scriptsize~\cite{kerbl20233d}} & 
\shortstack[c]{\scriptsize AGS-Mesh \\ \scriptsize~\cite{ren2025ags}} & 
\shortstack[c]{\scriptsize OM-GSD$^{\ast}$ \\ \scriptsize~\cite{lu2025orientation}} & 
\shortstack[c]{\scriptsize RVG-GSD$^{\ast}$ \\ \scriptsize~\cite{chang2025reconviagen}} & 
\shortstack[c]{\scriptsize SAM3D-GSD$^{\ast}$ \\ \scriptsize~\cite{chen2025sam}} & 
\scriptsize Ours & 
\scriptsize GT \\
        
        \end{tabular} }
      \caption{\textbf{Qualitative Rendering Comparison on Challenging Real-World Samples.} The first two samples (top row) originate from the ScanNet++ dataset~\cite{yeshwanth2023scannet++}, while the remaining two (bottom row) are drawn from ShapeR Evaluation Dataset~\cite{siddiqui2026shaper}. Our method achieves superior appearance reconstruction with high-fidelity texture consistency. In contrast, baselines exhibit significant observational gaps and produce disjointed textures that fail to maintain surface coherence across the object.} 
    \label{fig:main_rendering_realworlddata}
    \vspace{-2mm}
\end{figure*}

As summarized in Tab.~\ref{tab:main_rendering_scannetpp}, our method outperforms competing approaches in NVS tasks characterized by severe occlusions and highly sparse observations across both datasets. This advantage is further illustrated in Fig.~\ref{fig:main_rendering_realworlddata}: while traditional 3DGS suffers from significant artifacts in unobserved regions and diffusion-based methods struggle with texture consistency, \OURS{} synthesizes spatially coherent and photorealistic results.

The advantage of our approach is emphasized when compared to generative baselines, which struggle with scale and translation ambiguities. Our dual-space guided method enforces stringent 3D constraints by anchoring the flow trajectory to physical measurements at inference time. This training-free, zero-shot generative strategy ensures precise scale alignment between the generative prior and the target scene. Our approach effectively recovers occluded objects by leveraging these priors, while maintaining the characteristic appearance of 3DGS, even under extreme viewpoint scarcity and severe occlusions.

\subsection{Ablation Analysis}

\paragraph{\textbf{Dual-space Flow Steering.}}
We investigate the impact of different guidance mechanisms within our dual-space guidance pipeline on the 3D-FRONT dataset. Our results in Tab.~\ref{tab:all_ablations}~(a) show that latent-space constraints significantly influence the overall geometric completeness. Conversely, explicit-space constraints act as a fine-grained geometric polisher, further rectifying local surface details and aligning the generated output with sparse physical measurements. The synergy of both spaces yields the most consistent and accurate reconstructions.

\paragraph{\textbf{Robustness to View Sparsity.}}
To evaluate the stability of our method under varying degrees of observational incompleteness, we conduct experiments with 3, 5, and 10 input views on the 3D-FRONT~\cite{fu20213d} dataset at Tab.~\ref{tab:all_ablations}~(b). Even under 3 views, which is extremely sparse, our framework maintains high structural integrity by relying on the robust generative prior.

\paragraph{\textbf{Efficacy of 3DGS Refinement.}}
We ablate the refinement stage by comparing novel view synthesis results under two configurations: (i) raw generative output directly from the flow-matching trajectory (without 3DGS refinement) Sec.~\ref{sect:obj_rec}; and (ii) full radiance refinement using our staged optimization Sec.~\ref{sect:gs_opt}. As summarized in Tab.~\ref{tab:all_ablations}~(c), this step effectively anchors the generative prior to the photorealistic details of the target instance, bridging the domain gap between ``synthetic-looking'' outputs and high-fidelity real-world observations.

\paragraph{\textbf{Evaluation under Geometric and Segmentation Uncertainty.}} To demonstrate the robustness of our method in the absence of object meshes or scans, we evaluate its performance using predicted inputs. Specifically, we leverage the state-of-the-art Depth Anything 3 (DA3)~\cite{lin2025depth} model for depth estimation on ShapeR Evaluation Dataset and SAM 2~\cite{ravi2025sam} for mask
prediction on Scannet++ dataset. Fig.~\ref{fig:all_ablations}~(a)-(b) showcase that our method maintains high-fidelity reconstructions even with estimated geometric observations.

\begin{table}[t]
    \centering
    \begin{minipage}[t]{0.32\linewidth}
        \centering
        \small
        (a) Dual-space Guidance\\
        
        \setlength{\tabcolsep}{2pt}
        \scalebox{0.8}{
        \begin{tabular}{cc|ccc}
            \toprule
            Impl. & Expl. & CD $\downarrow$ & Comp. $\uparrow$ & F-Score $\uparrow$ \\
            \midrule
            \checkmark & & 2.61 & 69.87 & 69.57 \\
             & \checkmark & 1.88 & 75.37 &  39.17 \\
            \checkmark & \checkmark & \textbf{1.59} & \textbf{77.80} & \textbf{70.20} \\
            \bottomrule
        \end{tabular}
        }
    \end{minipage}
    \begin{minipage}[t]{0.32\linewidth}
        \centering
        \small
        (b) Sparsity Robustness\\

        \setlength{\tabcolsep}{2pt}
        \scalebox{0.8}{
        \begin{tabular}{c|ccc}
            \toprule
            Views & CD $\downarrow$ & Comp. $\uparrow$ & F-Score $\uparrow$ \\
            \midrule
            3 & 2.38 & 67.79 & 65.98 \\
            5 & 1.81 & 73.73 & 70.00 \\
            10 & \textbf{1.59} & \textbf{77.80} & \textbf{70.20} \\
            \bottomrule
        \end{tabular}
        }
    \end{minipage}
    \begin{minipage}[t]{0.3\linewidth}
        \centering
        \small
        (c) 3DGS Refinement\\
        \vspace{2mm}

    \setlength{\tabcolsep}{2pt}
    \scalebox{0.8}{
        \begin{tabular}{l|ccc}
            \toprule
            Strategy & PSNR $\uparrow$ & SSIM $\uparrow$ & LPIPS $\downarrow$ \\
            \midrule
            w/o Refine. & 22.32 & 0.937 & 7.66 \\
            w/ Refine. & \textbf{25.76} & \textbf{0.956} & \textbf{6.78} \\
            \bottomrule
        \end{tabular}
        }
    \end{minipage}
    \vspace{1mm}
     \caption{\textbf{Ablation Studies.} (a) Analysis of dual-space guidance on 3D-FRONT (CD $\times 10^{2}$); (b) Reconstruction performance under sparse input views on 3D-FRONT (CD $\times 10^{2}$); (c) Effect of 3DGS refinement on image quality on ScanNet++ (LPIPS $\times 10^2$).}
     \vspace{-5mm}
    \label{tab:all_ablations}
\end{table}

\paragraph{\textbf{Generalization to Alternative Latent Priors.}}
To demonstrate the versatility of our method, we integrate the proposed dual-space guidance method into UniLat3D~\cite{wu2025unilat3d}, a one-stage 3D generative framework. This experiment evaluates our mechanism's capability to steer alternative latent manifolds toward physical reality. As illustrated in Fig.~\ref{fig:all_ablations}~(c), our guidance enforces more precise alignments between the generated 3DGS and sparse observations compared with vanilla UniLat3D. This confirms that our dual-space steering serves as a backbone-agnostic scheme, capable of precisely calibrating diverse latent-based 3D diffusion models to match sparse physical inputs.

\begin{table}[b]
    \centering
    \hspace{-5mm}
    \vspace{-1mm}
    \begin{minipage}[t]{0.32\linewidth}
        \centering

        \label{fig:pred_depth}
        \setlength{\tabcolsep}{2pt}
        \scalebox{0.9}{
        \begin{tabular}{ccc}
        \includegraphics[trim={10.0cm 2.0cm 15.0cm 2.5cm},clip,width=0.32\linewidth]{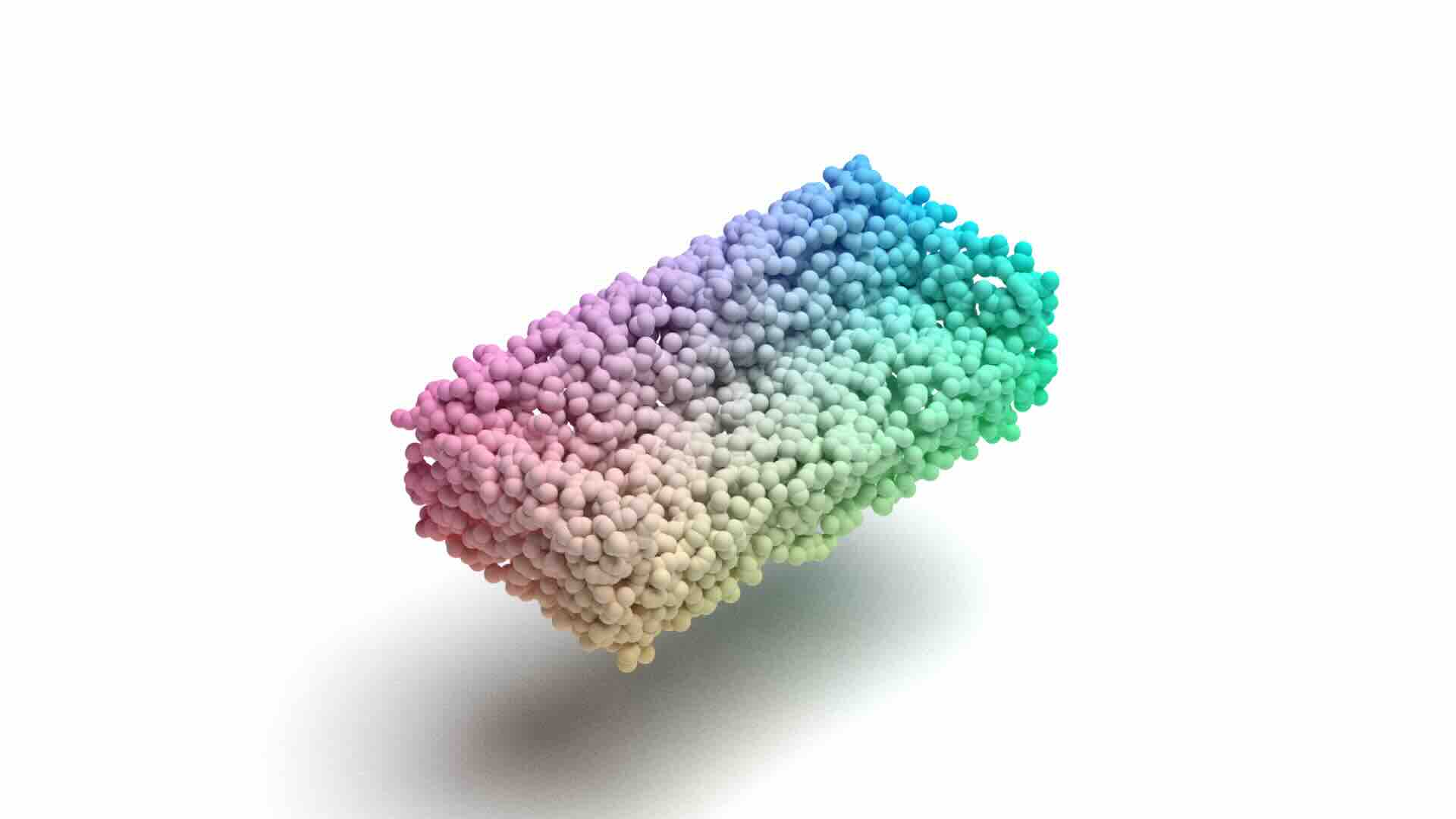} &
        \includegraphics[trim={10.0cm 2.0cm 15.0cm 2.5cm},clip,width=0.32\linewidth]{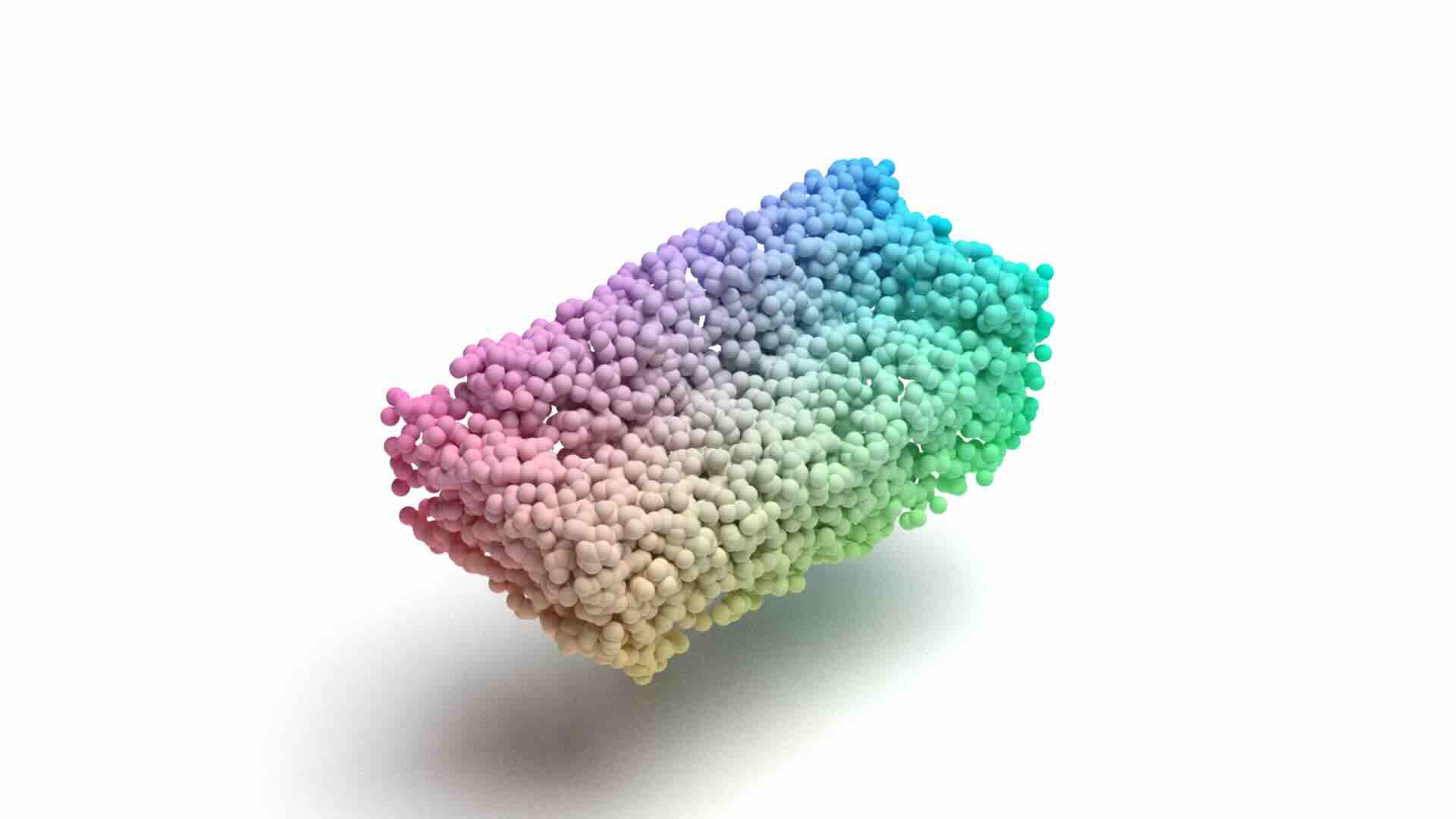} &
        \includegraphics[trim={10.0cm 2.0cm 15.0cm 2.5cm},clip,width=0.32\linewidth]{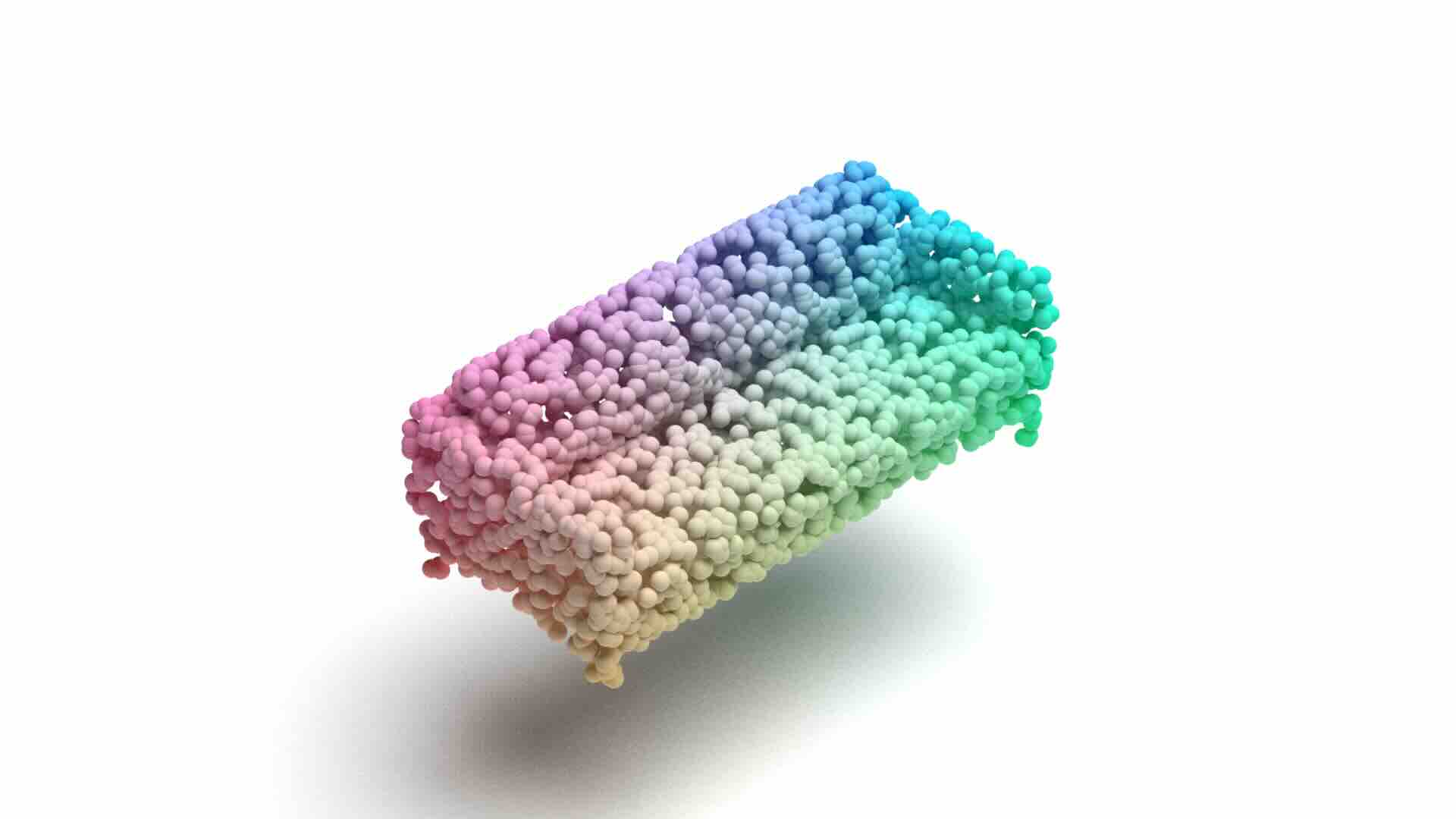} \\
        
        \includegraphics[trim={10.0cm 1.0cm 10.0cm 1cm},clip,width=0.32\linewidth]{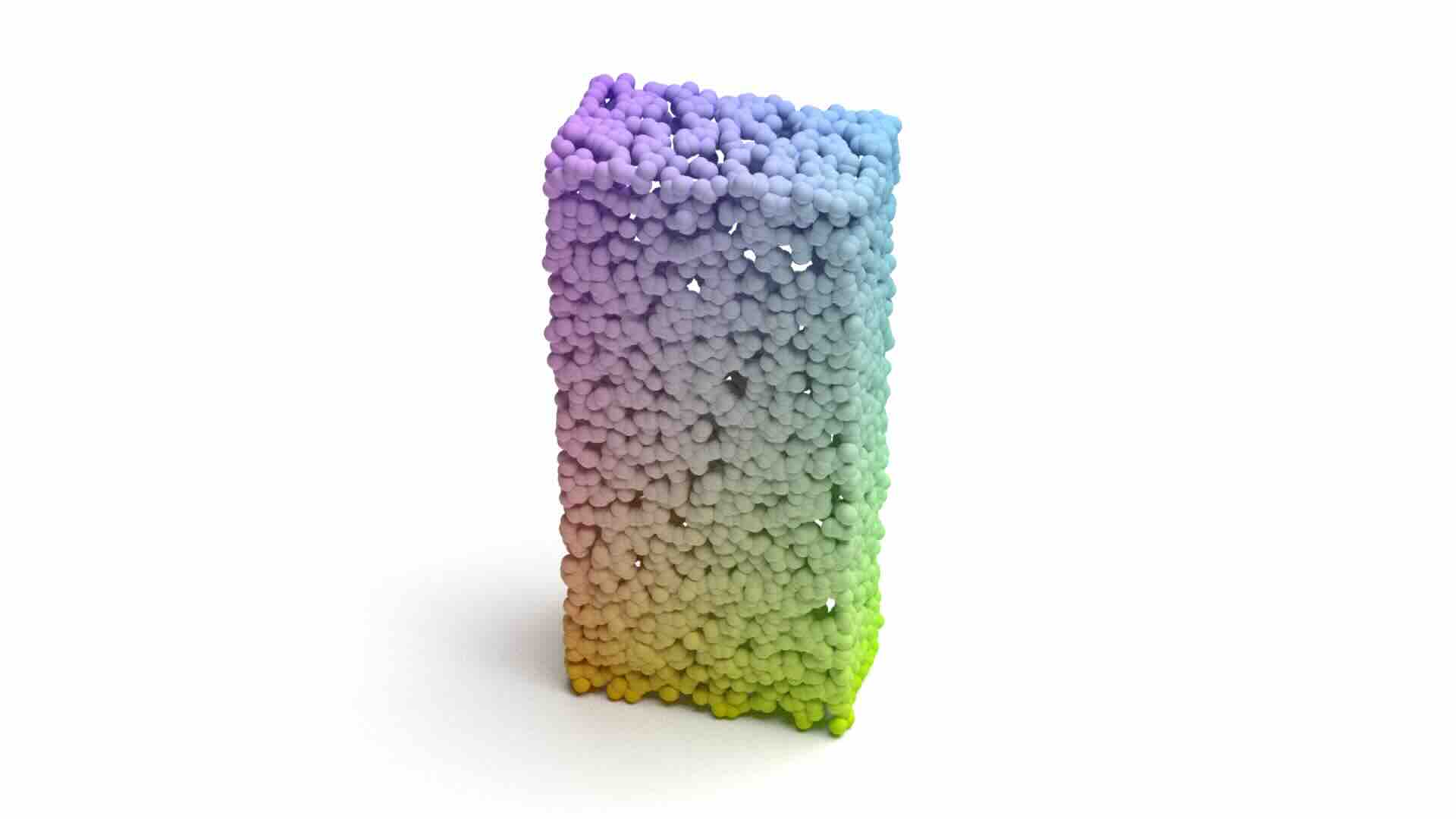} &
        \includegraphics[trim={10.0cm 1.0cm 10.0cm 1cm},clip,width=0.32\linewidth]{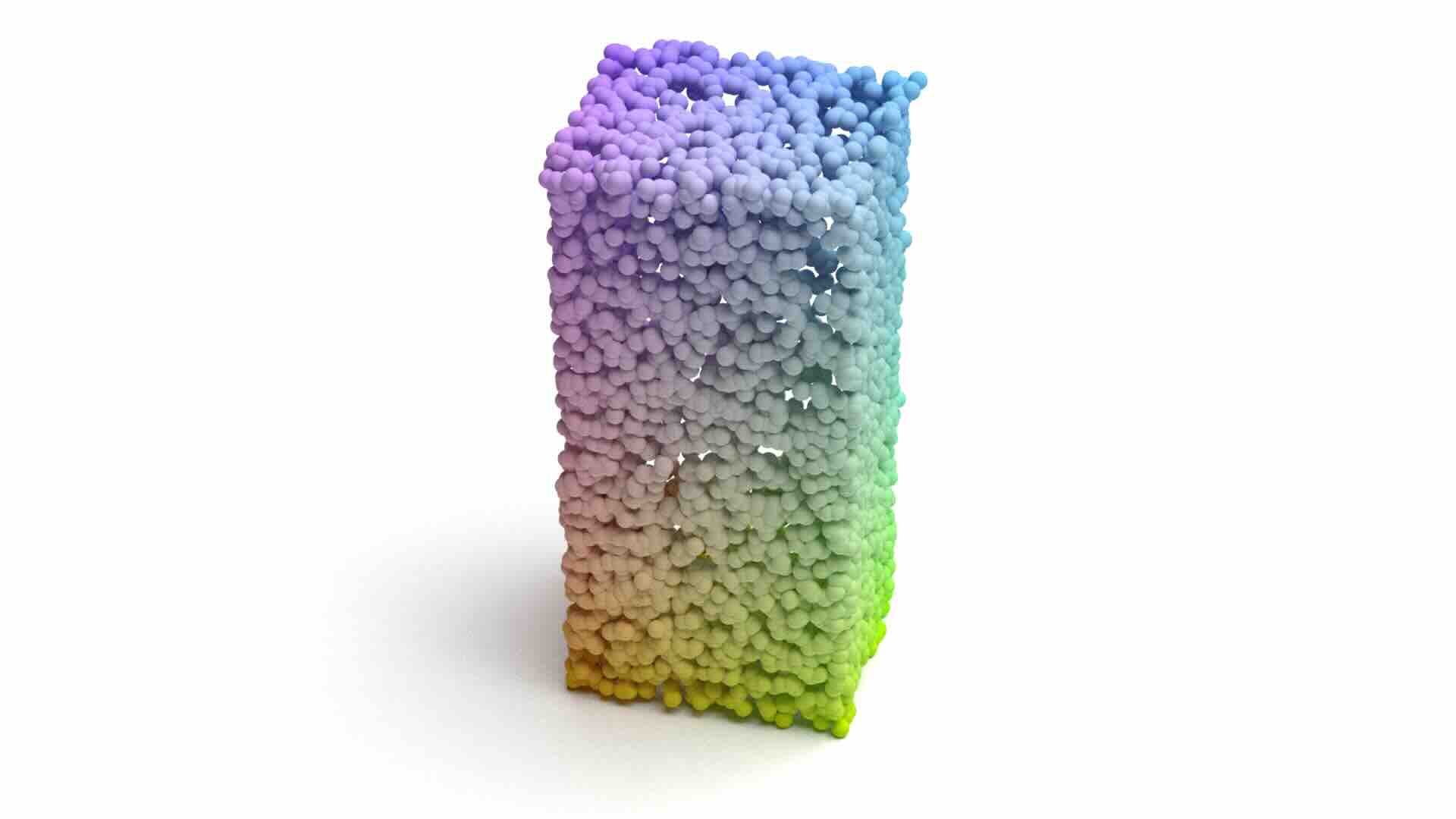} &
        \includegraphics[trim={10.0cm 1.0cm 10.0cm 1cm},clip,width=0.32\linewidth]{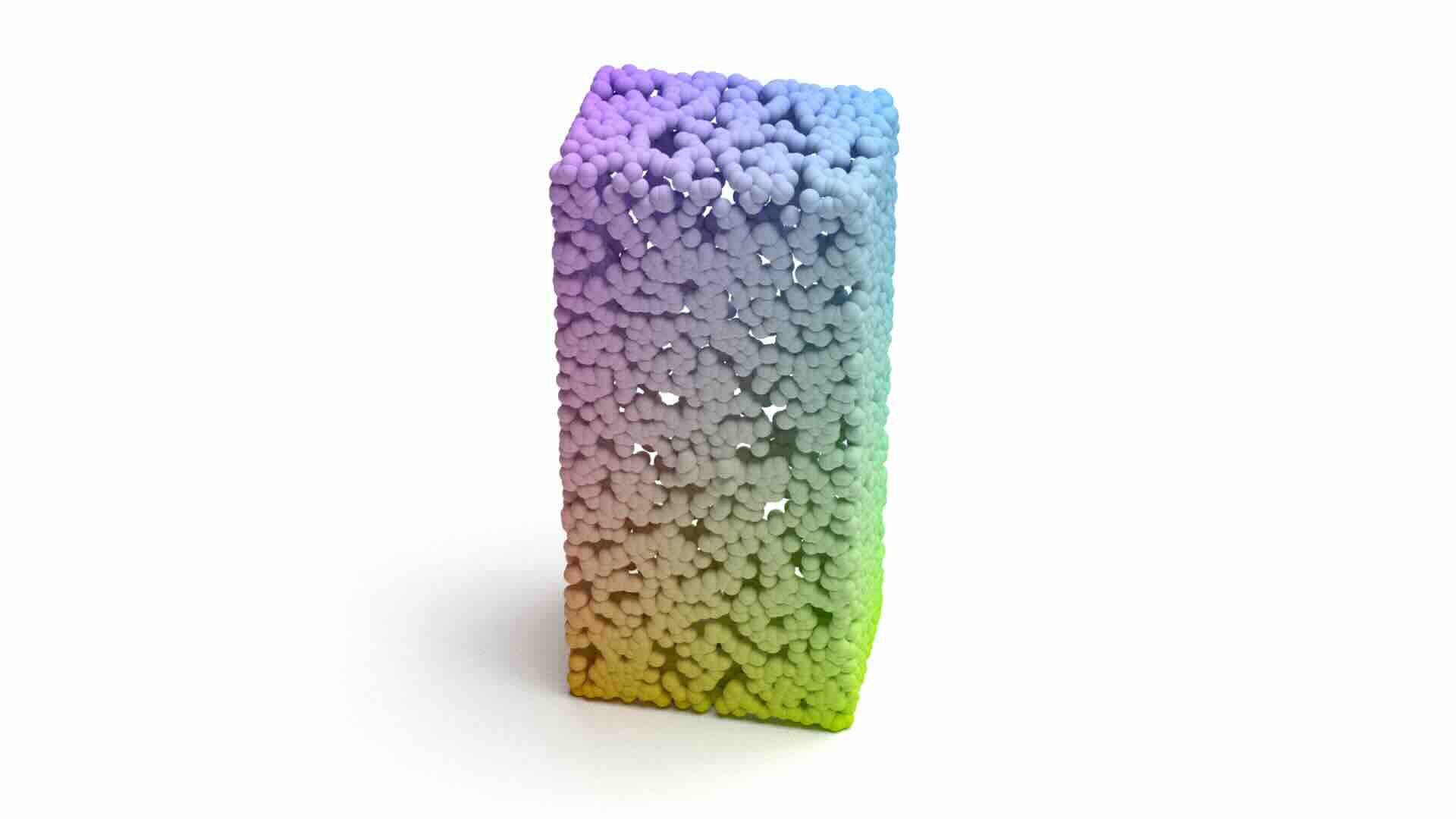} \\
        \scriptsize Ours & \scriptsize DA3 Depth & \scriptsize GT \\
    \end{tabular}
        } \\
                \small
        (a) Geometric Robustness\\
    \end{minipage}
    \hspace{-2mm}
    \begin{minipage}[t]{0.32\linewidth}
        \centering
        
        \label{fig:mask_depth}
        \setlength{\tabcolsep}{2pt}
        \scalebox{0.9}{
        \begin{tabular}{ccc}
        \includegraphics[trim={10.0cm 2.0cm 15.0cm 2.5cm},clip,width=0.32\linewidth]{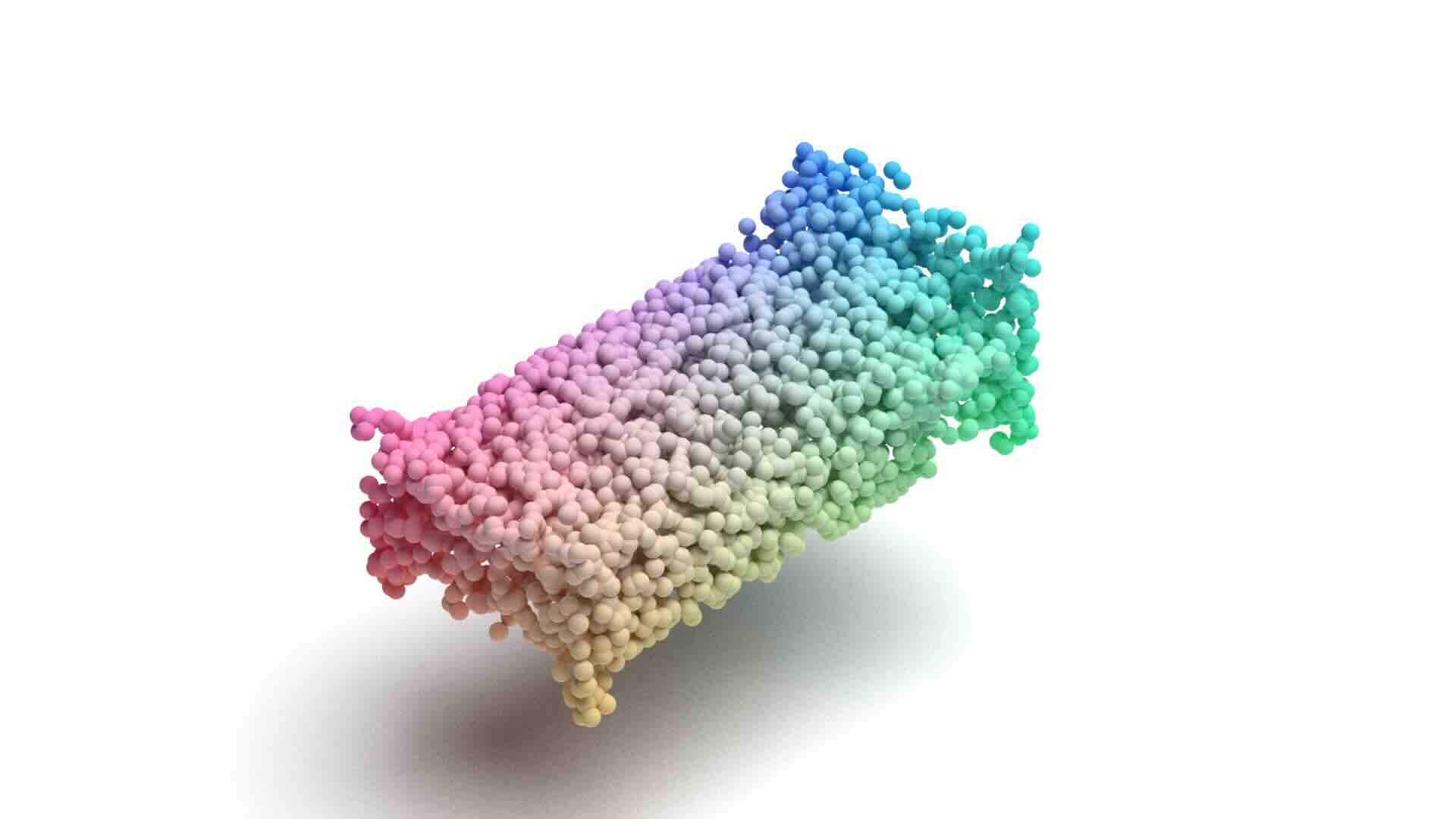} &
        \includegraphics[trim={10.0cm 2.0cm 15.0cm 2.5cm},clip,width=0.32\linewidth]{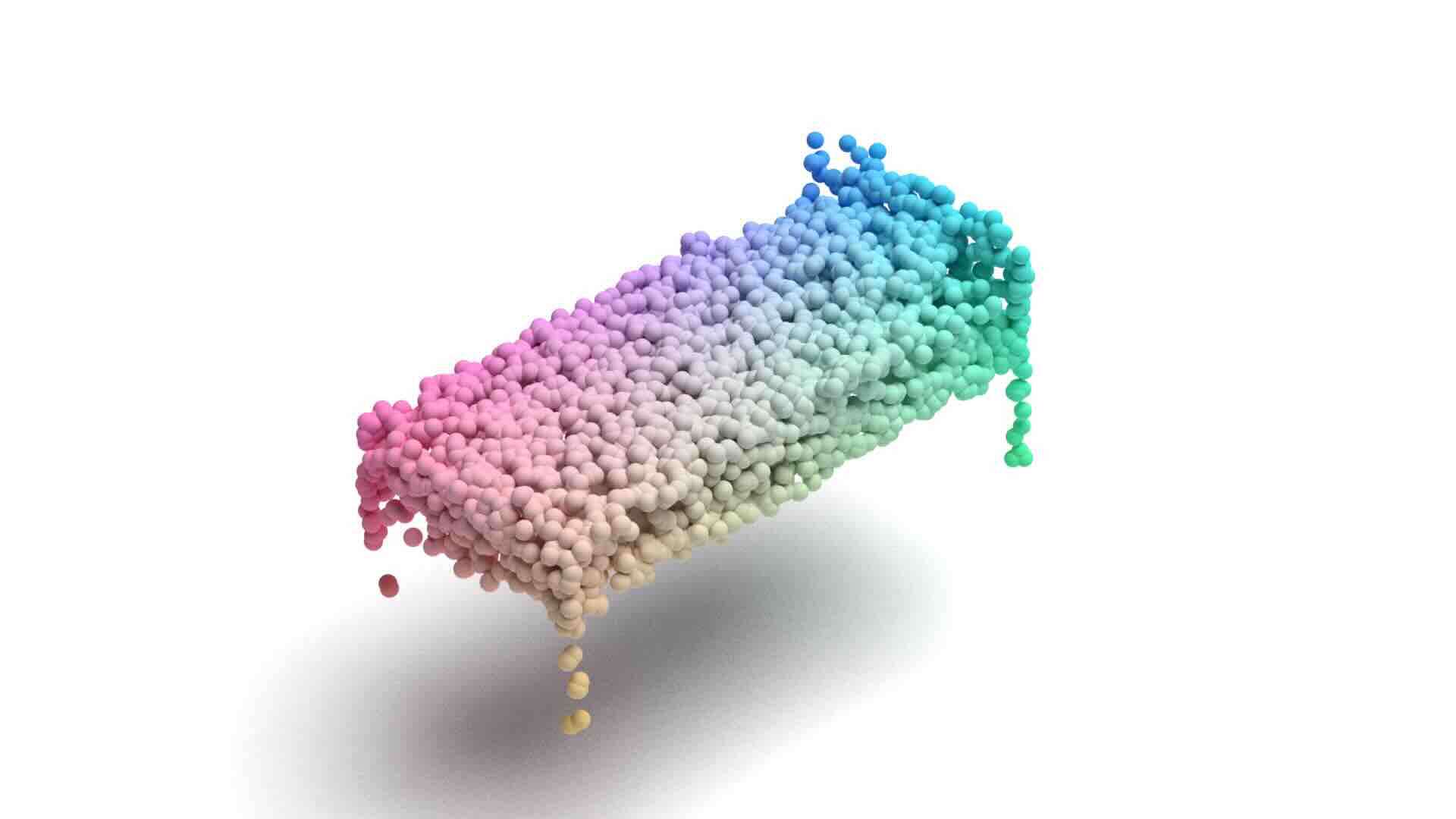} &
        \includegraphics[trim={10.0cm 2.0cm 15.0cm 2.5cm},clip,width=0.32\linewidth]{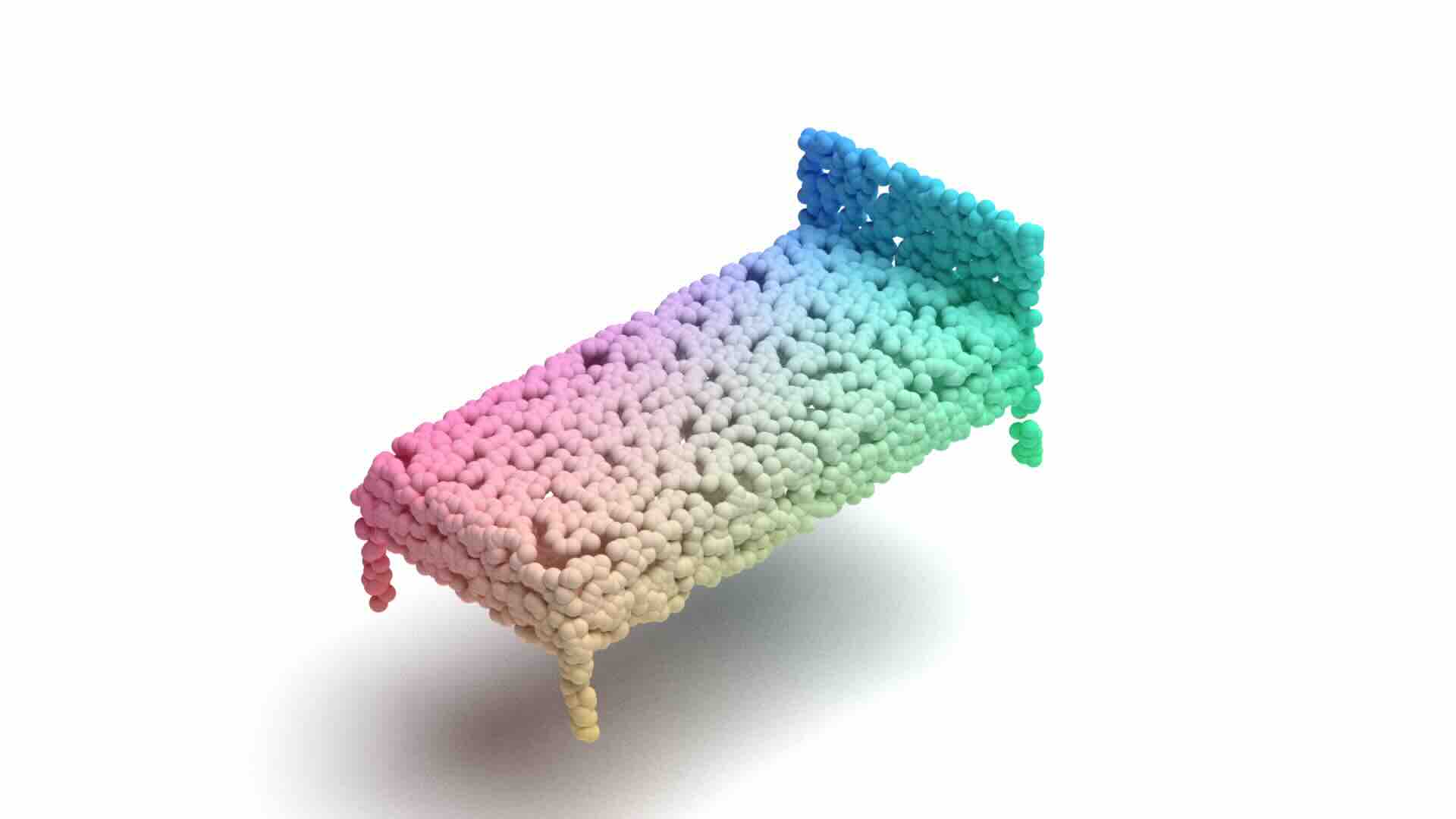} \\
        \includegraphics[trim={10.0cm 1.0cm 10.0cm 1cm},clip,width=0.32\linewidth]{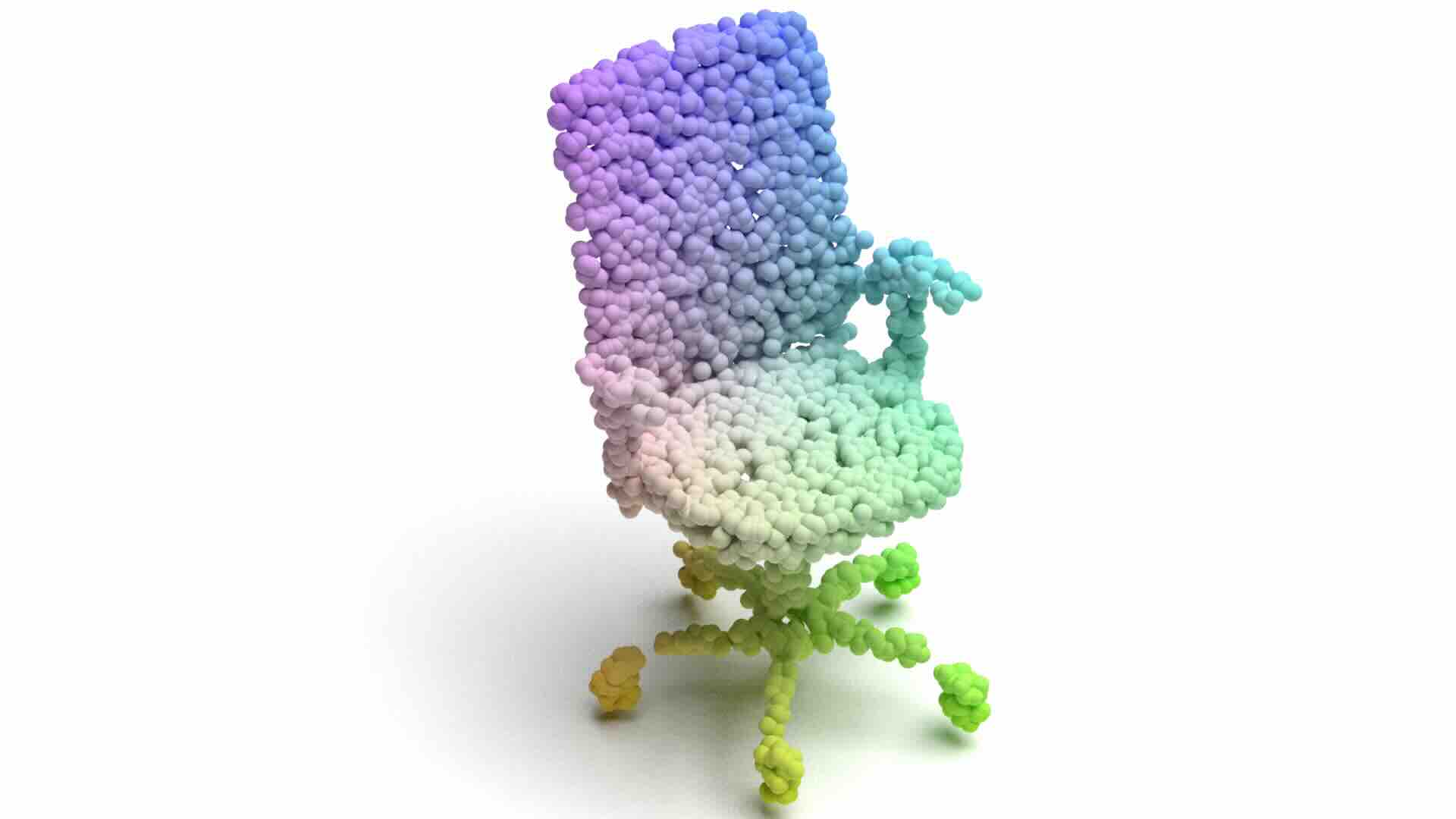} &
        \includegraphics[trim={10.0cm 1.0cm 10.0cm 1cm},clip,width=0.32\linewidth]{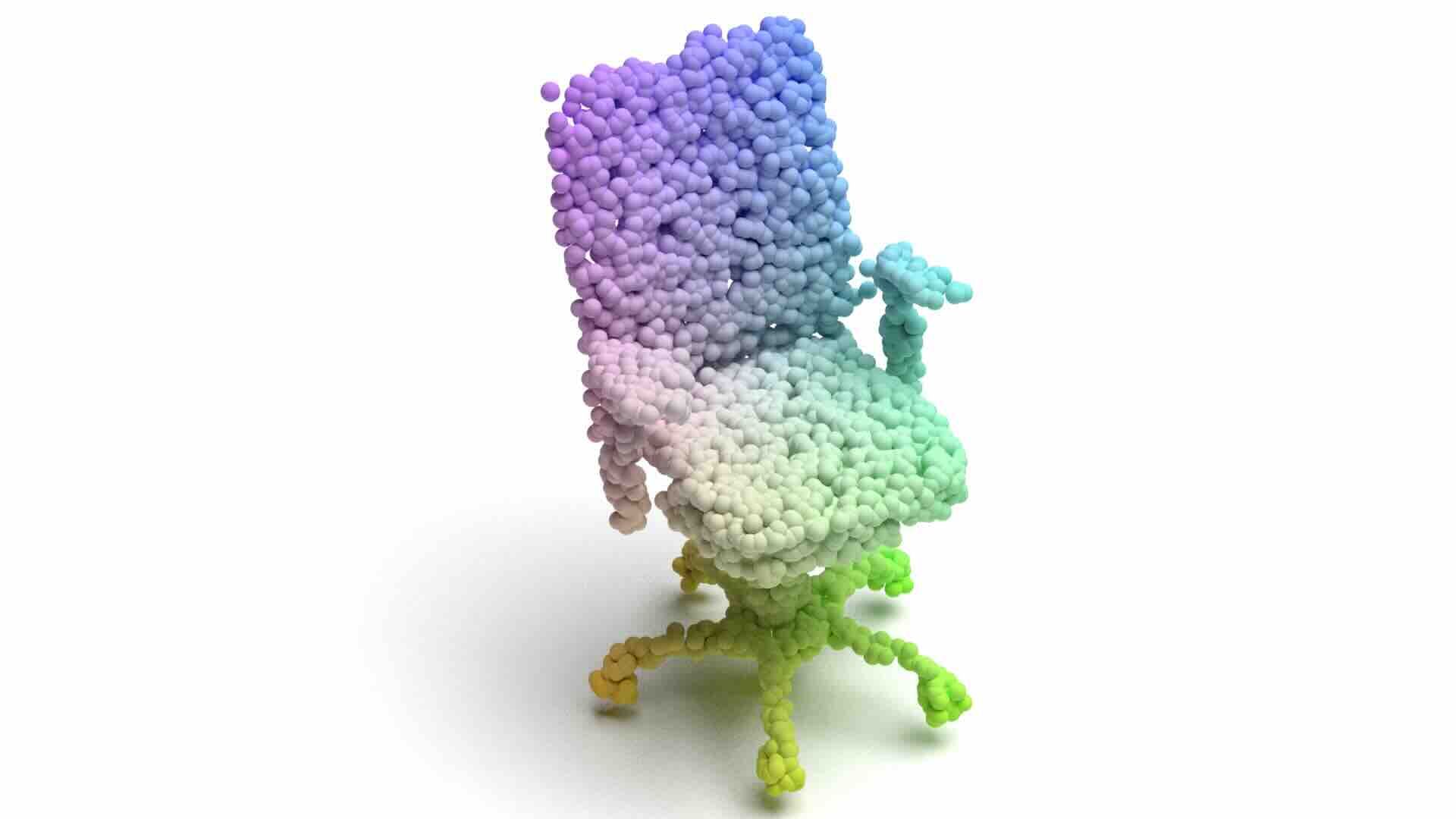} &
        \includegraphics[trim={10.0cm 1.0cm 10.0cm 1cm},clip,width=0.32\linewidth]{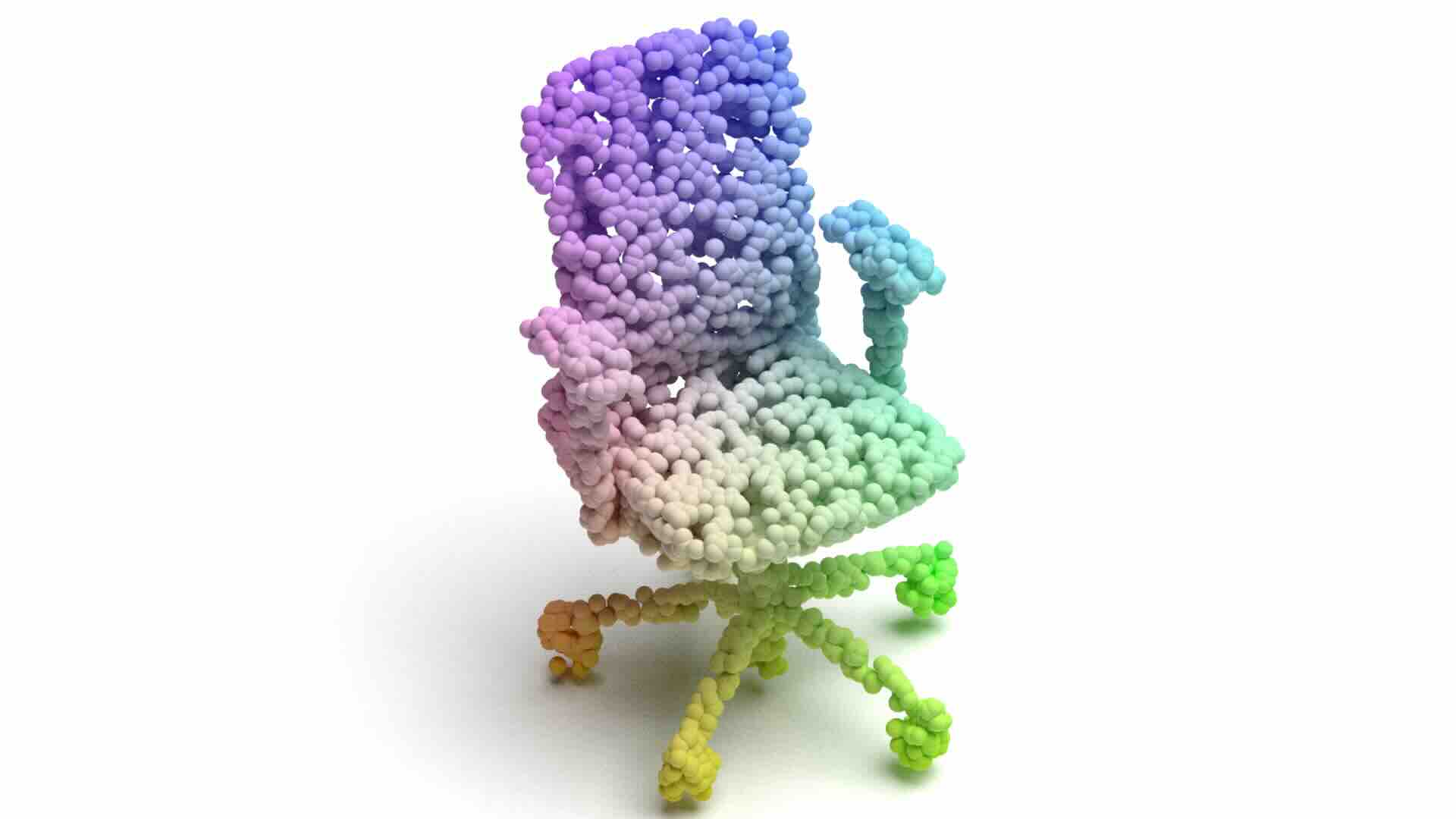} \\
        
        \scriptsize Ours & \scriptsize SAM 2 Mask & \scriptsize GT \\
    \end{tabular}
        } \\
        \small
        (b) Segmentation Robustness\\
    \end{minipage}
    \begin{minipage}[t]{0.3\linewidth}
        \centering
    \label{fig:unilat3d}
    \setlength{\tabcolsep}{2pt}
    \scalebox{0.9}{
        \begin{tabular}{ccc}
        \includegraphics[width=0.32\linewidth]{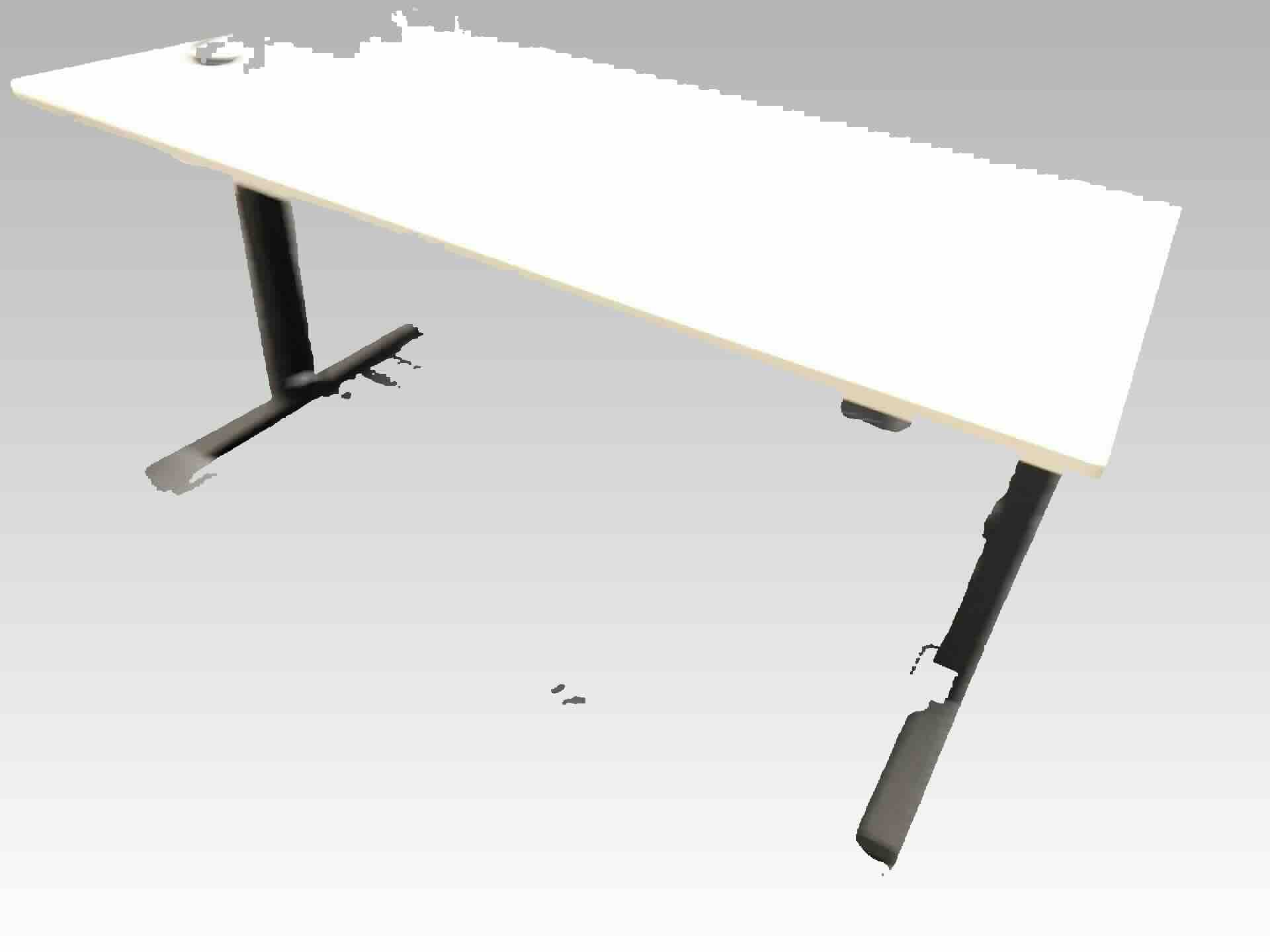} &
        \includegraphics[trim={0.0cm 2.0cm 0.0cm 2.5cm},clip,width=0.32\linewidth]{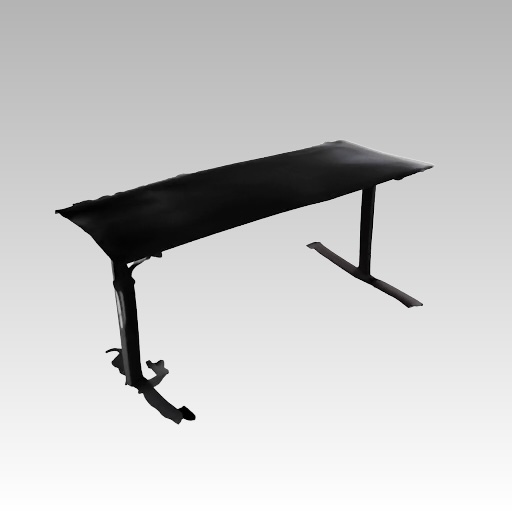} &
        \includegraphics[trim={0.0cm 2.0cm 0.0cm 2.5cm},clip,width=0.32\linewidth]{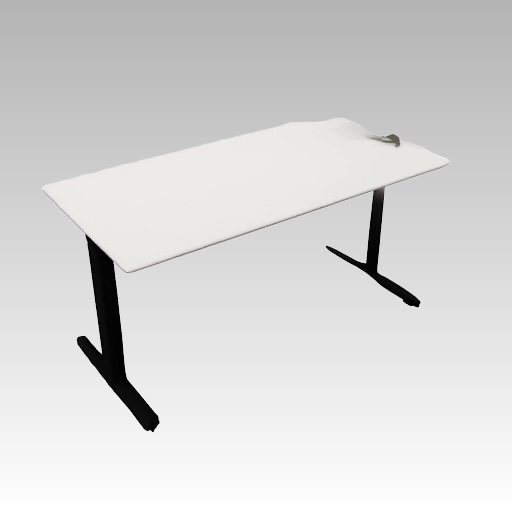} \\
        \includegraphics[trim={0.0cm 1.0cm 20cm 7cm},clip,width=0.32\linewidth]{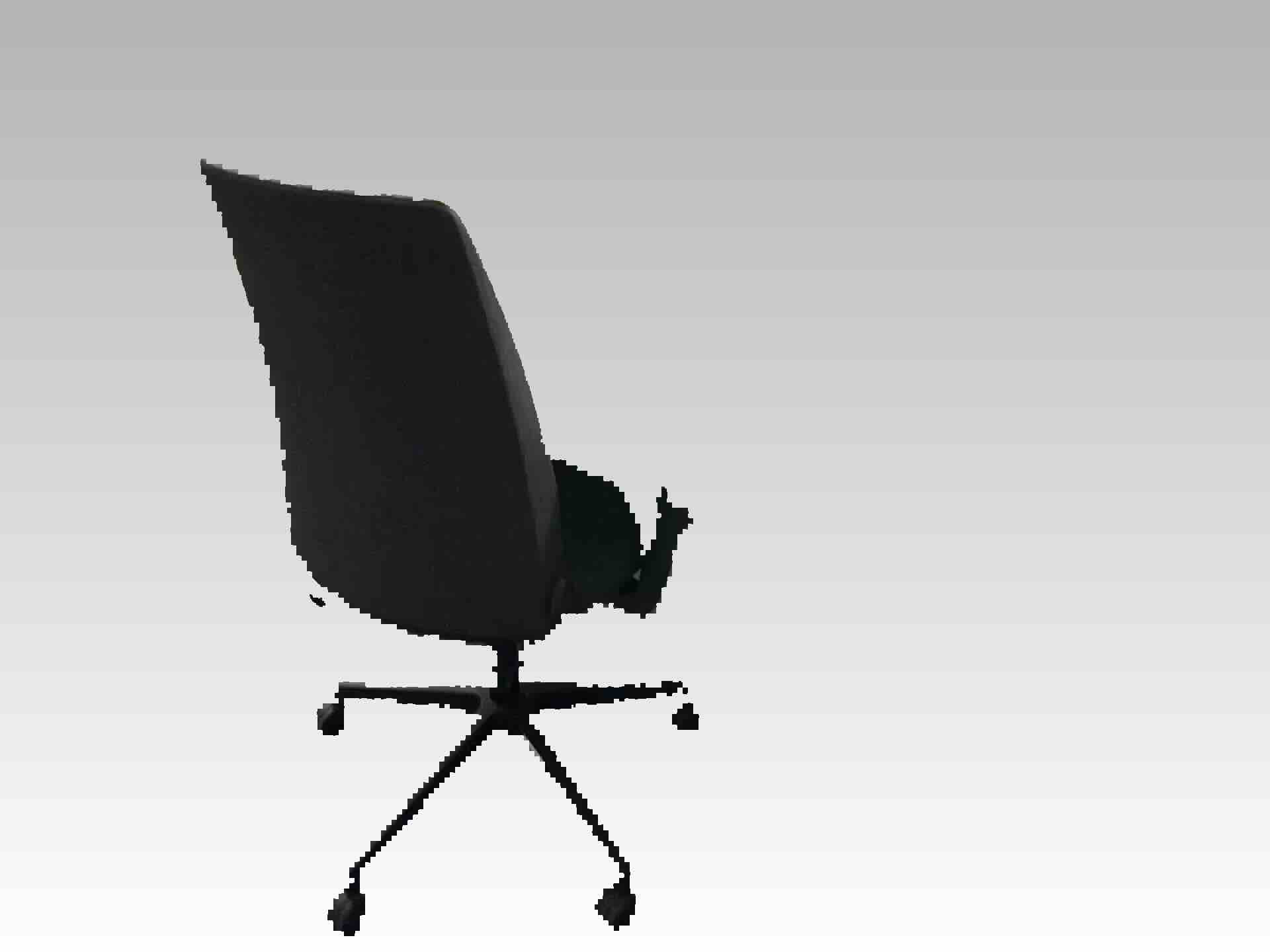} &
        \includegraphics[trim={0.0cm 1.0cm 0.0cm 1cm},clip,width=0.32\linewidth]{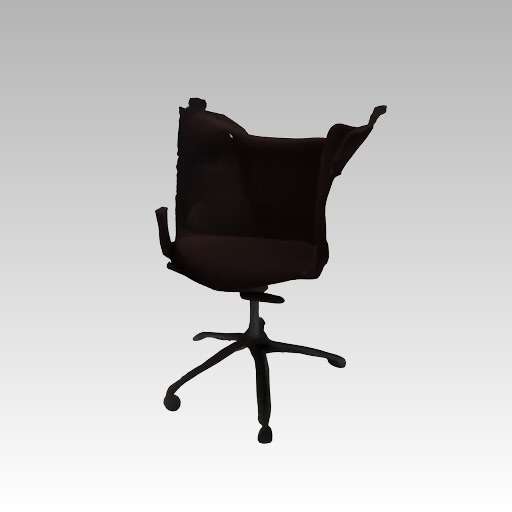} &
        \includegraphics[trim={0.0cm 1.0cm 0.0cm 1cm},clip,width=0.32\linewidth]{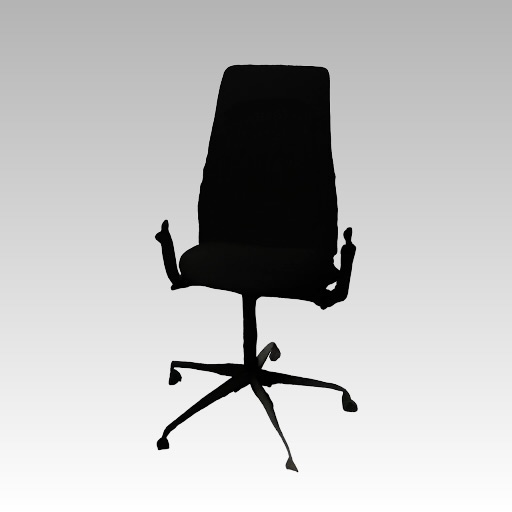} \\
        \scriptsize Input & \scriptsize Vanilla UniLat3D & \scriptsize w/ Our Guid. \\
    \end{tabular}
        }\\
        \small
        (c) Guidance Effectiveness\\
    \end{minipage}
    \vspace{2mm}
     \caption{\textbf{Ablation Studies.} (a) Visualizing the geometric robustness of our method when driven by Depth Anything 3 (DA3) estimated depth versus projected depth. (b) Visualizing the geometric robustness of our method when driven by SAM 2 estimated masks versus projected masks. (c)  Rendering comparison between Vanilla UniLat3D and our guided refinement.}
    \label{fig:all_ablations}
\end{table}

\subsection{Limitations}
Despite these advancements, challenges remain for future work. First, our reconstruction fidelity is inherently constrained by the pretrained diffusion prior; significant domain gaps between the prior and real-world observations still necessitate sensitive prompt tuning. Second, a fundamental geometry-texture trade-off persists during the 3DGS refinement stage: while optimization significantly enhances visible details, the lack of photometric constraints on unobserved surfaces can occasionally lead to over-smoothing or minor structural inconsistencies in those regions. Finally, our feature volume construction relies on aggregating view-dependent 2D DINO features, which can introduce blurring in overlapping voxels, transitioning to 3D-consistent feature spaces remains an open direction to address this challenge.

%% file: sections/6_conclusion.tex
\section{Conclusion}

We presented \OURS, a framework that reformulates sparse-view 3D reconstruction as a guided inverse problem to reconstruct complete 3D models despite the occlusions present in the observations.
By integrating a 3D Gaussian Splatting refinement stage, \OURS goes beyond the "synthetic look" of pure generative outputs, achieving visually faithful photorealism even under extremely sparse (e.g., 3-view) conditions. 
Beyond reconstruction, our object-aware representation can facilitate seamless scene editing, establishing a robust paradigm for leveraging 3D generative priors in high-fidelity real-world modeling.

\section{Acknowledgements}
We thank Sinisa Stekovic for his valuable insights and constructive discussions during the early stages of this research.

%% file: X_suppl.tex
\clearpage
\setcounter{section}{0}
\setcounter{figure}{0}
\setcounter{table}{0}
\setcounter{page}{1}

\renewcommand{\thesection}{S\arabic{section}}
\renewcommand{\thefigure}{S\arabic{figure}}
\renewcommand{\thetable}{S\arabic{table}}

\makeatletter
\@addtoreset{section}{part}
\maketitlesupplementary

\begin{abstract}
We provide more experimental results to give detailed comparisons between our approach and the baseline methods in Sec.~\ref{supp:aer}. Furthermore, in Sec.~\ref{supp:aid}, we include additional implementation details for our approach.  Finally, we also include an extra supplementary video, and we will make our code publicly available upon acceptance to benefit future works in this field. 
\end{abstract}

\section{Additional Experimental Results}
\label{supp:aer}

\paragraph{Additional Quantitative Results for Object Reconstruction in all the Datasets.} By measuring distances specifically from the ground-truth (GT) surface to the predicted points (Unidirectional Chamfer Distance), we accurately quantify the reconstruction's completeness and surface coverage without being penalized by extraneous outliers or boundary artifacts in the baseline results. We also provide additional quantitative evaluations for object reconstruction, including Dual-directional Chamfer Distance (CD) and Precision across all datasets, in Tab.~\ref{tab:additional_recon_eval}. These results further demonstrate the robustness of \OURS in reconstructing complete geometry from sparse observations, consistently outperforming state-of-the-art baselines across both synthetic and real-world scenes.

\begin{table}[b]
\centering
\vspace{-0.3cm}
\scalebox{0.9}{
\begin{tabular}{l cc cc cc}
\toprule
& \multicolumn{2}{c}{\makebox[0.25\textwidth][c]{\textbf{3D-FRONT}~\cite{fu20213d}}} & \multicolumn{2}{c}{\makebox[0.25\textwidth][c]{\textbf{ScanNet++}~\cite{yeshwanth2023scannet++}}} & \multicolumn{2}{c}{\makebox[0.25\textwidth][c]{\textbf{ShapeR}~\cite{siddiqui2026shaper}}}\\
\cmidrule(lr){2-3} \cmidrule(lr){4-5} \cmidrule(lr){6-7}

Method & 
\makebox[0.095\textwidth][c]{CD\textdownarrow} & 
\makebox[0.095\textwidth][c]{Precision\textuparrow} & \makebox[0.095\textwidth][c]{CD\textdownarrow} & 
\makebox[0.095\textwidth][c]{Precision\textuparrow} & \makebox[0.095\textwidth][c]{CD\textdownarrow} & 
\makebox[0.095\textwidth][c]{Precision\textuparrow} \\

\midrule
3DGS-D~\cite{kerbl20233d}       & 6.13  & 44.77 & 6.32 & 33.27 & 7.35 & 57.48 \\
AGS-Mesh~\cite{ren2025ags}        & 6.17  & 50.62 & 6.62 & 30.23  & 6.69 & 60.69 \\
OM-GSD$^{\ast}$~\cite{lu2025orientation} & 29.07 & 30.01 &12.50 & 28.99 & 11.29 & 23.39 \\
RVG-GSD$^{\ast}$~\cite{chang2025reconviagen} & 9.76 & 23.99 & 10.10 & 21.85 & 12.65 & 24.79 \\
SAM3D-GSD$^{\ast}$~\cite{chen2025sam} & 6.08 & 32.99 & 4.67 & 29.18 & 4.60 & 33.25\\
\textbf{Ours} & \textbf{3.38} & \textbf{64.51} & \textbf{3.53} & \textbf{43.97} & \textbf{3.45} & \textbf{63.55}  \\
\bottomrule
\end{tabular}
}
\caption{\textbf{Additional Quantitative Evaluation of Geometric Fidelity.} We report Duel-directional Chamfer Distance (CD $\times 10^2$) and Precision (\%) on 3D-FRONT (synthetic data), ScanNet++ and ShapeR Evaluation Dataset~(real data) as additional evaluations for 3D reconstruction. Threshold $\tau$ is set to $1\%$ of $d_{diag}$. ($^{\ast}$: Results by OM-GSD and RVG-GSD are aligned via uniform scaling and rigid transformation and rotation. Results by SAM3D-GSD are obtained using its multi-view implementation.))}
\label{tab:additional_recon_eval}
\end{table}

\paragraph{Quantitative Results for All Evaluated Objects.} Consistent with the discussion in the main manuscript, we further present the aggregate metrics averaged across our entire experimental set in Tab.~\ref{tab:all_object_rendering_scannetpp}. Most methods exhibit competitive performance on these aggregate statistics, as simple, well-observed samples tend to dominate the average. However, our framework maintains a more stable performance profile across both simple and highly occluded objects. While the average metrics appear comparable, our method shows superior 3D structural consistency in complex, unconstrained settings, effectively recovering details that remain ambiguous for standard optimization-based baselines.

\begin{table}[t]
\centering
\scalebox{0.9}{
\begin{tabular}{l ccc ccc}
\toprule 
& \multicolumn{3}{c}{\makebox[0.33\textwidth][c]{\textbf{ScanNet++}~\cite{yeshwanth2023scannet++}}} 
& \multicolumn{3}{c}{\makebox[0.33\textwidth][c]{\textbf{ShapeR}~\cite{siddiqui2026shaper}}}\\
\cmidrule(lr){2-4} \cmidrule(lr){5-7}

Method & 
\makebox[0.1\textwidth][c]{PSNR \textuparrow} & 
\makebox[0.1\textwidth][c]{SSIM \textuparrow} & 
\makebox[0.1\textwidth][c]{LPIPS \textdownarrow} & 
\makebox[0.1\textwidth][c]{PSNR \textuparrow} & 
\makebox[0.1\textwidth][c]{SSIM \textuparrow} & 
\makebox[0.1\textwidth][c]{LPIPS \textdownarrow}  \\

\midrule
3DGS-D~\cite{kerbl20233d} & 24.13 & \textit{0.961} & \underline{5.11} & 28.95 & \underline{0.975} & \textit{3.33} \\
        AGS-Mesh~\cite{ren2025ags} & \textit{24.70} & \textbf{0.965} & \textbf{4.80} & \textbf{30.01} & \textbf{0.976} & \underline{3.30} \\
        OM-GSD$^{\ast}$~\cite{lu2025orientation} & 24.26 & 0.953 & 5.65 & \textit{29.40} & \textit{0.974} & \textit{3.33} \\
        RVG-GSD$^{\ast}$~\cite{chang2025reconviagen} & 24.18 & 0.956 & 5.41 & \textit{29.40} & \underline{0.975} & 3.42 \\
        SAM3D-GSD$^{\ast}$~\cite{chen2025sam} & \underline{25.10} & \underline{0.963} & \textit{5.16} & \textit{29.40} & \textit{0.974} & \textit{3.33} \\
        \textbf{Ours} & \textbf{25.11} & 0.958 & 5.42 & \underline{29.99} & \underline{0.975} & \textbf{3.28}\\
\bottomrule
\end{tabular}
}
\caption{\textbf{Quantitative Rendering Comparison on all the ScanNet++~\cite{yeshwanth2023scannet++} and ShapeR Evaluation Dataset~\cite{siddiqui2026shaper} Samples.} We evaluate novel-view synthesis performance particularly on incomplete observations and occluded regions, reporting metrics averaged across all evaluated objects within the ScanNet++ dataset and ShapeR Evaluation Dataset. \textbf{Bold}, \underline{underlined}, and \textit{italicized} values indicate the first, second, and third best results, respectively. (LPIPS $\times 10^2$)}
\label{tab:all_object_rendering_scannetpp}
\vspace{-5mm}
\end{table}

\begin{figure*}[t] 
    \centering
    \vspace{-5mm}
    \setlength{\tabcolsep}{2pt}
    \resizebox{\textwidth}{!}{
    \begin{tabular}{cccccccc}
        \includegraphics[trim={15cm 0.0cm 15cm 0.0cm},clip,width=0.12\textwidth]{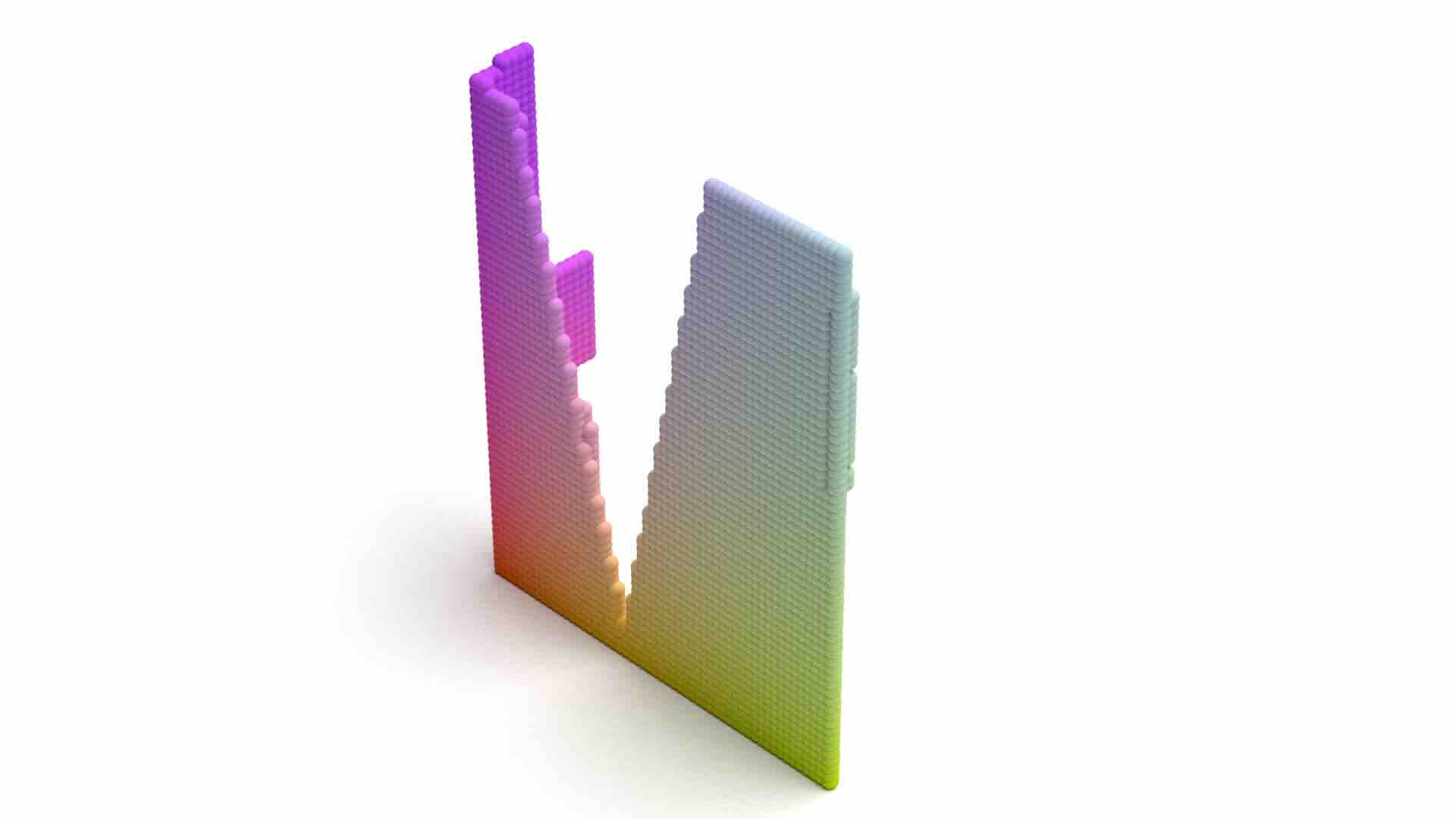} &
        \includegraphics[trim={15cm 0.0cm 15cm 0.0cm},clip, width=0.12\textwidth]{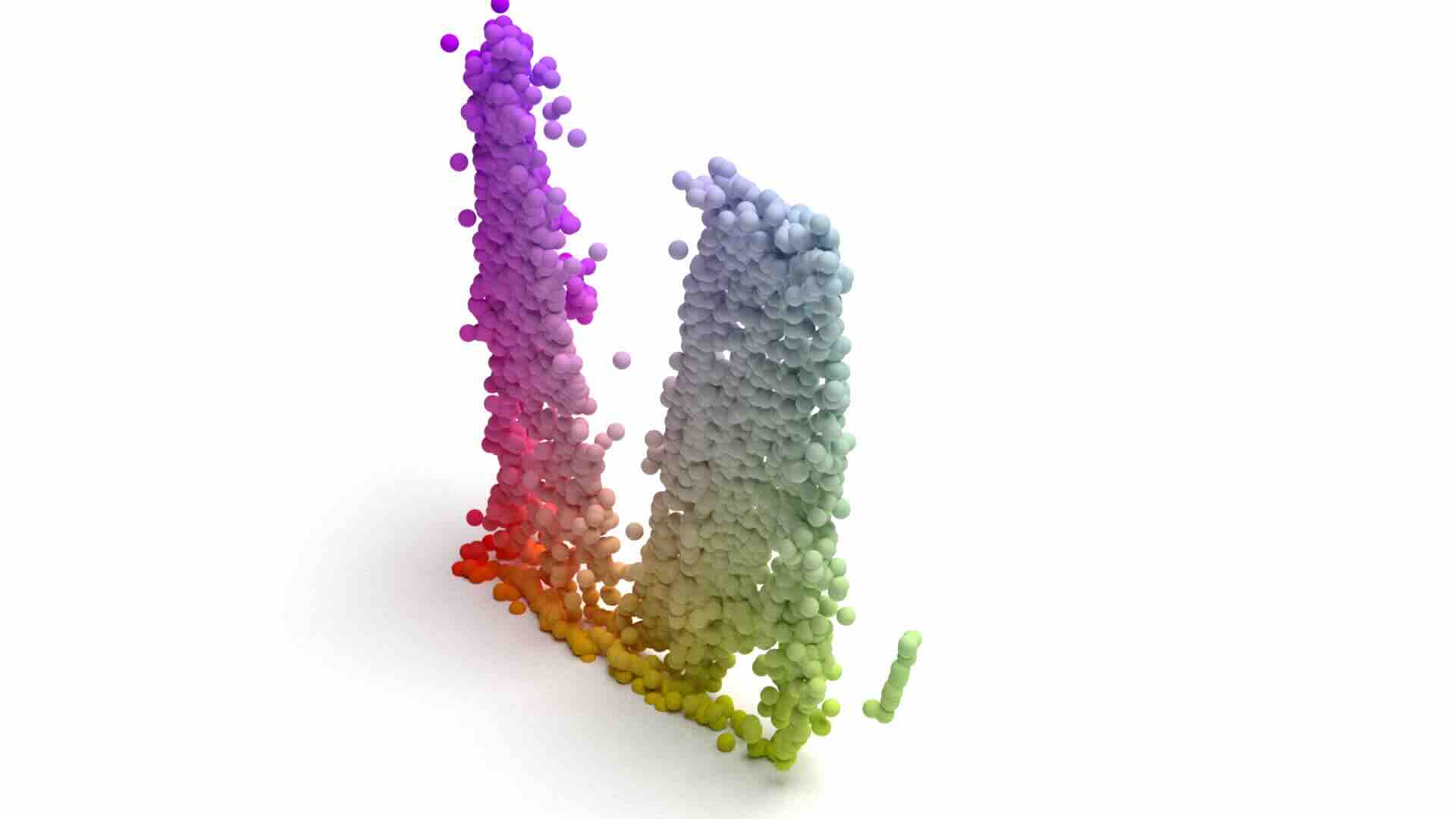} &
        \includegraphics[trim={15cm 0.0cm 15cm 0.0cm},clip, width=0.12\textwidth]{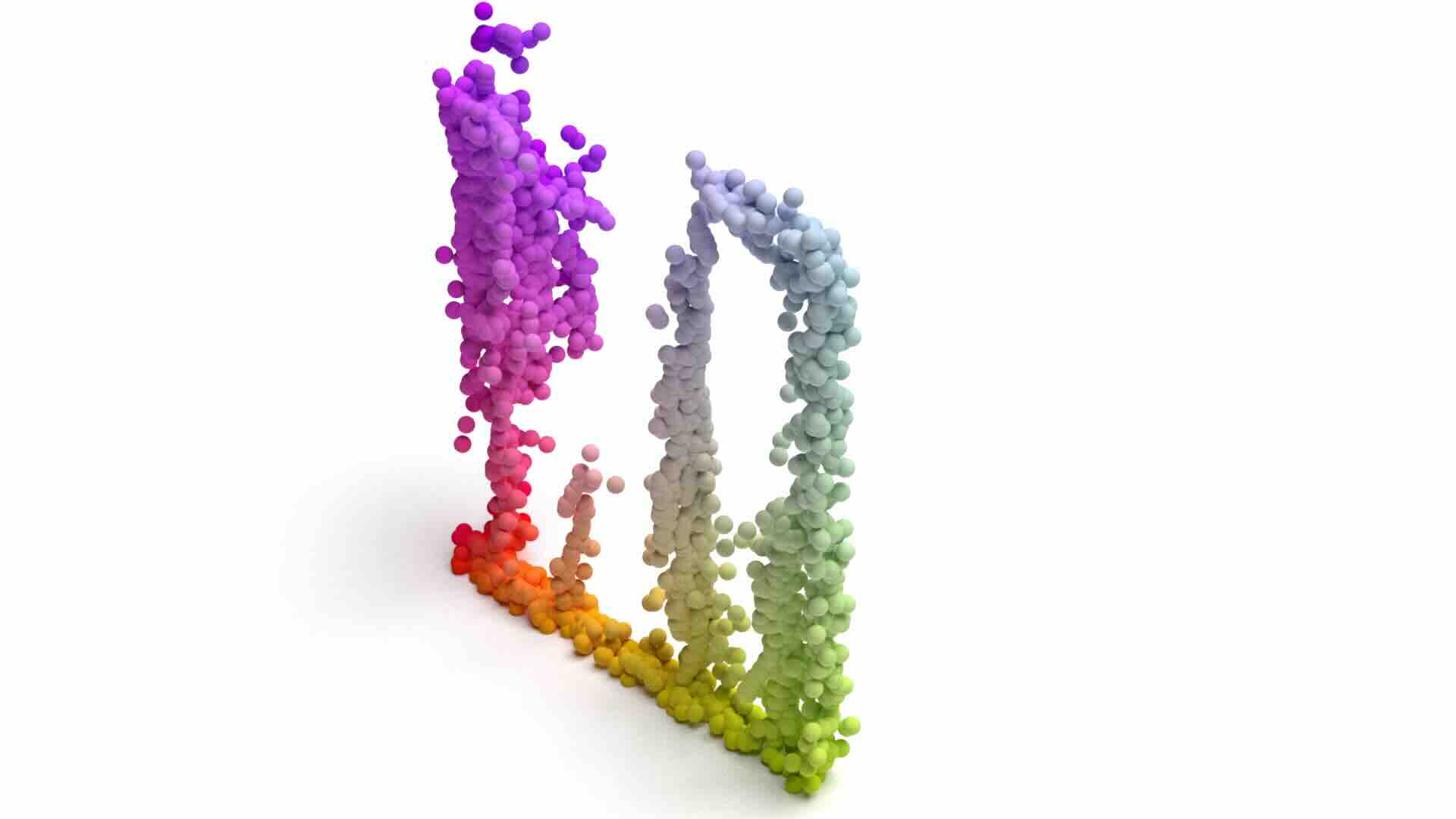} &
        \includegraphics[trim={15cm 0.0cm 15cm 0.0cm},clip, width=0.12\textwidth]{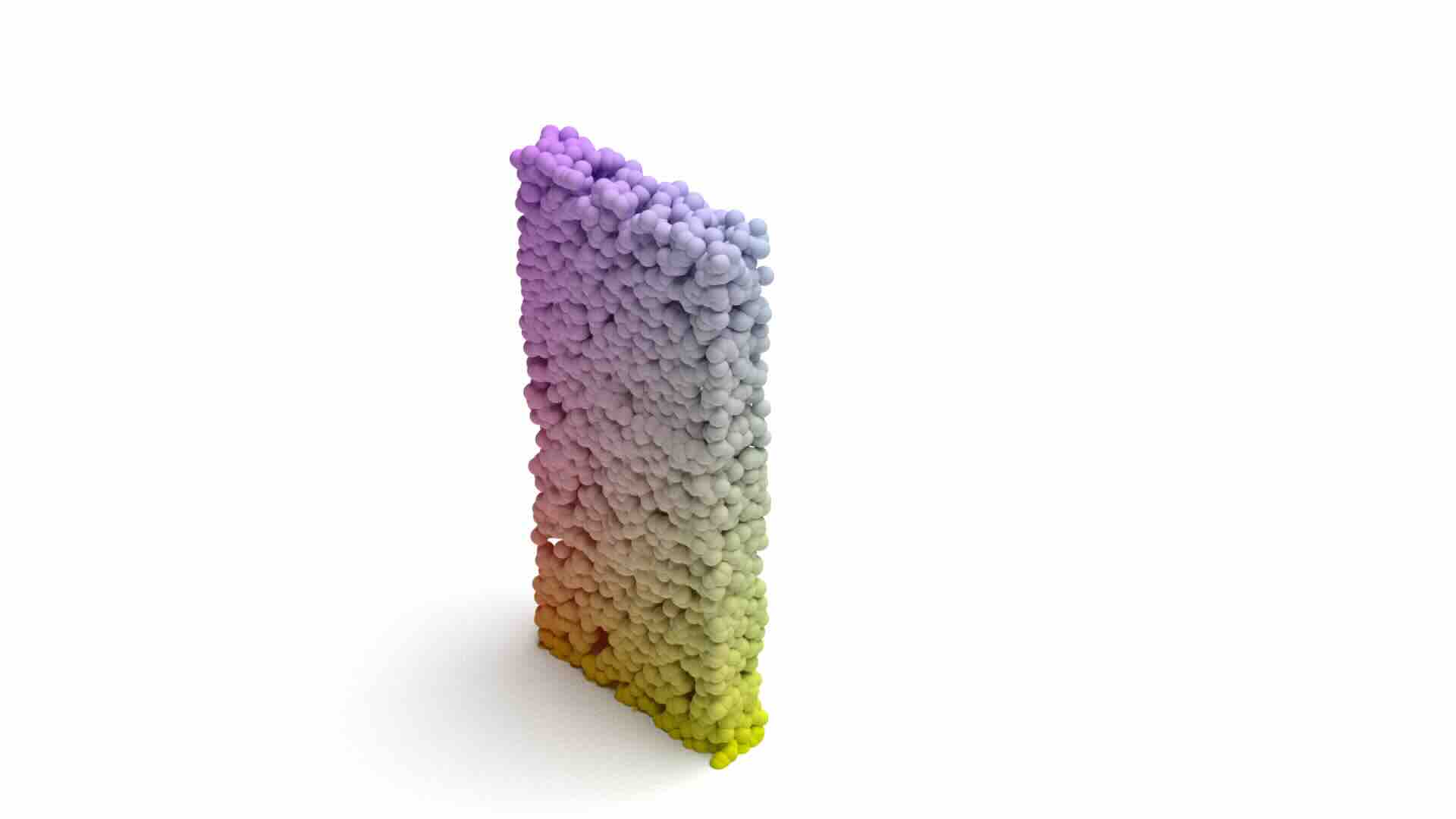} &
        \includegraphics[trim={15cm 0.0cm 15cm 0.0cm},clip, width=0.12\textwidth]{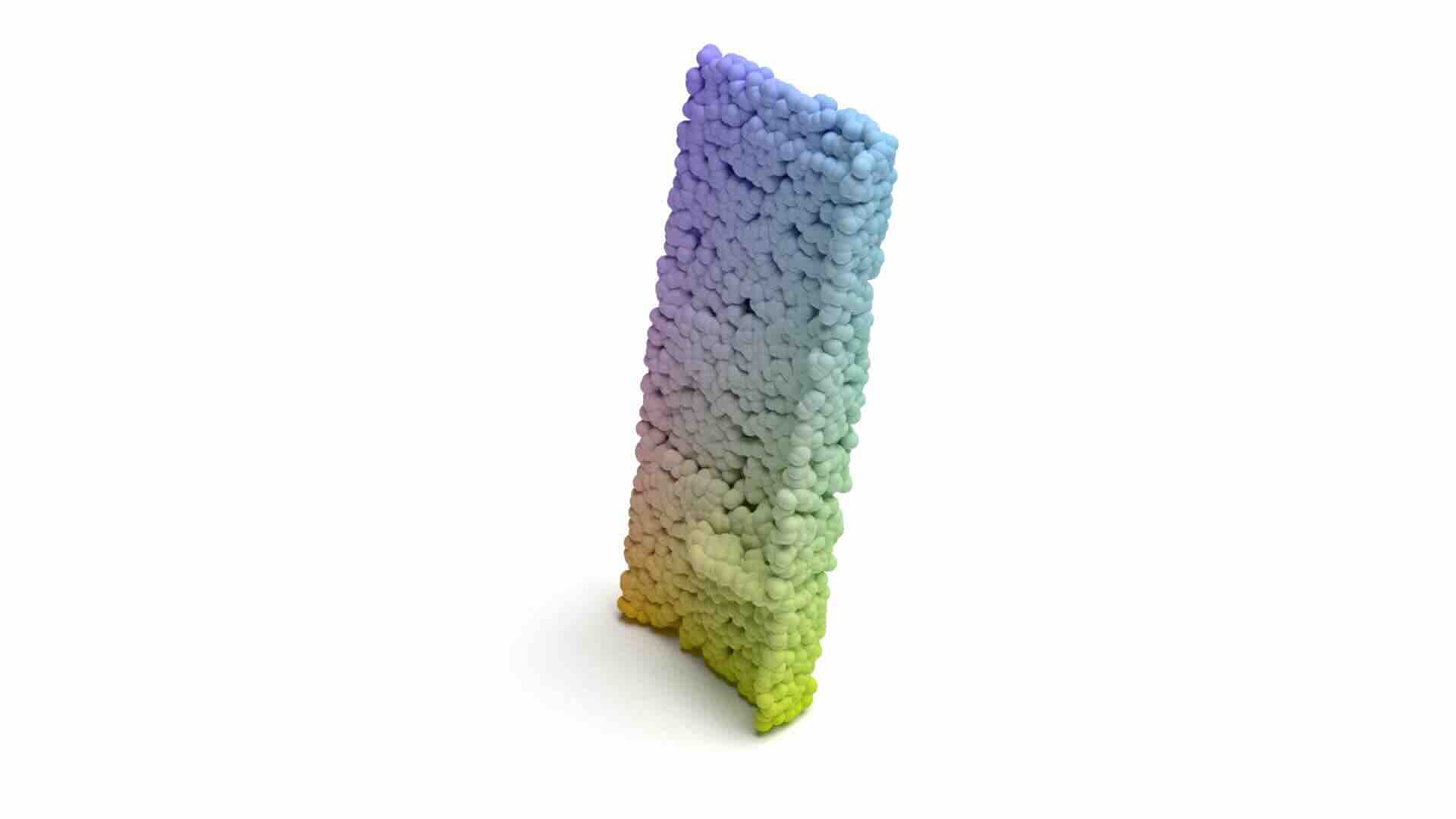}
        &
        \includegraphics[trim={15cm 0.0cm 15cm 0.0cm},clip, width=0.12\textwidth]{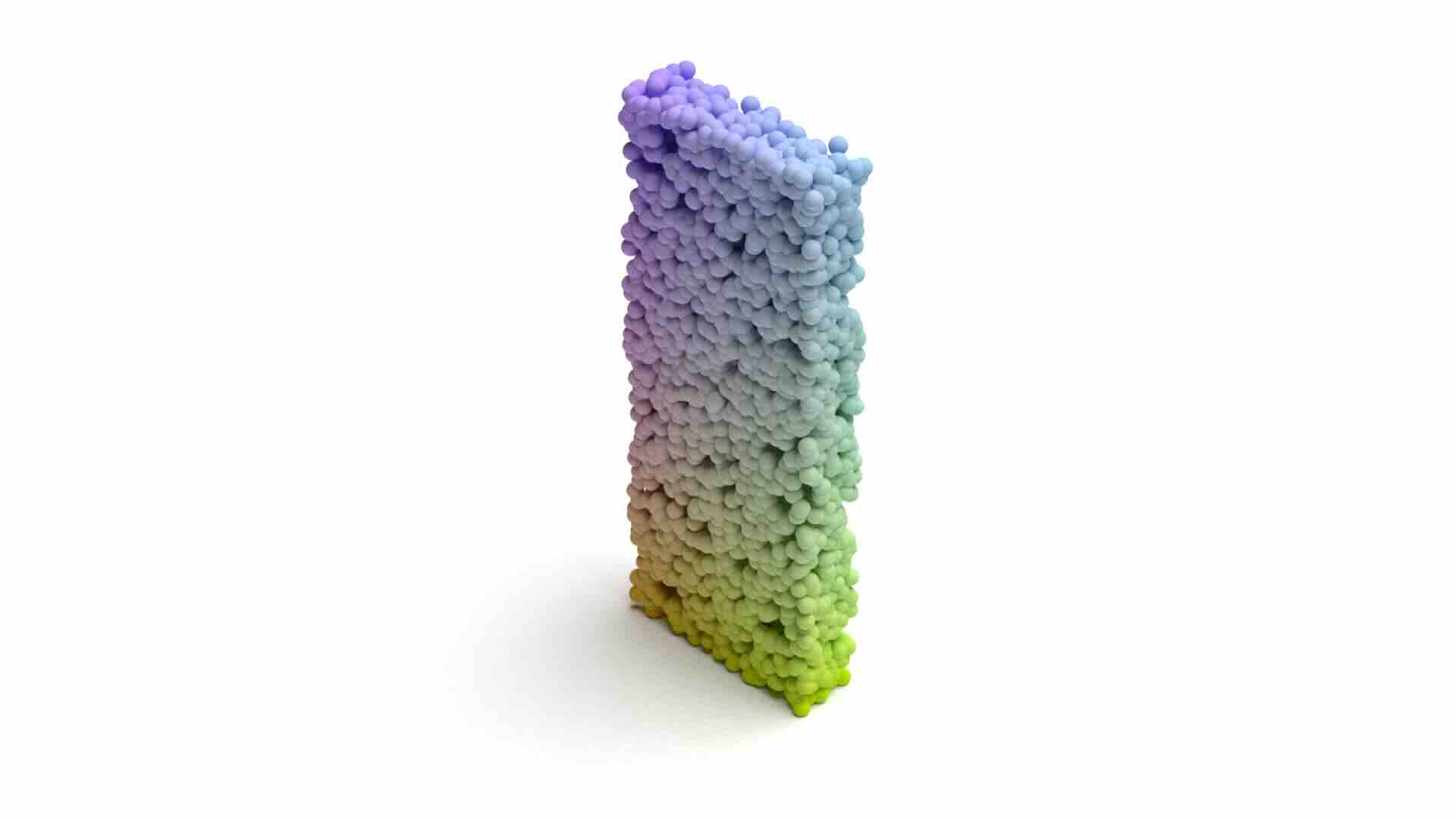}
        &
        \includegraphics[trim={15cm 0.0cm 15cm 0.0cm},clip,width=0.12\textwidth]{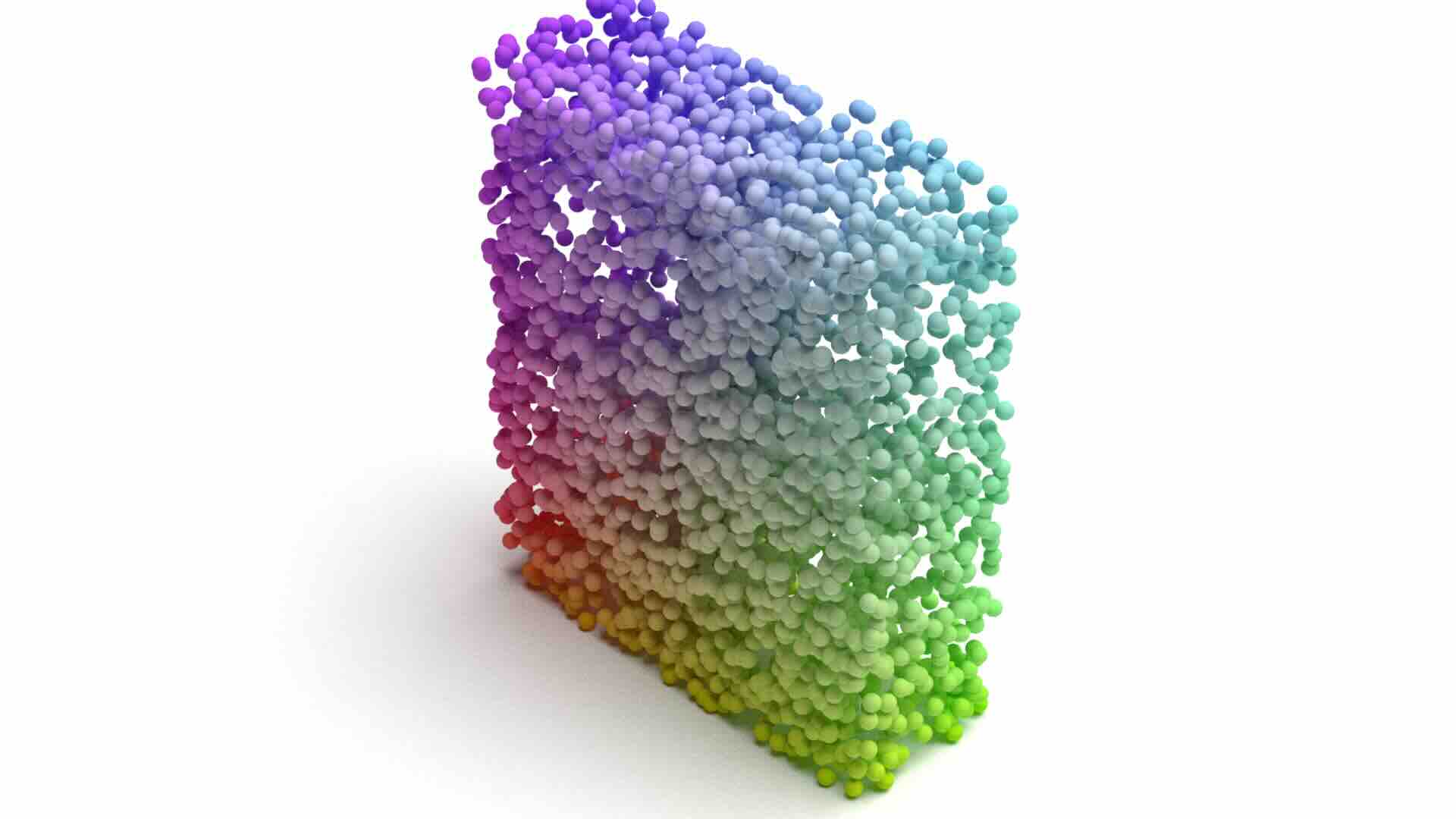} &
        \includegraphics[trim={15cm 0.0cm 15cm 0.0cm},clip,width=0.12\textwidth]{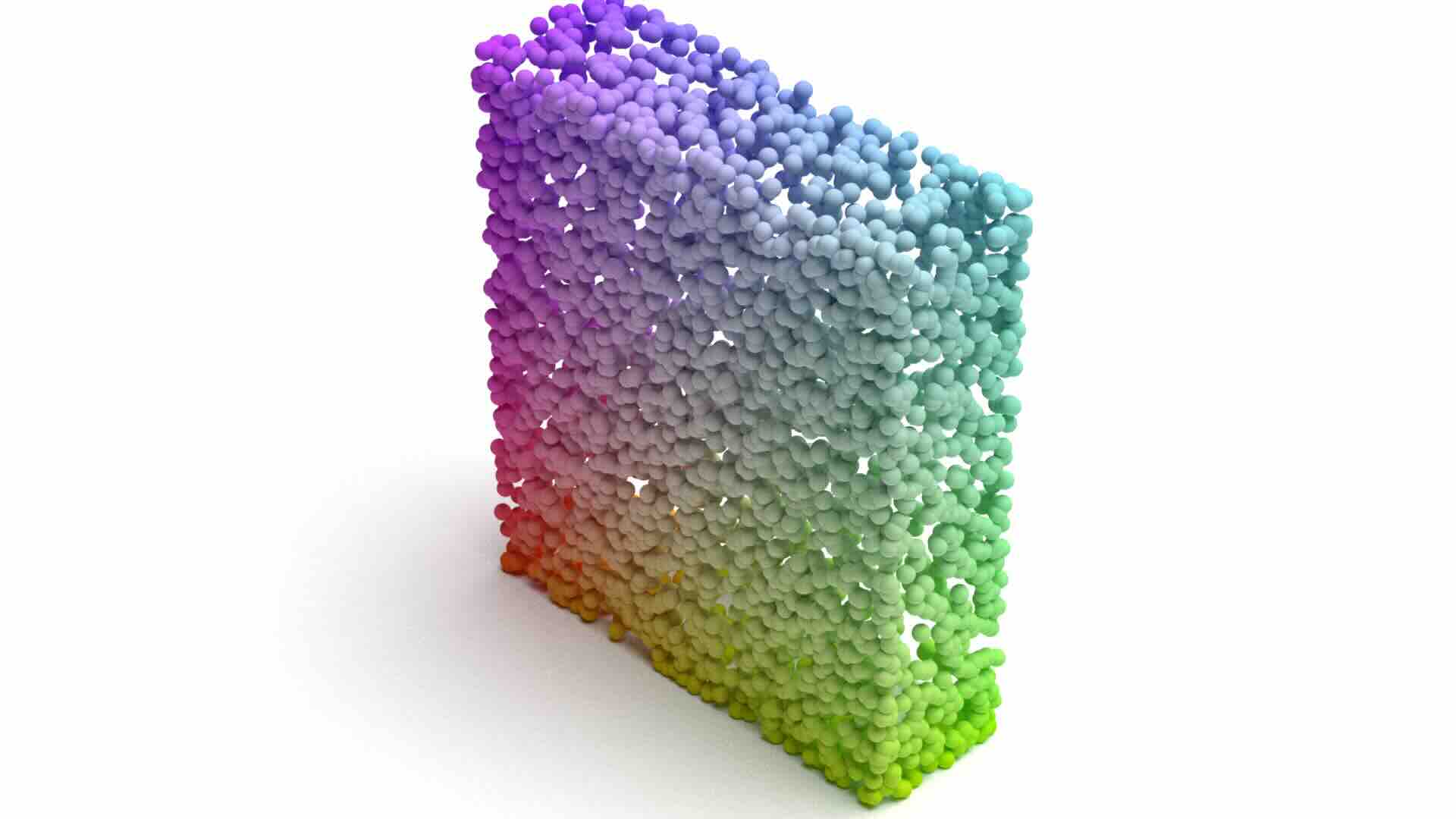}
        \\
        \includegraphics[trim={15cm 0.0cm 15cm 0.0cm},clip,width=0.12\textwidth]{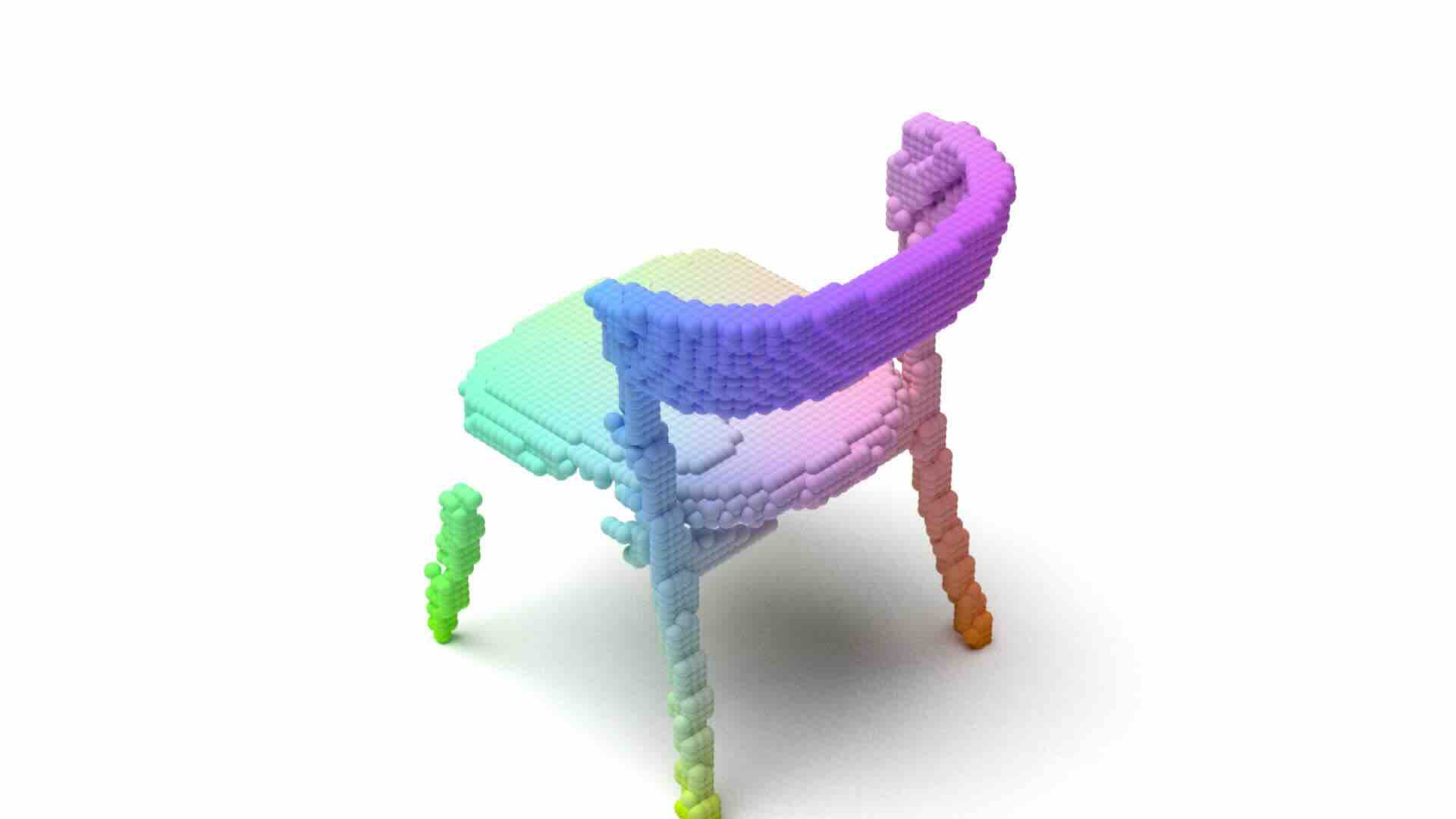} &
        \includegraphics[trim={15cm 0.0cm 15cm 0.0cm},clip, width=0.12\textwidth]{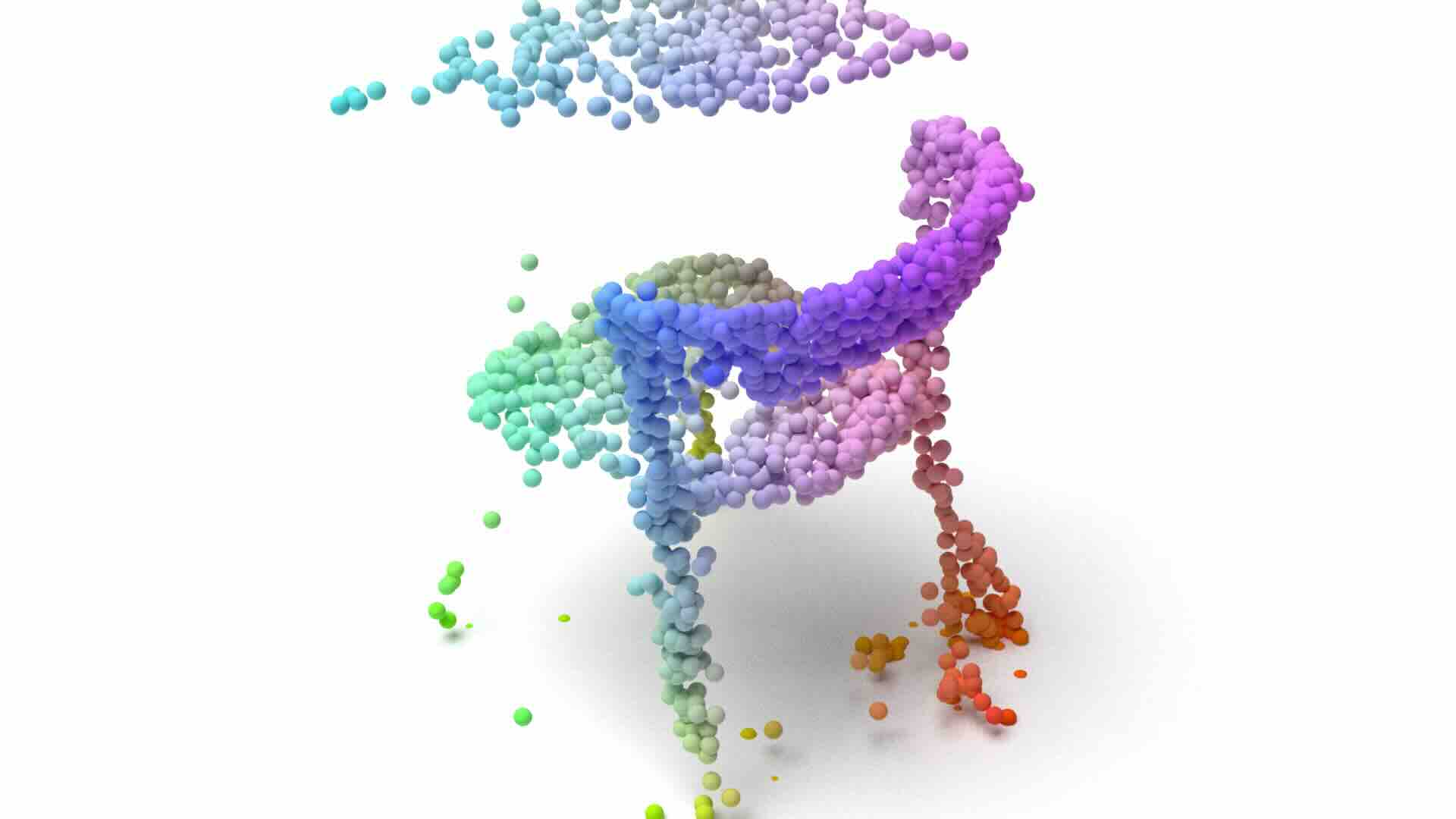} &
        \includegraphics[trim={15cm 0.0cm 15cm 0.0cm},clip, width=0.12\textwidth]{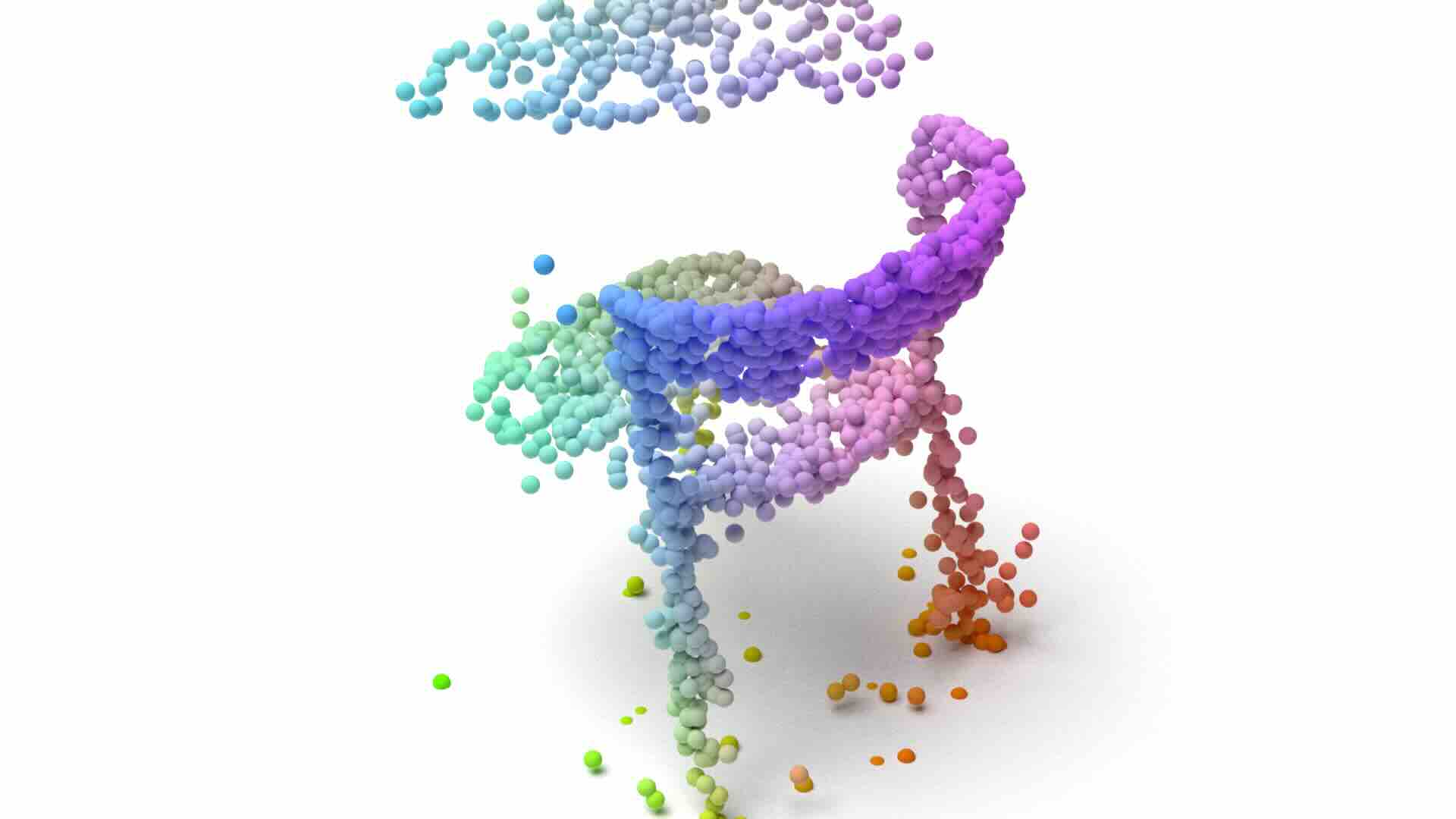} &
        \includegraphics[trim={15cm 0.0cm 15cm 0.0cm},clip, width=0.12\textwidth]{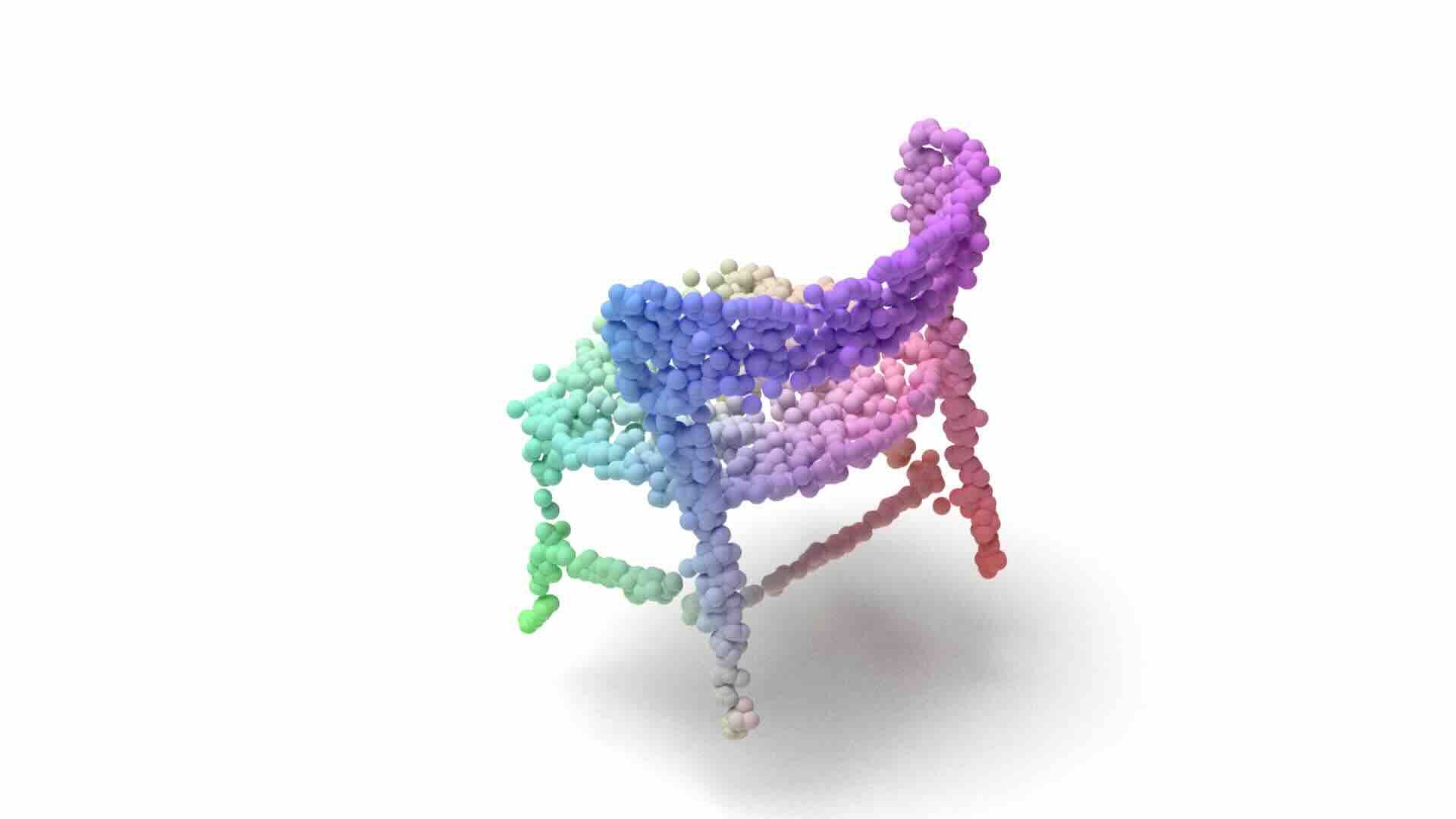} &
        \includegraphics[trim={15cm 0.0cm 15cm 0.0cm},clip, width=0.12\textwidth]{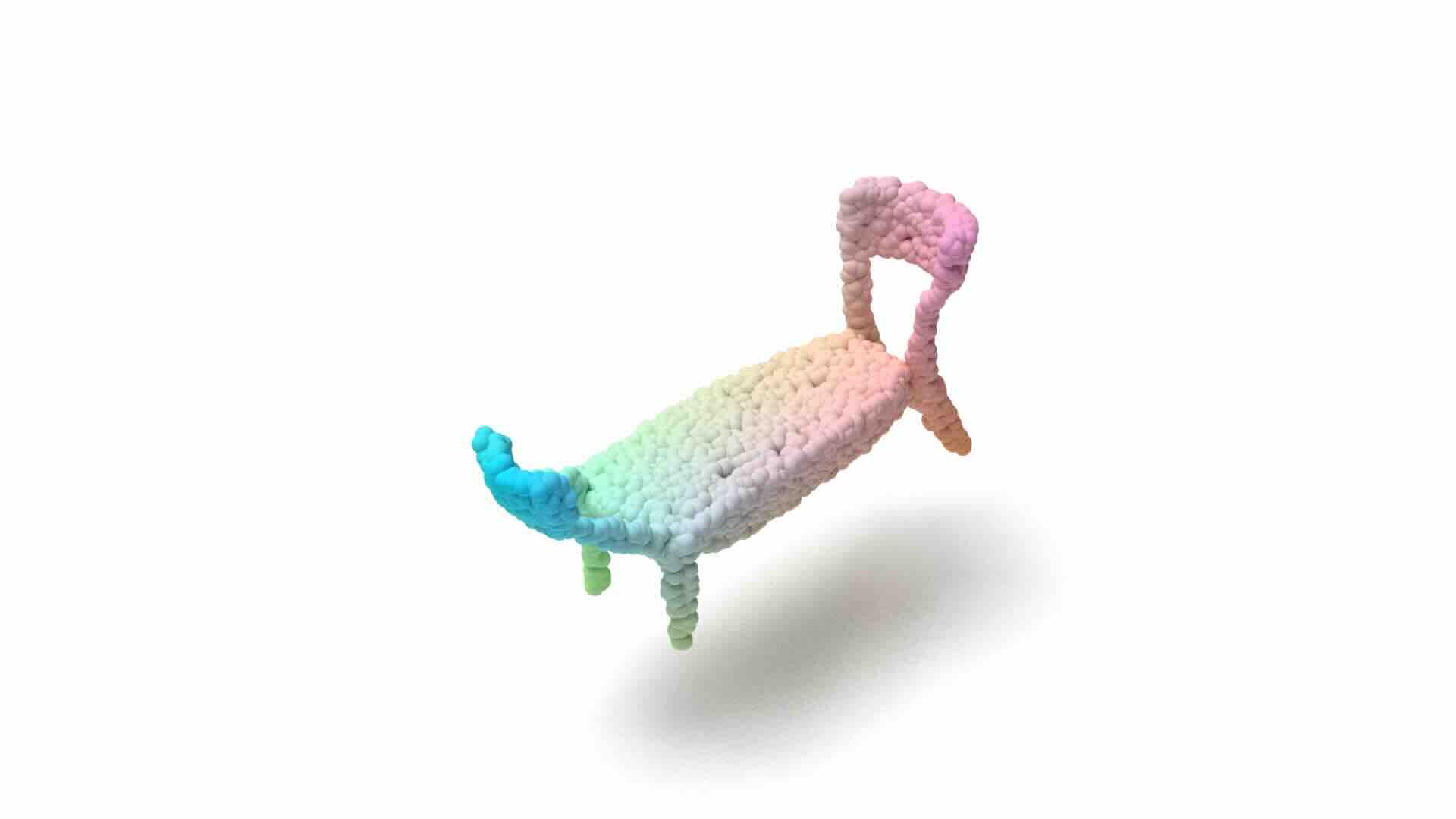}
        &
        \includegraphics[trim={15cm 0.0cm 15cm 0.0cm},clip, width=0.12\textwidth]{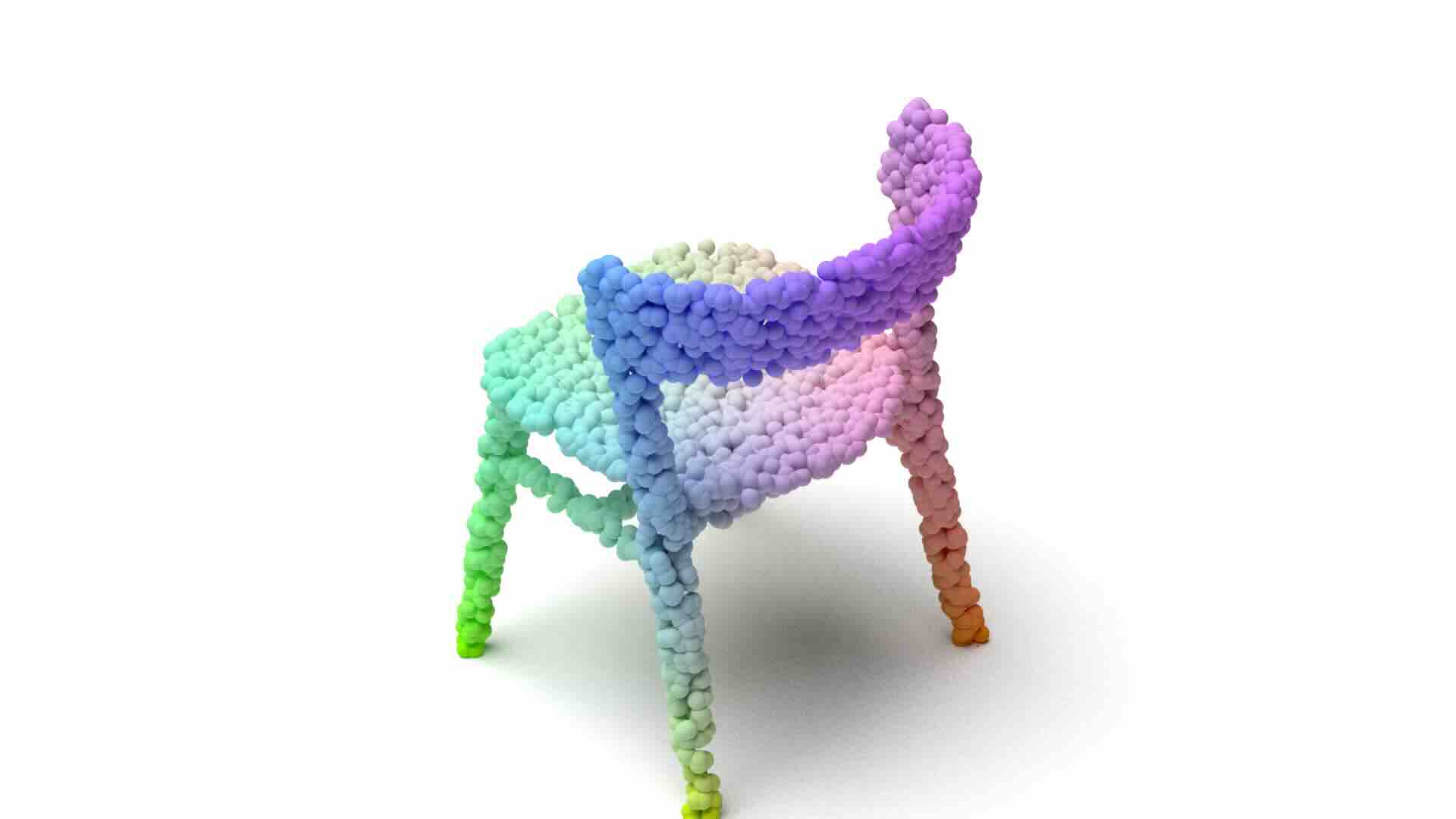}
        &
        \includegraphics[trim={15cm 0.0cm 15cm 0.0cm},clip, width=0.12\textwidth]{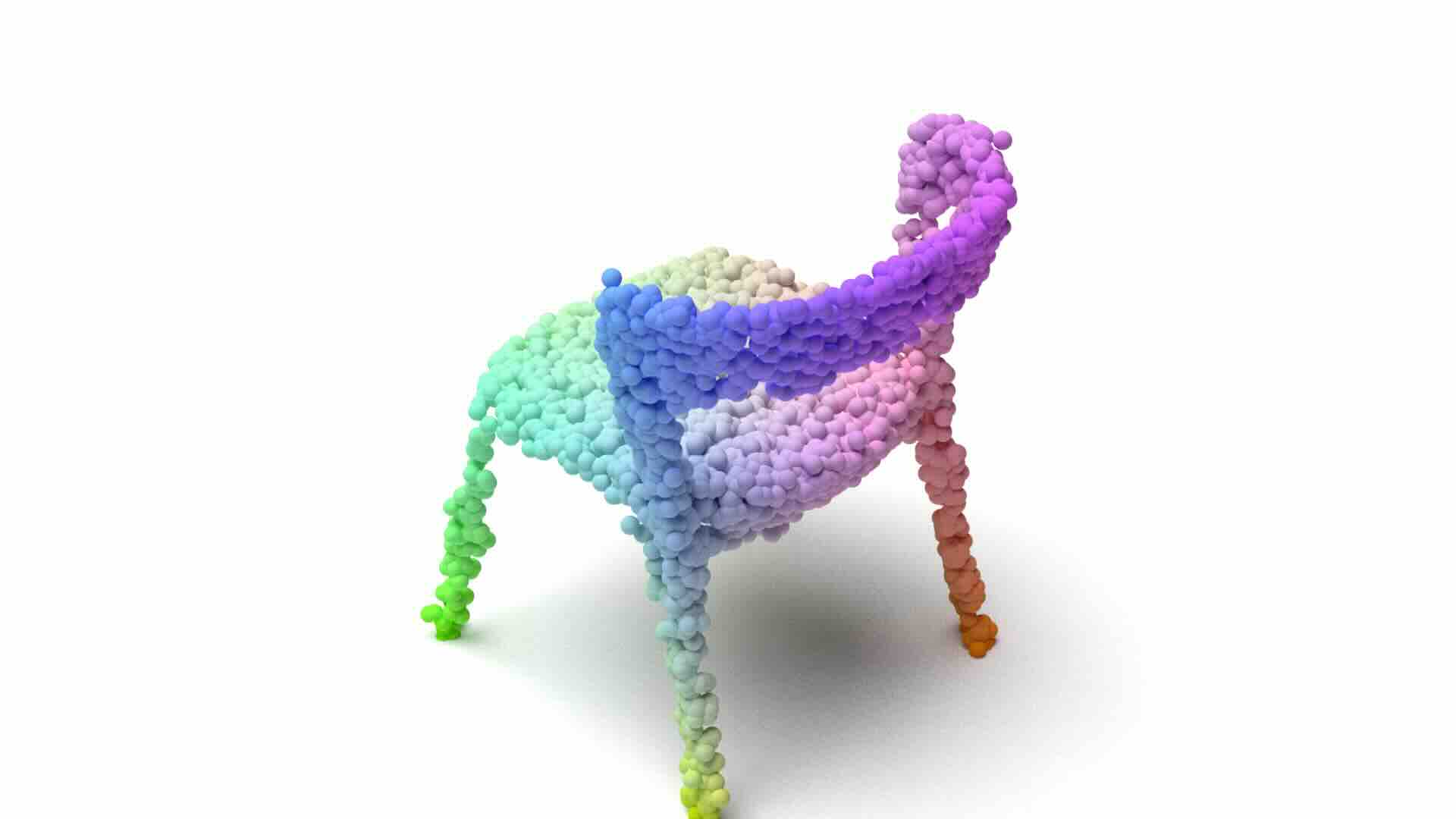} &
        \includegraphics[trim={15cm 0.0cm 15cm 0.0cm},clip,width=0.12\textwidth]{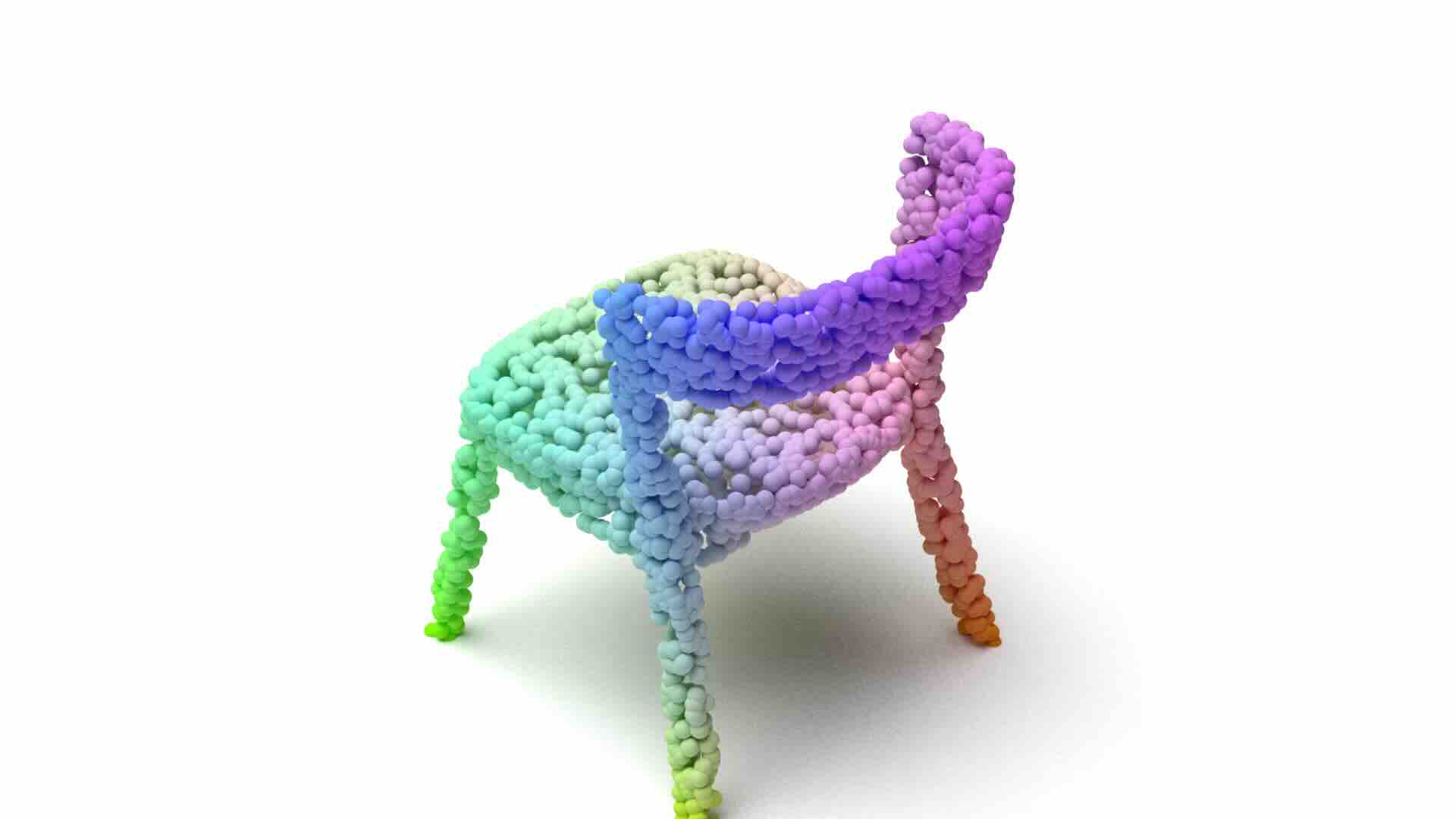} \\

        \includegraphics[trim={15cm 0.0cm 15cm 0.0cm},clip,width=0.12\textwidth]{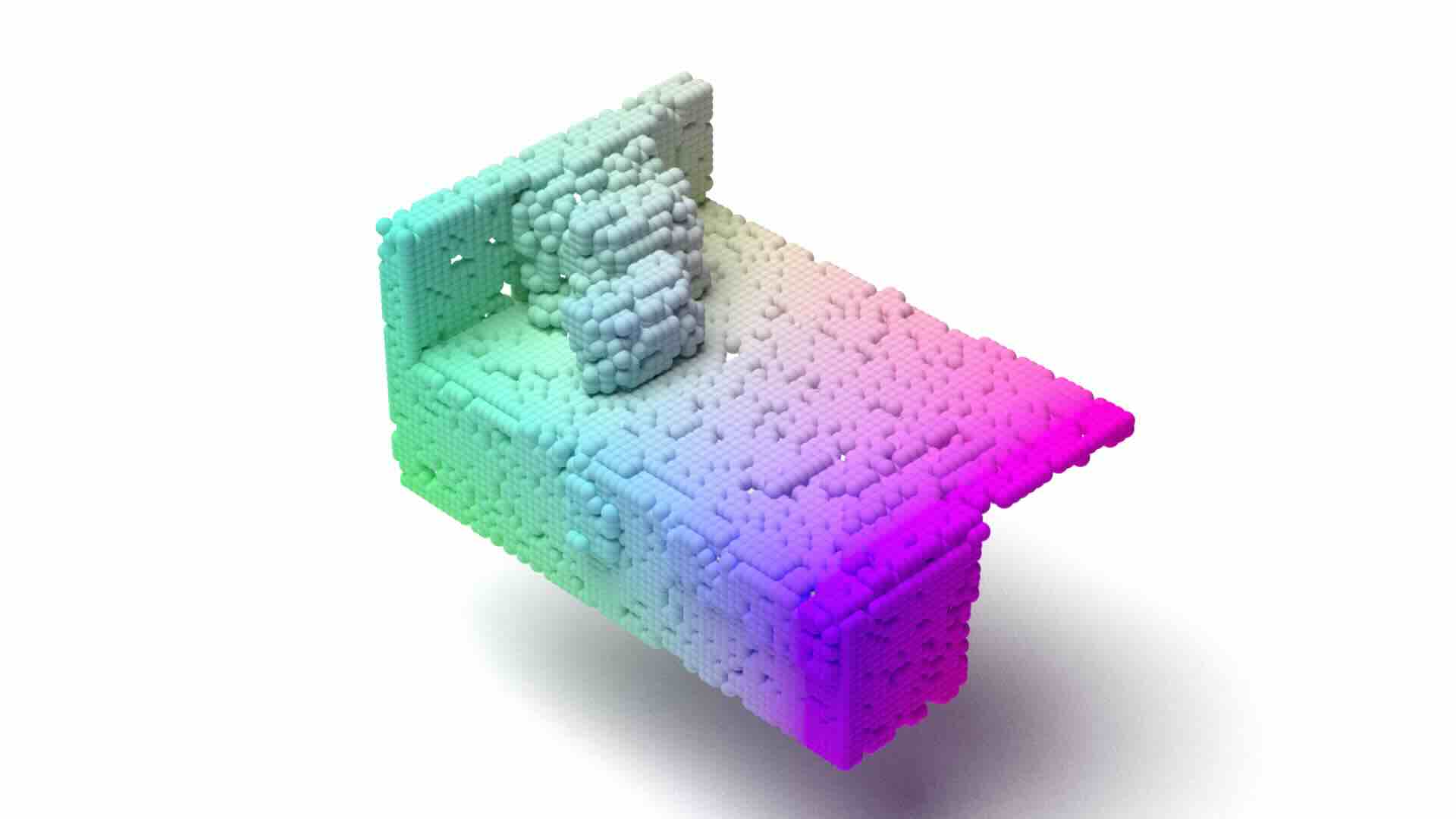} &
        
        \includegraphics[trim={15cm 0.0cm 15cm 0.0cm},clip,width=0.12\textwidth]{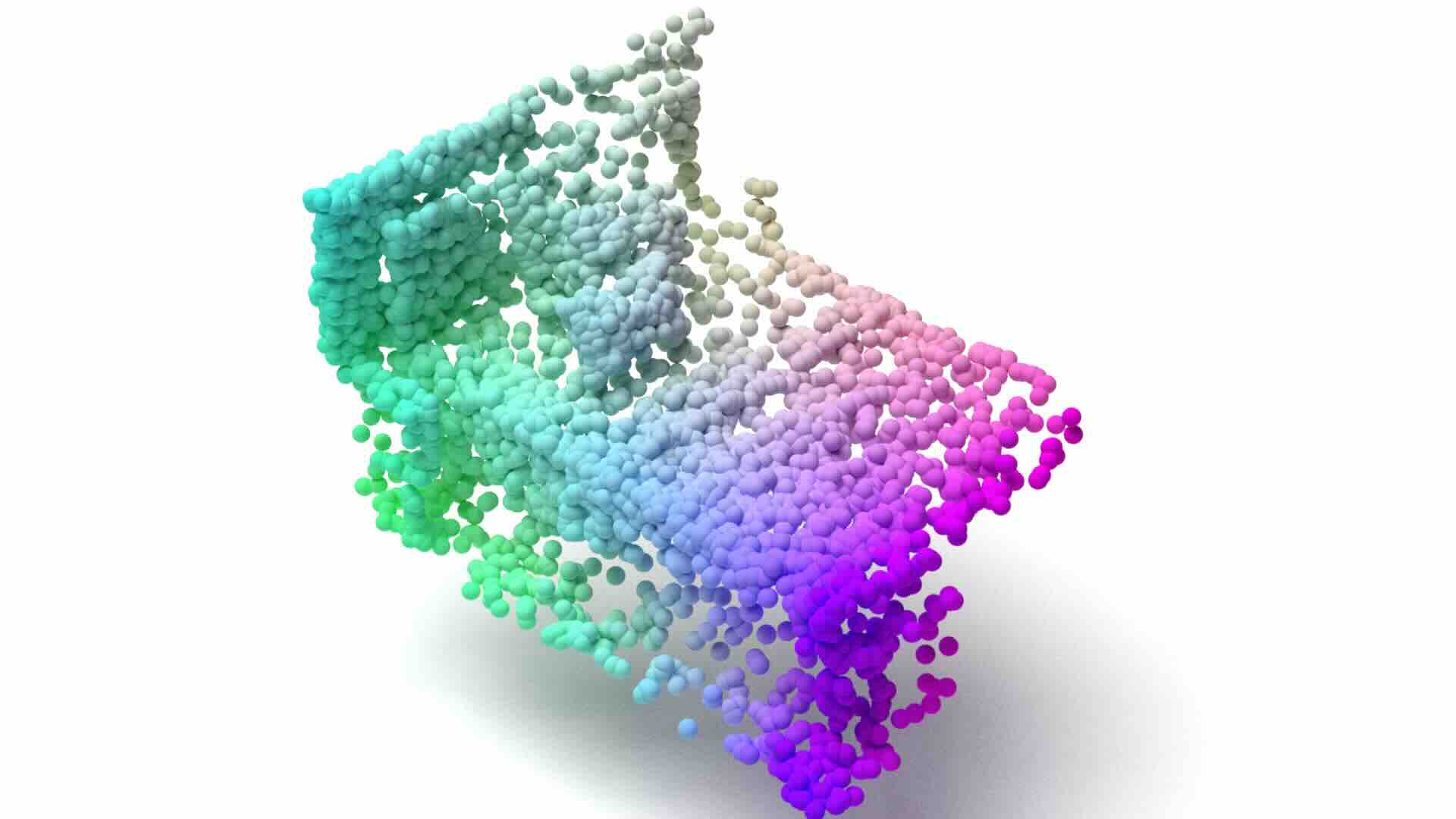} &
        \includegraphics[trim={15cm 0.0cm 15cm 0.0cm},clip, width=0.12\textwidth]{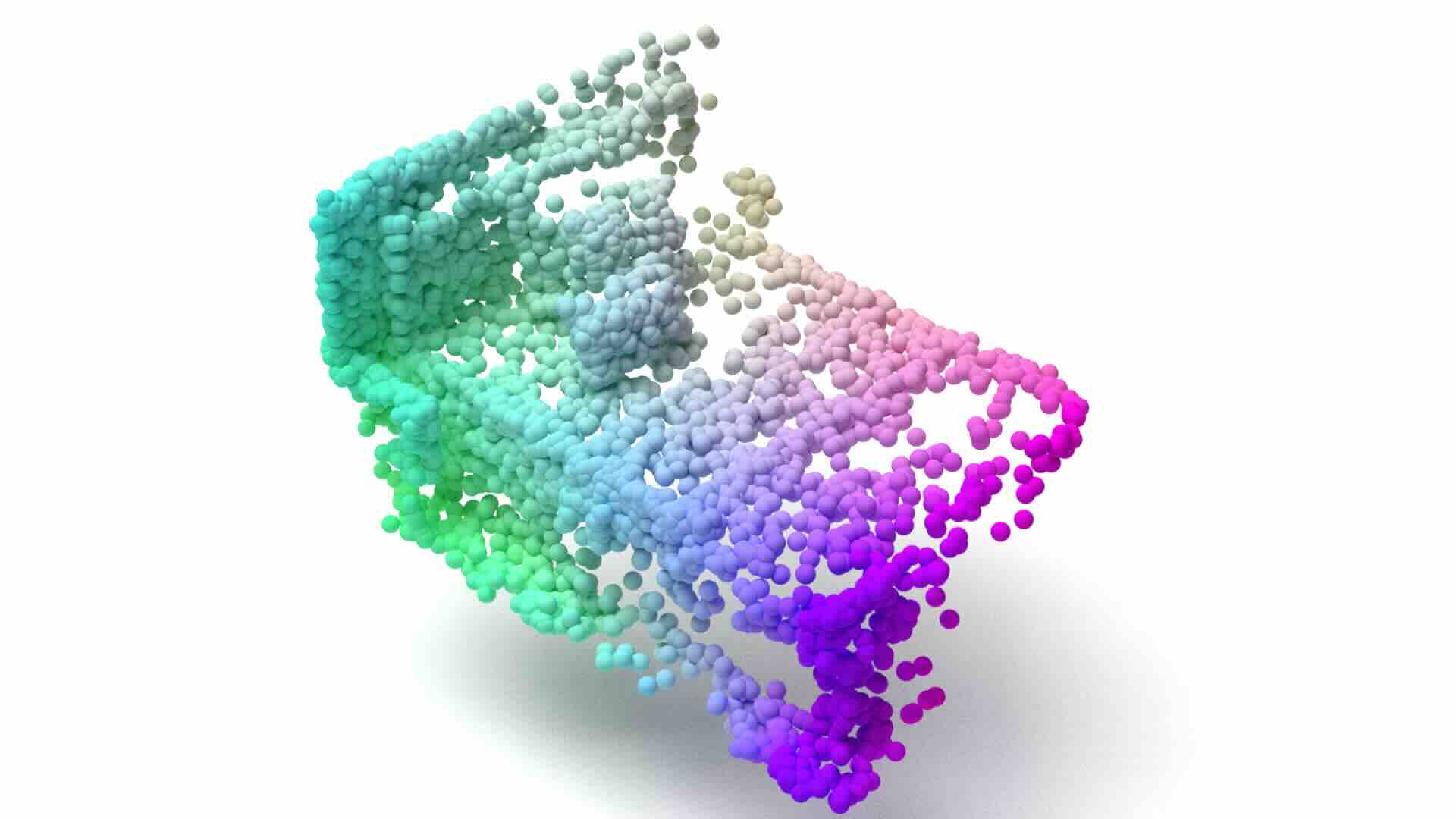} &
        \includegraphics[trim={15cm 0.0cm 15cm 0.0cm},clip, width=0.12\textwidth]{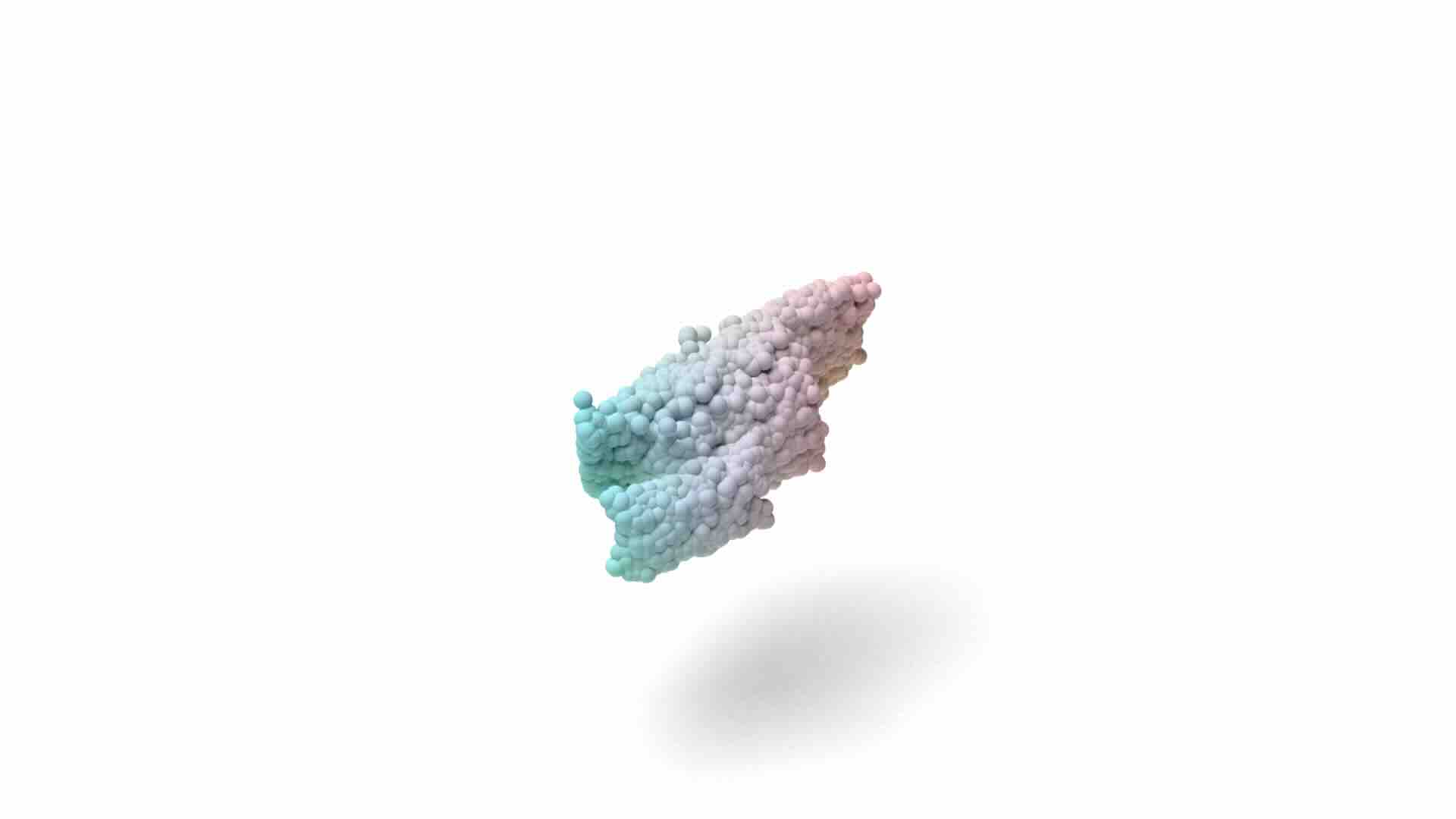} &
        \includegraphics[trim={15cm 0.0cm 15cm 0.0cm},clip, width=0.12\textwidth]{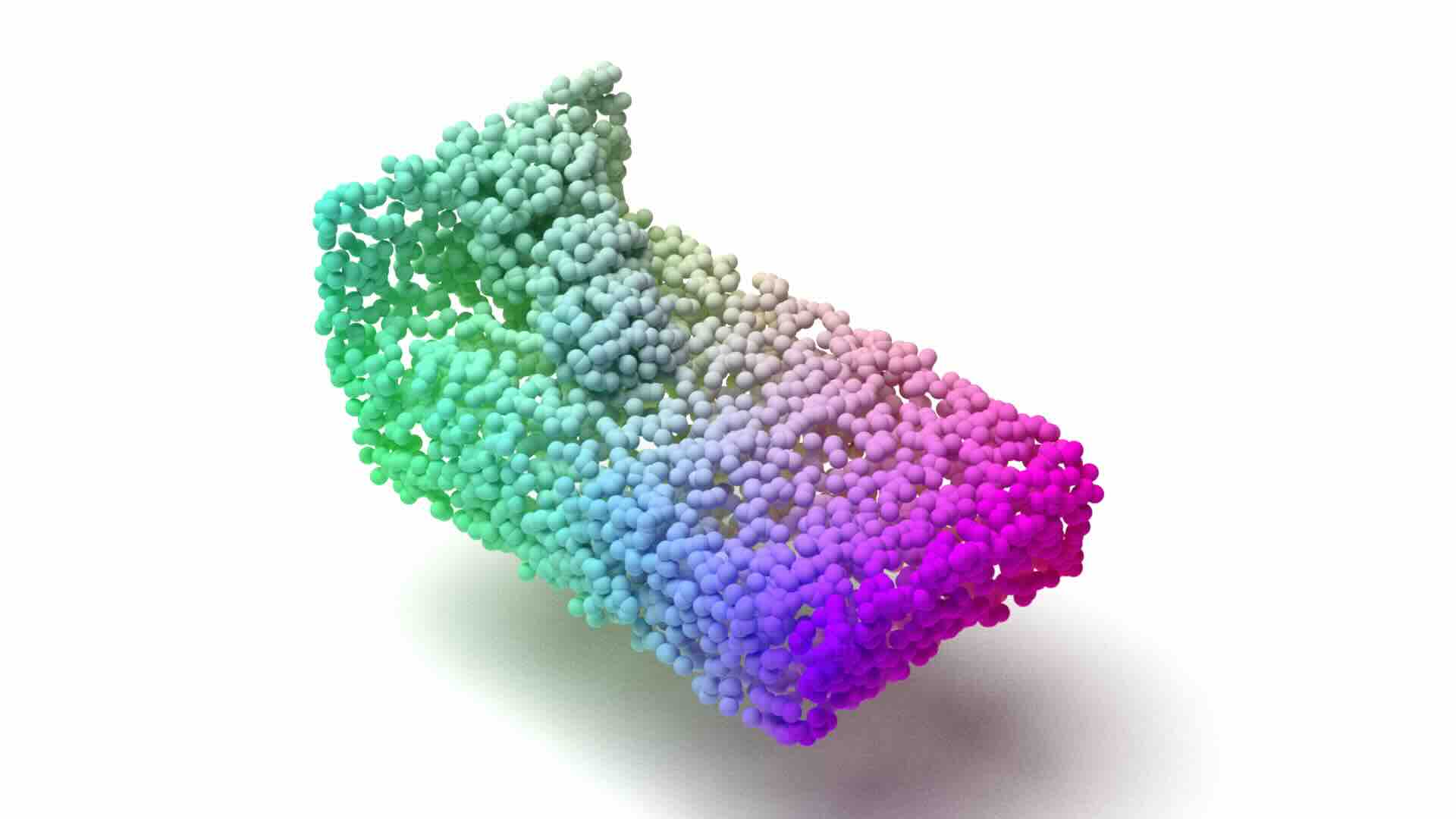}
        &
        \includegraphics[trim={15cm 0.0cm 15cm 0.0cm},clip, width=0.12\textwidth]{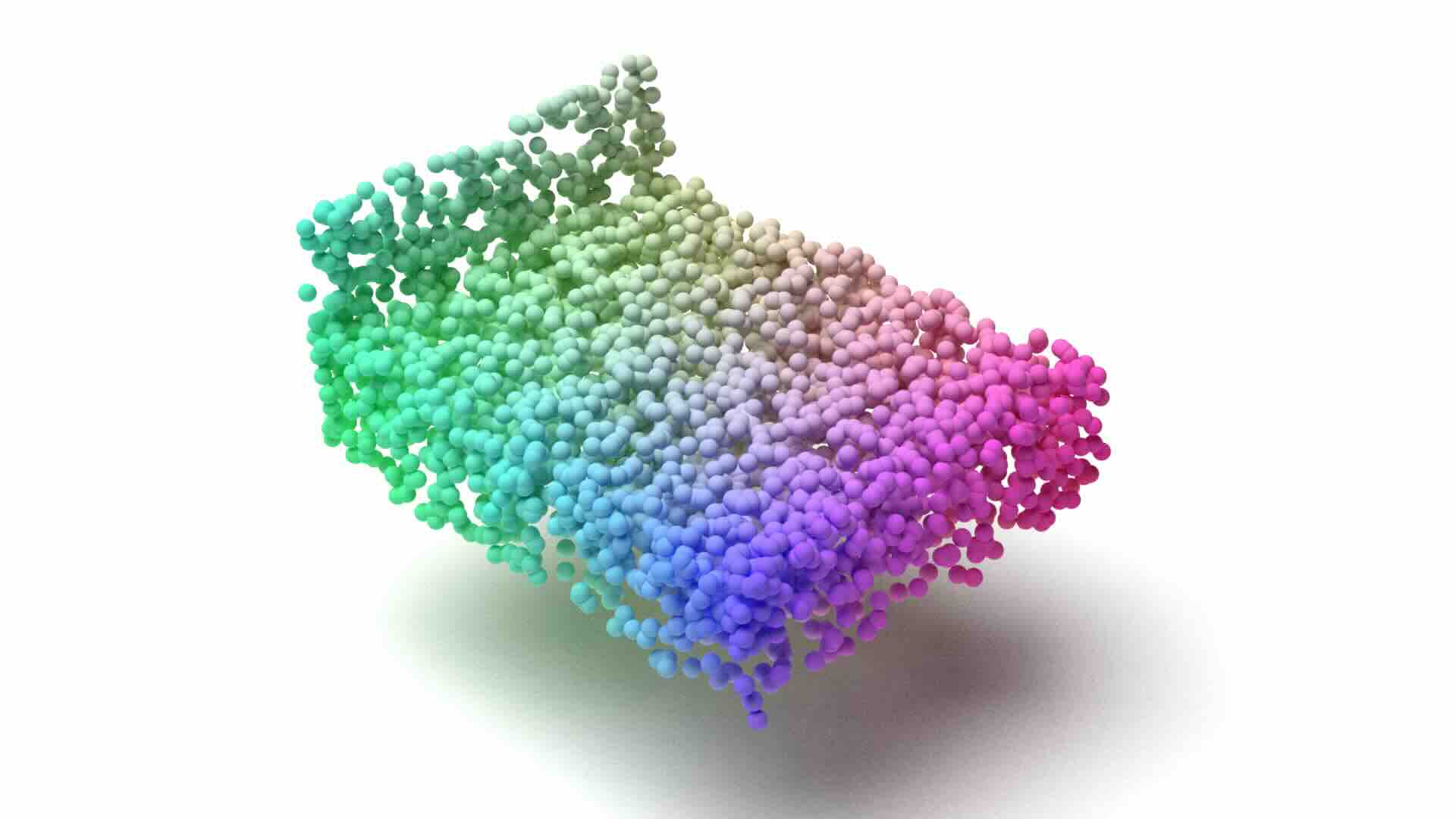}
        &
        \includegraphics[trim={15cm 0.0cm 15cm 0.0cm},clip, width=0.12\textwidth]{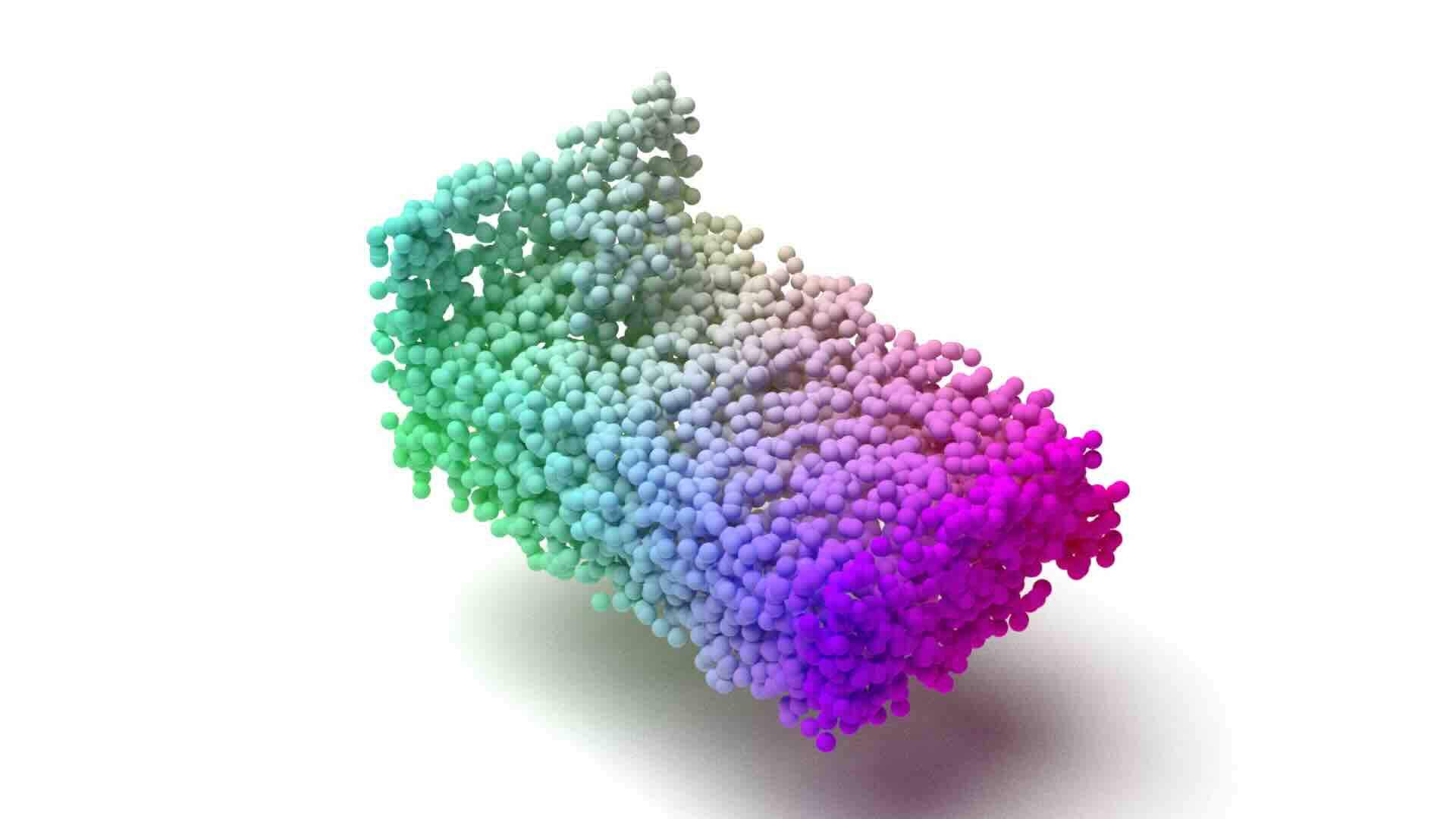} &
        \includegraphics[trim={15cm 0.0cm 15cm 0.0cm},clip,width=0.12\textwidth]{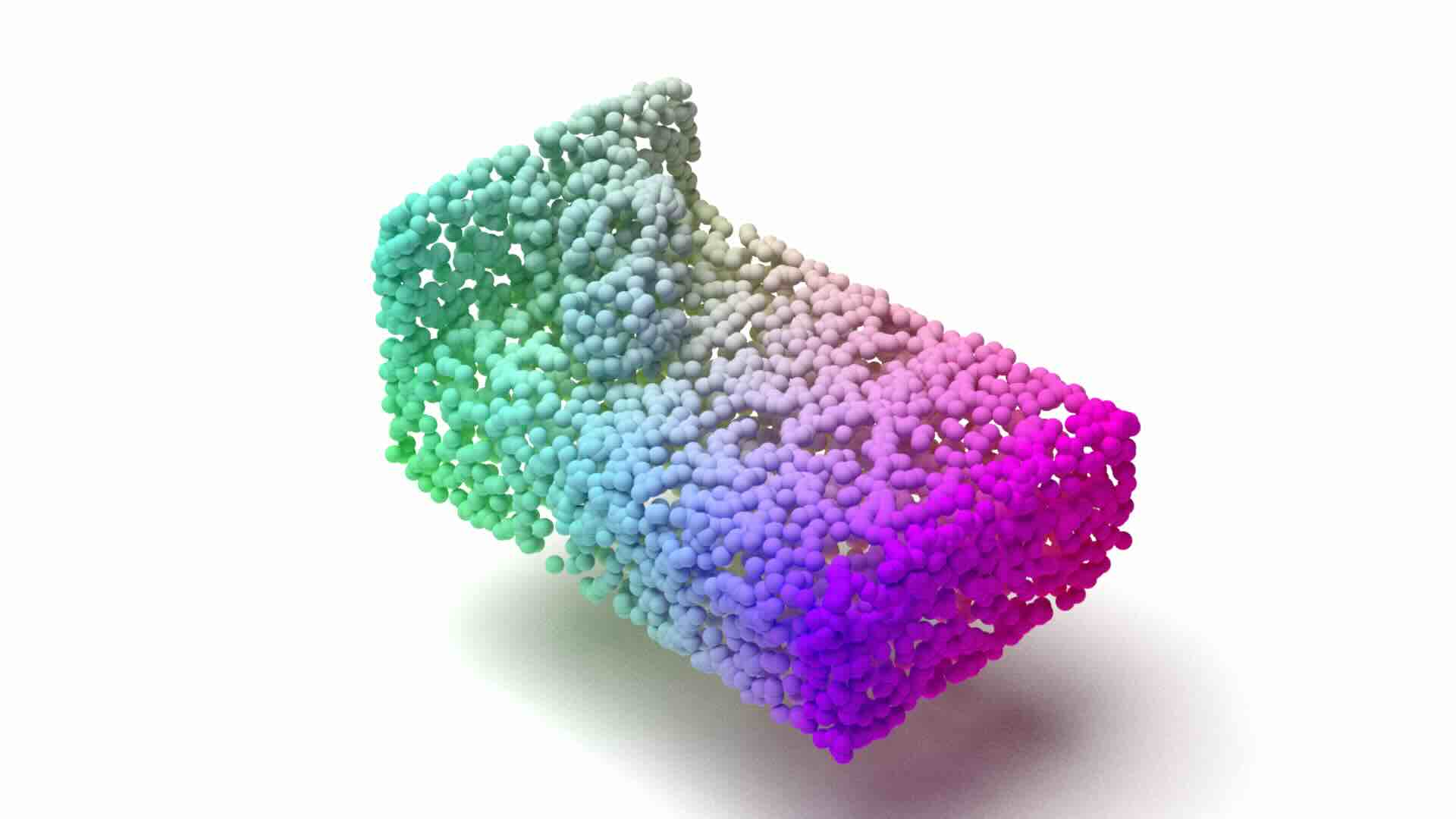} \\
        \includegraphics[trim={15cm 0.0cm 15cm 0.0cm},clip,width=0.12\textwidth]{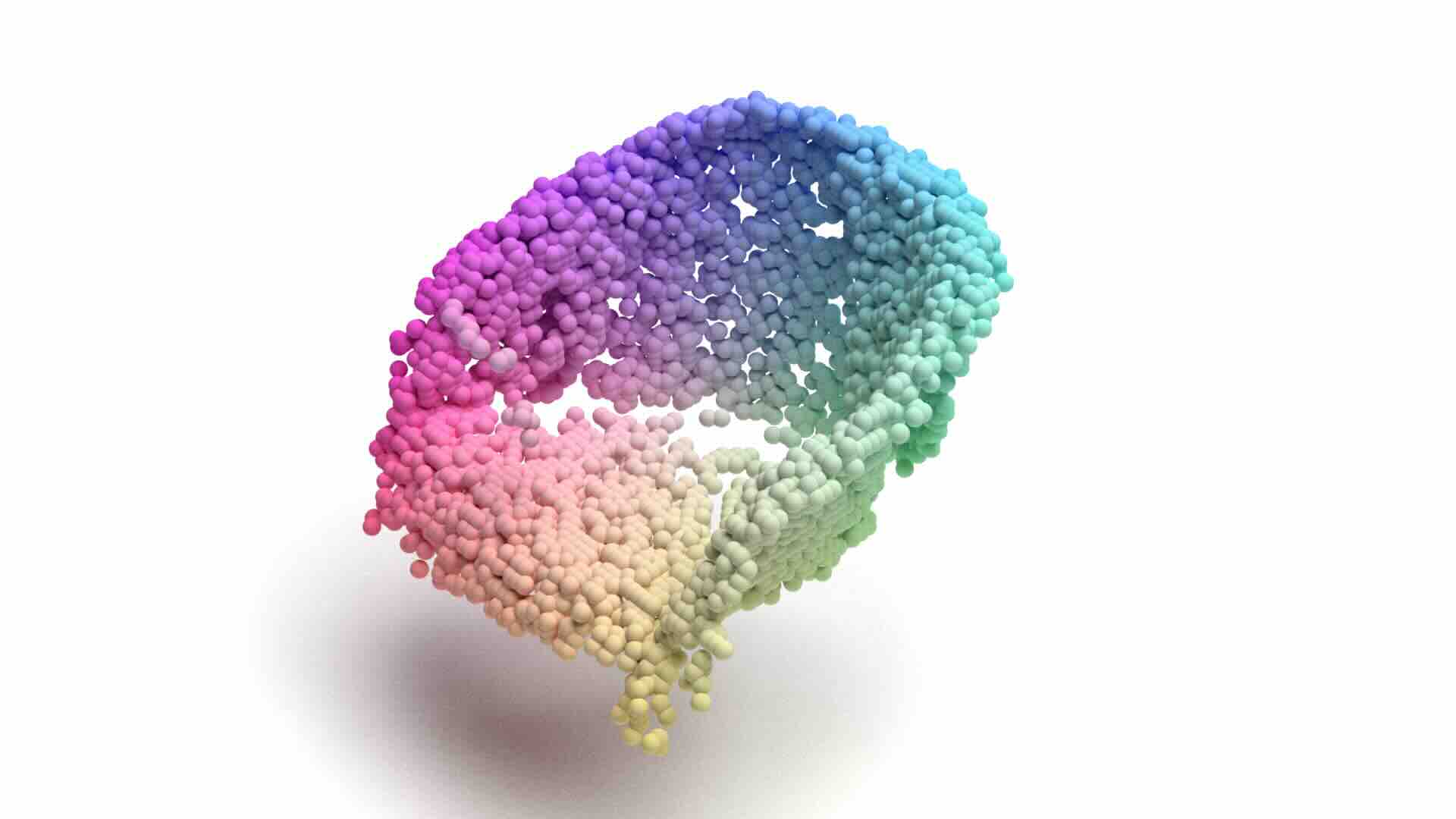} &
        \includegraphics[trim={15cm 0.0cm 15cm 0.0cm},clip, width=0.12\textwidth]{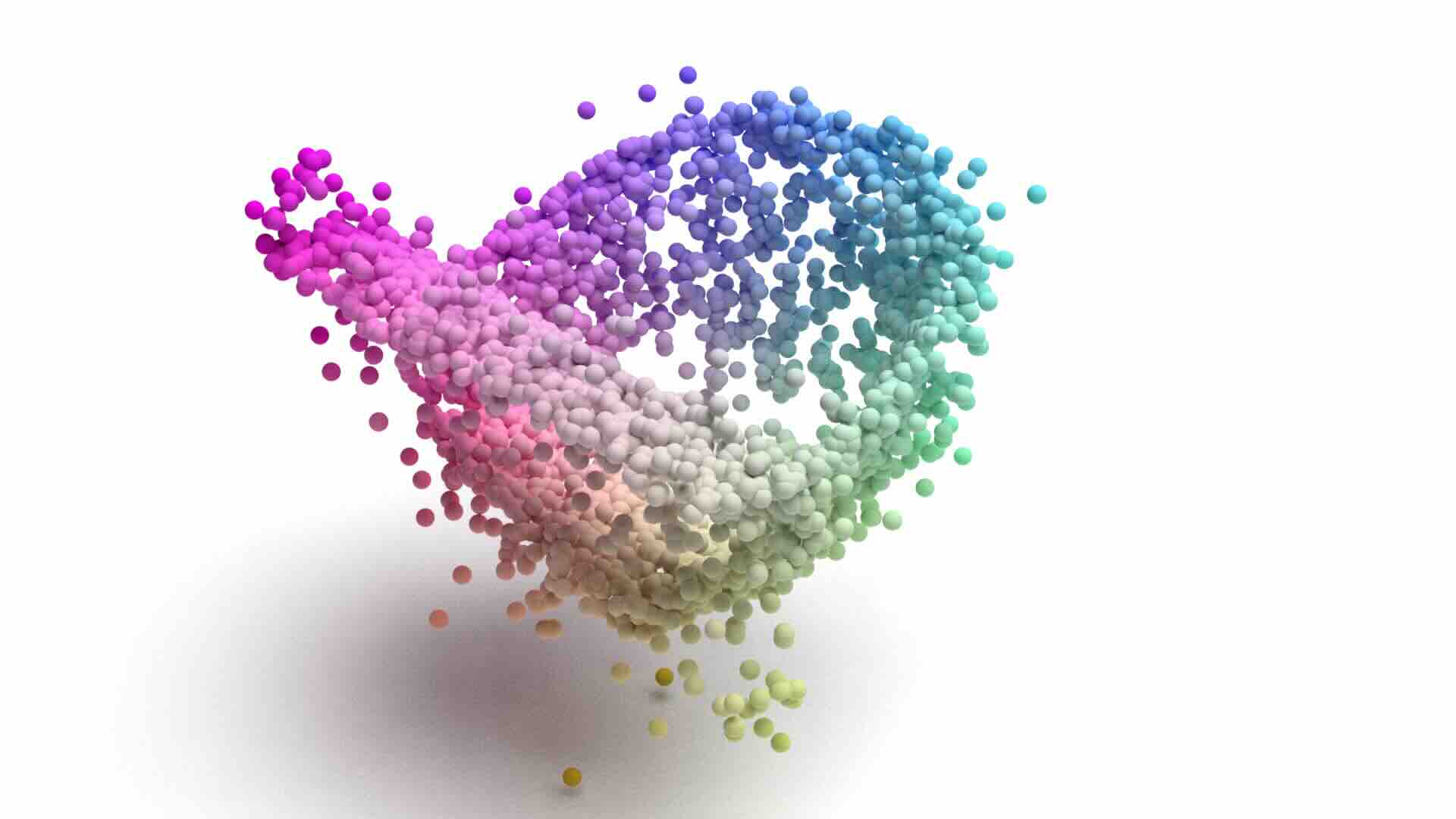} &
        \includegraphics[trim={15cm 0.0cm 15cm 0.0cm},clip, width=0.12\textwidth]{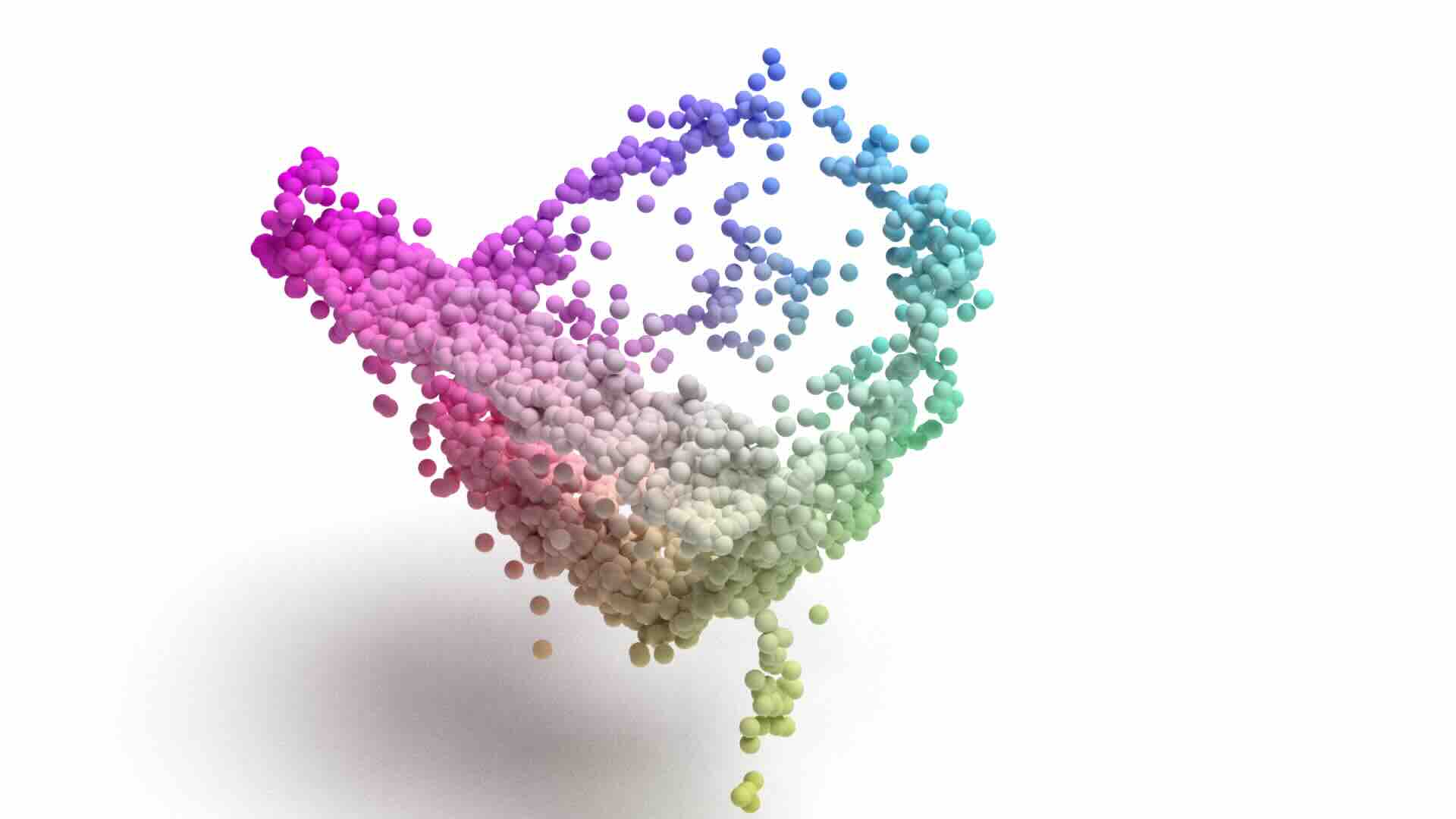} &
        \includegraphics[trim={15cm 0.0cm 15cm 0.0cm},clip, width=0.12\textwidth]{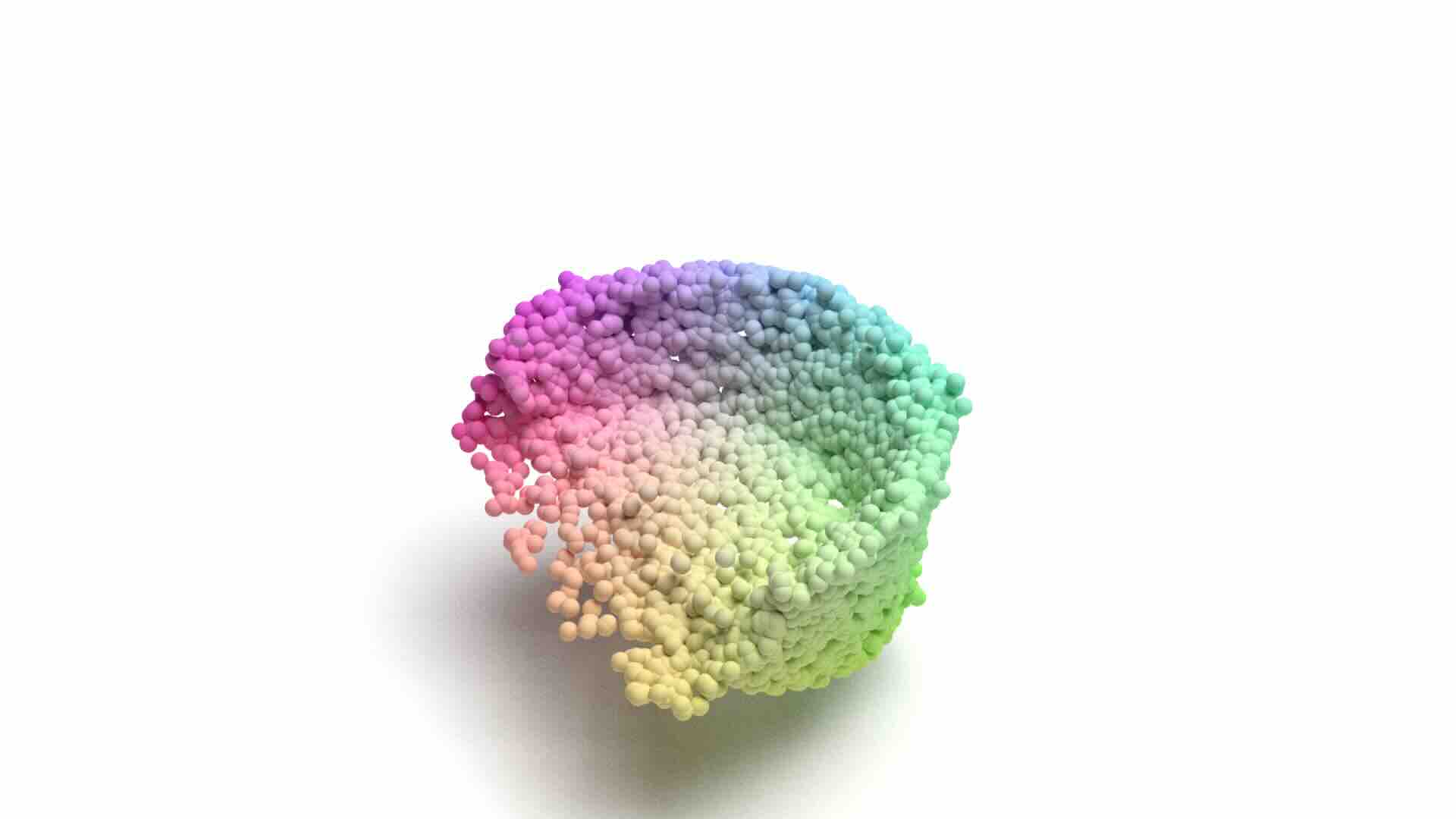} &
        \includegraphics[trim={15cm 0.0cm 15cm 0.0cm},clip, width=0.12\textwidth]{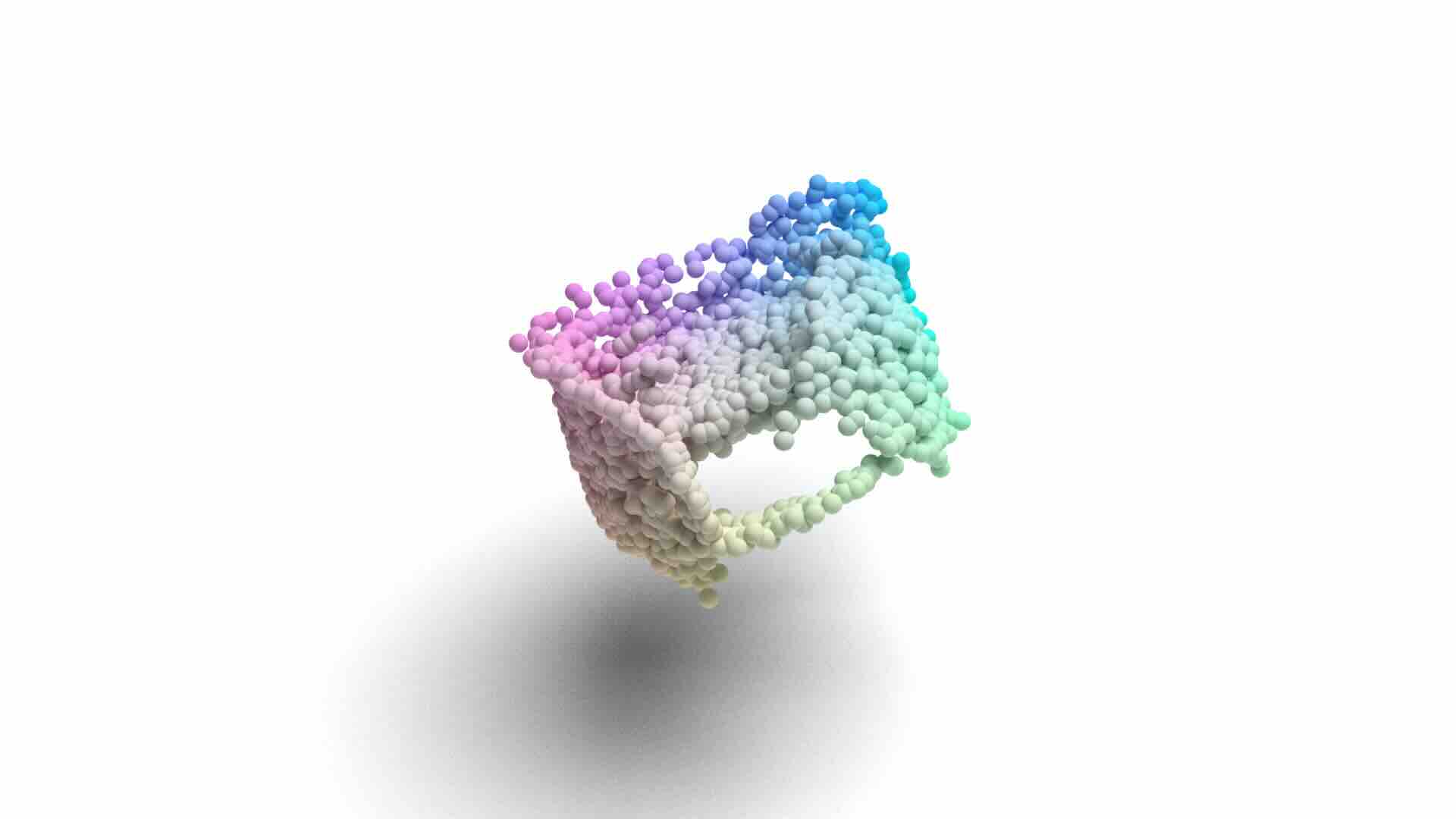} 
        &
        \includegraphics[trim={15cm 0.0cm 15cm 0.0cm},clip, width=0.12\textwidth]{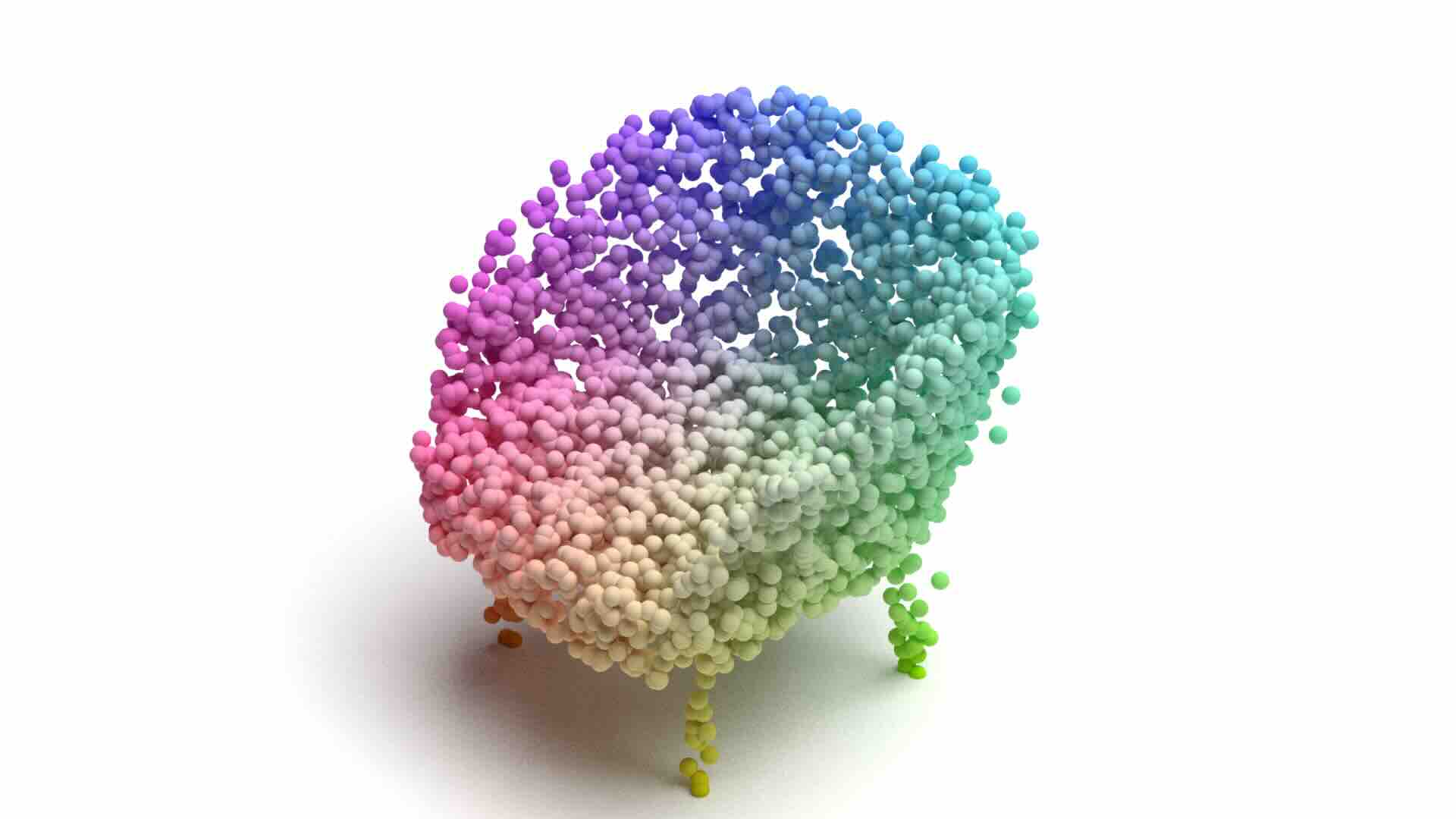} 
        &
        \includegraphics[trim={15cm 0.0cm 15cm 0.0cm},clip,width=0.12\textwidth]{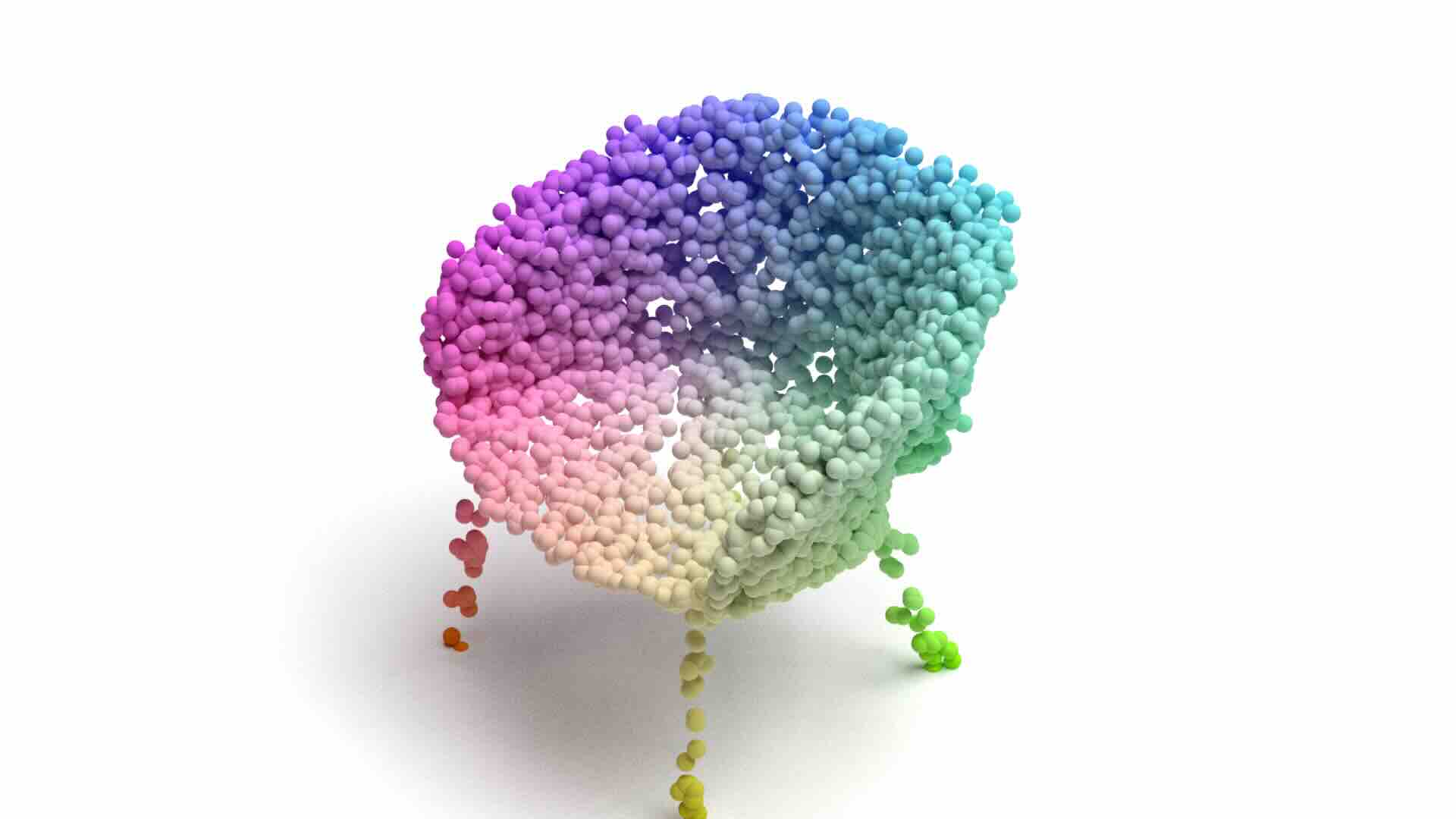} &
        \includegraphics[trim={15cm 0.0cm 15cm 0.0cm},clip,width=0.12\textwidth]{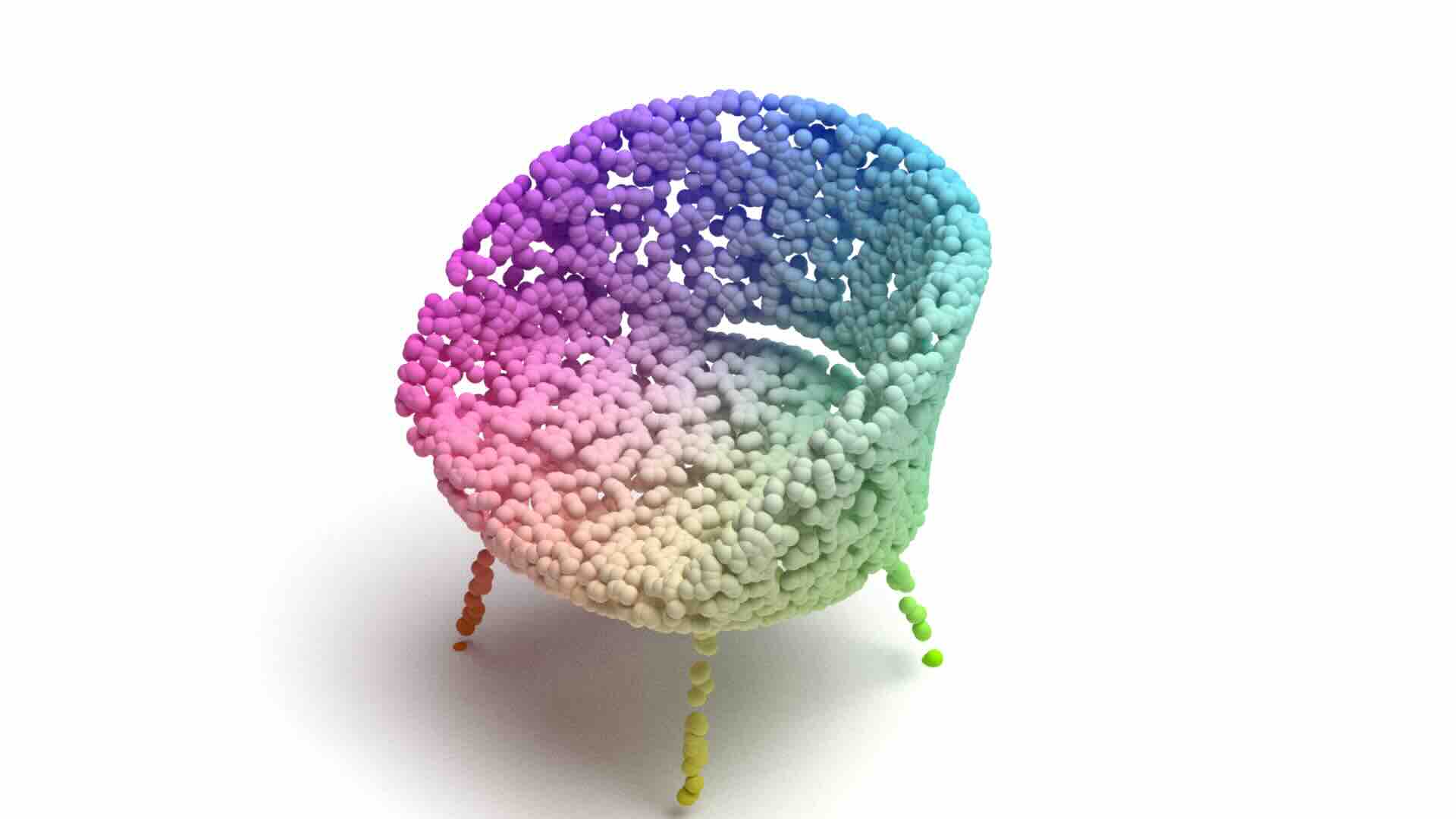}
        \\
        \includegraphics[trim={15cm 0.0cm 15cm 0.0cm},clip,width=0.12\textwidth]{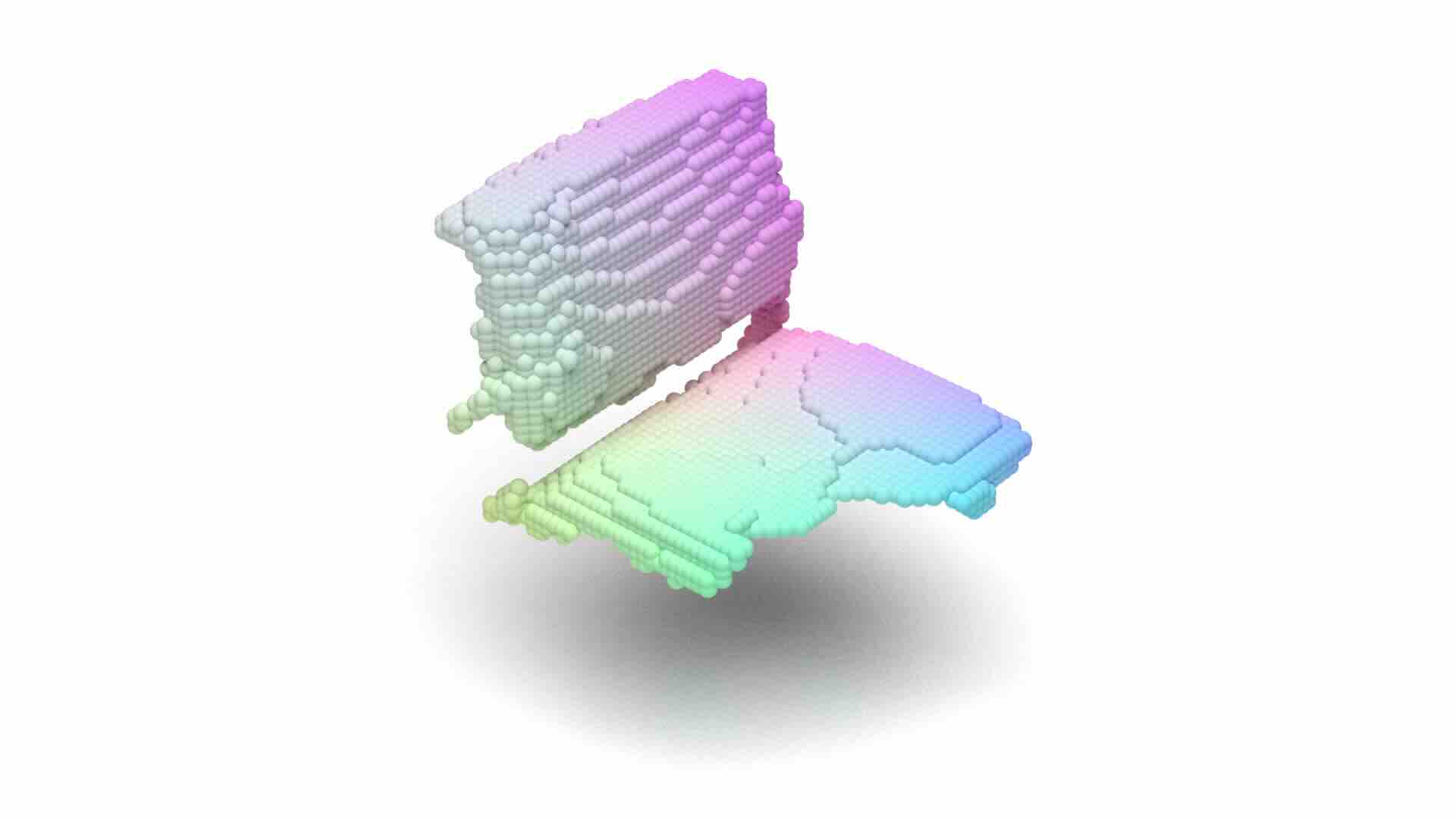} &
        \includegraphics[trim={15cm 0.0cm 15cm 0.0cm},clip, width=0.12\textwidth]{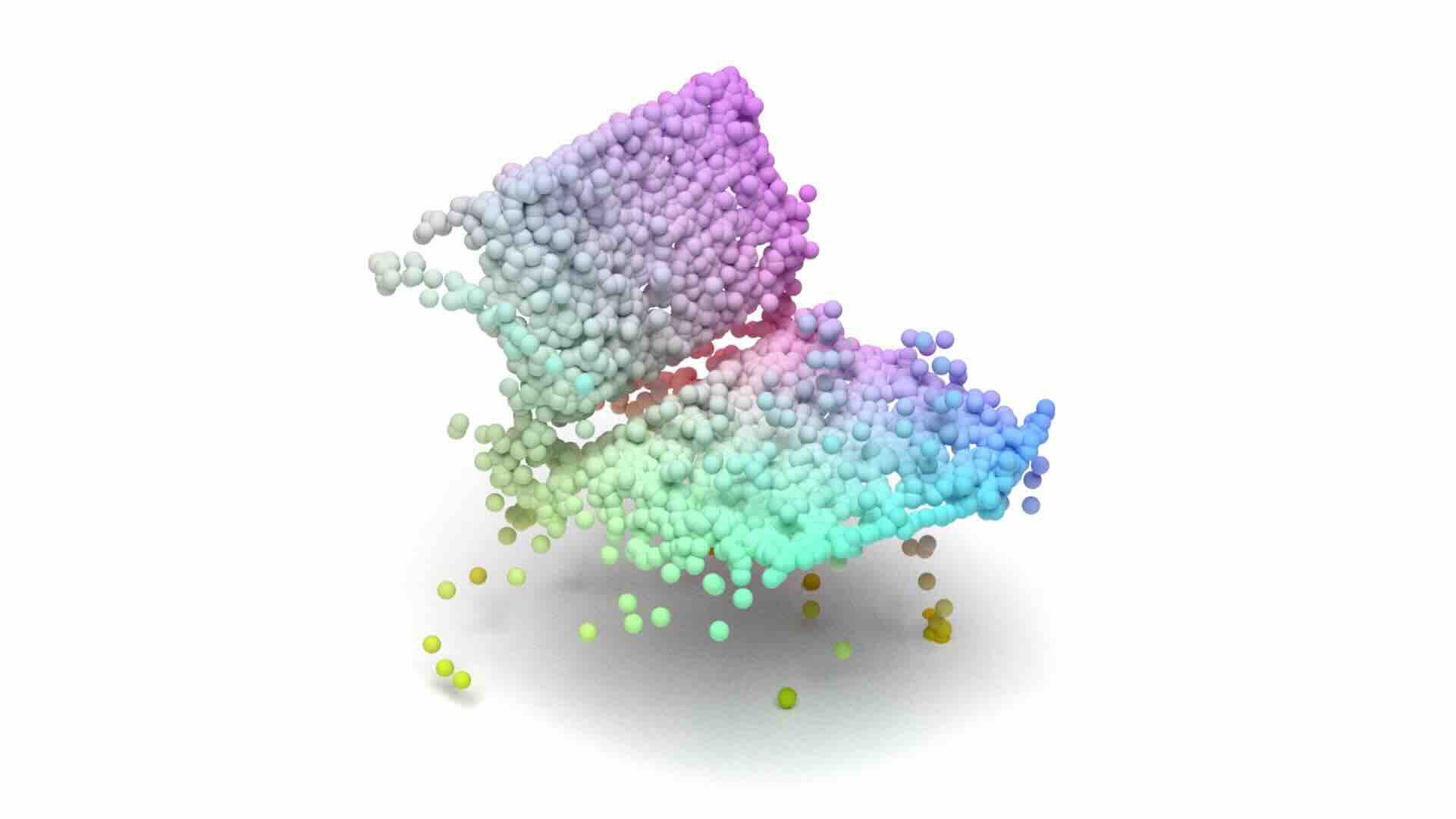} &
        \includegraphics[trim={15cm 0.0cm 15cm 0.0cm},clip, width=0.12\textwidth]{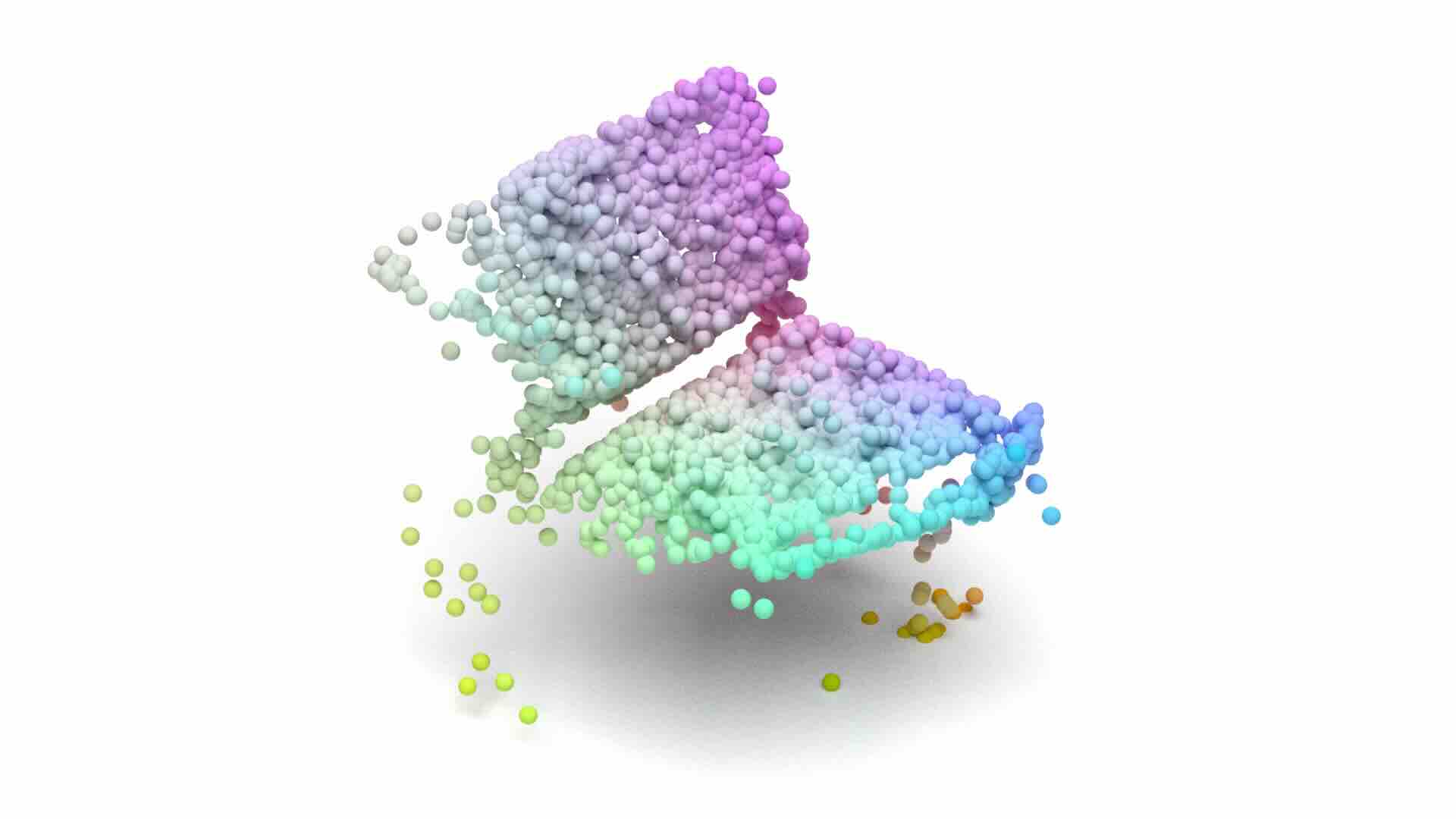} &
        \includegraphics[trim={15cm 0.0cm 15cm 0.0cm},clip, width=0.12\textwidth]{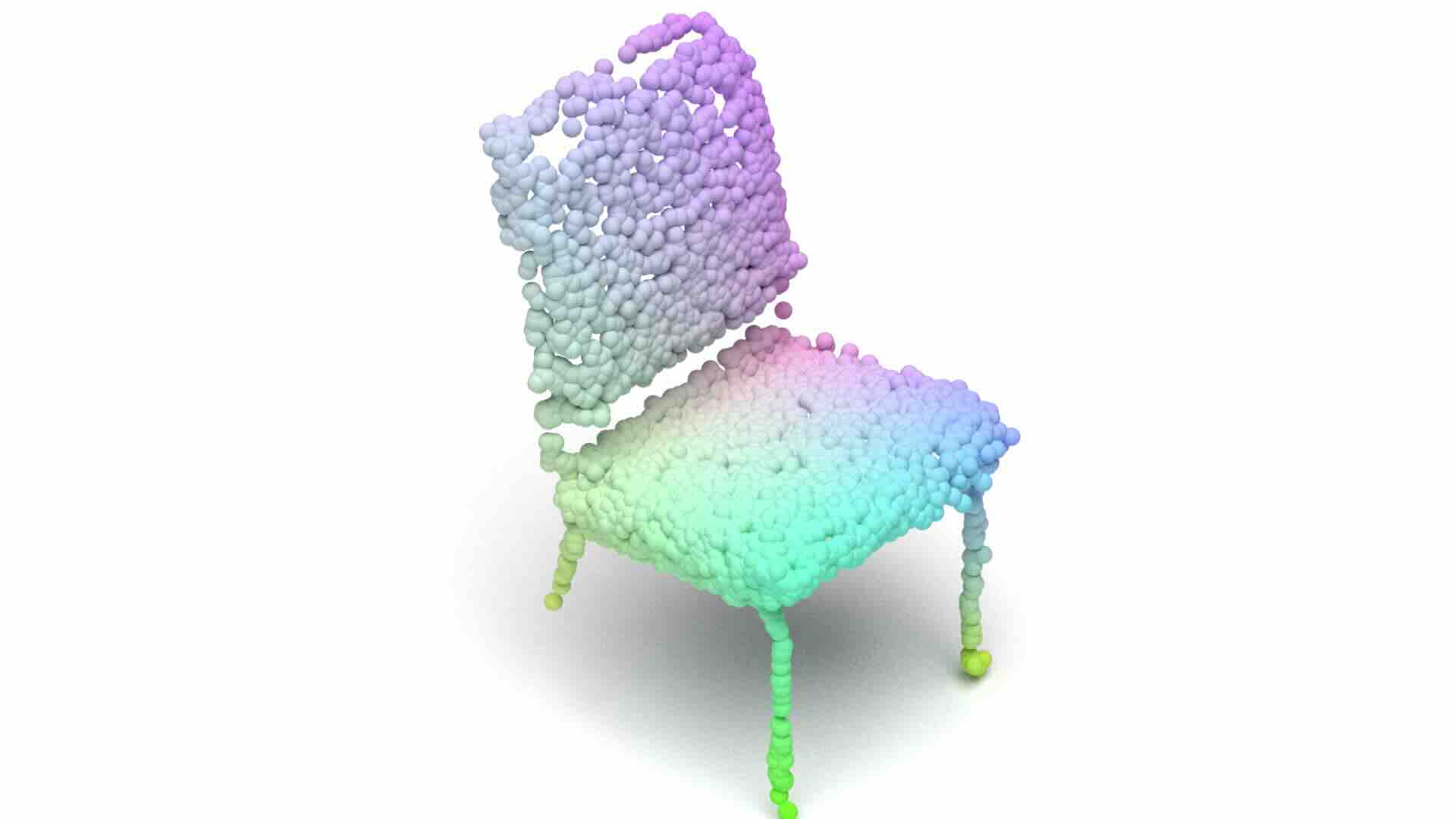} 
        &
        \includegraphics[trim={15cm 0.0cm 15cm 0.0cm},clip, width=0.12\textwidth]{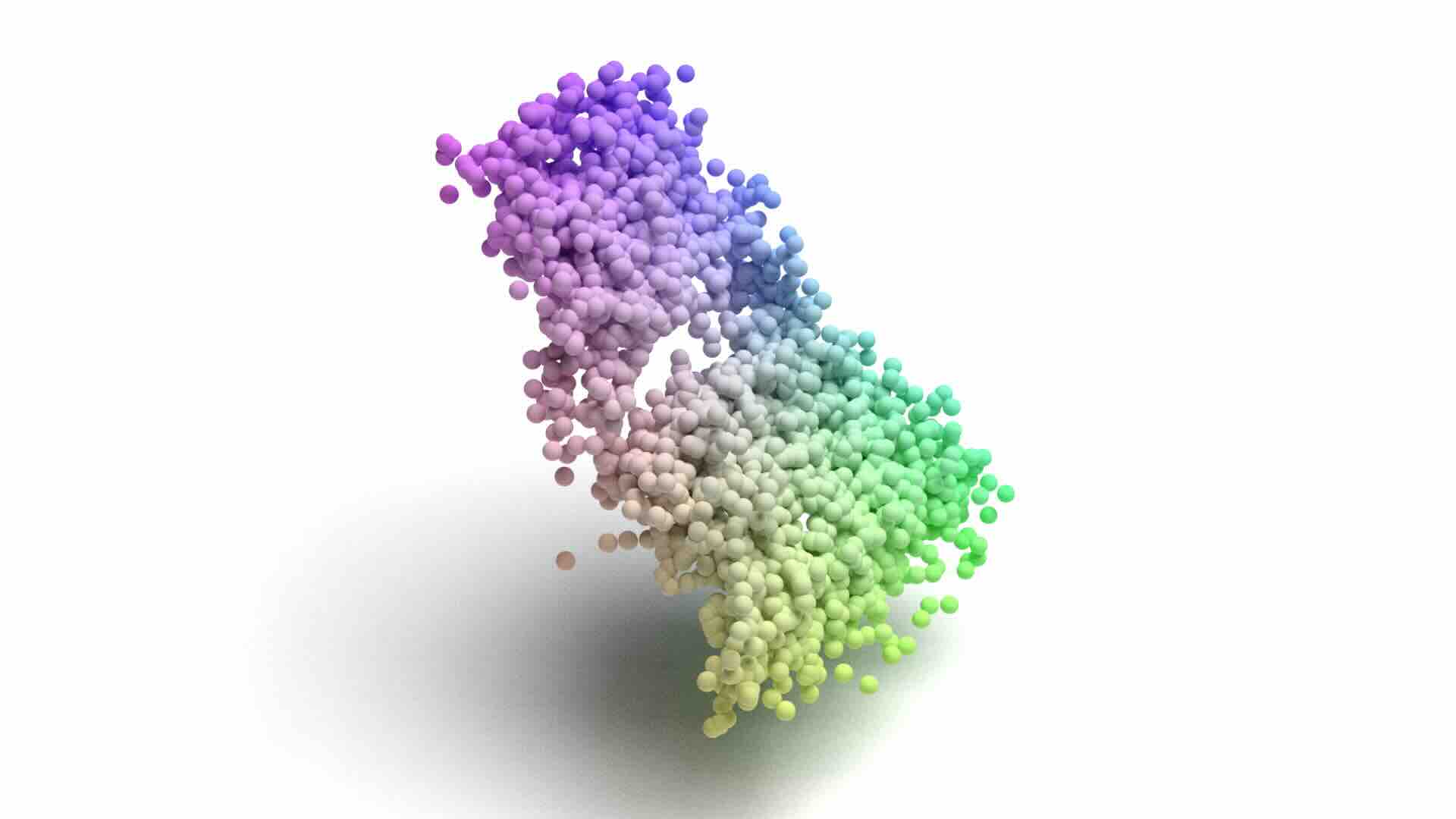}
        &
        \includegraphics[trim={15cm 0.0cm 15cm 0.0cm},clip, width=0.12\textwidth]{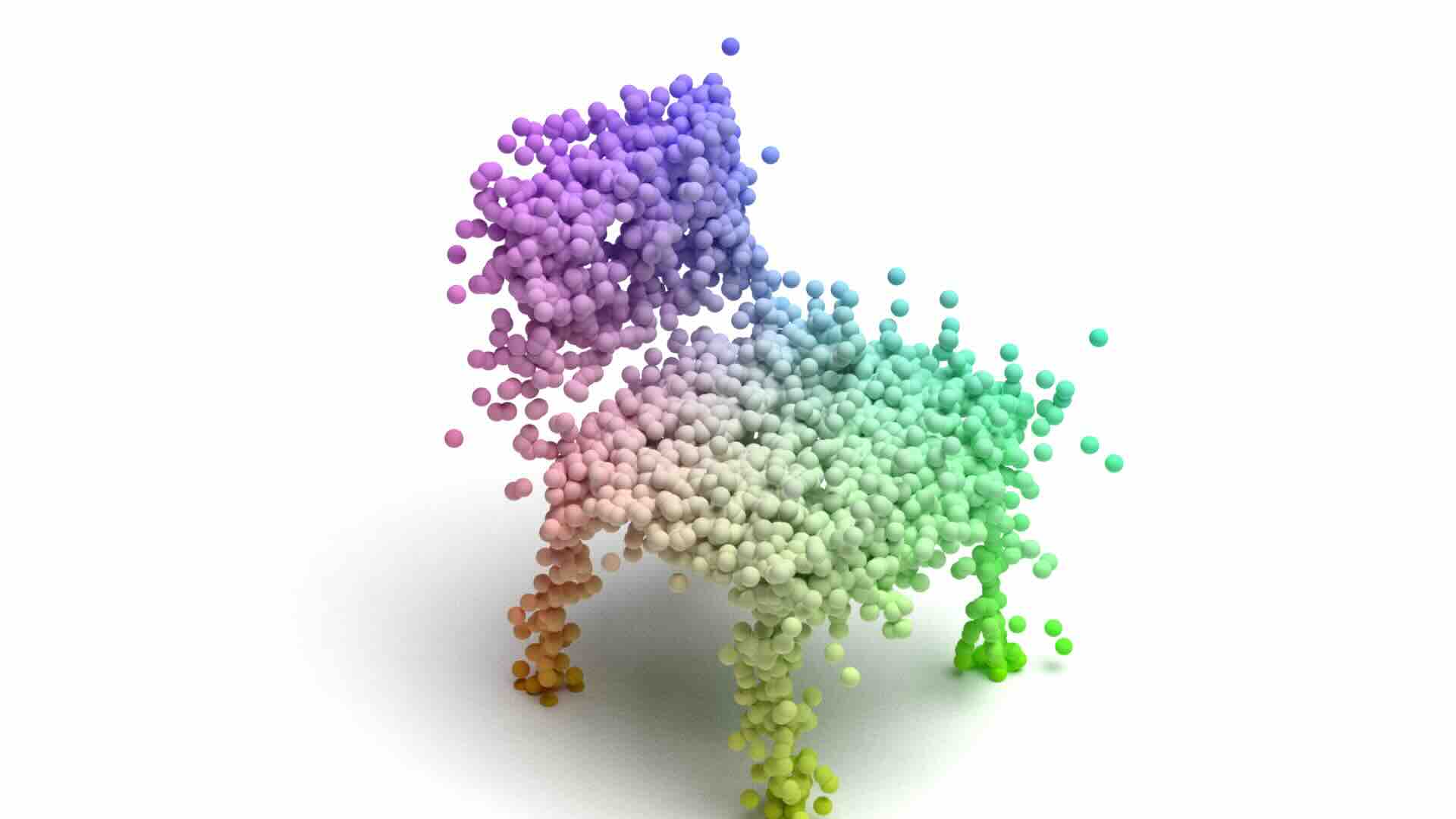}&
        \includegraphics[trim={15cm 0.0cm 15cm 0.0cm},clip, width=0.12\textwidth]{figures/scannetpp_pts/15_ours.jpg} &
        \includegraphics[trim={15cm 0.0cm 15cm 0.0cm},clip,width=0.12\textwidth]{figures/scannetpp_pts/15_gt.jpg} 
        \\
         \includegraphics[trim={15cm 0.0cm 15cm 0.0cm},clip,width=0.12\textwidth]{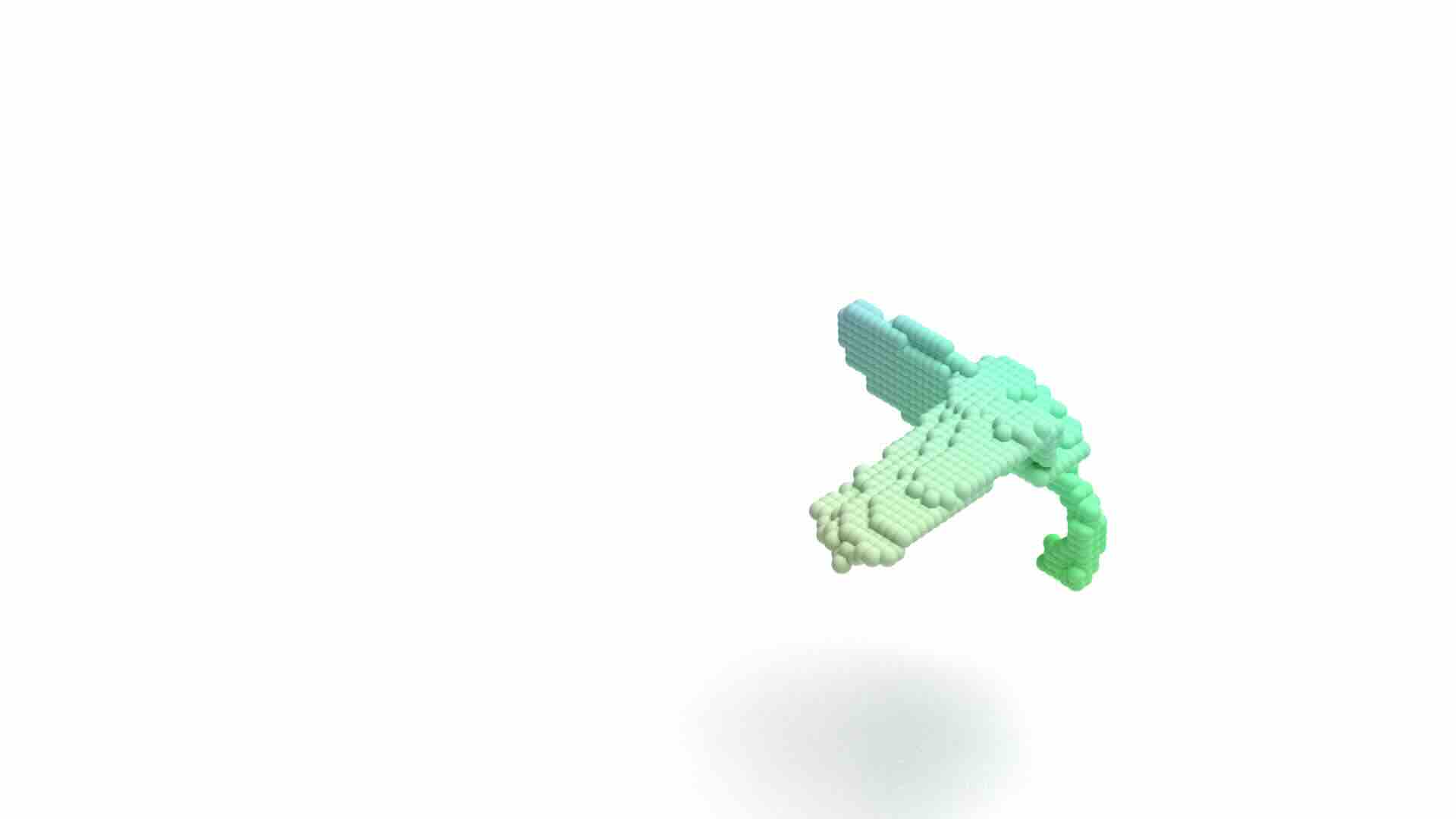} 
        &
        \includegraphics[trim={15cm 0.0cm 15cm 0.0cm},clip,width=0.12\textwidth]{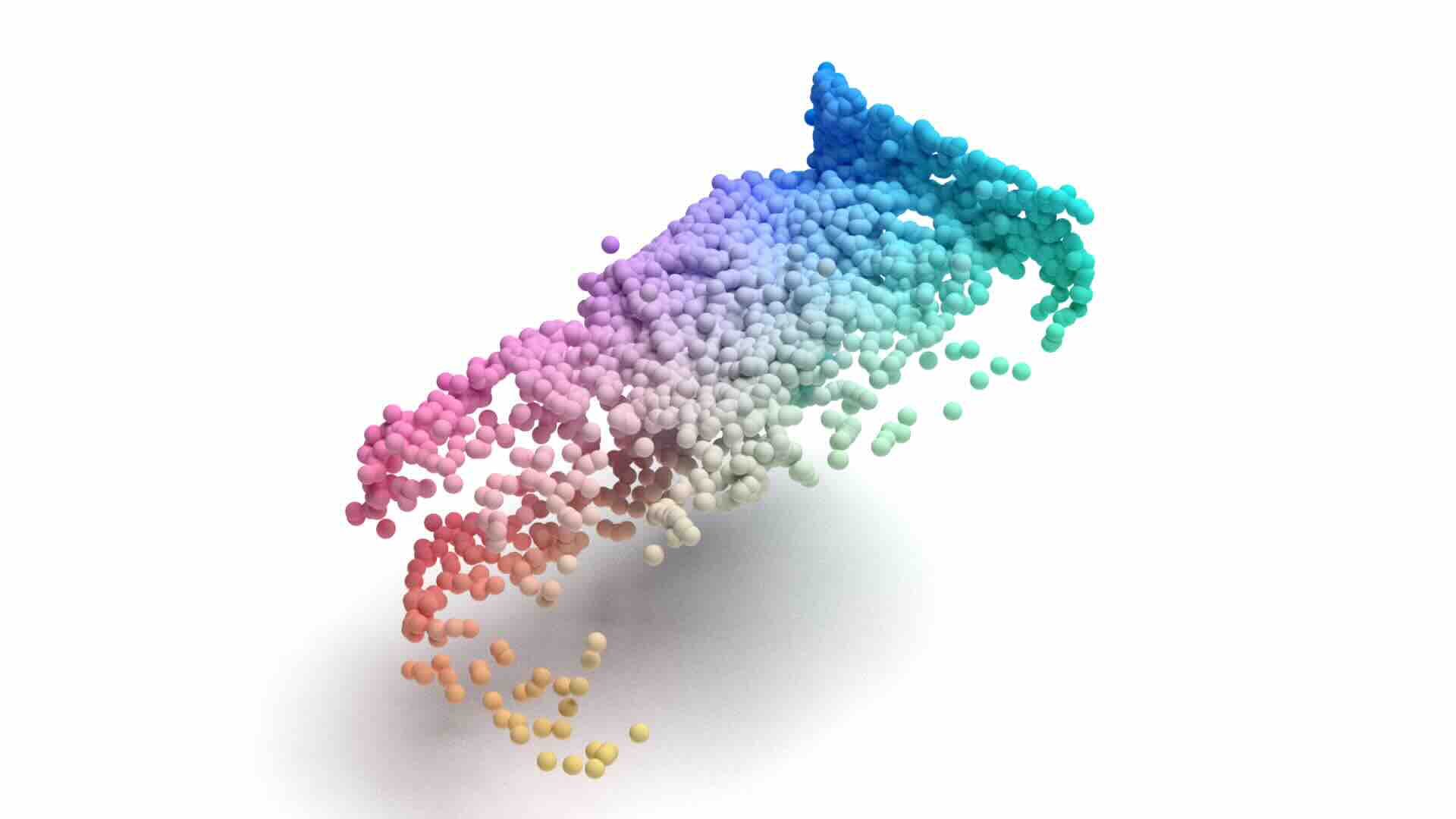} &
        \includegraphics[trim={15cm 0.0cm 15cm 0.0cm},clip, width=0.12\textwidth]{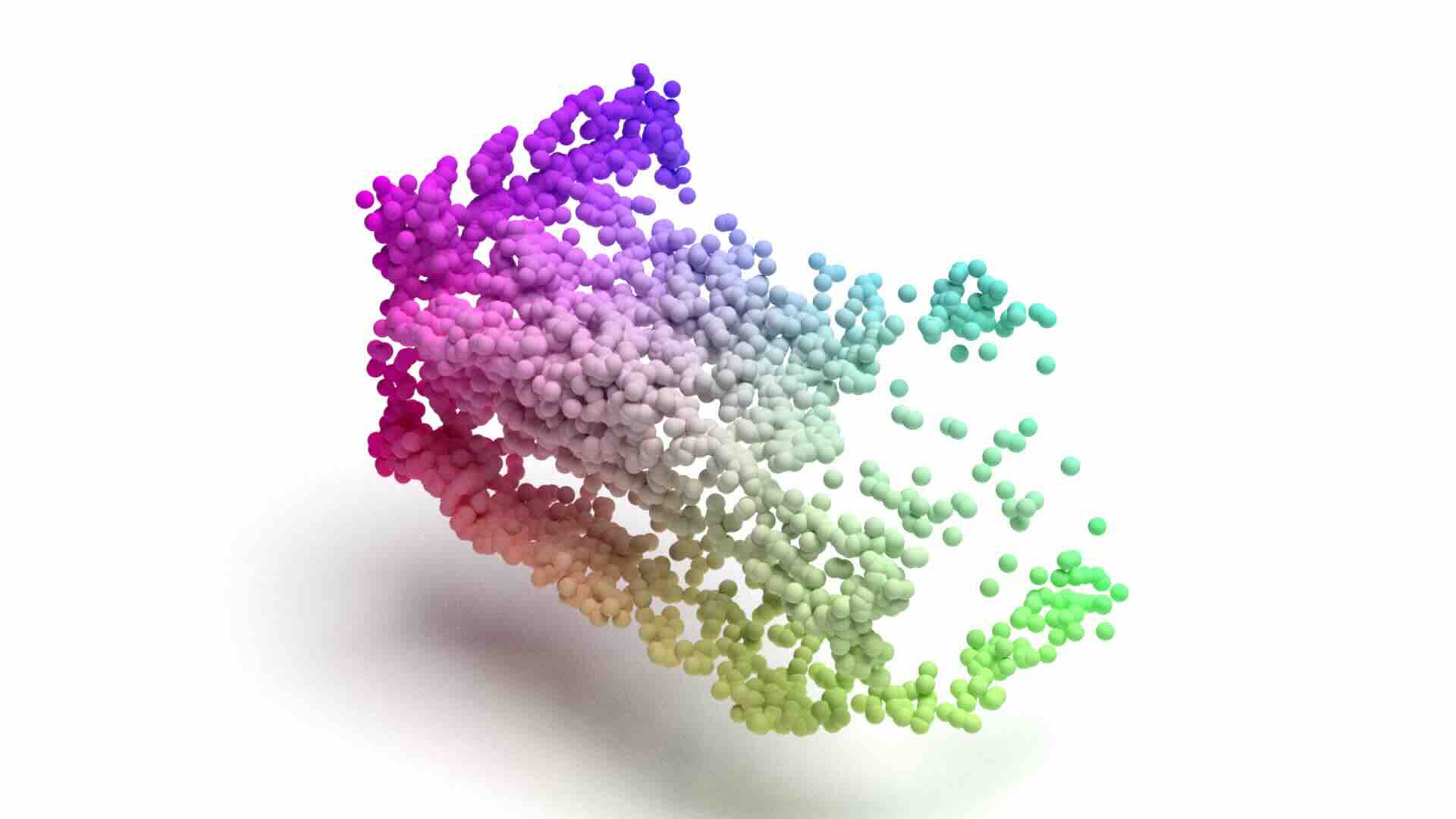} 
        &
        \includegraphics[trim={15cm 0.0cm 15cm 0.0cm},clip, width=0.12\textwidth]{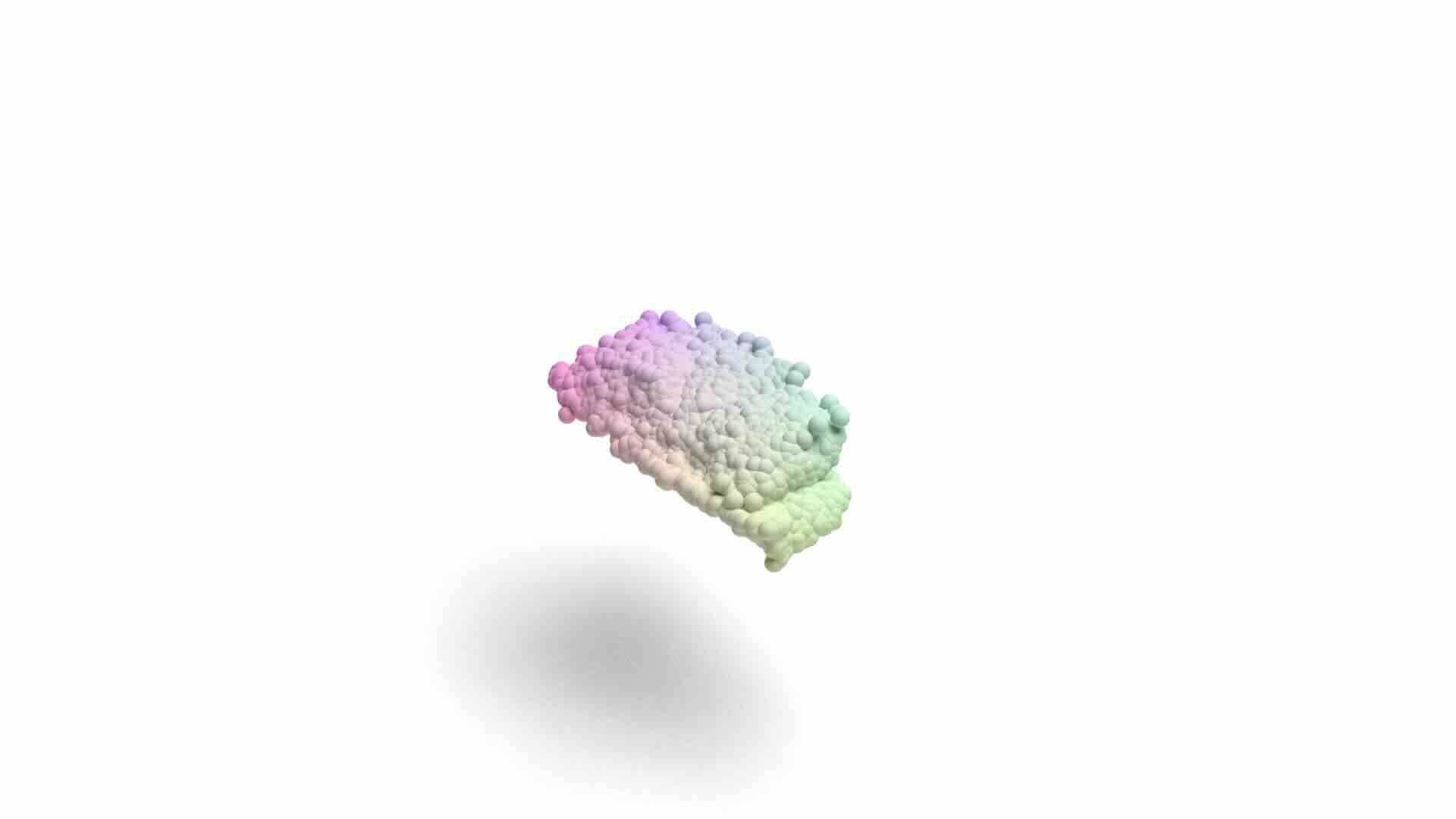} &
        \includegraphics[trim={15cm 0.0cm 15cm 0.0cm},clip, width=0.12\textwidth]{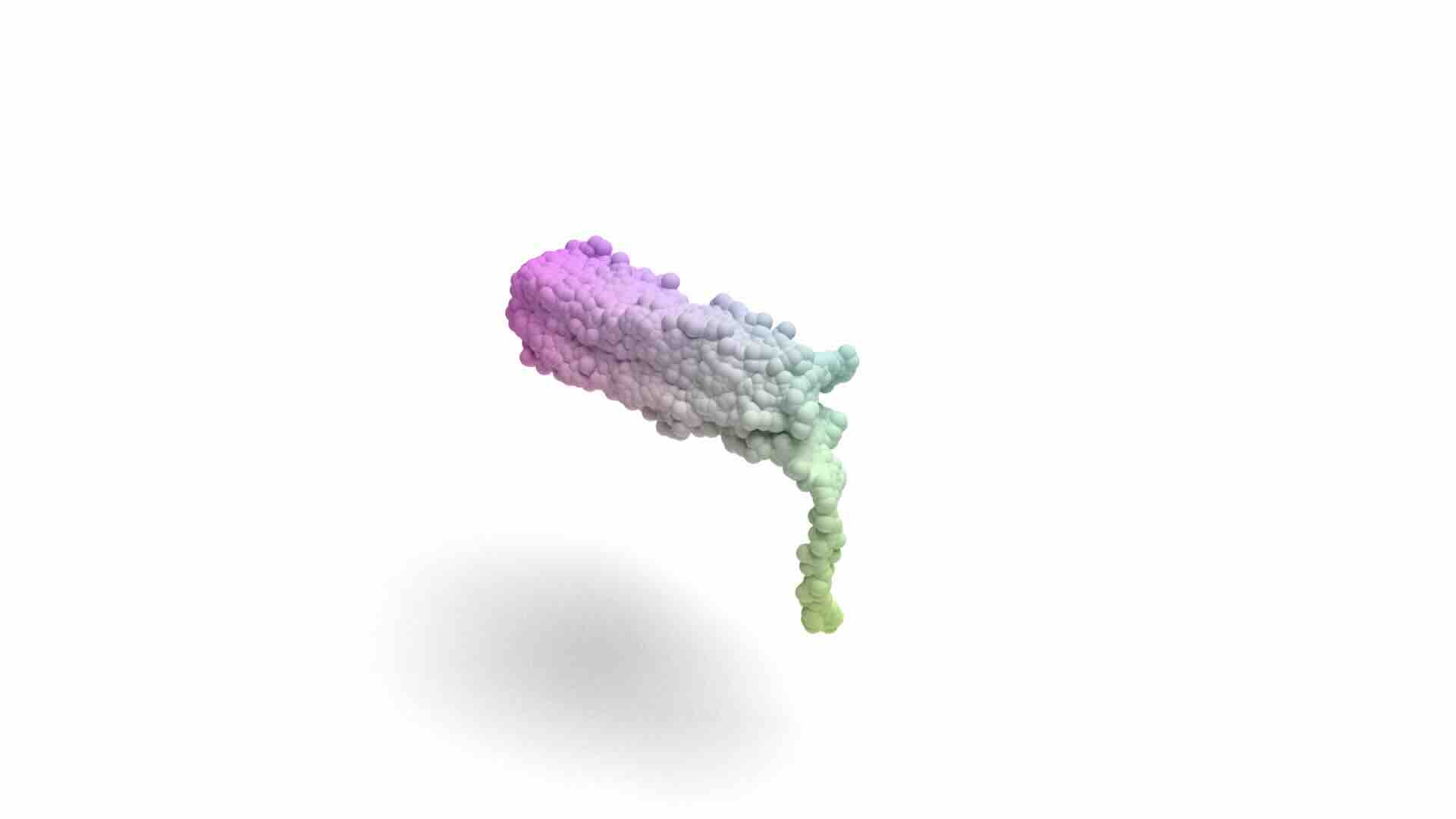} &
        \includegraphics[trim={15cm 0.0cm 15cm 0.0cm},clip, width=0.12\textwidth]{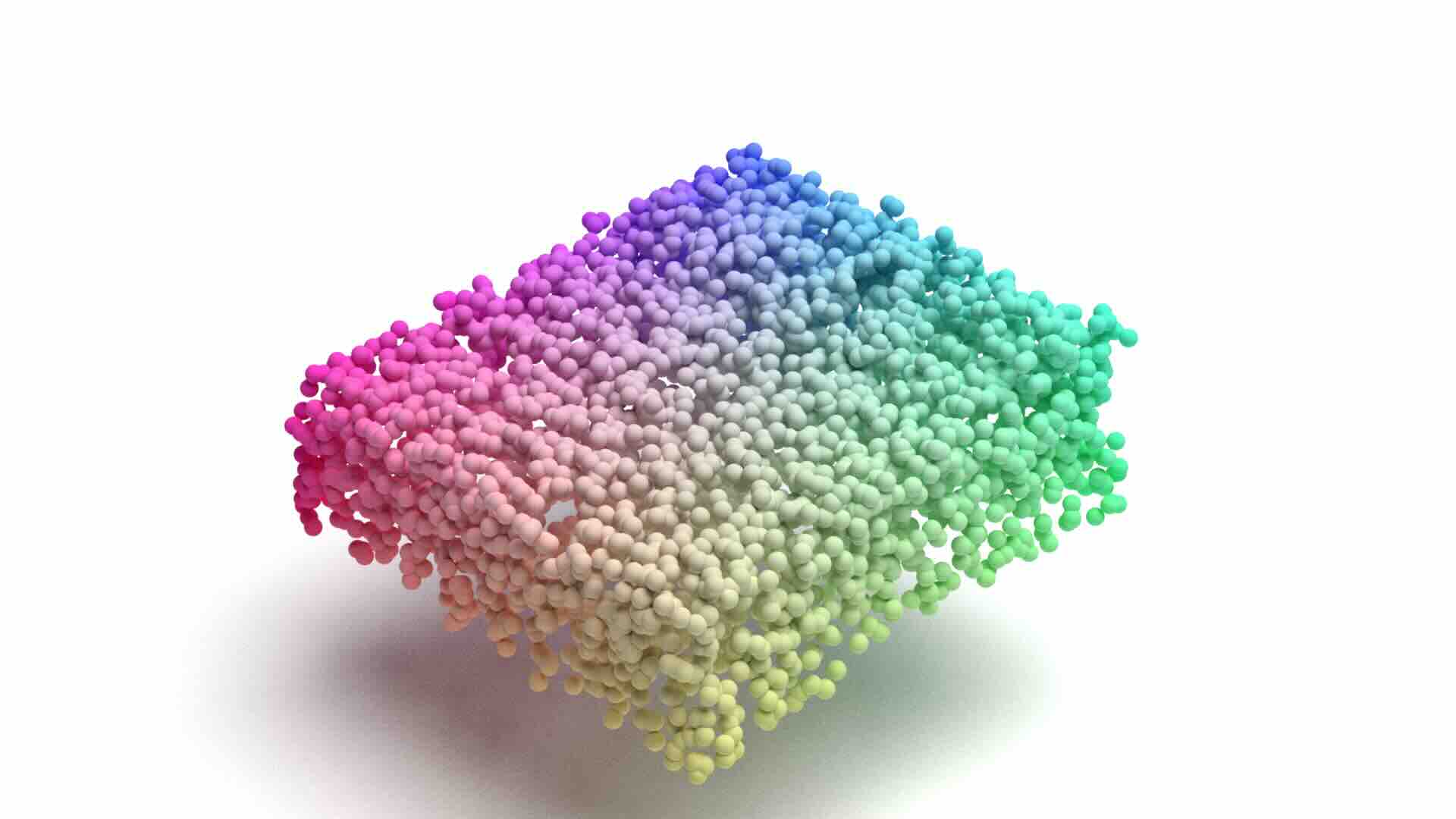} &
        \includegraphics[trim={15cm 0.0cm 15cm 0.0cm},clip, width=0.12\textwidth]{figures/scannetpp_pts/133_ours.jpg} &
        \includegraphics[trim={15cm 0.0cm 15cm 0.0cm},clip,width=0.12\textwidth]{figures/scannetpp_pts/133_gt.jpg} \\
        \includegraphics[trim={15cm 0.0cm 15cm 0.0cm},clip,width=0.12\textwidth]{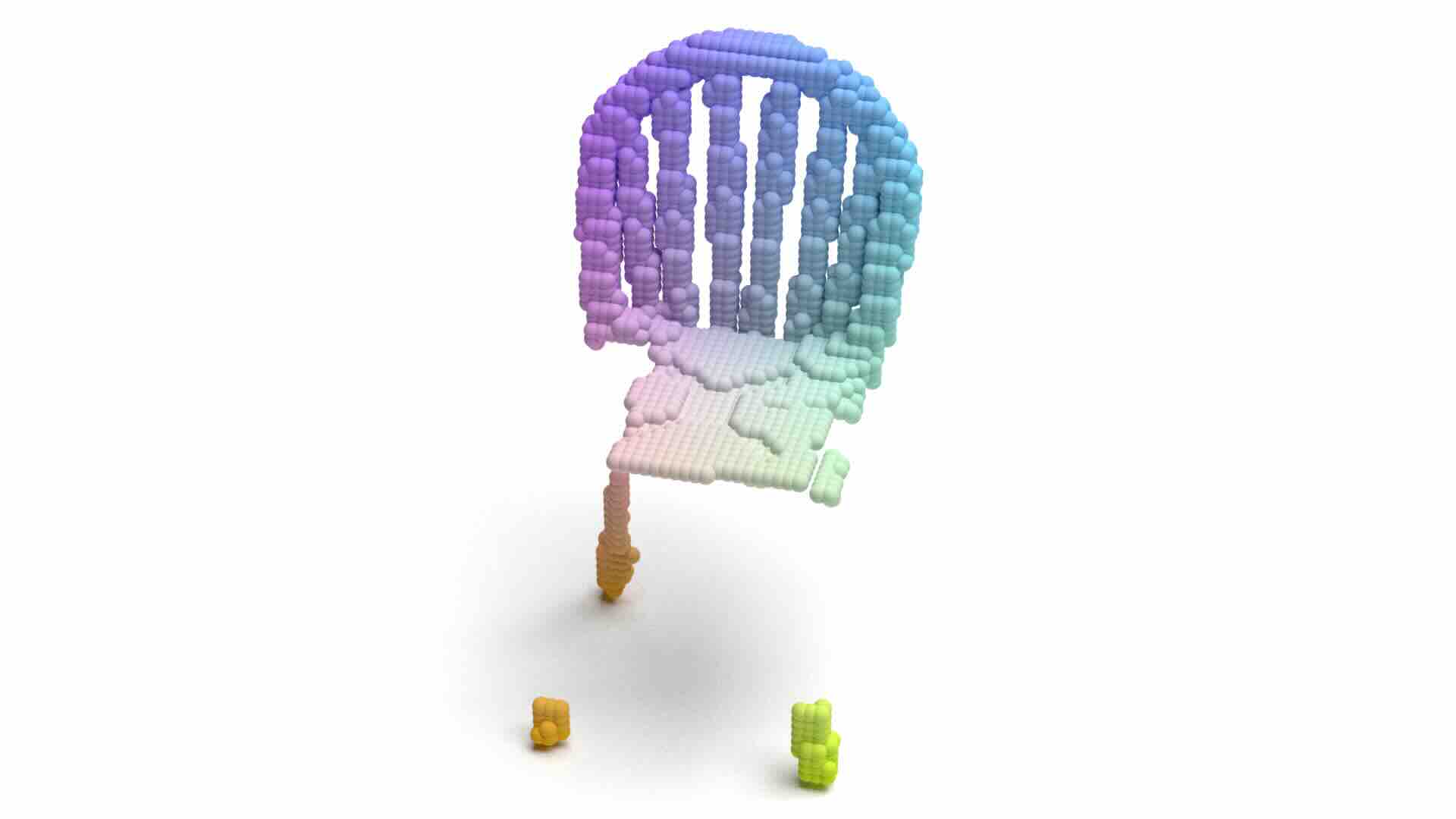} 
        &
        \includegraphics[trim={15cm 0.0cm 15cm 0.0cm},clip, width=0.12\textwidth]{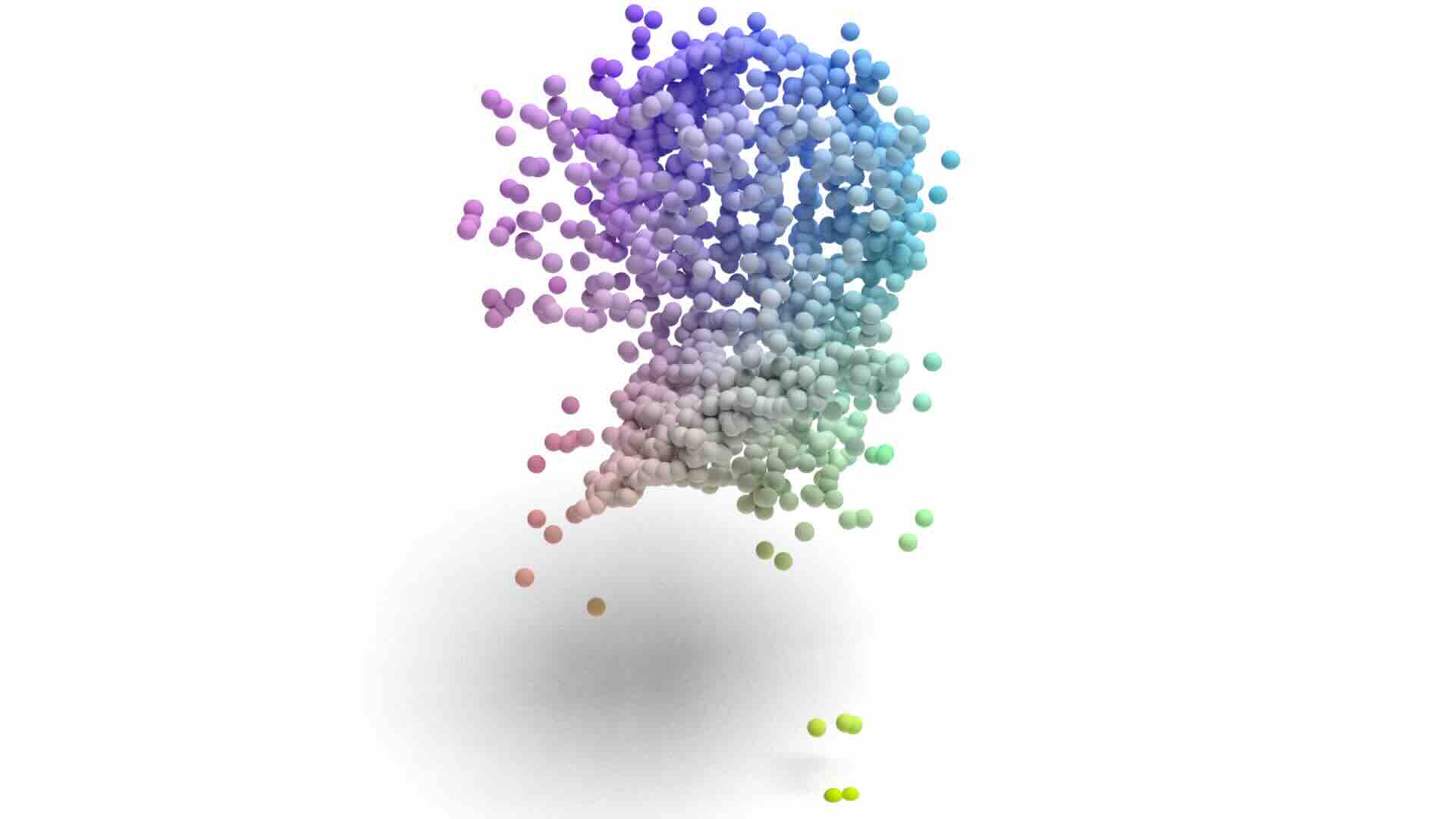} &
        \includegraphics[trim={15cm 0.0cm 15cm 0.0cm},clip, width=0.12\textwidth]{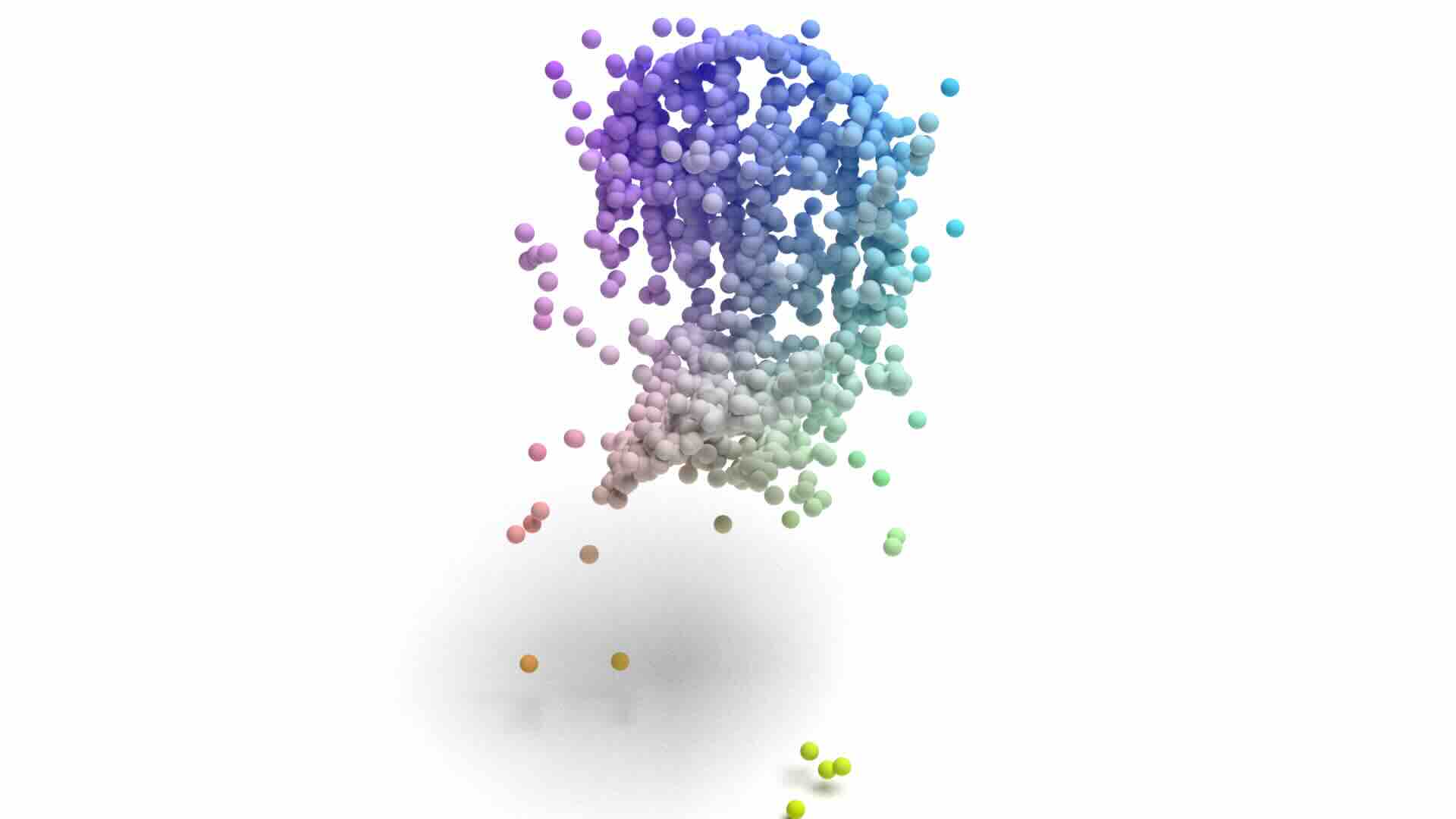} &
        \includegraphics[trim={15cm 0.0cm 15cm 0.0cm},clip, width=0.12\textwidth]{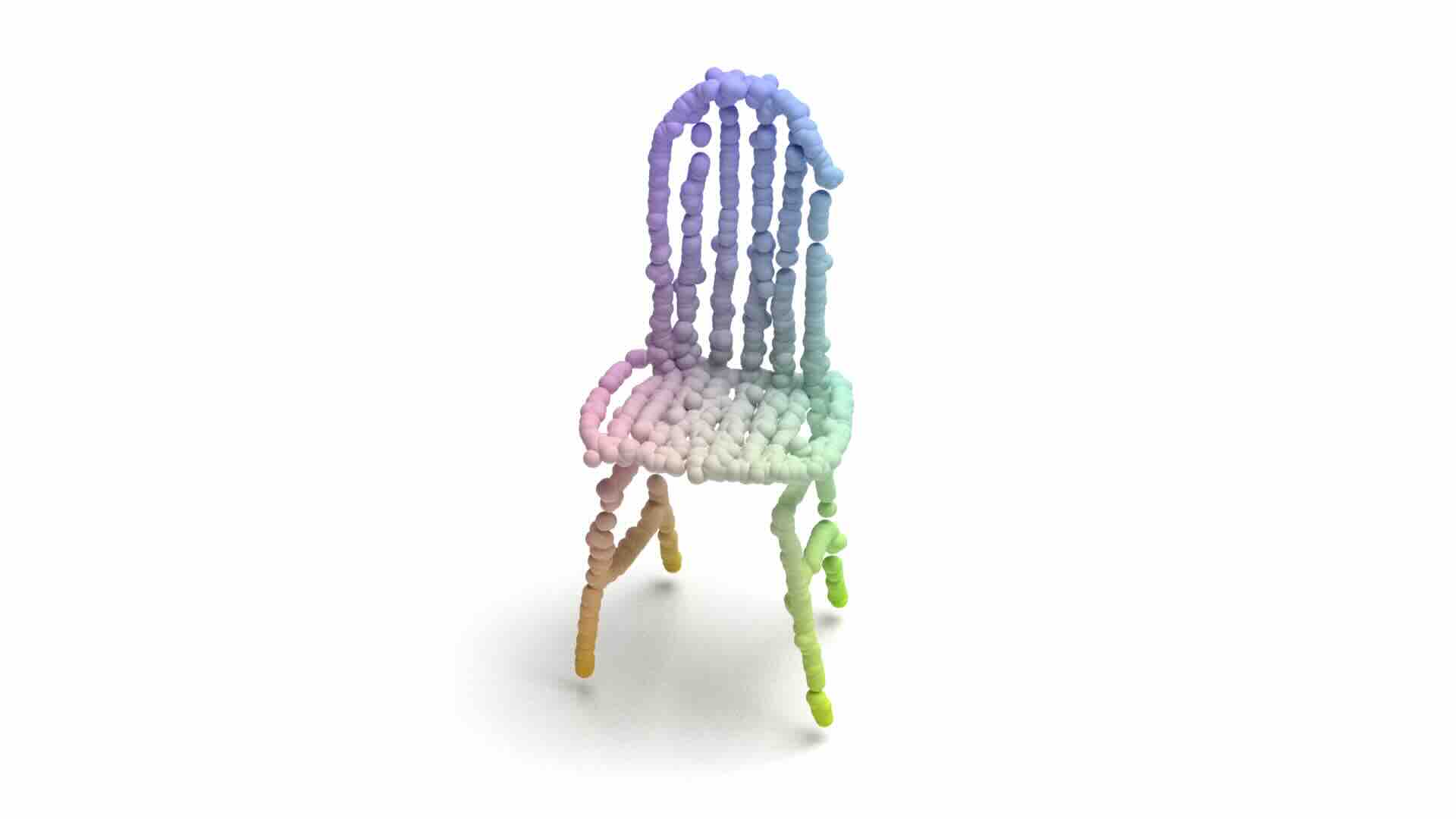} &
        \includegraphics[trim={15cm 0.0cm 15cm 0.0cm},clip,width=0.12\textwidth]{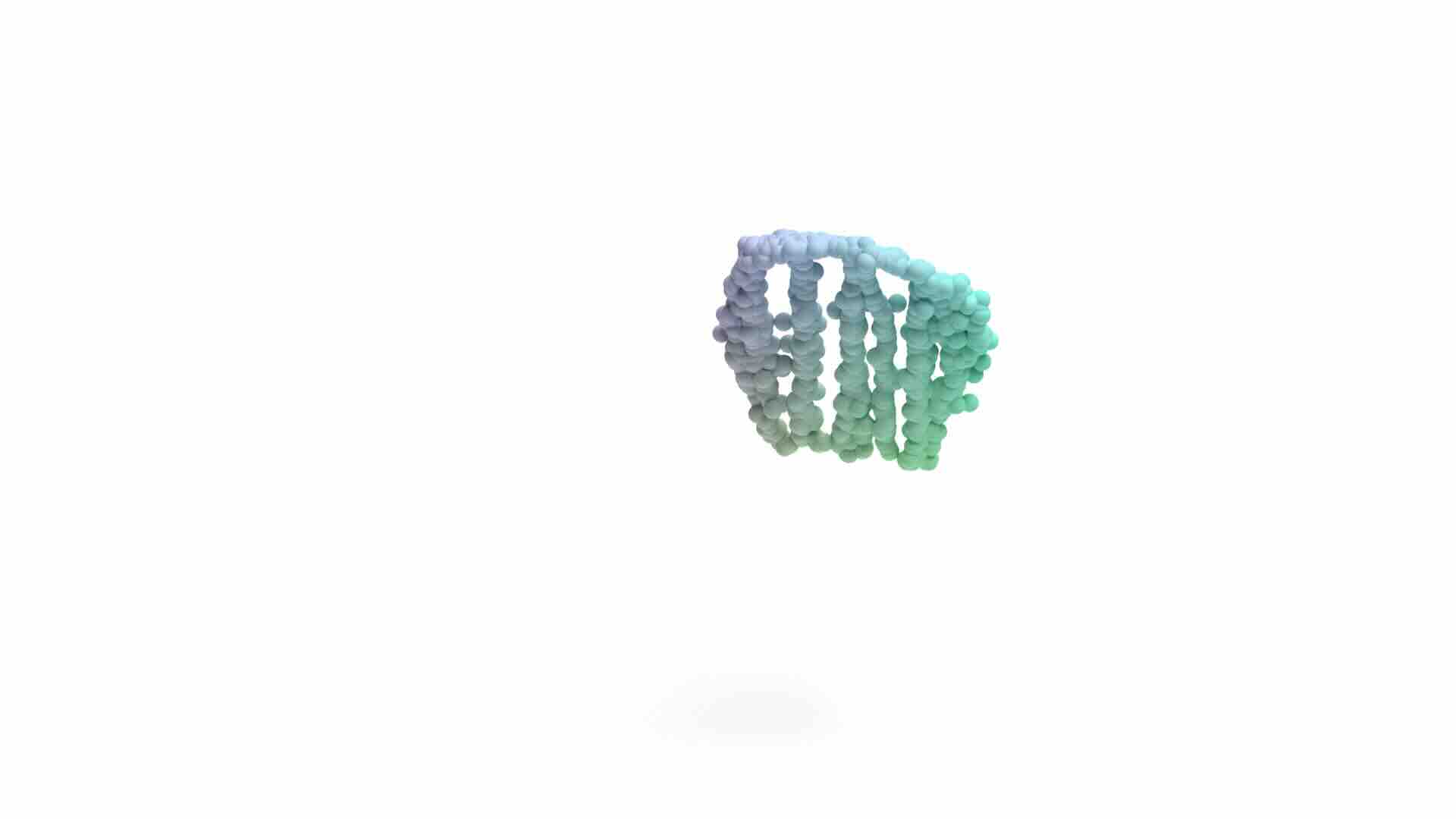} &
        \includegraphics[trim={15cm 0.0cm 15cm 0.0cm},clip,width=0.12\textwidth]{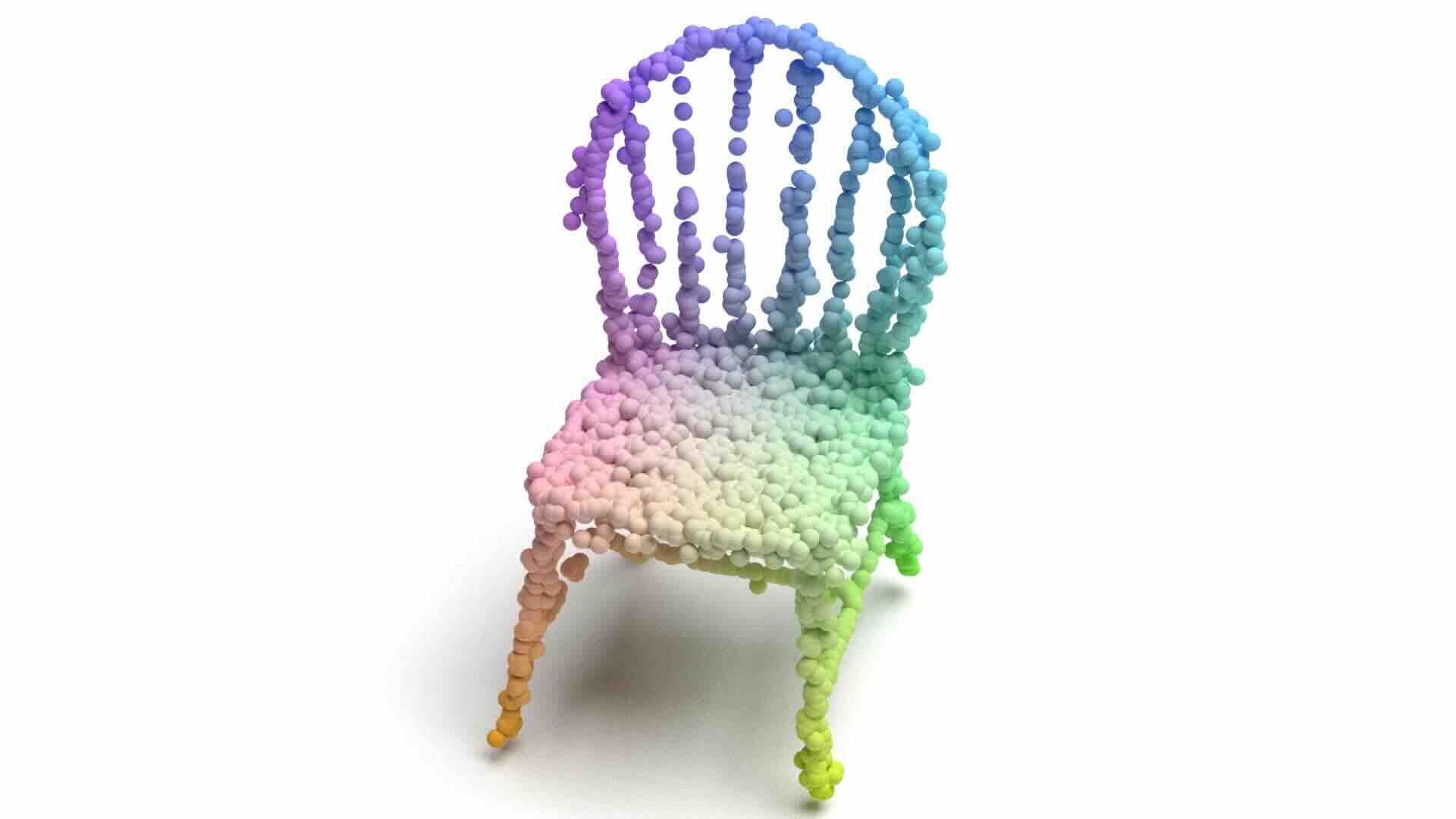} &
        \includegraphics[trim={15cm 0.0cm 15cm 0.0cm},clip,width=0.12\textwidth]{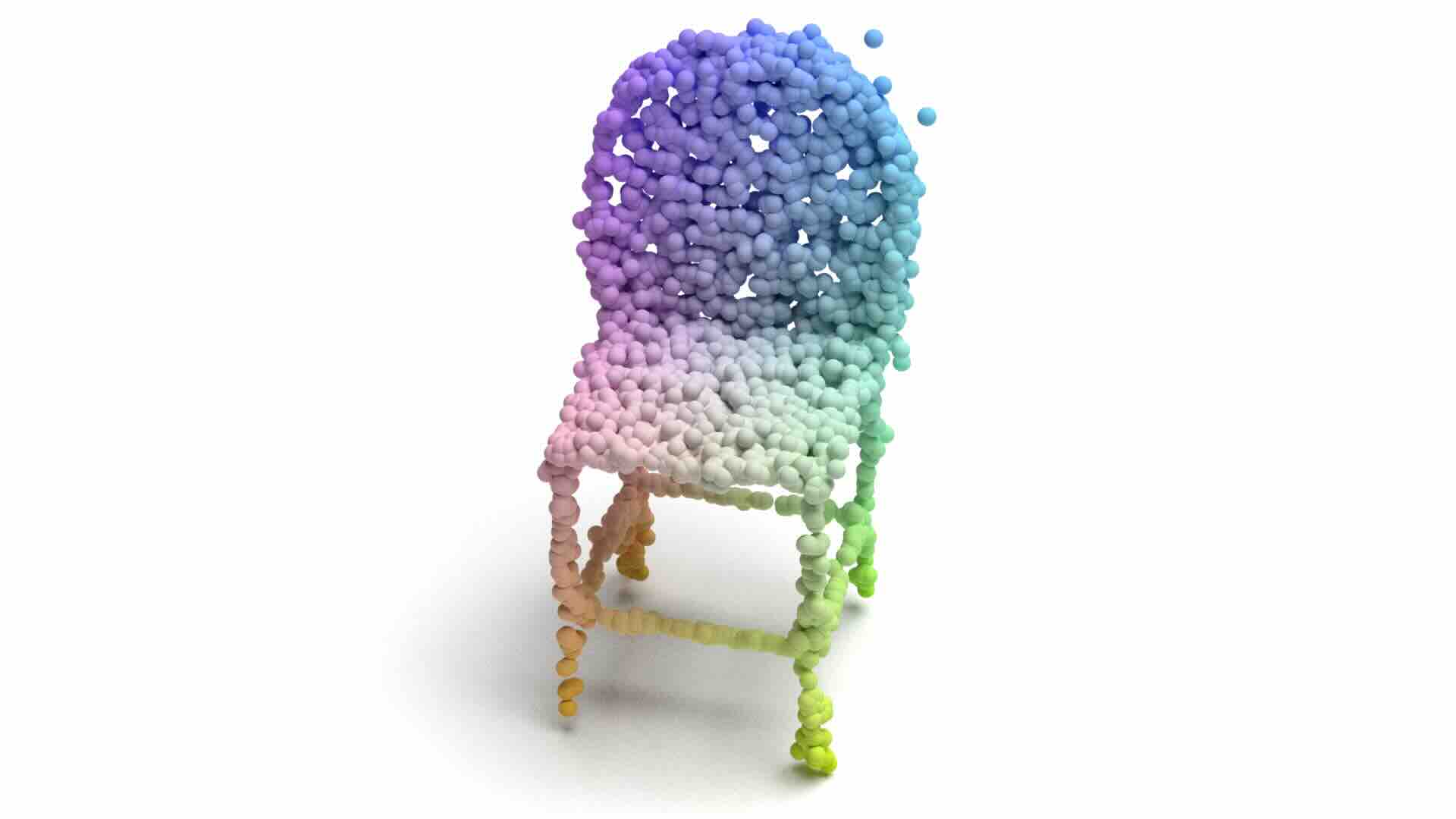} &
        \includegraphics[trim={15cm 0.0cm 15cm 0.0cm},clip,width=0.12\textwidth]{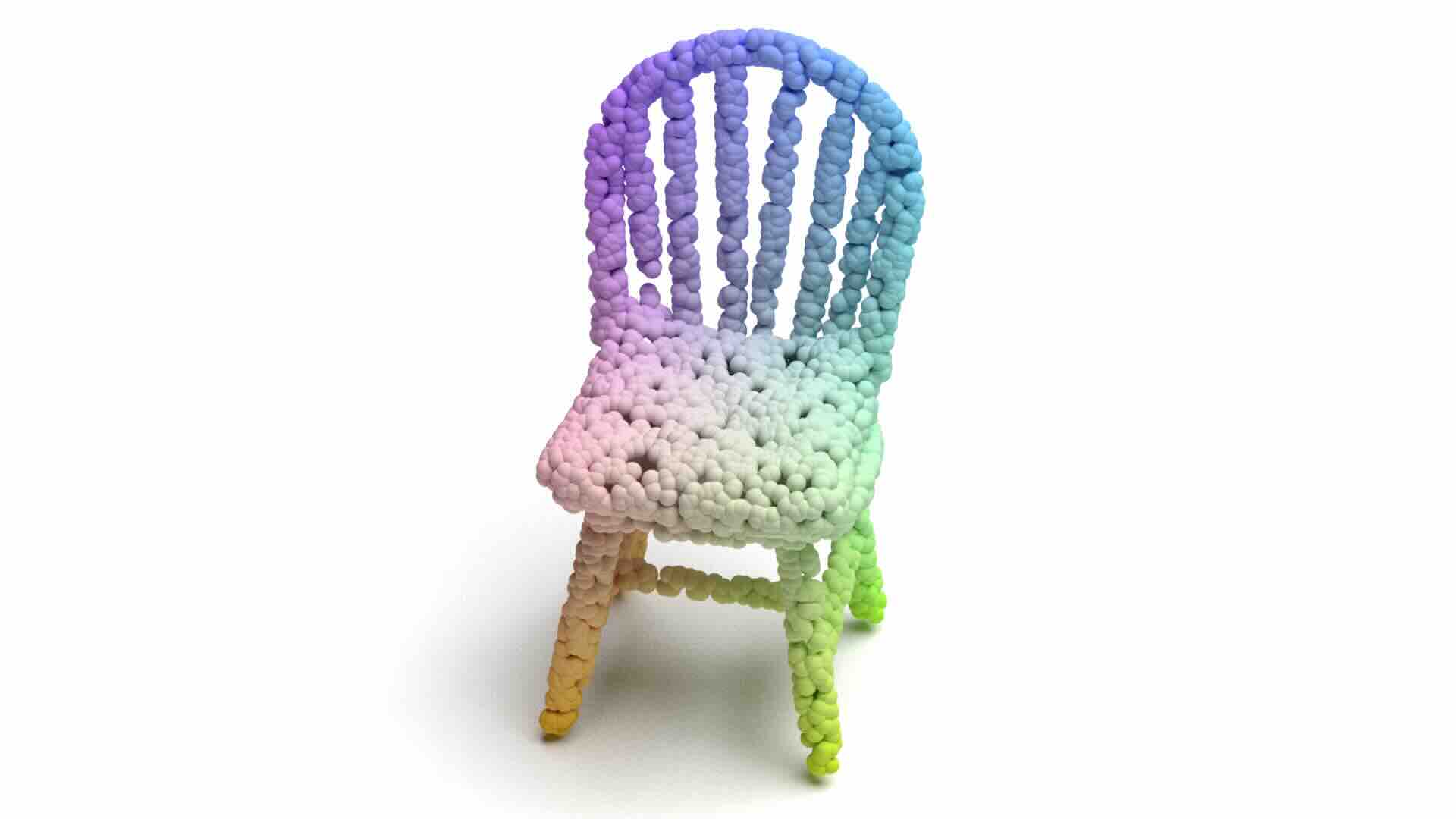}
        \\
        \includegraphics[trim={15cm 0.0cm 15cm 0.0cm},clip,width=0.12\textwidth]{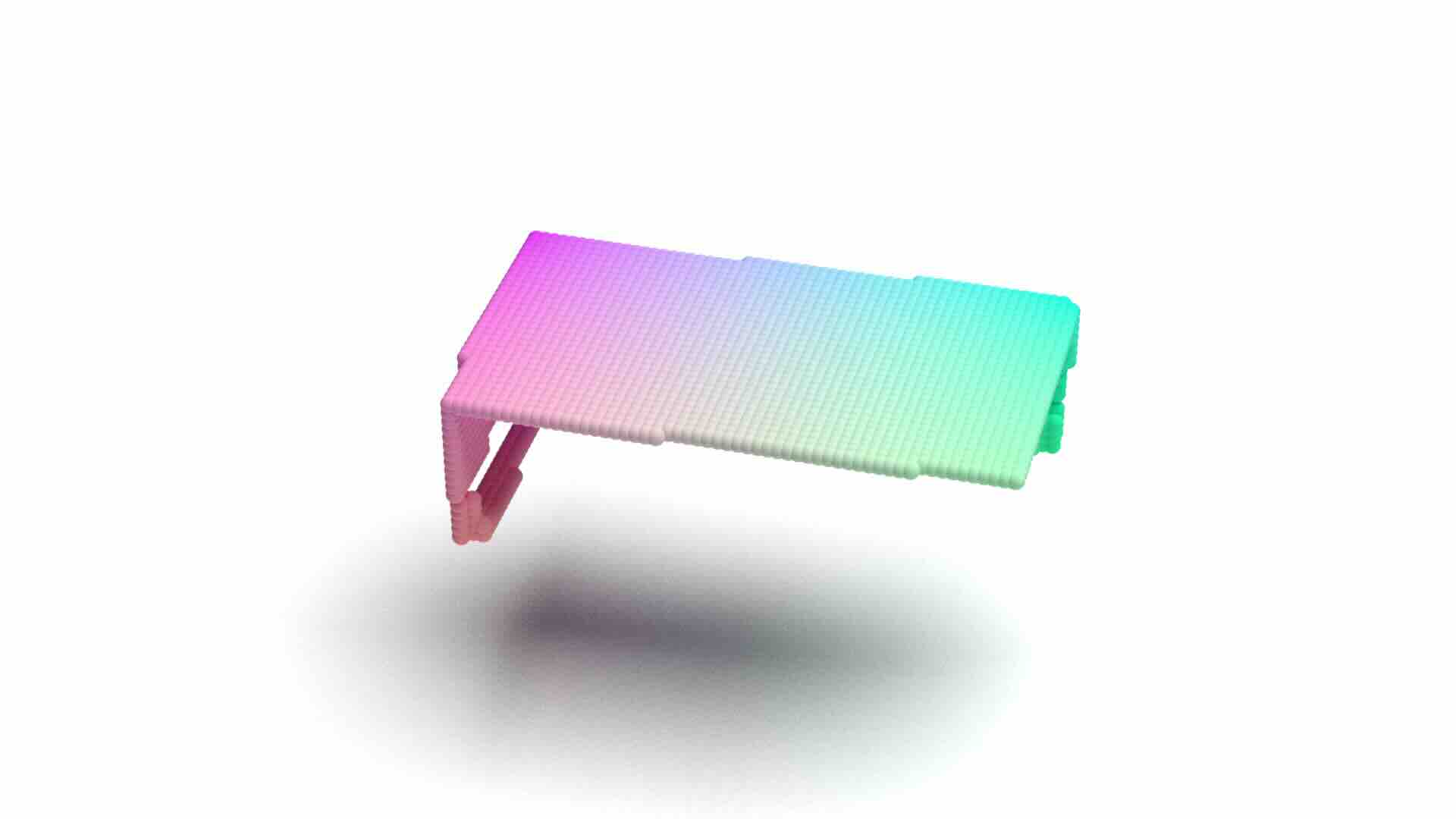} 
        &
        \includegraphics[trim={15cm 0.0cm 15cm 0.0cm},clip, width=0.12\textwidth]{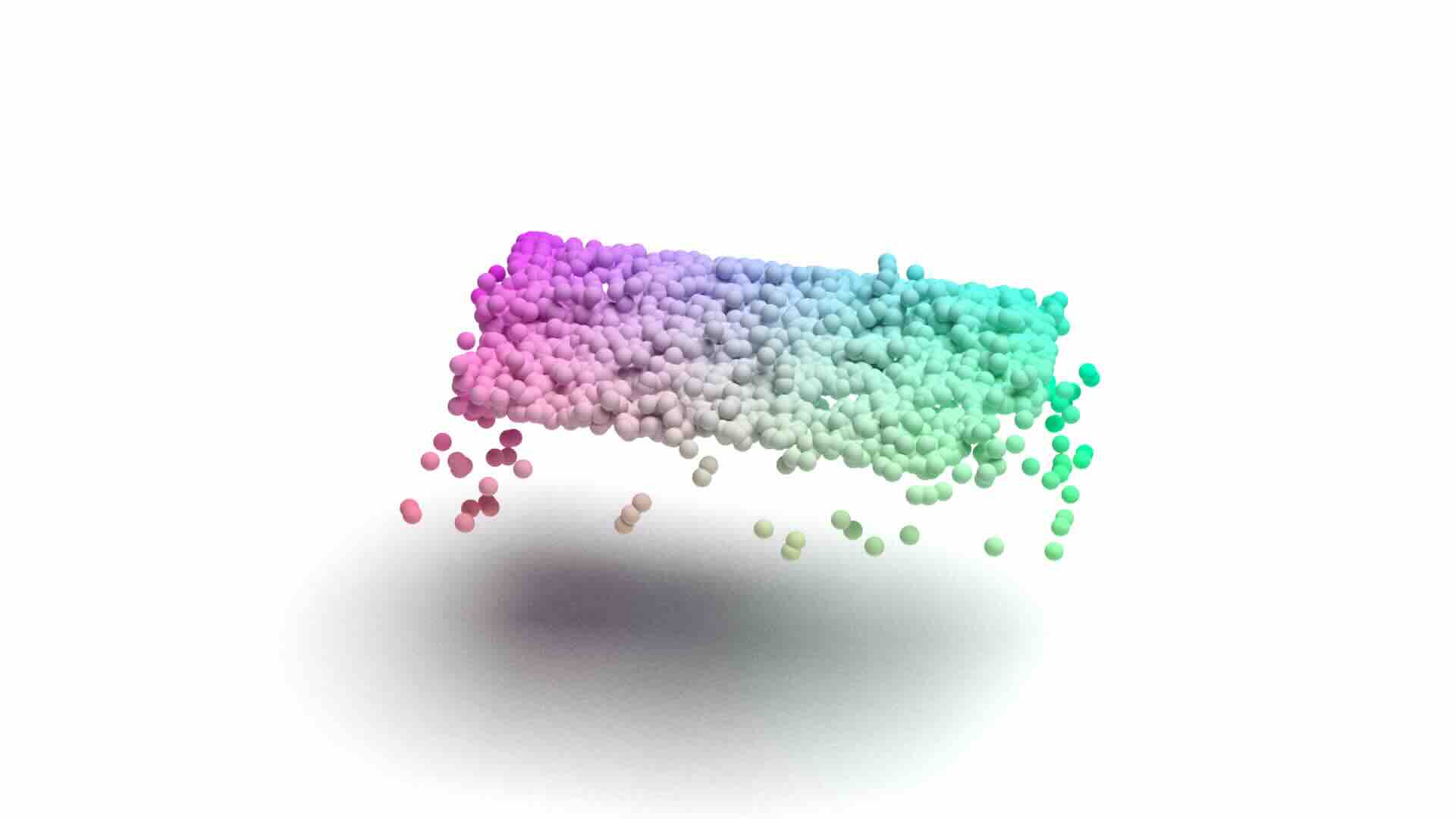} &
        \includegraphics[trim={15cm 0.0cm 15cm 0.0cm},clip, width=0.12\textwidth]{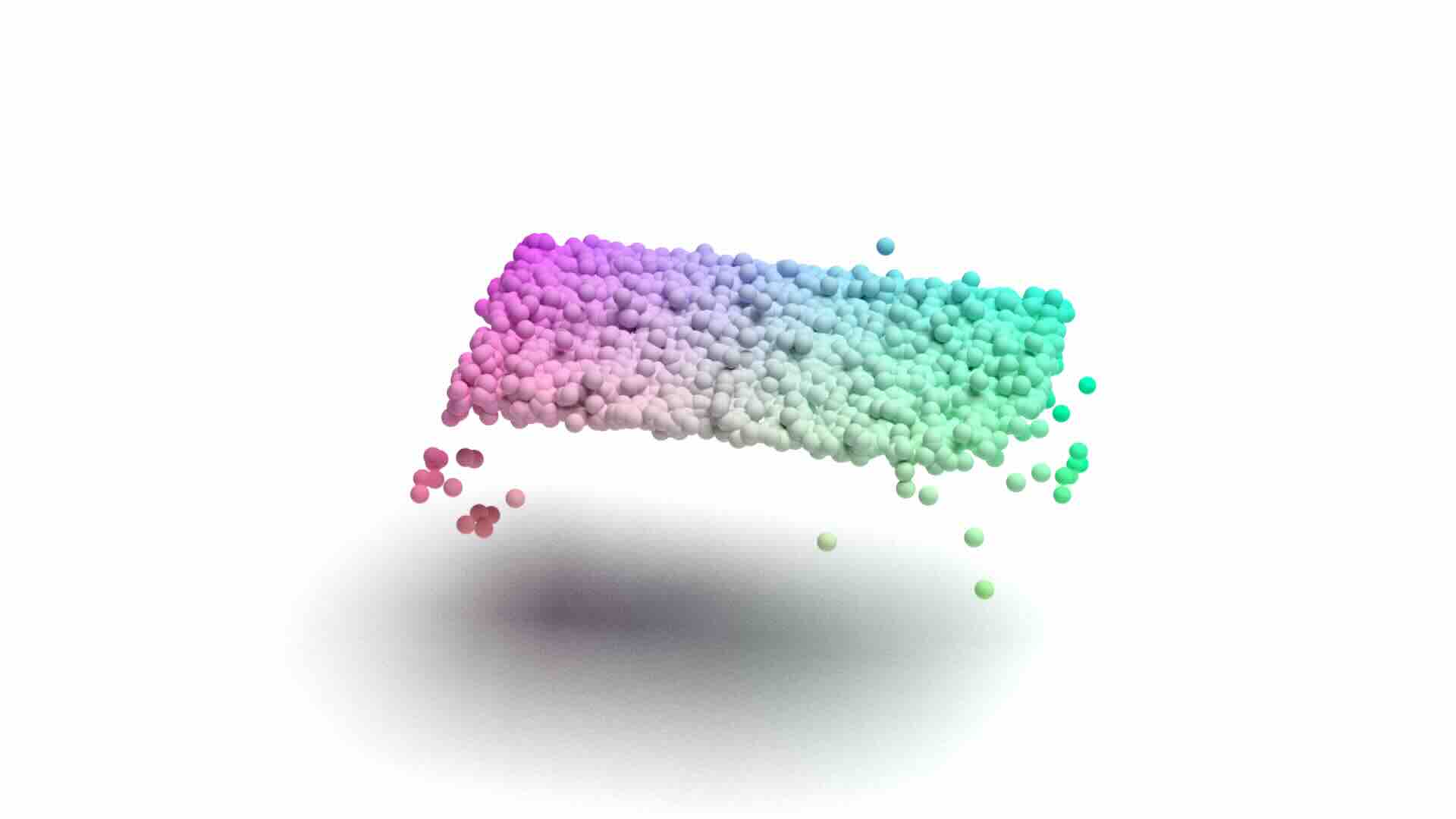} &
        \includegraphics[trim={15cm 0.0cm 15cm 0.0cm},clip, width=0.12\textwidth]{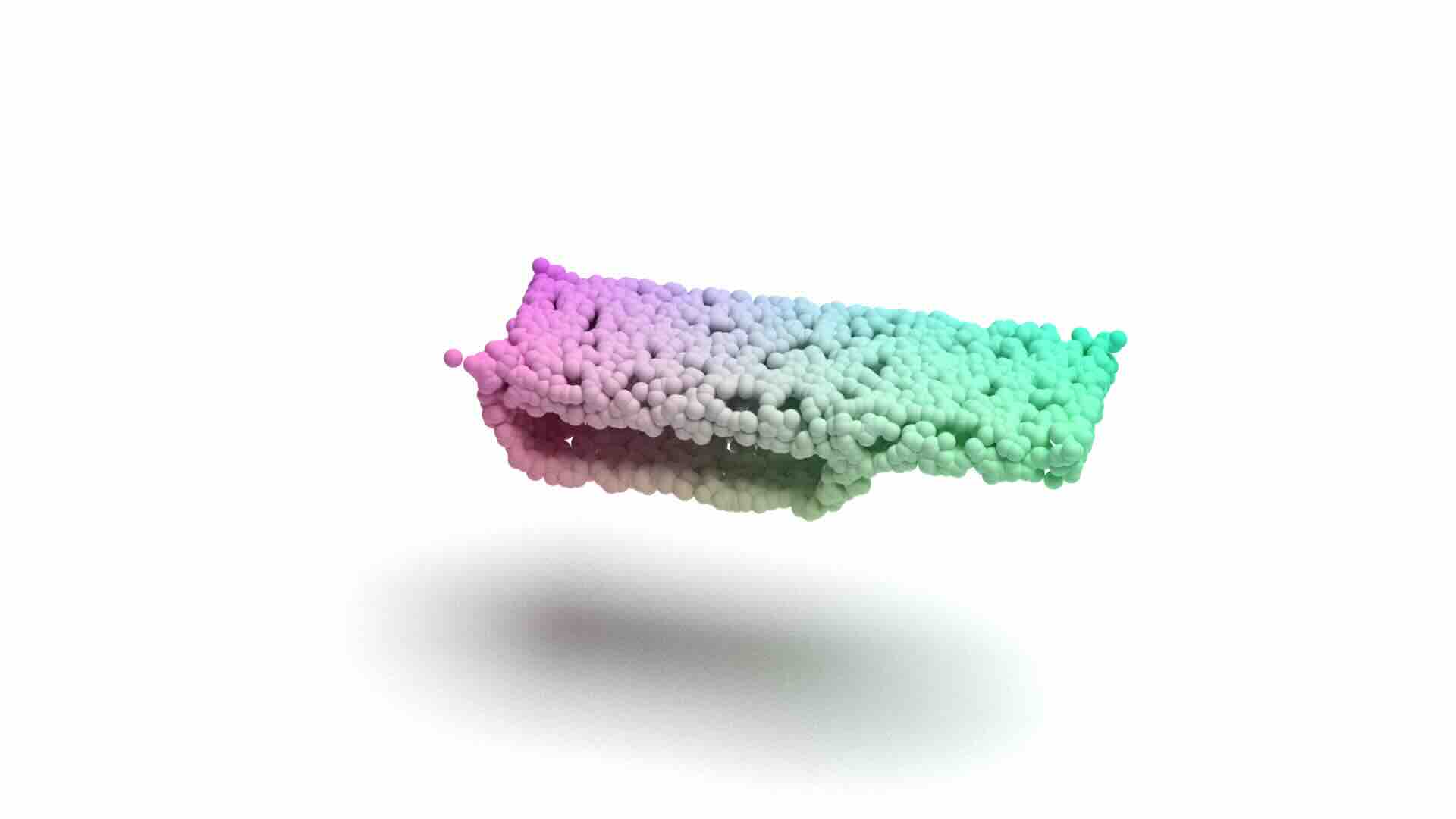} &
        \includegraphics[trim={15cm 0.0cm 15cm 0.0cm},clip,width=0.12\textwidth]{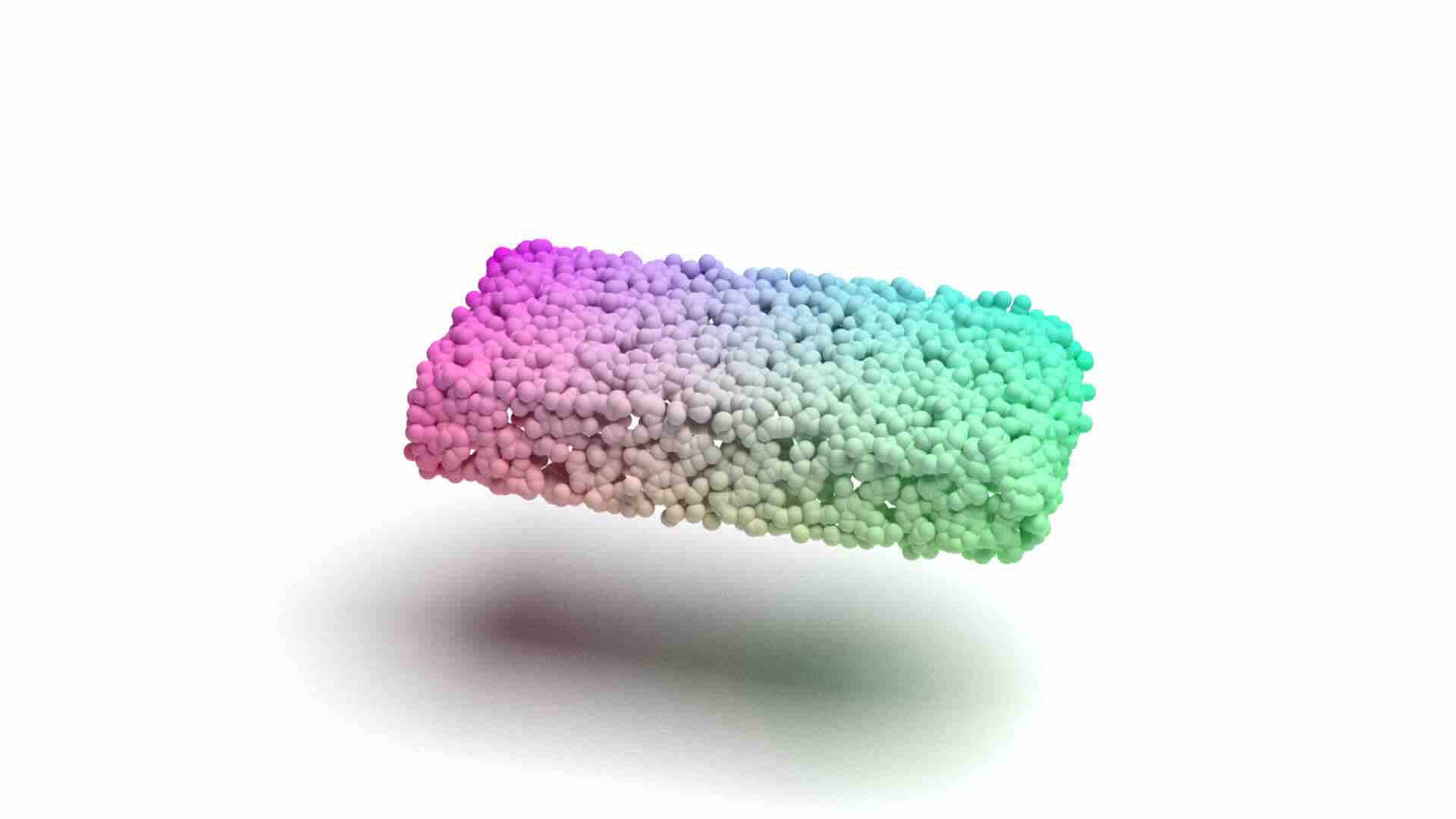} &
        \includegraphics[trim={15cm 0.0cm 15cm 0.0cm},clip,width=0.12\textwidth]{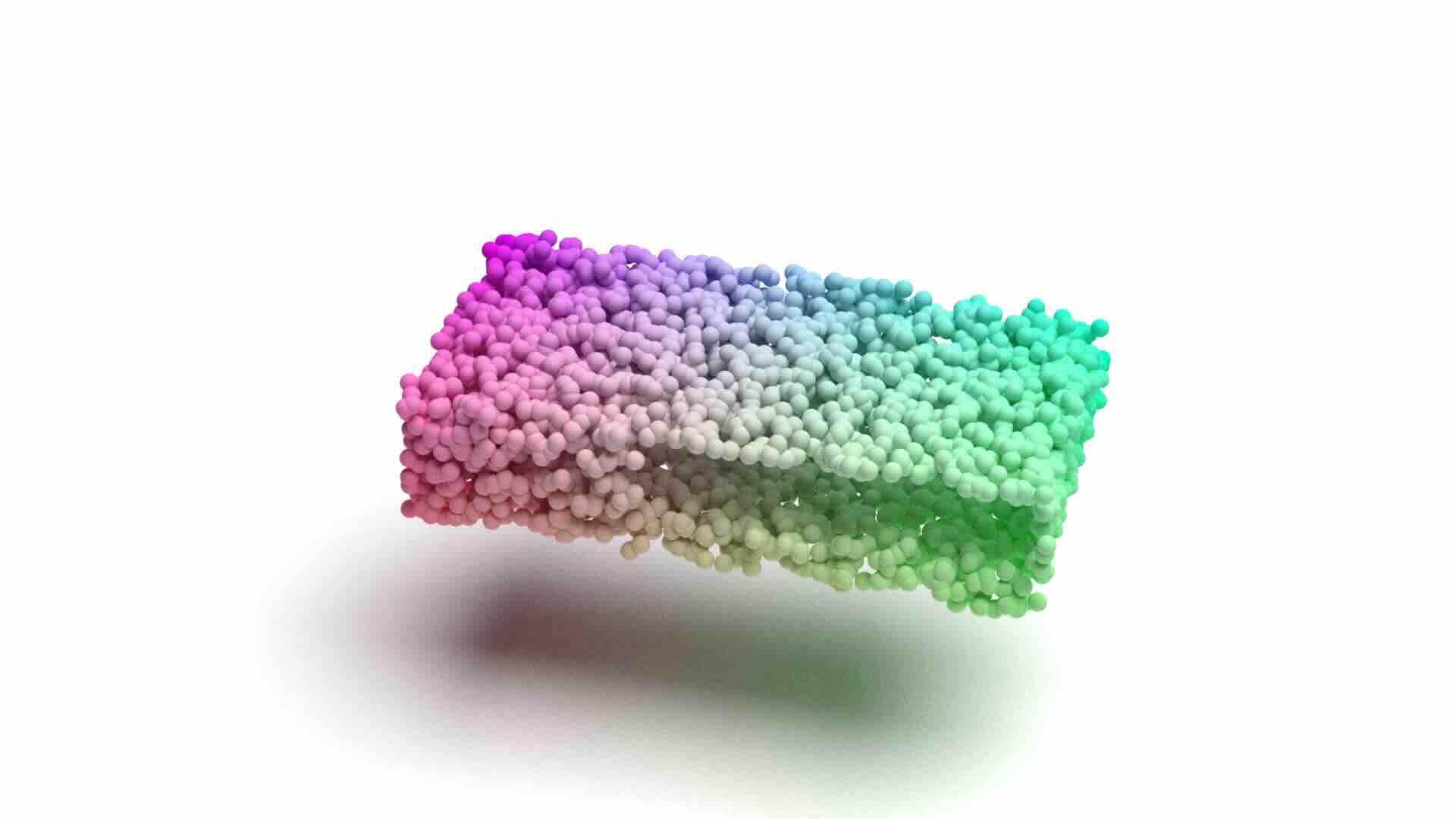} &
        \includegraphics[trim={15cm 0.0cm 15cm 0.0cm},clip,width=0.12\textwidth]{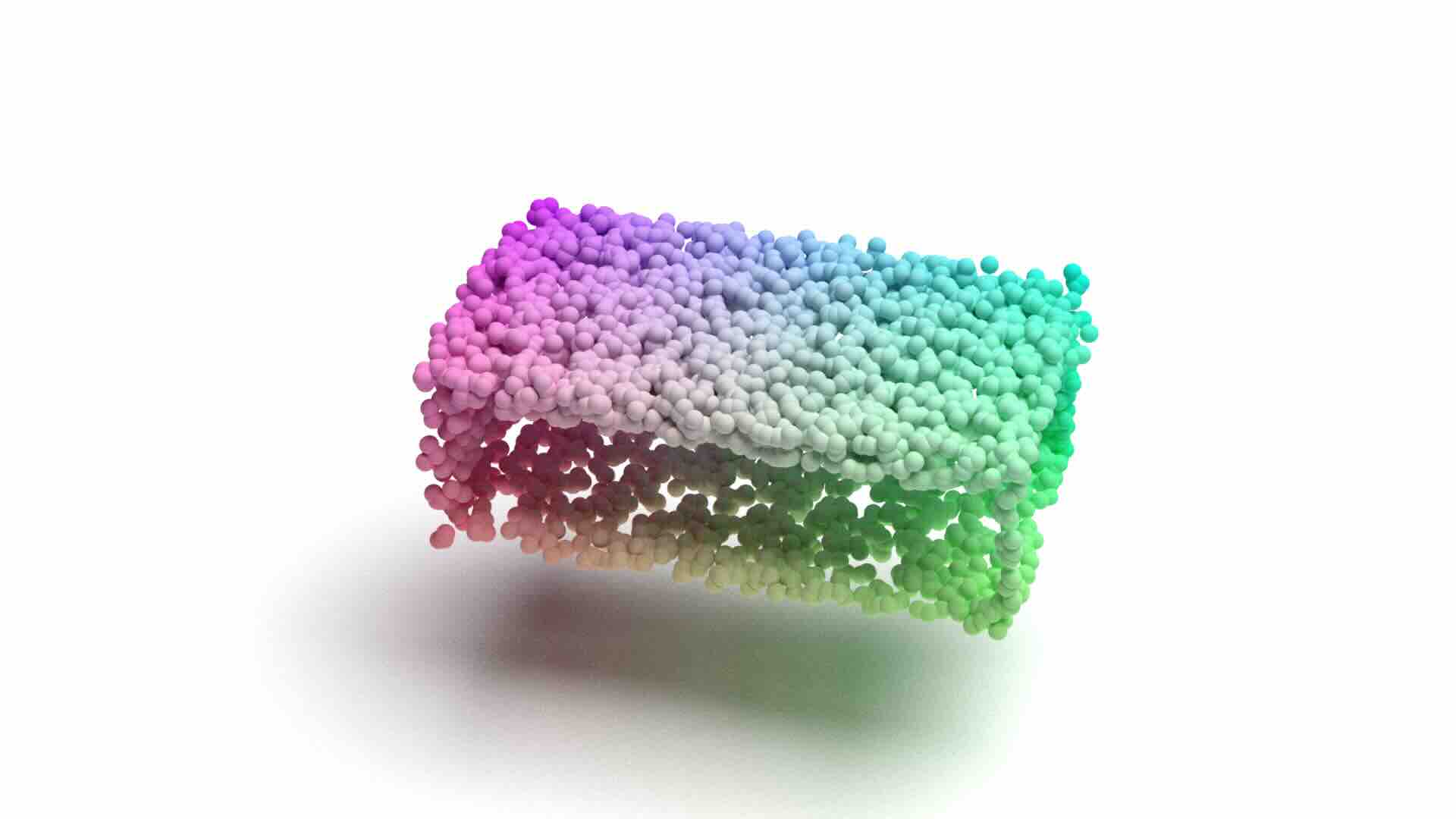} &
        \includegraphics[trim={15cm 0.0cm 15cm 0.0cm},clip,width=0.12\textwidth]{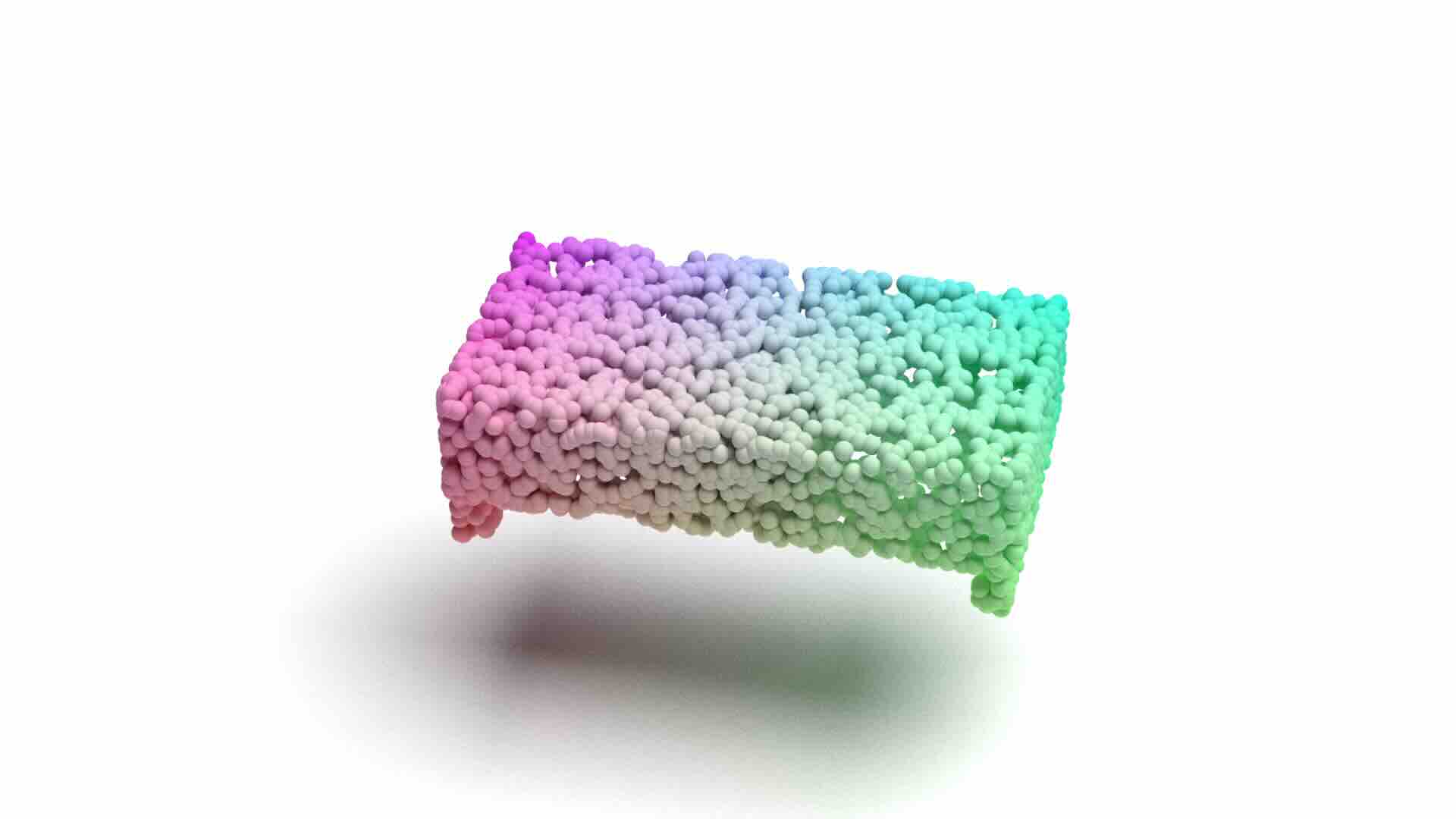}
        \\
        \includegraphics[trim={15cm 0.0cm 15cm 0.0cm},clip,width=0.12\textwidth]{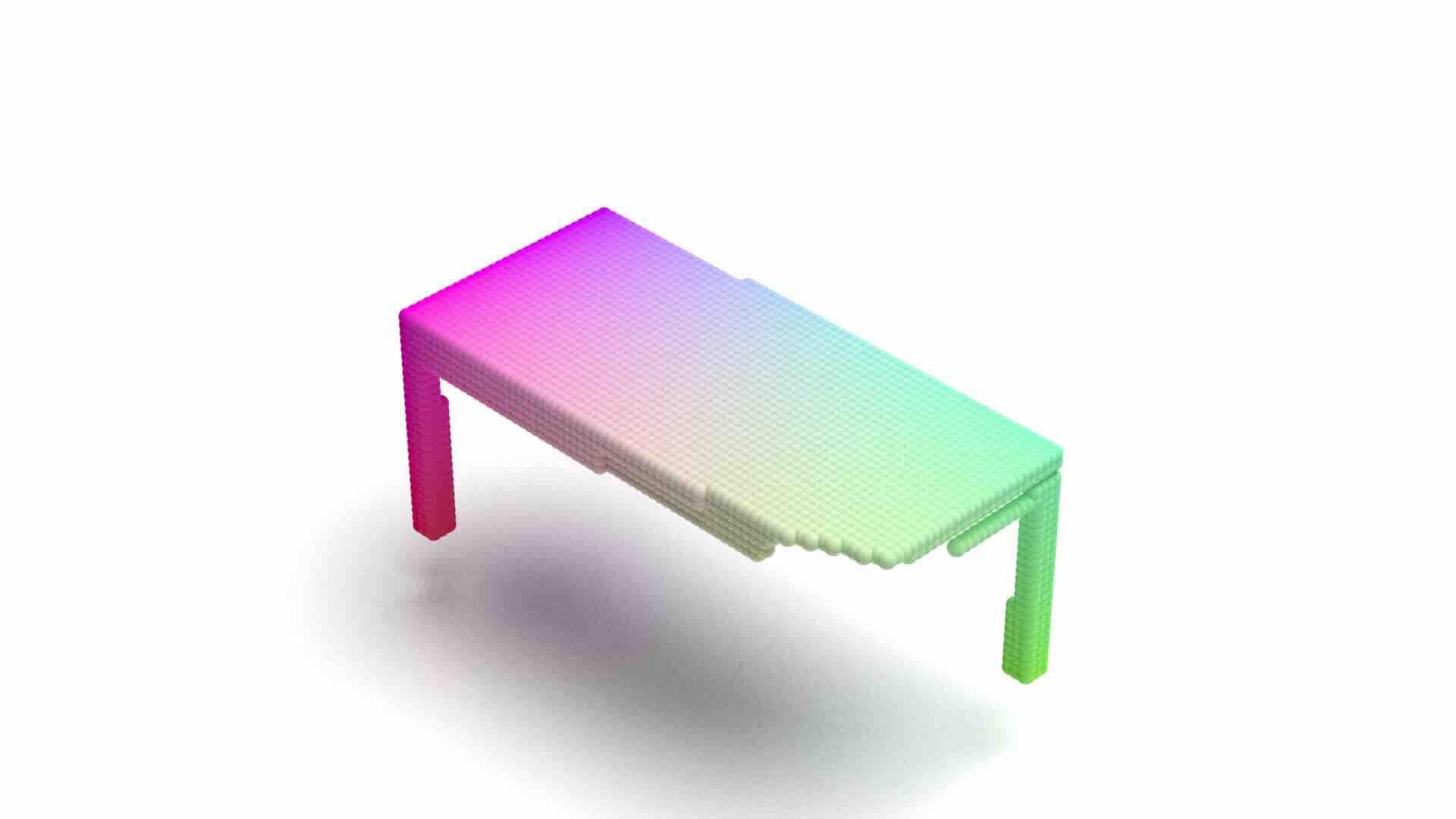} 
        &
        \includegraphics[trim={15cm 0.0cm 15cm 0.0cm},clip, width=0.12\textwidth]{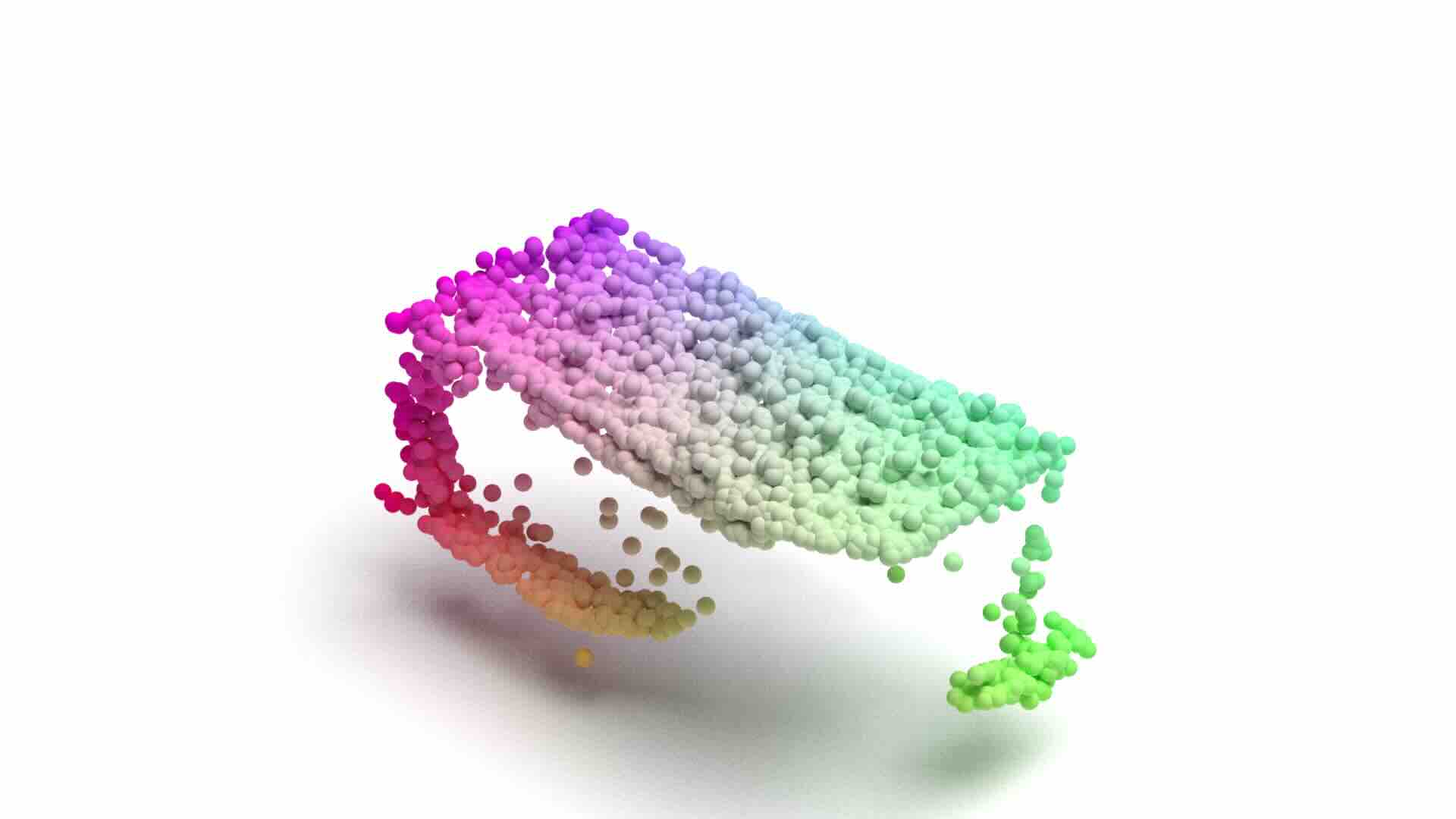} &
        \includegraphics[trim={15cm 0.0cm 15cm 0.0cm},clip, width=0.12\textwidth]{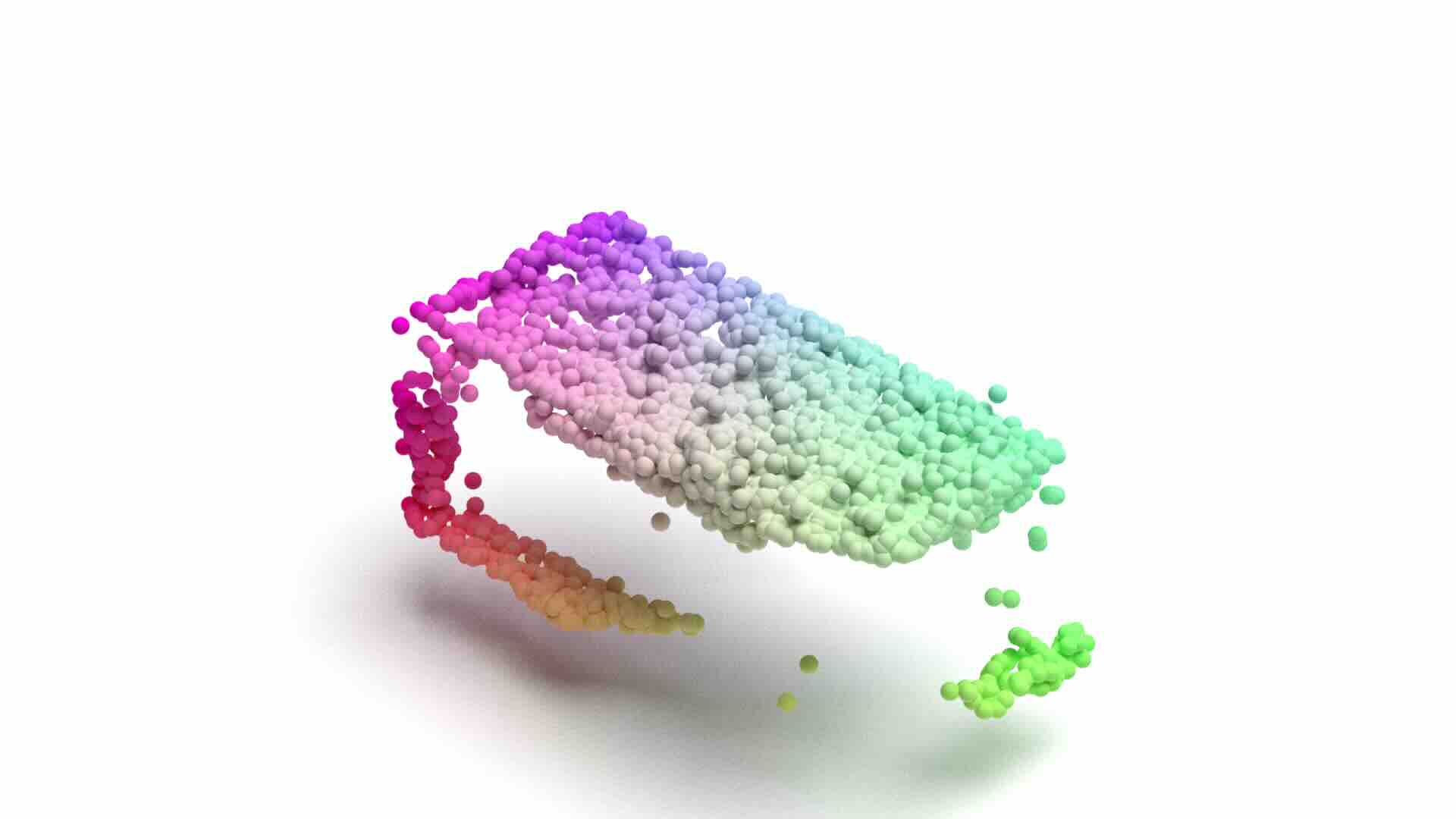} &
        \includegraphics[trim={15cm 0.0cm 15cm 0.0cm},clip, width=0.12\textwidth]{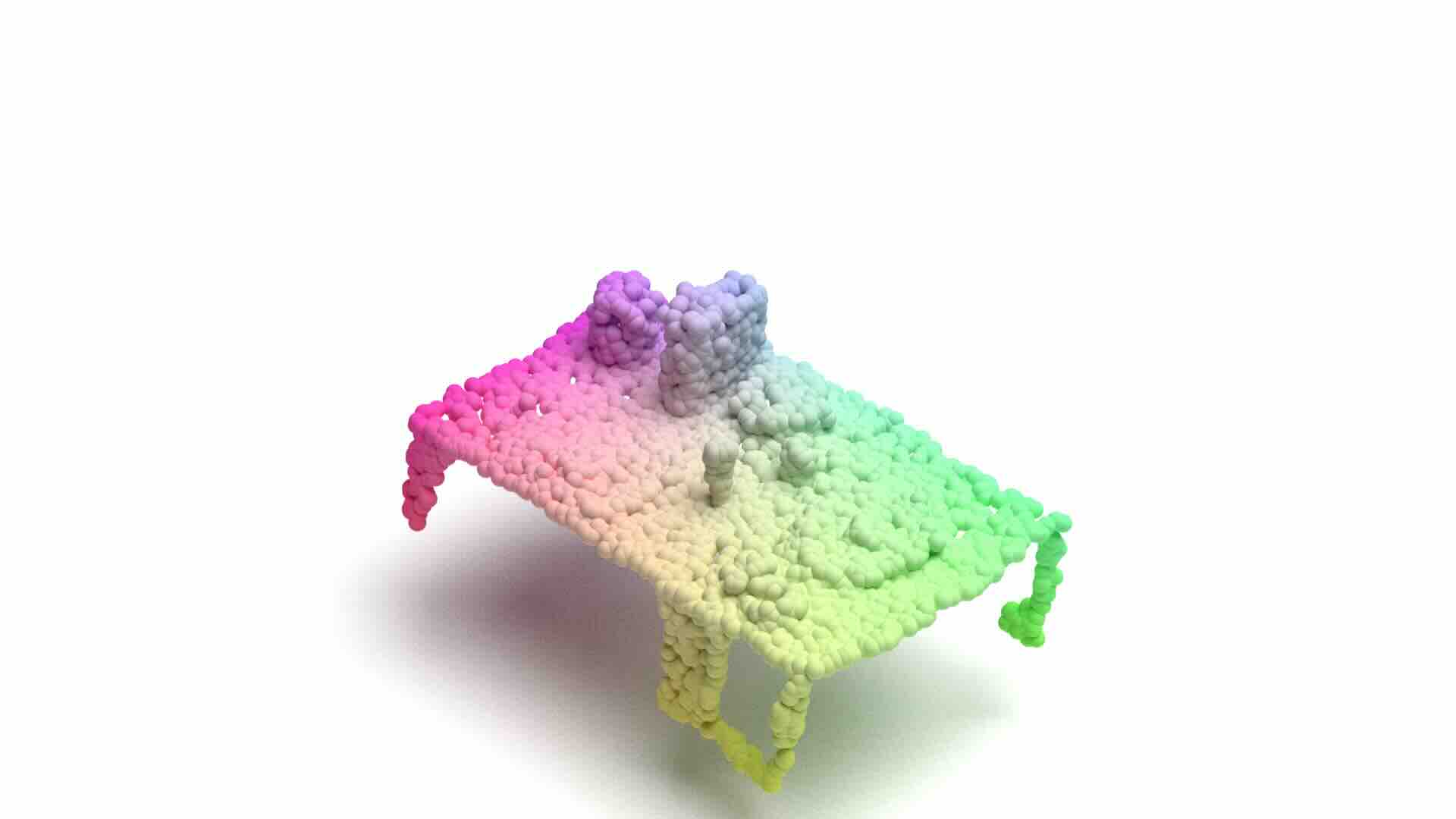} &
        \includegraphics[trim={15cm 0.0cm 15cm 0.0cm},clip,width=0.12\textwidth]{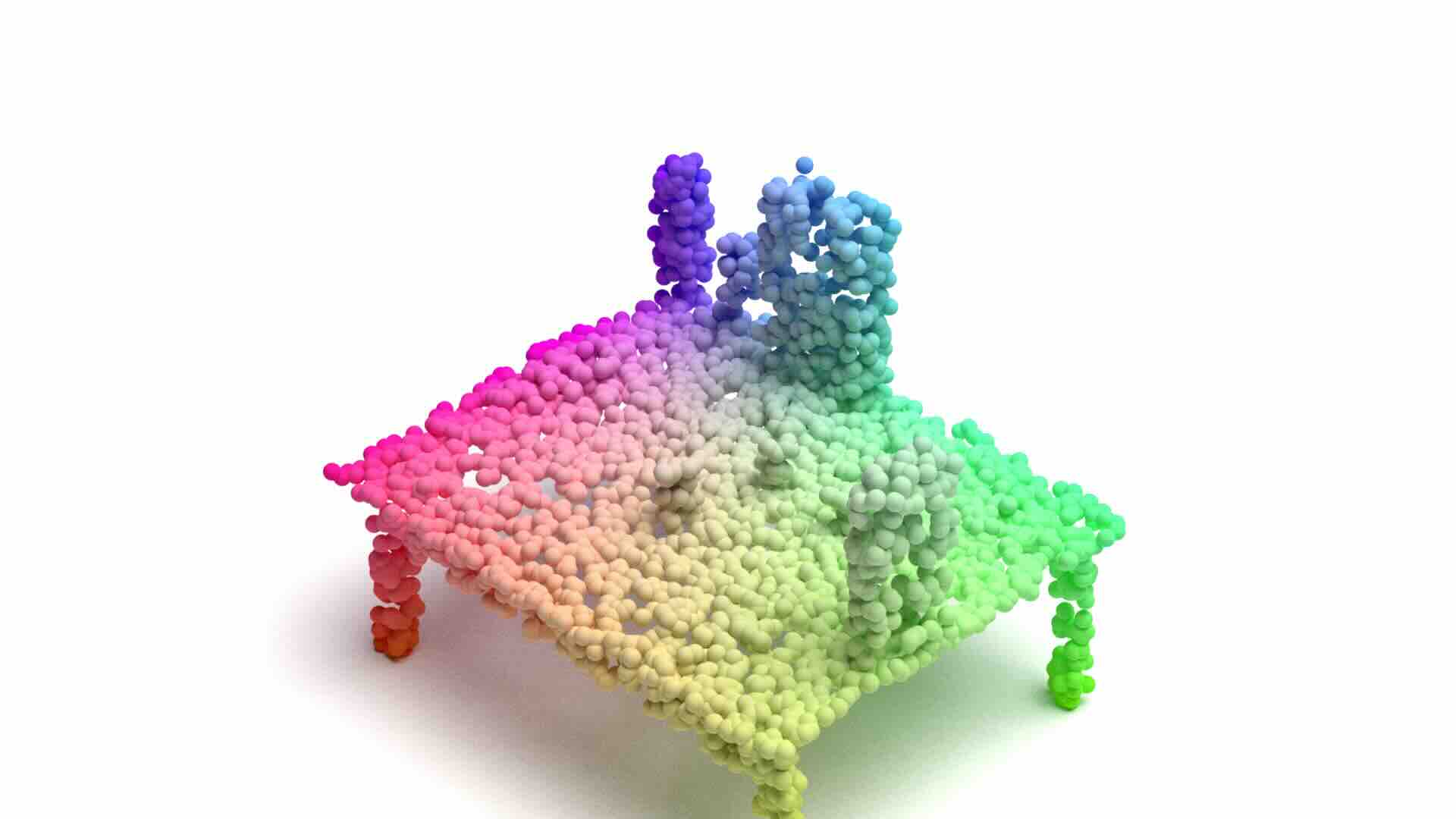} &
        \includegraphics[trim={15cm 0.0cm 15cm 0.0cm},clip,width=0.12\textwidth]{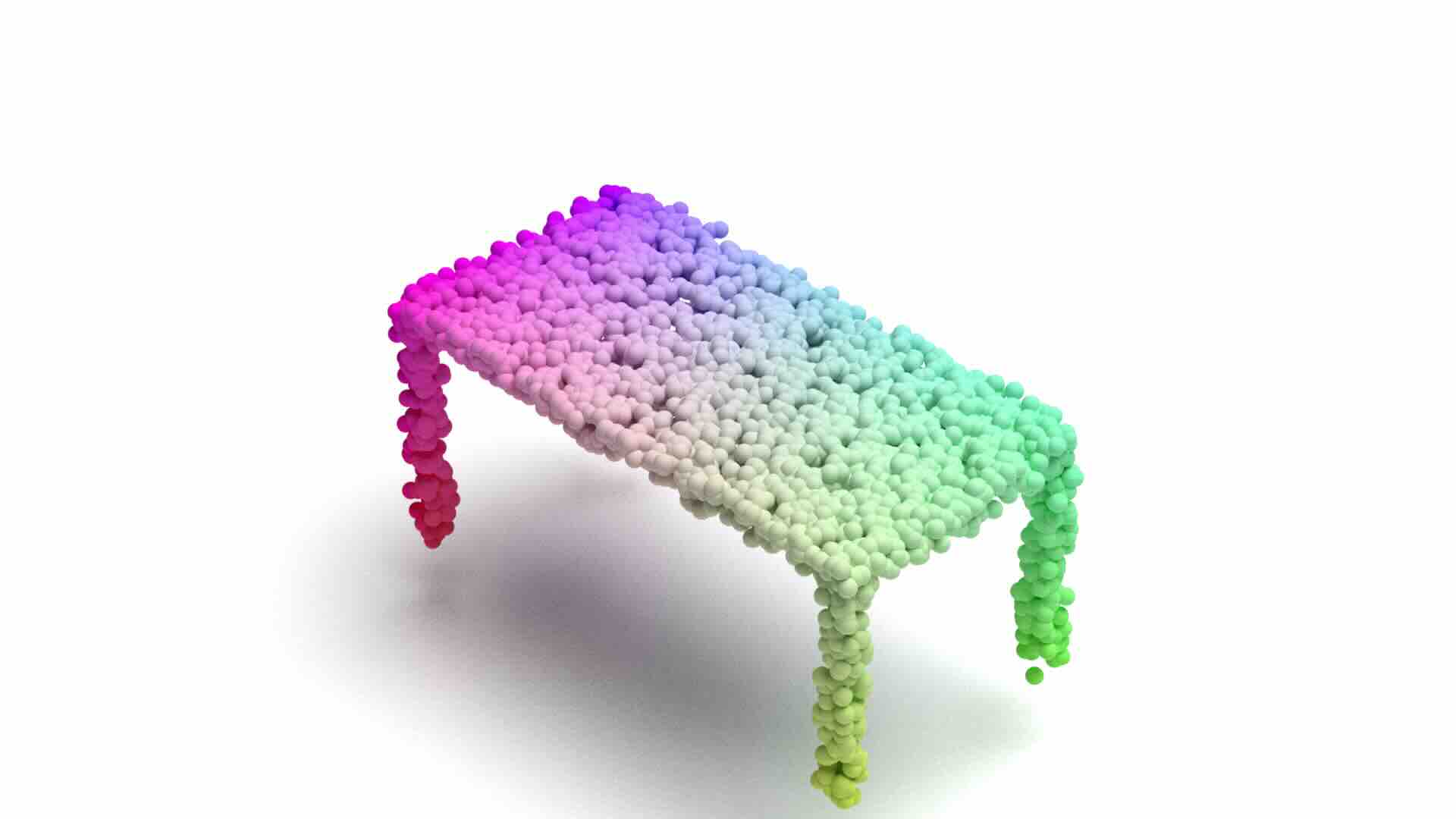} &
        \includegraphics[trim={15cm 0.0cm 15cm 0.0cm},clip,width=0.12\textwidth]{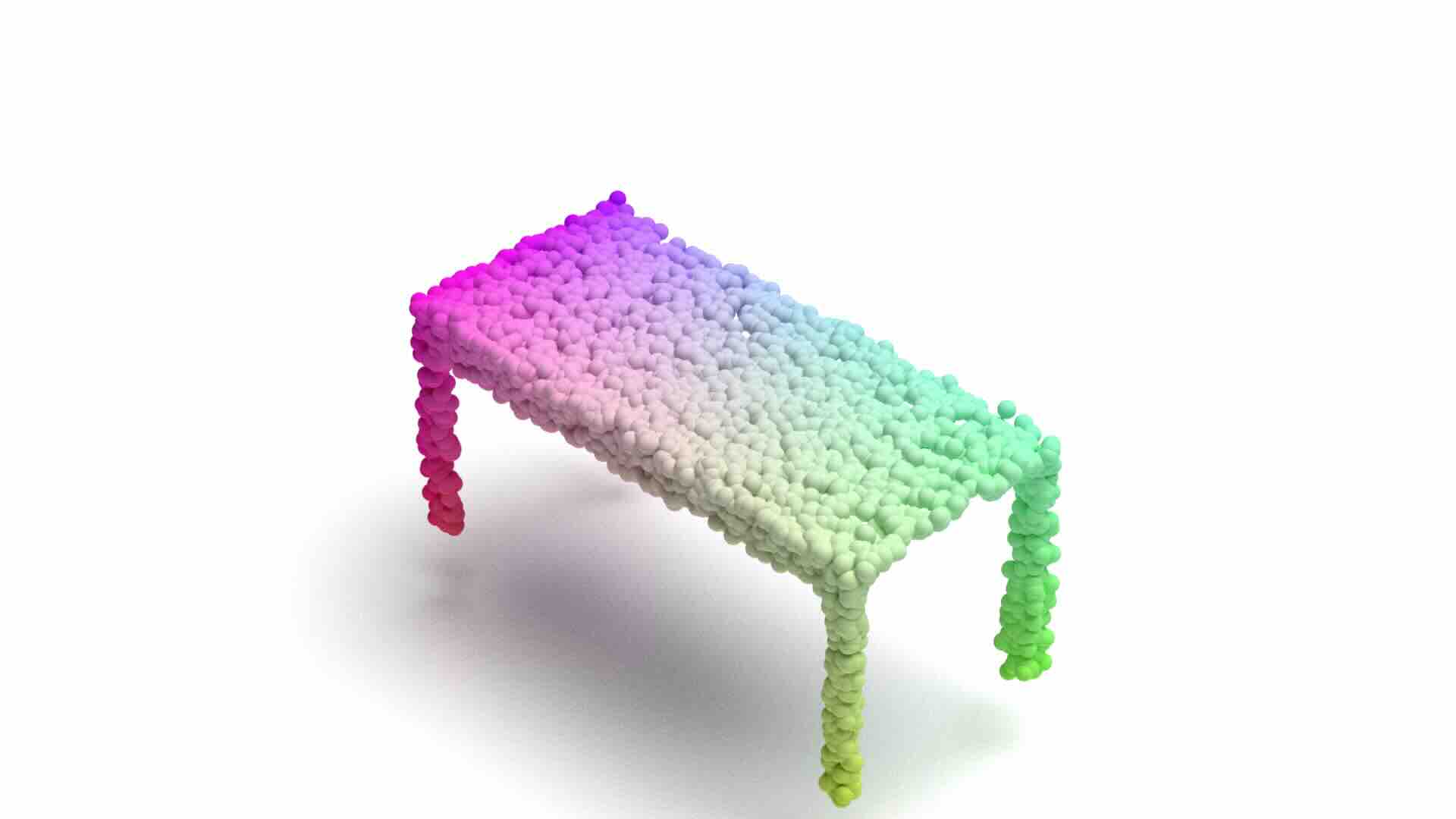} &
        \includegraphics[trim={15cm 0.0cm 15cm 0.0cm},clip,width=0.12\textwidth]{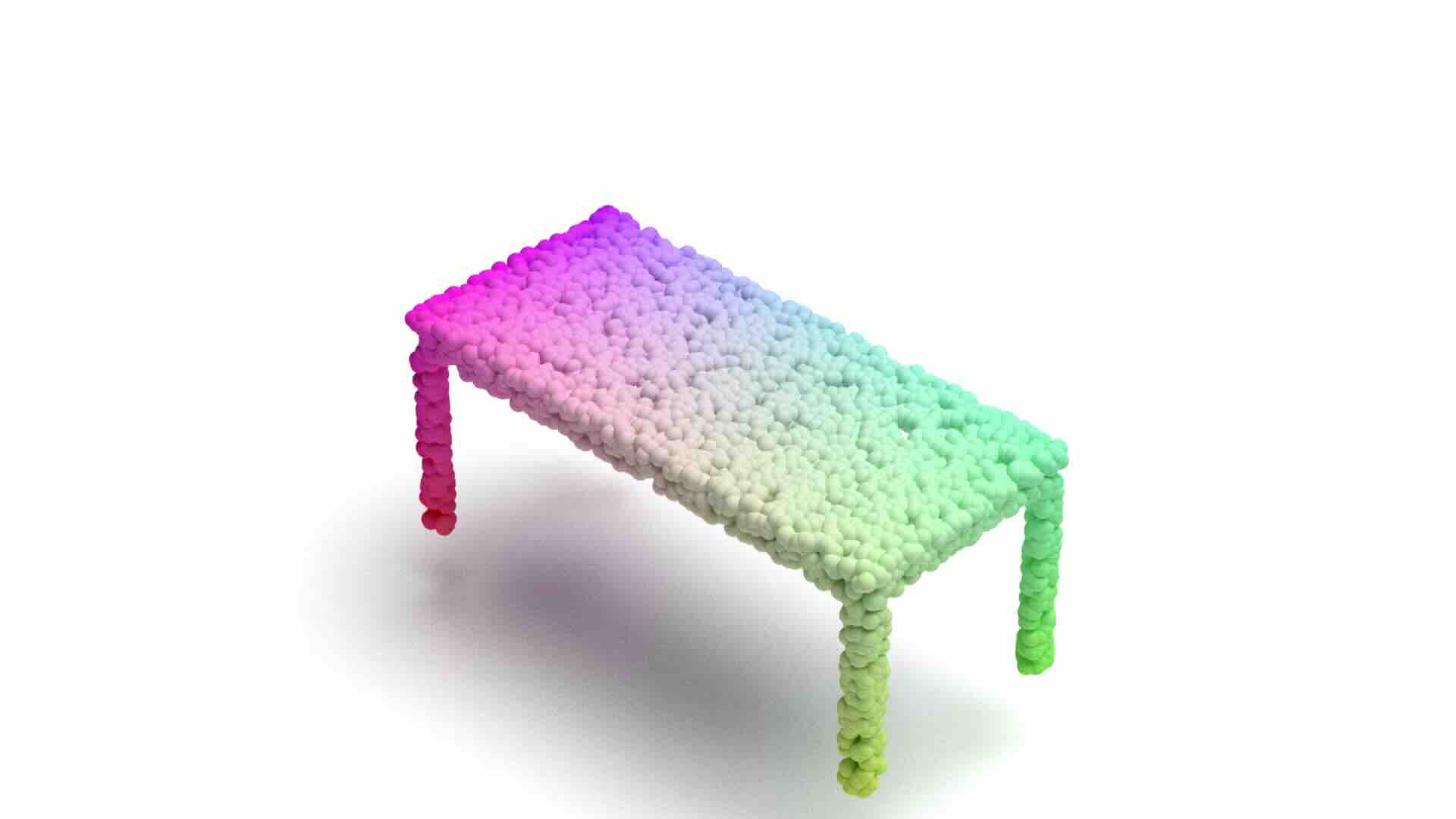} \\
        \scriptsize Fused Depths & 
\shortstack[c]{\scriptsize 3DGS-D \\ \scriptsize~\cite{kerbl20233d}} & 
\shortstack[c]{\scriptsize AGS-Mesh \\ \scriptsize~\cite{ren2025ags}} & 
\shortstack[c]{\scriptsize OM-GSD$^{\ast}$ \\ \scriptsize~\cite{lu2025orientation}} & 
\shortstack[c]{\scriptsize RVG-GSD$^{\ast}$ \\ \scriptsize~\cite{chang2025reconviagen}} & 
\shortstack[c]{\scriptsize SAM3D-GSD$^{\ast}$ \\ \scriptsize~\cite{chen2025sam}} & 
\scriptsize Ours & 
\scriptsize GT  \\
        \end{tabular}
        }
    \caption{\textbf{Qualitative Object Reconstruction Results.} The first three samples (top row) originate from the 3D-FRONT dataset~\cite{fu20213d}, the subsequent three (middle row) are drawn from ScanNet++~\cite{yeshwanth2023scannet++}, and the remaining three (bottom row) are drawn from ShapeR Evaluation Dataset~\cite{siddiqui2026shaper}. Compared to GS-based and purely generative baselines, our method reconstructs structurally sound geometry in heavily occluded regions while maintaining consistent metric scale.} 
    \label{fig:3d_qualitative}
    \vspace{-5mm}
\end{figure*}

\paragraph{Qualitative Results for Object Reconstruction in all the datasets.}
We provide additional visualizations of the reconstructed 3D geometries for all the datasets. By rendering the point-based representation directly from the 3D Gaussian centers, we further demonstrate the efficacy of our method in capturing high-fidelity object structures~(see Fig.~\ref{fig:3d_qualitative}). The first three samples originate from the 3D-FRONT dataset~\cite{fu20213d}, the subsequent three are drawn from ScanNet++~\cite{yeshwanth2023scannet++}, and the remaining three are drawn from ShapeR Evaluation Dataset~\cite{siddiqui2026shaper}. As illustrated in the visualizations, our method effectively reconstructs heavily occluded regions by leveraging generative priors to synthesize plausible structures. In contrast, standard GS-based methods fail to recover geometry in areas with limited visibility. Furthermore, purely generative baselines struggle with scale ambiguity, frequently failing to align with the metric dimensions of real-world objects or producing unrealistic geometric hallucinations. Our dual-space guidance successfully addresses these challenges by anchoring generative priors within the reconstructed spatial context.

\begin{figure*}[ht!] 
    \centering
    \centering
    \setlength{\tabcolsep}{2pt}
    \resizebox{\textwidth}{!}{
    \begin{tabular}{cccccccc}
        \includegraphics[trim={0.0cm 0.0cm 0.0cm 0.7cm},clip,width=0.12\textwidth]{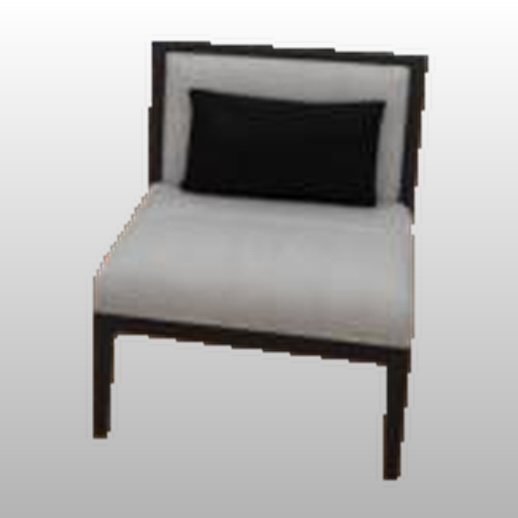} &
        \includegraphics[trim={8.0cm 2.0cm 7.0cm 7.3cm},clip,width=0.12\textwidth]{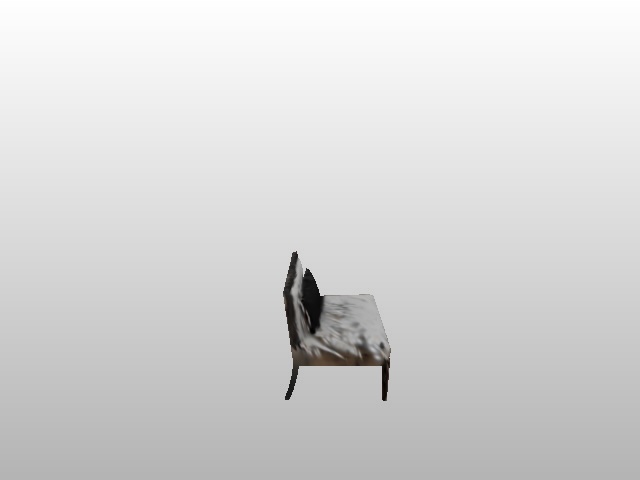} &
        \includegraphics[trim={8.0cm 2.0cm 7.0cm 7.3cm},clip,width=0.12\textwidth]{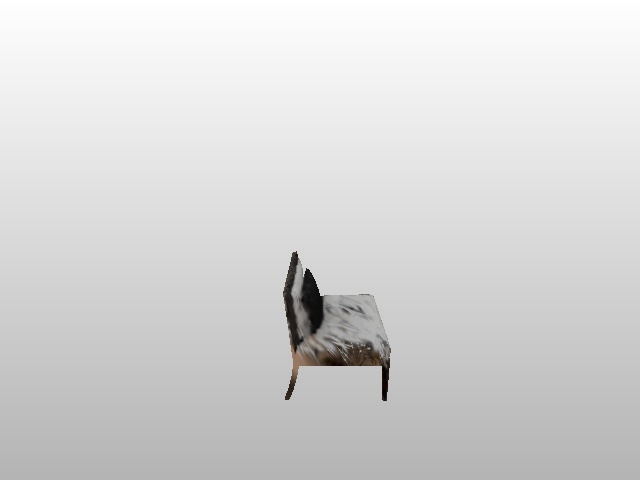} &
        \includegraphics[trim={8.0cm 2.0cm 7.0cm 7.3cm},clip,width=0.12\textwidth]{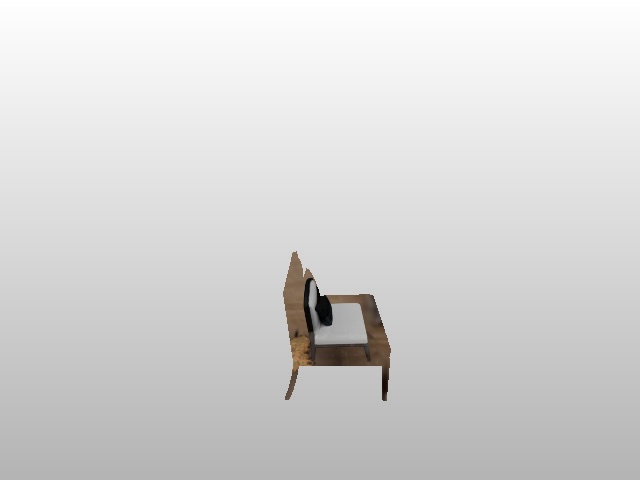} &
        \includegraphics[trim={8.0cm 2.0cm 7.0cm 7.3cm},clip,width=0.12\textwidth]{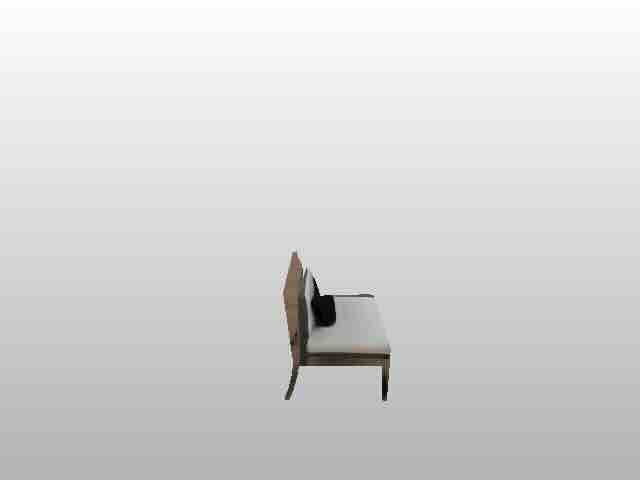} 
        &
        \includegraphics[trim={8.0cm 2.0cm 7.0cm 7.3cm},clip,width=0.12\textwidth]{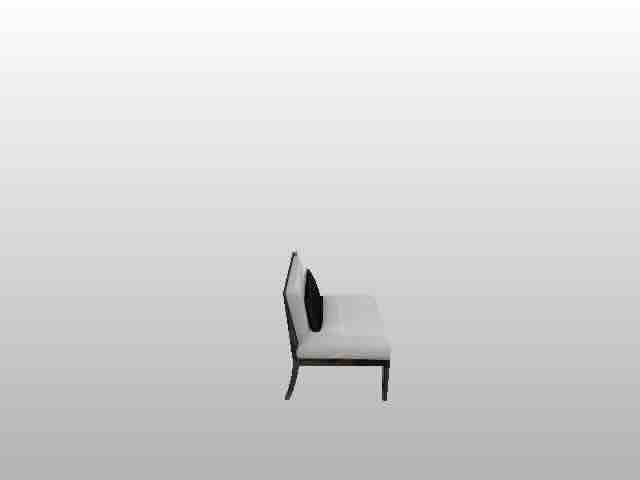} &
        \includegraphics[trim={8.0cm 2.0cm 7.0cm 7.3cm},clip,width=0.12\textwidth]{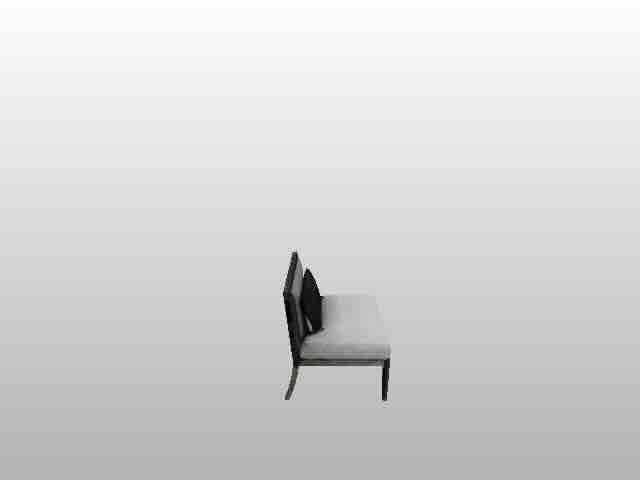} &
        \includegraphics[trim={8.0cm 2.0cm 7.0cm 7.3cm},clip,width=0.12\textwidth]{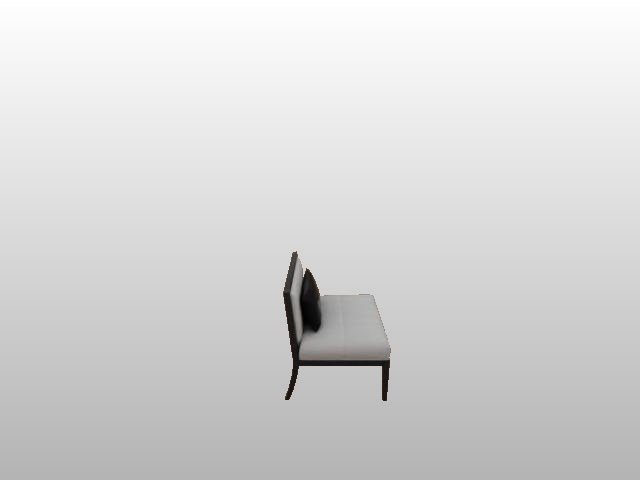}
       
        \\
        \includegraphics[width=0.12\textwidth]{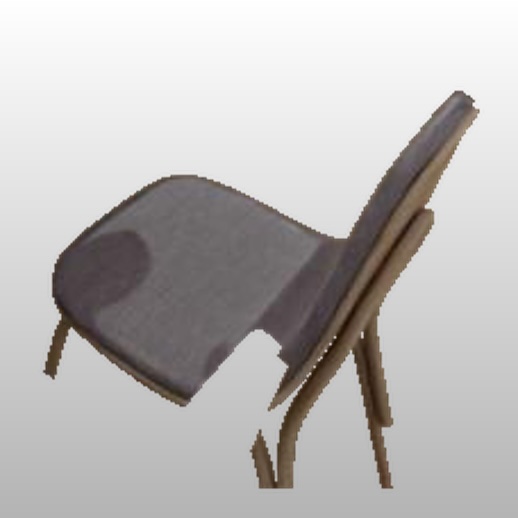} &
        \includegraphics[trim={5.0cm 1.0cm 10.0cm 7.3cm},clip,width=0.12\textwidth]{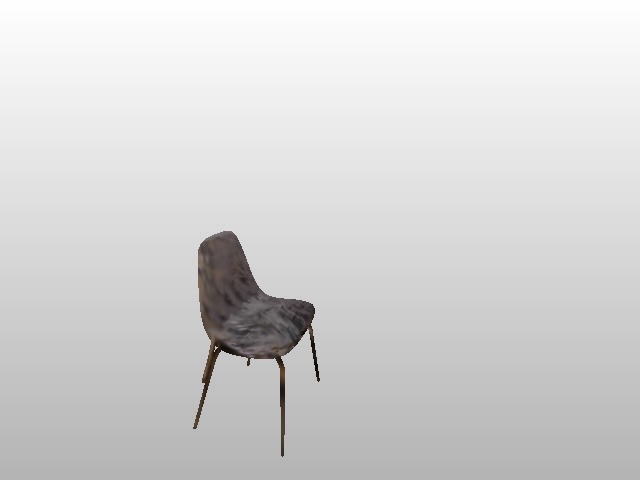} &
        \includegraphics[trim={5.0cm 1.0cm 10.0cm 7.3cm},clip,width=0.12\textwidth]{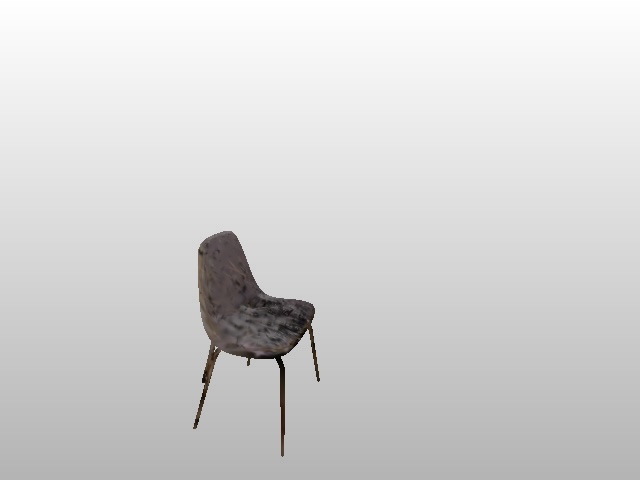} &
        \includegraphics[trim={5.0cm 1.0cm 10.0cm 7.3cm},clip,width=0.12\textwidth]{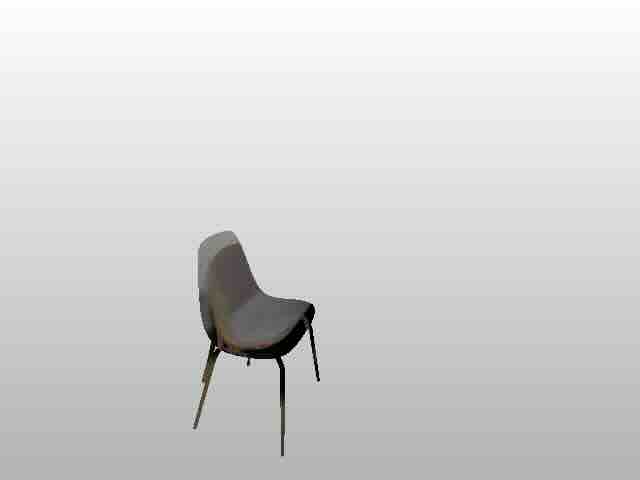} &
        \includegraphics[trim={5.0cm 1.0cm 10.0cm 7.3cm},clip,width=0.12\textwidth]{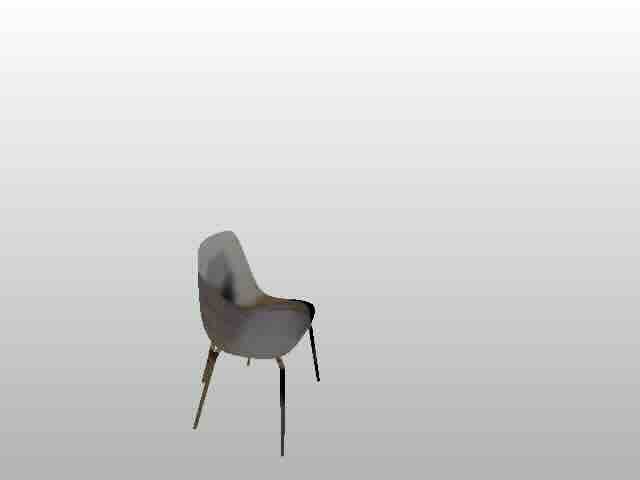}
        &
        \includegraphics[trim={5.0cm 1.0cm 10.0cm 7.3cm},clip,width=0.12\textwidth]{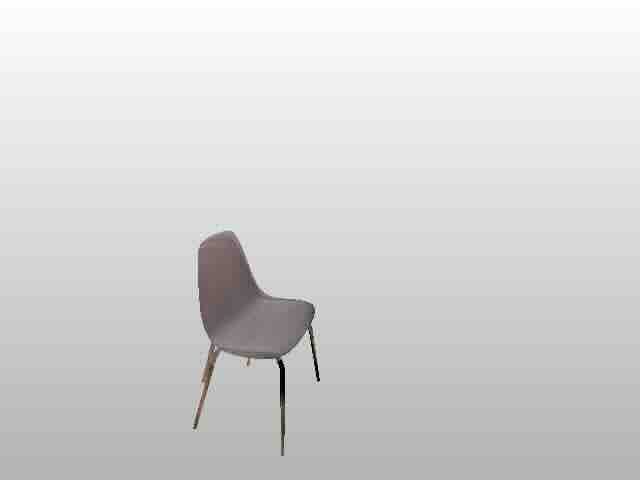}
        &
        \includegraphics[trim={5.0cm 1.0cm 10.0cm 7.3cm},clip,width=0.12\textwidth]{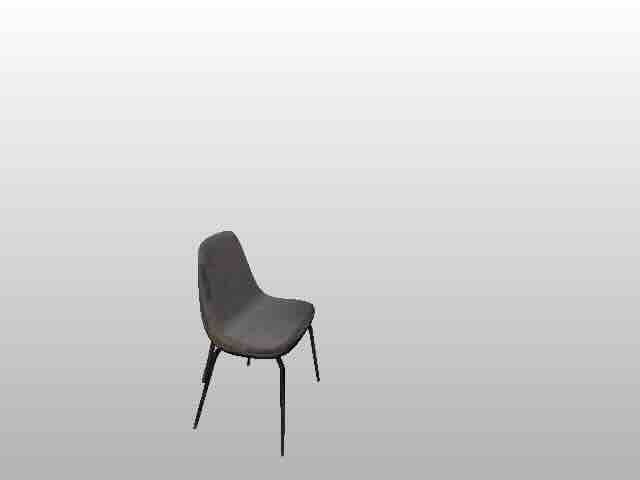} &
        \includegraphics[trim={5.0cm 1.0cm 10.0cm 7.3cm},clip,width=0.12\textwidth]{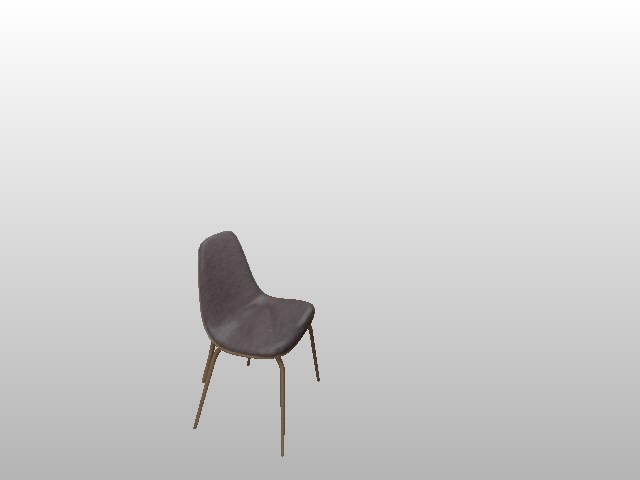}
        \\
        
        \includegraphics[trim={0.0cm 0.0cm 0.0cm 2.7cm},clip,width=0.12\textwidth]{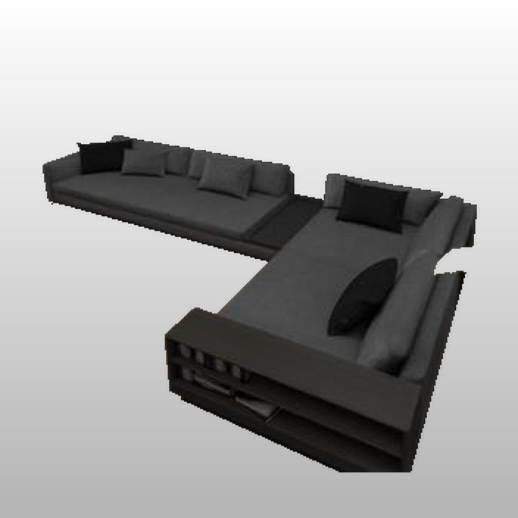} &
        \includegraphics[trim={5.0cm 0.0cm 0.0cm 2.0cm},clip,width=0.12\textwidth]{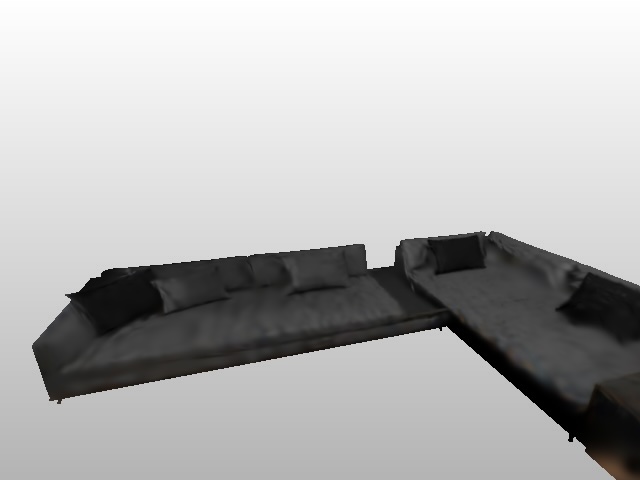} &
        \includegraphics[trim={5.0cm 0.0cm 0.0cm 2.0cm},clip,width=0.12\textwidth]{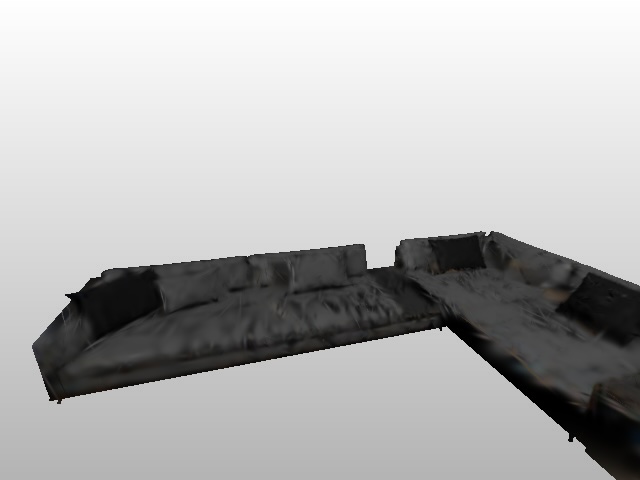} &
        \includegraphics[trim={5.0cm 0.0cm 0.0cm 2.0cm},clip,width=0.12\textwidth]{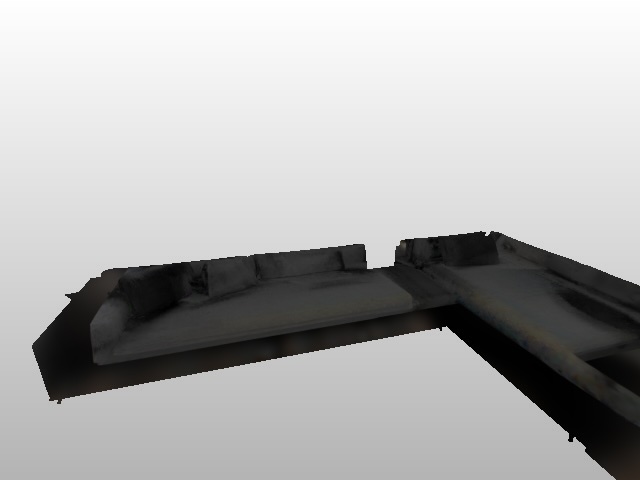} &
        \includegraphics[trim={5.0cm 0.0cm 0.0cm 2.0cm},clip,width=0.12\textwidth]{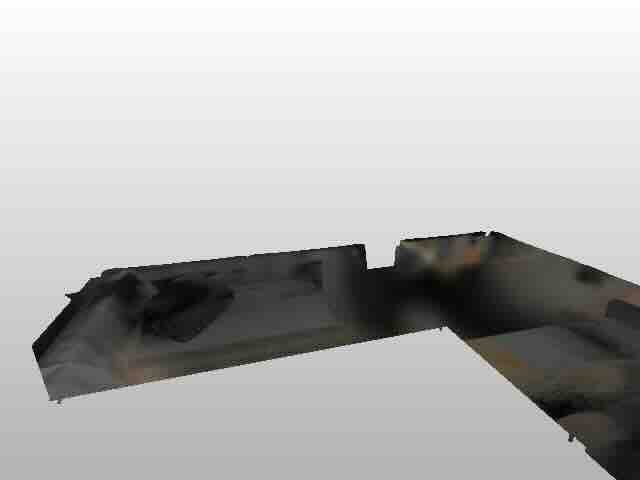}
        &
        \includegraphics[trim={5.0cm 0.0cm 0.0cm 2.0cm},clip,width=0.12\textwidth]{figures/3dfront_rendering/24_3dgs.jpg}
        &
        \includegraphics[trim={5.0cm 0.0cm 0.0cm 2.0cm},clip,width=0.12\textwidth]{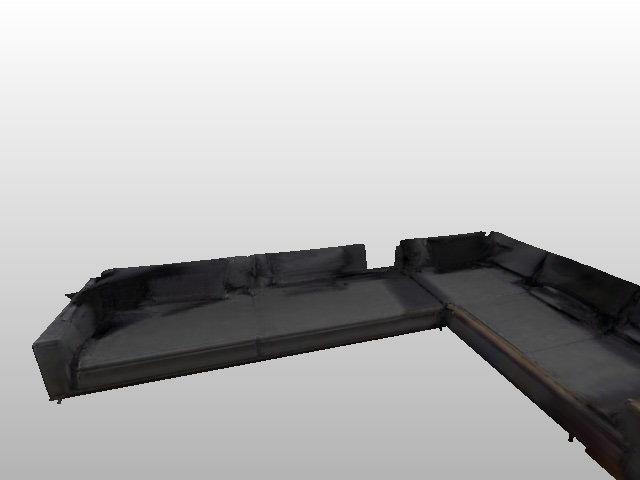} &
        \includegraphics[trim={5.0cm 0.0cm 0.0cm 2.0cm},clip,width=0.12\textwidth]{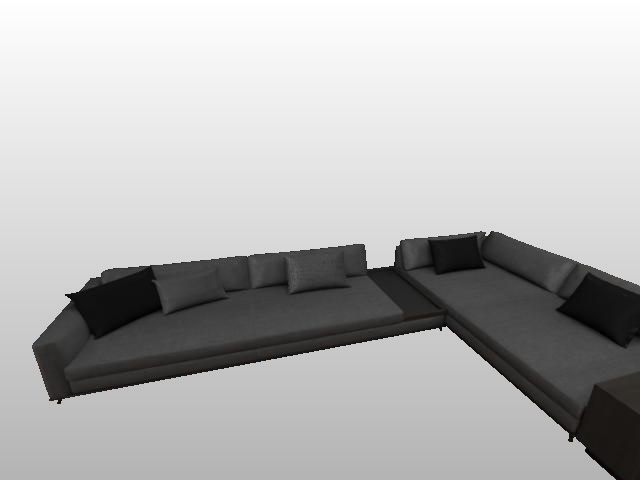}
        \\
        
        \includegraphics[trim={0.0cm 0.0cm 20.0cm 15.0cm},clip,width=0.12\textwidth]{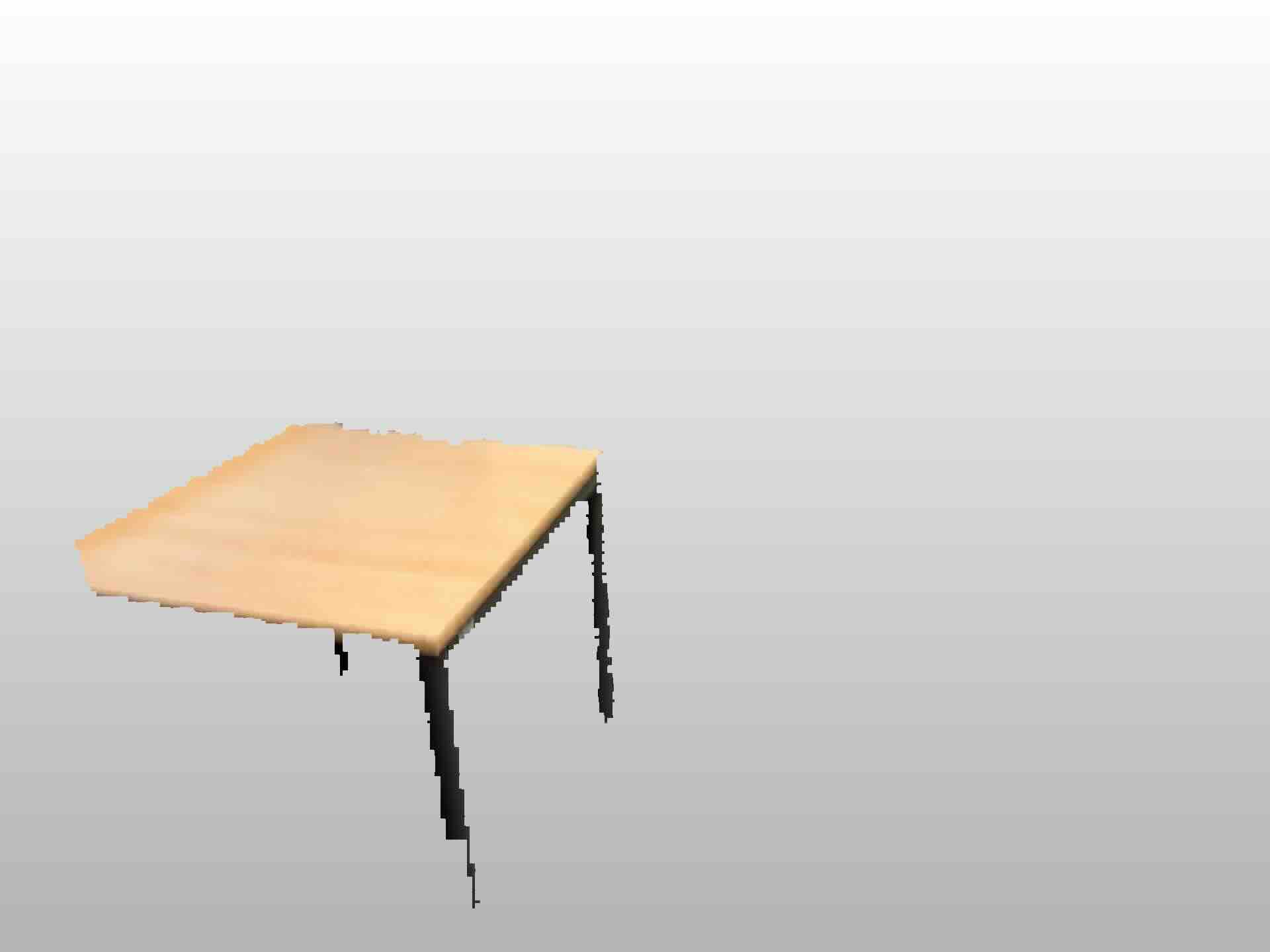} &
        \includegraphics[width=0.12\textwidth]{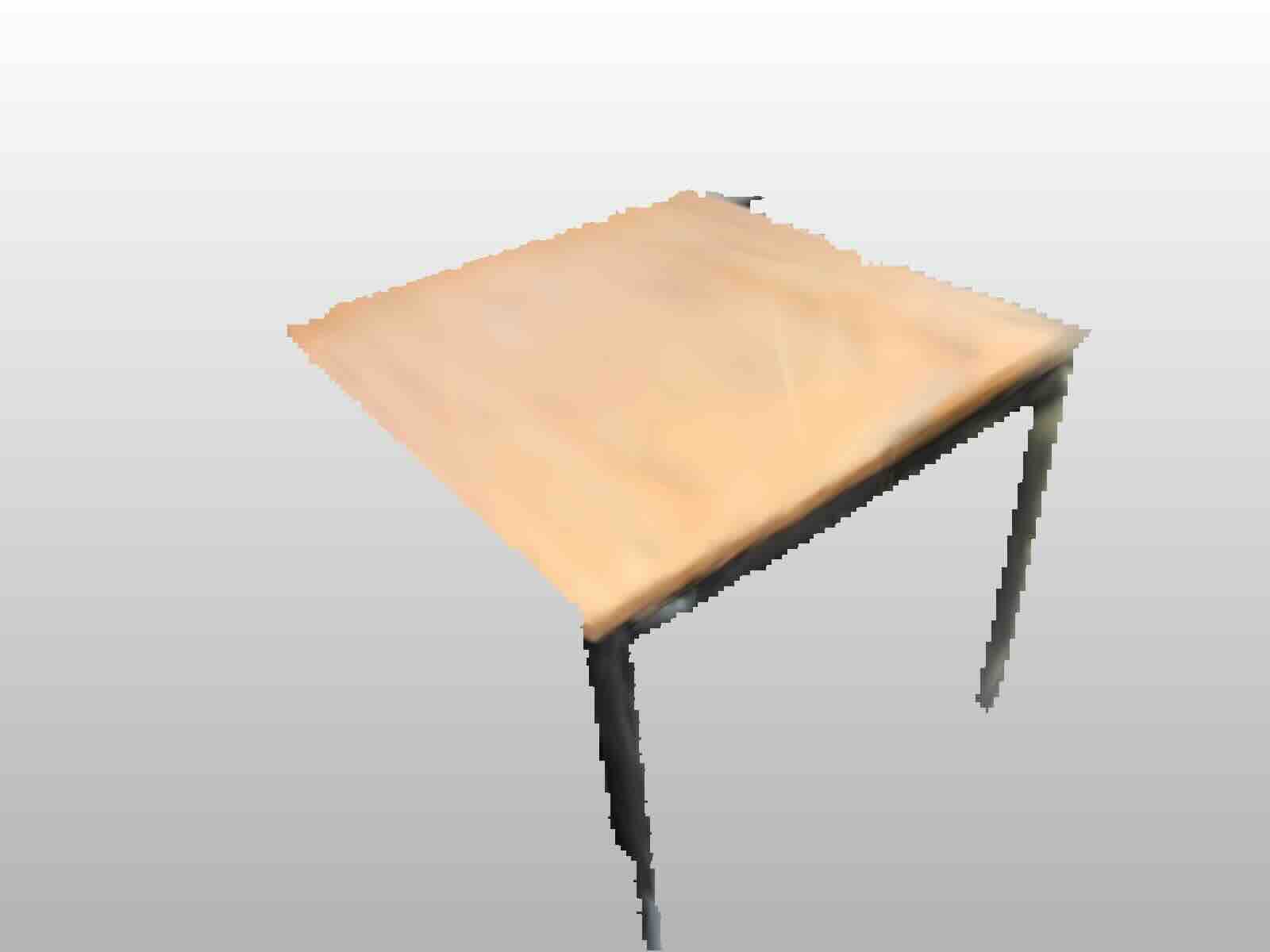} &
        \includegraphics[width=0.12\textwidth]{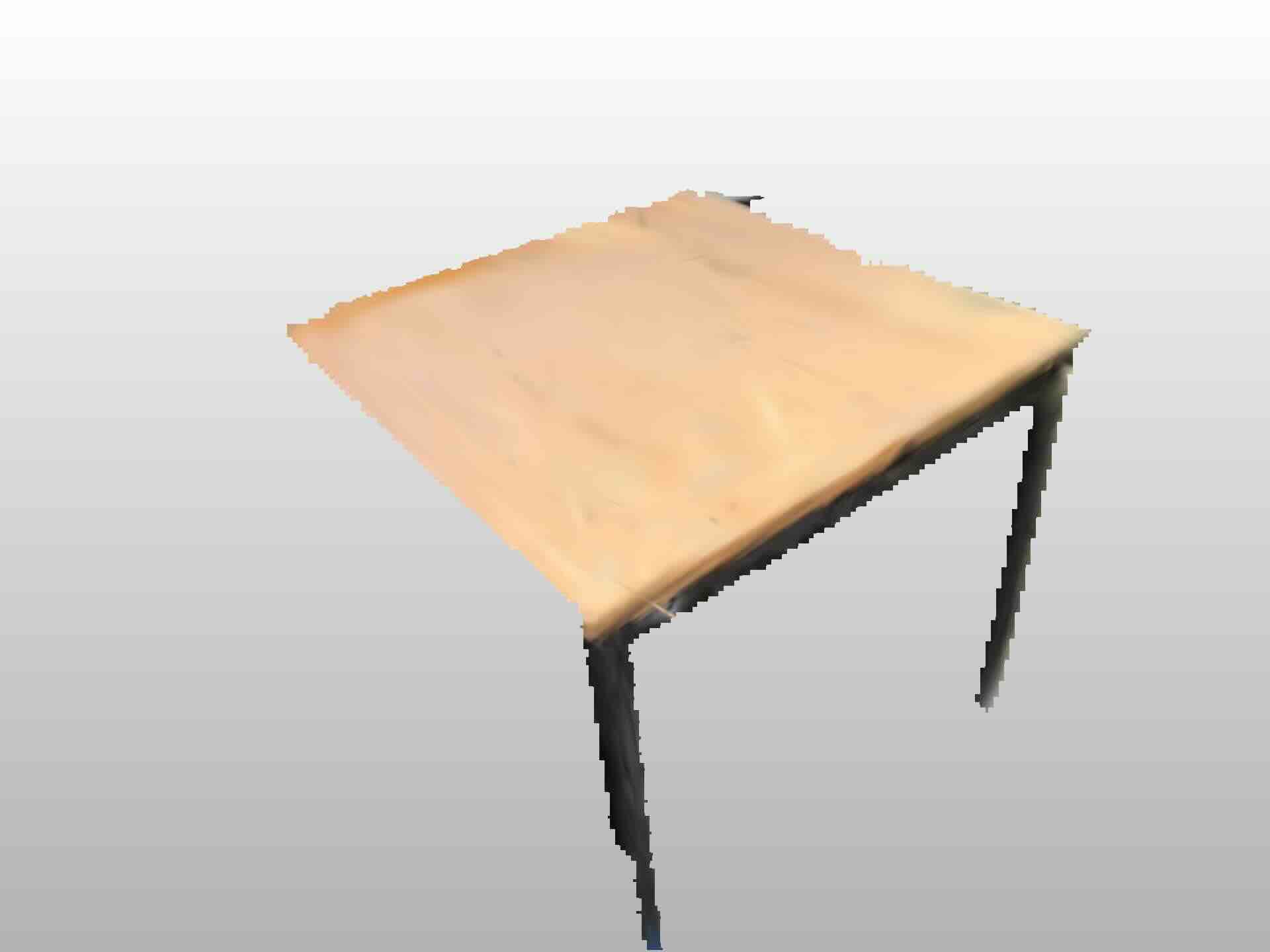} &
        \includegraphics[width=0.12\textwidth]{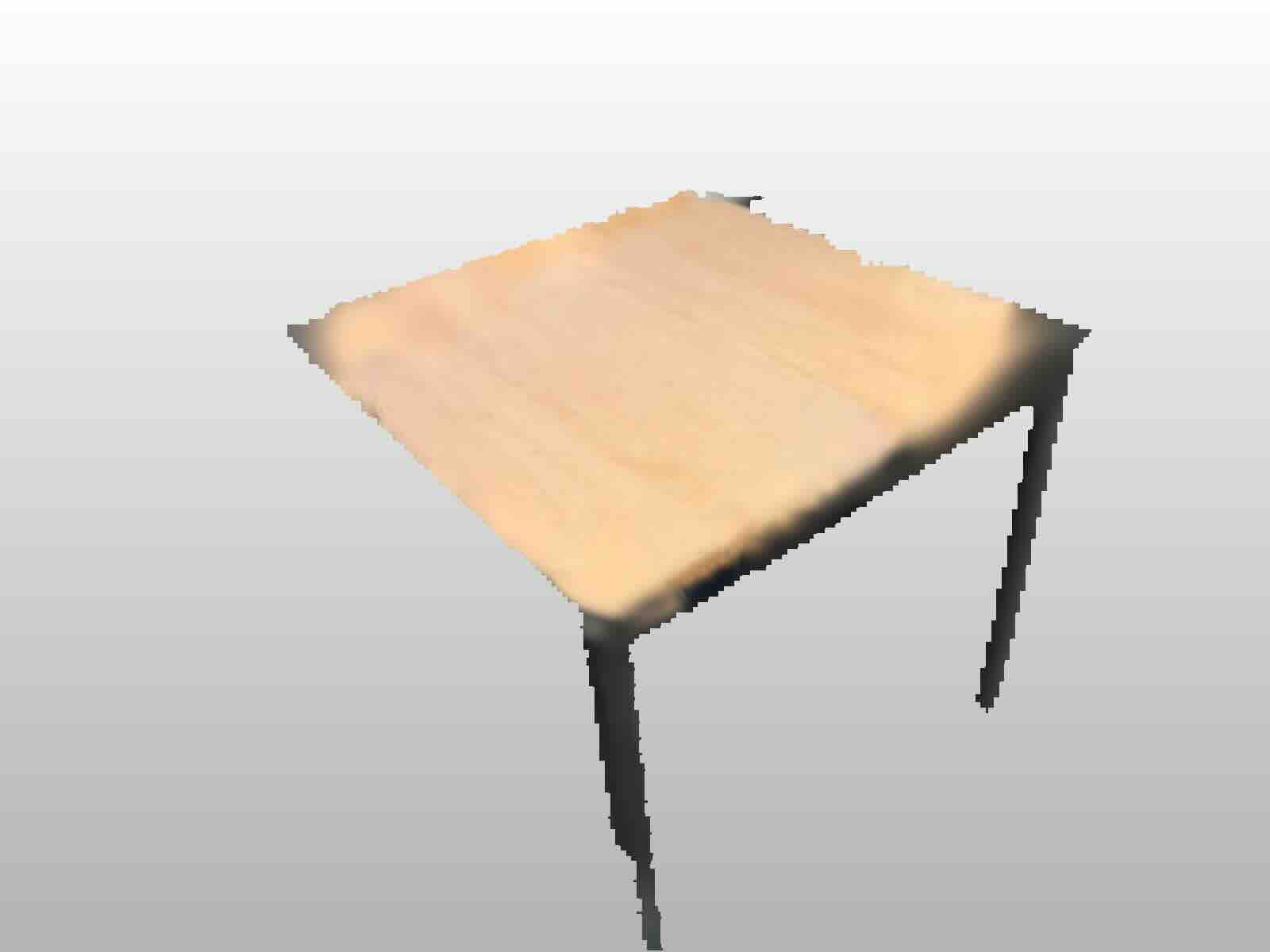}
        &
        \includegraphics[width=0.12\textwidth]{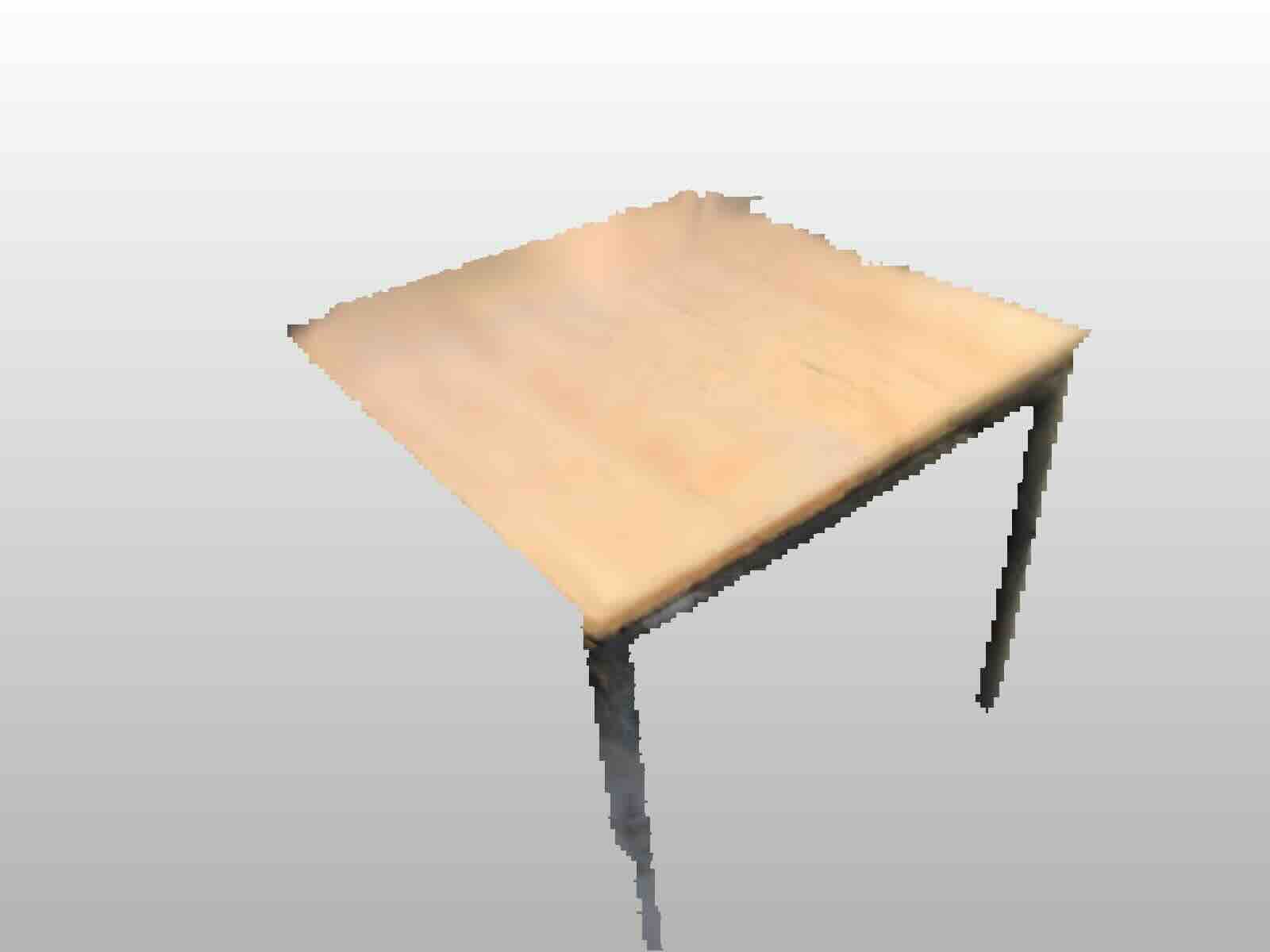}
        &
        \includegraphics[width=0.12\textwidth]{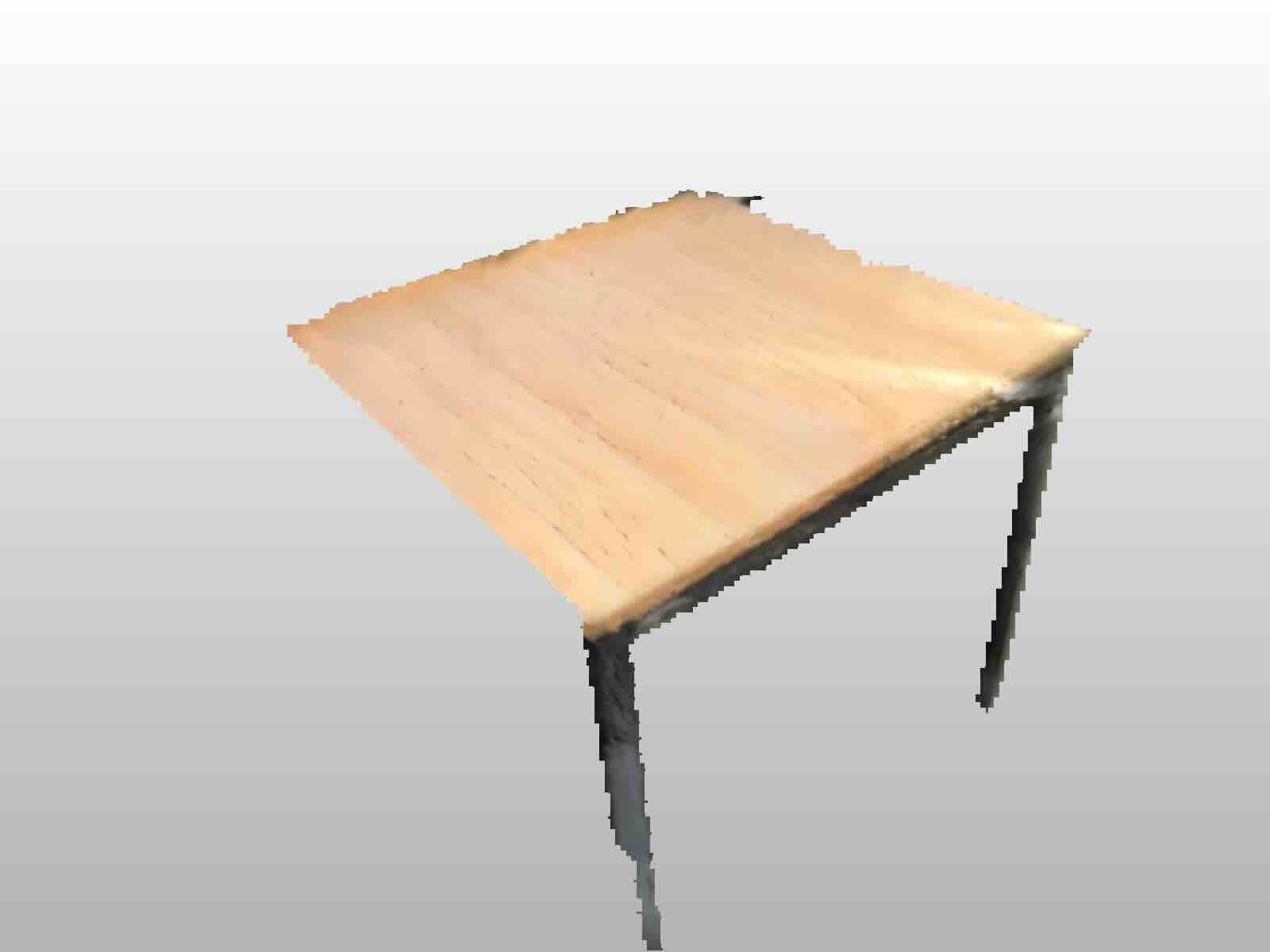} &
        \includegraphics[width=0.12\textwidth]{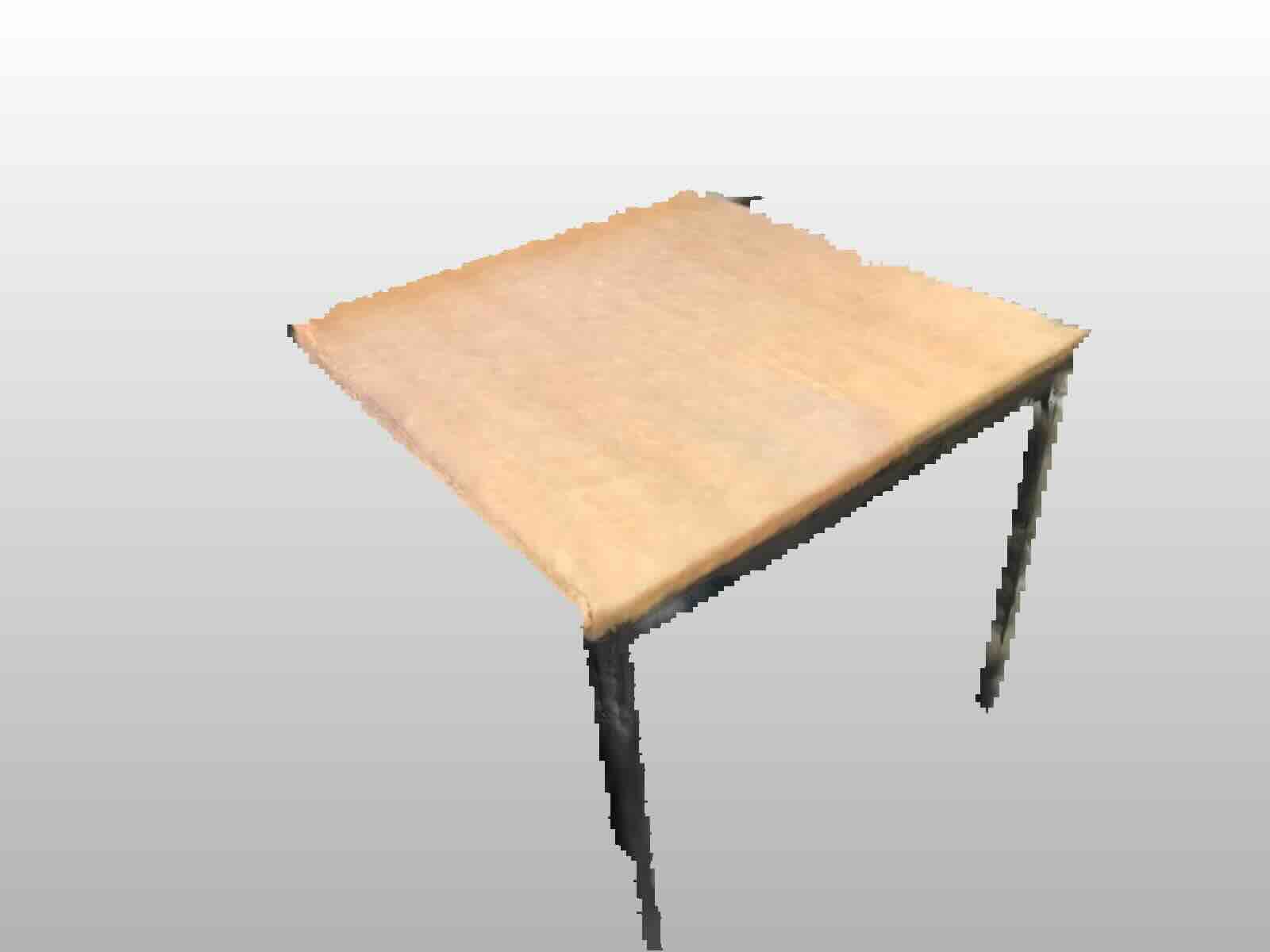} &
        \includegraphics[width=0.12\textwidth]{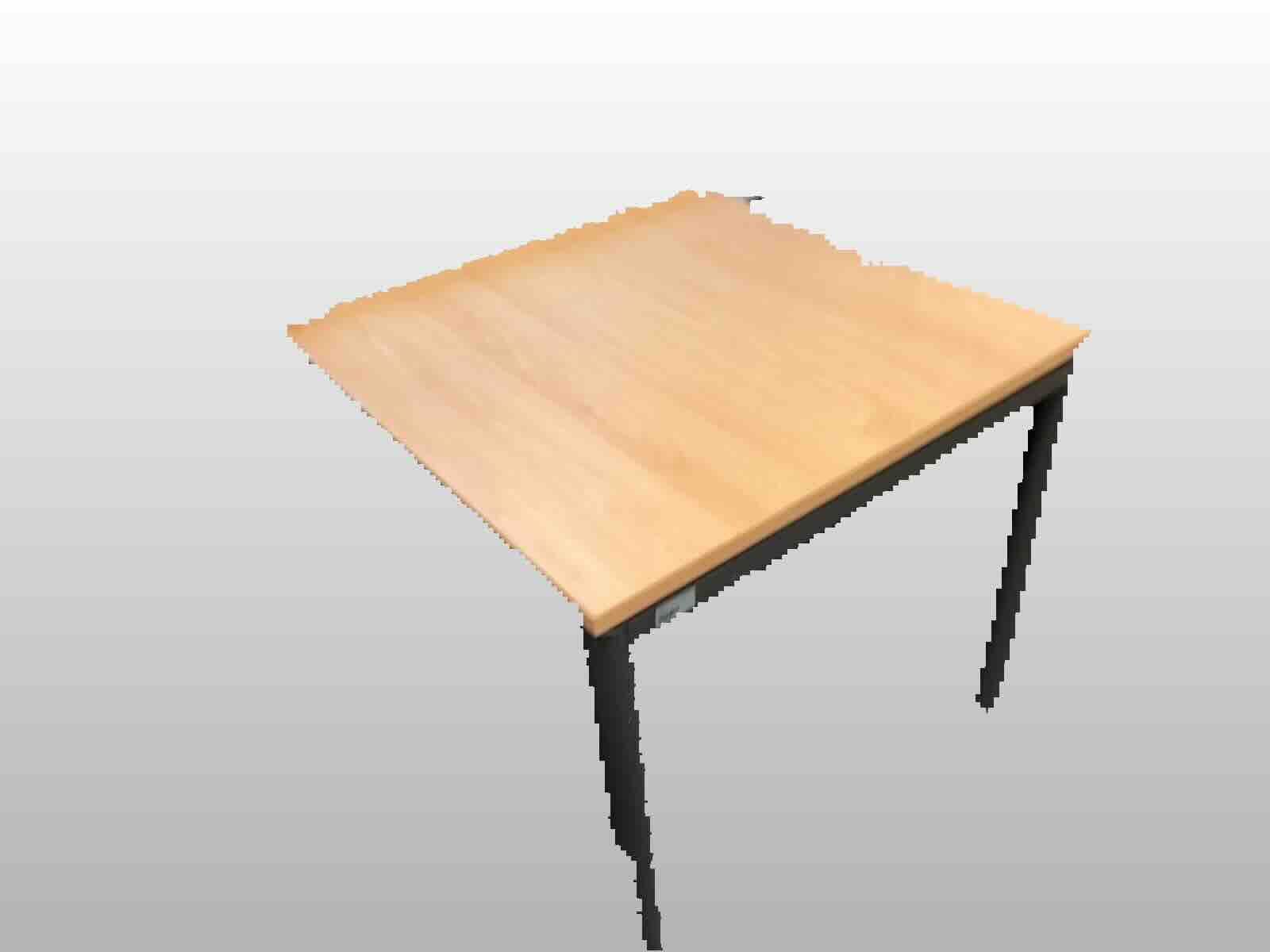}
        
        \\
        \includegraphics[trim={2.0cm 5.0cm 2.0cm 3.5cm},clip,width=0.12\textwidth]{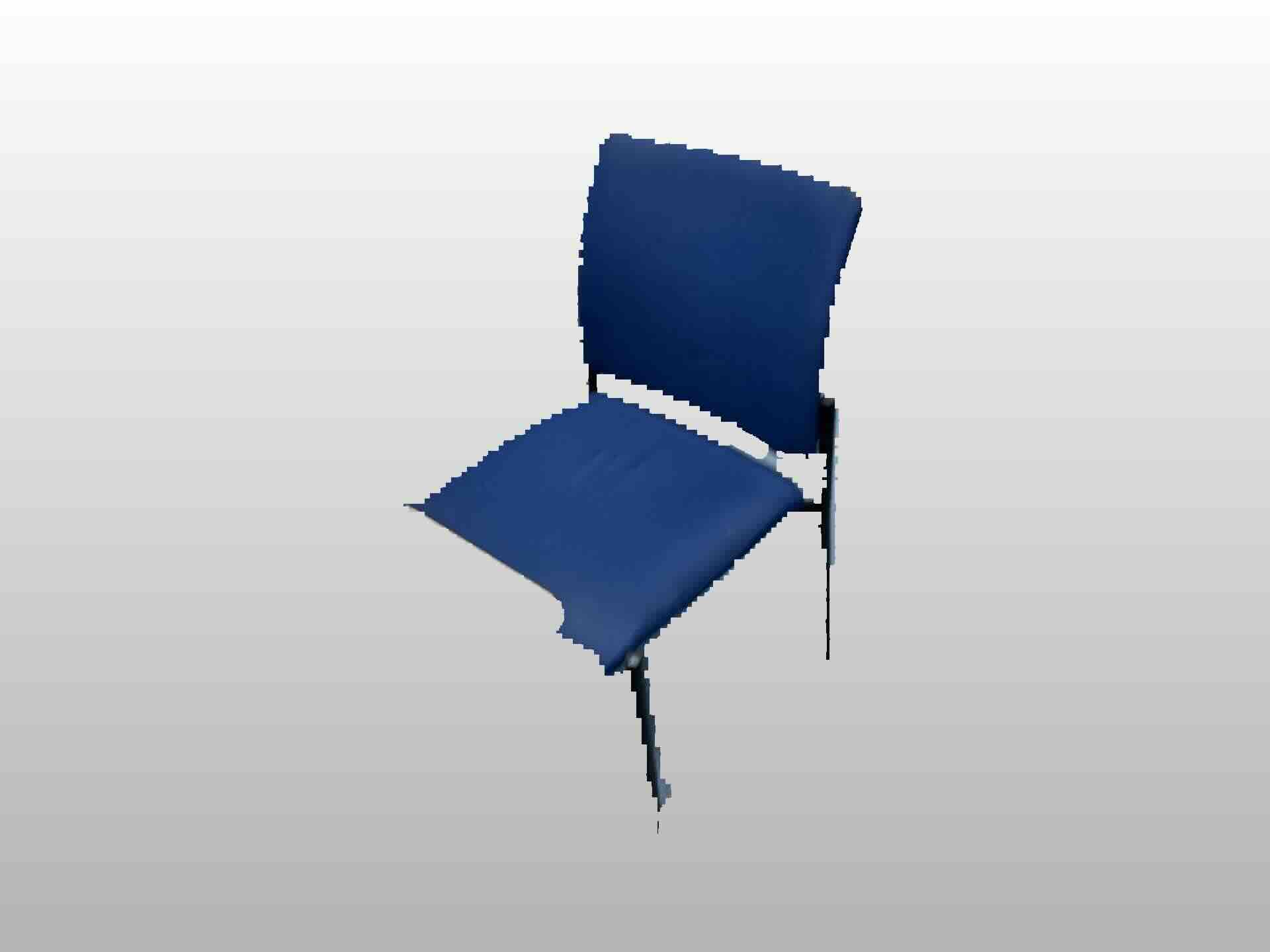} &
        \includegraphics[trim={3.0cm 0.0cm 1.0cm 8.0cm},clip,width=0.12\textwidth]{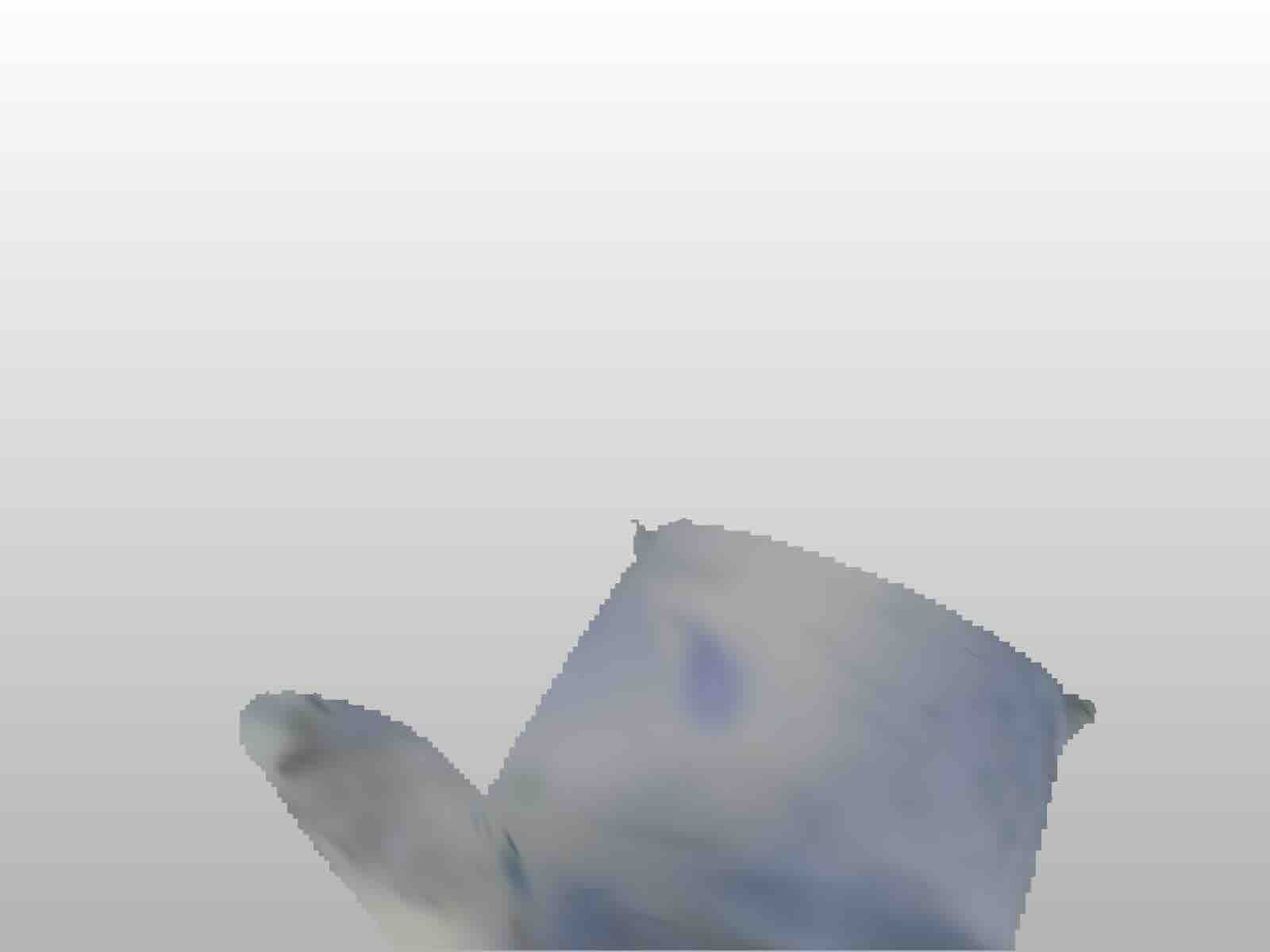} &
        \includegraphics[trim={3.0cm 0.0cm 1.0cm 8.0cm},clip,width=0.12\textwidth]{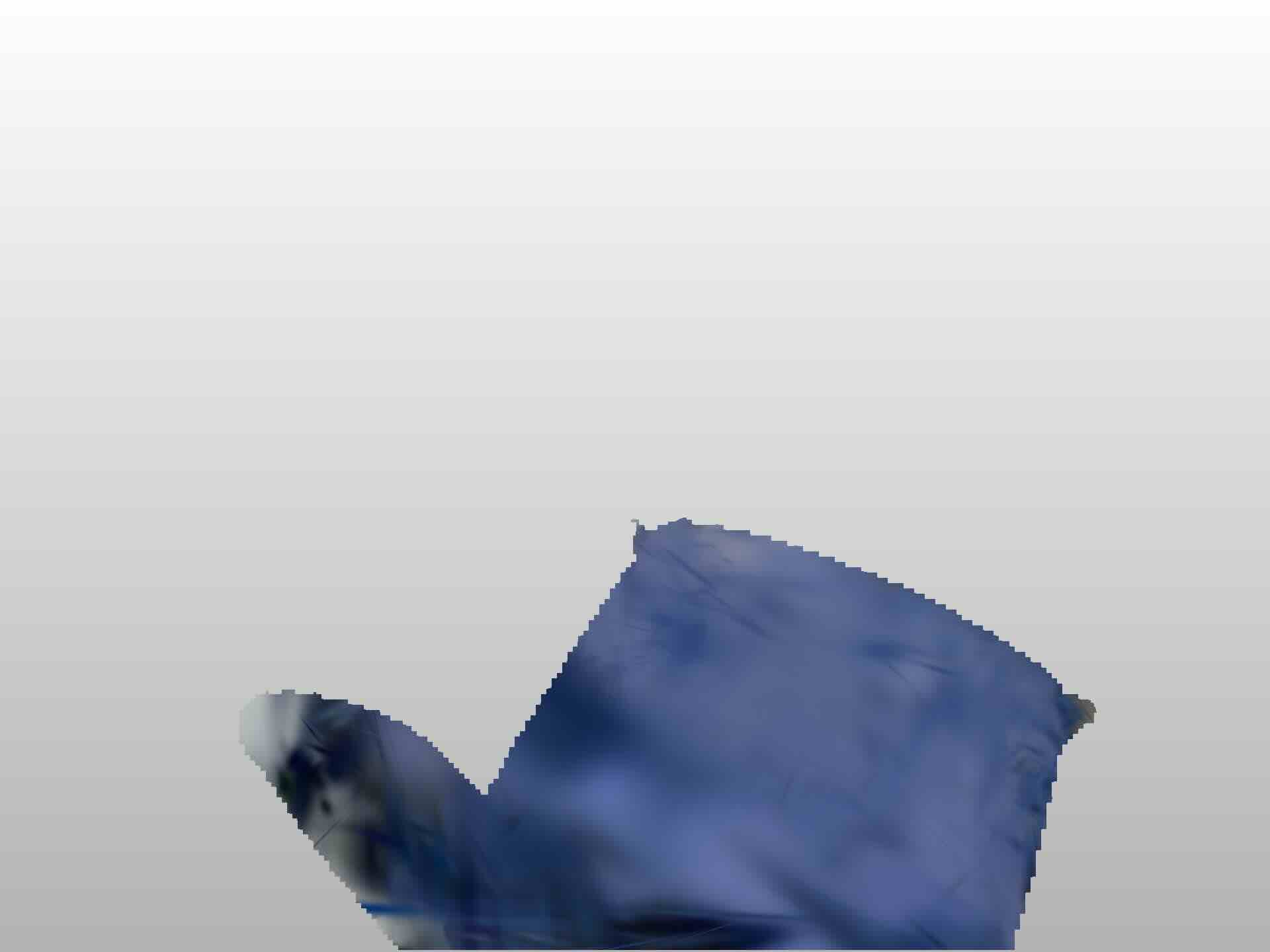} &
        \includegraphics[trim={3.0cm 0.0cm 1.0cm 8.0cm},clip,width=0.12\textwidth]{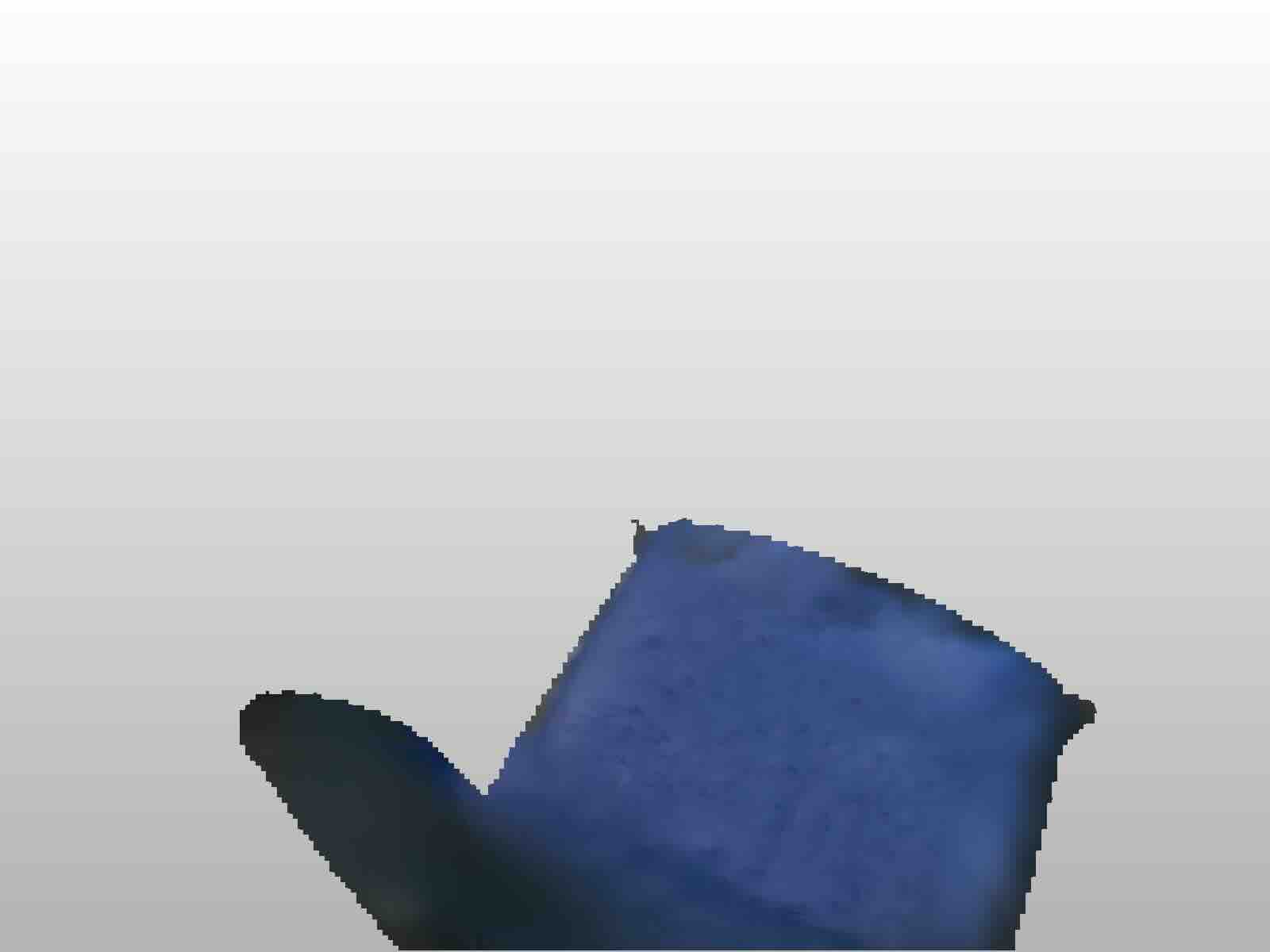} &
        \includegraphics[trim={3.0cm 0.0cm 1.0cm 8.0cm},clip,width=0.12\textwidth]{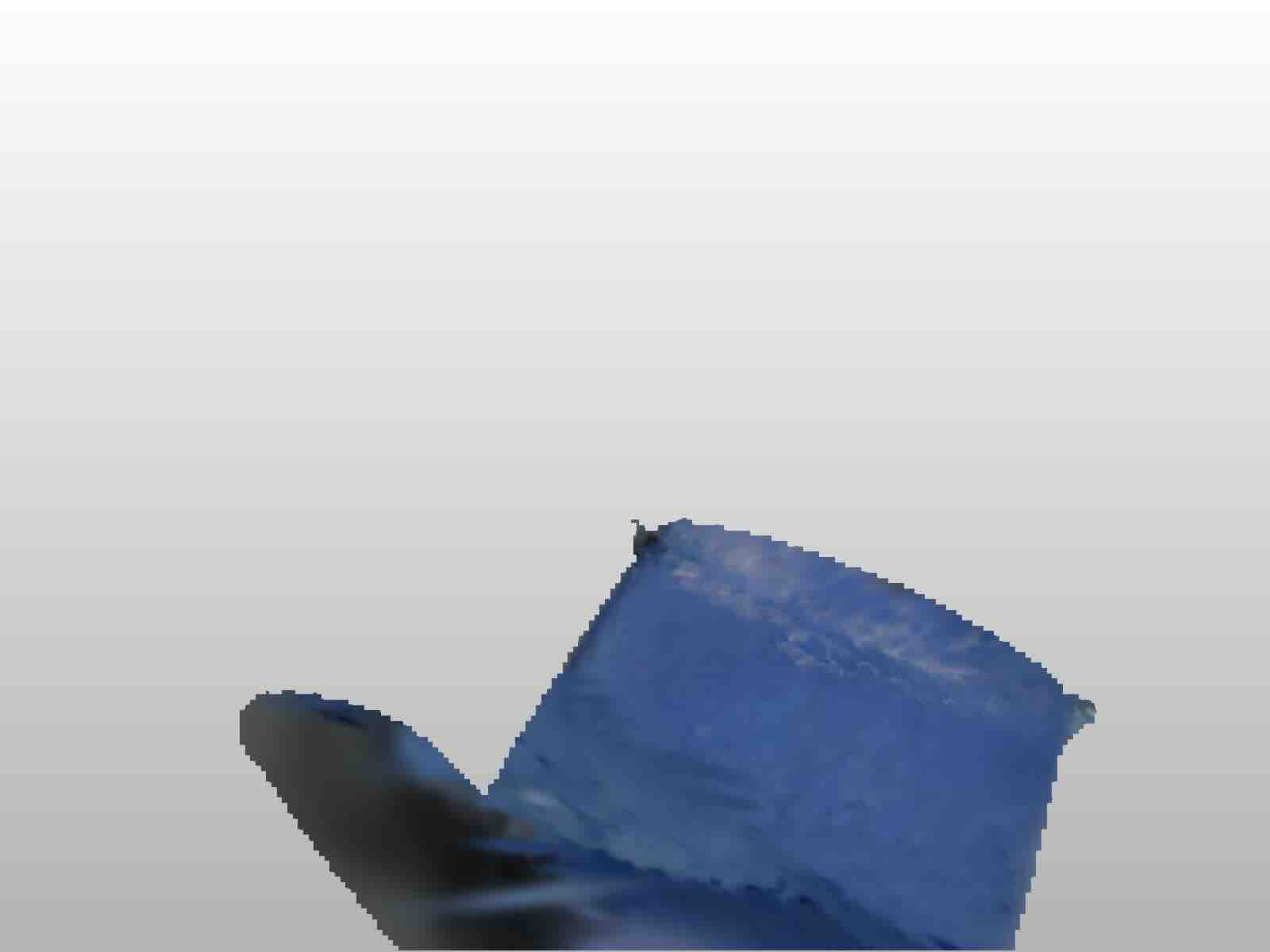} 
        &
        \includegraphics[trim={3.0cm 0.0cm 1.0cm 8.0cm},clip,width=0.12\textwidth]{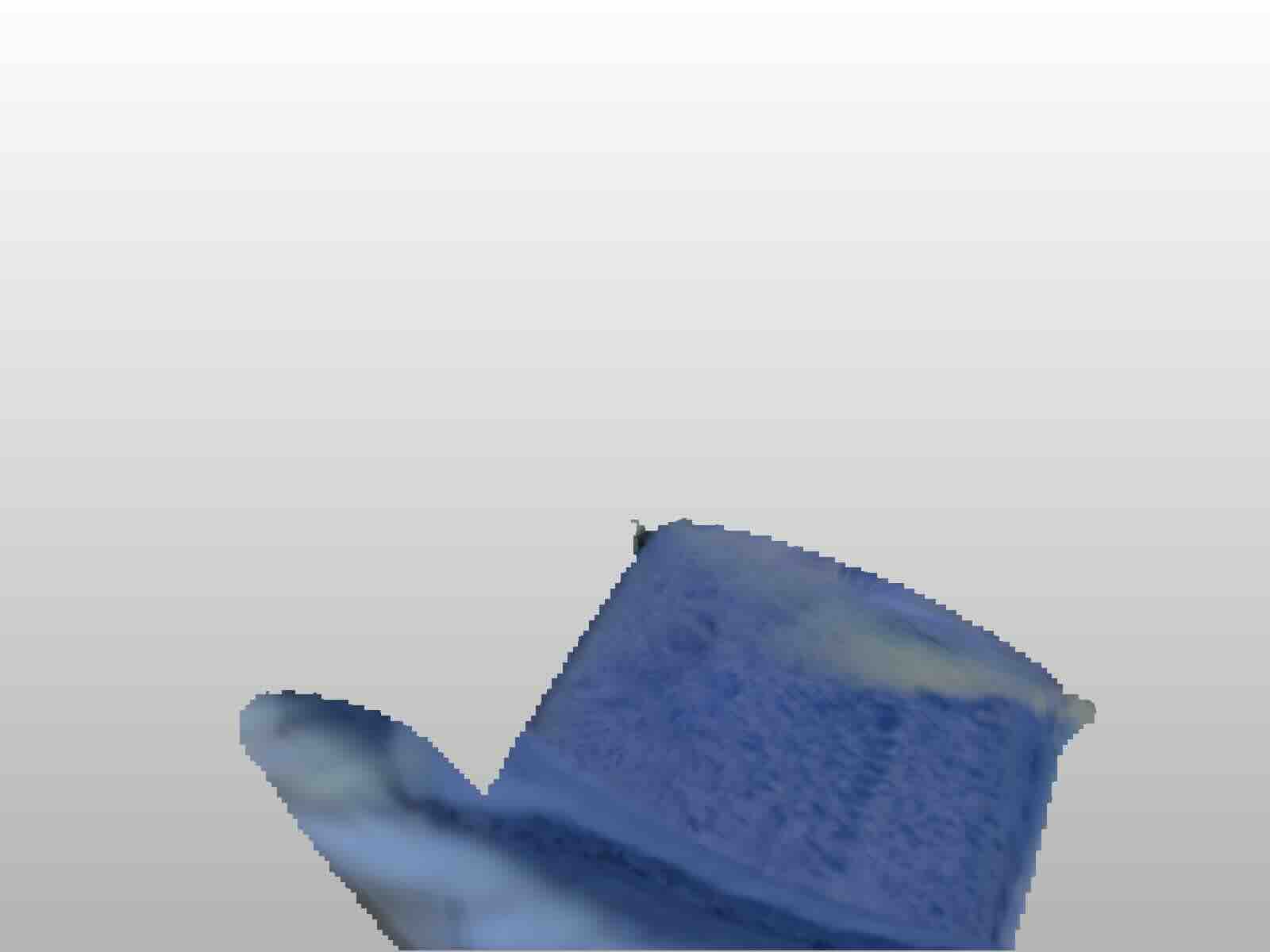} 
        &
        \includegraphics[trim={3.0cm 0.0cm 1.0cm 8.0cm},clip,width=0.12\textwidth]{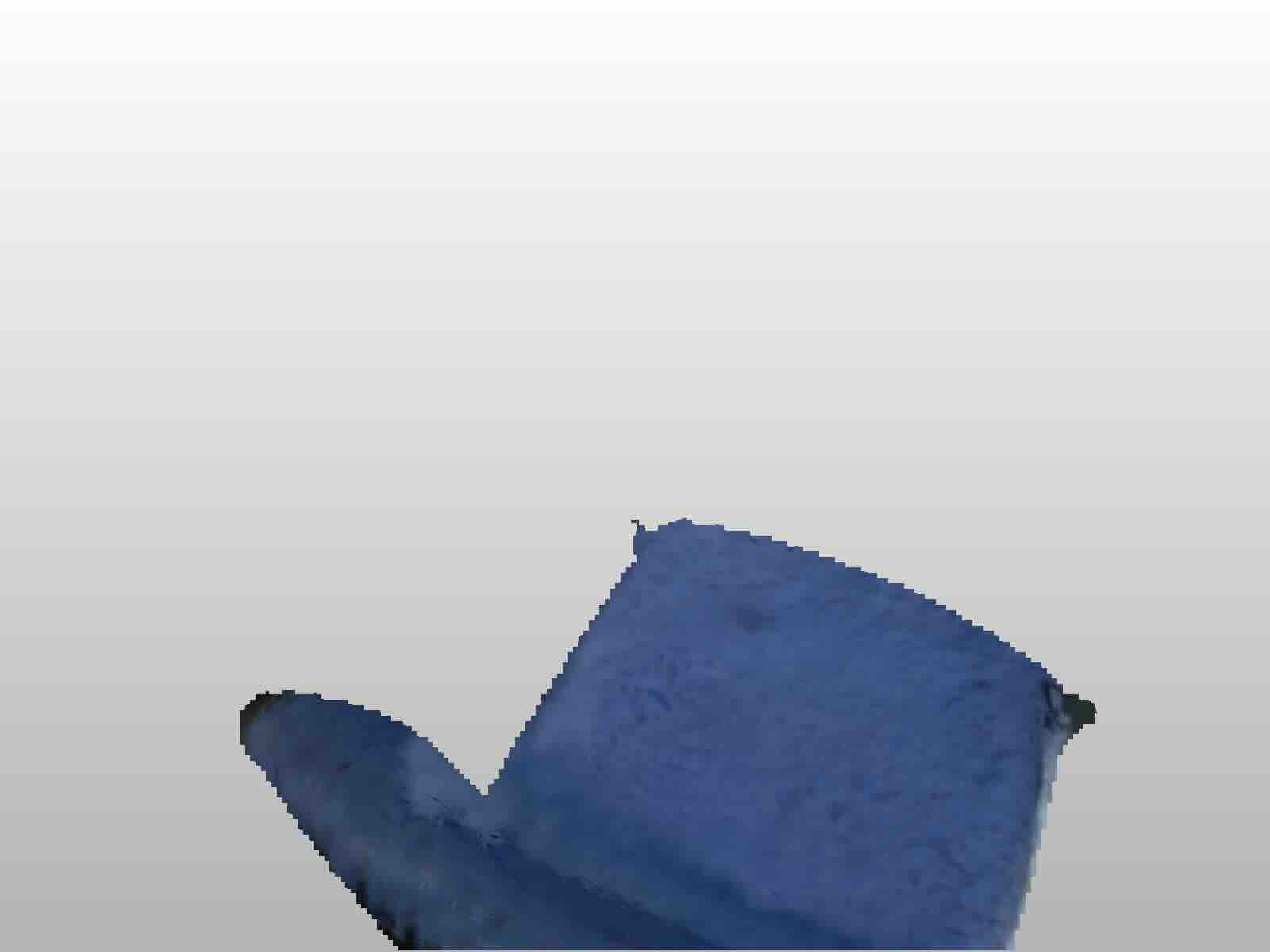} &
        \includegraphics[trim={3.0cm 0.0cm 1.0cm 8.0cm},clip,width=0.12\textwidth]{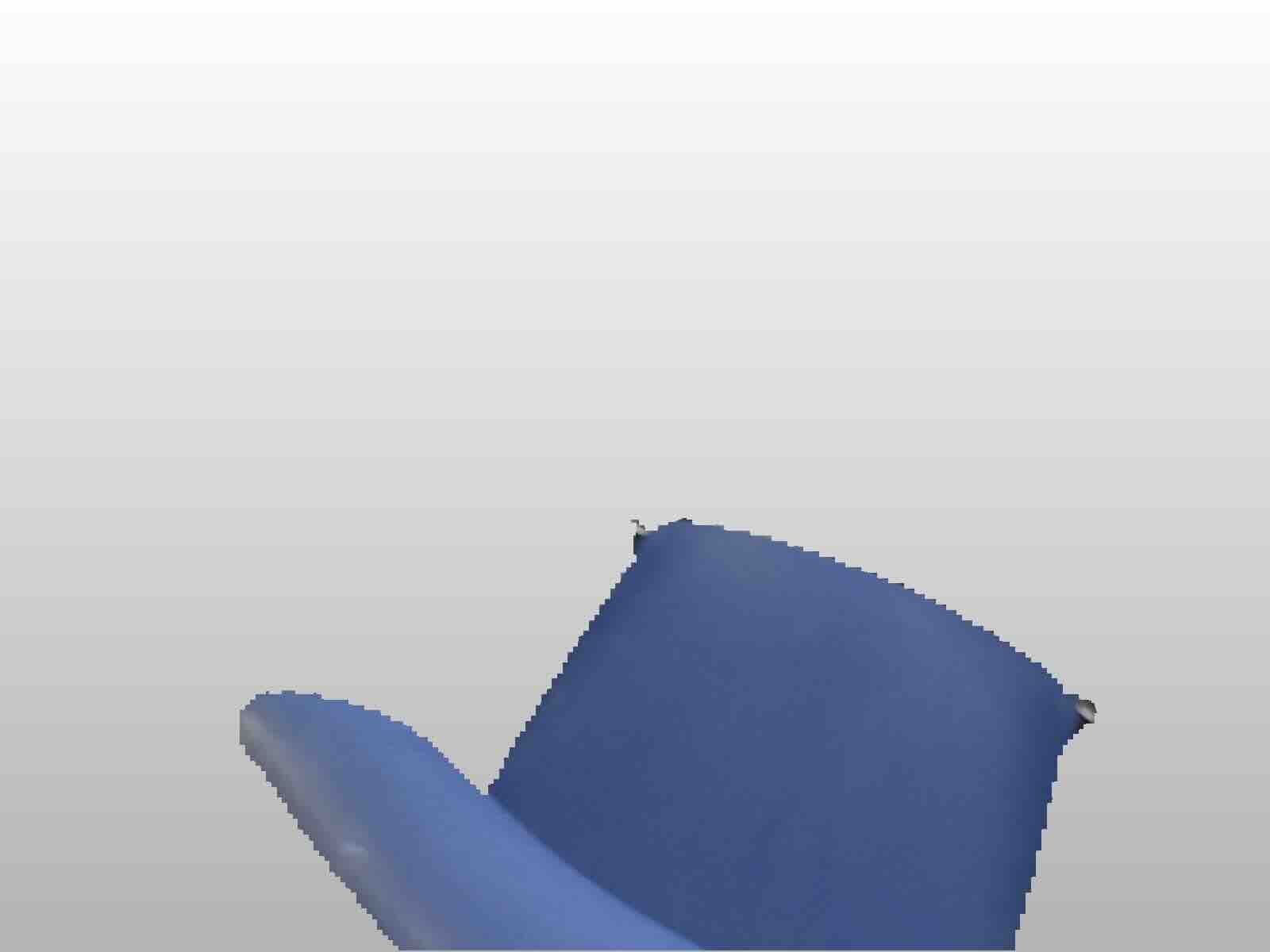}
 
        \\
        \includegraphics[trim={0.0cm 2.0cm 5.0cm 2.0cm},clip,width=0.12\textwidth]{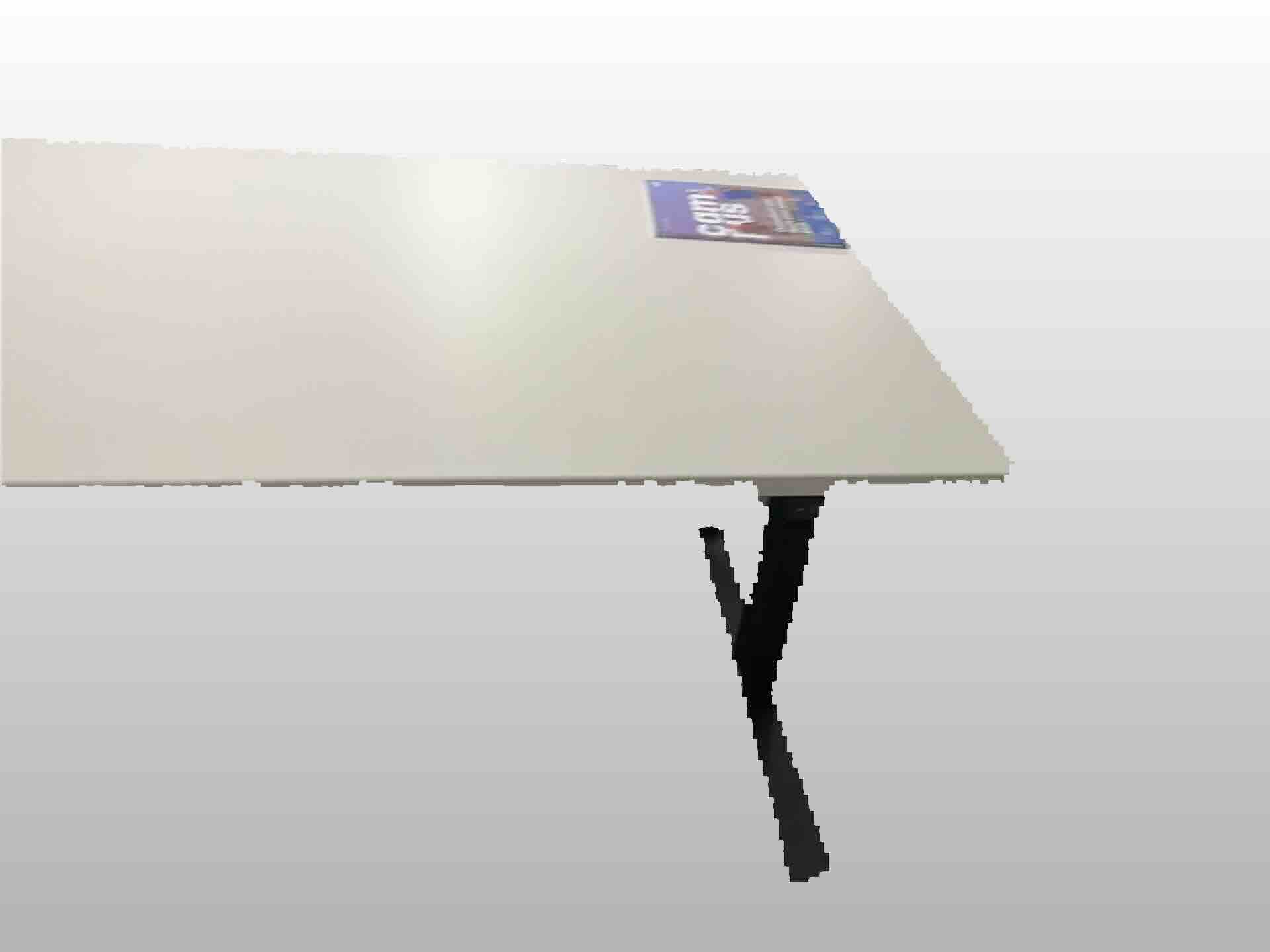} &
        \includegraphics[trim={5.0cm 1.0cm 7.0cm 7.5cm},clip,width=0.12\textwidth]{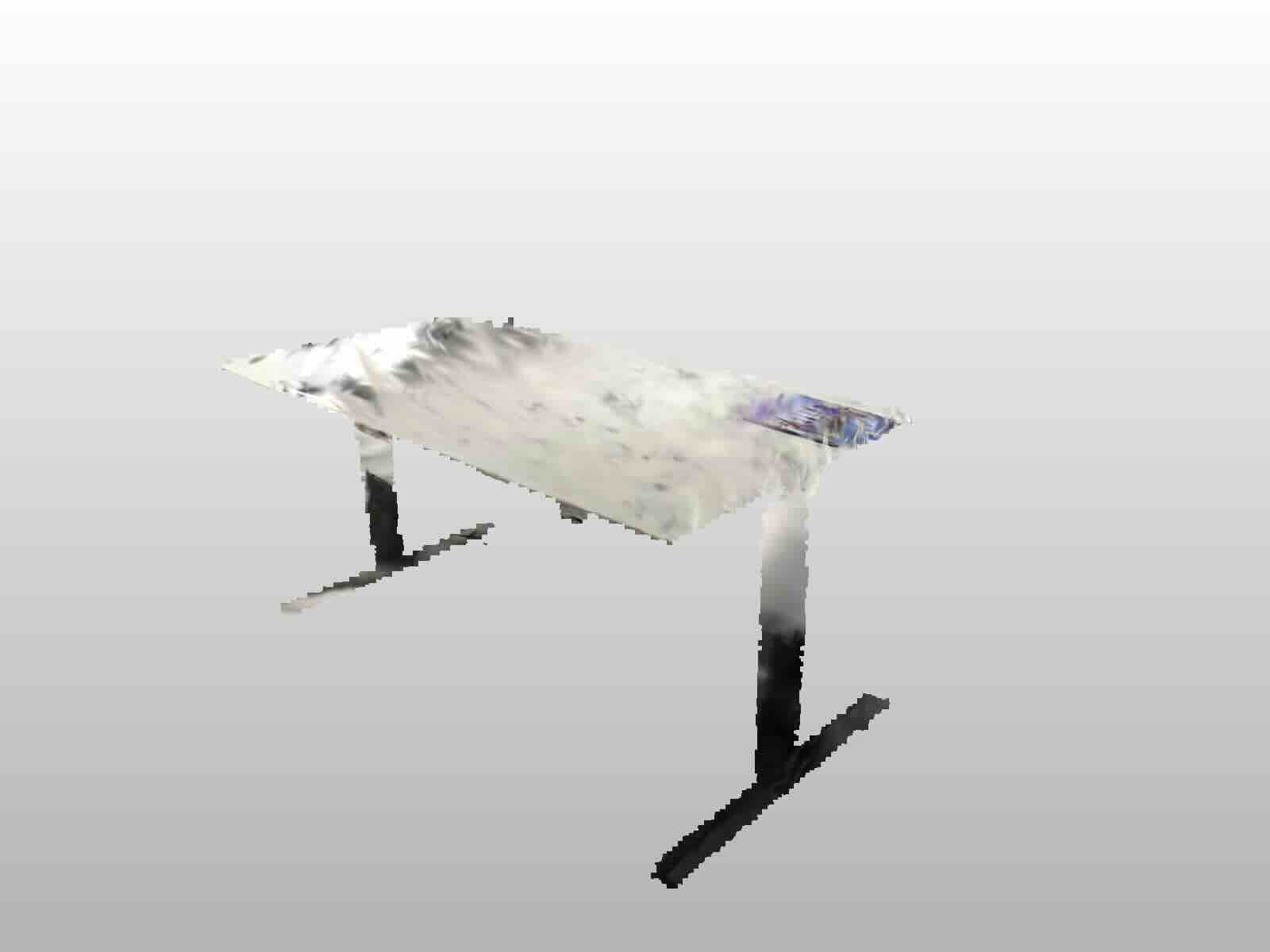} &
        \includegraphics[trim={5.0cm 1.0cm 7.0cm 7.5cm},clip,width=0.12\textwidth]{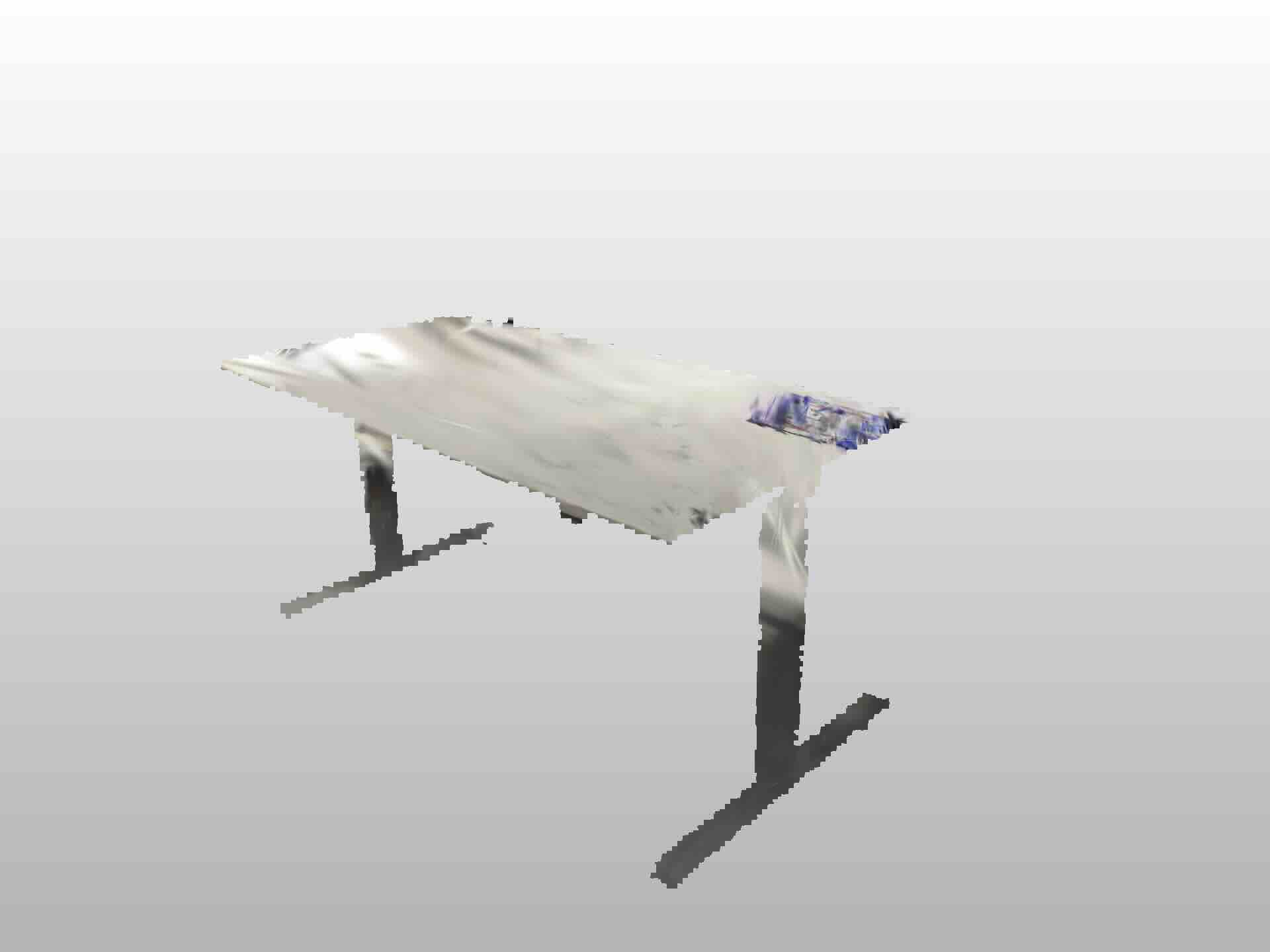} &
        \includegraphics[trim={5.0cm 1.0cm 7.0cm 7.5cm},clip,width=0.12\textwidth]{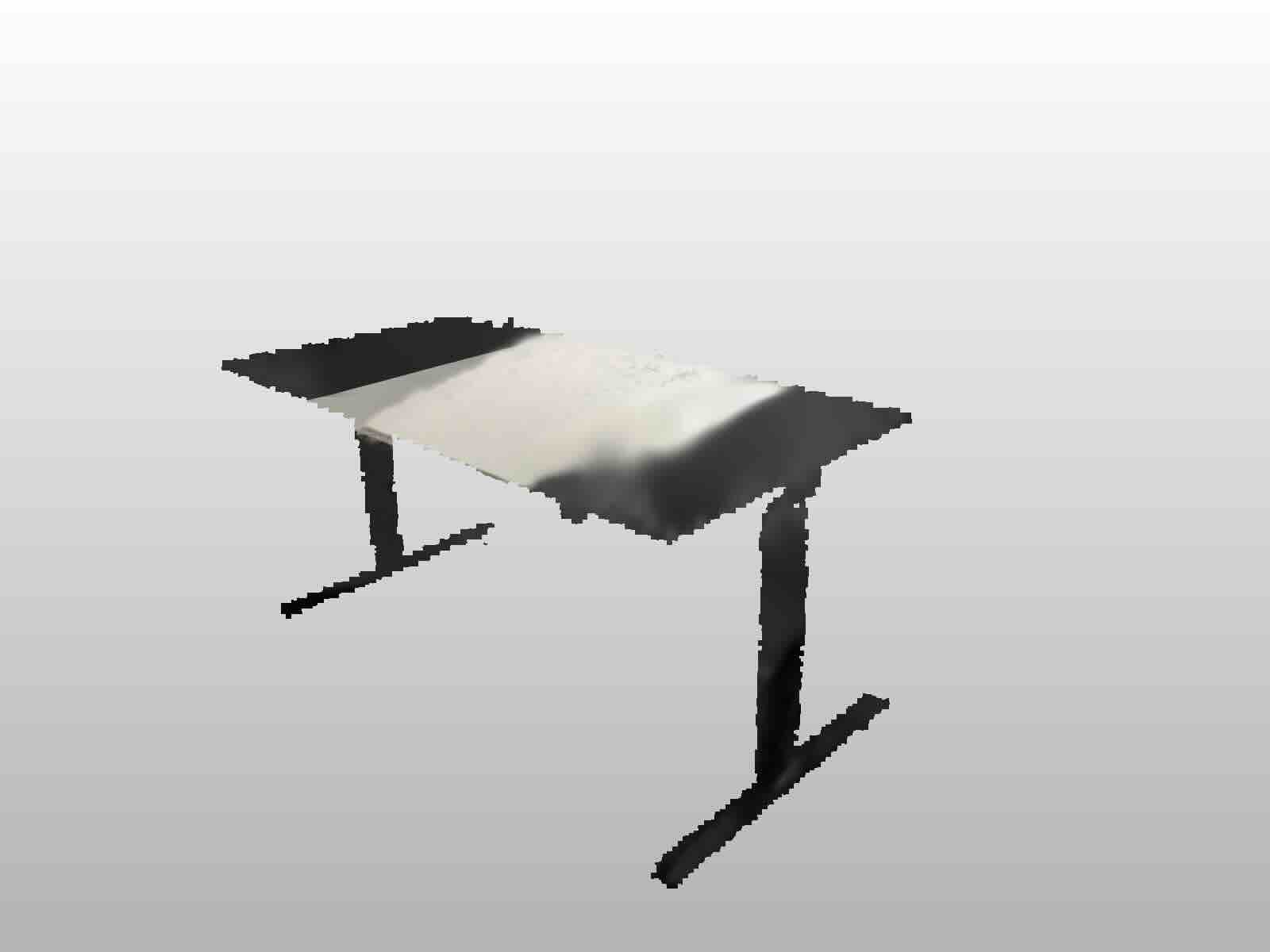}&
        \includegraphics[trim={5.0cm 1.0cm 7.0cm 7.5cm},clip,width=0.12\textwidth]{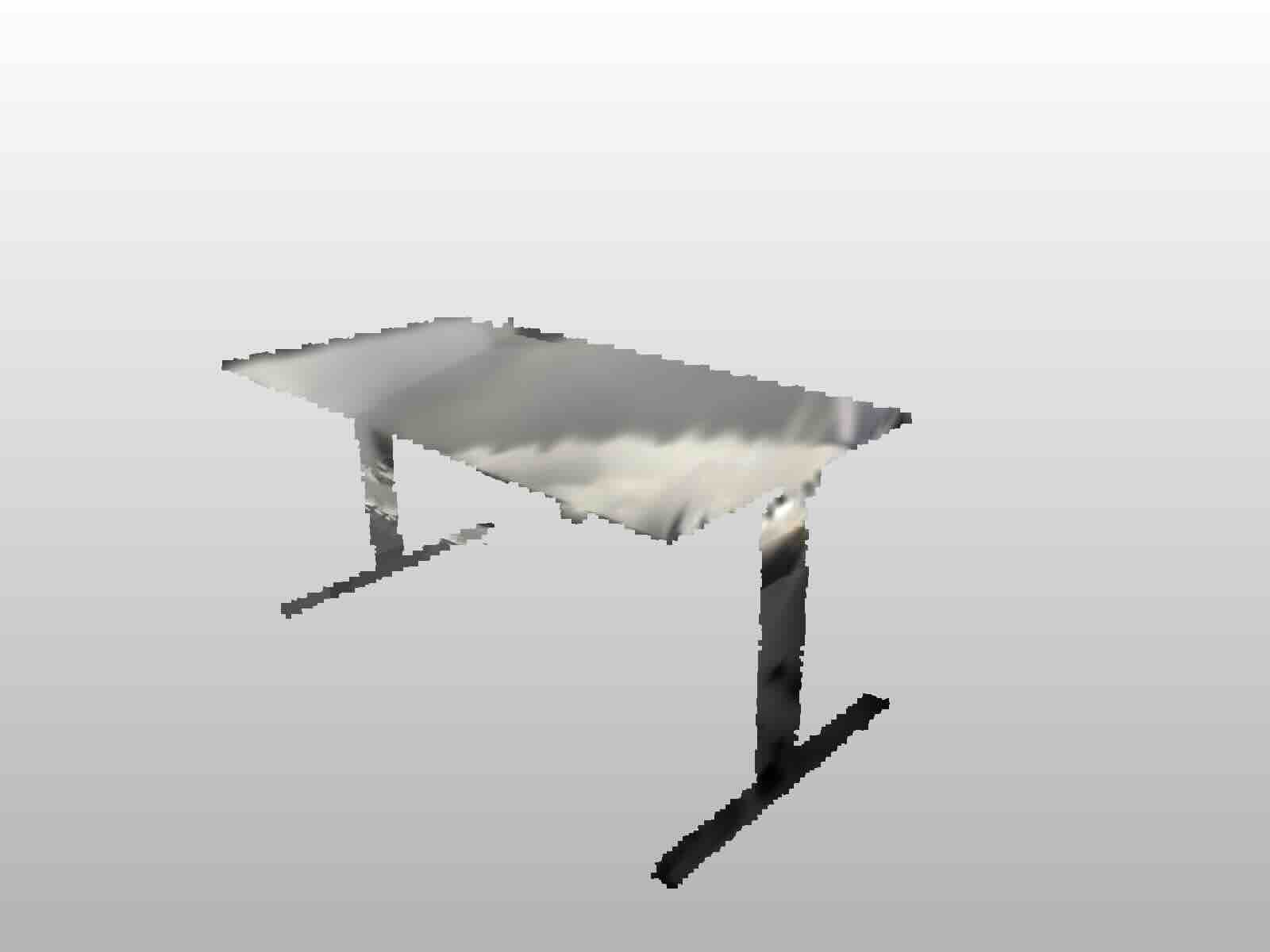} &
        \includegraphics[trim={5.0cm 1.0cm 7.0cm 7.5cm},clip,width=0.12\textwidth]{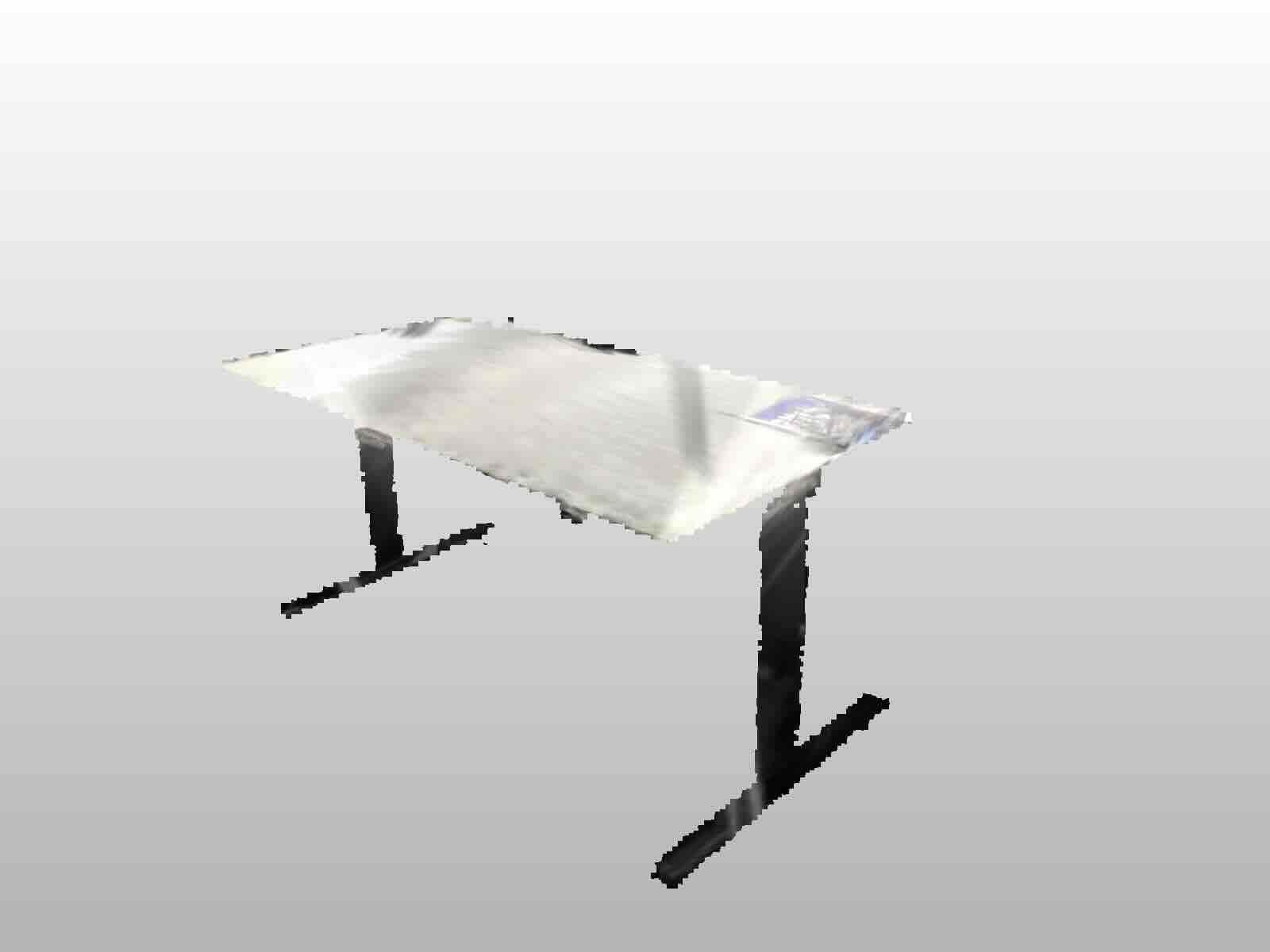} 
        &
        \includegraphics[trim={5.0cm 1.0cm 7.0cm 7.5cm},clip,width=0.12\textwidth]{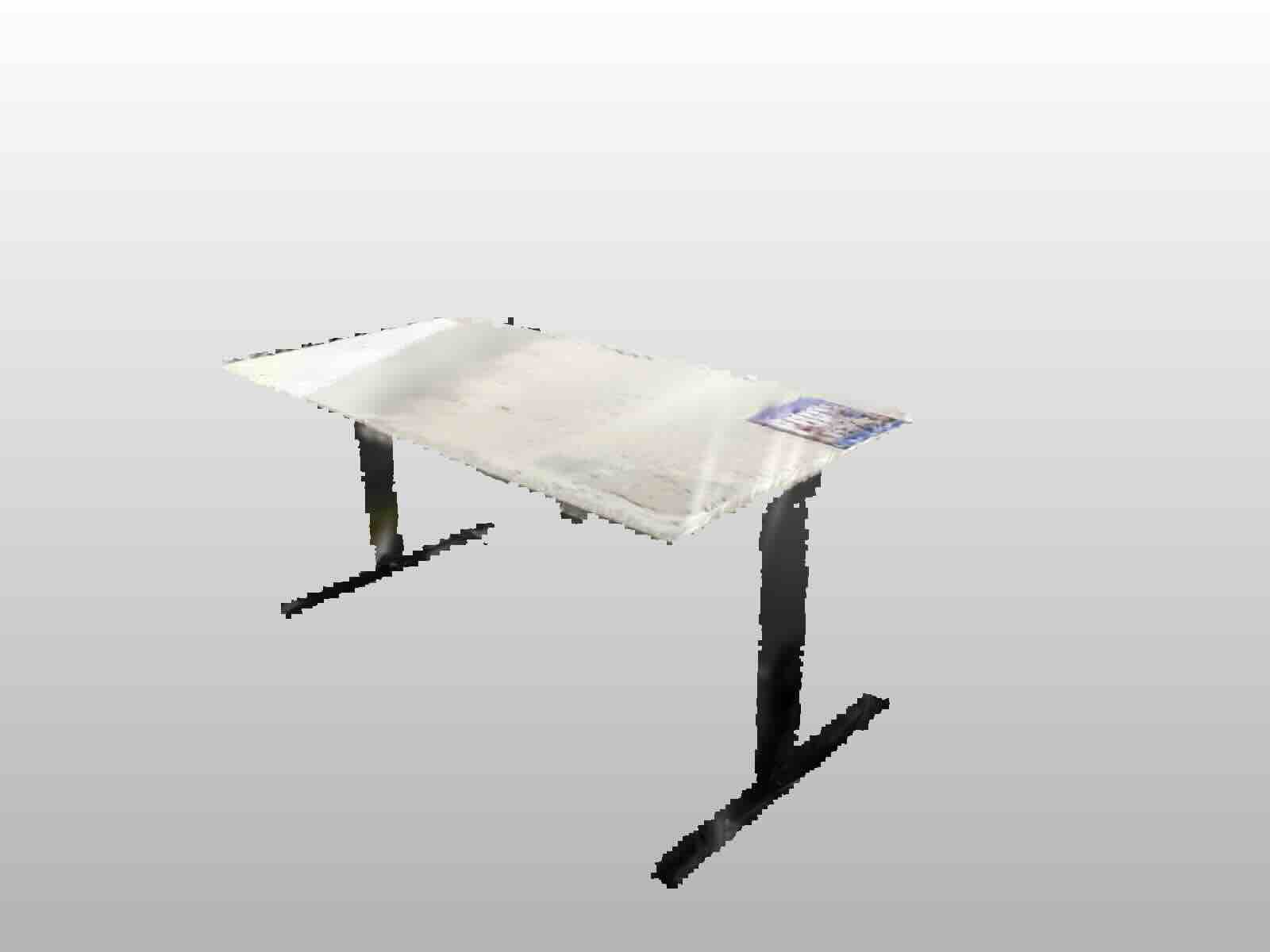} &
        \includegraphics[trim={5.0cm 1.0cm 7.0cm 7.5cm},clip,width=0.12\textwidth]{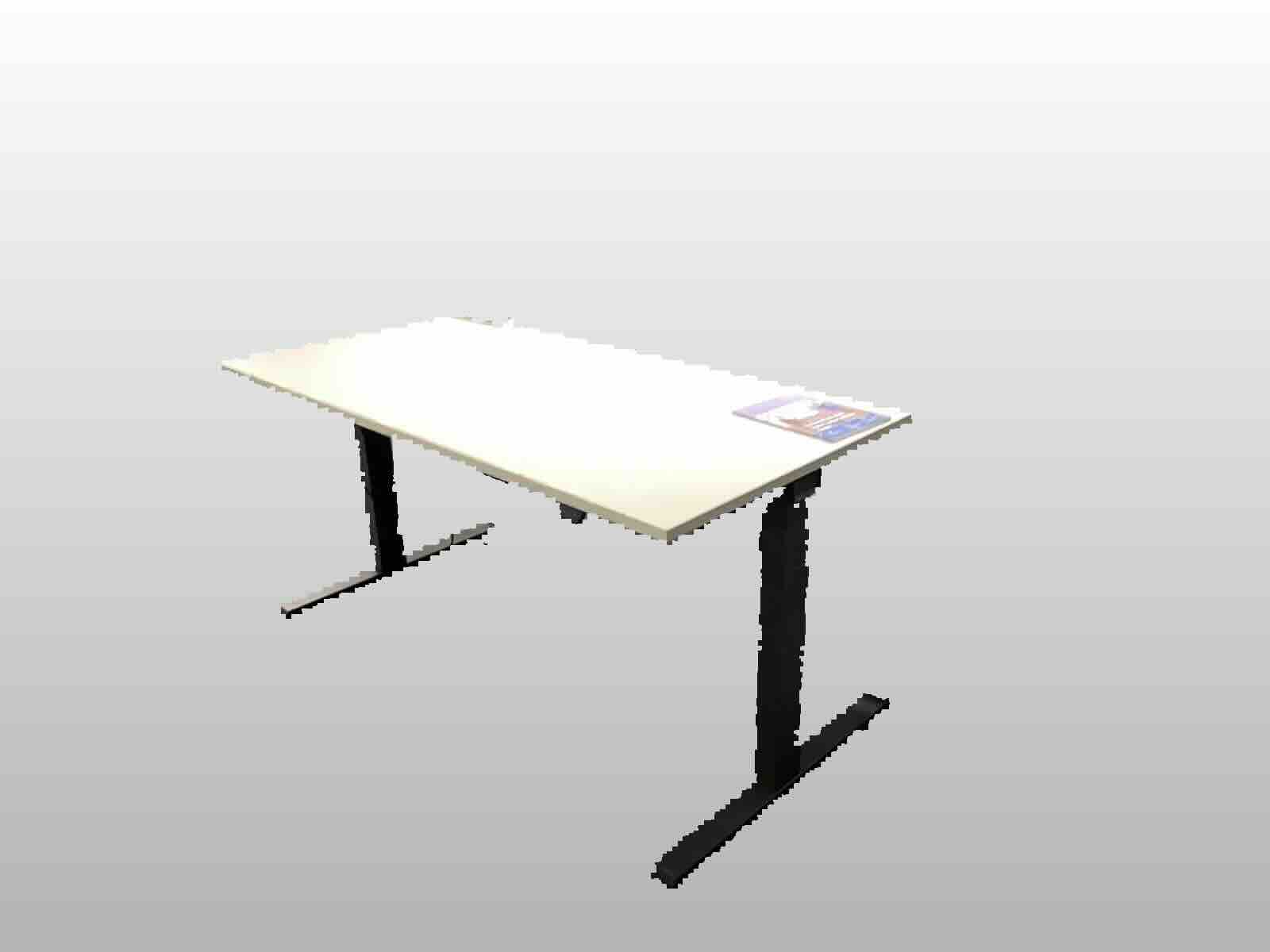}
        \\
        \includegraphics[trim={5.0cm 18cm 30.0cm 20cm},clip,width=0.12\textwidth]{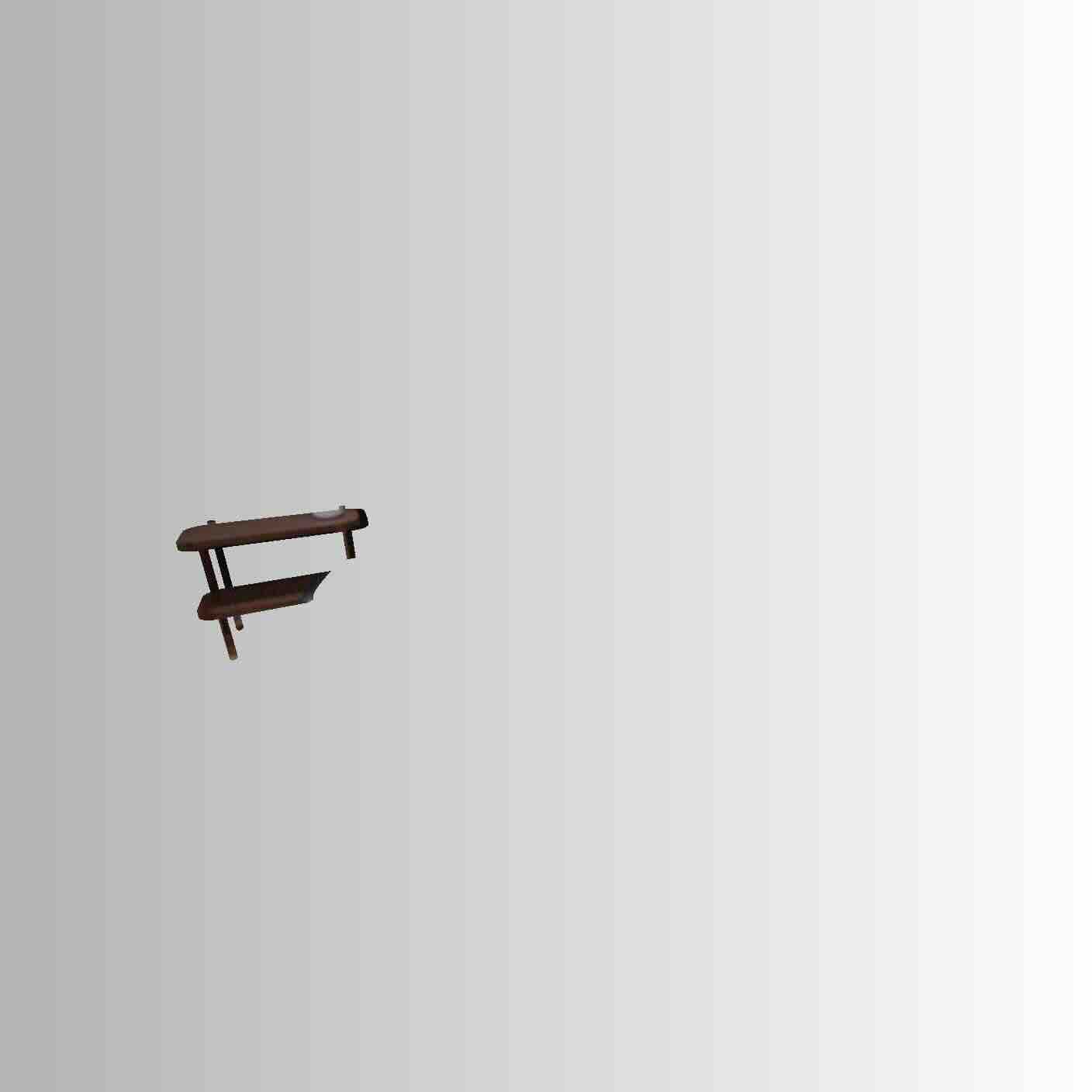} 
        &
        \includegraphics[trim={25cm 10cm 0cm 20cm},clip,width=0.12\textwidth]{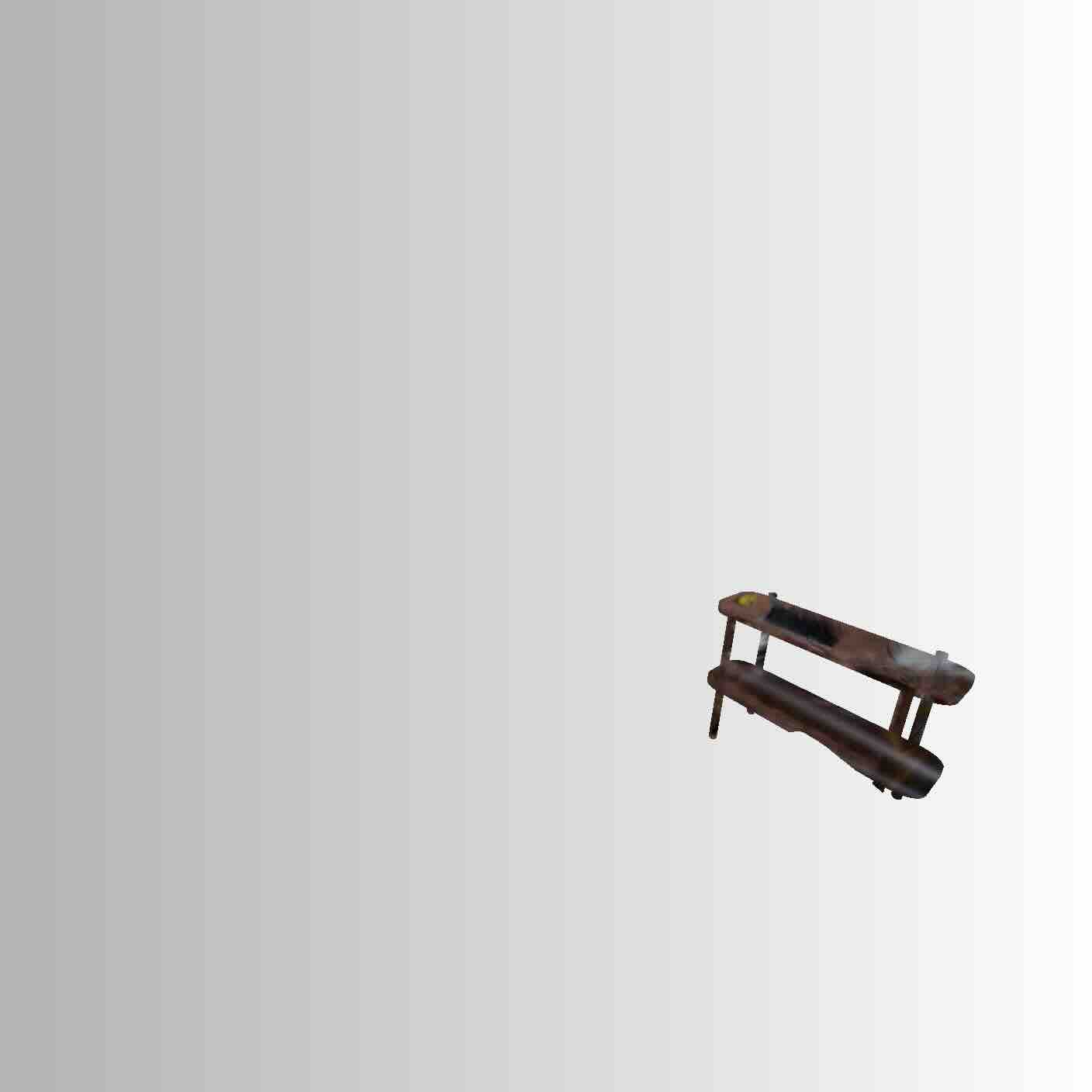} &
        \includegraphics[trim={25cm 10cm 0cm 20cm},clip,width=0.12\textwidth]{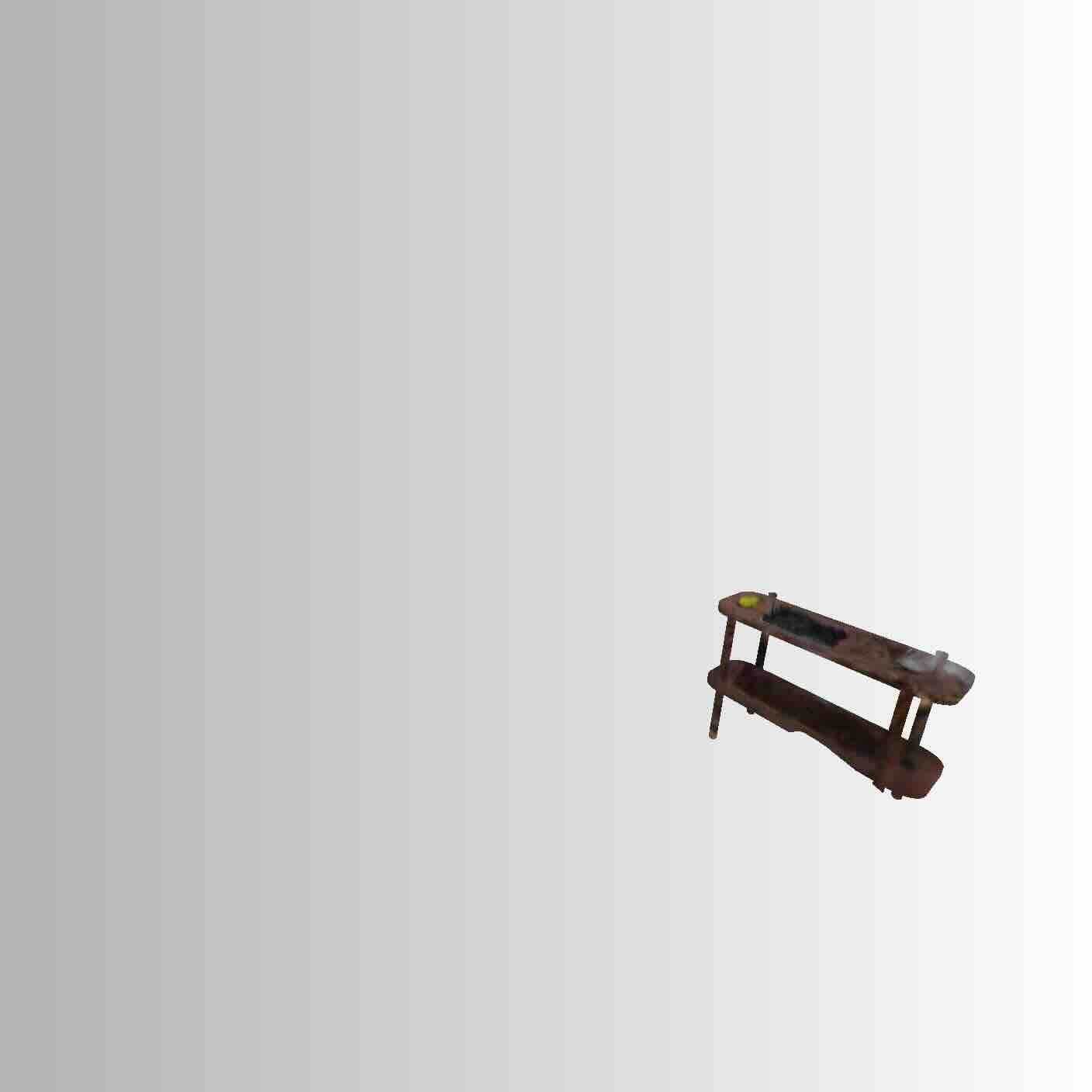} &
        \includegraphics[trim={25cm 10cm 0cm 20cm},clip,width=0.12\textwidth]{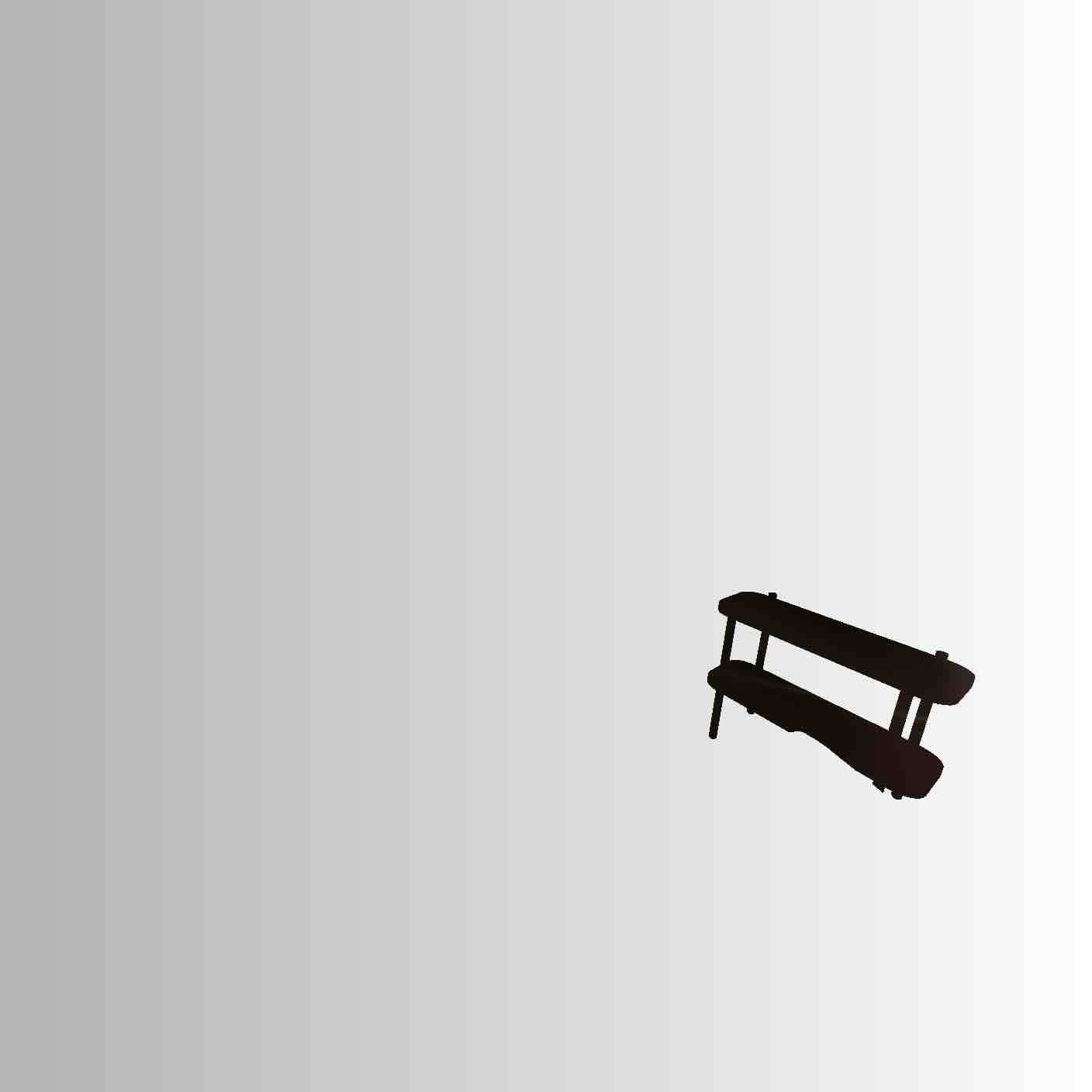} &
        \includegraphics[trim={25cm 10cm 0cm 20cm},clip,width=0.12\textwidth]{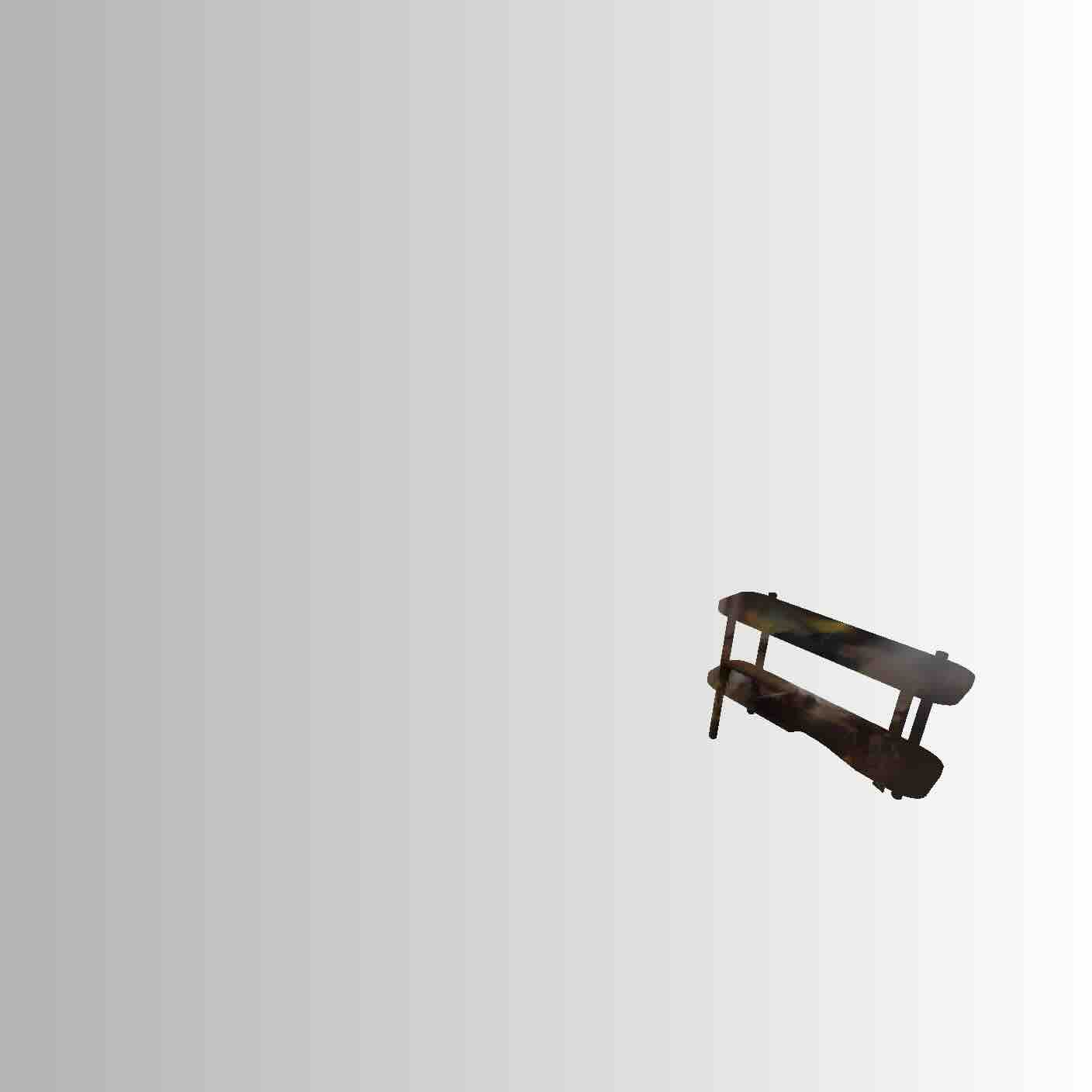} &
        \includegraphics[trim={25cm 10cm 0cm 20cm},clip,width=0.12\textwidth]{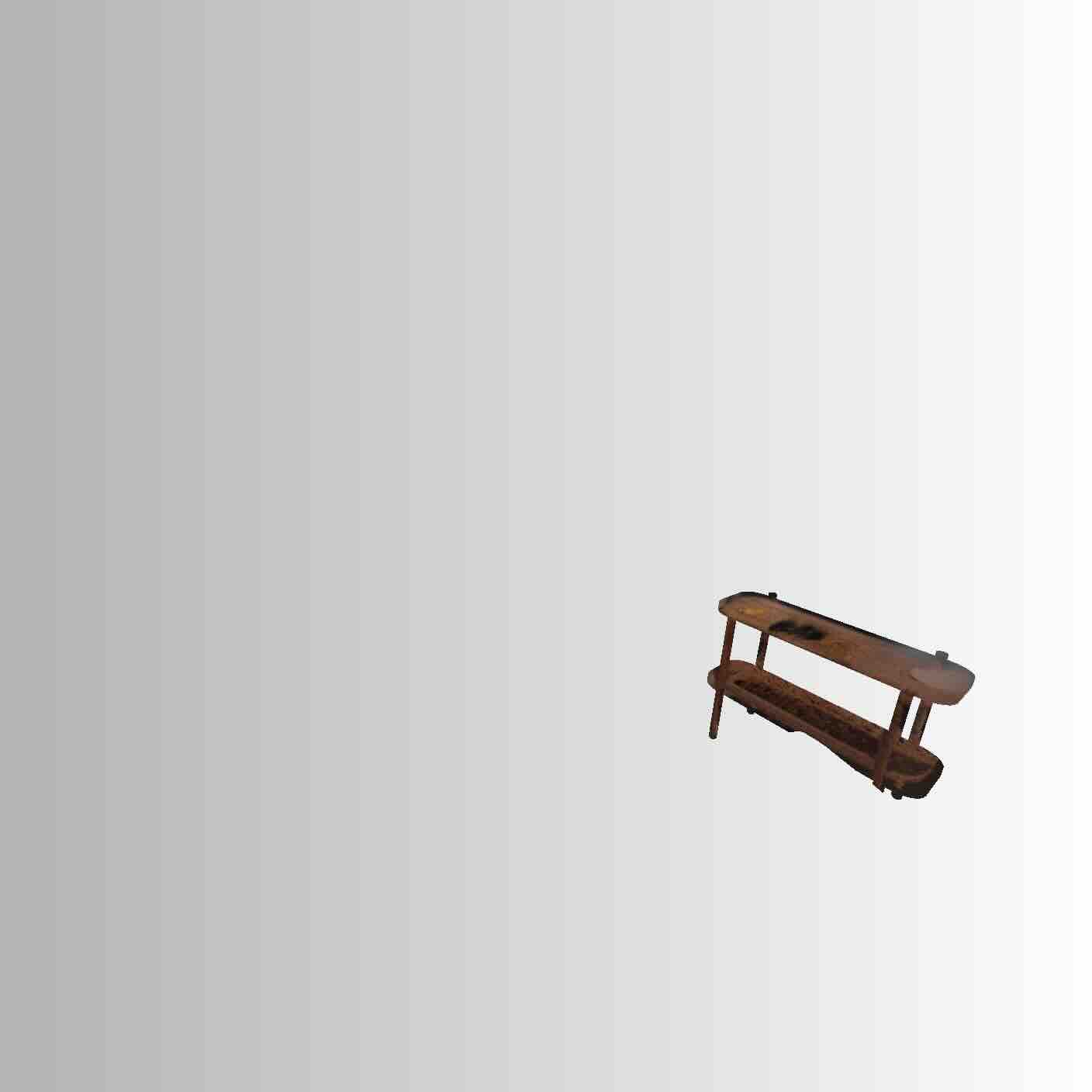} &
        \includegraphics[trim={25cm 10cm 0cm 20cm},clip,width=0.12\textwidth]{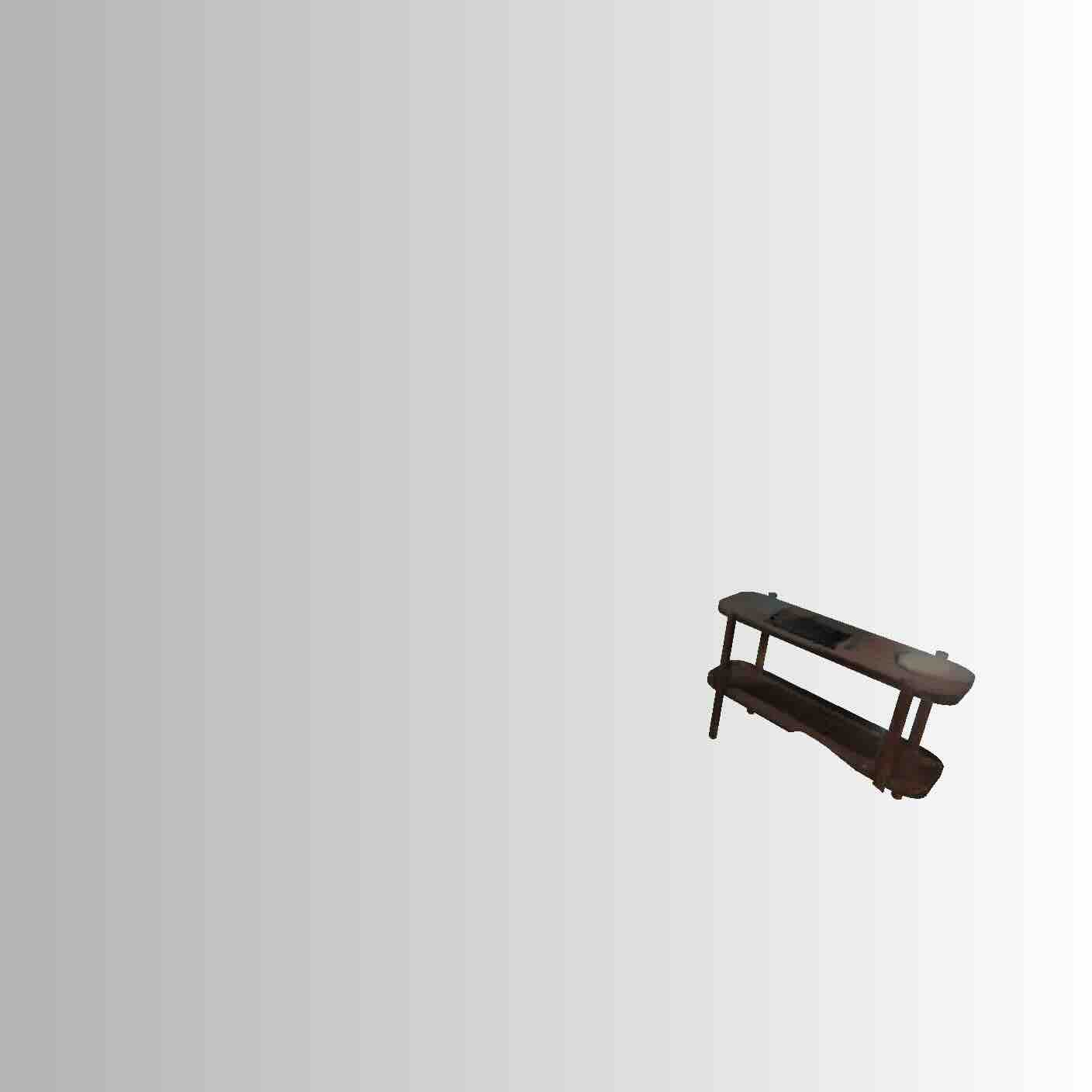} &
        \includegraphics[trim={25cm 10cm 0cm 20cm},clip,width=0.12\textwidth]{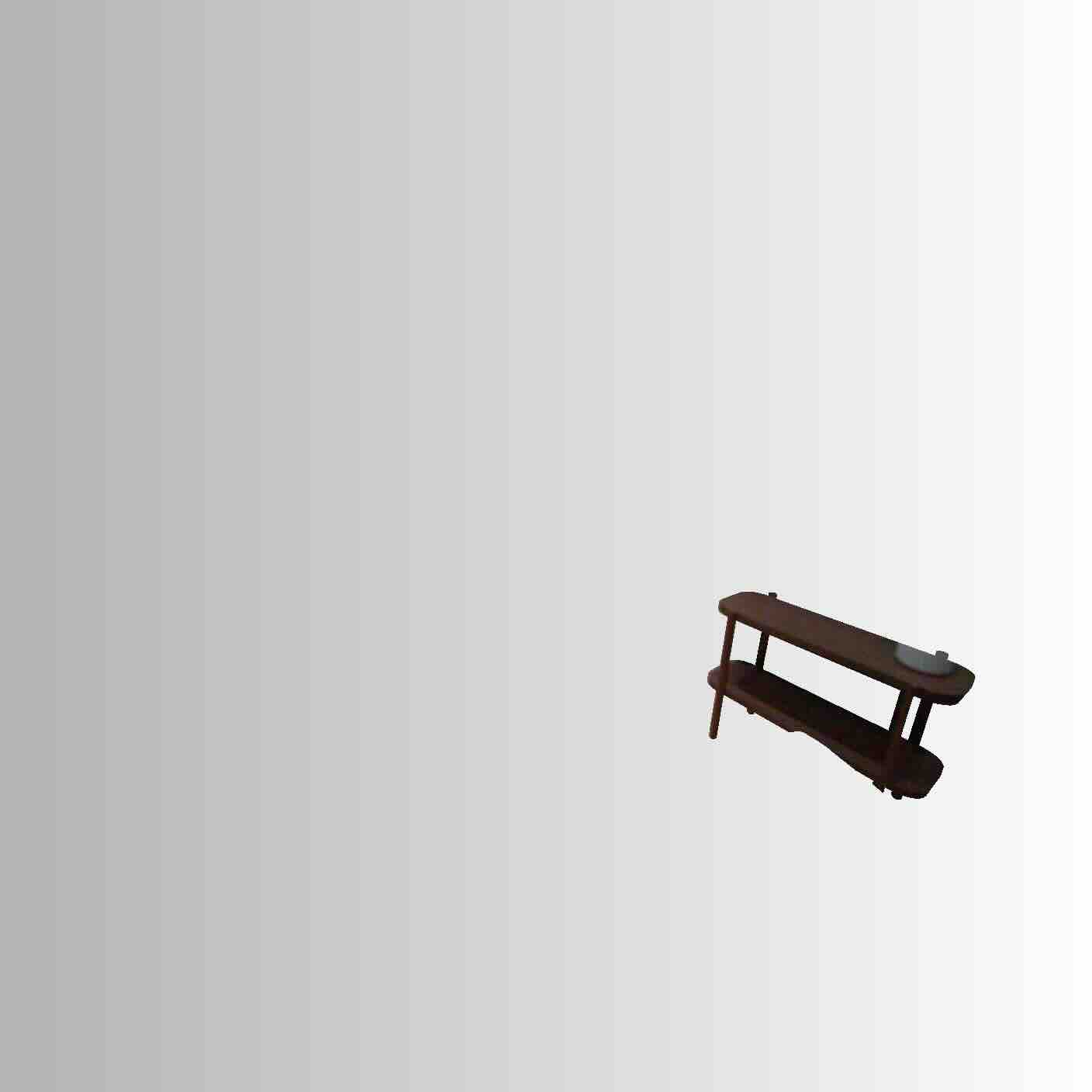}

        \\
         \includegraphics[trim={17.5cm 15cm 5cm 10cm},clip,width=0.12\textwidth]{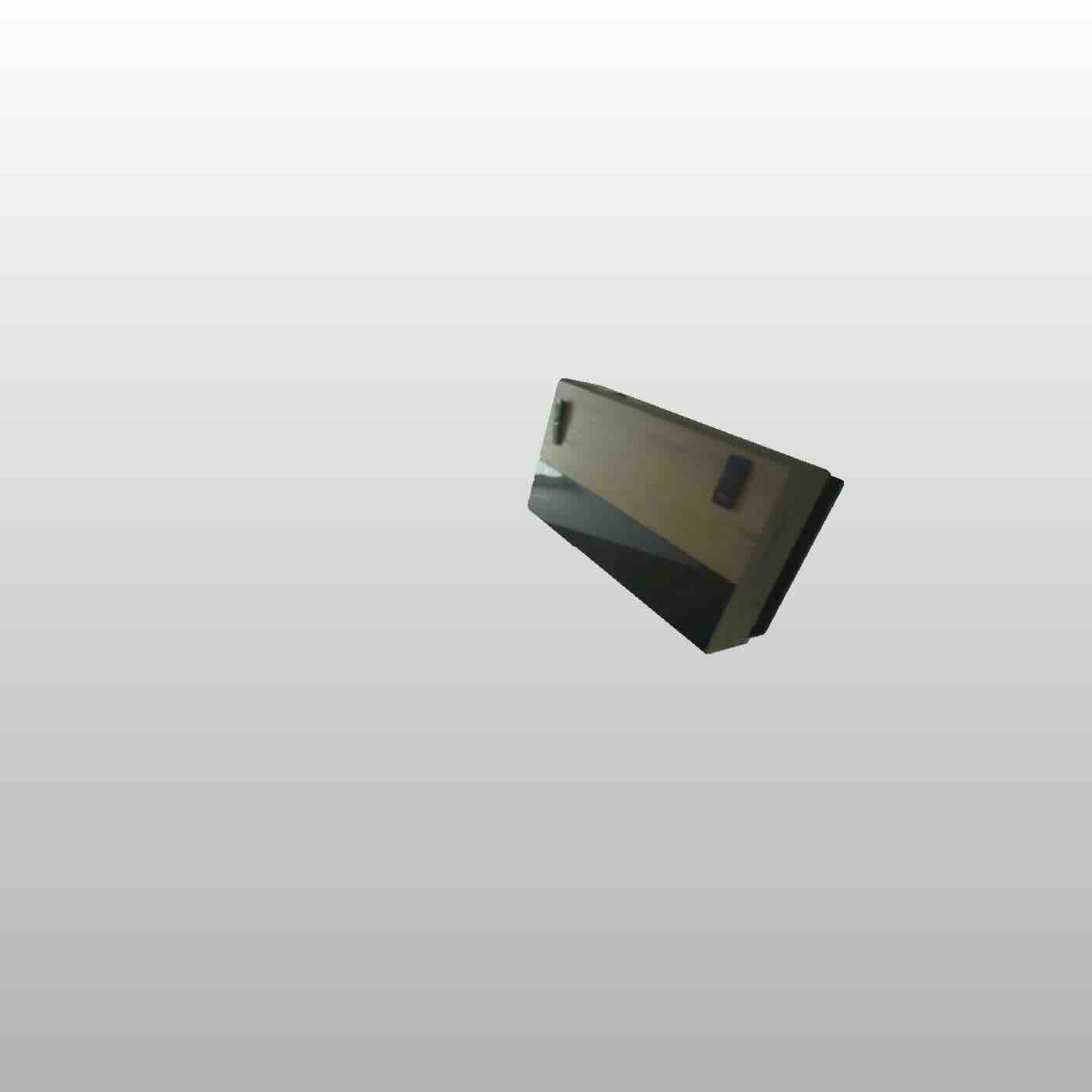} 
        &
        \includegraphics[trim={17.5cm 2.5cm 5cm 22.5cm},clip,width=0.12\textwidth]{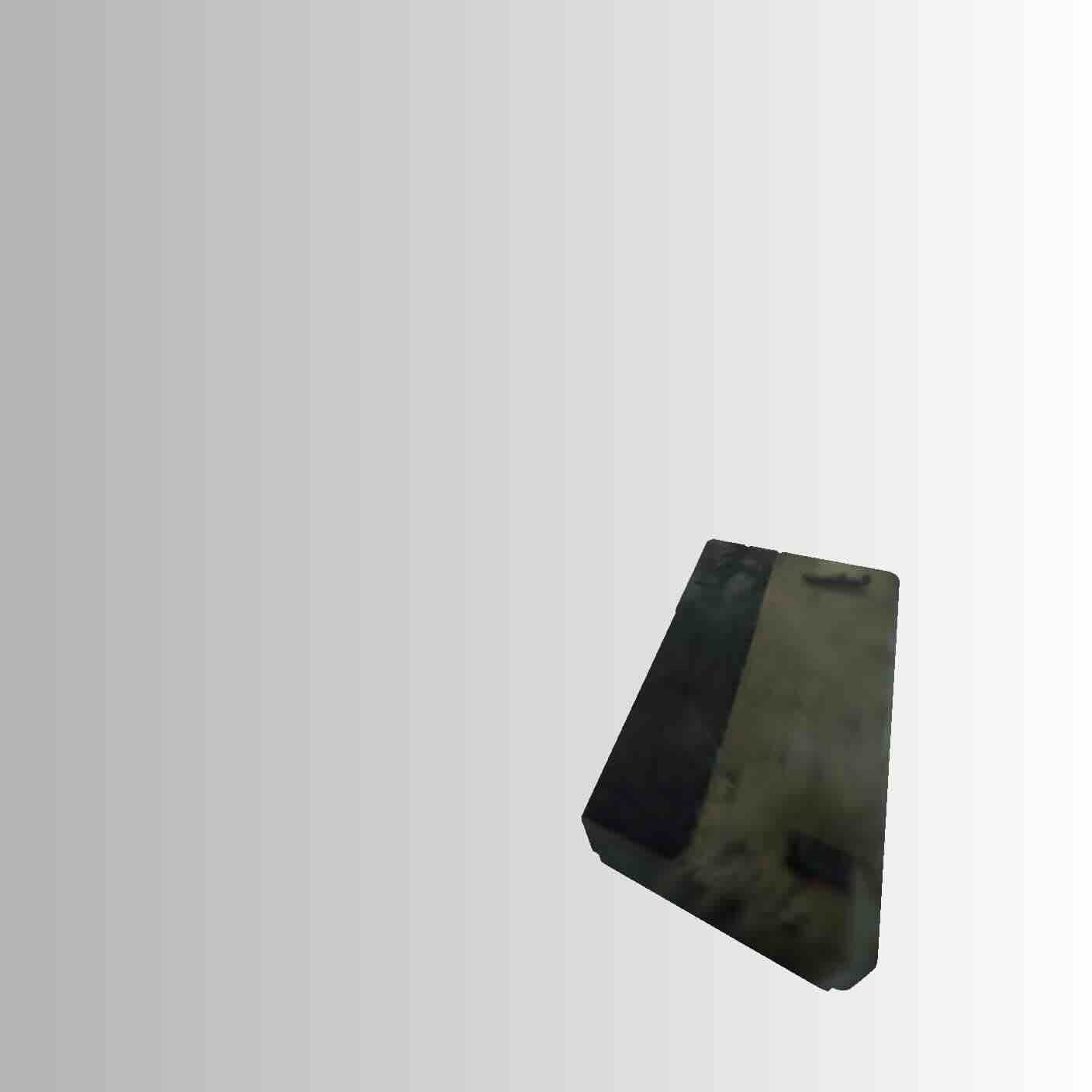} &
        \includegraphics[trim={17.5cm 2.5cm 5cm 22.5cm},clip,width=0.12\textwidth]{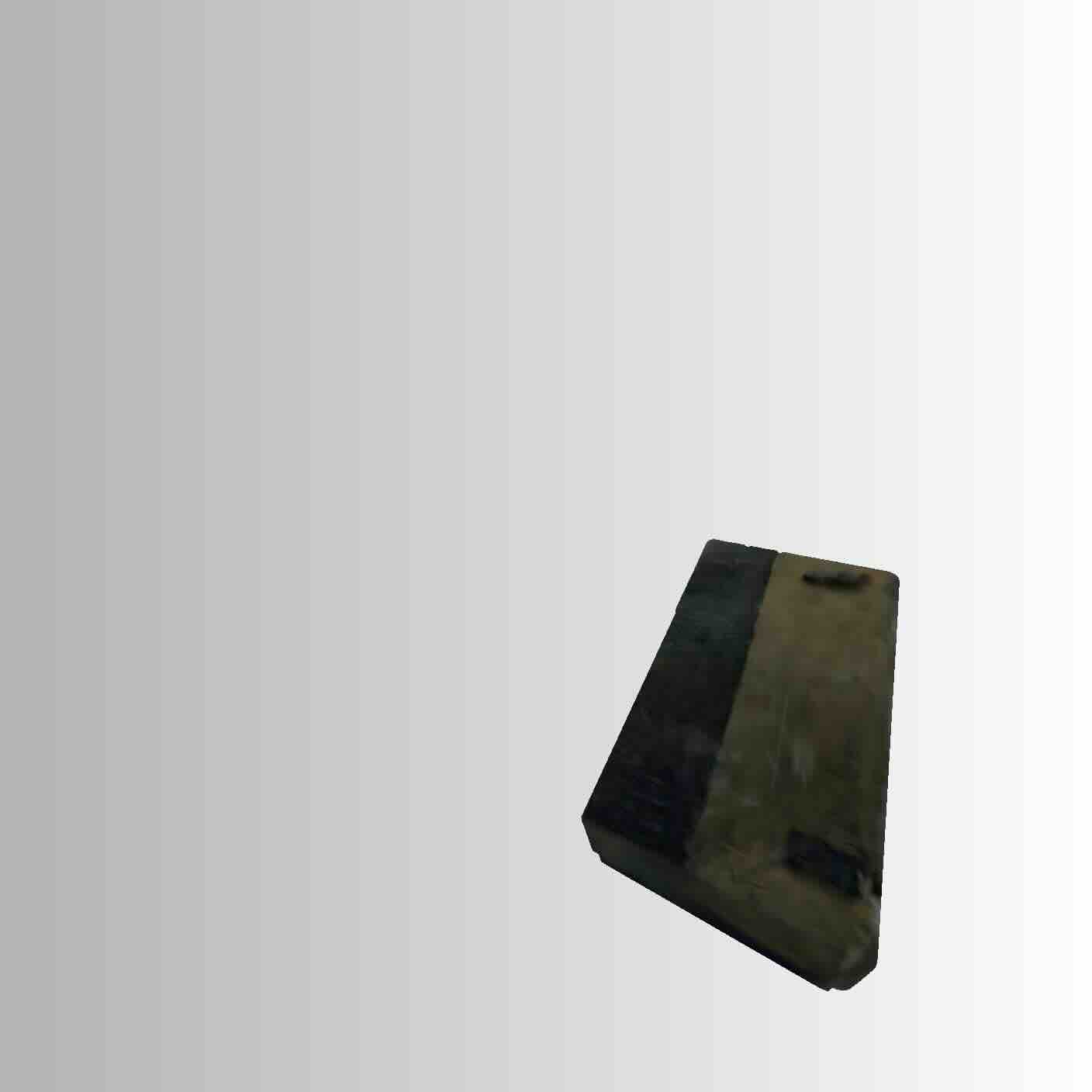} 
        &
        \includegraphics[trim={8.75cm 1.25cm 2.5cm 11.25cm},clip,width=0.12\textwidth]{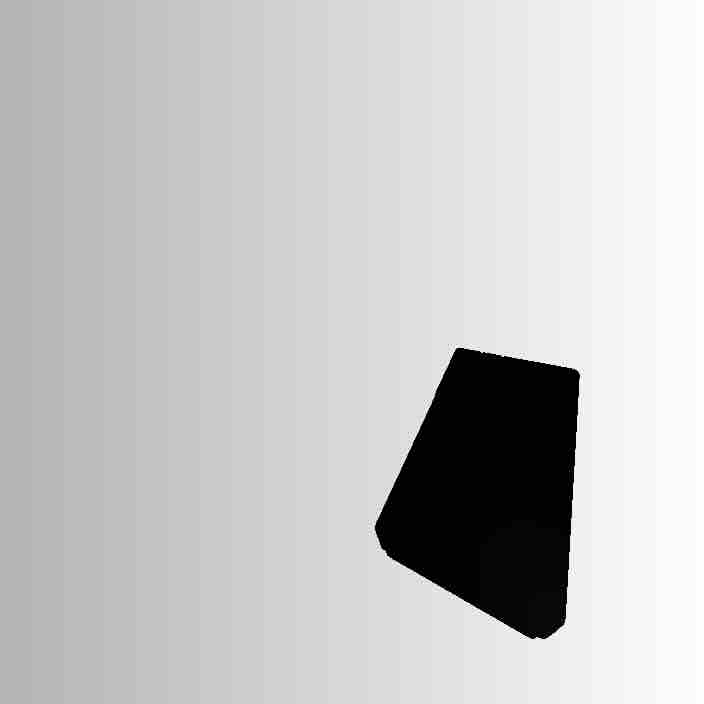}
        &
        \includegraphics[trim={17.5cm 2.5cm 5cm 22.5cm},clip,width=0.12\textwidth]{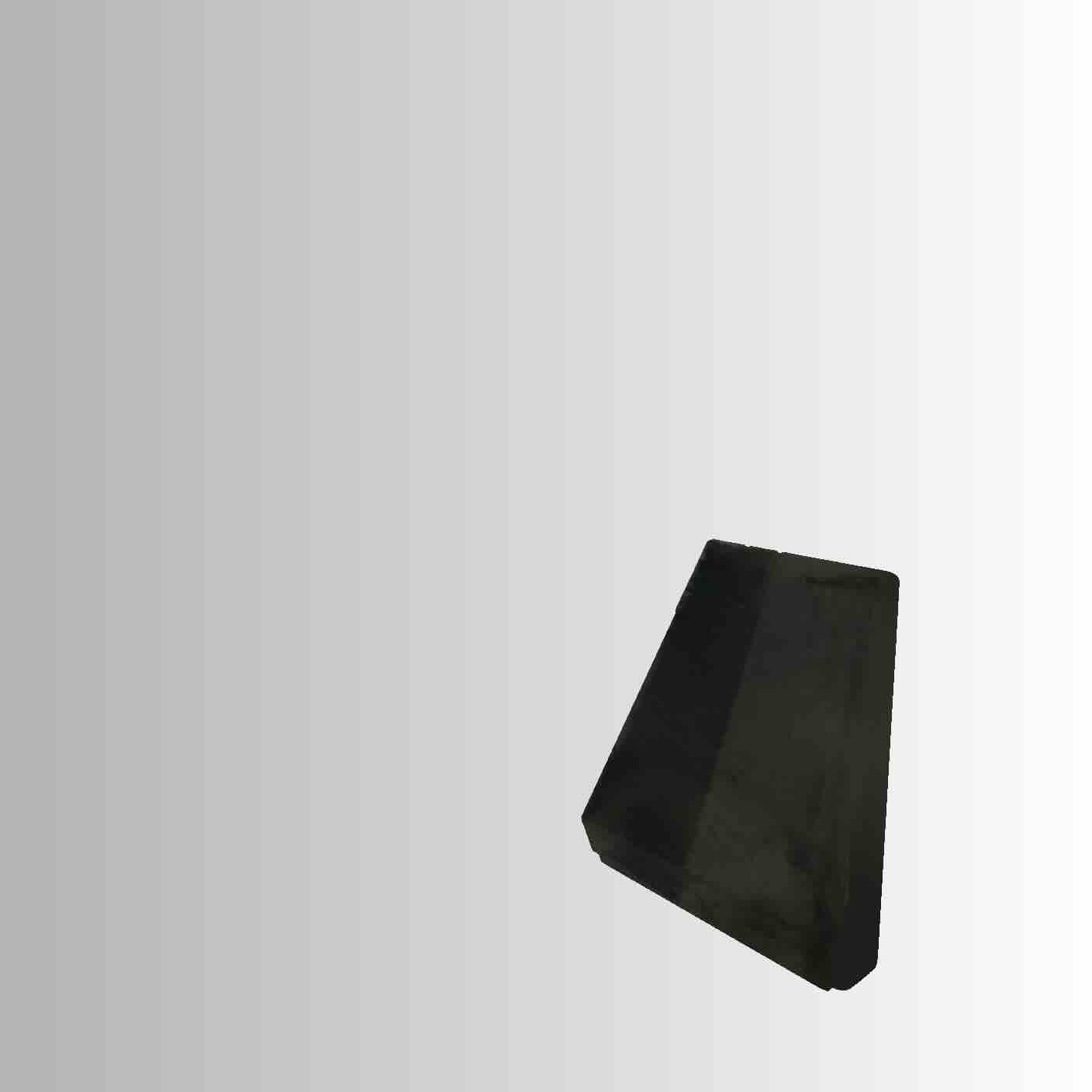} &
        \includegraphics[trim={17.5cm 2.5cm 5cm 22.5cm},clip,width=0.12\textwidth]{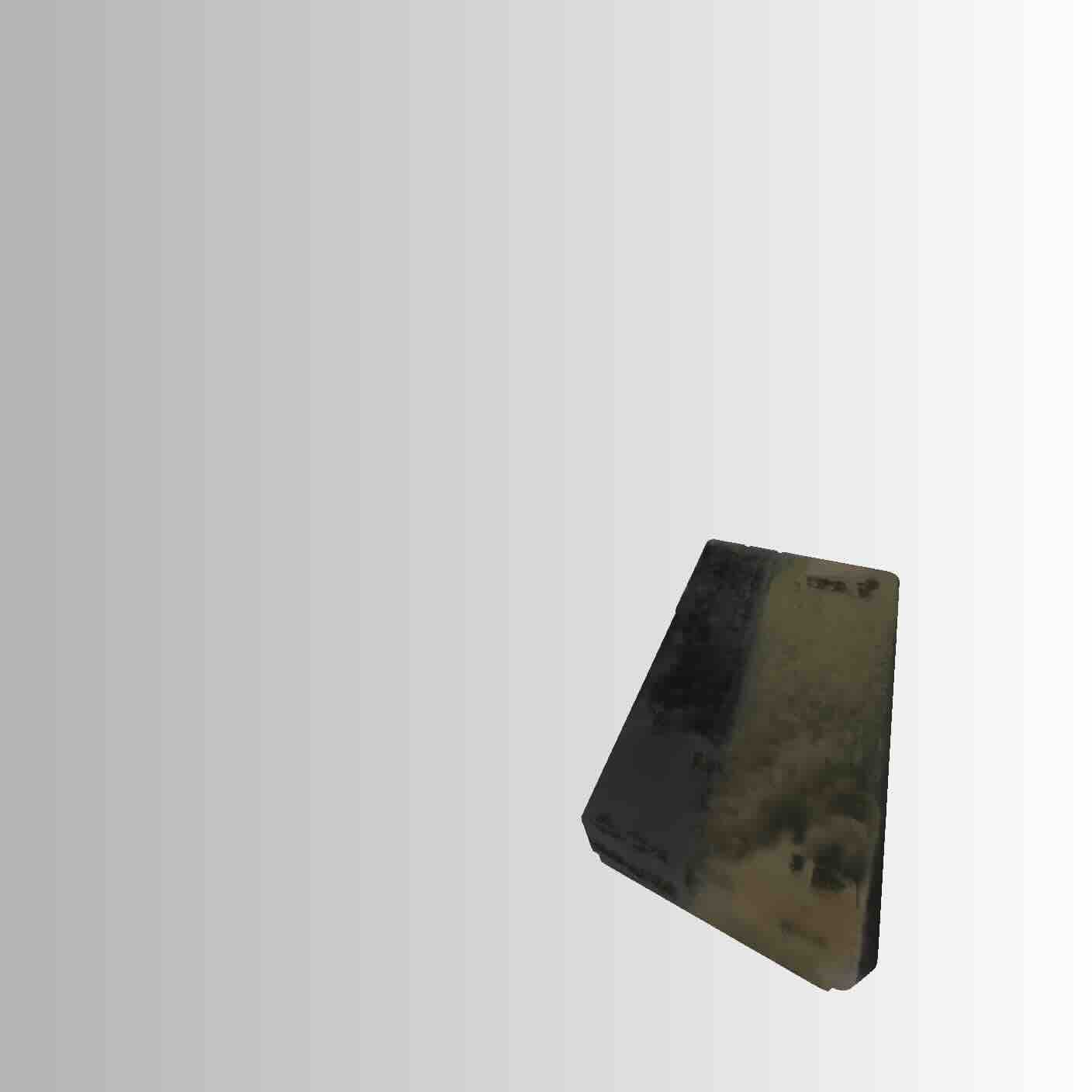} &
        \includegraphics[trim={17.5cm 2.5cm 5cm 22.5cm},clip,width=0.12\textwidth]{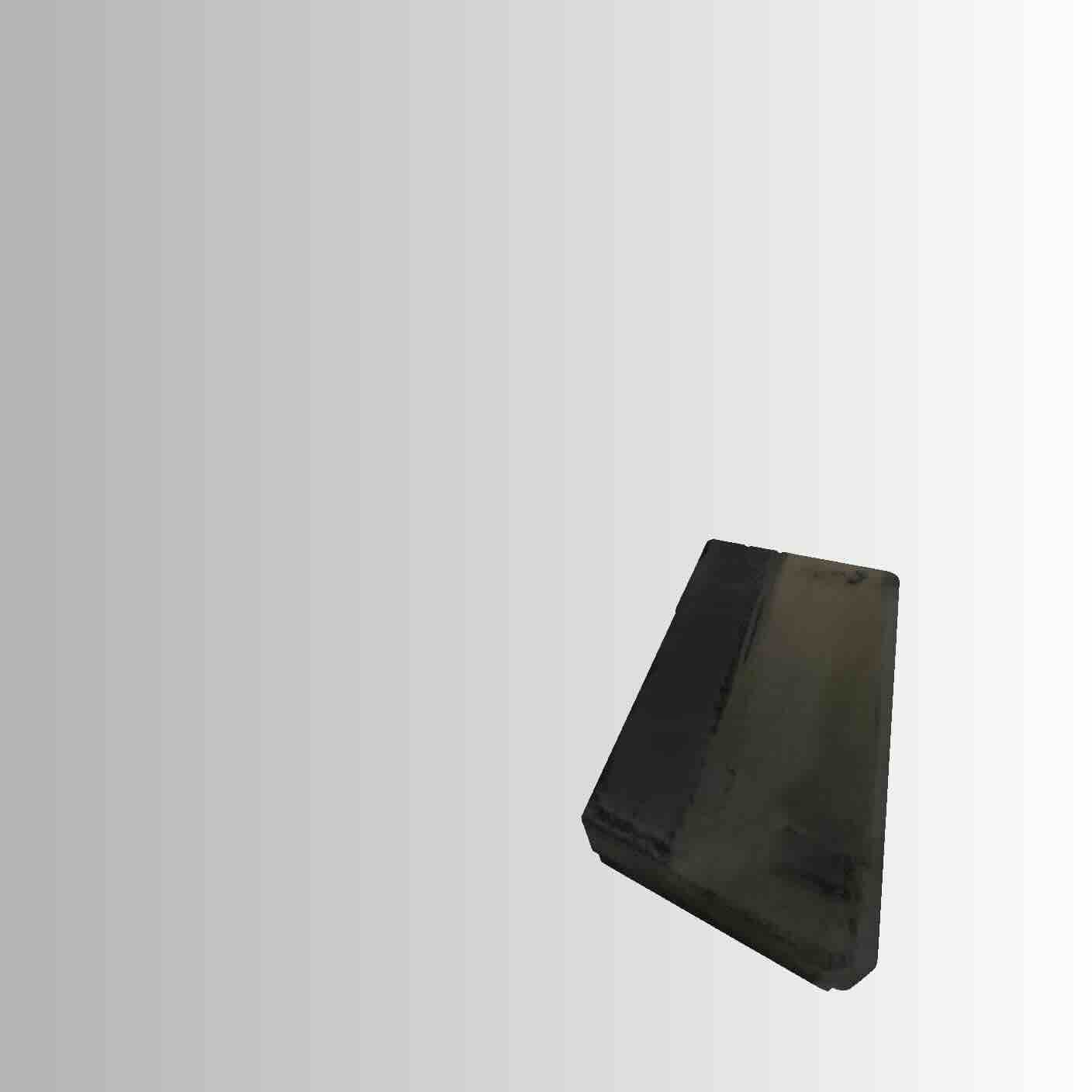} &
        \includegraphics[trim={17.5cm 2.5cm 5cm 22.5cm},clip,width=0.12\textwidth]{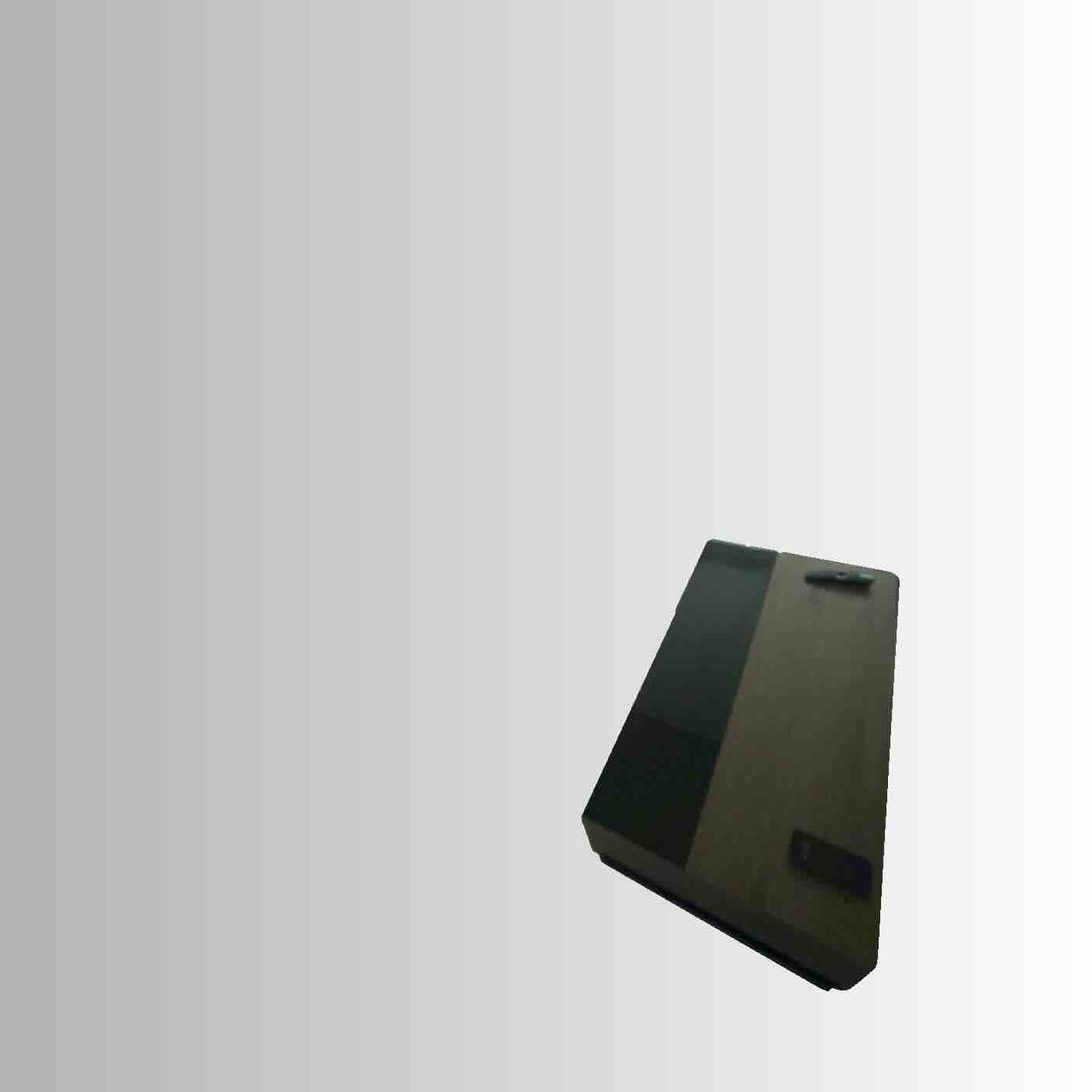}

        \\
        \includegraphics[trim={25cm 10cm 0cm 21cm},clip,width=0.12\textwidth]{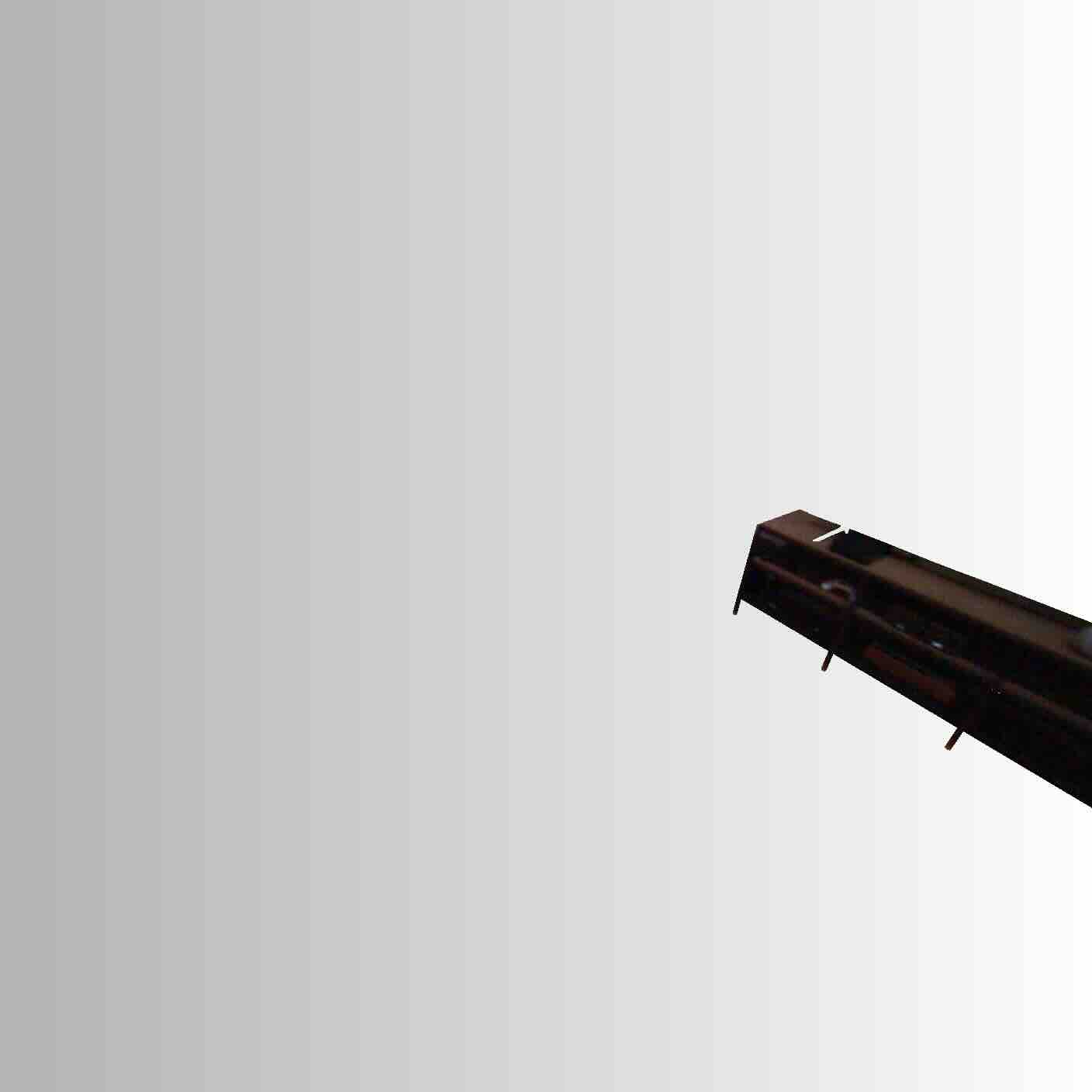} 
        &
        \includegraphics[trim={2.5cm 10cm 17.5cm 17.5cm},clip,width=0.12\textwidth]{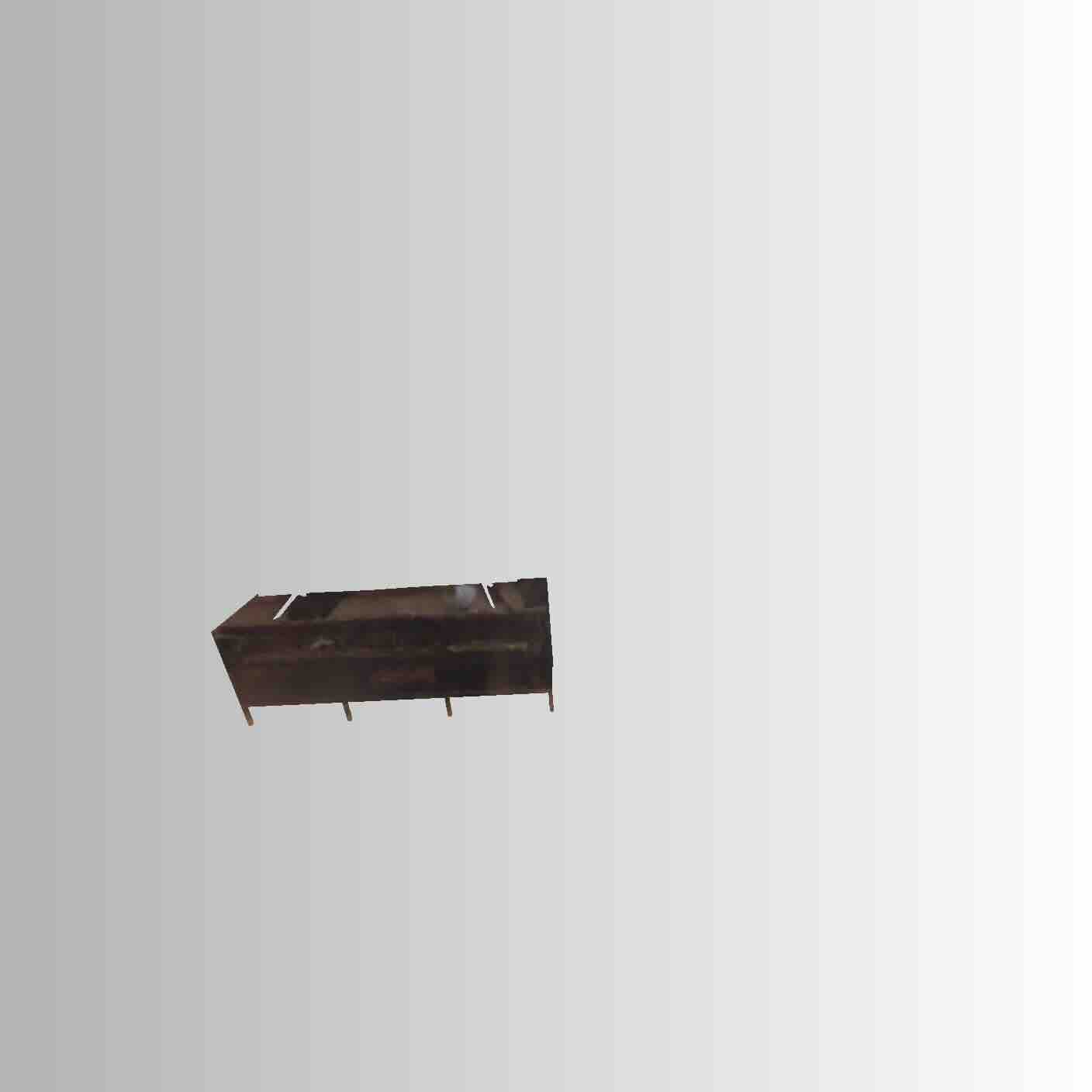} &
        \includegraphics[trim={2.5cm 10cm 17.5cm 17.5cm},clip,width=0.12\textwidth]{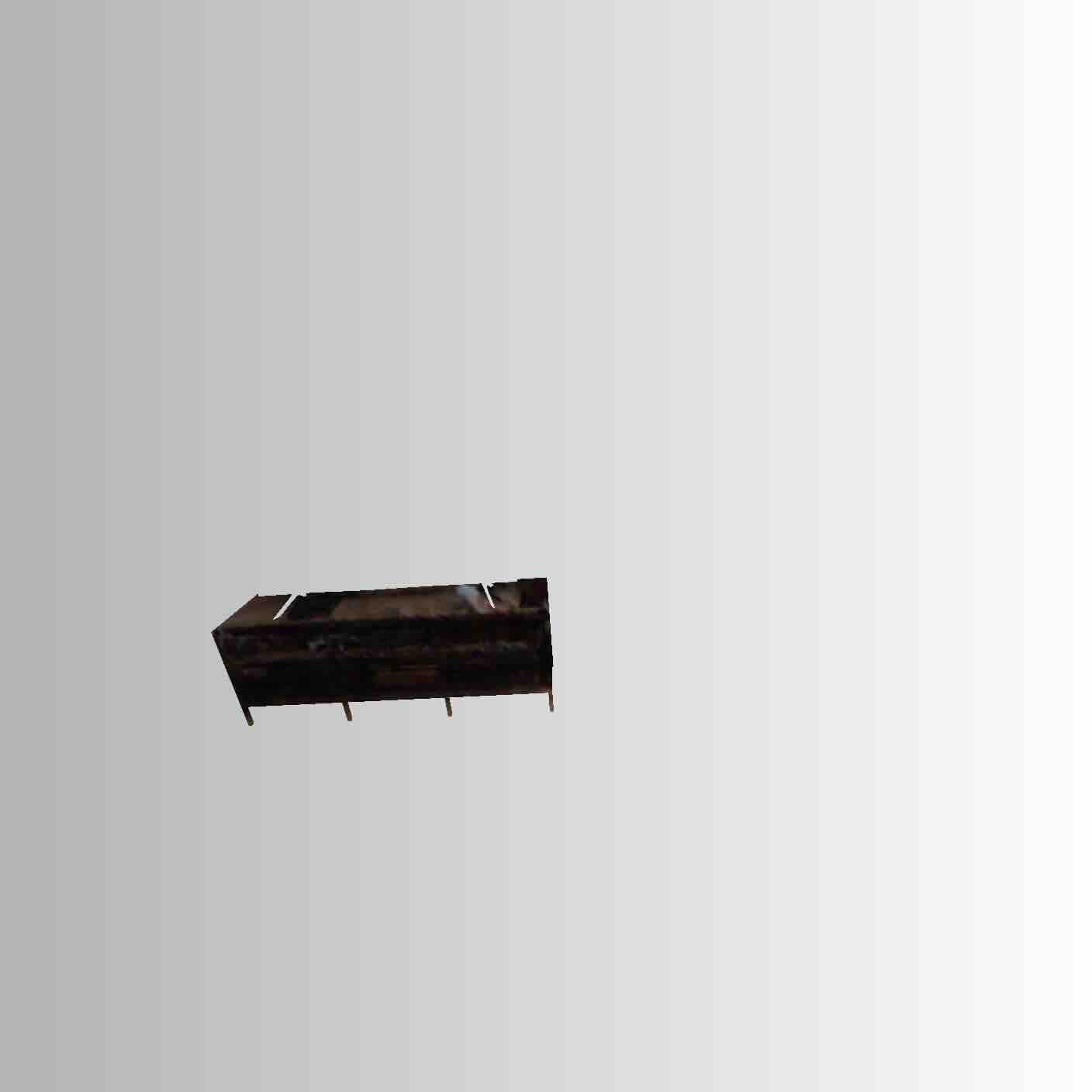} &
        \includegraphics[trim={1.25cm 5cm 8.75cm 8.75cm},clip,width=0.12\textwidth]{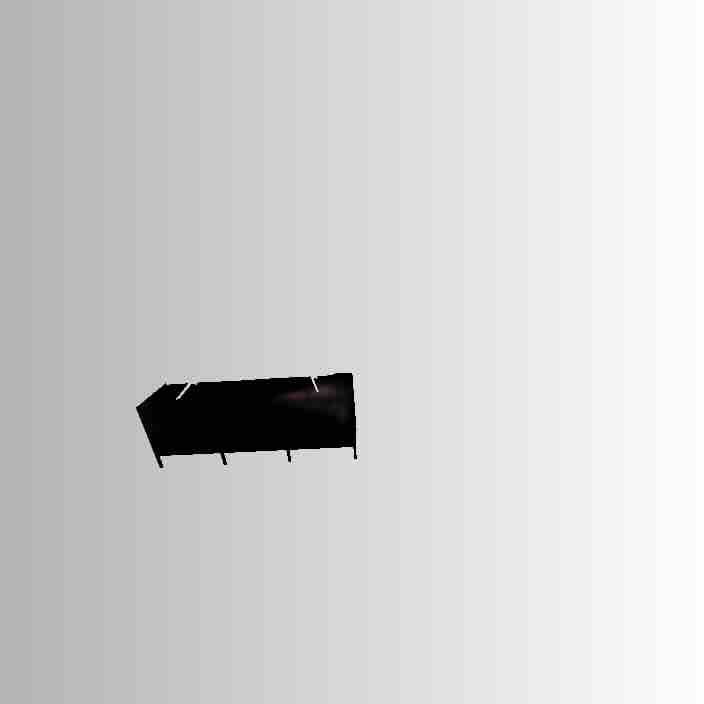} &
        \includegraphics[trim={2.5cm 10cm 17.5cm 17.5cm},clip,width=0.12\textwidth]{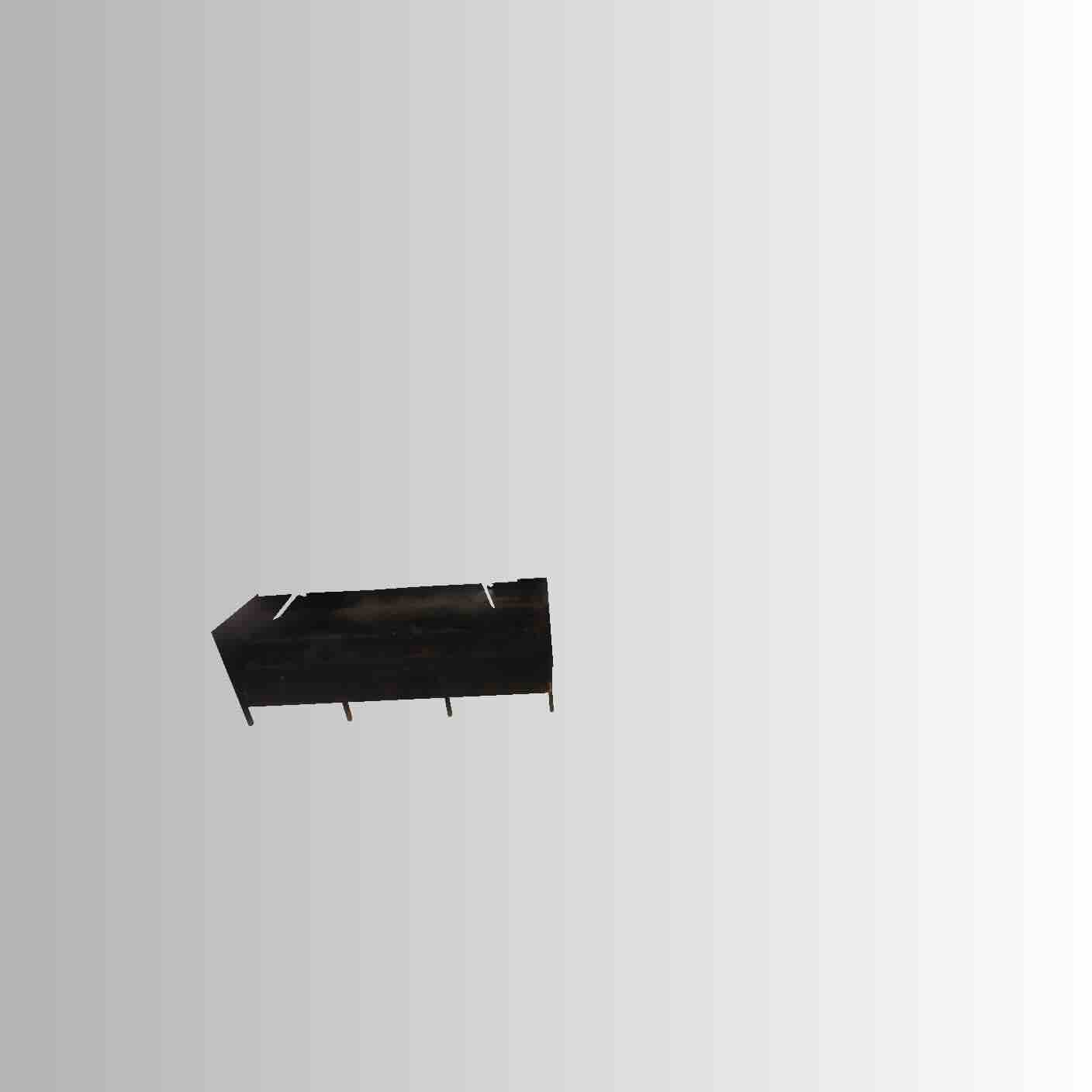} &
        \includegraphics[trim={2.5cm 10cm 17.5cm 17.5cm},clip,width=0.12\textwidth]{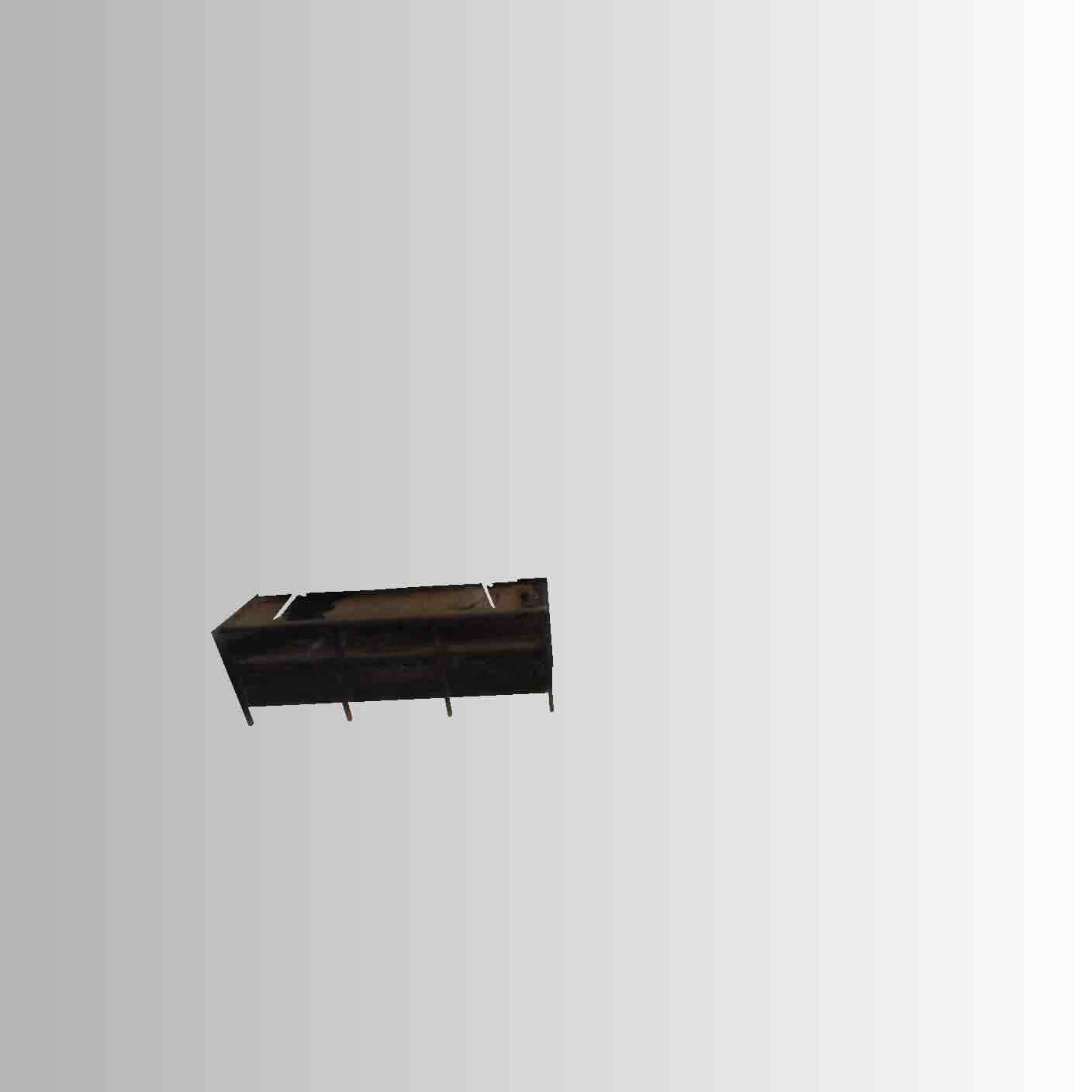} &
        \includegraphics[trim={2.5cm 10cm 17.5cm 17.5cm},clip,width=0.12\textwidth]{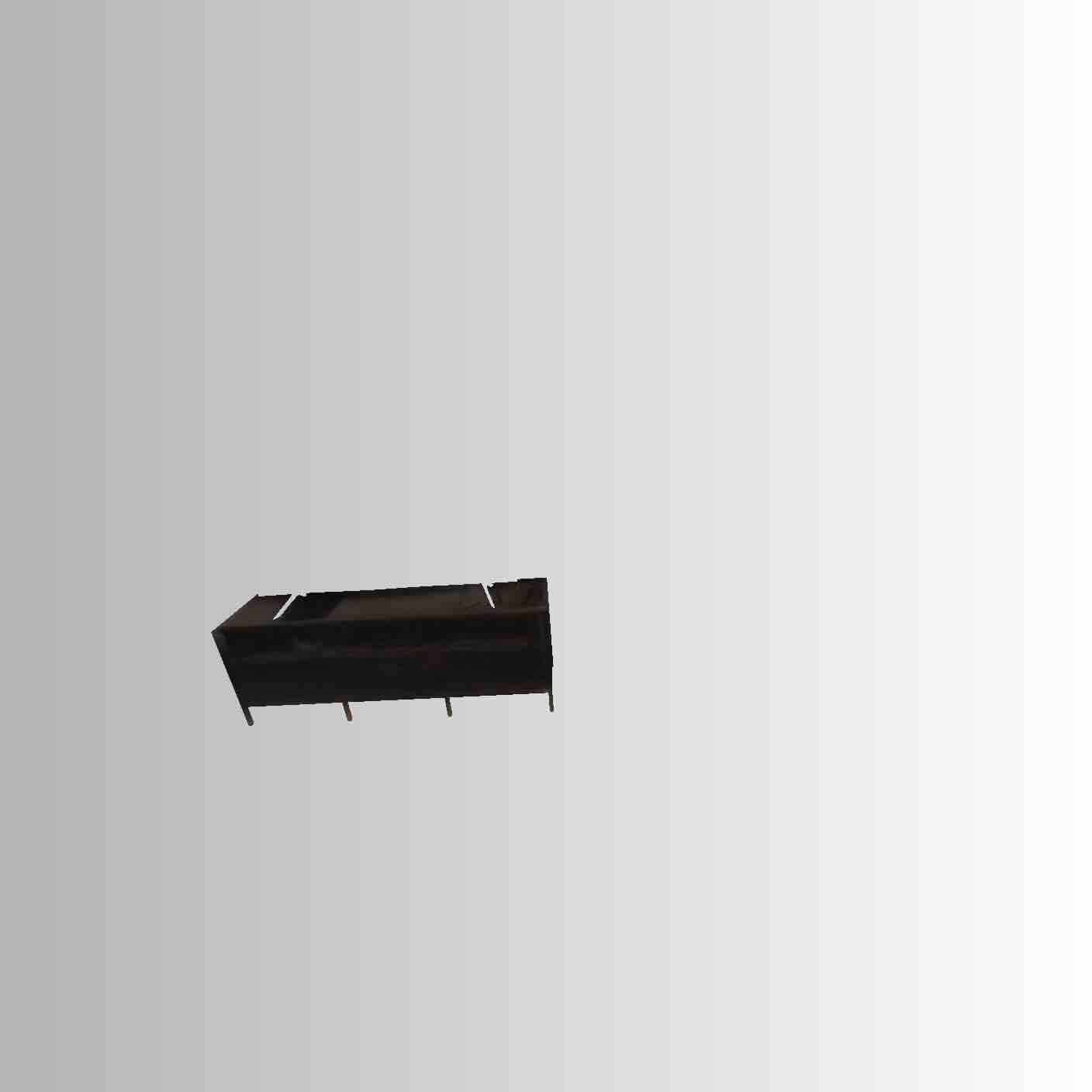} &
        \includegraphics[trim={2.5cm 10cm 17.5cm 17.5cm},clip,width=0.12\textwidth]{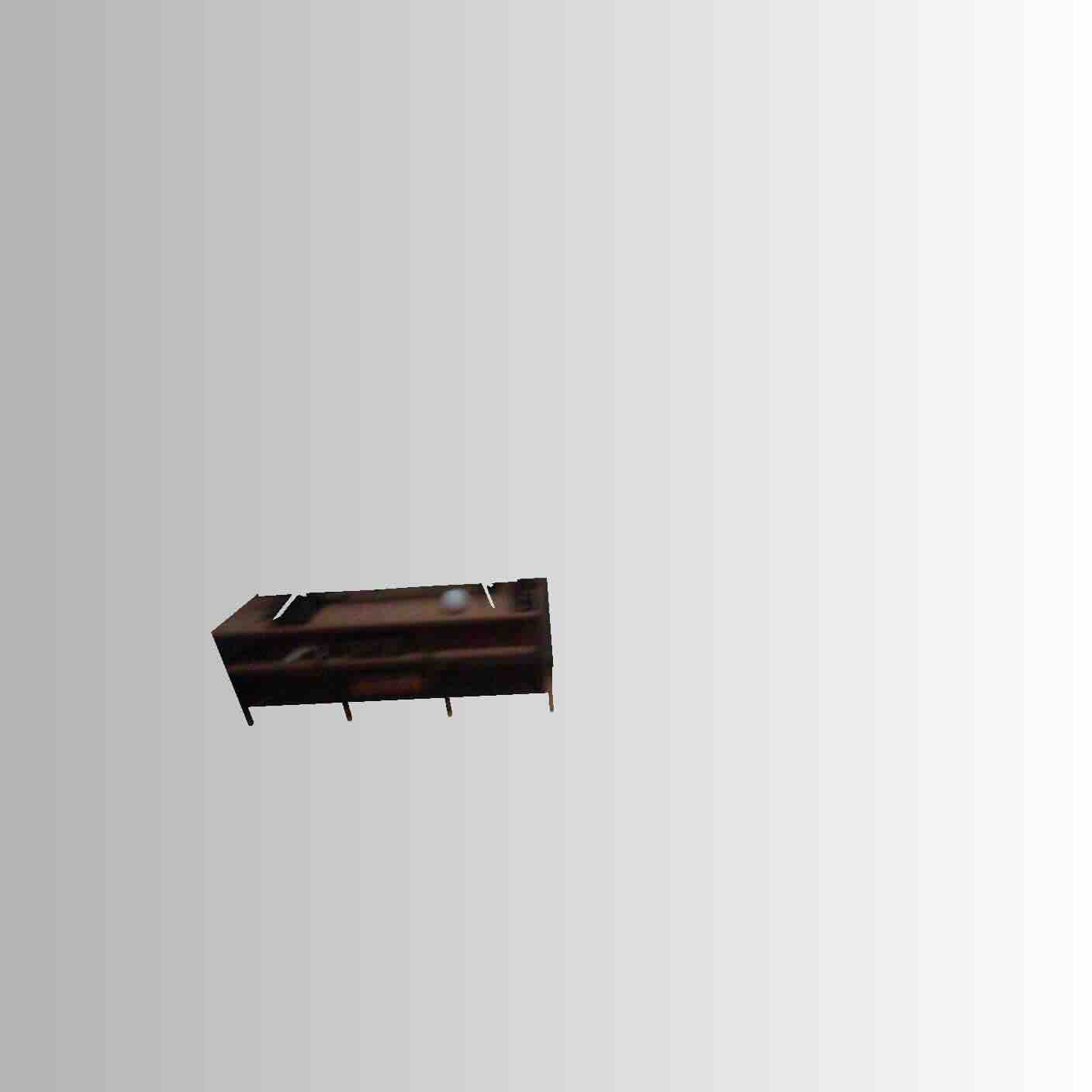}

        \\
        \scriptsize Input Images & 
\shortstack[c]{\scriptsize 3DGS-D \\ \scriptsize~\cite{kerbl20233d}} & 
\shortstack[c]{\scriptsize AGS-Mesh \\ \scriptsize~\cite{ren2025ags}} & 
\shortstack[c]{\scriptsize OM-GSD$^{\ast}$ \\ \scriptsize~\cite{lu2025orientation}} & 
\shortstack[c]{\scriptsize RVG-GSD$^{\ast}$ \\ \scriptsize~\cite{chang2025reconviagen}} & 
\shortstack[c]{\scriptsize SAM3D-GSD$^{\ast}$ \\ \scriptsize~\cite{chen2025sam}} & 
\scriptsize Ours & 
\scriptsize GT  \\
        \end{tabular} }
    \caption{\textbf{Qualitative Rendering Comparison.} The first three samples (top row) originate from the 3D-FRONT dataset~\cite{fu20213d}, the subsequent three (middle row) are drawn from ScanNet++~\cite{yeshwanth2023scannet++}, and the remaining three (bottom row) are drawn from ShapeR Evaluation Dataset~\cite{siddiqui2026shaper}. Our method achieves superior appearance reconstruction with high-fidelity texture consistency across multi-view renders. In contrast, baseline approaches exhibit significant observational gaps and produce disjointed textures that lack surface coherence, often leading to inconsistent color transitions.} 
    \label{fig:suppl_rendering}
    \vspace{-5mm}
\end{figure*}

\paragraph{Qualitative Evaluation of Novel View Synthesis.} We present extensive visualizations of novel view synthesis (NVS) across all the benchmarks in Fig.~\ref{fig:suppl_rendering}. The first three samples are sourced from the 3D-FRONT dataset~\cite{fu20213d}, while the subsequent three are drawn from ScanNet++~\cite{yeshwanth2023scannet++}, and the final three are drawn from ShapeR Evaluation Dataset~\cite{siddiqui2026shaper}. By evaluating our approach across these diverse synthetic and real-world environments, we demonstrate its superior capability in recovering high-fidelity appearance and maintaining structural integrity, particularly within heavily occluded regions that challenge standard reconstruction pipelines.

\section{Additional Implementation Details}
\label{supp:aid}

\paragraph{Guided Object Reconstruction.} 
During inference, our method optimizes object latents without necessitating fine-tuning of the generative model. All experiments were conducted on a single NVIDIA RTX 4090 GPU. For each object, we employ 300 iterations for both the voxel (at $64^3$ resolution) and 3D Gaussian generation stages. 
On average, the entire inference process, from input to the final generated 3D Gaussian Splatting (3DGS) representation, requires 59.86 seconds per object. Note that our method performs instance-specific optimization at inference time. Therefore, comparing our runtime directly to feed-forward generative baselines is not entirely equitable, as our approach invests additional computation to refine structural consistency on a per-instance basis without the need for prior training.

\paragraph{Photorealistic Radiance Refinement.} 
Benefiting from the high-quality 3D Gaussian initialization produced in the previous stage, our refinement process is significantly more efficient than standard baselines. Specifically, our method requires only 300 iterations for scene-level 3DGS optimization, whereas conventional 3DGS-based baselines typically require over 1,000 iterations to achieve convergence.

\begin{wrapfigure}{r}{0.48\columnwidth}
    \centering
    \vspace{-0.5cm}

    \setlength{\tabcolsep}{1pt}
    \begin{tabular}{ccc}
        
        \includegraphics[trim={15cm 0.0cm 15cm 0.0cm},clip,width=0.32\linewidth]{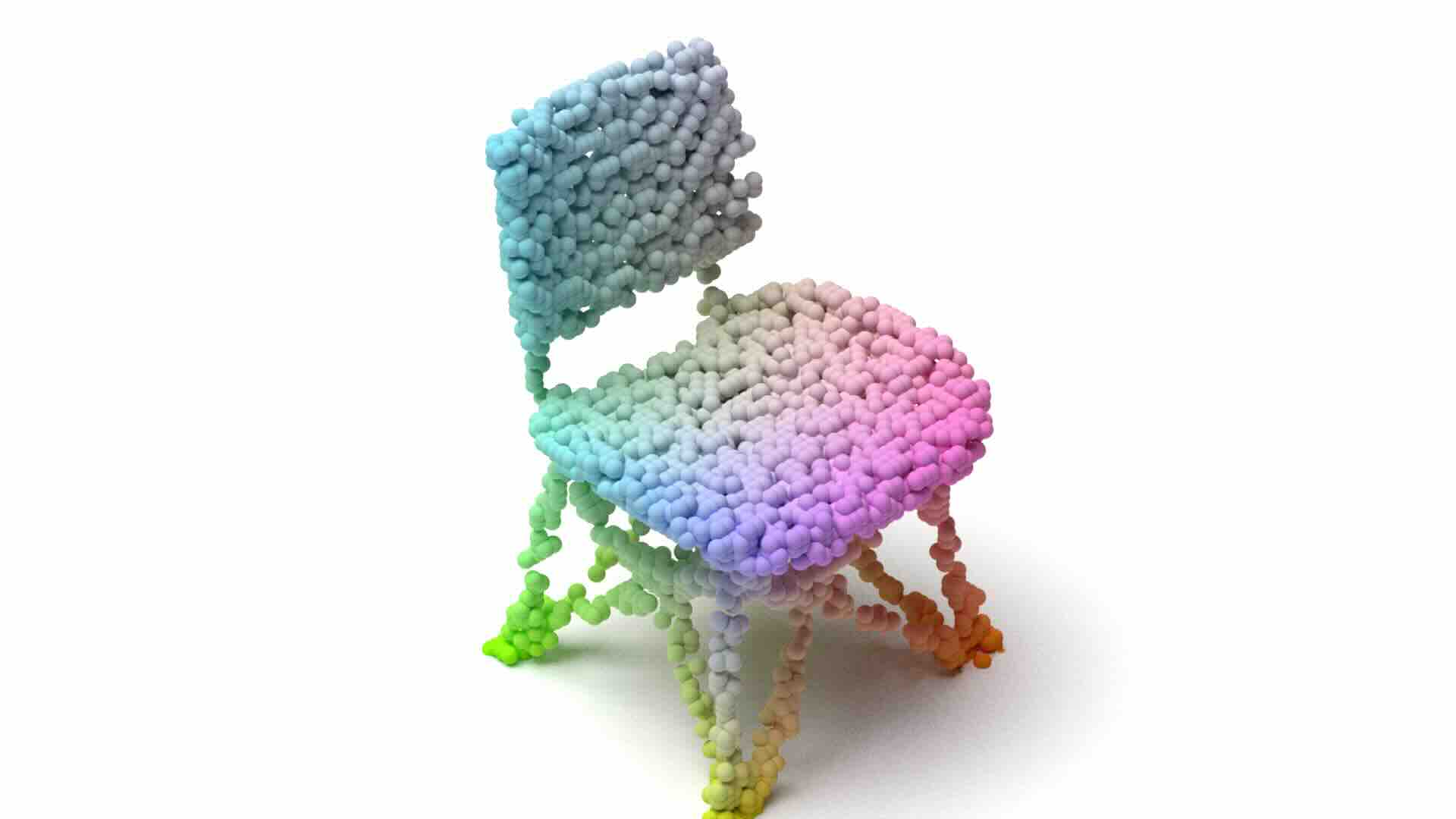} &
        \includegraphics[trim={15cm 0.0cm 15cm 0.0cm},clip,width=0.32\linewidth]{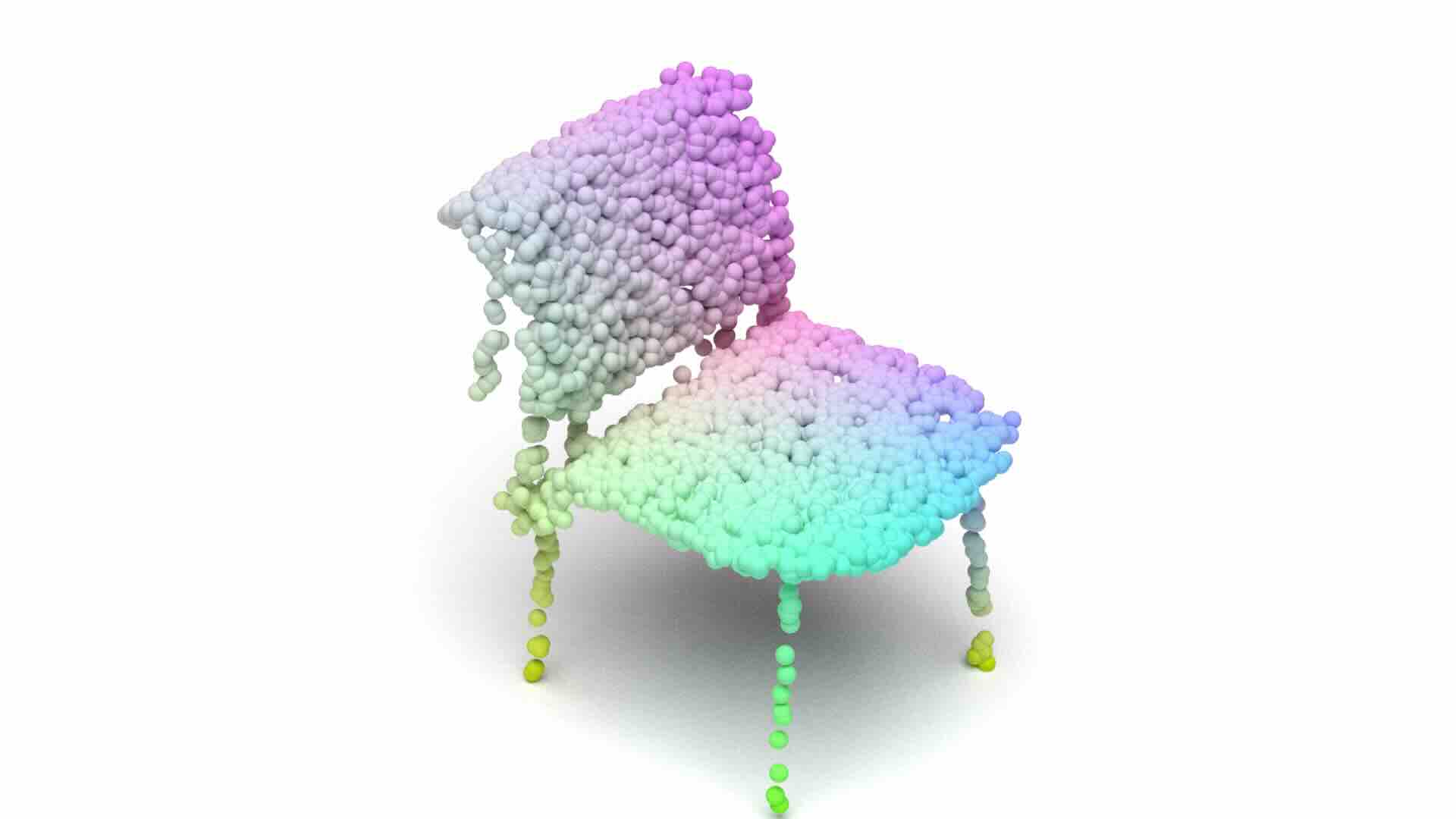} &
        \includegraphics[trim={15cm 0.0cm 15cm 0.0cm},clip,width=0.32\linewidth]{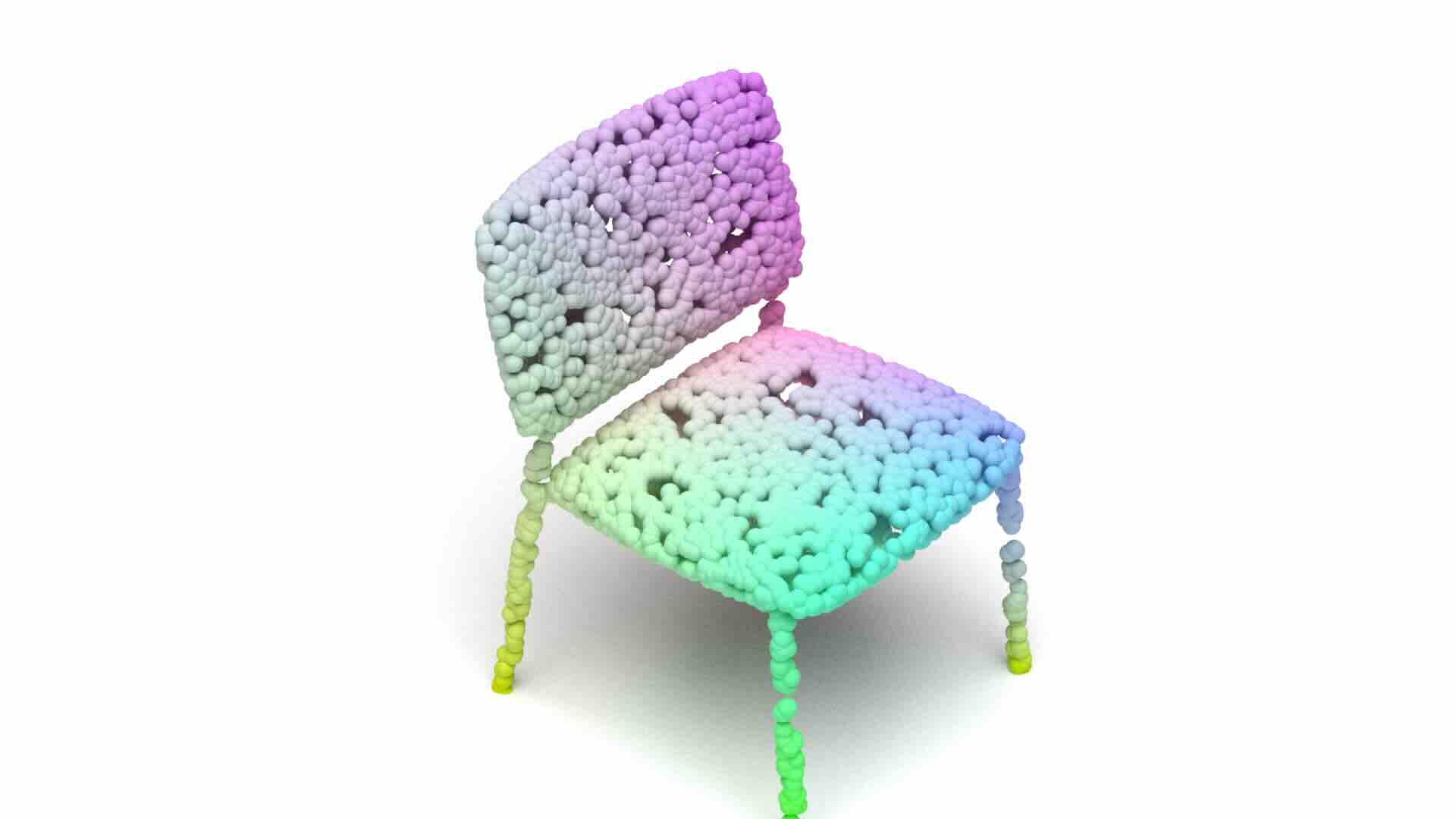} \\
        
        \includegraphics[width=0.32\linewidth]{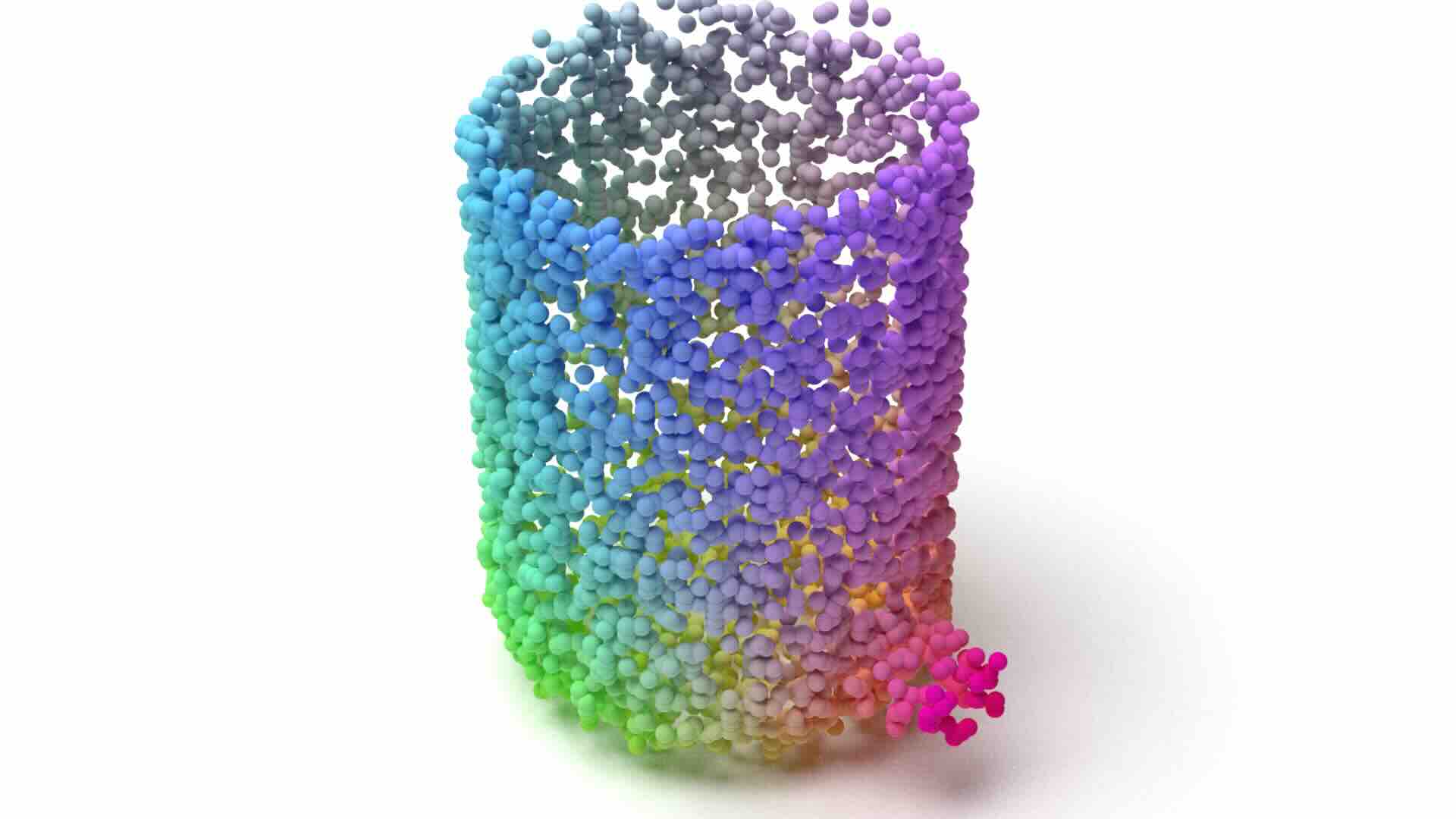} &
        \includegraphics[trim={0.0cm 1.0cm 0.0cm 1cm},clip,width=0.32\linewidth]{figures/scannetpp_pts/6_ours.jpg} &
        \includegraphics[trim={0.0cm 1.0cm 0.0cm 1cm},clip,width=0.32\linewidth]{figures/scannetpp_pts/6_gt.jpg} \\
        
    \scriptsize \makecell[t]{Vanilla \\ TRELLIS~\cite{xiang2025structured}} & 
    \scriptsize \makecell[t]{Ours \\ (w / Guid.)} & 
    \scriptsize GT \\
    \end{tabular}
    
    \caption{\textbf{Qualitative Comparison for voxel generation.} Our dual-space guidance ensures structural integrity and high-fidelity geometric consistency in occluded regions. While Vanilla TRELLIS~\cite{xiang2025structured} is capable of generating complete 3D volumes, it often produces geometric hallucinations that fail to adhere to real-world observations.}
    \label{fig:vox_generation}
    
    \vspace{-0.3cm} 
\end{wrapfigure}

\paragraph{3D-FRONT Dataset~\cite{fu20213d}.} 
In our experiments, we utilize four large-scale scenes (\textit{6a0e73bc}, \textit{a8c505d9}, \textit{bb9f9b20} and \textit{fccb9d09}), spanning 12 distinct rooms including dining rooms, living rooms, and bedrooms. This evaluation set comprises 56 unique objects across various categories such as tables, chairs, beds, nightstands, sofas, and cabinets. To ensure a rigorous evaluation, we exclude rendering comparisons on synthetic datasets where GS-based baselines benefit from ideal RGB-D initialization. Such comparisons would fail to accurately reflect the robust reconstruction capabilities of our method compared to these well-initialized baselines.

To ensure the topological integrity of the initial voxel grids, we utilize category-specific prompts. For \textit{beds}, we use: 
\textit{``A solid, monolithic and water-tight bed with a 360-degree fully enclosed frame, integrated headboard and matching side rails, complete posterior structure, ensuring a fully intact 3D volumetric geometry.''} 
For other categories, the prompt is dynamically constructed as: 
\textit{``A water-tight and fully manifold 3D mesh of a [category] in [room type], no redundant parts, no missing parts, no holes, complete surface.''} 
These constraints are essential to suppress the generation of redundant artifacts or incomplete geometries, ensuring high-fidelity structural initialization for the subsequent 3D Gaussian Splatting stage.

\paragraph{ScanNet++ Dataset~\cite{yeshwanth2023scannet++}.} 
We further evaluate our approach on real-world data using four scenes from ScanNet++ (\textit{116456116b}, \textit{30966f4c6e}, \textit{32280ecbca} and \textit{1a130d092a}). Following the object annotations from SCANnotate++~\cite{rao2025leveraging}, we tested a total of 51 objects. Masks are generated via projection of the 3D scans into each camera frame. This subset includes a diverse range of categories, such as tables, chairs, sofas, trash bins, cabinets, printers, and displays. We also sample training and testing views within different spatial regions relative to the object center. Additionally, we employ PromptDA~\cite{lin2025prompting} to enhance the resolution of the depth data captured by the iPhone.

To handle the noise and heavy occlusions in ScanNet++~\cite{yeshwanth2023scannet++} datasets, we employ highly descriptive, category-specific prompts to guide the voxel generation stage. These prompts enforce geometric constraints such as symmetry, manifoldness, and specific structural components (e.g., I-beam legs for tables). A complete list of the prompts used for various object categories is provided in Tab.~\ref{tab:prompts}. 
However, textual prompts alone are insufficient to overcome the lack of physical grounding; we observe that the Vanilla TRELLIS~\cite{xiang2025structured} framework still struggles to produce geometrically consistent results compared with the observed complex scenes, as illustrated in Fig.~\ref{fig:vox_generation}. Specifically, without our spatial anchoring, the Vanilla TRELLIS~\cite{xiang2025structured} suffers from significant scale ambiguity and spatial misalignment, resulting in structures that do not accurately reflect the object's true metric dimensions within the scene. This highlights the inherent limitations of purely text-guided generation and underscores the necessity of our proposed dual-space guidance for accurate spatial anchoring.

\paragraph{ShapeR Evaluation Dataset~\cite{siddiqui2026shaper}.} 
We also evaluate our approach on real-world data using three scenes from ShapeR Evaluation Dataset (\textit{ADT1292}, \textit{BNB1045} and \textit{HWD0045}). To ensure robust evaluation, we selected 20 objects with a frame density high enough to allow for disjoint training and testing splits. This subset includes a diverse range of categories, such as tables, chairs, sofas, tvstand, nightstand, cabinets, and bed. We primarily utilize the original captions provided by the dataset, applying manual rotations to GT meshes that are not forward-facing in their canonical space.

\begin{table}[h]
\centering
\vspace{-0.3cm}
\small
\begin{tabular}{lp{9cm}}
\toprule
\makebox[2cm][l]{\textbf{Category}} & \makebox[7cm][l]{\textbf{Core Prompting Strategy }} \\
\midrule
Chair & Enforce bilateral symmetry, specific support structures, and consistent armrest connectivity. \\
Table & Specify leg geometries to ensure floor contact and eliminate floating artifacts. \\
Trash Bin & Emphasis on material properties and manifold topology for watertight reconstruction. \\
Sofa \& Cabinet & Prioritize monolithic volumetric occupancy and structural integrity to prevent fragmented geometry. \\
Other & Direct category-level guidance: \textit{``A [category]''}. \\
\bottomrule
\end{tabular}
\caption{Selected category-specific prompts for voxel generation in ScanNet++.}
\label{tab:prompts}
\vspace{-0.7cm}
\end{table}

\paragraph{OM and RVG Post-processing Details.} OrientationMatters (OM)~\cite{lu2025orientation} and ReconViaGen (RVG)~\cite{chang2025reconviagen} are flow-matching-based reconstruction frameworks derived from TRELLIS~\cite{xiang2025structured}, designed to provide consistent, object-centric orientations from sparse multi-view inputs. Since image-based diffusion models often suffer from scale and translation ambiguities, we spatially align the results of baselines (3) and (4) to our coordinate system through uniform re-scaling and rigid alignment. This step is essential to ensure that our geometric comparisons remain fair and consistent. Specifically, we rescale the outputs of these two baselines by matching the minimum dimensions across all three axes relative to our results. We then employ RANSAC to estimate the optimal translation vectors. Finally, for a subset of objects where the model predicts wrong orientations, we perform a manual rotational correction to align the coordinate frames for consistent evaluation.

\paragraph{Video Supplement and Code Release.} The attached video provides a comprehensive overview of our project, showcasing geometric reconstructions and novel-view synthesis for various scene categories, which demonstrate our method's robustness in handling occlusions and noise in real-world environments. Our source code and implementation details will be made publicly available upon acceptance to facilitate future research.

